\documentclass{article}

\usepackage[utf8]{inputenc} %

\usepackage[preprint]{jmlr_custom}
\makeatletter

\long\def\@makecaption#1#2{%
  \vskip 8pt%
  \begingroup
    \footnotesize
    \setbox\@tempboxa\hbox{\textbf{#1}: #2}%
    \ifdim \wd\@tempboxa >\hsize
      \noindent \textbf{#1}: #2\par
    \else
      \hbox to\hsize{\hfil\box\@tempboxa\hfil}%
    \fi
  \endgroup
}

\makeatother

\usepackage{graphicx} %
\usepackage{enumitem}
\setlist[enumerate]{label=(\roman*), leftmargin=*, itemsep=1pt, topsep=2pt}
\usepackage{microtype}
\usepackage{pdflscape}
\usepackage{amsmath}

\usepackage{etoolbox}
\usepackage{longtable}

\usepackage{amsfonts}
\usepackage{amssymb,amsthm}
\usepackage{bm}
\usepackage{bbm}
\usepackage{parskip}
\usepackage{rotating}
\usepackage{mathtools}
\usepackage{mathrsfs}
\usepackage{mathbbol}
\usepackage{xcolor}
\usepackage[dvipsnames]{xcolor}
\usepackage{adjustbox}
\usepackage[capitalize]{cleveref}
\usepackage{mathtools}
\usepackage[only,llbracket,rrbracket]{stmaryrd}

\usepackage{booktabs}  %
\usepackage{array}     %
\usepackage[most]{tcolorbox}  %

\usepackage{wrapfig}
\usepackage{tikz}
\usetikzlibrary{shapes.geometric, arrows.meta, positioning, calc}

\definecolor{intropath}{RGB}{93, 178, 96}
\definecolor{advancedpath}{RGB}{218, 153, 76}
\definecolor{expertpath}{RGB}{172, 109, 218}

\renewcommand{\arraystretch}{1.3}  %
\definecolor{rowgray}{gray}{0.95}  %

\definecolor{mybg}{HTML}{EAE9FF} 
\definecolor{outline}{HTML}{B5B2F8} 
\definecolor{mytitle}{HTML}{CAC8FF} 

\newtcolorbox{purplebox}[1][]{%
  enhanced,
  breakable,
  frame hidden,
  boxrule=0pt,
  parbox=false,
  borderline={1pt}{0pt}{outline},
  colback=mybg,
  colbacklower=mybg,
  arc=2mm,
  title={#1},
  fonttitle=\bfseries,
  coltitle=black, %
  attach boxed title to top left={xshift=2mm,yshift*=-2mm},
  varwidth boxed title,
  boxed title style={
    frame hidden,
    boxrule=0pt,
    colback=mytitle,
    arc=1mm,
    left=2mm, right=2mm, top=0.6mm, bottom=0.6mm
  }
}

\newcommand{\purple}[2][]{%
  \begin{purplebox}[#1]
    #2
  \end{purplebox}
}

\tcbset{
  purpletheoremstyle/.style={
    enhanced,
    breakable,
    frame hidden,
    boxrule=0pt,
    parbox=false,
    borderline={1pt}{0pt}{outline},
    colback=mybg,
    colbacklower=mybg,
    arc=2mm,
    fonttitle=\bfseries,
    coltitle=black,
    attach boxed title to top left={xshift=2mm,yshift*=-2mm},
    varwidth boxed title,
    boxed title style={
      frame hidden,
      boxrule=0pt,
      colback=mytitle,
      arc=1mm,
      left=2mm,
      right=2mm,
      top=0.6mm,
      bottom=0.6mm
    }
  }
}

\definecolor{pinkbg}{HTML}{F7EFFF}
\definecolor{pinkoutline}{HTML}{EABBFF}
\definecolor{redbg}{HTML}{FFE7F3}
\definecolor{redoutline}{HTML}{FABBDB}
\definecolor{greenbg}{HTML}{D7F4FF}
\definecolor{greenoutline}{HTML}{86D2F0}

\tcbset{
  pinktheoremstyle/.style={
    enhanced,
    breakable,
    frame hidden,
    boxrule=0pt,
    parbox=false,
    borderline={1pt}{0pt}{pinkoutline},
    colback=pinkbg,
    colbacklower=pinkbg,
    arc=2mm,
    fonttitle=\bfseries,
    coltitle=black,
    attach boxed title to top left={xshift=2mm,yshift*=-2mm},
    varwidth boxed title,
    boxed title style={
      frame hidden,
      boxrule=0pt,
      colback=mytitle,
      arc=1mm,
      left=2mm,
      right=2mm,
      top=0.6mm,
      bottom=0.6mm
    }
  }
}

\tcbset{
  redtheoremstyle/.style={
    enhanced,
    breakable,
    frame hidden,
    boxrule=0pt,
    parbox=false,
    borderline={1pt}{0pt}{redoutline},
    colback=redbg,
    colbacklower=redbg,
    arc=2mm,
    fonttitle=\bfseries,
    coltitle=black,
    attach boxed title to top left={xshift=2mm,yshift*=-2mm},
    varwidth boxed title,
    boxed title style={
      frame hidden,
      boxrule=0pt,
      colback=mytitle,
      arc=1mm,
      left=2mm,
      right=2mm,
      top=0.6mm,
      bottom=0.6mm
    }
  }
}
\tcbset{
  greentheoremstyle/.style={
    enhanced,
    breakable,
    frame hidden,
    boxrule=0pt,
    parbox=false,
    borderline={1pt}{0pt}{greenoutline},
    colback=greenbg,
    colbacklower=greenbg,
    arc=2mm,
    fonttitle=\bfseries,
    coltitle=black,
    attach boxed title to top left={xshift=2mm,yshift*=-2mm},
    varwidth boxed title,
    boxed title style={
      frame hidden,
      boxrule=0pt,
      colback=mytitle,
      arc=1mm,
      left=2mm,
      right=2mm,
      top=0.6mm,
      bottom=0.6mm
    }
  }
}

\usepackage{tcolorbox}
\tcbuselibrary{theorems, breakable}
\usepackage{amsthm}

\tcolorboxenvironment{definition}{
  pinktheoremstyle,
}

\tcolorboxenvironment{theorem}{
  greentheoremstyle
}
\tcolorboxenvironment{lemma}{
  pinktheoremstyle
}

\tcolorboxenvironment{corollary}{
  pinktheoremstyle
}

\tcolorboxenvironment{proposition}{
  greentheoremstyle
}

\tcolorboxenvironment{assumption}{
  greentheoremstyle
}

\tcolorboxenvironment{remark}{
  redtheoremstyle
}

\definecolor{MyPurple}{HTML}{6B67EE}

\newcommand{\boldtext}[1]{\textbf{\textcolor{MyGreen}{#1}}}

\definecolor{customblue}{HTML}{4A90E2}
\newcommand{\bluetext}[1]{\textcolor{customblue}{#1}}

\definecolor{customgreen}{HTML}{AA61D4}
\newcommand{\greentext}[1]{\textcolor{customgreen}{#1}}

\definecolor{custompink}{HTML}{0096A5}
\newcommand{\pinktext}[1]{\textcolor{custompink}{#1}}

\definecolor{generalbg}{HTML}{FDF1DB}   
\definecolor{generaltitle}{HTML}{F3DEAA}

\newtcolorbox{generalbox}[1][]{%
  enhanced,
  breakable,
  frame hidden,
  boxrule=0pt,
  parbox=false,
  colback=generalbg,
  colbacklower=generalbg!4!gray,
  arc=2mm,
  title={#1},
  fonttitle=\bfseries,
  coltitle=black,
  attach boxed title to top left={xshift=2mm,yshift*=-2mm},
  varwidth boxed title,
  boxed title style={
    frame hidden,
    boxrule=0pt,
    colback=generaltitle,
    arc=1mm,
    left=2mm, right=2mm, top=0.6mm, bottom=0.6mm
  }
}

\usepackage{listings}
\usepackage{xcolor}

\definecolor{codebg}{rgb}{0.97,0.97,0.97}
\definecolor{codekw}{HTML}{50B2D7}
\definecolor{codestring}{HTML}{CD76FF}
\definecolor{codecomment}{HTML}{6B67EE}

\definecolor{figblue}{HTML}{3A62B4}
\definecolor{figred}{HTML}{C8143C}
\definecolor{figyellow}{HTML}{FDF2DC}
\definecolor{MyPurple}{HTML}{6B67EE}
\definecolor{MyRed}{HTML}{D7006B}
\definecolor{MyPink}{HTML}{CD76FF}
\definecolor{MyGreen}{HTML}{50B2D7}

\title{Foundations of Schrödinger Bridges for Generative Modeling}

\author{%
\textbf{Sophia Tang}
\\[8pt]
{\normalfont Department of Computer and Information Science}\\
{\normalfont University of Pennsylvania}\\
{\normalfont Correspondence to: \href{sophtang@seas.upenn.edu}{\texttt{sophtang@seas.upenn.edu}}}
}

\begin{document}

\maketitle

\begin{abstract}
At the core of modern generative modeling frameworks—including diffusion models, score-based models, and flow matching—is the task of transforming a simple prior distribution into a complex target distribution through stochastic paths in probability space. Schrödinger bridges provide a unifying principle underlying these approaches, framing the problem as determining an optimal stochastic bridge between marginal distribution constraints with minimal-entropy deviations from a pre-defined reference process. This guide develops the mathematical foundations of the Schrödinger bridge problem, drawing on optimal transport, stochastic control, and path-space optimization, and focuses on its dynamic formulation with direct connections to modern generative modeling. We build a comprehensive toolkit for constructing Schrödinger bridges from first principles, and show how these constructions give rise to generalized and task-specific computational methods.
\looseness=-1
\end{abstract}

\begin{center}
    \includegraphics[width=0.98\linewidth]{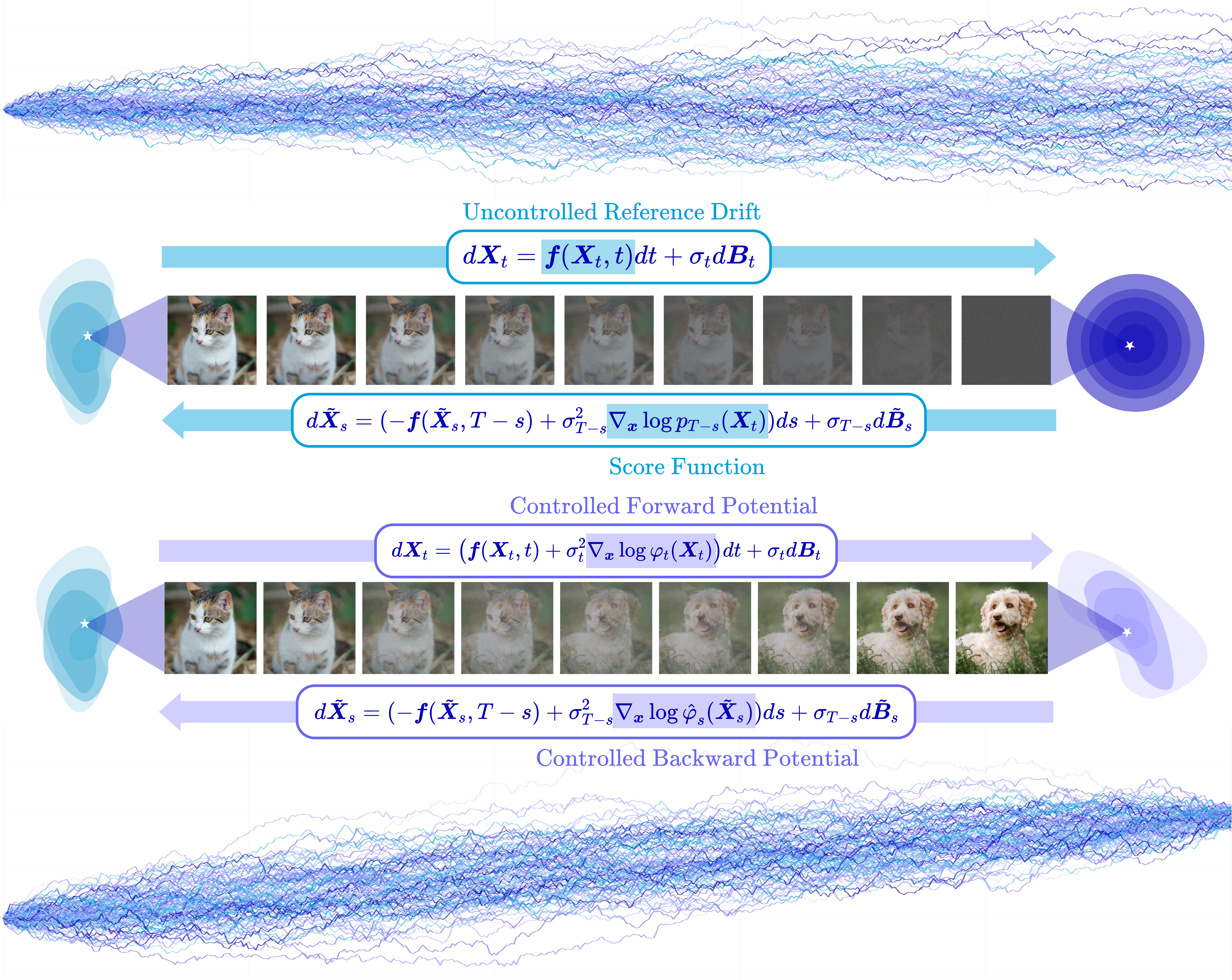}
    \label{fig:cover}
\end{center}

\newpage

\tableofcontents

\newpage

\section*{Preface}

The field of generative modeling has experienced rapid and transformative progress in recent years. Advances range from foundational theoretical developments—such as diffusion models and flow matching—to algorithmic improvements in sampling speed and generation quality, as well as significant applications spanning language, video, and scientific domains. While this breadth of progress has driven remarkable innovation, it has also made the field increasingly complex and difficult to navigate, particularly from a foundational perspective.

In this guide, we introduce \textbf{Schrödinger bridges as a unifying theoretical framework for generative modeling}. This perspective generalizes a broad class of modern approaches—including diffusion models, flow-matching methods, and stochastic control formulations—while providing a principled and flexible foundation for addressing specialized scientific problems. The body of work that has built the Schrödinger bridge framework spans multiple fields, from theoretical developments that formalize path-space optimization and entropy-regularized transport, to principled extensions in diverse problem settings, to algorithmic advances that enable scalable training and inference. Despite this progress, the literature remains fragmented, and the relationships between different formulations are often difficult to identify.

The goal of this guide is to build both intuition and a deep mathematical understanding of the core principles of Schrödinger bridges, from its origins in optimal transport to its dynamic path space formulation, which underlies modern generative modeling frameworks. At a high level, we begin from a \textbf{single unifying principle}: optimal stochastic bridges between distributions can be characterized as minimal-entropy deviations from a reference process subject to marginal constraints. From this perspective, we develop the mathematical tools spanning optimal transport, probability theory, and stochastic calculus required to understand the static and dynamic formulations of the Schrödinger bridge problem, before naturally extending these ideas to diverse problem settings and modern generative modeling techniques. Throughout this guide, we construct a principled toolkit for building Schrödinger bridges from first principles and show how these constructions give rise to theoretically-grounded and scalable computational methods for simulating the optimal Schrödinger bridge dynamics.

This guide contains eight primary sections, which are organized as follows:

\paragraph{Section \ref{sec:static-sb}} introduces the \textbf{static Schrödinger bridge (SB) problem}, which provides the theoretical foundations for optimal transport between probability distributions.
\begin{itemize}
    \item Section \ref{subsec:optimal-mass-transport} traces back to the classical \textbf{optimal mass transport} (OMT) problem posed by Monge and Kantorovich, which seeks an optimal static coupling between probability distributions that minimizes total transport cost. 
    \item Section \ref{subsec:entropy-probability-spaces} introduces the foundational properties of \textbf{entropy} on probability spaces, including the fundamental properties of the Kullback-Leibler (KL) divergence.
    \item Section \ref{subsec:entropic-ot} introduces the \textbf{entropic optimal transport} (EOT) problem, which leverages entropy as a method of regularizing the OMT problem with a pre-defined reference coupling, which yields a unique solution.
    \item Section \ref{subsec:static-sbp} extends the EOT problem to formalize the \textbf{static Schrödinger bridge problem} and its corresponding \textbf{dual problem}, which introduces a pair of Schrödinger potentials that uniquely yield a clean form of the optimal solution. 
    \item Section \ref{subsec:sinkhorn-algorithm} breaks down the classic algorithm for solving the static SB problem, known as \textbf{Sinkhorn's algorithm}, which alternates between optimizing the dual Schrödinger potentials.
\end{itemize}

\paragraph{Section \ref{sec:dynamic-sb}} lifts the static SB problem into the space of stochastic path measures defined as the \textbf{dynamic Schrödinger bridge (SB) problem}.
\begin{itemize}
    \item Section \ref{subsec:dynamic-ot-problem} starts by redefining the static SB problem as learning a continuous-time deterministic flow between distributions, known as the \textbf{dynamic optimal transport} (OT) problem. 
    \item Section \ref{subsec:nonlinear-sbp} introduces the \textbf{dynamic Schrödinger bridge} (SB) problem, which reformulates the SB problem as an entropy minimization over stochastic processes. 
    \item Section \ref{subsec:path-measure-ito-processes} provides the key definitions and theory required for understanding path measures as stochastic differential equations (SDEs) that can be \textit{steered} via control drifts and \textit{transformed} through functions. 
    \item Section \ref{subsec:fp-equation} derives the \textbf{Fokker-Planck equation}, governing how the probability density over the stochastic paths generated from an SDE evolves forward in time, and the \textbf{Feynman-Kac equation}, governing how functions evaluated at the end of a stochastic process evolve backward in time.
    \item Section \ref{subsec:girsanov} derives Girsanov's theorem from first principles, which allows us to define changes in measure and KL divergences on path space.
    \item Section \ref{subsec:path-measure-rnd-kl} explicitly breaks down the theory of \textbf{Radon-Nikodym derivatives} between path measures, which is used to explicitly define the \textit{relative-entropy} between path measures.
    \
\end{itemize}

\paragraph{Section \ref{sec:sb-optimal-control}} reformulates the SB problem through the lens of \textbf{stochastic optimal control (SOC)} theory.
\begin{itemize}
    \item Section \ref{subsec:stochastic-optimal-control} introduces the general framework of \textbf{stochastic optimal control}, including Bellman's Principle of Optimality given a terminal constraint and deriving the value function that defines the optimal control.
    \item Section \ref{subsec:sb-soc} connects SB to SOC, showing that the optimal bridge corresponds to an \textbf{optimal control drift}.
    \item Section \ref{subsec:soc-objectives} develops \textbf{practical objectives} for solving the SOC problem.
\end{itemize}

\paragraph{Section \ref{sec:building-stochastic-bridge}} several complementary mechanisms for \textbf{building Schrödinger bridges}.
\begin{itemize}
    \item Section \ref{subsect:mixture-bridges} introduces \textbf{mixtures of conditional bridges} constructed with the reference process and a pre-defined endpoint coupling.
    \item Section \ref{subsec:time-reversal} derives the \textbf{time-reversal formula} of SDEs, which is fundamental to backward dynamics in SB.
    \item Section \ref{subsec:forward-backward-sde} introduces \textbf{forward-backward stochastic differential equations (FBSDEs)}, providing a coupled characterization of SB with respect to the time-dependent Schrödigner potentials.
    \item Section \ref{subsec:doob-transform} presents \textbf{Doob’s $h$-transform}, which constructs conditioned stochastic processes by tilting the reference process using $h$-function.
    \item Section \ref{subsec:markov-reciprocal-proj} formalizes \textbf{Markovian and reciprocal projections}, which perform entropy-minimizing projections in path space that converge to the optimal SB measure.
    \item Section \ref{subsec:stochastic-interpolants} introduces \textbf{stochastic interpolants}, providing a practical way to construct bridges between distributions as deterministic interpolants with Gaussian noise.
\end{itemize}

\paragraph{Section \ref{sec:variations-of-sb}} explores important \textbf{variations of the Schrödinger bridge problem} for different modeling settings.
\begin{itemize}
    \item Section \ref{subsec:gaussian-sb} studies the \textbf{Gaussian SB problem}, which admits a closed-form solutions.
    \item Section \ref{subsec:generalized-sb} introduces the \textbf{generalized SB problem} which generalizes dynamic SB to model mean-field interactions.
    \item Section \ref{subsec:multi-marginal-sb} introduces the \textbf{multi-marginal SB problem}, which extends SB to settings with multiple intermediate marginal constraints.
    \item Section \ref{subsec:unbalanced-sbp} develops the \textbf{unbalanced SB problem}, allowing mass creation and destruction along the stochastic trajectories.
    \item Section \ref{subsec:branched-sbp} introduces the \textbf{branched SB problem}, enabling modeling of diverging trajectories to multiple distinct terminal modes.
    \item Section \ref{subsec:fractional-sbp} studies \textbf{fractional SB problems}, incorporating long-range temporal dependencies through fractional Brownian motion.
\end{itemize}

\paragraph{Section \ref{sec:generative-modeling}} connects SB theory to \textbf{modern generative modeling} frameworks.
\begin{itemize}
    \item Section \ref{subsec:primer-sgm} provides a primer on \textbf{score-based generative modeling}, which learns gradients of log-densities.
    \item Section \ref{subsec:likelihood-training} introduces \textbf{likelihood training of forward-backward SDEs}, which trains the forward and backward potential drifts with likelihood objectives.
    \item Section \ref{subsec:diffusionsbm} develops \textbf{diffusion Schrödinger bridge matching}, which parameterizes the Iterative Markovian Fitting procedure with a learned Markov drift. 
    \item Section \ref{subsec:score-and-flow} introduces \textbf{simulation-free score and flow matching}, enabling training of SB without simulating full trajectories.
    \item Section \ref{subsec:adjoint-matching} presents \textbf{adjoint matching}, which learns the optimal SB while avoiding explicit sampling from target distributions.
\end{itemize}

\paragraph{Section \ref{sec:discrete-state-space}} extends SB to \textbf{discrete state spaces}.
\begin{itemize}
    \item Section \ref{subsec:ctmc} introduces \textbf{continuous-time Markov chains (CTMCs)} as discrete analogues of stochastic processes.
    \item Section \ref{subsec:discrete-sbp} formulates the \textbf{discrete Schrödinger bridge problem}, introducing Radon-Nikodym derivatives and KL divergences for CTMCs.
    \item Section \ref{subsec:soc-ctmcs} develops \textbf{stochastic optimal control for CTMCs} as a method of learning the optimal CTMC path measure with terminal constraint. 
    \item Section \ref{subsec:discrete-sb-soc} connects discrete SB with \textbf{stochastic optimal control} and introduces practical algorithms and objectives. 
    \item Section \ref{subsec:discrete-markov-reciprocal} introduces \textbf{Markovian and reciprocal projections in discrete spaces}.
    \item Section \ref{subsec:ddsbm} develops \textbf{discrete diffusion Schrödinger bridge matching}, which parameterizes the Markovian and reciprocal generators to solve the discrete SB.
\end{itemize}

\paragraph{Section \ref{sec:applications}} highlights diverse \textbf{applications of generative modeling with Schrödinger bridges}.
\begin{itemize}
    \item Section \ref{subsec:data-translation} applies SB to \textbf{data translation} tasks, which map between structured data distributions. 
    \item Section \ref{subsec:cell-modeling} studies \textbf{single-cell state dynamics}, showing how SB formulations can model cell population dynamics and responses to perturbation.
    \item Section \ref{subsec:sampling} applies SB to \textbf{sampling Boltzmann distributions}, showing how the SB frameworks can generate from unnormalized energy distributions without explicit samples.
\end{itemize}

This guide is designed to establish both an intuitive and logical flow between foundational concepts, paired with rigorous derivations that solidify abstract ideas with principled theory. Despite the mathematically rigorous exploration of Schrödinger bridges in this guide, it does \textit{not} require prior background on stochastic calculus, Schrödinger bridges, or generative modeling, as all necessary concepts are clearly defined and explained within the guide. Given this, the guide is designed for researchers, practitioners, and students interested in gaining a deep foundational understanding of generative modeling and its connections with recent developments in the field. This guide is not intended to be an exhaustive survey of all algorithms, architectures, or empirical applications involving Schrödinger bridges. Instead, our goal is to provide readers with a conceptual and mathematical foundation from which diverse models and future developments can be understood.

\newpage
\section*{Notation}
Throughout this guide, we often refer to the \textbf{control drift} and \textbf{velocity field} of an SDE. While it will become more apparent while reading, the \textbf{control drift} refers to added term to the reference drift which is scaled by the diffusion coefficient, generally denoted $\sigma_t\boldsymbol{u}(\boldsymbol{x},t)$, and the \textbf{velocity field} refers to the entire non-diffusion term in the SDE that appears before $dt$ including both the reference drift and the control drift, generally denoted $\boldsymbol{v}(\boldsymbol{x},t)$. 

A key notational convention we adopt for clarity is using the $\boldsymbol{X}^u_{0:T}=(\boldsymbol{X}_t)_{t\in [0,T]}^u$ when we want to emphasize that the path measure is generated under a \textit{specific} control drift $\boldsymbol{u}(\boldsymbol{x},t)$, which appears in the sections on stochastic optimal control. In other sections, we omit this superscript when the underlying path measure is clear from context. \textbf{In the table below, we list the common notation used throughout this guide, which we will often define repeatedly throughout the text for clarity.} 

\vspace{0.5em}
\begingroup
\renewcommand{\arraystretch}{1.25}
\begin{center}
\begin{longtable}{@{\extracolsep{\fill}}>{\centering\arraybackslash}p{1.8in} p{4.0in}@{}}
\hline
\textbf{Notation} & \textbf{Meaning} \label{table:notation1}
\\
\hline
\multicolumn{2}{l}{\textit{Basic Notation}} \\
\hline
$\text{KL}(\cdot\|\cdot)$ & Kullback-Leibler divergence \\
$\inf$ & infimum or the largest lower bound of a set \\
$\sup$ & supremum or the smallest upper bound of a set \\
$\partial_t$ & partial derivative with respect to $t$ \\
$d, dt, d\boldsymbol{x}$ & differential, time differential, state differential \\
$\nabla$ & gradient operator, shorthand for $\nabla $ unless specified otherwise \\
$\nabla \cdot$ & divergence operator, shorthand for $\nabla \cdot$ unless specified otherwise \\
$\Delta=\nabla\cdot\nabla=\nabla^2$ & Laplacian operator, shorthand for $\Delta_{\boldsymbol{x}}\cdot$ unless specified otherwise \\
$\phi\in C^{2}(\mathbb{R}^d)$ & functions that are twice continuously differentiable with respect to $\boldsymbol{x}\in \mathbb{R}^d$ \\
$\phi\in C^{2,1}(\mathbb{R}^d\times [0,T])$ & functions that are twice continuously differentiable with respect to $\boldsymbol{x}\in \mathbb{R}^d$ and continuously differentiable with respect to $t\in [0,T]$ \\
$\phi\in L^2$ & square integrable functions with finite $L^2$ norm \\
$\boldsymbol{I}_d$ & $d$-dimensional Identity matrix \\
$\boldsymbol{1}_{\boldsymbol{x}=\boldsymbol{y}}$ & indicator function that returns 1 when $\boldsymbol{x}= \boldsymbol{y}$ and 0 otherwise \\
$\frac{\delta\mathcal{L}}{\delta \boldsymbol{u}}$ & functional derivative of the functional $\mathcal{L}$ with respect to $\boldsymbol{u}$ \\
\hline
\multicolumn{2}{l}{\textit{Matrix and Vector Operations}} \\
\hline
$\langle\cdot, \cdot\rangle$ & inner sum \\
$\|\cdot\|^2$ & $L^2$-norm unless specified otherwise \\
$\boldsymbol{A}^\top$ & vector or matrix transpose \\
$\boldsymbol{A}^{-1}$ & matrix inverse \\
\hline
\multicolumn{2}{l}{\textit{State Spaces}} \\
\hline
$\mathcal{X},\mathcal{Y}$ & arbitrary state spaces (in Section \ref{sec:static-sb}, $\mathcal{X}$ is the source state space and $\mathcal{Y}$ is the target state space and in Section \ref{sec:discrete-state-space}, $\mathcal{X}:=\{1, \dots, d\}$ is used to denote a finite state space) \\
$\mathcal{X}\times\mathcal{Y}$ & product space of $\mathcal{X}$ and $\mathcal{Y}$ \\
$\mathbb{R}^d$ & $d$-dimensional state space \\
$\mathcal{P}(\cdot)$ & probability space \\
$\mathcal{P}(\mathbb{R}^d)$ & space of probability distributions over the state space $\mathbb{R}^d$ \\
$C([0,T]; \mathbb{R}^d)$ & path space over time horizon $t\in [0,T]$ and state space $\mathbb{R}^d$\\
$\mathcal{P}(C([0,T]; \mathbb{R}^d))$ & space of probability distributions over the path space \\
$\mathcal{U}$ & space of feasible control drifts, defined in Definition \ref{def:dynamic-sb}\\
\hline
\multicolumn{2}{l}{\textit{Probability Densities}} \\
\hline
$p_t$ & marginal density of path measure at time $t$, typically the controlled path measure $\mathbb{P}^u$ \\
$p_0, p_T$ & initial and terminal marginals generated from path measure $\mathbb{P}$ \\
$p^\star_t$ & marginal density of optimal path measure $\mathbb{P}^\star$ at time $t$ \\
$p^\star_0, p^\star_T$ & initial and terminal marginals generated from optimal path measure $\mathbb{P}$ \\
$q_t$ & marginal density of reference path measure $\mathbb{Q}$ at time $t$ \\
$q_0, q_T$ & initial and terminal marginals generated from reference path measure $\mathbb{Q}$ \\
$\tilde{p}_s$ & marginal density with respect to reverse time coordinate $s:=T-t$ \\
$\pi_0$ & marginal distribution constraint at time $t=0$ \\
$\pi_T$ & marginal distribution constraint at time $t=T$ \\
$\pi_{0,T}$ & endpoint law or coupling distribution \\
$\pi_0 \otimes \pi_T$ & product measure \\
$\pi^\star_{0,T}$ & optimal OT or SB endpoint law or coupling distribution \\
\hline
\multicolumn{2}{l}{\textit{Stochastic Processes and Path Measures}} \\
\hline
$\boldsymbol{x}$ & realized state in $\mathbb{R}^d$ \\
$\boldsymbol{X}_t$ & random variable in $\mathbb{R}^d$ \\
$\boldsymbol{X}_{0:T}, (\boldsymbol{X}_t)_{t\in [0,T]}$ & forward stochastic process over time $t\in [0,T]$ in $\mathbb{R}^d$ \\
$\tilde{\boldsymbol{X}}_{0:T}, (\tilde{\boldsymbol{X}}_s)_{s\in [0,T]}$ & backward stochastic process over reverse time $s=T-t\in [0,T]$ in $\mathbb{R}^d$ \\
$\boldsymbol{X}^u_{0:T}, (\boldsymbol{X}^u_t)_{t\in [0,T]}$ & forward stochastic process generated from the controlled SDE with control $\boldsymbol{u}$ \\
$\boldsymbol{B}_t$ & Brownian motion random variable \\
$\mathbb{P}, \mathbb{Q}$ & path measure in $\mathcal{P}(C([0,T]; \mathbb{R}^d))$ \\
$\mathbb{P}_{0,T}, \mathbb{Q}_{0,T}, \mathbb{P}^\star_{0,T}, \mathbb{P}^u_{0,T}, \Pi_{0,T}, \mathbb{M}_{0,T}$ & endpoint law of path measure \\
$\mathbb{P}^u$ & controlled path measure in $\mathcal{P}(C([0,T]; \mathbb{R}^d))$ with control drift $\boldsymbol{u}$ \\
$\mathbb{Q}$ & reference path measure that defines the prior dynamics \\
$\sigma\mathbb{B}$ & pure Brownian motion path measure with SDE $d\boldsymbol{X}_t=\sigma_td\boldsymbol{B}_t$ \\
$\mathbb{P}^\star$ & path measure of the optimal solution \\
$\Pi$ & mixture of bridges under the reference process defined as $\Pi:=\Pi_{0,T}\mathbb{Q}_{\cdot |0,T}$ \\
$\mathcal{R}(\mathbb{Q})$ & reciprocal class of $\mathbb{Q}$ containing all mixtures of bridges \\
$\mathbb{M}$ & Markov path measure \\
$\mathcal{M}$ & space of Markov measures \\
$\mathbb{M}^\star:=\text{proj}_{\mathcal{M}}(\cdot)$ & Markov projection of a reciprocal measure\\
$\Pi^\star:=\text{proj}_{\mathcal{R}(\mathbb{Q})}(\cdot)$ & reciprocal projection of a reciprocal measure\\
$\mathbb{P}^{\tilde{u}}\ll \mathbb{P}^u$ & absolute continuity, where  $\mathbb{P}^{\tilde{u}}$ is absolutely continuous with respect to $\mathbb{P}^u$ \\
$\mathbb{P}^{\tilde{u}}\sim \mathbb{P}^u$ & mutual absolute continuity, where  $\mathbb{P}^{\tilde{u}}$ is absolutely continuous with respect to $\mathbb{P}^u$, and vice versa \\
$\mathbb{P}_{\tau |t}(\boldsymbol{x}_\tau|\boldsymbol{x})$ or $\mathbb{P}(\boldsymbol{X}_\tau=\boldsymbol{x}_\tau|\boldsymbol{X}_t=\boldsymbol{x})$ & transition density from state $\boldsymbol{x}$ at time $t$ to state $\boldsymbol{x}_\tau$ at time $\tau$ \\
\hline
\multicolumn{2}{l}{\textit{Schrödinger Bridge and Generative Modeling Theory}} \\
\hline
$\boldsymbol{f}(\boldsymbol{x},t):\mathbb{R}^d \times [0,T] \to \mathbb{R}^d$ & uncontrolled drift of reference process $\mathbb{Q}$ \\
$\varphi:\mathcal{X}\to \mathbb{R},\hat\varphi:\mathcal{Y}\to \mathbb{R}$ & static Schrödinger potentials defined on $\mathcal{X}$ and $\mathcal{Y}$ \\
$\varphi\oplus\hat\varphi$ & separable sum of functions defined on two coordinates \\
$\varphi_t(\boldsymbol{x}):\mathbb{R}^d\times [0,T]\to \mathbb{R}$ & forward  Schrödinger bridge potential \\
$\hat{\varphi}_t(\boldsymbol{x}):\mathbb{R}^d\times [0,T]\to \mathbb{R}$ & backward  Schrödinger bridge potential \\
$\boldsymbol{u}(\boldsymbol{x},t):\mathbb{R}^d\times [0,T] \to \mathbb{R}^d$ & control drift \\
$\boldsymbol{u}^\star(\boldsymbol{x},t):\mathbb{R}^d\times [0,T] \to \mathbb{R}^d$ & Schrödinger bridge drift or optimal control drift \\
$\bar{\boldsymbol{u}}(\boldsymbol{x},t):\mathbb{R}^d\times [0,T] \to \mathbb{R}^d$ & non-gradient-tracking control drift $\bar{\boldsymbol{u}}=\texttt{stopgrad}(\boldsymbol{u})$ \\
$\boldsymbol{v}(\boldsymbol{x},t):\mathbb{R}^d\times [0,T] \to \mathbb{R}^d$ & velocity field or arbitrary control drift depending on context \\
$\boldsymbol{\Sigma}_t: [0,T] \to \mathbb{R}^{d\times d}$ & diffusion covariance matrix (simplified with scalar $\sigma_t$ throughout the guide) \\
$\sigma_t: [0,T] \to \mathbb{R}_{\geq 0}$ & scalar diffusion coefficient \\
$c(\boldsymbol{x},\boldsymbol{y}):\mathcal{X}\times \mathcal{Y} \to \mathbb{R}$ & used in Section \ref{sec:static-sb} to denote the transport cost \\
$c(\boldsymbol{x},t):\mathbb{R}^d\times [0,T] \to \mathbb{R}$ & used in remaining sections to denote state cost or potential \\
$V_t(\boldsymbol{x}):\mathbb{R}^d \to \mathbb{R}$ & value function defined as the optimal cost-to-go (aim to minimize) \\
$\psi_t(\boldsymbol{x}):\mathbb{R}^d \to \mathbb{R}$ & Lagrange multiplier which defines the value (aim to maximize) \\
$\Phi(\boldsymbol{x}):\mathbb{R}^d \to \mathbb{R}$ & terminal cost \\
$\mathcal{A}$ & uncontrolled generator \\
$\mathcal{A}^u$ & controlled generator \\
$t$ & time coordinate ranging $t\in [0,T]$\\
$s$ & another time coordinate, sometimes used to denote the reverse time coordinate $s=T-t$ ranging $s\in [0,T]$ when specified\\
$\tau$ & another time coordinate, typically used for integration \\
$\Delta t$ & finite time step \\
$\theta, \phi$ & neural network parameters \\
$M$ or $M_t$ & transport map that maps distributions to new distributions via the push-forward $M_{\#}p=p'$ \\
\hline
\end{longtable}
\end{center}
\endgroup
\vspace{0.5em}

\newpage

\section{The Static Schrödinger Bridge Problem}
\label{sec:static-sb}
We start our deep dive into Schrödinger bridges from its origins in the classical optimal mass transport (OMT) problem in both Monge’s and Kantorovich’s formulations. This perspective naturally frames the static Schrödinger bridge problem as a probabilistic regularization of the OMT problem, linking deterministic transport theory with stochastic processes and entropy minimization\footnote{As the focus of this guide is on Schrödinger bridges for generative modeling, we analyze optimal transport and \textbf{entropic optimal transport} with the intent of connecting it with the dynamic problem in Section \ref{sec:dynamic-sb}. For a more comprehensive exposition of entropic OT, we refer the readers to \citet{nutz2021introduction}.}.

\subsection{The Optimal Mass Transport Problem}
\label{subsec:optimal-mass-transport}
The origins of the Schrödinger bridge problem can be traced back to the problem of \boldtext{optimal mass transport} (OMT) \citep{villani2021topics}, which defines an optimal mapping between points in one distribution to another. In this section, we introduce Monge and Kantorovich's formulation of the OMT problem, which serves as a starting point for understanding how to \textit{optimally} transport mass between probability distributions. 

Consider two probability distributions $\pi_0\in \mathcal{P}(\mathcal{X})$ and $\pi_0\in \mathcal{P}(\mathcal{Y})$ and a transport cost function $c(\boldsymbol{x},\boldsymbol{y}): \mathcal{X}\times \mathcal{Y}\to\mathbb{R}$ which determines the cost of transporting a unit of mass from $\boldsymbol{x}\in \mathcal{X}$ to $\boldsymbol{y}\in \mathcal{Y}$. We also define the space of \boldtext{transport maps} $M: \mathcal{X}\to \mathcal{Y}$ that generate $\pi_T$ as the pushforward of $\pi_0$ (i.e., $M_{\#}\pi_0=\pi_T$) as: 
\begin{align}
    \mathcal{T}(\pi_0, \pi_T)=\{M: \mathcal{X}\to \mathcal{Y}\;|\;M_{\#}\pi_0=\pi_T\}
\end{align}
Among all transport maps, Monge's OMT problem seeks the optimal map $M^\star$ that minimizes the transport cost.

\begin{definition}[Monge's Optimal Mass Transport Problem]
    Given probability distributions $\pi_0\in \mathcal{P}(\mathcal{X})$ and $\pi_0\in \mathcal{P}(\mathcal{Y})$ and cost function $c: \mathcal{X}\times \mathcal{Y}\to\mathbb{R}$, Monge's OMT problem aims to find the \textbf{optimal transport map} $M^\star$ that minimizes:
    \begin{align}
    \inf_{M\in \mathcal{T}(\pi_0, \pi_T)}\left\{\int_{\mathcal{X}}c(\boldsymbol{x}, M(\boldsymbol{x}))d\pi_0(\boldsymbol{x})\bigg|M:\mathcal{X}\to \mathcal{Y}\text{ and }M_{\#}\pi_0=\pi_T\right\}\tag{Monge's OMT Problem}\label{eq:monge-problem}
    \end{align}
\end{definition}
Crucially, (\ref{eq:monge-problem}) can be an ill-posed problem and yield no solution when the mass of the distribution $\pi_0$ is concentrated at a single point (e.g., a dirac delta) and $\pi_T$ is concentrated at multiple points (e.g., two dirac deltas with fractional mass), which means mass must be \textit{split} to reach both targets, and the space of deterministic transport maps $\mathcal{T}(\pi_0, \pi_T)$ is empty. To avoid the possibility of ill-posedness, Kantorovich defined the OMT problem as an optimization over the space of \textbf{optimal couplings} $\pi_{0,T}\in \Pi(\pi_0, \pi_T)$ defined as: 
\begin{align}
    \Pi(\pi_0, \pi_T)=\left\{\pi_{0,T}\in\mathcal{P}( \mathcal{X}\times \mathcal{Y})\;\big|(\text{proj}_{\mathcal{X}})_{\#}\pi_{0,T}=\pi_0, (\text{proj}_{\mathcal{Y}})_{\#}\pi_{0,T}=\pi_T\right\}\label{eq:space-of-couplings}  
\end{align}
where $(\text{proj}_{\mathcal{X}})_{\#}\pi_{0,T}$ is the $\mathcal{X}$-marginal of $\pi_{0,T}$ and vice versa. $\Pi(\pi_0, \pi_T)$ is never empty since it always contains the product measure $\pi_0\otimes \pi_T$. 

\begin{definition}[Kantorovich's Optimal Mass Transport Problem]
    Given probability distributions $\pi_0\in \mathcal{P}(\mathcal{X})$ and $\pi_0\in \mathcal{P}(\mathcal{Y})$ and cost function $c: \mathcal{X}\times \mathcal{Y}\to\mathbb{R}$, Kantorovich's OMT problem aims to find the \textbf{optimal coupling} $\pi^\star_{0,T}$ that minimizes:
    \begin{align}
    \inf_{\pi_{0,T}}\bigg\{\int_{\mathcal{X}\times \mathcal{Y}}c(\boldsymbol{x}, \boldsymbol{y})d\pi_{0,T}(\boldsymbol{x}, \boldsymbol{y})\bigg|\pi_{0,T}\in \Pi(\pi_0, \pi_T)\bigg\}\tag{Kantorovich's OMT Problem}\label{eq:kant-omt-problem}
    \end{align}
\end{definition}

While Kantorovich’s formulation of the OMT problem guarantees existence of an optimal coupling, the resulting problem remains a linear optimization problem in $\pi_{0,T}\in \mathcal{X}\times \mathcal{Y}$ that is susceptible to non-unique and deterministic mappings, where for each state $\boldsymbol{x}\sim \pi_0$, the distribution of $\pi_{0,T}(\boldsymbol{x}, \cdot)\in \mathcal{P}(\mathcal{Y})$ is sparse and concentrated at few points. To obtain a smoother and more statistically meaningful coupling, we will consider \textbf{entropy regularization} as a strategy for optimizing a \textit{stochastic coupling} where each state $\boldsymbol{x}\sim \pi_0$ yields a smooth distribution $\pi_{0,T}(\boldsymbol{x},\cdot)$ over possible mappings.

\subsection{Entropy on Probability Spaces}
\label{subsec:entropy-probability-spaces}
Before we introduce \textbf{entropic optimal transport}, which leads to the \textbf{static Schrödinger bridge} problem, we need to establish the concept of \boldtext{entropy}. Conceptually, entropy measures the \textit{uncertainty} of one probability measure with respect to another probability measure. We formalize this in the following definition\footnote{In this section, we use $p, \tilde{p}\in \mathcal{P}(\mathcal{X})$ denote arbitrary probability measures on $\mathcal{X}$ and $q, \tilde{q}\in \mathcal{P}(\mathcal{Y})$ denote arbitrary probability measures on $\mathcal{Y}$ rather than their definitions in the preface.}.

\begin{figure}
    \centering
    \includegraphics[width=0.8\linewidth]{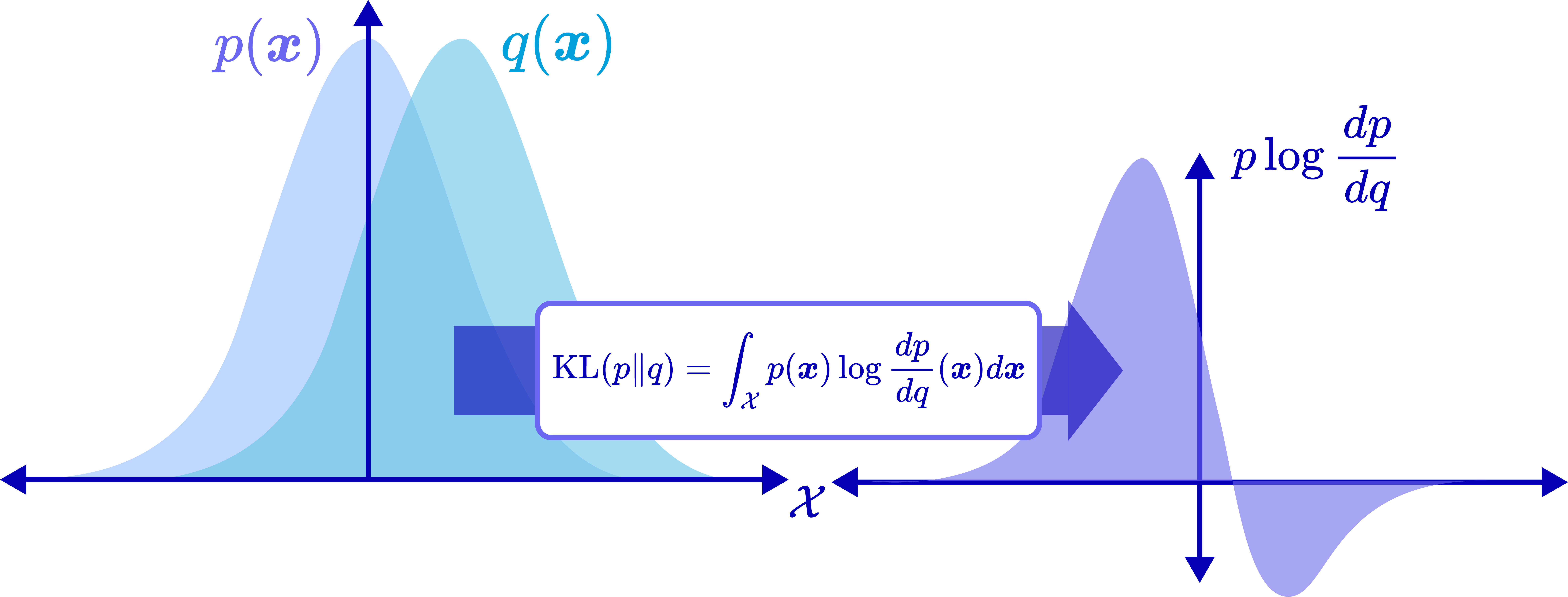}
    \caption{\textbf{Illustration of Relative Entropy or Kullback-Leibler (KL) Divergence.} The KL divergence $\text{KL}(p\|q)$ between two 1-dimensional probability distributions $p, q\in \mathcal{P}(\mathcal{X})$ on the state space $\mathcal{X}\subseteq\mathbb{R}$. The \textbf{left} shows the independent distributions and the \textbf{right} shows the signed contributions of $p\log \frac{dp}{dq}$ which yields the KL divergence when integrated, quantifying how well $q$ approximates $p$.}
    \label{fig:kl-divergence}
\end{figure}

\begin{definition}[Entropy Between Probability Measures]\label{def:entropy}
    Consider two probability measures $p, q\in \mathcal{P}(\mathcal{X})$ on a measureable state space $\mathcal{X}\subseteq\mathbb{R}^d$. The \textbf{entropy} of $p$ \textit{relative to} the measure $q$ is defined as:
    \begin{align}
        \text{KL}(p\|q):=\begin{cases}
            \mathbb{E}_{p}[\log \frac{dp}{dq}]&q\ll p\\
            \infty &q\not\ll p
        \end{cases}
    \end{align}
    where $q\ll p$ means that $q$ is absolutely continuous with respect to $p$ such that for all measurable sets $A\subseteq \mathcal{X}$, we have $p(A)=0\implies q(A)=0$. $\text{KL}(\cdot\|\cdot)$ is also known as the \textbf{Kullback-Leibler (KL) divergence}.
\end{definition}

Intuitively, the Kullback-Leibler (KL) divergence $\text{KL}(p\|q)$ measures the \textit{expected log-likelihood ratio under $p$}, which indicates the excess \textit{distributional mismatch} incurred when data generated under $p$ is modeled under $q$ (Figure \ref{fig:kl-divergence}). In contrast to typical distance measures, the KL divergence is \textbf{asymmetric}, where $\text{KL}(p\|q)$ is a measure of \textit{how well $q$ approximates $p$}, which is not equivalent to how well $p$ approximates $q$. Therefore, we use KL divergence in entropy-regularized OT and Schrödinger bridges to penalize deviations from the \textit{reference} coupling or path measure. 

We now establish the \boldtext{chain rule for KL divergence}, which decomposes the KL divergence of a joint measure into the divergence between its marginals and the expected conditional divergence. This property allows us to decompose the KL divergence under marginal constraints into a constant and a variable term.

\begin{lemma}[Chain Rule of KL Divergences]\label{lemma:chain-rule-kl}
    Given two joint probability measures $\pi_{X,Y}, \pi'_{X,Y}\in \mathcal{P}(\mathcal{X}\times \mathcal{Y})$ that are absolutely continuous $\pi_{0,T}\ll \pi'_{0,T}$. Denote the $\mathcal{X}$-marginals as $\pi_X:= \int_{\mathcal{Y}}\pi_{0,T}d\boldsymbol{y}$ and $\pi'_X:= \int_{\mathcal{Y}}\pi'_{0,T}d\boldsymbol{y}$ and the conditional distribution on $\mathcal{Y}$ given $\boldsymbol{x}\in \mathcal{X}$ as $\pi_{Y|X}$ and $\pi'_{Y|X}$. Then, the \textbf{KL divergence decomposes into}:
    \begin{align}
        \text{KL}(\pi_{X,Y}\|\pi'_{X,Y})=\text{KL}(\pi_X\|\pi_Y)+\mathbb{E}_{\boldsymbol{x}\sim \pi_X}\left[\text{KL}\left(\pi_{Y|X}(\cdot|\boldsymbol{x})\|\pi'_{Y|X}(\cdot|\boldsymbol{x})\right)\right]\tag{KL Divergence Chain Rule}\label{eq:chain-rule-kl}
    \end{align}
\end{lemma}

\textit{Proof. }This proof follows simply from the definition of KL divergence:
\begin{small}
\begin{align}
    \text{KL}(\pi_{X,Y}\|\pi'_{X,Y})&=\int_{\mathcal{X}\times \mathcal{Y}}\log \frac{d\pi_{X,Y}}{d\pi'_{X,Y}}=\int_{\mathcal{X}\times \mathcal{Y}}\log \frac{d\pi_{X}}{d\pi'_{X}}\bluetext{\frac{d\pi_{Y|X}}{d\pi_{Y|X}'}}d\pi_{X,Y}\nonumber\\
    &=\underbrace{\int_{\mathcal{X}\times \mathcal{Y}}\log \frac{d\pi_{X}}{d\pi'_{X}}d\pi_{X,Y}}_{\text{only dependent on }\boldsymbol{x}}+\int_{\mathcal{X}}\bluetext{\underbrace{\int_{\mathcal{Y}}\log\bluetext{\frac{d\pi_{Y|X}}{d\pi_{Y|X}'}}d\pi_{Y|X}}_{=\text{KL}(\pi_{Y|X}\|\pi_{Y|X}')}}d\pi_{X}\nonumber\\
    &=\int_{\mathcal{X}}\log \frac{d\pi_{X}}{d\pi'_{X}}d\pi_{X}+\int_{\mathcal{X}}\text{KL}(\pi_{Y|X}\|\pi_{Y|X}')d\pi_{X}\nonumber\\
    &=\boxed{\text{KL}(\pi_X\|\pi'_X)+\mathbb{E}_{\pi_X}\left[\text{KL}(\pi_{Y|X}\|\pi_{Y|X}')\right]}
\end{align}
\end{small}
which concludes the proof of the KL chain rule. \hfill $\square$

The KL chain rule reveals how relative entropy decomposes across joint and conditional distributions. An immediate consequence of this decomposition is the \boldtext{data processing inequality}, which formalizes the idea that applying the same stochastic transformation to two probability measures cannot increase their divergence. In other words, when two distributions are passed through the same Markov kernel, the resulting distributions become at least as indistinguishable as the originals.

\begin{lemma}[Data Processing Inequality]\label{lemma:data-processing-inequality}
    Given two probability measures $p,q\in \mathcal{P}(\mathcal{X})$ and a Markov kernel $\mathcal{K}:\mathcal{X}\to \mathcal{P}(\mathcal{Y})$ that maps states $\boldsymbol{x}\in \mathcal{X}$ to probability measures on $\mathcal{Y}$, define $\tilde{q}, \tilde{p}\in \mathcal{P}(\mathcal{Y})$ defined as:
    \begin{align}
        \tilde{q}(\boldsymbol{y}):=\int_{\mathcal{X}}q(\boldsymbol{x})\mathcal{K}(\boldsymbol{x},\boldsymbol{y})d\boldsymbol{x}, \quad \tilde{p}(\boldsymbol{y}):=\int_{\mathcal{X}}p(\boldsymbol{x})\mathcal{K}(\boldsymbol{x},\boldsymbol{y})d\boldsymbol{x}
    \end{align}
    Then, they satisfy $\text{KL}(\tilde{p}\|\tilde{q})\leq \text{KL}(p\|q)$.
\end{lemma}

\textit{Proof.} We prove this by applying the KL chain rule from Lemma \ref{lemma:chain-rule-kl}. Let the joint probability measures transformed via the Markov kernel $\mathcal{K}$ be denoted as:
\begin{small}
\begin{align}
    P(d\boldsymbol{x},d\boldsymbol{y}):=p(d\boldsymbol{x})\mathcal{K}(\boldsymbol{x},d\boldsymbol{y}), \quad Q(d\boldsymbol{x},d\boldsymbol{y}):=p(d\boldsymbol{x})\mathcal{K}(\boldsymbol{x},d\boldsymbol{y}) 
\end{align}
\end{small}
where $\tilde{p}$ and $\tilde{q}$ are the $\mathcal{Y}$-marginals of $P$ and $Q$, respectively:
\begin{small}
\begin{align}
    \tilde{p}(\boldsymbol{y})=\int_{\mathcal{X}} P(d\boldsymbol{x},d\boldsymbol{y})d\boldsymbol{x},\quad  \tilde{q}(\boldsymbol{y})=\int_{\mathcal{X}} Q(d\boldsymbol{x},d\boldsymbol{y})d\boldsymbol{x}
\end{align}
\end{small}
Applying the KL chain rule from Lemma \ref{lemma:chain-rule-kl} to the joint measures, we get:
\begin{small}
\begin{align}
    \text{KL}(P\|Q)=\text{KL}(\tilde{p}\|\tilde{q})+\underbrace{\mathbb{E}_{\tilde{p}}[\text{KL}(P_{X|Y}\|Q_{X|Y})]}_{\geq 0}\geq \text{KL}(\tilde{p}\|\tilde{q})\label{eq:data-process-proof1}
\end{align}
\end{small}
Since the KL divergence is always non-negative, it follows that $\text{KL}(\tilde{p}\|\tilde{q})\leq \text{KL}(P\|Q)$. Since $P$ and $Q$ are obtained by applying the same kernel $\mathcal{K}$ to $p$ and $q$, the conditional law $\mathcal{K}(\boldsymbol{x},d\boldsymbol{y})$ cancels in the KL divergence as:
\begin{small}
\begin{align}
    \text{KL}(P\|Q)&=\int_{\mathcal{X}\times \mathcal{Y}}\log \left(\bluetext{\frac{dP}{dQ}(\boldsymbol{x},\boldsymbol{y})}\right)P(d\boldsymbol{x},d\boldsymbol{y})=\int_{\mathcal{X}\times \mathcal{Y}}\log \bigg(\bluetext{\frac{p(d\boldsymbol{x})}{q(d\boldsymbol{x})}\underbrace{\frac{\mathcal{K}(\boldsymbol{x},d\boldsymbol{y})}{\mathcal{K}(\boldsymbol{x},d\boldsymbol{y})}}_{=1}}\bigg)\pinktext{P(d\boldsymbol{x},d\boldsymbol{y})}\nonumber\\
    &=\int_{\mathcal{X}\times \mathcal{Y}}\log \left(\bluetext{\frac{dp}{dq}(\boldsymbol{x})}\right)\pinktext{p(d\boldsymbol{x})\mathcal{K}(\boldsymbol{x},d\boldsymbol{y})}=\int_{\mathcal{X}\times \mathcal{Y}}\log \left(\frac{dp}{dq}(\boldsymbol{x})\right)p(d\boldsymbol{x})\underbrace{\pinktext{\int_{ \mathcal{Y}}\mathcal{K}(\boldsymbol{x},d\boldsymbol{y})}}_{=1}=\text{KL}(p\|q)\label{eq:data-process-proof2}
\end{align}
\end{small}
Combining (\ref{eq:data-process-proof1}) and (\ref{eq:data-process-proof2}), we have:
\begin{small}
\begin{align}
    \boxed{\text{KL}(\tilde{p}\|\tilde{q})\leq \text{KL}(p\|q)}
\end{align}
\end{small}
which proves the data processing inequality.\hfill $\square$

Now that we have developed the notion of entropy on probability spaces as a measure of uncertainty and relative entropy as a measure of distinguishability between distributions, we are ready to introduce the \textbf{entropic optimal transport} (EOT) problem. In this formulation, entropy is used to regularize the classical optimal mass transport problem by penalizing deviation from a reference coupling, yielding a transport objective that admits a \textit{stable} and \textit{unique} solution while remaining close, in relative entropy, to a prescribed baseline law.

\subsection{Entropic Optimal Transport Problem}
\label{subsec:entropic-ot}
Having defined the \textbf{optimal mass transport} (OMT) problem as finding a transport plan between marginal distributions that minimizes a cost function, we now extend these ideas to define the \boldtext{entropic optimal transport} (EOT) problem \citep{nutz2021introduction, leonard2013survey}, which moves closer to the formulation of the Schrödinger bridge problem.

Just like the OMT problem, the EOT problem seeks an optimal transport plan $\pi_{0,T}^\star\in \Pi(\pi_0, \pi_T)$ between marginals $\pi_0$ and $\pi_T$. However, unlike the OMT problem, \textit{optimality} is no longer determined solely by the transport cost $c(\boldsymbol{x},\boldsymbol{y})$ but also an \textbf{entropy regularization term} $\text{KL}(\pi_{0,T}\|q)$, where $q\in \mathcal{P}(\mathcal{X}\times\mathcal{Y})$ is a reference coupling measure defined on the product space.

\begin{definition}[Entropic Optimal Transport (EOT) Problem]\label{def:eot-problem}
    Consider two probability distributions $\pi_0\in \mathcal{P}(\mathcal{X})$ and $\pi_T\in \mathcal{P}(\mathcal{Y})$ and a cost function $c(\boldsymbol{x},\boldsymbol{y}):\mathcal{X}\times \mathcal{Y}\to [0,\infty)$ that defines the cost of transporting units of mass from states $\boldsymbol{x}\in \mathcal{X}$ to states $\boldsymbol{y}\in \mathcal{Y}$. The \textbf{entropic optimal transport} (EOT) problem aims to find the optimal transport plan $\pi^\star_{0,T}$ that minimizes:
    \begin{align}
        \inf_{\pi_{0,T}\in \Pi(\pi_0, \pi_T)}\left\{\int_{\mathcal{X}\times \mathcal{Y}}c(\boldsymbol{x},\boldsymbol{y})d\pi_{0,T}(\boldsymbol{x},\boldsymbol{y})+\alpha \text{KL} (\pi_{0,T}\|q)\right\}\tag{Entropic OT Problem}\label{eq:entropic-ot-problem}
    \end{align}
    where $\text{KL}(\cdot \|\cdot)$ is the \textbf{KL divergence} (Definition \ref{def:entropy}) between a transport plan and a fixed probability measure $q\in \mathcal{P}(\mathcal{X}\times\mathcal{Y})$.
\end{definition}

From (\ref{eq:entropic-ot-problem}), we start to observe characteristics of the Schrödinger problem emerge. Specifically, we highlight the KL divergence as a measure of distance from a pre-defined, \textit{reference} coupling, that prevents the optimal transport coupling from diverging too far from it. This penalty is scaled by a constant $\alpha\in \mathbb{R}$, where small values result in couplings that allow solutions to diverge more, and large values penalize even small deviations from the reference dynamics. First, we can define the cost functional minimized in (\ref{eq:entropic-ot-problem}) with the reference measure set to $dq:=d(\pi_0\otimes \pi_T)$ as:
\begin{align}
    \mathcal{F}(\pi_{0,T}):=\int_{\mathcal{X}\times\mathcal{Y}}c(\boldsymbol{x},\boldsymbol{y})d\pi_{0,T}(\boldsymbol{x},\boldsymbol{y})+\text{KL}(\pi_{0,T}\|\pi_0\otimes \pi_T)\tag{EOT Functional}\label{eq:eot-functional}
\end{align}
Now, we want to define a new reference measure that reduces the (\ref{eq:entropic-ot-problem}) into a simple KL-minimization problem. To do this, we need a way to absorb the cost integral into the KL minimization objective. Since the reference measure in the KL divergence is defined in the denominator of the logarithm, we scale the product reference measure by the exponential of the negative cost function and some unknown constant $\alpha$, which transforms into an additive cost when taking the KL. Concretely, we define a \textit{tilted} reference measure as:
\begin{align}
    d\tilde{q}:=\frac{e^{-c}}{\alpha}d(\pi_0\otimes \pi_T), \quad \alpha:=\int_{\mathcal{X}\times\mathcal{Y}} d\tilde{q}=\int_{\mathcal{X}\times\mathcal{Y}}e^{-c}d(\pi_0\otimes \pi_T)
\end{align}
where $\alpha$ is defined such that $dq$ integrates to one. Then, we can expand $\text{KL}(\pi_{0,T}\|q)$ as:
\begin{small}
\begin{align}
    \text{KL}(\pi_{0,T}\|\tilde{q})&=\int_{\mathcal{X}\times\mathcal{Y}}\log \left(\frac{d\pi_{0,T}}{d\tilde{q}}\right)d\pi_{0,T}=\int_{\mathcal{X}\times \mathcal{Y}}\log \left(\frac{d\pi_{0,T}}{\bluetext{\frac{e^{-c}}{\alpha}d(\pi_0\otimes \pi_T)}}\right)d\pi_{0,T}\nonumber\\
    &=\int_{\mathcal{X}\times \mathcal{Y}}\left[\log \left(\frac{d\pi_{0,T}}{\bluetext{d(\pi_0\otimes \pi_T)}}\right)+\bluetext{\log (\alpha e^c)}\right]d\pi_{0,T}\nonumber\\
    &= \underbrace{\int_{\mathcal{X}\times \mathcal{Y}}\log \left(\frac{d\pi_{0,T}}{\bluetext{d(\pi_0\otimes \pi_T)}}\right)d\pi_{0,T}}_{\text{KL}(\pi_{0,T}\|(\pi_0\otimes\pi_T))}+\int_{\mathcal{X}\times \mathcal{Y}}c(\boldsymbol{x}, \boldsymbol{y})d\pi_{0,T}+ \underbrace{\int_{\mathcal{X}\times \mathcal{Y}} \log \alpha d\pi_{0,T}}_{= \log \alpha}\nonumber\\
    &=\text{KL}(\pi_{0,T}\|(\pi_0\otimes\pi_T))+\int_{\mathcal{X}\times \mathcal{Y}}c(\boldsymbol{x}, \boldsymbol{y})d\pi_{0,T}+\log \alpha
\end{align}
\end{small}
where we separate the integrands and use the fact that integrating a constant $\log \alpha$ over a density returns the constant. Now, we can rearrange to recover an equivalent expression for $\mathcal{F}(\pi_{0,T})$ in (\ref{eq:eot-functional}) defined as:
\begin{small}
\begin{align}
    \text{KL}(\pi_{0,T}\|\tilde{q})-\underbrace{\log \alpha}_{\text{constant}}&=\text{KL}(\pi_{0,T}\|(\pi_0\otimes\pi_T))+\int_{\mathcal{X}\times \mathcal{Y}}c(\boldsymbol{x}, \boldsymbol{y})d\pi_{0,T}\bluetext{=:\mathcal{F}(\pi_{0,T})}
\end{align}
\end{small}
by our definition in (\ref{eq:eot-functional}). Since $\log\alpha$ is a constant independent of $\pi_{0,T}$, it does not affect the minimization problem, and the (\ref{eq:entropic-ot-problem}) reduces to a KL minimization problem:
\begin{align}
    \inf_{\pi_{0,T}\in \Pi(\pi_0, \pi_T)}\left\{\bluetext{\int_{\mathcal{X}\times \mathcal{Y}}c(\boldsymbol{x},\boldsymbol{y})d\pi_{0,T}(\boldsymbol{x},\boldsymbol{y})+ \text{KL} (\pi_{0,T}\|q)}\right\}=\inf_{\pi_{0,T}\in \Pi(\pi_0, \pi_T)}\bluetext{\text{KL} (\pi_{0,T}\|\tilde{q})}\label{eq:entropic-ot-kl-proj}
\end{align}
From this result, we can interpret the \textbf{entropic OT problem} as a KL projection of a reference measure $d\tilde{q}$ onto the set of couplings with the prescribed marginals $\Pi(\pi_0, \pi_T)$. This is exactly the variational structure underlying the static Schrödinger bridge (SB) problem. 

Before we move on to defining the SB problem, we first consider \textit{why} we need entropy regularization in the first place over simply minimizing a cost functional, as in the optimal mass transport problem. 

\purple[Motivations for Entropy Regularization]{
Although we have established the definition of \textbf{entropy} and how it can be used to extend the optimal mass transport problem to an \textit{entropy-regularized} optimal transport problem, a question remains: \textit{Why do we need entropy-regularization?} The answer can be condensed into the following points:
\begin{enumerate}
    \item[(i)] \textbf{Stochastic Coupling.} The entropy term penalizes large probabilities under the coupling $\pi_{0,T}$ which produces a large value for $\pi_{0,T}\log \pi_{0,T}$. While OMT is susceptible to convergence on a singular plan that is concentrated on a lower-dimensional set, such that for each initial point $\boldsymbol{x}_0\sim \pi_0$, all mass is transported to a single or small set of terminal points $\boldsymbol{y}$ rather than a smooth density over $\pi_T$. Entropy regularization ensures that the solution is a \textbf{stochastic coupling} in which mass is spread smoothly over the joint probability space $\mathcal{X}\times\mathcal{Y}$, rather than a deterministic map.
    \item[(ii)] \textbf{Strict Convexity and Uniqueness.} The (\ref{eq:monge-problem}) is linear in $\pi_{0,T}$, such that multiple minimizers can exist with the same total cost. This makes the problem extremely sensitive to initialization, as different initial conditions can yield vastly different solutions. However, since the entropy function defined as $\pi_{0,T}\mapsto \int \pi_{0,T}\log \pi_{0,T}$ is \textit{convex in $\pi_{0,T}$}, the entropic OT problem has a \textbf{unique minimizer} $\pi^\star_{0,T}$ regardless of the initialization.
    \item[(iii)] \textbf{Generalization of Optimal Mass Transport.} As the regularization constant $\varepsilon\to 0$, the problem reduces to the optimal mass transport problem. Therefore, the entropic OT problem can be seen as a generalization of the OMT problem with tunable entropic regularization, where $\varepsilon\to \infty$ makes the cost function negligible, and the solution is equal to the reference coupling $q$.
\end{enumerate}
}

Now that we have built the foundations of optimal transport of distributions and entropy regularization, we are finally ready to begin our discussion on Schrödinger bridges, starting with the \textbf{static Schrödinger bridge problem}.

\subsection{Static Schrödinger Bridge Problem}
\label{subsec:static-sbp}
In this section, we will formally define the \boldtext{static Schrödinger bridge (SB) problem}, which is closely related to the entropic OT problem discussed in Section \ref{subsec:entropic-ot}. We introduce the notion of \boldtext{Schrödinger potentials} $(\varphi, \hat{\varphi})$\footnote{Note that in this section and in Section~\ref{subsec:sinkhorn-algorithm} we use the notation $(\varphi^\star,\hat{\varphi}^\star)$ for the Schrödinger potentials while discussing optimization over $\varphi$ and $\hat{\varphi}$. In the remainder of the guide, $(\varphi^\star,\hat{\varphi}^\star)$ denotes the unique optimal Schrödinger potentials.} that \textit{uniquely} solve a pair of equations, called the \textbf{Schrödinger system}, and simultaneously define the unique static Schrödinger bridge solution. These ideas will form the foundation of our discussion of Sinkhorn's algorithm in Section \ref{subsec:sinkhorn-algorithm}.

\begin{definition}[Static Schrödinger Bridge Problem]\label{def:static-sb-problem}
    Given two marginal distribution constraints $\pi_0\in \mathcal{P}(\mathcal{X})$ and $\pi_T\in \mathcal{P}(\mathcal{Y})$, define the set of all couplings $ \Pi(\pi_0, \pi_T)\subset\mathcal{P}(\mathcal{X}\times \mathcal{Y})$ with $\pi_0$ and $\pi_T$ as its $\mathcal{X}$- and $\mathcal{Y}$-marginals, respectively. Given a reference measure $q\sim \pi_0\otimes \pi_T$\footnote{the symbol $\sim$ denotes mutual absolute continuity where for $q(S)=0\iff (\pi_0\otimes \pi_T)(S)=0$ for all measurable sets $S\subseteq \mathcal{X}\times\mathcal{Y}$}. Then, the \textbf{static Schrödinger bridge} (SB) problem defined as:
    \begin{align}
        \pi^\star_{0,T}=\underset{\pi_{0,T}\in \Pi(\pi_0, \pi_T)}{\arg\min}\text{KL}(\pi_{0,T}\|q)\tag{Static SB Problem}\label{eq:static-sb}
    \end{align}
    where the minimizer $\pi^\star_{0,T}$ is \textit{unqiue} and is called the \textbf{static Schrödinger bridge} between $\pi_0$ and $\pi_T$. Furthermore, when the reference coupling takes the form:
    \begin{small}
    \begin{align}
        q(\boldsymbol{x},\boldsymbol{y}):=\frac{e^{-c(\boldsymbol{x},\boldsymbol{y})}}{\alpha}(\pi_0\otimes\pi_T) (\boldsymbol{x},\boldsymbol{y})
    \end{align}
    \end{small}
    The static SB problem coincides with the entropic optimal transport (EOT) problem:
    \begin{small}
    \begin{align}
        \pi^\star_{0,T}=\underset{\pi_{0,T}\in \Pi(\pi_0, \pi_T)}{\arg\min}\left\{\int_{\mathcal{X}\times \mathcal{Y}}c(\boldsymbol{x},\boldsymbol{y})d\pi_{0,T}(\boldsymbol{x},\boldsymbol{y})+\alpha \text{KL} (\pi_{0,T}\|q)\right\}
    \end{align}
    \end{small}
\end{definition}

The static Schrödinger bridge problem can therefore be viewed as the entropy projection of a reference coupling onto the set of couplings with fixed marginals, and is equivalent to the form of the entropic OT derived in (\ref{eq:entropic-ot-kl-proj}).

Although the static Schrödinger bridge problem is formulated as an optimization over couplings, its solution admits a simple multiplicative structure. In particular, the optimal coupling can be written as a reweighted version of the reference measure using two functions known as the \boldtext{Schrödinger potentials} $(\varphi, \hat{\varphi})$ which together solve the \boldtext{Schrödinger system}. Next, we will define the Schrödinger potentials $(\varphi, \hat{\varphi})$ and show that the pair that solves the Schrödinger system is \textit{unique} and satisfies the marginal constraints.

\begin{proposition}[Schrödinger Potentials]
    Consider a reference measure $q\ll \pi_0\otimes \pi_T$, which implies that the Radon-Nikodym derivative $\frac{dq}{d(\pi_0\otimes \pi_T)}$ is well-defined and given some cost function $c:\mathcal{X}\times\mathcal{Y} \to \mathbb{R}$ as:
    \begin{align}
        \frac{dq}{d(\pi_0\otimes \pi_T)}=e^{-c(\boldsymbol{x},\boldsymbol{y})}
    \end{align}
    Then given two functions $\varphi:\mathcal{X}\to \mathbb{R}$ and $\hat{\varphi}:\mathcal{Y}\to \mathbb{R}$, called the \textbf{Schrödinger potentials}, we define the \textbf{Schrödinger system} as the pair of equations:
    \begin{align}
    \begin{cases}
        \varphi(\boldsymbol{x})=-\log\int_{\mathcal{Y}}e^{\hat{\varphi}(\boldsymbol{y})-c(\boldsymbol{x},\boldsymbol{y})}\pi_T(d\boldsymbol{y})\\
        \hat{\varphi}(\boldsymbol{y})=-\log\int_{\mathcal{X}}e^{\varphi(\boldsymbol{x})-c(\boldsymbol{x},\boldsymbol{y})}\pi_0(d\boldsymbol{x})\tag{Schrödinger System}\label{eq:schrodinger-system}
    \end{cases}
    \end{align}
    where the solution $\hat{\pi}_{0,T}$ solves the \textbf{static Schrödinger bridge problem} (Definition \ref{def:static-sb-problem}) and satisfy:
    \begin{align}
        &d\hat{\pi}_{0,T}(d\boldsymbol{x}, d\boldsymbol{y})=e^{\varphi(\boldsymbol{x})+\hat{\varphi}(\boldsymbol{y})-c(\boldsymbol{x}, \boldsymbol{y})}d(\pi_0\otimes \pi_T)\\
        &\hat{\pi}_0(d\boldsymbol{x})=\pi_0(d\boldsymbol{x}), \quad \hat{\pi}_T(d\boldsymbol{y})=\pi_T(d\boldsymbol{y})
    \end{align}
    where $\hat{\pi}_0(\boldsymbol{x}):= \int_{\mathcal{Y}}\hat{\pi}_{0,T}(d\boldsymbol{x}, d\boldsymbol{y})$ is the first marginal of $\hat{\pi}_{0,T}$ and $\hat{\pi}_{T}(\boldsymbol{y}):=\int_{\mathcal{X}}\hat{\pi}_{0,T}(d\boldsymbol{x}, d\boldsymbol{y})$ is the second marginal. Furthermore, the potentials $(\varphi, \hat{\varphi})$ are \textbf{unique} up to an additive constant.
\end{proposition}

\textit{Proof.} Rather than a direct proof, we present an intuitive derivation of the Schrödinger equations starting from the definition of the static SB problem and its Lagrangian form in Step 1. In Step 2, we prove the uniqueness of $(\varphi, \hat{\varphi})$. 

\textbf{Step 1: Derivation of Schrödinger System from Static SB Problem.}
First, recall the form \textbf{static SB} problem from Definition \ref{def:static-sb-problem} given by:
\begin{small}
\begin{align}
    \min_{\hat{\pi}_{0,T}\in \Pi(\pi_0, \pi_T)}\text{KL}(\hat{\pi}_{0,T}\|q)=\min_{\hat{\pi}_{0,T}\in \Pi(\pi_0, \pi_T)}\int_{\mathcal{X}\times\mathcal{Y}}\log \frac{d\hat{\pi}_{0,T}}{dq}d\hat{\pi}_{0,T}(d\boldsymbol{x},d\boldsymbol{y}), \quad dq:=e^{-c(\boldsymbol{x},\boldsymbol{y})}d(\pi_0\otimes \pi_T)
\end{align}
\end{small}
We can rewrite this objective using Lagrangian multipliers $\varphi(\boldsymbol{x})$ and $\hat{\varphi}(\boldsymbol{y})$ to enforce the marginal constraints as:
\begin{align}
    \mathcal{L}(\hat{\pi}_{0,T}):=\int_{\mathcal{X}\times\mathcal{Y}}\log \left(\frac{d\hat{\pi}_{0,T}}{dq}\right)d\hat{\pi}_{0,T}+\int_{\mathcal{X}}\varphi(\boldsymbol{x})(\pi_0-\hat{\pi}_0)(d\boldsymbol{x})+\int_{\mathcal{Y}}\varphi(\boldsymbol{y})(\pi_T-\hat{\pi}_T)(d\boldsymbol{y})
\end{align}
where we define $\hat{\pi}_0(\boldsymbol{x}):= \int_{\mathcal{Y}}\hat{\pi}_{0,T}(d\boldsymbol{x}, d\boldsymbol{y})$ as the first marginal and $\hat{\pi}_{T}(\boldsymbol{y}):=\int_{\mathcal{X}}\hat{\pi}_{0,T}(d\boldsymbol{x}, d\boldsymbol{y})$ is the second marginal. To obtain the \textbf{optimality conditions}, we take the functional derivative of $\mathcal{L}(\hat{\pi}_{0,T})$ with respect to $\hat{\pi}_{0,T}$ and set the first variation to zero to get:
\begin{small}
\begin{align}
    &\frac{\delta}{\delta\hat{\pi}_{0,T}}\mathcal{L}(\hat{\pi}_{0,T})=\log \left(\frac{d\hat{\pi}_{0,T}}{dq}\right)+1-\varphi(\boldsymbol{x})-\hat{\varphi}(\boldsymbol{y})=0\implies \log \left(\frac{d\hat{\pi}_{0,T}}{dq}\right)=-1+\varphi(\boldsymbol{x})+\hat{\varphi}(\boldsymbol{y})\nonumber\\
    &\implies \frac{d\hat{\pi}_{0,T}}{dq}=e^{-1}e^{\varphi(\boldsymbol{x})+\hat{\varphi}(\boldsymbol{y})}\nonumber\\
    &\implies d\hat{\pi}_{0,T}(d\boldsymbol{x},d\boldsymbol{y})=e^{\varphi(\boldsymbol{x})+\hat{\varphi}(\boldsymbol{y})-c(\boldsymbol{x},\boldsymbol{y})}\bluetext{d(\pi_0\otimes\pi_T)}=e^{\varphi(\boldsymbol{x})+\hat{\varphi}(\boldsymbol{y})-c(\boldsymbol{x},\boldsymbol{y})}\bluetext{\pi_0(d\boldsymbol{x})\pi_T(d\boldsymbol{y})}
\end{align}
\end{small}
where the constant $e^{-1}$ can be absorbed into the Lagrange multipliers. Now, we can compute the expression for each marginal of $\hat{\pi}_{0,T}$ by integration:
\begin{small}
\begin{align}
    &\hat{\pi}_0(d\boldsymbol{x}) =\int_{\mathcal{Y}} d\hat{\pi}_{0,T}(d\boldsymbol{x},d\boldsymbol{y})=\int_{\mathcal{Y}}e^{\varphi(\boldsymbol{x})+\hat{\varphi}(\boldsymbol{y})-c(\boldsymbol{x},\boldsymbol{y})}\pi_0(d\boldsymbol{x})\pi_T(d\boldsymbol{y})=e^{\varphi(\boldsymbol{x})}\pi_0(d\boldsymbol{x})\int_{\mathcal{Y}}e^{\hat{\varphi}(\boldsymbol{y})-c(\boldsymbol{x},\boldsymbol{y})}\pi_T(d\boldsymbol{y})\nonumber\\
    &\implies e^{\varphi(\boldsymbol{x})}\bluetext{\pi_0(d\boldsymbol{x})}\int_{\mathcal{Y}}e^{\hat{\varphi}(\boldsymbol{y})-c(\boldsymbol{x},\boldsymbol{y})}\pi_T(d\boldsymbol{y})=\bluetext{\pi_0(d\boldsymbol{x})}\implies \bluetext{\log }\left(e^{\varphi(\boldsymbol{x})}\int_{\mathcal{Y}}e^{\hat{\varphi}(\boldsymbol{y})-c(\boldsymbol{x},\boldsymbol{y})}\pi_T(d\boldsymbol{y})\right)=\bluetext{\log (1)}\nonumber\\
    &\implies  \bluetext{\varphi(\boldsymbol{x})=-\log \int_{\mathcal{Y}}e^{\hat{\varphi}(\boldsymbol{y})-c(\boldsymbol{x},\boldsymbol{y})}\pi_T(d\boldsymbol{y})}\label{eq:schrodinger-system1}
\end{align}
\end{small}
Similarly, for the second marginal, we have:
\begin{small}
\begin{align}
    &\hat{\pi}_T(d\boldsymbol{y}) =\int_{\mathcal{X}} d\hat{\pi}_{0,T}(\boldsymbol{x},\boldsymbol{y})=\int_{\mathcal{X}}e^{\varphi(\boldsymbol{x})+\hat{\varphi}(\boldsymbol{y})-c(\boldsymbol{x},\boldsymbol{y})}\pi_0(d\boldsymbol{x})\pi_T(d\boldsymbol{y})=e^{\hat{\varphi}(\boldsymbol{y})}\pi_T(d\boldsymbol{y})\int_{\mathcal{X}}e^{\varphi(\boldsymbol{x})-c(\boldsymbol{x},\boldsymbol{y})}\pi_0(d\boldsymbol{x})\nonumber\\
    &\implies e^{\hat{\varphi}(\boldsymbol{y})}\pi_T(d\boldsymbol{y})\int_{\mathcal{X}}e^{\varphi(\boldsymbol{y})-c(\boldsymbol{x},\boldsymbol{y})}\pi_0(d\boldsymbol{x})=\bluetext{\pi_T(d\boldsymbol{y})}\implies \bluetext{\log }\left(e^{\hat{\varphi}(\boldsymbol{x})}\int_{\mathcal{X}}e^{\varphi(\boldsymbol{x})-c(\boldsymbol{x},\boldsymbol{y})}\pi_0(d\boldsymbol{x})\right)=\bluetext{\log (1)}\nonumber\\
    &\implies  \bluetext{\hat{\varphi}(\boldsymbol{y})=-\log \int_{\mathcal{Y}}e^{\varphi(\boldsymbol{x})-c(\boldsymbol{x},\boldsymbol{y})}\pi_0(d\boldsymbol{x})}\label{eq:schrodinger-system2}
\end{align}
\end{small}
Together, (\ref{eq:schrodinger-system1}) and (\ref{eq:schrodinger-system2}) recover the (\ref{eq:schrodinger-system}) and establishes the proof that the Schrödinger equation for $\varphi(\boldsymbol{x})$ enforces the first marginal constraint $\hat{\pi}_0=\pi_0$ and the Schrödinger equation for $\hat{\varphi}(\boldsymbol{x})$ enforces the second marginal constraint $\hat{\pi}_T=\pi_T$.

\textbf{Step 2: Proving Uniqueness of Schrödinger Potentials.} 
From our derivation in Step 1, we show that the solution to the Lagrangian form of the static SB problem can be written as:
\begin{align}
    \forall (\boldsymbol{x}, \boldsymbol{y}) \in \mathcal{X}\times\mathcal{Y}, \quad\frac{d\hat{\pi}_{0,T}}{dq}=e^{\varphi(\boldsymbol{x})+\hat{\varphi}(\boldsymbol{y})}\implies \log \left(\frac{d\hat{\pi}_{0,T}}{dq}\right)=\varphi(\boldsymbol{x})+\hat{\varphi}(\boldsymbol{y})
\end{align}
To prove that $(\varphi, \hat{\varphi})$ are \textit{unique} for the solution $\hat{\pi}_{0,T}$, consider two new functions $\varphi':\mathcal{X}\to \mathbb{R}$ and $\hat{\varphi}':\mathcal{Y}\to \mathbb{R}$ which also satisfy:
\begin{small}
\begin{align}
     \quad\log \left(\frac{d\hat{\pi}_{0,T}}{dq}\right)=\varphi(\boldsymbol{x})+\hat{\varphi}(\boldsymbol{y})=\varphi'(\boldsymbol{x})+\hat{\varphi}'(\boldsymbol{y})\implies \varphi(\boldsymbol{x})-\varphi'(\boldsymbol{x})=\hat{\varphi}'(\boldsymbol{y})-\hat{\varphi}(\boldsymbol{y}), \quad \bluetext{q\text{-a.s.}}\label{eq:sb-potential-unique1}
\end{align}
\end{small}
Since the identity holds for $q\text{-a.s.}$\footnote{holds for all $(\boldsymbol{x},\boldsymbol{y})\in \mathcal{X}\times \mathcal{Y}$ outside the null set where $q(N)=0$} on $\mathcal{X}\times \mathcal{Y}$, by Fubini's theorem\footnote{which states that integrating over the product space is equal to integrating over each space sequentially}, there exist $\boldsymbol{x}^\star\in \mathcal{X}$ and $\boldsymbol{y}^\star\in \mathcal{Y}$, such that the identity (\ref{eq:sb-potential-unique1}) holds for $\pi_0\text{-a.e.}$ (almost everywhere) in $\boldsymbol{x}$ for fixed $\boldsymbol{y}=\boldsymbol{y}^\star$ and $\pi_T\text{-a.e.}$ in $\boldsymbol{y}$ for fixed $\boldsymbol{x}=\boldsymbol{x}^\star$. Since the left-hand side is dependent \textit{only on $\boldsymbol{x}$} and the right-hand side is dependent \textit{only on $\boldsymbol{y}$}, fixing $\boldsymbol{y}=\boldsymbol{y}^\star$ and defining $a:=\hat{\varphi}'(\boldsymbol{y}^\star)-\hat{\varphi}(\boldsymbol{y}^\star)$ yields:
\begin{align}
    &\varphi(\boldsymbol{x})-\varphi'(\boldsymbol{x}) =\hat{\varphi}'(\boldsymbol{y}^\star)-\hat{\varphi}(\boldsymbol{y}^\star)=:a, \quad \pi_0\text{-a.s.}\\
    &\implies \begin{cases}
        \varphi(\boldsymbol{x})=\varphi'(\boldsymbol{x})+a & \bluetext{\pi_0\text{-a.s.}}\\
        \hat{\varphi}(\boldsymbol{y})=\hat{\varphi}'(\boldsymbol{y})-a & \bluetext{\pi_T\text{-a.s.}}
    \end{cases}
\end{align}
which proves that $(\varphi, \hat{\varphi})$ are \textit{unique up to a constant} $a\in \mathbb{R}$.\hfill $\square$

Using these results, we can show that the Schrödinger potentials $(\varphi, \hat{\varphi})$ also define the \boldtext{unique optimal coupling} $\pi^\star_{0,T}$ that solves (\ref{eq:static-sb}).

\begin{proposition}[Solution to Static SB Problem]\label{prop:solution-static-sb}
    Given two marginals $\pi_0\in \mathcal{P}(\mathcal{X})$ and $\pi_T\in \mathcal{P}(\mathcal{Y})$ and a reference coupling $q\sim \pi_0\otimes \pi_T$, assume the set of finite entropy couplings $\Pi(\pi_0, \pi_T)\neq \varnothing$. Let $\hat{\pi}_{0,T}$ be a coupling that satisfies:
    \begin{align}
        \log \left(\frac{d\hat{\pi}_{0,T}}{dq}\right)=\varphi\oplus\hat{\varphi}, \quad q\text{-a.s.}
    \end{align}
    for measurable functions $\varphi:\mathcal{X}\to \mathbb{R}$ and $\hat{\varphi}:\mathcal{Y}\to \mathbb{R}$. Then $\hat{\pi}_{0,T}=\pi^\star_{0,T}$ that solves (\ref{eq:static-sb}) and yields a \textbf{constant} map:
    \begin{align}
        \quad \pi_{0,T}\mapsto \mathbb{E}_{\pi_{0,T}}\left[\log \frac{d\pi^\star_{0,T}}{dq}\right], \quad \forall \pi_{0,T}\in \Pi(\pi_0, \pi_T)\cup \{\pi_{0,T}^\star\}
    \end{align}
\end{proposition}
\textit{Proof.} \textbf{Step 1: Prove the Constant Map.} 
Assume $\varphi \in L^1(\pi_0)$ and $\hat{\varphi} \in L^1(\pi_T)$ are measurable and integrable. Consider the optimal coupling $\pi^\star_{0,T}$ which satisfies:
\begin{small}
\begin{align}
    \log \left(\frac{d\pi^\star_{0,T}}{dq}\right)=\varphi\oplus\hat{\varphi}:=\varphi(\boldsymbol{x})+\hat\varphi(\boldsymbol{y}), \quad q\text{-a.s.}\label{eq:static-sb-proof1}
\end{align}
\end{small}
Given (\ref{eq:static-sb-proof1}) and absolute continuity $\pi_{0,T}\ll q$, we can write:
\begin{small}
\begin{align}
    \mathbb{E}_{\pi_{0,T}}\left[\log \frac{d\pi_{0,T}^\star}{dq}\right]=\mathbb{E}_{\pi_{0,T}}\left[\varphi(\boldsymbol{x})+\hat\varphi(\boldsymbol{y})\right]&=\int_{\mathcal{X}\times \mathcal{Y}}\varphi(\boldsymbol{x})\pi_{0,T}(d\boldsymbol{x},d\boldsymbol{y})+\int_{\mathcal{X}\times \mathcal{Y}}\hat\varphi(\boldsymbol{y})\pi_{0,T}(d\boldsymbol{x},d\boldsymbol{y})\nonumber\\
    &=\underbrace{\int_{\mathcal{X}}\varphi(\boldsymbol{x})\pi_{0}(d\boldsymbol{x})+\int_{\mathcal{Y}}\hat\varphi(\boldsymbol{y})\pi_{T}(d\boldsymbol{y})}_{\text{independent on coupling }\pi_{0,T}}
\end{align}
\end{small}
where the last equality follows from the fact that $\pi_{0,T}$ has marginals $\pi_0$ and $\pi_T$. Since the final equality is only dependent on $\pi_0$ and $\pi_T$ and not the coupling $\pi_{0,T}$, we have that the map:
\begin{align}
    \pi_{0,T}\mapsto \mathbb{E}_{\pi_{0,T}}\left[\log \frac{d\pi_{0,T}^\star}{dq}\right]\label{eq:static-sb-proof2}
\end{align}
is constant over $\Pi(\pi_0, \pi_T)$. 

\textbf{Step 2: Prove Optimality.} 
Now, we aim to show that $\pi^\star_{0,T}$ uniquely minimizes $\text{KL}(\pi_{0,T}\|q)$ over $\Pi$. For any coupling $\pi_{0,T}\in \Pi(\pi_0, \pi_T)$, we can decompose the KL divergence $\text{KL}(\pi_{0,T}\|q)$ with respect to $\pi^\star_{0,T}$ as:
\begin{small}
\begin{align}
    \text{KL}(\pi_{0,T}\|q)&=\mathbb{E}_{\pi_{0,T}}\left[\log \frac{d\pi_{0,T}}{dq}\right]=\underbrace{\mathbb{E}_{\pi_{0,T}}\left[\log \frac{d\pi_{0,T}}{d\pi^\star_{0,T}}\right]}_{=\text{KL}(\pi_{0,T}\|\pi_{0,T}^\star)}+\underbrace{\mathbb{E}_{\pi_{0,T}}\left[\log \frac{d\pi^\star_{0,T}}{dq}\right]}_{\text{constant in }\pi_{0,T}\text{ given (\ref{eq:static-sb-proof2})}}\nonumber\\
    &=\bluetext{\text{KL}(\pi_{0,T}\|\pi_{0,T}^\star)}+C
\end{align}
\end{small}
where we use the constant map property from Step 1 (\ref{eq:static-sb-proof2}). Since the KL divergence is non-negative with equality $\text{KL}(\pi_{0,T}\|\pi_{0,T}^\star)=0$ if and only if $\pi_{0,T}=\pi^\star_{0,T}$, we have that $\pi^\star_{0,T}$ is the unqiue solution to (\ref{eq:static-sb}).\hfill $\square$

From this proof, we observe that $(\varphi, \hat\varphi)$ depend only on the marginal constraints and thus act as Lagrange multipliers on the marginal constraints. This naturally leads to an alternative \boldtext{dual formulation of the static SB problem}, which aims to maximize the potentials to enforce the marginal constraints.

\begin{theorem}[Dual Formulation of Static SB Problem]\label{theorem:dual-static-sb}
    The equivalent \textbf{dual formulation} of (\ref{eq:static-sb}) is defined as:
    \begin{small}
    \begin{align}
        \underbrace{\inf_{\pi_{0,T}\in \Pi(\pi_0, \pi_T)}\text{KL}(\pi_{0,T}\|q)}_{\text{primal problem}}=\underbrace{\sup_{\varphi, \hat{\varphi}}\bigg\{\int_{\mathcal{X}}\varphi d\pi_0+\int_{\mathcal{Y}}\hat{\varphi} d\pi_T-\int_{\mathcal{X}\times\mathcal{Y}} e^{\varphi\oplus \hat{\varphi}}dq+1\bigg\}}_{\text{dual problem}}\tag{Strong Duality}\label{eq:strong-duality}
    \end{align}
    \end{small}
    where the \textbf{supremum} is achieved by the Schrödinger potentials $(\varphi^\star, \hat{\varphi}^\star)$ which define associated Schrödinger bridge coupling $\pi^\star_{0,T}$ and satisfy:
    \begin{small}
    \begin{align}
        \text{KL}(\pi^\star_{0,T}\|q)=\inf_{\pi_{0,T}\in \Pi(\pi_0, \pi_T)}\text{KL}(\pi_{0,T}\|q)=\int_{\mathcal{X}}\varphi^\star d\pi_0+\int_{\mathcal{Y}}\hat{\varphi}^\star d\pi_T, \quad \frac{d\pi^\star_{0,T}}{dq}=e^{\varphi^\star\oplus\hat{\varphi}^\star} \quad q\text{-a.s.}
    \end{align}
    \end{small}
    where $\varphi^\star\in L^1(\pi_0), \hat{\varphi}^\star\in L^1(\pi_T)$ and $\varphi^\star\oplus \hat{\varphi}^\star$ is unique.
\end{theorem}

\textit{Proof.} First, we will prove \boldtext{weak duality} using Fenchel's inequality before we prove strong duality, which establishes the relationship between the optimal Schrödinger potentials $(\varphi^\star, \hat{\varphi}^\star)$ and the Schrödinger bridge coupling $\pi_{0,T}^\star$.

\textbf{Step 1: Prove Weak Duality.} 
Weak duality states that for any measurable functions $\varphi\in L^1(\pi_0)$, $\hat{\varphi}\in L^1(\pi_T)$ and coupling $\pi_{0,T}\in \Pi(\pi_0, \pi_T)$, the following inequality holds:
\begin{align}
    \text{KL}(\pi_{0,T}\|q)\geq\int_{\mathcal{X}}\varphi d\pi_0+\int_{\mathcal{Y}}\hat{\varphi}d\pi_T-\int_{\mathcal{X}\times \mathcal{Y}}e^{\varphi\oplus\hat{\varphi}}dq+1\tag{Weak Duality}\label{eq:weak-duality}
\end{align}
First, we define the \textbf{Fenchel's inequality} which states that for any convex function $f$ and any $\alpha, \beta$, the inequality is satisfied:
\begin{align}
    f(\alpha)\geq \alpha \beta-f^\star(\beta), \quad\text{where}\quad f^\star(\beta):=\sup_{\alpha\geq 0}\{\alpha \beta-f(\alpha)\}
\end{align}
where $f^\star(\beta)$ is the \textbf{Fenchel conjugate} of $f(\alpha)$ which defines all supporting affine functions for $f(\alpha)$\footnote{the expression $\alpha \beta-f(\alpha)$ arises from defining any affine function $\alpha\beta-\gamma\leq f(\alpha), \forall \alpha$ that lies below the convex function at all points $\alpha$, rearranging to get $\gamma \geq \alpha \beta-h(\alpha), \forall \alpha$, and defining the Fenchel conjugate as the smallest possible $c$ to get $f^\star(\alpha):=\sup_{\alpha\geq 0}\{\alpha\beta-f(\alpha)\}$}. Since the KL divergence can be written as an integral over $\alpha\log \alpha$, we define the convex function $f(\alpha)=\alpha\log \alpha-\alpha$ which yields the clean convex conjugate $f^\star(\alpha)=\sup_{\alpha\geq 0}\{\alpha\beta-\alpha\log \alpha+\alpha\}=e^{\beta}$, which can be derived by computing the derivative of $\alpha\beta-\alpha\log \alpha+\alpha$ and setting it to zero\footnote{explicitly, we let $g(\alpha):=\alpha\beta-\alpha\log \alpha+\alpha$, take the derivative $g'(\alpha) =\beta-\log \alpha-1+1$, and set to zero to get $\beta=\log \alpha\implies \alpha=e^{\beta}$, which is the supremum since $f$ is convex.}. Then, by Fenchel's inequality, we have:
\begin{align}
    \alpha\log \alpha-\alpha\geq \beta\alpha-e^{\beta}\implies \alpha\log \alpha-\beta \alpha \geq \alpha-e^{\beta}, \quad \forall \alpha\geq 0, \beta \in \mathbb{R}
\end{align}
By setting $\alpha:=\frac{d\pi_{0,T}}{dq}$ and $\beta:=\varphi \oplus \hat{\varphi}$, we see that $\alpha\log \alpha$ recovers the KL divergence and $e^{\beta}$ becomes $e^{\varphi\oplus \hat{\varphi}}$. Then, we can expand the KL divergence as:
\begin{small}
\begin{align}
    \text{KL}(\pi_{0,T}\|q)&=\int_{\mathcal{X}\times\mathcal{Y}}\frac{d\pi_{0,T}}{dq}\log \left(\frac{d\pi_{0,T}}{dq}\right)dq\nonumber\\
    &=\int_{\mathcal{X}\times\mathcal{Y}}\frac{d\pi_{0,T}}{dq}\log \left(\frac{d\pi_{0,T}}{dq}\right)dq\bluetext{+\int_{\mathcal{X}\times\mathcal{Y}}(\varphi\oplus \hat{\varphi})\frac{d\pi_{0,T}}{dq}dq-\int_{\mathcal{X}\times\mathcal{Y}}(\varphi\oplus \hat{\varphi})\frac{d\pi_{0,T}}{dq}dq}\nonumber\\
    &=\int_{\mathcal{X}\times\mathcal{Y}}(\varphi\oplus \hat{\varphi})\frac{d\pi_{0,T}}{dq}dq+\int_{\mathcal{X}\times\mathcal{Y}}\pinktext{\underbrace{\left[\frac{d\pi_{0,T}}{dq}\log \left(\frac{d\pi_{0,T}}{dq}\right)-(\varphi\oplus\hat{\varphi})\frac{d\pi_{0,T}}{dq}\right]}_{=:\alpha\log\alpha-\beta \alpha }}dq\nonumber\\
    &\geq\int_{\mathcal{X}\times\mathcal{Y}}(\varphi\oplus \hat{\varphi})d\pi_{0,T}+\int_{\mathcal{X}\times\mathcal{Y}}\pinktext{\left[\frac{d\pi_{0,T}}{dq}-e^{\varphi\oplus \hat{\varphi}}\right]}dq\nonumber\\
    &\geq \int_{\mathcal{X}}\varphi d\pi_0+\int_{\mathcal{Y}}\hat{\varphi} d\pi_T+\underbrace{\int_{\mathcal{X}\times\mathcal{Y}}\pinktext{d\pi_{0,T}}}_{=1}-\int_{\mathcal{X}\times\mathcal{Y}}\pinktext{e^{\varphi\oplus \hat{\varphi}}}dq\nonumber\\
    &\geq\int_{\mathcal{X}}\varphi d\pi_0+\int_{\mathcal{Y}}\hat{\varphi} d\pi_T-\int_{\mathcal{X}\times\mathcal{Y}}e^{\varphi\oplus \hat{\varphi}}dq+1
\end{align}
\end{small}
which concludes our proof of weak duality. 

\textbf{Step 2: Prove Strong Duality. }
To prove \boldtext{strong duality}, we recall the definition of the optimal Schrödinger potentials $(\varphi^\star, \hat{\varphi}^\star)$ and the optimal Schrödinger bridge coupling $\pi^\star_{0,T}\in \Pi(\pi_0, \pi_T)$ which satisfy:
\begin{align}
    \frac{d\pi^\star_{0,T}}{dq}=e^{\varphi^\star \oplus \hat{\varphi}^\star}\quad q\text{-a.s.}\quad \text{where}\quad \varphi^\star\in L^1(\pi_0), \hat{\varphi}\in L^1(\pi_T)\label{eq:sb-duality-proof1}
\end{align}
Given the inequality in (\ref{eq:weak-duality}), we can take the infimum on the left-hand side and the supremum on the right-hand side, which preserves the inequality to get:
\begin{align}
    \bluetext{\inf_{\pi_{0,T}\in\Pi(\pi_0, \pi_T)}}\text{KL}(\pi_{0,T}\|q)\geq\pinktext{\sup_{\varphi\in L^1(\pi_0), \hat{\varphi}\in L^1(\pi_T)}}\underbrace{\left\{\int_{\mathcal{X}}\varphi d\pi_0+\int_{\mathcal{Y}}\hat{\varphi}d\pi_T-\int_{\mathcal{X}\times \mathcal{Y}}e^{\varphi\oplus\hat{\varphi}}dq+1\right\}}_{=:G(\varphi, \hat{\varphi})}
\end{align}
where we let $G(\varphi, \hat{\varphi})$ define the dual objective. Now, we will expand both sides of the inequality for the \textbf{optimal} $\varphi^\star, \hat{\varphi}^\star, \pi^\star_{0,T}$. For the \textbf{left-hand side}, we apply (\ref{eq:sb-duality-proof1}) to derive the KL divergence for the optimal coupling $\pi^\star_{0,T}$ as: 
\begin{align}
    \text{KL}(\pi^\star_{0,T}\|q)&=\int_{\mathcal{X}\times \mathcal{Y}}\bluetext{\log \left(\frac{d\pi^\star_{0,T}}{dq}\right)}d\pi_{0,T}^\star=\int_{\mathcal{X}\times\mathcal{Y}}\bluetext{(\varphi^\star\oplus\hat{\varphi})}d\pi^\star_{0,T}\nonumber\\
    &=\int_{\mathcal{X}}\bluetext{\varphi^\star d\pi^\star_{0,T}}+\int_{\mathcal{Y}}\bluetext{\hat{\varphi}^\star d\pi^\star_{0,T}}=\int_{\mathcal{X}}\bluetext{\varphi^\star d\pi_0}+\int_{\mathcal{Y}}\bluetext{\hat{\varphi}^\star d\pi_T}\label{eq:sb-duality-proof2}
\end{align}
For the \textbf{right-hand side}, we use the definition $d\pi^\star_{0,T}=e^{\varphi^\star\oplus\hat{\varphi}}dq$ to get:
\begin{align}
    G(\varphi^\star, \hat{\varphi}^\star)&=\int_{\mathcal{X}}\varphi d\pi_0+\int_{\mathcal{Y}}\hat{\varphi}d\pi_T-\bluetext{\int_{\mathcal{X}\times \mathcal{Y}}e^{\varphi\oplus\hat{\varphi}}dq}+1\nonumber\\
    &=\int_{\mathcal{X}}\varphi d\pi_0+\int_{\mathcal{Y}}\hat{\varphi}d\pi_T-\bluetext{\underbrace{\int_{\mathcal{X}\times \mathcal{Y}}d\pi_{0,T}^\star}_{=1}}+1=\int_{\mathcal{X}}\varphi d\pi_0+\int_{\mathcal{Y}}\hat{\varphi}d\pi_T\label{eq:sb-duality-proof3}
\end{align}
Since (\ref{eq:sb-duality-proof2}) and (\ref{eq:sb-duality-proof3}) are equal for the optimal $\varphi^\star, \hat{\varphi}^\star, \pi_{0,T}^\star$, it follows that the infimum is $\text{KL}(\pi^\star_{0,T}\|q)$ and equal to the supremum at optimality:
\begin{align}
    \bluetext{\inf_{\pi_{0,T}\in\Pi(\pi_0, \pi_T)}}\text{KL}(\pi_{0,T}\|q)=\text{KL}(\pi^\star_{0,T}\|q)=G(\varphi^\star, \hat{\varphi}^\star)=\pinktext{\sup_{\varphi\in L^1(\pi_0), \hat{\varphi}\in L^1(\pi_T)}}G(\varphi, \hat{\varphi})
\end{align}
which proves (\ref{eq:strong-duality}) and establishes that equality between objectives occurs exactly at the Schrödinger bridge solution $\varphi^\star, \hat{\varphi}^\star, \pi_{0,T}^\star$. Since $-e^{\varphi\oplus\hat{\varphi}}$ is strictly concave in $(\varphi\oplus \hat{\varphi})$, the sum $(\varphi^\star\oplus \hat{\varphi}^\star)$ is the \textit{unique} maximizer, which also implies that $\varphi$ and $\hat{\varphi}$ are \textit{unique up to a constant}.  \hfill $\square$

In this section, we reformulated the static SB problem from the problem of optimizing an optimal transport coupling $\pi^\star_{0,T}$ on the high-dimensional product space $\mathcal{X}\times\mathcal{Y}$ to a problem of determining two scalar SB potential functions $(\varphi, \hat\varphi)$ defined on the marginals $\pi_0$ and $\pi_T$ which \textit{uniquely} characterize the optimal bridge. Leveraging this simplified structure, we can now turn to establishing a tractable algorithm for solving the static SB problem.

\subsection{Sinkhorn's Algorithm}
\label{subsec:sinkhorn-algorithm}
\begin{figure}
    \centering
    \includegraphics[width=\linewidth]{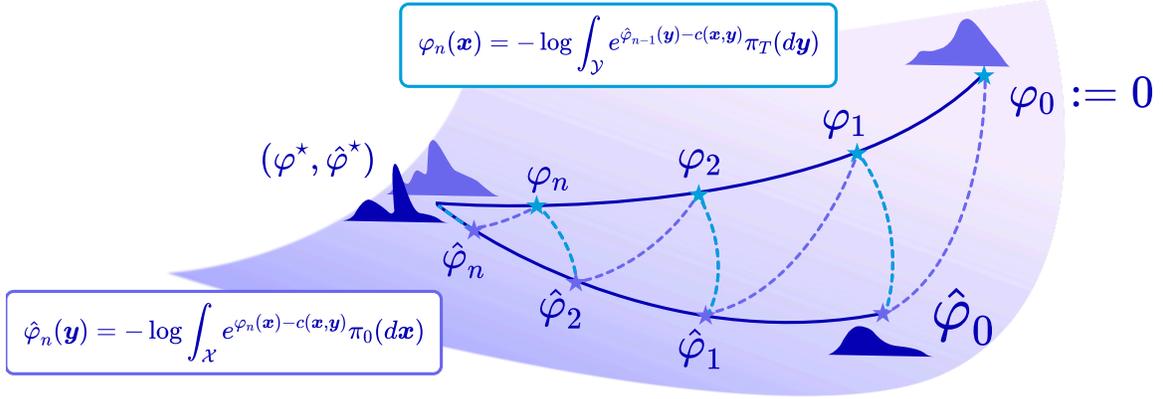}
    \caption{\textbf{Illustration of Sinkhorn's Algorithm.} Starting from an initial potential (e.g. $\varphi_0:=0$), Sinkhorn's algorithm alternates updates of the dual potentials $\varphi_n$ and $\hat\varphi_n$ via log-integral transforms involving the transport cost $c(\boldsymbol{x},\boldsymbol{y})$ and the marginal constraints $\pi_0$ and $\pi_T$. Each alternating step enforces one marginal constraint while preserving the entropic structure, and the sequence $(\varphi_n, \hat\varphi_n)$ converges to the optimal dual pair $(\varphi^\star, \hat\varphi^\star)$ that uniquely defines the static Schrödinger bridge coupling $\pi^\star_{0,T}$.}
    \label{fig:sinkhorn}
\end{figure}

We now introduce the classical algorithm used to solve the static Schrödinger system defined in (\ref{eq:schrodinger-system}), known as \textbf{Sinkhorn's algorithm} in optimal transport \citep{sinkhorn1967diagonal, knight2008sinkhorn, nutz2021introduction, cuturi2013sinkhorn} or the \textbf{Iterative Proportional Fitting} (IPF) procedure in statistics \citep{fortet1940resolution, kullback1968probability, ruschendorf1995convergence}.

Recall that (\ref{eq:schrodinger-system}) consists of two equations for the pair of Schrödinger potentials $(\varphi, \hat{\varphi})$ defined as:
\begin{align}
    \varphi(\boldsymbol{x})=-\log\int_{\mathcal{Y}}e^{\hat{\varphi}(\boldsymbol{y})-c(\boldsymbol{x},\boldsymbol{y})}\pi_T(d\boldsymbol{y})\tag{First Potential}\label{eq:first-potential}\\
    \hat{\varphi}(\boldsymbol{y})=-\log\int_{\mathcal{X}}e^{\varphi(\boldsymbol{x})-c(\boldsymbol{x},\boldsymbol{y})}\pi_0(d\boldsymbol{x})\tag{Second Potential}\label{eq:second-potential}
\end{align}
Optimizing both $\varphi$ and $\hat{\varphi}$ simultaneously would lead to a mismatch, as they depend on each other. Therefore, it is natural to consider an \textbf{alternating optimization scheme} that optimizes one of the potentials with the other potential fixed. This is exactly the intuition behind \boldtext{Sinkhorn's algorithm} (Figure \ref{fig:sinkhorn}). The algorithm starts by initializing the first potential at some $\varphi=\varphi_0$ and defines an alternating optimization sequence $\{\varphi_n, \hat{\varphi}_n\}_{n\geq 0}$. We can also consider this alternating sequence as maximizing the \textbf{dual problem} defined in (\ref{eq:strong-duality}) with the objective:
\begin{align}
    G(\varphi, \hat{\varphi}):=\int_{\mathcal{X}}\varphi d\pi_0+\int_{\mathcal{Y}}\hat{\varphi} d\pi_T-\int_{\mathcal{X}\times\mathcal{Y}} e^{\varphi\oplus \hat{\varphi}}dq+1\tag{Dual Objective}\label{eq:dual-objective}
\end{align}
Combining both the Schrödinger system equations and the dual problem, we define the alternating optimization sequence $\{\varphi_n, \hat{\varphi}_n\}_{n\geq 0}$ with $\varphi_0:=0$ as:
\begin{enumerate}
    \item [(i)] Solve $\hat{\varphi}_n$ using (\ref{eq:second-potential}) with $\varphi:=\varphi_n$. Equivalently, solve $\hat{\varphi}_n:=\arg\max G(\varphi_n, \cdot)$ using (\ref{eq:dual-objective}).
    \item [(ii)] Solve $\varphi_{n+1}$ using (\ref{eq:first-potential}) with $\hat{\varphi}:=\hat{\varphi}_n$. Equivalently, solve $\varphi_{n+1}:=\arg\max G(\cdot , \hat\varphi_n)$ using (\ref{eq:dual-objective}).
\end{enumerate}
Since (\ref{eq:dual-objective}) is \textbf{strictly concave} with respect to both $\varphi$ and $\hat{\varphi}$, each iteration strictly increases the objective $G(\varphi_n, \hat{\varphi}_n)< G(\varphi_{n+1}, \hat{\varphi}_n)< G(\varphi_{n+1}, \hat{\varphi}_{n+1})$, unless the optimal pair $(\varphi^\star, \hat{\varphi}^\star)$ is reached, where $G(\varphi_n, \hat{\varphi}_n)=G(\varphi_{n+1}, \hat{\varphi}_n)=G(\varphi_{n+1}, \hat{\varphi}_{n+1})$. Furthermore, the coupling at each iteration is defined as:
\begin{align}
    &d\pi_{0,T}(\varphi, \hat{\varphi}):=e^{\varphi\oplus\hat{\varphi}}dq=e^{\varphi\oplus\hat{\varphi}-c}d(\pi_0\otimes \pi_T)\quad \text{define: } \begin{cases}
        \pi^{2n}_{0,T}:=\pi_{0,T}(\varphi_{n}, \hat{\varphi}_n)\\
        \pi_{0,T}^{2n-1}:=\pi_{0,T}(\varphi_{n}, \hat{\varphi}_{n-1})\\
        \pi^{-1}_{0,T}=\pi_{0,T}(0,0)=e^{-c}d(\pi_0\otimes \pi_T)=:dq
    \end{cases}\label{eq:sinkhorn-density2}
\end{align}

Given our definition above, we can now define a few \textbf{properties of Sinkhorn's algorithm} which will aid us in proving its convergence. 

\begin{lemma}[Properties of Sinkhorn's Algorithm]\label{lemma:sinkhorn-properties}
    The sequence of potentials $\{\varphi_n, \hat{\varphi}_n\}_{n\geq 0}$, where $\varphi_n\in L^1(\pi_0)$ and $\hat{\varphi}_n\in L^1(\pi_T)$ are integrable, and coupled densities $\{\pi_{0,T}^{2n}, \pi^{2n-1}_{0,T}\}_{n\geq 0}$ defined by the Sinkhorn iterations satisfy the following properties:
    \begin{enumerate}
        \item [(i)] Each KL step equals a difference of potentials:
        \begin{small}
        \begin{align}
            \text{KL}\left(\pi_{0,T}^{2n}\|\pi_{0,T}^{2n-1}\right)=\int_{\mathcal{Y}}(\hat{\varphi}_n-\hat{\varphi}_{n-1})\pi_T(d\boldsymbol{y}), \quad \text{KL}\left(\pi_{0,T}^{2n+1}\|\pi_{0,T}^{2n}\right)=\int_{\mathcal{X}}(\varphi_{n+1}-\varphi_n)\pi_0(d\boldsymbol{x})\nonumber
        \end{align}
        \end{small}
        \item[(ii)] The total dual potential equals the total accumulated KL cost:
        \begin{small}
        \begin{align}
            \pi_T(\hat{\varphi}_n)=\sum_{k=0}^{n}\text{KL}\left(\pi_{0,T}^{(2k)}\|\pi_{0,T}^{(2k-1)}\right), \quad \pi_0(\varphi_n)=\sum_{k=0}^{n-1}\text{KL}\left(\pi_{0,T}^{(2k+1)}\|\pi_{0,T}^{(2k)}\right) \nonumber
        \end{align}
        where $\pi_0(\varphi_n)$ and $\pi_T(\hat{\varphi}_n)$ are non-negative and increasing.
        \end{small}
    \end{enumerate}
\end{lemma}

\textit{Proof.} We will prove each part of the Lemma in steps.

\textbf{Step 1: Proof of Property (i). } 
To prove this, we start with the definition of the KL divergence:
\begin{small}
\begin{align}
    \text{KL}\left(\pi_{0,T}^{2n}\|\pi_{0,T}^{2n-1}\right)=\int_{\mathcal{X}\times\mathcal{Y}}\log \left(\bluetext{\frac{d\pi^{2n}_{0,T}}{d\pi^{2n-1}_{0,T}}}\right)d\pi^{2n}_{0,T}, \quad \text{KL}\left(\pi_{0,T}^{2n+1}\|\pi_{0,T}^{2n}\right)=\int_{\mathcal{X}\times\mathcal{Y}}\log \left(\pinktext{\frac{d\pi^{2n+1}_{0,T}}{d\pi^{2n}_{0,T}}}\right)d\pi^{2n+1}_{0,T}\label{eq:sinkhorn-density4}
\end{align}
\end{small}
From (\ref{eq:sinkhorn-density2}), we can decompose the log ratio of the density updates as:
\begin{small}
\begin{align}
    \bluetext{\frac{d\pi^{2n}_{0,T}}{d\pi^{2n-1}_{0,T}}}&=\frac{e^{\varphi_n(\boldsymbol{x})+\bluetext{\hat{\varphi}_{n}(\boldsymbol{y})}-c(\boldsymbol{x}, \boldsymbol{y})}d(\pi_0\otimes \pi_T)}{e^{\varphi_n(\boldsymbol{x})+\bluetext{\hat{\varphi}_{n-1}(\boldsymbol{y})}-c(\boldsymbol{x}, \boldsymbol{y})}d(\pi_0\otimes \pi_T)}=\frac{e^{\hat{\varphi}_n(\boldsymbol{y})}}{e^{\hat{\varphi}_{n-1}(\boldsymbol{y})}}=\bluetext{e^{\hat{\varphi}_n(\boldsymbol{y})-\hat{\varphi}_{n-1}(\boldsymbol{y})}}\label{eq:sinkhorn-density3}\\
    \pinktext{\frac{d\pi^{2n+1}_{0,T}}{d\pi^{2n}_{0,T}}}&=\frac{e^{\pinktext{\varphi_{n+1}(\boldsymbol{x})}+\hat{\varphi}_{n}(\boldsymbol{y})-c(\boldsymbol{x}, \boldsymbol{y})}d(\pi_0\otimes \pi_T)}{e^{\pinktext{\varphi_{n}(\boldsymbol{x})}+\hat{\varphi}_n(\boldsymbol{y})-c(\boldsymbol{x}, \boldsymbol{y})}d(\pi_0\otimes \pi_T)}=\frac{e^{\varphi_{n+1}(\boldsymbol{x})}}{e^{\varphi_{n}(\boldsymbol{x})}}=\pinktext{e^{\varphi_{n+1}(\boldsymbol{x})-\varphi_{n}(\boldsymbol{x})}}\label{eq:sinkhorn-density5}
\end{align}
\end{small}
Plugging the result (\ref{eq:sinkhorn-density3}) into (\ref{eq:sinkhorn-density4}), we get:
\begin{small}
\begin{align}
    \text{KL}\left(\pi_{0,T}^{2n}\|\pi_{0,T}^{2n-1}\right)&=\int_{\mathcal{X}\times\mathcal{Y}}\log \left(\bluetext{e^{\hat{\varphi}_n(\boldsymbol{y})-\hat{\varphi}_{n-1}(\boldsymbol{y})}}\right)d\pi^{2n}_{0,T}\nonumber\\
    &=\int_{\mathcal{X}\times\mathcal{Y}} \left(\bluetext{\hat{\varphi}_n(\boldsymbol{y})-\hat{\varphi}_{n-1}(\boldsymbol{y})}\right)d\pi^{2n}_{0,T}=\int_{\mathcal{Y}}\left(\bluetext{\hat{\varphi}_n(\boldsymbol{y})-\hat{\varphi}_{n-1}(\boldsymbol{y})}\right)\pi_T(d\boldsymbol{y}) \label{eq:sinkhorn-density6}
\end{align}
\end{small}
Similarly, plugging in (\ref{eq:sinkhorn-density5}) into (\ref{eq:sinkhorn-density4}), we have:
\begin{small}
\begin{align}
    \text{KL}\left(\pi_{0,T}^{2n+1}\|\pi_{0,T}^{2n}\right)&=\int_{\mathcal{X}\times\mathcal{Y}}\log \left(\pinktext{e^{\varphi_{n+1}(\boldsymbol{x})-\varphi_{n}(\boldsymbol{x})}}\right)d\pi^{2n+1}_{0,T}\nonumber\\
    &=\int_{\mathcal{X}\times\mathcal{Y}} \left(\pinktext{\varphi_{n+1}(\boldsymbol{x})-\varphi_{n}(\boldsymbol{x})}\right)d\pi^{2n+1}_{0,T}=\int_{\mathcal{X}}\left(\pinktext{\varphi_{n+1}(\boldsymbol{x})-\varphi_{n}(\boldsymbol{x})}\right)\pi_0(d\boldsymbol{x}) \label{eq:sinkhorn-density7}
\end{align}
\end{small}
which recovers the KL steps defined in (i). Since the KL iterates are finite, we can conclude that the potentials are integrable over the marginal densities such that $\varphi_n\in L^1(\pi_0)$ and $\hat\varphi_n\in L^1(\pi_T)$. In the next step, we show that the Sinkhorn iterates admit an additive structure, where the total dual potentials are equivalent to the sum of KL divergences between iterations.

\textbf{Step 2: Proof of Property (ii). }
We can sum up the KL divergence expressions derived in (\ref{eq:sinkhorn-density6}) and (\ref{eq:sinkhorn-density7}) to recover the total accumulated KL at each iteration:
\begin{small}
\begin{align}
    \sum_{k=0}^{n}\text{KL}\left(\pi_{0,T}^{(2k)}\|\pi_{0,T}^{(2k-1)}\right)&=\sum_{k=0}^{n}\int_{\mathcal{Y}}\left(\hat{\varphi}_{k}-\hat{\varphi}_{k-1}\right)\pi_T=\int_{\mathcal{Y}}\bluetext{\underbrace{\sum_{k=0}^{n}\left(\hat{\varphi}_{k}-\hat{\varphi}_{k-1}\right)}_{=(\hat{\varphi}_{n}-\hat{\varphi}_{-1})}}\pi_T\overset{(\bigstar)}{=}\int_{\mathcal{Y}}\hat    \varphi_n\pi_T=:\pi_T(\hat{\varphi}_n)\\
    \sum_{k=0}^{n-1}\text{KL}\left(\pi_{0,T}^{(2k+1)}\|\pi_{0,T}^{(2k)}\right)&=\sum_{k=0}^{n-1}\int_{\mathcal{X}}\left(\varphi_{k+1}-\varphi_{k}\right)\pi_0=\int_{\mathcal{X}}\pinktext{\underbrace{\sum_{k=0}^{n-1}\left(\varphi_{k+1}-\varphi_{k}\right)}_{=(\varphi_{n}-\varphi_0)}}\pi_0\overset{(\bigstar)}{=}\int_{\mathcal{X}}\varphi_n\pi_0=:\pi_0(\varphi_n)
\end{align}
\end{small}
where the equalities $(\bigstar)$ follow from applying the \textbf{telescoping trick} to cancel intermediate terms and substituting the initialization of $\hat{\varphi}_{-1}=0$ and $\varphi_0=0$. This recovers a clean expression for the total accumulated KL after $n$ iterations of Sinkhorn's algorithm. \hfill $\square$

To show convergence of Sinkhorn iterations, we aim to show that the \textbf{KL divergence with the optimal Schrödinger bridge coupling converges to zero}, i.e., $\text{KL}(\pi^\star_{0,T}\|\pi_{0,T}^{(n)})\to 0$, and achieves the correct marginals $\pi_0$ and $\pi_T$ as $n\to \infty$ iterations, which implies that the potentials converge to the \textbf{Schrödinger potentials} $(\varphi, \hat{\varphi})\to(\varphi^\star, \hat{\varphi}^\star)$. 

\begin{proposition}[Marginal Convergence of Sinkhorn Iterations]\label{prop:sinkhorn-marginal-convergence}
    Each iteration of Sinkhorn's algorithm results in a decrease in KL divergence, such that for all $n\geq -1$, the KL with the optimal coupling $\pi^\star_{0,T}$ is decreasing in $n$:
    \begin{align}
        \text{KL}\left(\pi^\star_{0,T}\|\pi^{(n)}_{0,T}\right)&=\text{KL}\left(\pi^\star_{0,T}\|q\right)-\sum_{k=0}^n\text{KL}\left(\pi^{(k)}_{0,T}\|\pi^{(k-1)}_{0,T}\right)\label{eq:sinkhorn-kl-decrease}
    \end{align}
\end{proposition}

\textit{Proof.} Using the properties of Sinkhorn iterations defined in Lemma \ref{lemma:sinkhorn-properties}, we can expand the expression for the KL divergence with the optimal coupling $\pi^\star_{0,T}$ as:
\begin{small}
\begin{align}
    \text{KL}(\pi^\star_{0,T}\|\pi^{2n}_{0,T})&=\mathbb{E}_{\pi^\star_{0,T}}\left[\log \frac{d\pi^\star_{0,T}}{d\pi^{2n}_{0,T}}\right]=\mathbb{E}_{\pi^\star_{0,T}}\left[\log \frac{d\pi^\star_{0,T}}{dq}\frac{dq}{d\pi^{2n}_{0,T}}\right]=\underbrace{\mathbb{E}_{\pi^\star_{0,T}}\left[\log \frac{d\pi^\star_{0,T}}{dq}\right]}_{\text{KL}(\pi^\star_{0,T}\|q)}-\mathbb{E}_{\pi^\star_{0,T}}\left[\log \frac{d\pi^{2n}_{0,T}}{dq}\right]\nonumber\\
    &=\text{KL}(\pi^\star_{0,T}\|q)-\mathbb{E}_{\pi^\star_{0,T}}\left[\bluetext{\sum_{k=0}^n\log \frac{d\pi^{(2k)}_{0,T}}{d\pi^{(2k-1)}_{0,T}}}+\pinktext{\sum_{k=0}^{n-1}\log \frac{d\pi^{(2k+1)}_{0,T}}{d\pi^{(2k)}_{0,T}}}\right]\nonumber\\
    &=\text{KL}(\pi^\star_{0,T}\|q)-\mathbb{E}_{\pi^\star_{0,T}}\left[\bluetext{\sum_{k=0}^n \left(\hat{\varphi}_k-\hat{\varphi}_{k-1}\right)}+\pinktext{\sum_{k=0}^{n-1}\left(\varphi_{k+1}-\varphi_k\right) }\right]\nonumber\\
    &=\text{KL}(\pi^\star_{0,T}\|q)-\mathbb{E}_{\pi^\star_{0,T}}\left[\bluetext{ \left(\hat{\varphi}_n-\hat{\varphi}_{-1}\right)}+\pinktext{\left(\varphi_{n}-\varphi_0\right) }\right]\nonumber\\
    &=\text{KL}(\pi^\star_{0,T}\|q)-\mathbb{E}_{\pi^\star_{0,T}}\left[\bluetext{\hat{\varphi}_n}+\pinktext{\varphi_{n} }\right]\nonumber\\
    &=\text{KL}(\pi^\star_{0,T}\|q)-\left(\pi_T(\hat\varphi_n)+\pi_0(\varphi_n)\right)
\end{align}
\end{small}
Substituting the result from Lemma \ref{lemma:sinkhorn-properties} (ii), we have:
\begin{small}
\begin{align}
    \text{KL}(\pi^\star_{0,T}\|\pi^{2n}_{0,T})&=\text{KL}(\pi^\star_{0,T}\|q)-\left(\pi_T(\hat\varphi_n)+\pi_0(\varphi_n)\right)\nonumber\\
    &=\text{KL}(\pi^\star_{0,T}\|q)-\left(\sum_{k=0}^{n}\text{KL}\left(\pi_{0,T}^{(2k)}\|\pi_{0,T}^{(2k-1)}\right)+\sum_{k=0}^{n-1}\text{KL}\left(\pi_{0,T}^{(2k+1)}\|\pi_{0,T}^{(2k)}\right)\right)\nonumber\\
    &=\text{KL}(\pi^\star_{0,T}\|q)-\sum_{k=0}^{2n}\text{KL}\left(\pi_{0,T}^{(k)}\|\pi_{0,T}^{(k-1)}\right)\nonumber\\
    \implies &\boxed{\text{KL}(\pi^\star_{0,T}\|\pi^{(n)}_{0,T})=\text{KL}(\pi^\star_{0,T}\|q)-\sum_{k=0}^{n}\text{KL}\left(\pi_{0,T}^{(k)}\|\pi_{0,T}^{(k-1)}\right)}
\end{align}
\end{small}
which proves that the KL divergence with the optimal coupling is decreasing as the number of iterations $n$ increases.\hfill $\square$

We can now establish the convergence of Sinkhorn's algorithm, which states that as the number of iterations increases to infinity $n\to \infty$, the KL divergence with the optimal coupling converges to zero.

\begin{corollary}[Convergence of Sinkhorn's Algorithm]
    The KL divergence between the marginals satisfies the inequality:
    \begin{align}
        \text{KL}(\pi^{(k)}_{0}\|\pi_{0})+\text{KL}(\pi^{(k)}_{T}\|\pi_{T})\leq \text{KL}(\pi^{(k)}_{0,T}\|\pi_{0,T}^{(k-1)})
    \end{align}
    and the sum of the right-hand side for $n$ iterations is bounded by:
    \begin{small}
    \begin{align}
        &\sum_{k=1}^n \text{KL}(\pi^{(k)}_{0,T}\|\pi_{0,T}^{(k-1)})\leq  \text{KL}(\pi^\star_{0,T}\|q)\\
        \implies  \text{KL}(\pi^{(k)}_{0,T}\|\pi_{0,T}^{(k-1)})\to 0, &\quad \text{KL}(\pi^{(k)}_{0}\|\pi_{0})\to 0, \quad \text{KL}(\pi^{(k)}_{T}\|\pi_{T})\to 0, \quad \text{for} \quad n\to \infty
    \end{align}
    \end{small}
    which also implies that $\pi_0^{(k)}\to \pi_0$ and $\pi_T^{(k)}\to \pi_T$ as $n\to \infty$.
\end{corollary}

\textit{Proof.} First, we recall that the marginals are correct for $\pi_0$ on the odd iterations $k\geq 1$, such that $\pi^{(k)}_0=\pi_0$, and correct for $\pi_T$ for the even iterations $k\geq 2$, such that $\pi^{(k)}_{T}=\pi_T$. Therefore, we can rewrite the sum of the marginal KL divergences as:
\begin{align}
    \text{KL}(\pi^{(2k)}_{0}\|\pi_{0})+\underbrace{\text{KL}(\pi^{(2k)}_{T}\|\pi_{T})}_{=0}&=\bluetext{\text{KL}(\pi^{(2k)}_{0}\|\pi_{0})\leq \text{KL}(\pi^{(2k)}_{0,T}\|\pi_{0,T}^{(2k-1)})}\\
    \underbrace{\text{KL}(\pi^{(2k-1)}_{0}\|\pi_{0})}_{=0}+\text{KL}(\pi^{(2k-1)}_{T}\|\pi_{T})&=\pinktext{\text{KL}(\pi^{(2k-1)}_{T}\|\pi_{T})\leq \text{KL}(\pi^{(2k-1)}_{0,T}\|\pi_{0,T}^{(2k-2)})}
\end{align}
where the inequalities can be obtained by applying the \textbf{data processing inequality} (Lemma \ref{lemma:data-processing-inequality}). For $n$ iterations, we can rearrange (\ref{eq:sinkhorn-kl-decrease}) to get an upper bound for the sum of KL divergences for $k\geq 1$ as:
\begin{small}
\begin{align}
    \sum_{k=0}^n\text{KL}\left(\pi^{(k)}_{0,T}\|\pi^{(k-1)}_{0,T}\right)&=\text{KL}\left(\pi^\star_{0,T}\|q\right)-\underbrace{\text{KL}\left(\pi^\star_{0,T}\|\pi^{(n)}_{0,T}\right)}_{\geq 0}\leq \text{KL}\left(\pi^\star_{0,T}\|q\right)\\
    \implies \sum_{\bluetext{k=1}}^n\text{KL}\left(\pi^{(k)}_{0,T}\|\pi^{(k-1)}_{0,T}\right)&=\sum_{k=0}^n\text{KL}\left(\pi^{(k)}_{0,T}\|\pi^{(k-1)}_{0,T}\right)-\text{KL}\left(\pi^{(0)}_{0,T}\|\bluetext{\pi^{(-1)}_{0,T}}\right)\nonumber\\
    &\leq \text{KL}\left(\pi^\star_{0,T}\|q\right)-\text{KL}\left(\pi^{(0)}_{0,T}\|\bluetext{q}\right)
\end{align}
\end{small}
where $\pi^{(-1)}_{0,T}=dq$ as in (\ref{eq:sinkhorn-density2}). Therefore, for $n \to \infty$, we have: 
\begin{small}
\begin{align}
    \sum_{k=1}^\infty\bluetext{\text{KL}(\pi^{(2k)}_{0}\|\pi_{0})}\leq \sum_{k=1}^\infty\text{KL}\left(\pi^{(k)}_{0,T}\|\pi^{(k-1)}_{0,T}\right)\leq \text{KL}\left(\pi^\star_{0,T}\|q\right)-\text{KL}\left(\pi^{(0)}_{0,T}\|q\right)&\implies\bluetext{\text{KL}(\pi^{(n)}_{0}\|\pi_{0})\to 0} \\
    \sum_{k=1}^\infty\pinktext{\text{KL}(\pi^{(2k-1)}_{T}\|\pi_{T})}\leq \sum_{k=1}^\infty\text{KL}\left(\pi^{(k)}_{0,T}\|\pi^{(k-1)}_{0,T}\right)\leq \text{KL}\left(\pi^\star_{0,T}\|q\right)-\text{KL}\left(\pi^{(0)}_{0,T}\|q\right)&\implies \pinktext{\text{KL}(\pi^{(n)}_T\|\pi_T)\to 0}
\end{align}
\end{small}
Applying Pinkser's inequality, which states that the total variation between distributions $p, q$ satisfy $\|p-q\|_{\text{TV}}\leq \sqrt{2\text{KL}(p\|q)}$, we conclude that $\pi^{(n)}_0\to \pi_0$ and $\pi^{(n)}_T\to \pi_T$ in variation. \hfill $\square$

While Proposition \ref{prop:sinkhorn-marginal-convergence} proves that the \textbf{marginals} generated by the Sinkhorn iterations converge to the marginal constraints, it \textbf{does not directly imply strong convergence} to the optimal coupling $\pi_{0,T}^\star$ or the Schrödinger potentials $(\varphi, \hat{\varphi})\to (\varphi^\star, \hat{\varphi}^\star)$. However, strong convergence is only guaranteed for certain conditions on $c$, which we establish with the following Theorem.

\begin{theorem}[Strong Convergence of Sinkhorn's Algorithm]\label{theorem:sinkhorn-strong}
    Given a cost function $c:\mathcal{X}\times \mathcal{Y}\to \mathbb{R}$ that is bounded from below and satisfies exponential integrability, such that:
    \begin{align}
        \exists r > 1, \quad \int e^{r c(\boldsymbol{x},\boldsymbol{y})}d(\pi_0\otimes \pi_T)< \infty\tag{Integrability Condition}\label{eq:cost-integrability-cond}
    \end{align}
    Then, the Sinkhorn iterates converge to the true Schrödinger potentials $\varphi_n \to \varphi^\star$ and $\hat\varphi_n \to \hat\varphi^\star$, and the induced couplings converge to the optimal Schrödinger bridge, $\text{KL}(\pi^\star_{0,T}\|\pi^{(n)}_{0,T}) \to 0$ and $\pi^{(n)}_{0,T}\to \pi^\star_{0,T}$ in variation.
\end{theorem}

\textit{Proof Sketch.} Under the (\ref{eq:cost-integrability-cond}) on the cost, the Sinkhorn iterates remain uniformly integrable and cannot develop singular behavior. Since the iterates are uniformly integrable, functional analysis establishes that there exists a subsequence where $e^{\varphi_n}$ converges weakly in $L^1$, which passes the limit and forces convergence of $\hat\varphi_n \to \hat\varphi^\star$ under the coupled update condition:
\begin{small}
\begin{align}
    &\underbrace{\int_{\mathcal{X}} e^{\varphi_n(\boldsymbol{x})}e^{-c(\boldsymbol{x},\boldsymbol{y})}d\pi_0(\boldsymbol{x})\to \int_{\mathcal{X}} e^{\varphi^\star(\boldsymbol{x})}e^{-c(\boldsymbol{x},\boldsymbol{y})}d\pi_0(\boldsymbol{x})}_{\text{converges weakly in }L^1(\pi_0)}\nonumber\\
    \implies &-\log \lim_{n\to \infty}\int_{\mathcal{X}} e^{\varphi_n(\boldsymbol{x})}e^{-c(\boldsymbol{x},\boldsymbol{y})}d\pi_0(\boldsymbol{x})= \boxed{-\log \int_{\mathcal{X}} e^{\varphi^\star(\boldsymbol{x})}e^{-c(\boldsymbol{x},\boldsymbol{y})}d\pi_0(\boldsymbol{x})=:\hat\varphi^\star(\boldsymbol{y})}
\end{align}
\end{small}

We have shown in Theorem \ref{theorem:dual-static-sb} that $(\varphi\oplus \hat\varphi)$ is unique, which implies that all subsequences of iterations converge to the same limit. Therefore, the full sequence of Sinkhorn iterates converges, and the induced couplings converge in KL and total variation to the Schrödinger bridge\footnote{For rigorous proof, see Theorem 6.15 in \citet{nutz2021introduction}.}.\hfill $\square$

We have now established the \textbf{first method to tractably solve the Schrödinger bridge problem under the static formulation}, which is equivalent to solving the entropic optimal transport problem. From our analysis, we have shown that the intuitive idea of \textit{alternating} between solving (\ref{eq:first-potential}) and (\ref{eq:second-potential}) leads to a theoretically grounded algorithm which establishes Sinkhorn iterations as alternating KL projections onto the marginal constraint sets, or equivalently as coordinate ascent on the dual Schrödinger potentials. These ideas lay the foundation of computational optimal transport and modern generative modeling techniques.

\subsection{Closing Remarks for Section \ref{sec:static-sb}}
In this section, we traced the origins of the Schrödinger bridge problem to the classical \textbf{optimal mass transport} (OMT) problem introduced by Monge and Kantorovich, which seeks an optimal \textit{coupling} between probability distributions that minimizes a prescribed transport cost. Motivated by the non-uniqueness and instability that can arise in the solutions of the OMT problem, we introduced the formulation of \textbf{entropic optimal transport} (EOT), in which an entropy-regularization term that measures deviation from a reference coupling is added to the transport objective. The resulting entropy-regularized problem admits a \textit{unique} and computationally stable solution. By reparameterizing the reference coupling, we showed that the EOT problem is equivalent to the \textbf{static Schrödinger bridge} (SB) problem, where the objective is to find a coupling that remains closest in relative entropy to a reference endpoint law while satisfying the prescribed marginal distributions. Finally, we introduced the classical Sinkhorn algorithm, an efficient iterative scheme for solving the static SB problem in practice.

While the static formulation provides the foundation for understanding how probability mass can be optimally transported between distributions, its connection to modern generative modeling emerges through the \textbf{dynamic formulation} of the Schrödinger bridge problem. The dynamic viewpoint lifts the static coupling problem to the space of \textit{stochastic path measures}, describing how probability flows through time under controlled stochastic dynamics. In the next section, we develop this perspective by introducing the theory of stochastic processes and path measures, defining KL divergences over path space, and deriving the optimality conditions of the dynamic Schrödinger bridge in the form of both nonlinear and linear partial differential equations (PDEs).

\newpage
\section{The Dynamic Schrödinger Bridge Problem}
\label{sec:dynamic-sb}
Having established the connection between optimal transport and entropy-regularized couplings, we now turn to the \textit{dynamic} formulation of the Schrödinger bridge (SB) problem. While optimal transport focuses on static couplings between distributions, the dynamic SB lifts the problem to the level of \textbf{stochastic processes}. Instead of directly transporting mass, we seek the most likely evolution of a stochastic system that transforms an initial distribution into a target distribution over time. In this section, we formalize this dynamic viewpoint and develop the stochastic calculus tools, including path measures, Itô processes, and change-of-measure techniques, necessary to analyze and solve the dynamic SB problem.

\subsection{Dynamic Optimal Transport Problem}
\label{subsec:dynamic-ot-problem}
Just like how the static Schrödinger bridge problem could be traced back to the Monge-Kantorovich optimal mass transport problems, we begin our discussion of the dynamic Schrödinger bridge problem with \boldtext{Bernamou-Brenier (dynamic) optimal transport problem} \citep{benamou2000computational}. The \textbf{key idea} that differentiates the \textit{dynamic} from the \textit{static} problems is the intuition behind transporting mass with a \textit{continuous flow over a time interval} rather than static couplings between marginals.

Concretely, given an initial distribution $\pi_0\in  \mathcal{P}(\mathbb{R}^d)$ at time $t=0$ and a target distribution $\pi_T\in  \mathcal{P}(\mathbb{R}^d)$ at time $t=T$, the dynamic formulation aims to determine the continuous-time evolution of a marginal probability density $p_t\in \mathcal{P}(\mathbb{R}^d)$ that evolves mass from $\pi_0\to \pi_T$ over $t\in [0,T]$. 

\begin{figure}
    \centering
    \includegraphics[width=\linewidth]{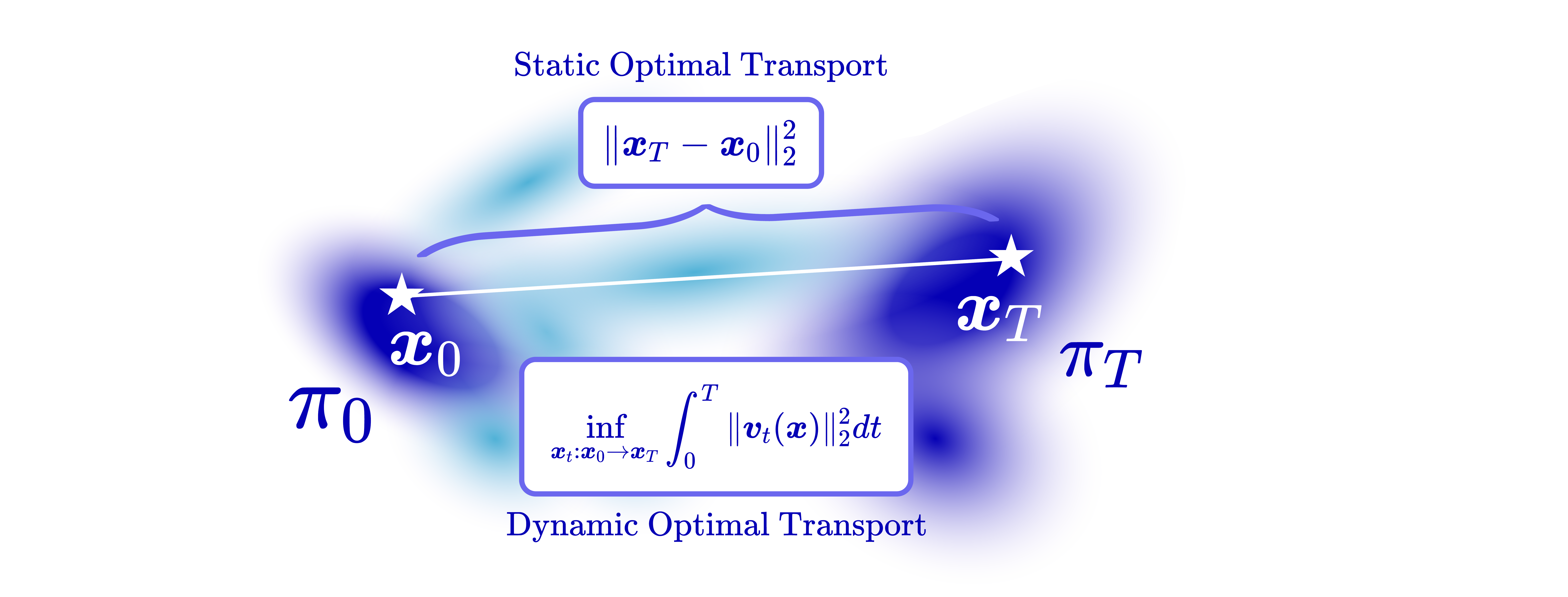}
    \caption{\textbf{Comparison Between Static and Dynamic Optimal Transport.} Illustration of the relationship between the static and dynamic formulations of optimal transport between two marginal distributions $\pi_0$ and $\pi_T$. The static formulation minimizes the transport cost directly between endpoints via the quadratic cost $\|\boldsymbol{x}_T-\boldsymbol{x}_0\|^2 $, while the dynamic (Benamou–Brenier) formulation instead seeks a time-dependent velocity field $\boldsymbol{v}_t(\boldsymbol{x})$ that continuously transports mass from $\pi_0$ to $\pi_T$ while minimizing the kinetic energy $\int_0^T\|\boldsymbol{v}_t(\boldsymbol{x})\|^2 dt$. The white trajectory represents a particle path under the optimal flow, illustrating how dynamic OT realizes the same optimal coupling as static OT through continuous mass evolution.}
    \label{fig:dynamic-ot}
\end{figure}

Recall that (\ref{eq:kant-omt-problem}) aims to minimize the cost of transporting states $\boldsymbol{x}_0\sim \pi_0$ to $\boldsymbol{x}_T\sim \pi_T$ defined as $c(\boldsymbol{x}_0,\boldsymbol{x}_T)$. A natural cost function is the Euclidean distance $c(\boldsymbol{x}_0,\boldsymbol{x}_T):=\|\boldsymbol{x}_T-\boldsymbol{x}_0\|^2 $, which is the \textbf{straight-line distance} between $\boldsymbol{x}_0$ and $\boldsymbol{x}_T$. To reformulate this into a \textit{dynamic} transport problem, we can consider transporting the state continuous state $\boldsymbol{x}$ over the straight line which travels a total distance of $\|\boldsymbol{x}_T-\boldsymbol{x}_0\|^2 $. Concretely, the quadratic transport cost from $\boldsymbol{x}_0\to \boldsymbol{x}_T$ is equivalent to the integrating energy-minimizing velocity field $\boldsymbol{v}_t(\boldsymbol{x}):=\frac{d}{dt}\boldsymbol{x}_t$ over $t\in [0,T]$ defined as:
\begin{align}
    \|\boldsymbol{x}_T-\boldsymbol{x}_0\|^2 =\inf_{\boldsymbol{x}_t:\boldsymbol{x}_0\to \boldsymbol{x}_T}\int_0^T\|\boldsymbol{v}_t(\boldsymbol{x})\|^2 dt
\end{align}
Leveraging this identity, we can reframe the static OT problem as minimizing the \textbf{total kinetic energy} of transporting particles from the distribution $\pi_0$, where particles at location $\boldsymbol{x}$ move with velocity $\boldsymbol{v}_t(\boldsymbol{x})$, weighted by the probability mass of the particle $p_t(\boldsymbol{x})$ (Figure \ref{fig:dynamic-ot}). Jointly minimizing the velocity and marginal probability path $(\boldsymbol{v}_t, p_t)$ yields the \boldtext{dynamic OT problem}.

\begin{definition}[Dynamic Optimal Transport (OT) Problem]\label{def:dynamic-ot-problem}
    Given two marginal constraints $\pi_0, \pi_T\in \mathcal{P}(\mathbb{R}^d)$, the \textbf{dynamic optimal transport (OT) problem} aims to find the optimal probability flow $p^\star_t:\mathbb{R}^d\times[0,T]\to \mathbb{R}\in \mathcal{P}(\mathbb{R}^d)$ and velocity field $\boldsymbol{v}^\star_t:\mathbb{R}^d\times[0,T]\to \mathbb{R}^d$ for $t\in [0,T]$ that minimizes:
    \begin{small}
    \begin{align}
        \inf_{(p_t, \boldsymbol{v}_t)}\left\{\int_0^T\int_{\mathbb{R}^d}\|\boldsymbol{v}_t(\boldsymbol{x})\| ^2p_t(\boldsymbol{x})d\boldsymbol{x}dt\right\} \quad\text{s.t.}\quad\begin{cases}
            \partial_tp_t+\nabla\cdot(p_t\boldsymbol{v}_t)=0\\
            p_0=\pi_0, \quad p_T=\pi_T
        \end{cases}\tag{Dynamic OT Problem}\label{eq:dynamic-ot-prob}
    \end{align}
    \end{small}
    whre the continuity equation $\partial_tp_t+\nabla\cdot(p_t\boldsymbol{v}_t)=0$ enforces conservation of probability mass over the continuous flow.
\end{definition}
The solution to (\ref{eq:dynamic-ot-prob}) is a time-interpolation of the \boldtext{optimal transport map} $M^\star$ defined as:
\begin{align}
    M^\star_t(\boldsymbol{x}_0) =(1-t) \boldsymbol{x}_0+tM^\star(\boldsymbol{x}_0), \quad  \boldsymbol{x}_0\sim \pi_0, \quad t\in [0,T]\tag{Optimal Transport Map}
\end{align}
which yields the optimal marginal density $p^\star_t$ as the pushforward of $\pi_0$ via the transport map at time $t\in [0,T]$ denoted $p^\star_t=(M_t^\star)_{\#}\pi_0$.

This Bernamou-Brenier (dynamic) formulation characterizes optimal transport minimization of kinetic energy over \textit{straight}, \textit{deterministic} flows satisfying the continuity equation. However, most real-world systems do not naturally evolve in straight lines, but rather stochastic paths that traverse non-linear manifolds. This naturally leads us to the \textbf{dynamic Schrödinger bridge problem}, which considers the optimal transport problem where the underlying dynamics are \textit{stochastic} rather than deterministic. 

\subsection{Dynamic Schrödinger Bridge Problem}
\label{subsec:sb-problem}

While the dynamic optimal transport (OT) formulation characterizes the most efficient deterministic flow that transports probability mass between two distributions, many real-world systems evolve under intrinsic stochasticity. This motivates a \textbf{stochastic generalization of dynamic OT}, leading to the \boldtext{dynamic Schrödinger bridge (SB) problem}. Instead of searching over deterministic velocity fields, the SB formulation asks: \textit{Among all stochastic evolutions that transform an initial distribution into a target distribution over a time horizon $[0,T]$, which one is \textbf{most likely} relative to a given reference dynamics?}

To answer this, the dynamic SB problem introduces a \boldtext{reference path measure} $\mathbb{Q}$ which describes the baseline stochastic dynamics of a system. The dynamic Schrödinger bridge problem then selects, among all processes \textbf{matching prescribed initial and terminal marginals}, the one that deviates minimally from the reference process in relative entropy. 

\begin{definition}[Dynamic Schrödinger Bridge Problem]\label{def:dynamic-sb-problem}
Let $\pi_0, \pi_T\in \mathcal{P}(\mathbb{R}^d)$ be probability measures on state space $\mathbb{R}^d$ and let $\mathbb{Q}\in \mathcal{P}(C([0,T]; \mathbb{R}^d))$ be a \textbf{reference path measure}, where $\mathcal{P}(C([0,T]; \mathbb{R}^d))$ is the space of probability paths over $\mathbb{R}^d$. 

The \textbf{dynamic Schrödinger bridge (SB) problem} seeks a new path measure $\mathbb{P}\in \mathcal{P}(C([0,T]; \mathbb{R}^d))$ with initial and terminal marginals matching $p_0=\pi_0$ and $p_T=\pi_T$ which minimizes the relative entropy with respect to $\mathbb{Q}$:
\begin{align}
    \mathbb{P}^\star=\underset{\mathbb{P}\in \mathcal{P}(C([0,T]; \mathbb{R}^d))}{\arg\min}\big\{\text{KL}(\mathbb{P}\|\mathbb{Q}):p_0=\pi_0, p_T=\pi_T\big\}\tag{Dynamic SB Problem}\label{eq:dynamic-sb-problem}
\end{align}
where $\text{KL}(\cdot \|\cdot )$ denotes the KL divergence on path space given by:
\begin{align}
    \text{KL}(\mathbb{P}\|\mathbb{Q})=\mathbb{E}_{\mathbb{P}}\left[\log \frac{\mathrm{d}\mathbb{P}}{\mathrm{d}\mathbb{Q}}\right]
\end{align}
\end{definition}

This perspective reveals the Schrödinger bridge as an entropy-regularized analogue of the (\ref{eq:dynamic-ot-prob}). In the limit of vanishing noise, the problem recovers classical optimal transport, while for positive noise, it admits a rich stochastic structure. In particular, when the reference dynamics is Brownian motion, the optimal path measure inherits a Markov structure and can be characterized through a time-dependent drift correction of the reference process. In this section, we will break down these ideas, starting with path measures and Itô processes, then deriving the path-space KL divergence, and finally exploring the dynamic SB problem through the lens of the Hamilton-Jacobi-Bellman and Fokker-Planck system or equations and the Hopf-Cole transform.

\subsection{Path Measures and Itô Processes}
\label{subsec:path-measure-ito-processes}
Before diving deeper into Schrödinger Bridge theory, let's first establish the foundations of \boldtext{path measures and Itô processes} that will help us solve the SB problem. First, we define a \boldtext{stochastic process} as a random variable $\boldsymbol{X}_t\in \mathbb{R}^d$ that evolves over a time horizon $t\in [0, T]$, denoted $\boldsymbol{X}_{0:T}:=(\boldsymbol{X}_t)_{t\in [0,T]}$. The distribution of many such random variables follows a time-dependent marginal probability distribution, denoted $p_t\in \mathcal{P}(\mathbb{R}^d)$. 

\begin{figure}
    \centering
    \includegraphics[width=\linewidth]{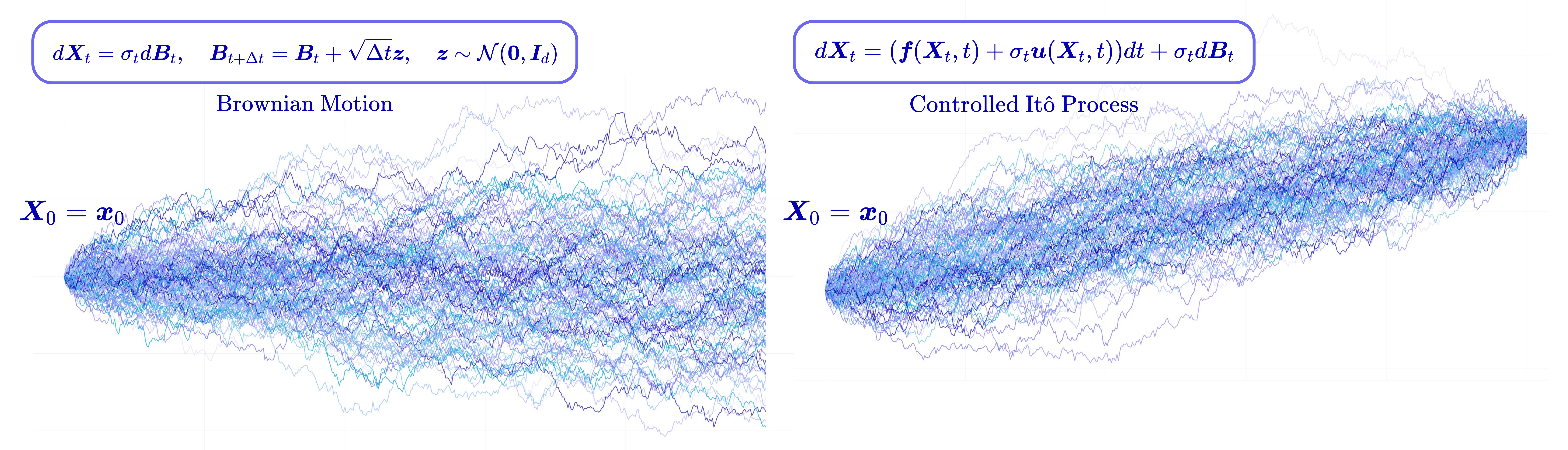}
    \caption{\textbf{Brownian Motion and Controlled Itô Processes.} Illustration of one-dimensional stochastic trajectories generated by two stochastic differential equations (SDEs) starting from $\boldsymbol{X}_0=0$ over the time interval $t\in [0,1]$. \textbf{Left:} Sample paths of pure Brownian motion of the form $d\boldsymbol{X}_t=\sigma_td\boldsymbol{B}_t$, where increments are Gaussian with variance proportional to the timestep $\boldsymbol{B}_{t+\Delta t}=\boldsymbol{B}_t+\sqrt{\Delta t}\boldsymbol{z}$ with $\boldsymbol{z}\sim \mathcal{N}(0,\boldsymbol{I}_d)$. \textbf{Right:} Sample paths of a controlled Itô process of the form $d\boldsymbol{X}_t=(\boldsymbol{f}(\boldsymbol{X}_t,t)+\sigma_t\boldsymbol{u}(\boldsymbol{X}_t,t))+\sigma_td\boldsymbol{B}_t$, where we set $\boldsymbol{f}\equiv 0$ and $\boldsymbol{u}(\boldsymbol{x},t):=\frac{2-\boldsymbol{x}}{T-t+\epsilon}$ which pulls the process to $\boldsymbol{X}_T=2$.}
    \label{fig:brownian}
\end{figure}

Specifically, we want to consider only stochastic processes $\boldsymbol{X}_{0:T}$ that \textit{depend} only on the \textbf{past} and \textbf{present} states of the system, which are formally said to be \textbf{adapted to the filtration} $(\mathcal{F}_t)_{t\in [0,T]}$. The filtration $(\mathcal{F}_t)_{t\in [0,T]}$ is simply a sequence of events $\mathcal{F}_t$ that is \textbf{ordered} such that all events in $\mathcal{F}_s$ for $s\leq t$ are in $\mathcal{F}_t$, i.e., $\mathcal{F}_s \subseteq \mathcal{F}_t$ for $s\leq t$. This definition allows us to define a \boldtext{$\mathcal{F}_t$-adapted process} which depends only on the past and present events.

\begin{definition}[$\mathcal{F}_t$-Adapted Process]\label{def:filtration-adapted-process}
A stocastic process $\boldsymbol{X}_{0:T}$ is said to be $\mathcal{F}_t$-adapted if the random variable $\boldsymbol{X}_t$ is $\mathcal{F}_t$-measurable for all $t\in [0,T]$, where:
\begin{align}
    \mathcal{F}_t:=\sigma(\boldsymbol{X}_\tau: 0\leq \tau\leq t)
\end{align}
which is the sigma-algebra\footnote{This is not to be confused with the $\sigma$ notation for the diffusion coefficient. We will only use $\sigma$ to denote sigma-algebra sparingly.} generated by the history of the process up to time $t$ or all the information available from the process up to time $t$. Equivalently, this means that for all measurable sets $S\in \mathcal{S}(\mathbb{R}^d)$ over the state space, the process $\boldsymbol{X}_{0:T}$ satisfies:
\begin{align}
    \{\boldsymbol{X}_t\in S\}\in \mathcal{F}_t
\end{align}
which is a formal way of stating that all values $\boldsymbol{X}_t\in S$ can be determined using only information up to time $t$.
\end{definition}

A specific stochastic process that we will see often is \boldtext{Brownian motion}, or \textbf{$\mathcal{F}_t$-adapted Wiener process}, which begins at the origin and evolves via independent Gaussian steps with variance that is proportional to the time increment.

\begin{definition}[Brownian Motion]\label{def:brownian-motion}
Brownian motion $(\boldsymbol{B}_t)_{t\in [0,T]}$ is a type of stochastic process that starts at $\boldsymbol{B}_0=\boldsymbol{0}$ and evolves via \textbf{independent Gaussian increments} defined as
\begin{align}
    \boldsymbol{B}_{t+\Delta t}=\boldsymbol{B}_t+\sqrt{\Delta t}\boldsymbol{z}, \quad \boldsymbol{z}\sim \mathcal{N}(\boldsymbol{0}, \boldsymbol{I}_d)
\end{align}
where $\boldsymbol{z}$ is sampled independently from a unit isotropic Gaussian with zero-mean across all time steps. This means that each increment is Gaussian $\boldsymbol{B}_{t+\Delta t}-\boldsymbol{B}_t\sim \mathcal{N}(\boldsymbol{0}, \Delta t\boldsymbol{I}_d)$ and independent of $\mathcal{F}_t$. 
\end{definition}

Now that we have defined the notion of stochastic processes and Brownian motion, we can define an \boldtext{Itô process} which is the class of stochastic processes that are the foundation of most generative modeling frameworks, from diffusion to flow matching to Schrödinger bridge matching.

\begin{definition}[Itô Process]
An Itô Process is a stochastic process $(\boldsymbol{X}_t)_{t\in[0,T]}$ whose state $\boldsymbol{X}_t$ can be written as
\begin{align}
    \boldsymbol{X}_t=\boldsymbol{X}_0+\int_0^t\boldsymbol{f}(\boldsymbol{X}_t, t)ds+\int_0^t\boldsymbol{\Sigma}_td\boldsymbol{B}_t\label{eq:ito-sde2}
\end{align}
which can be equivalently defined as the solution of an \textbf{stochastic differential equation} (SDE) of the form
\begin{align}
    d\boldsymbol{X}_t=\boldsymbol{f}(\boldsymbol{X}_t,t)dt+\boldsymbol{\Sigma}_td\boldsymbol{B}_t\label{eq:ito-sde}
\end{align}
where $\boldsymbol{f}(\boldsymbol{X}_t, t):\mathbb{R}^d\times [0,T] \to \mathbb{R}^d$ is known as the \textbf{drift} and $\boldsymbol{\Sigma}_t\in \mathbb{R}^{d\times d}$ is the \textbf{diffusion} coefficient matrix. In most applications, the diffusion coefficient is a constant or time-dependent scalar $\sigma_t:[0,T] \to \mathbb{R}_{\geq 0}$\footnote{While we introduce the theory of path measures with the general covariance matrix $\boldsymbol{\Sigma}_t$, we simplify to the scalar time-dependent diffusion coefficient $\sigma_t$ in the remainder of this guide as it is most commonly used throughout the literature.}, which simplifies the SDE into $d\boldsymbol{X}_t=\boldsymbol{f}(\boldsymbol{X}_t,t)dt+\sigma_td\boldsymbol{B}_t$.
\end{definition}

Next we consider stochastic dynamics where there exists some external influences that perturb the drift of the process via some control drift. These \boldtext{controlled Itô processes} can be modeled with a controlled stochastic differential equation (SDE), where the time-dependent control velocity $\boldsymbol{u}(\boldsymbol{x},t):\mathbb{R}^d\times [0,T]\to \mathbb{R}^d$ enters the drift in the directions spanned by the diffusion coefficient $\sigma_t$. 

\begin{definition}[Controlled Itô Process]\label{def:controlled-ito}
    A controlled Itô process is obtained by introducing a control $\boldsymbol{u}(\boldsymbol{x},t):\mathbb{R}^d\times [0,T]\to \mathbb{R}^d$\footnote{In the remainder of this guide, we omit explicitly writing the domain and codomain ($\mathbb{R}^d \times [0,T] \to \mathbb{R}^d$) when it is clear from context.} that modifies the drift in the directions spanned by the diffusion coefficient, which yields the \textbf{controlled stochastic differential equation} (SDE) of the form: 
    \begin{small}
    \begin{align}
        d\boldsymbol{X}^u_t=\left(\boldsymbol{f}(\boldsymbol{X}_t^u,t)+\boldsymbol{\Sigma}_t\boldsymbol{u}(\boldsymbol{X}_t^u,t)\right)dt+\boldsymbol{\Sigma}_td\boldsymbol{B}_t\tag{Controlled SDE}\label{eq:controlled-sde}
    \end{align}
    \end{small}
    where $\boldsymbol{f}(\boldsymbol{x},t):\mathbb{R}^d \times[0,T] \to \mathbb{R}^d$ is the reference drift and $\boldsymbol{\Sigma}_t:[0,T] \to \mathbb{R}^{d\times d}$ is the diffusion coefficient. Equivalently, the controlled process can be written as:
    \begin{small}
    \begin{align}
        \boldsymbol{X}^u_t=\boldsymbol{X}_0+\int_0^t\left(\boldsymbol{f}(\boldsymbol{X}_\tau^u,\tau)+\boldsymbol{\Sigma}_\tau\boldsymbol{u}(\boldsymbol{X}_\tau^u,\tau)\right)d\tau+\int_0^t\boldsymbol{\Sigma}_\tau d\boldsymbol{B}_\tau
    \end{align}
    \end{small}
\end{definition}

Observe that the control is scaled by the diffusion coefficient $\boldsymbol{\Sigma}_t\boldsymbol{u}$ rather than added independently. This parameterization of the controlled SDE guarantees that the controlled process remains \textbf{absolutely continuous} with respect to the reference path measure $\mathbb{Q}$ defined using the diffusion term $\sigma_td\boldsymbol{B}_t$. This property ensures that the relative entropy used in the dynamic SB objective takes a tractable form, which we will show explicitly in Section \ref{subsec:girsanov}. Intuitively, scaling with the diffusion coefficient ensures that the control drift \textbf{only steers the process in directions within the space supported by the stochastic noise}, whereas adding the control drift independently without scaling could push the process in directions where the noise has no support.

Having developed an understanding of uncontrolled and controlled Itô processes, we now derive a \textbf{key property} of Itô processes, which states that applying some function to the state produces another Itô process over the transformed coordinates. This property is captured via \boldtext{Itô's formula}.

\begin{theorem}[Itô's Formula in $\mathbb{R}^d$]\label{thm:ito-formula}
Consider an Itô process $(\boldsymbol{X}_t\in \mathbb{R}^d)_{t\in [0,T]}$ and a scalar function $\phi(\boldsymbol{x},t):\mathbb{R}^d\times[0,T]\to \mathbb{R}$ that twice continuously differentiable $\phi\in C^{2,1}(\mathbb{R}^d\times [0,T])$\footnote{functions in the Hessian are continuous and produces a square integrable random variable $\phi(\boldsymbol{X}_t,t)\in L_2$ (i.e., $\mathbb{E}[|\phi(\boldsymbol{X}_t,t)|^2]< \infty$}. Then, the transformed random variable $\phi(\boldsymbol{X}_t,t)$ is also an Itô process that evolves via the SDE given by
\begin{align}
    d\phi(\boldsymbol{X}_t,t)=\partial_t\phi(\boldsymbol{X}_t,t)+\nabla \phi(\boldsymbol{X}_t,t)d\boldsymbol{X}_t+\frac{1}{2}d\boldsymbol{X}_t^{\top}(\nabla ^2\phi(\boldsymbol{X}_t,t))d\boldsymbol{X}_t\label{eq:ito-1}
\end{align}
Substituting $d\boldsymbol{X}_t=\boldsymbol{f}(\boldsymbol{X}_t, t)dt+\boldsymbol{\Sigma}_td\boldsymbol{B}_t$, we rewrite the formula as:
\begin{small}
\begin{align}
    d\boldsymbol{Y}_t&=\bigg[\partial_t\phi(\boldsymbol{X}_t,t)+\boldsymbol{f}(\boldsymbol{X}_t,t)^\top\nabla \phi(\boldsymbol{X}_t,t)+\frac{1}{2}\text{Tr}\left(\boldsymbol{\Sigma}_t\boldsymbol{\Sigma}_t^\top\nabla ^2\phi(\boldsymbol{X}_t,t)\right)\bigg]dt+\nabla \phi(\boldsymbol{X}_t,t)^\top \boldsymbol{\Sigma}d\boldsymbol{B}_t\tag{Itô's Formula}\label{eq:ito-2}
\end{align}
\end{small}
which means that the space of Itô processes is \textbf{closed under twice-differentiable functions}. If $\phi(\boldsymbol{x}):\mathbb{R}^d \to \mathbb{R}\in C^2(\mathbb{R}^d)$ is a scalar function dependent only on $\boldsymbol{x}$, then the partial derivative $\partial_t\phi$ disappears, and $\phi(\boldsymbol{X}_t)$ follows the SDE:
\begin{small}
\begin{align}
    d\phi(\boldsymbol{X}_t)=\left[\boldsymbol{f}(\boldsymbol{X}_t,t)^\top \nabla \phi(\boldsymbol{X}_t,t)+\frac{\sigma_t^2}{2}\Delta \phi(\boldsymbol{X}_t)\right]dt+ \nabla \phi(\boldsymbol{X}_t)^\top \sigma_td\boldsymbol{B}_t
\end{align}
\end{small}
\end{theorem}

\textit{Proof.} In this proof, we will break down the steps from (\ref{eq:ito-1}) to (\ref{eq:ito-2}). After substituting $d\boldsymbol{X}_t=\boldsymbol{f}_tdt+\boldsymbol{\Sigma}_td\boldsymbol{B}_t$ where we denote $\boldsymbol{f}_t\equiv \boldsymbol{f}(\boldsymbol{X}_t,t)$ for simplicity, we get:
\begin{small}
\begin{align}
    d\phi(\boldsymbol{X}_t,t)=\partial_t\phi(\boldsymbol{X}_t,t)+\nabla \phi(\boldsymbol{X}_t,t)^\top(\boldsymbol{f}_tdt+\boldsymbol{\Sigma}_td\boldsymbol{B}_t)+\frac{1}{2}(\boldsymbol{f}_tdt+\boldsymbol{\Sigma}_td\boldsymbol{B}_t)^{\top}(\nabla ^2\phi(\boldsymbol{X}_t,t))(\boldsymbol{f}_tdt+\boldsymbol{\Sigma}_td\boldsymbol{B}_t)
\end{align}
\end{small}
Itô's calculus defines $dtdt=\boldsymbol{B}_t=0$ and $(dt)^2=0$, so we can simplify the second term to: 
\begin{small}
\begin{align}
    d\phi(\boldsymbol{X}_t,t)=\partial_t\phi(\boldsymbol{X}_t,t)+\nabla \phi(\boldsymbol{X}_t,t)^\top(\boldsymbol{f}_tdt+\boldsymbol{\Sigma}_td\boldsymbol{B}_t)+\frac{1}{2}\underbrace{(\boldsymbol{\Sigma}_td\boldsymbol{B}_t)^{\top}(\nabla ^2\phi(\boldsymbol{X}_t,t))(\boldsymbol{\Sigma}_td\boldsymbol{B}_t)}_{(\bigstar)}\label{eq:ito-intermediate}
\end{align}
\end{small}
Now, we will simplify the ($\bigstar$) term in the equation. Recall that $\boldsymbol{\Sigma}_t\in \mathbb{R}^{d\times d}$, $d\boldsymbol{B}_t\in \mathbb{R}^d$, and $\nabla^2\phi(\boldsymbol{X}_t,t)\in \mathbb{R}^{d\times d}$ by decomposing it into component form. Substituting the matrix-vector product in component form $(\boldsymbol{\Sigma}_td\boldsymbol{B}_t)_i=\sum_{k=1}^d\boldsymbol{\Sigma}_t^{ik}d\boldsymbol{B}_t^k$ into ($\bigstar$), we have: 
\begin{small}
\begin{align}
    (\bluetext{\boldsymbol{\Sigma}_td\boldsymbol{B}_t})^{\top}(\nabla ^2\phi(\boldsymbol{X}_t,t))(\bluetext{\boldsymbol{\Sigma}_td\boldsymbol{B}_t})&=\sum_{i,j=1}^d\left[\left(\bluetext{\sum_{k=1}^d\boldsymbol{\Sigma}_t^{ik}d\boldsymbol{B}_t^k}\right)(\nabla ^2\phi(\boldsymbol{X}_t,t))_{ij}\left(\bluetext{\sum_{\ell=1}^d\boldsymbol{\Sigma}_t^{j\ell}d\boldsymbol{B}_t^\ell}\right)\right]\nonumber\\
    &=\sum_{i,j,k,\ell}\boldsymbol{\Sigma}_t^{ik}\boldsymbol{\Sigma}_t^{j\ell}(\nabla ^2\phi(\boldsymbol{X}_t,t))_{ij}\bluetext{\underbrace{d\boldsymbol{B}_t^kd\boldsymbol{B}_t^\ell}_{\delta_{k\ell dt}}}=\sum_{i,j,k}\pinktext{\underbrace{\boldsymbol{\Sigma}_t^{ik}\boldsymbol{\Sigma}_t^{jk}}_{(\boldsymbol{\Sigma}\boldsymbol{\Sigma}^\top)_{ij}}}(\nabla ^2\phi(\boldsymbol{X}_t,t))_{ij}dt\nonumber\\
    &=\bluetext{\underbrace{\sum_{i,j}(\nabla^2\phi(\boldsymbol{X}_t,t))_{ij}(\boldsymbol{\Sigma}\boldsymbol{\Sigma}^\top)_{ij}}_{\text{Tr}(\boldsymbol{\Sigma}\boldsymbol{\Sigma}^\top\nabla ^2\phi(\boldsymbol{X}_t,t))}}dt=\text{Tr}(\boldsymbol{\Sigma}\boldsymbol{\Sigma}^\top\nabla ^2\phi(\boldsymbol{X}_t,t))dt
\end{align}
\end{small}
Substituting the simplified ($\bigstar$) back into (\ref{eq:ito-intermediate}), we get the final form of the SDE for the transformed Itô process:
\begin{small}
\begin{align}
    d\phi(\boldsymbol{X}_t,t)&=\partial_t\phi(\boldsymbol{X}_t,t)+\nabla \phi(\boldsymbol{X}_t,t)^\top(\boldsymbol{f}_tdt+\boldsymbol{\Sigma}_td\boldsymbol{B}_t)+\frac{1}{2}\text{Tr}\left(\boldsymbol{\Sigma}_t\boldsymbol{\Sigma}_t^\top\nabla ^2\phi(\boldsymbol{X}_t,t)\right)dt\nonumber\\
    &=\bigg[\partial_t\phi(\boldsymbol{X}_t,t)+\boldsymbol{f}(\boldsymbol{X}_t,t)^\top\nabla \phi(\boldsymbol{X}_t,t)+\frac{1}{2}\text{Tr}\left(\boldsymbol{\Sigma}_t\boldsymbol{\Sigma}_t^\top\nabla ^2\phi(\boldsymbol{X}_t,t)\right)\bigg]dt+\nabla \phi(\boldsymbol{X}_t,t)^\top \boldsymbol{\Sigma}_td\boldsymbol{B}_t
\end{align}
\end{small}
which is Itô's formula for an arbitrary twice-continuous function $\phi(\boldsymbol{x},t):\mathbb{R}^d \times [0,T ] \to \mathbb{R}\in C^{2,1}(\mathbb{R}^d\times [0,T])$ and SDE $d\boldsymbol{X}_t=\boldsymbol{f}(\boldsymbol{X}_t,t)dt+ \boldsymbol{\Sigma}_td\boldsymbol{B}_t$. 

When $\phi$ doesn't depend on $t$ and is only a function of the state $\phi(\boldsymbol{x}):\mathbb{R}^d\to \mathbb{R}$, we have $\partial_t\phi(\boldsymbol{x}) =0$. Additionally, when the diffusion is an isotropic Gaussian $\boldsymbol{\Sigma}_t=\sigma_t\boldsymbol{I}_d$, then the trace term reduces to:
\begin{align}
    \text{Tr}\left(\bluetext{\boldsymbol{\Sigma}_t\boldsymbol{\Sigma}_t^\top}\nabla ^2\phi(\boldsymbol{X}_t,t)\right)=\text{Tr}\left(\bluetext{\sigma_t^2\boldsymbol{I}_d\boldsymbol{I}_d^\top}\nabla ^2\phi(\boldsymbol{X}_t,t)\right)=\bluetext{\sigma^2_t}\text{Tr}\left(\nabla ^2\phi(\boldsymbol{X}_t,t)\right)=\bluetext{\sigma_t^2}\Delta \phi(\boldsymbol{X}_t,t)
\end{align}
With these simplifications, the \textbf{Itô formula} becomes:
\begin{align}
    d\phi(\boldsymbol{X}_t,t)&=\bigg[\boldsymbol{f}(\boldsymbol{X}_t,t)^\top\nabla \phi(\boldsymbol{X}_t,t)+\frac{\sigma_t^2}{2}\Delta \phi(\boldsymbol{X}_t  )\bigg]dt+\nabla \phi(\boldsymbol{X}_t,t)^\top\sigma_td\boldsymbol{B}_t\tag{Simplified Itô Process}\label{eq:simplified-ito}
\end{align}
which is the \textbf{key ingredient} to several derivations and proofs that we will study in this guide. \hfill $\square$

Since the Schrödinger bridge problem studies \boldtext{controlled stochastic processes}, we establish the following Corollary which is a direct consequence of our proof for (\ref{eq:ito-2}), but simply replaces the drift of the uncontrolled SDE $\boldsymbol{f}(\boldsymbol{X}_t,t)$ with the drift of the controlled SDE $(\boldsymbol{f}(\boldsymbol{X}_t,t)+\sigma_t\boldsymbol{u}(\boldsymbol{X}_t,t))$ from Definition \ref{def:controlled-ito}.

\begin{corollary}[Itô Formula for Controlled SDEs]\label{corollary:ito-controlled}
    Consider a controlled stochastic process $(\boldsymbol{X}_t\in \mathbb{R}^d)_{t\in [0,T]}$ generated by the controlled SDE:
    \begin{align}
        d\boldsymbol{X}_t=(\boldsymbol{f}(\boldsymbol{X}_t,t)+\sigma_t\boldsymbol{u}(\boldsymbol{X}_t,t))dt+ \sigma_td\boldsymbol{B}_t\tag{Controlled SDE}
    \end{align}
    where $\boldsymbol{u}(\boldsymbol{x},t)$ is the control drift and $\sigma_t$ is the scalar diffusion coefficient. For any scalar function $\phi(\boldsymbol{x},t):\mathbb{R}^d\times[0,T] \to \mathbb{R}\in C^{2,1}(\mathbb{R}^d\times [0,T])$\footnote{where $C^{1,2}$ denotes that the function is continuously differentiable with respect to $t$ and twice continuously differentiable with respect to $\boldsymbol{x}$}, the transformed stochastic process defined by $\phi(\boldsymbol{X}^u_t,t)$ follows the SDE:
    \begin{small}
    \begin{align}
        d\phi(\boldsymbol{X}^u_t,t)=\left[\partial_t\phi(\boldsymbol{X}^u_t,t)+(\boldsymbol{f}+\sigma\boldsymbol{u})(\boldsymbol{X}^u_t,t)^\top \nabla \phi(\boldsymbol{X}^u_t,t)+\frac{\sigma_t^2}{2}\Delta \phi(\boldsymbol{X}^u_t,t)\right]dt+ \nabla \phi(\boldsymbol{X}^u_t,t)^\top \sigma_td\boldsymbol{B}_t\tag{Controlled Itô Formula}\label{eq:control-ito-formula}
    \end{align}
    \end{small}
    If $\phi(\boldsymbol{x}):\mathbb{R}^d \to \mathbb{R}\in C^2(\mathbb{R}^d)$ is a scalar function dependent only on $\boldsymbol{x}$, then the partial derivative $\partial_t\phi$ disappears, and $\phi(\boldsymbol{X}^u_t)$ follows the SDE:
    \begin{small}
    \begin{align}
        d\phi(\boldsymbol{X}^u_t)=\left[(\boldsymbol{f}+\sigma\boldsymbol{u})(\boldsymbol{X}^u_t)^\top \nabla \phi(\boldsymbol{X}^u_t,t)+\frac{\sigma_t^2}{2}\Delta \phi(\boldsymbol{X}^u_t)\right]dt+ \nabla \phi(\boldsymbol{X}^u_t)^\top \sigma_td\boldsymbol{B}_t
    \end{align}
    \end{small}
\end{corollary}

The controlled Itô formula shows that, for any sufficiently smooth test function, the drift term of the transformed process is determined by a \textbf{linear differential operator} acting on that function. This operator is called the \boldtext{infinitesimal generator of the SDE} $\mathcal{A}_t$, which describes the instantaneous rate of change of the expected value of $\phi(\boldsymbol{X}_t,t)$ at time $t$, explicitly written as:
\begin{align}
    (\mathcal{A}_t\phi)(\boldsymbol{x},t)=\lim_{\Delta t\to 0}\left[\frac{\mathbb{E}[\phi(\boldsymbol{X}_{t+\Delta t},t+\Delta t)|\boldsymbol{X}_t=\boldsymbol{x}]-\phi(\boldsymbol{x},t)}{\Delta t}\right]\tag{Infinitesmal Generator}
\end{align}
which plays a central role in describing the local evolution of observables.

\begin{definition}[Infinitesimal Generator of Itô Process]\label{def:infinitesmal generator}
    For any test function $\phi(\boldsymbol{x},t)\in C^{2,1}(\mathbb{R}^d\times [0,T])$, the \textbf{infinitesimal generator of the uncontrolled SDE} is the operator $\mathcal{A}_t$ defined as:
    \begin{small}
    \begin{align}
        (\mathcal{A}_t\phi)(\boldsymbol{x},t):=\boldsymbol{f}(\boldsymbol{x},t)^\top \nabla \phi(\boldsymbol{x},t)+\frac{\sigma^2_t}{2}\Delta \phi(\boldsymbol{x},t)\tag{Uncontrolled Generator}\label{eq:ito-uncontrolled-generator}
    \end{align}
    \end{small}
    and the \textbf{infinitesimal generator of the controlled SDE} is the operator $\mathcal{A}_t^u$ defined as:
    \begin{small}
    \begin{align}
        (\mathcal{A}^u_t\phi)(\boldsymbol{x},t):=(\boldsymbol{f}(\boldsymbol{x},t)+\sigma_t\boldsymbol{u}(\boldsymbol{x},t))^\top \nabla\phi(\boldsymbol{x},t)+\frac{\sigma^2_t}{2}\Delta \phi(\boldsymbol{x},t)\tag{Controlled Generator}\label{eq:ito-controlled-generator}
    \end{align}
    \end{small}
    Equivalently, the controlled generator can be written in terms of the uncontrolled generator by expanding the divergence as:
    \begin{small}
    \begin{align}
        (\mathcal{A}^u_t\phi)(\boldsymbol{x},t)&=\bluetext{\boldsymbol{f}(\boldsymbol{x},t)^\top \nabla \phi(\boldsymbol{x},t)}+\frac{\sigma^2_t}{2}\Delta \phi(\boldsymbol{x},t)+\sigma_t\boldsymbol{u}(\boldsymbol{x},t)^\top \nabla\phi(\boldsymbol{x},t)\nonumber\\
        &=\bluetext{(\mathcal{A}_t\phi)(\boldsymbol{x},t)}+\boldsymbol{u}(\boldsymbol{x},t)^\top\nabla\phi(\boldsymbol{x},t)\label{eq:generator-control-to-uncontrolled}
    \end{align}
    \end{small}
\end{definition}

This definition of the infinitesimal generator can be used to rewrite Itô's formula for uncontrolled and controlled SDEs as:
\begin{align}
    d\phi(\boldsymbol{X}_t,t)&=(\partial_t+\mathcal{A}_t)\phi(\boldsymbol{X}_t,t)dt+\sigma_t\nabla \phi(\boldsymbol{X}_t,t)^\top d\boldsymbol{B}_t\tag{Uncontrolled Itô Formula with $\mathcal{A}_t$}\label{eq:uncontrolled-ito-formula-generator}\nonumber\\
    d\phi(\boldsymbol{X}^u_t,t)&=(\partial_t+\mathcal{A}^u_t)\phi(\boldsymbol{X}^u_t,t)dt+\sigma_t\nabla \phi(\boldsymbol{X}^u_t,t)^\top d\boldsymbol{B}_t\tag{Controlled Itô Formula with $\mathcal{A}^u_t$}\label{eq:controlled-ito-formula-generator}
\end{align}

Using the infinitesimal generator, we can define a specific family of stochastic processes, called \boldtext{martingales}.

\begin{definition}[Martingales]\label{def:martingale}
    A $\mathcal{F}_t$-adapted stochastic process $\boldsymbol{Y}_{0:T}:=(\boldsymbol{Y}_t)_{t\in [0,T]}$ is called a \textbf{martingale} if it is integrable for all $t\in [0,T]$ (i.e., $\mathbb{E}[|\boldsymbol{Y}_t|]< \infty$) and satisfies the \textbf{martingale property}:
    \begin{align}
        \mathbb{E}[\boldsymbol{Y}_t|\mathcal{F}_s]=\boldsymbol{Y}_s\implies \mathbb{E}[\boldsymbol{Y}_t]=\mathbb{E}[\boldsymbol{Y}_s], \quad \forall 0\leq s\leq t\tag{Martingale Property}\label{eq:martingale-property}
    \end{align}
    which means that the expectation is \textit{constant over time}. Furthermore, we have that if $\boldsymbol{Y}_t:= \phi(\boldsymbol{X}_t,t)$ is the martingale process generated by the function $\phi(\boldsymbol{x},t)\in C^{2,1}(\mathbb{R}^d\times[0,T])$, the uncontrolled generator vanishes, such that:
    \begin{align}
        (\mathcal{A}_t\phi)(\boldsymbol{X}_t,t)=\boldsymbol{f}(\boldsymbol{X}_t,t)^\top \nabla \phi(\boldsymbol{X}_t,t)+\frac{\sigma^2_t}{2}\Delta \phi(\boldsymbol{X}_t,t)=0
    \end{align}
\end{definition}

Given the definition of Itô processes as stochastic differential equations with drift and diffusion terms, we observe that there are infinitely many possible paths under the same SDE due to the non-deterministic diffusion term, which injects randomness into each trajectory; however, certain paths are more likely than others under a given drift. 

Sampling an infinite number of paths under a specific SDE forms a \textbf{probability distribution in the path space}, known as a \boldtext{path measure} $\mathbb{P}\in \mathcal{P}(C([0,T]; \mathbb{R}^d))$, which given an Itô process $\boldsymbol{X}_{0:T}\in C([0,T]; \mathbb{R}^d)$ returns the probability that the process was generated under that SDE. Specifically, we will consider the general space of \boldtext{controlled path measures} with a control drift $\boldsymbol{u}(\boldsymbol{x},t)$, where $\boldsymbol{u}\equiv0$ yields the \boldtext{reference path measure} $\mathbb{Q}$.

\begin{definition}[Controlled Path Measure]\label{def:path-measure}
A \textbf{controlled path measure} $\mathbb{P^u}\in C([0,T]; \mathbb{R}^d)$ is a probability measure on the path space $C([0,T]; \mathbb{R}^d)$ induced by a stochastic differential equation (SDE) with control $\sigma_t\boldsymbol{u}(\boldsymbol{x},t)$ defined in (\ref{eq:controlled-sde}). For any measurable set of trajectories $S\subseteq C([0,T]; \mathbb{R}^d)$, $\mathbb{P}^u(S)$ is the probability that the trajectories in the set are generated under the controlled SDE:
\begin{align}
    \mathbb{P}^u(S)=\text{Pr}\left(\boldsymbol{X}_{0:T}\in S\text{ under the controlled SDE of }\mathbb{P}^u\right) 
\end{align}
Throughout this guide, we will use a slight abuse of notation and denote the probability of a stochastic path under $\mathbb{P}^u$ as $\mathbb{P}^u(\boldsymbol{X}_{0:T})$. 
\end{definition}

Next, we will define how the marginal densities evolve over the continuous time interval via the \textbf{Fokker-Planck equation}, which provides an alternative representation of path measures as deterministic \textbf{partial differential equations} instead of non-deterministic SDEs.

\subsection{Fokker-Planck and Feynman-Kac Equations}
\label{subsec:fp-equation}
Defining a path measure with an SDE only captures the evolution of single particles along the trajectories supported by the path measure, but tells us nothing about the evolution of the distribution generated by many particles under a path measure. The \boldtext{Fokker-Planck equation} allows us to define the behavior of the \textbf{probability distribution} of particles $p_t$ at each time point $t\in [0,T]$, such that individual states simulated by the SDE in (\ref{eq:ito-sde}) are samples $\boldsymbol{X}_t\sim p_t$.

\begin{theorem}[Fokker-Planck Equation]\label{thm:fokker-planck}
    Let $\boldsymbol{X}_{0:T}$ be a stochastic process governed by the SDE:
    \begin{align}
        d\boldsymbol{X}_t=\boldsymbol{f}(\boldsymbol{X}_t,t)dt+ \sigma_td\boldsymbol{B}_t
    \end{align}
    where $\boldsymbol{f}(\boldsymbol{x},t)$ is the drift and $\sigma_t\in \mathbb{R}$ is the scalar diffusion coefficient. Then, the marginal density $p_t\in \mathcal{P}(\mathbb{R}^d)$ of the particles generated by the SDE evolves via the \textbf{Fokker-Planck equation} defined as:
    \begin{align}
        \partial_tp_t(\boldsymbol{x}) =-\nabla\cdot (\boldsymbol{f}(\boldsymbol{x},t)p_t(\boldsymbol{x}))+\frac{\sigma_t^2}{2}\Delta p_t(\boldsymbol{x})\tag{Fokker-Planck Equation}\label{eq:fokker-planck equation}
    \end{align}
    where $\boldsymbol{X}_t \sim p_t$. Equivalently, $p_t$ is probability law of $\boldsymbol{X}_t$, denoted $\text{Law}(\boldsymbol{X}_t)=p_t$.
\end{theorem}

\textit{Intuition.} Before we derive the Fokker-Planck equation explicitly, let us first consider the problem of characterizing the distribution over particles generated from an SDE without direct access to the density $p_t$. From Itô's formula (Theorem \ref{thm:ito-formula}), we can define the evolution of samples from the SDE via twice-differentiable functions over the state space $\mathbb{R}^d$. 

With the capability to study the random variable with arbitrary smooth functions, known as \textbf{test functions} $\phi(\boldsymbol{x}):\mathbb{R}^d\to \mathbb{R}$, we can extract different pieces of information about the distribution $p_t$ by taking the expectation $\mathbb{E}_{p_t}[\phi(\boldsymbol{X}_t)]$. For instance, the expectation of the identity function $\phi(\boldsymbol{x})=\boldsymbol{x}$ measures how the mean of the distribution evolves over time, and the squared function $\phi(\boldsymbol{x})=\boldsymbol{x}^2$ measures how the variance evolves over time. Unlike the test function used to derive Itô's formula, we don't require it to depend on time, since we use it only to probe the spatial structure of the density at each time. Intuitively, by analyzing the evolution of \textit{any} arbitrary smooth functions on $\boldsymbol{X}_t$, we can derive the evolution of the density $p_t$.

\textit{Derivation.} Consider a test function $\phi(\boldsymbol{x})\in C_c^2(\mathbb{R}^d)$ that is twice differentiable and has compact support such that the boundary terms vanish in integration by parts. By Itô's formula in (\ref{eq:simplified-ito}), the random variable $\phi(\boldsymbol{X}_t)$ is an Itô process that evolves via the SDE:
\begin{align}
    d\phi(\boldsymbol{X}_t)=\bigg[\underbrace{\boldsymbol{f}(\boldsymbol{X}_t,t)^\top \nabla \phi(\boldsymbol{X}_t)}_{\text{drift contribution}}+\underbrace{\frac{\sigma_t^2}{2}\Delta \phi(\boldsymbol{X}_t)}_{\text{diffusion contribution}}\bigg]dt+\underbrace{\sigma_t\nabla \phi(\boldsymbol{X}_t)^\top d\boldsymbol{B}_t}_{\text{stochastic term}}
\end{align}
Since we want to define the evolution on a distribution level, we take the expectation on both sides, which eliminates the stochastic term since the Itô integral has zero expectation:
\begin{small}
\begin{align}
    \mathbb{E}_{p_t}\left[d\phi(\boldsymbol{X}_t)\right]&=\mathbb{E}_{p_t}\left[\boldsymbol{f}(\boldsymbol{X}_t,t)^\top \nabla \phi(\boldsymbol{X}_t)+\frac{\sigma_t^2}{2}\Delta \phi(\boldsymbol{X}_t)\right]dt+\underbrace{\mathbb{E}_{p_t}\left[\sigma_t\nabla \phi(\boldsymbol{X}_t)^\top d\boldsymbol{B}_t\right]}_{=0}\nonumber\\
    &=\mathbb{E}_{p_t}\left[\boldsymbol{f}(\boldsymbol{X}_t,t)^\top \nabla \phi(\boldsymbol{X}_t)+\frac{\sigma_t^2}{2}\Delta \phi(\boldsymbol{X}_t)\right]dt
\end{align}
\end{small}
Dividing both sides by $dt$, we have:
\begin{small}
\begin{align}
    \partial_t\mathbb{E}_{p_t}\left[\phi(\boldsymbol{X}_t)\right]&=\mathbb{E}_{p_t}\left[\boldsymbol{f}(\boldsymbol{X}_t,t)^\top\nabla \phi(\boldsymbol{X}_t)+\frac{\sigma_t^2}{2}\Delta \phi(\boldsymbol{X}_t)\right]
\end{align}
\end{small}
By definition, the expectations can be written in integral form as $\mathbb{E}_{p_t(\boldsymbol{x})}[\cdot]=\int_{\mathbb{R}^d}(\cdot)p_td\boldsymbol{x}$ which gives:
\begin{small}
\begin{align}
    \partial_t\int_{\mathbb{R}^d}\phi(\boldsymbol{x})p_t(\boldsymbol{x})d\boldsymbol{x}&=\int_{\mathbb{R}^d}\left[\boldsymbol{f}(\boldsymbol{x},t)^\top\nabla \phi(\boldsymbol{x})+\frac{\sigma_t^2}{2}\Delta \phi(\boldsymbol{x})\right]p_t(\boldsymbol{x})d\boldsymbol{x}\\
    \int_{\mathbb{R}^d}\phi(\boldsymbol{x})\partial_tp_t(\boldsymbol{x})d\boldsymbol{x}&=\underbrace{\int_{\mathbb{R}^d}\boldsymbol{f}(\boldsymbol{x},t)^\top  \nabla \phi(\boldsymbol{x})p_t(\boldsymbol{x})d\boldsymbol{x}}_{\text{drift term}}+\underbrace{\frac{1}{2}\int_{\mathbb{R}^d}\sigma_t^2\Delta\phi(\boldsymbol{x})p_t(\boldsymbol{x})d\boldsymbol{x}}_{\text{diffusion term}}\label{eq:fp-intermediate}
\end{align}
\end{small}
For the \textbf{drift term}, $\boldsymbol{f}(\boldsymbol{x},t)\in \mathbb{R}^d$ and $\nabla \phi(\boldsymbol{x})\in \mathbb{R}^d$, we can apply integration by parts on each coordinate to get:
\begin{small}
\begin{align}
    \int_{\mathbb{R}^d}\boldsymbol{f}(\boldsymbol{x},t)\cdot \nabla \phi(\boldsymbol{x})p_t(\boldsymbol{x})d\boldsymbol{x}&=\int_{\mathbb{R}^d}\sum_{i=1}^d\boldsymbol{f}_i(\boldsymbol{x},t) \frac{\partial\phi(\boldsymbol{x})}{\partial \boldsymbol{x}_i}p_t(\boldsymbol{x})d\boldsymbol{x}\nonumber\\
    &=\int_{\mathbb{R}^d}\sum_{i=1}^d\underbrace{\frac{\partial\phi(\boldsymbol{x})}{\partial \boldsymbol{x}_i}}_{dv}\underbrace{\big(\boldsymbol{f}_i(\boldsymbol{x},t) p_t(\boldsymbol{x})\big)}_{u}d\boldsymbol{x}\tag{1st integration by parts}\\
    &=-\int_{\mathbb{R}^d}\sum_{i=1}^d\underbrace{\phi(\boldsymbol{x})}_{v}\underbrace{\frac{\partial}{\partial \boldsymbol{x}_i}\big(\boldsymbol{f}_i(\boldsymbol{x},t) p_t(\boldsymbol{x})\big)}_{du}d\boldsymbol{x}\tag{2nd integration by parts}\nonumber\\
    &=-\int_{\mathbb{R}^d}\phi(\boldsymbol{x})\underbrace{\sum_{i=1}^d\frac{\partial}{\partial \boldsymbol{x}_i}\big(\boldsymbol{f}_i(\boldsymbol{x},t) p_t(\boldsymbol{x})\big)}_{=\nabla\cdot (\boldsymbol{f}(\boldsymbol{x},t) p_t(\boldsymbol{x}))}d\boldsymbol{x}\nonumber\\
    &=\boxed{-\int_{\mathbb{R}^d}\phi(\boldsymbol{x})\nabla\cdot (\boldsymbol{f}(\boldsymbol{x},t) p_t(\boldsymbol{x}))d\boldsymbol{x}}\tag{drift term}
\end{align}
\end{small}
where all boundary terms vanish since $\phi$ has compact support. For the \textbf{diffusion term}, we apply integration by parts twice as follows:
\begin{small}
\begin{align}
    \frac{1}{2}\int_{\mathbb{R}^d}\sigma_t^2\Delta \phi(\boldsymbol{x})p_t(\boldsymbol{x})d\boldsymbol{x}&=\frac{\sigma_t^2}{2}\int_{\mathbb{R}^d}p_t(\boldsymbol{x})\sum_{i=1}^d\frac{\partial^2}{\partial\boldsymbol{x}_i \partial\boldsymbol{x}_i} \phi(\boldsymbol{x})d\boldsymbol{x}\nonumber\\
    &=\frac{\sigma_t^2}{2}\sum_{i=1}^d\int_{\mathbb{R}^d}\underbrace{p_t(\boldsymbol{x})}_{u}\underbrace{\frac{\partial}{\partial\boldsymbol{x}_i}\left( \frac{\partial}{\partial\boldsymbol{x}_i}\phi(\boldsymbol{x})\right)}_{dv}d\boldsymbol{x}\tag{1st integration by parts}\\
    &= -\frac{\sigma_t^2}{2}\sum_{i=1}^d\int_{\mathbb{R}^d}\underbrace{\frac{\partial}{\partial\boldsymbol{x}_i} p_t(\boldsymbol{x})}_{u}\underbrace{\frac{\partial}{\partial\boldsymbol{x}_i}\phi(\boldsymbol{x})}_{dv}d\boldsymbol{x}\tag{2nd integration by parts}\\
    &=\frac{\sigma_t^2}{2}\sum_{i=1}^d\int_{\mathbb{R}^d}\phi(\boldsymbol{x})\frac{\partial^2}{\partial\boldsymbol{x}_i\partial \boldsymbol{x}_i}p_t(\boldsymbol{x})d\boldsymbol{x}\nonumber\\
    &=\frac{\sigma_t^2}{2}\int_{\mathbb{R}^d}\phi(\boldsymbol{x})\underbrace{\sum_{i=1}^d\frac{\partial^2}{\partial\boldsymbol{x}_i\partial \boldsymbol{x}_i}p_t(\boldsymbol{x})}_{\Delta p_t(\boldsymbol{x})}d\boldsymbol{x}\nonumber\\
    &=\boxed{\frac{\sigma_t^2}{2}\int_{\mathbb{R}^d}\phi(\boldsymbol{x})\Delta p_td\boldsymbol{x}}\tag{diffusion term}
\end{align}
\end{small}
Substituting the simplified drift and diffusion terms back into (\ref{eq:fp-intermediate}), we obtain:
\begin{small}
\begin{align}
    \partial_t\int_{\mathbb{R}^d}\phi(\boldsymbol{x})p_t(\boldsymbol{x})d\boldsymbol{x}&=\underbrace{-\int_{\mathbb{R}^d}\phi(\boldsymbol{x})\nabla\cdot (\boldsymbol{f}(\boldsymbol{x},t) p_t(\boldsymbol{x}))d\boldsymbol{x}}_{\text{drift term}}+\underbrace{\frac{1}{2}\int_{\mathbb{R}^d}\phi(\boldsymbol{x})\sigma_t^2\Delta p_t(\boldsymbol{x})d\boldsymbol{x}}_{\text{diffusion term}}\nonumber\\
    &=\int_{\mathbb{R}^d}\phi(\boldsymbol{x})\bigg(-\nabla\cdot (\boldsymbol{f}(\boldsymbol{x},t) p_t(\boldsymbol{x}))+\frac{\sigma_t^2}{2}\Delta p_t(\boldsymbol{x})\bigg)d\boldsymbol{x}
\end{align}
\end{small}
Since we defined $\phi(\boldsymbol{x})$ such that this equality holds for \textit{any arbitrary test function}, then the integrands must be equal which recovers the \textbf{Fokker-Planck} (FP) equation:
\begin{align}
    \boxed{\partial_tp_t(\boldsymbol{x})=-\nabla\cdot (\boldsymbol{f}(\boldsymbol{x},t) p_t(\boldsymbol{x}))+\frac{\sigma_t^2}{2}\Delta p_t(\boldsymbol{x})}\tag{Fokker-Planck Equation}
\end{align}
Furthermore, the density $p_t$ that solves the FP equation is \textit{unique} givne a fixed initial condition\footnote{For rigorous proof of uniqueness, see \citet{bogachev2021uniqueness}}.\hfill $\square$

By testing against an arbitrary smooth test function and equating integrands, we recover the Fokker–Planck equation that governs the time evolution of the density $p_t(\boldsymbol{x})$. We observe that the drift term $\boldsymbol{f}(\boldsymbol{x},t)$ induces the transport of mass via divergence, while the diffusion coefficient $\sigma_t$ contributes to smoothing through the Laplacian operator. The Fokker–Planck equation therefore provides a forward-time description of the stochastic dynamics, complementing the underlying stochastic differential equation.

\begin{remark}[Fokker-Planck Equation Generalizes the Continuity Equation]\label{remark:fokker-planck-continuity}
The Fokker-Planck equation can be interpreted as a \textbf{stochastic generalization of the classic continuity equation} from the (\ref{eq:dynamic-ot-prob}), which ensures the conservation of probability mass, defined as:
\begin{align}
    \partial_t p_t(\boldsymbol{x}) + \nabla \cdot \big(p_t(\boldsymbol{x}) \boldsymbol{v}(\boldsymbol{x},t)\big) = 0\tag{Continuity Equation}\label{eq:continuity-equation-remark}
\end{align}
which expresses that probability mass is transported along the flow induced by the velocity $\boldsymbol{v}(\boldsymbol{x},t)$. The Fokker-Planck equation is equal to the continuity equation with an additional Laplacian term $\frac{\sigma_t^2}{2}\Delta p_t$ that captures the \textbf{spreading of mass caused by Brownian motion}. 

Intuitively, the Laplacian is the \textbf{divergence of the gradient} $\Delta p_t=\nabla\cdot\nabla p_t$ which measures whether the direction of steepest increase in probability density $\nabla p_t$ is spreading out (positive divergence) or converging (negative divergence), so the positive diffusion coefficient increases the \textbf{spreading out of probability density}.
\end{remark}

Using this intuition, we can easily extend the Fokker-Planck equation to controlled processes by introducing the scaled control $\sigma_t\boldsymbol{u}(\boldsymbol{x},t)$ into the drift contribution, yielding the \boldtext{controlled Fokker-Planck equation} that governs the evolution of the controlled marginal density.

\begin{corollary}[Controlled Fokker-Planck Equation]
\label{cor:controlled-fp}
Let $(\boldsymbol{X}^u_t)_{t\in [0,T]}$ be a stochastic process governed by the SDE:
\begin{align}
    d\boldsymbol{X}^u_t = \left(\boldsymbol{f}(\boldsymbol{X}^u_t,t) + \sigma_t \boldsymbol{u}(\boldsymbol{X}^u_t,t)\right)dt + \sigma_t d\boldsymbol{B}_t,
\end{align}
where $\boldsymbol{u}(\boldsymbol{x},t)$ is the control drift, $\boldsymbol{f}(\boldsymbol{x},t)$ is the reference drift and $\sigma_t\in \mathbb{R}$ is the scalar diffusion coefficient. Let $p_t\in \mathcal{P}(\mathbb{R}^d)$ denote the marginal density or probability law of $\boldsymbol{X}^u_t$, such that $\boldsymbol{X}^u_t \sim p_t$ and $\text{Law}(\boldsymbol{X}_t^u)=p_t$. Then $p_t$ evolves according to the \textbf{controlled Fokker-Planck equation}
\begin{small}
\begin{align}
    \partial_t p_t(\boldsymbol{x})=-\nabla \cdot\left(\left(\boldsymbol{f}(\boldsymbol{x},t)+\sigma_t\boldsymbol{u}(\boldsymbol{x},t)\right)p_t(\boldsymbol{x})\right)+\frac{\sigma_t^2}{2} \Delta p_t(\boldsymbol{x})\tag{Controlled Fokker-Planck Equation}\label{eq:controlled-fp-equation}
\end{align}
\end{small}
Equivalently, expanding the divergence term gives
\begin{align}
    \partial_t p_t(\boldsymbol{x})=-\nabla \cdot\left(\boldsymbol{f}(\boldsymbol{x},t)p_t(\boldsymbol{x})\right)-\nabla \cdot\bigl(\sigma_t\boldsymbol{u}(\boldsymbol{x},t)p_t(\boldsymbol{x})\bigr)
    + \frac{\sigma_t^2}{2} \Delta p_t(\boldsymbol{x})
\end{align}
\end{corollary}

We have shown that uncontrolled and controlled Itô processes can not only be written as stochastic differential equations (SDEs) that govern the forward-time evolution of individual trajectories $\boldsymbol{X}_t$, but also admit an \textbf{equivalent deterministic representation through the Fokker-Planck equation}, which is a deterministic partial differential equation (PDE) that governs the forward-time evolution of the \textit{marginal density} $p_t$ that characterizes the probability law of the SDE trajectories. 

However, the Schrödinger bridge problem is inherently a two-sided problem enforced by both initial \textit{and terminal} marginal distribution constraints. Therefore, we need to consider how constraints or functions evaluated on the terminal distribution $p_T$ evolve \textbf{backward in time} through the stochastic process to determine the probability of an intermediate state under the terminal constraint. This evolution is captured by the \boldtext{Feynman-Kac equation}, which we analyze next.

\begin{theorem}[Feynman-Kac Equation]\label{thm:feynman-kac}
    Consider a scalar function $r(\boldsymbol{x},t):\mathbb{R}^d\times[0,T]\to \mathbb{R}$ that solves the linear PDE with a terminal constraint $r(\boldsymbol{x},T)=\Phi(\boldsymbol{x})$ and running cost $c(\boldsymbol{x},t):\mathbb{R}^d\times[0,T] \to \mathbb{R}$\footnote{Note that from now on, $c(\boldsymbol{x},t)$ will be used to denote the running cost, which is distinct from $c(\boldsymbol{x},\boldsymbol{y})$ which is used in Section \ref{sec:static-sb} to denote the transport cost from $\boldsymbol{x}$ to $\boldsymbol{y}$} defined as:
    \begin{small}
    \begin{align}
        \partial_tr(\boldsymbol{x},t)+\langle\boldsymbol{f}(\boldsymbol{x},t), \nabla r(\boldsymbol{x},t)\rangle + \frac{\sigma_t^2}{2}\Delta r(\boldsymbol{x},t)-c(\boldsymbol{x},t)r(\boldsymbol{x},t)=0, \quad r(\boldsymbol{x},T)=\Phi(\boldsymbol{x})\tag{Feynman-Kac PDE}\label{eq:feynman-kac-pde}
    \end{align}
    \end{small}
    where $\boldsymbol{f}(\boldsymbol{x},t)$ is the reference drift and $\sigma_t$ is the diffusion coefficient of an SDE that generates a stochastic process $(\boldsymbol{X}_\tau)_{\tau\in [t,T]}$ defined by:
    \begin{align}
        d\boldsymbol{X}_\tau=\boldsymbol{f}(\boldsymbol{X}_\tau,\tau)d\tau+ \sigma_\tau d\boldsymbol{B}_\tau,\quad \boldsymbol{X}_t=\boldsymbol{x}\label{eq:feynman-kac-sde}
    \end{align}
    Then, the solution $r$ that solves the (\ref{eq:feynman-kac-pde}) can be written as the following expectation over stochastic paths:
    \begin{align}
        r(\boldsymbol{x},t)=\mathbb{E}\left[\exp \left(-\int_t^Tc(\boldsymbol{X}_s,s)ds\right)\Phi(\boldsymbol{X}_T)\bigg|\boldsymbol{X}_t=\boldsymbol{x}\right]\tag{Feynman-Kac Formula}\label{eq:feynman-kac-formula}
    \end{align}
    which is called the \textbf{Feynman-Kac formula}.
\end{theorem}

\textit{Derivation.} Intuitively, $r(\boldsymbol{x},t)$ is the expected future value of the terminal constraint $\Phi(\boldsymbol{X}_T)$ after discounting the running cost $c(\boldsymbol{X}_s,s)$ given the current state $\boldsymbol{X}_t=\boldsymbol{x}$ at time $t$. With this intuition in mind, we derive the Feynman-Kac formula in steps.

\textbf{Step 1: Define the Stochastic Processes. }
We can consider the following stochastic process that describes the evolution of the accumulated running cost over $[\tau,t]$:
\begin{small}
\begin{align}
    \boldsymbol{M}_\tau:=\underbrace{\exp \left(-\int_t^\tau c(\boldsymbol{X}_s,s)ds\right)}_{=:\boldsymbol{Z}_\tau}r(\boldsymbol{X}_\tau,\tau), \quad \tau\in [t,T]\label{eq:fp-proof4}
\end{align}
\end{small}
Since the exponential term is itself a stochastic process that changes with $\tau$, we can define $\boldsymbol{Z}_\tau:=\exp \left(-\int_t^\tau c(\boldsymbol{X}_s,s)ds\right)$ and $\boldsymbol{M}_\tau:=\boldsymbol{Z}_\tau r(\boldsymbol{X}_\tau,\tau)$. Now, we aim to show that $\boldsymbol{M}_\tau$ is a \textbf{martingale} whose expected value does not change over time (Definition \ref{def:martingale}). In other words, we aim to show that the \textit{expected future reward} $\mathbb{E}[\boldsymbol{M}_T]$ is equal to the current value $\boldsymbol{M}_t$, i.e., $\mathbb{E}[\boldsymbol{M}_T]=\boldsymbol{M}_t$.

\textbf{Step 2: Apply Itô's Formula. }
Applying \textbf{Itô's product rule} to $\boldsymbol{M}_s$ and $r(\boldsymbol{X}_s,s)$, we get the following expression for $d\boldsymbol{M}_s$:
\begin{small}
\begin{align}
    d\boldsymbol{M}_\tau=\boldsymbol{Z}_\tau dr(\boldsymbol{X}_\tau,\tau)+ r(\boldsymbol{X}_\tau,\tau)d\boldsymbol{Z}_s+\bluetext{\underbrace{d\boldsymbol{Z}_sdr(\boldsymbol{X}_\tau,\tau)}_{=O(d\tau^2)=0}}=\boldsymbol{Z}_\tau\underbrace{dr(\boldsymbol{X}_\tau,\tau)}_{(\bigstar)}+ r(\boldsymbol{X}_\tau,\tau)\underbrace{d\boldsymbol{Z}_\tau}_{(\diamond)}\label{eq:fp-proof3}
\end{align}
\end{small}
To expand $(\bigstar)$, we can apply (\ref{eq:ito-2}) to get:
\begin{small}
\begin{align}
    dr(\boldsymbol{X}_\tau,\tau)&=\bigg[\partial_\tau r(\boldsymbol{X}_\tau,\tau)+\bluetext{\underbrace{\langle \boldsymbol{f}(\boldsymbol{X}_\tau,\tau), \nabla r(\boldsymbol{X}_\tau,\tau)\rangle+ \frac{\sigma^2_t}{2}\Delta r(\boldsymbol{X}_\tau,\tau)}_{\text{uncontrolled generator applied to }r}}\bigg]d\tau+ \nabla r(\boldsymbol{X}_\tau,\tau)^\top \sigma_td\boldsymbol{B}_\tau\nonumber\\
    &=(\partial_\tau  r(\boldsymbol{X}_\tau,\tau)+\mathcal{A}_tr(\boldsymbol{X}_\tau,\tau))d\tau+\nabla r(\boldsymbol{X}_\tau,\tau)^\top \sigma_td\boldsymbol{B}_\tau\tag{$\bigstar$}\label{eq:fp-proof1}
\end{align}
\end{small}
where we recognize the unconditional generator $\mathcal{A}_t$ from (\ref{eq:ito-uncontrolled-generator}). To expand $(\diamond)$, we denote $\boldsymbol{A}_\tau := -\int_t^\tau c(\boldsymbol{X}_s,s)ds$ and apply the chain rule:
\begin{small}
\begin{align}
    d\boldsymbol{Z}_\tau=d(e^{\boldsymbol{A}_\tau})=\underbrace{e^{\boldsymbol{A}_\tau}}_{=\boldsymbol{Z}_\tau}\bluetext{d\boldsymbol{A}_\tau}=\boldsymbol{Z}_\tau\bluetext{\frac{d\boldsymbol{A}_\tau}{d\tau}}d\tau=\boldsymbol{Z}_\tau(\bluetext{-c(\boldsymbol{X}_\tau,\tau)}d\tau)=-c(\boldsymbol{X}_\tau,\tau)\boldsymbol{Z}_\tau d\tau\tag{$\diamond$}\label{eq:fp-proof2}
\end{align}
\end{small}
where the expression for $\frac{d\boldsymbol{A}_\tau}{d\tau}=-c(\boldsymbol{X}_\tau,\tau)$ follows from the fundamental theorem of calculus for integrals with variable upper limit. Substituting (\ref{eq:fp-proof1}) and (\ref{eq:fp-proof2}) back into (\ref{eq:fp-proof3}), we get: 
\begin{small}
\begin{align}
    d\boldsymbol{M}_\tau&=\boldsymbol{Z}_\tau\underbrace{\left[(\partial_\tau r+\mathcal{A}r)d\tau+\nabla r^\top \sigma_td\boldsymbol{B}_\tau\right]}_{(\bigstar)}+ r\underbrace{\left(-c\boldsymbol{Z}_\tau d\tau\right)}_{(\diamond)}\nonumber\\
    &=\boldsymbol{Z}_\tau\bluetext{\underbrace{\left(\partial_\tau r+\mathcal{A}r-cr\right)}_{=0}}d\tau+ \boldsymbol{Z}_\tau\nabla r^\top \sigma_td\boldsymbol{B}_\tau
\end{align}
\end{small}
where we observe that the drift term is exactly equal to the (\ref{eq:feynman-kac-pde}), which is equal to zero given the solution $r(\boldsymbol{x},t)$. Therefore, the drift disappears, and $d\boldsymbol{M}_\tau=\boldsymbol{Z}_\tau\nabla r^\top \sigma_td\boldsymbol{B}_\tau$ contains only a diffusion term and is a \textbf{martingale} with constant zero-expectation. Therefore, we have:
\begin{small}
\begin{align}
    &\boldsymbol{M}_T=\boldsymbol{M}_t+ \int _t^Td\boldsymbol{M}_\tau=\boldsymbol{M}_t+\int_t^T\boldsymbol{Z}_\tau\nabla r^\top \sigma_td\boldsymbol{B}_\tau\nonumber\\
    &\implies \mathbb{E}[\boldsymbol{M}_T]=\mathbb{E}[\boldsymbol{M}_t]+\underbrace{\mathbb{E}\left[\int_t^T\boldsymbol{Z}_\tau\nabla r^\top \sigma_td\boldsymbol{B}_\tau\right]}_{\text{Itô integral }=0}\nonumber\\
    &\implies \mathbb{E}[\boldsymbol{M}_T]=\mathbb{E}[\boldsymbol{M}_t]=\boldsymbol{M}_t
\end{align}
\end{small}
where the final equality follows from the fact that $\boldsymbol{M}_t$ is the starting time, and the accumulated cost over $[t,t]$ is 0.

\textbf{Step 3: Substitute Terminal Constraint. }
By our definition in (\ref{eq:fp-proof4}), we have:
\begin{small}
\begin{align}
    &\boldsymbol{M}_T=\exp \left(-\int_t^Tc(\boldsymbol{X}_s,s)ds\right)\underbrace{r(\boldsymbol{X}_T,T)}_{=\Phi(\boldsymbol{X}_T)}=\exp \left(-\int_t^Tc(\boldsymbol{X}_s,s)ds\right)\Phi(\boldsymbol{X}_T)
\end{align}
\end{small}
Given that $\boldsymbol{M}_t=r(\boldsymbol{x},t)$ and $\boldsymbol{M}_t=\mathbb{E}[\boldsymbol{M}_T]$, we have shown that the solution $r(\boldsymbol{x},t)$ takes the form:
\begin{small}
\begin{align}
    \boxed{r(\boldsymbol{x},t)=\mathbb{E}\left[\exp \left(-\int_t^Tc(\boldsymbol{X}_s,s)ds\right)\Phi(\boldsymbol{X}_T)\bigg|\boldsymbol{X}_t=\boldsymbol{x}\right]}\tag{Feynman-Kac Formula}
\end{align}
\end{small}
which is exactly the (\ref{eq:feynman-kac-formula}). \hfill $\square$

A special case of the Feynman-Kac formula when there is no running cost ($c\equiv 0$) is the \boldtext{Kolmogorov backward equation}, which describes the expected future reward as depending only on the expected value over terminal states. 

\begin{corollary}[Kolmogorov Backward Equation]\label{corollary:kolmogorov-backward}
    Let $\Phi:\mathbb{R}^d\to \mathbb{R}$ be a terminal constraint function and define $r(\boldsymbol{x},t):\mathbb{R}^d\times[0,T]\to \mathbb{R}^d$ as the expected future reward from a intermediate state $\boldsymbol{X}_t=\boldsymbol{x}$:
    \begin{small}
    \begin{align}
    r(\boldsymbol{x},t):=\mathbb{E}_{\boldsymbol{X}_{t:T}\sim \mathbb{Q}}\left[\Phi(\boldsymbol{X}_T)|\boldsymbol{X}_t=\boldsymbol{x}\right]\tag{Feynman-Kac Representation}\label{eq:feynman-kac-kolmogorov-bwd}
    \end{align}
    \end{small}
    where $\mathbb{Q}$ is the reference path measure defined by the SDE $d\boldsymbol{X}_t=\boldsymbol{f}(\boldsymbol{X}_t,t)ds+\sigma_t d\boldsymbol{B}_t$. Then $r(\boldsymbol{x},t)$ satisfies the \textbf{Kolmogorov backward equation} defined as:
    \begin{small}
    \begin{align}
        \partial_t r(\boldsymbol{x},t)+\langle\boldsymbol{f}(\boldsymbol{x},t), \nabla r(\boldsymbol{x},t)\rangle+\frac{\sigma_t^2}{2}\Delta r(\boldsymbol{x},t)=0, \quad r(\boldsymbol{x},T)=\Phi(\boldsymbol{x})\tag{Kolmogorov Backward Equation}\label{eq:kolmogorov-bwd-equation}
    \end{align}
    \end{small}
\end{corollary}

Together, the (\ref{eq:fokker-planck equation}) and (\ref{eq:feynman-kac-formula}) describe the two complementary ways in which stochastic differential equations evolve in time. The \textbf{Fokker–Planck equation} characterizes the forward evolution of probability densities $p_t$, describing how the distribution of particles transported by an SDE spreads and flows through state space. 

In contrast, the \textbf{Feynman-Kac formula} provides a backward evolution of functions along stochastic trajectories, expressing solutions to certain partial differential equations as expectations over future paths of the stochastic process. In this sense, the Fokker-Planck equation propagates probability mass forward in time, while the Feynman-Kac equation propagates value functions or potentials backward through the stochastic dynamics. This forward–backward structure will play a central role in the theory of Schrödinger bridges, where the optimal dynamics are characterized by the interaction between forward-evolving densities and backward-evolving potentials that together determine the controlled stochastic process connecting the prescribed endpoint distributions.

\subsection{Girsanov's Theorem}
\label{subsec:girsanov}
Now that we have established \textbf{path measures} and how they evolve both on an individual trajectory level via stochastic differential equations and on a density level via the Fokker-Planck partial differential equation, we will shift our focus to describing the \textbf{relationship between path measures}. This is the \textbf{core objective} of the Schrödinger bridge problem, which is to define a new path measure that minimally diverges from a reference path measure with marginal constraints.

We begin our discussion with \boldtext{Girsanov's theorem}, which characterizes the probability density ratio of a single stochastic process under different path measures. In this section, we derive Girsanov’s theorem in the special case where the reference path measure is standard Brownian motion with identity diffusion matrix. The key idea is to understand how a change of drift modifies the underlying path measure. To understand the intuition, we build the result from scratch using the discrete-time structure of Brownian motion\footnote{This follows closely from the derivation presented in \citet{sarkka2019applied}.}. For notational simplicity, we will consider a constant diffusion coefficient $\sigma$ over time, and omit the time subscript. 

Recall from Definition \ref{def:brownian-motion} that Brownian motion can be written as a sequence of \textbf{independent Gaussian increments} which can be written in discrete time with step size $\Delta t:=t_{k+1}-t_k$ as:
\begin{align}
    \sigma\mathbb{B}: \quad \boldsymbol{B}_{t_{k+1}}=\boldsymbol{B}_{t_k}+\sigma\sqrt{\Delta t}\boldsymbol{z}=:\boldsymbol{B}_{t_k}+ \sigma\Delta \boldsymbol{B}_{t_k}, \quad \boldsymbol{z}\sim \mathcal{N}(\boldsymbol{0}, \boldsymbol{I}_d)\label{eq:discrete-brownian}
\end{align}
where $\sigma$ is the diffusion coefficient and $\sigma\mathbb{B}$ to denotes the pure Brownian motion process with diffusion coefficient $\sigma$. Then, we can write the Brownian increment $\Delta \boldsymbol{B}_{t_k}$ as being sampled from a $d$-dimensional Gaussian with covariance $\Delta t\boldsymbol{I}_d$, denoted $\Delta\boldsymbol{B}_{t_k}\sim \mathcal{N}(\boldsymbol{0}, \Delta t\boldsymbol{I}_d)$. The \textbf{transition probability} under $\sigma\mathbb{B}$ is a Markov transition density $p(\boldsymbol{B}_{t+\Delta t}|\boldsymbol{B}_{0:t})=p(\boldsymbol{B}_{t+\Delta t}|\boldsymbol{B}_t)$ which is a $d$-dimensional Gaussian centered at $\boldsymbol{B}_{t_k}$ with covariance $\sigma^2\Delta t\boldsymbol{I}_d$ given by:
\begin{align}
    \sigma\mathbb{B}(\boldsymbol{B}_{t_{k+1}}|\boldsymbol{B}_{t_k})&=\mathcal{N}(\boldsymbol{B}_{t_{k+1}}|\boldsymbol{B}_{t_k})=\frac{1}{(2\pi\sigma^2\Delta t)^{d/2}}\exp\left(-\frac{\|\boldsymbol{B}_{t_{k+1}}-\boldsymbol{B}_{t_k}\|^2}{2\sigma^2\Delta t}\right)
\end{align}
Then, we can write the joint probability of the discrete-time Brownian process $(\boldsymbol{B}_{t_k})_{k\in \{1, \dots, K\}}=\boldsymbol{B}_{t_1}, \dots, \boldsymbol{B}_{t_n}$ where $t_{k+1}=t_k+\Delta t$ as:
\begin{small}
\begin{align}
    \sigma\mathbb{B}(\boldsymbol{B}_{t_1}, \dots , \boldsymbol{B}_{t_K})&=\prod_{k=1}^K \sigma\mathbb{B}(\boldsymbol{B}_{t_{k+1}}|\boldsymbol{B}_{t_k})=\prod_{k=1}^K \left[\frac{1}{(2\pi\sigma^2\Delta t)^{d/2}}\exp\left(-\frac{\|\boldsymbol{B}_{t_{k+1}}-\boldsymbol{B}_{t_k}\|^2}{2\sigma^2\Delta t}\right)\right]\nonumber\\
    &=\prod_{k=1}^K \left[\frac{1}{(2\pi\sigma^2\Delta t)^{d/2}}\exp\left(-\frac{\sigma^2\|\Delta \boldsymbol{B}_{t_k}\|^2}{2\sigma^2\Delta t}\right)\right]\nonumber\\
    &=\prod_{k=1}^K \left[\frac{1}{(2\pi\sigma^2\Delta t)^{d/2}}\exp\left(-\frac{\|\Delta \boldsymbol{B}_{t_k}\|^2}{2\Delta t}\right)\right]
\end{align}
\end{small}
Now, we consider the Itô process with a control drift $\boldsymbol{u}(\boldsymbol{X}_t,t)$ that evolves via the SDE given by (\ref{eq:ito-sde}) which evolves in discrete increments defined by:
\begin{align}
    \mathbb{P}: \quad \boldsymbol{X}_{t_{k+1}}=\boldsymbol{X}_{t_k}+\sigma \boldsymbol{u}(\boldsymbol{X}_{t_k},t_k)\Delta t+\sigma\Delta \boldsymbol{B}_{t_k}
\end{align}
which is the discrete Brownian process in (\ref{eq:discrete-brownian}) with an extra drift term that generates the path measure $\mathbb{P}$. Since the drift depends only on the current state $\boldsymbol{X}_t$, the transition probability is also a Markov transition density $\mathbb{P}(\boldsymbol{X}_{t_{k+1}}|(\boldsymbol{X}_{t_\ell})_{\ell\in \{1, \dots,k\}})=\mathbb{P}(\boldsymbol{X}_{t_{k+1}}|\boldsymbol{X}_{t_k})$ and can be written as a Gaussian centered at $\boldsymbol{X}_{t_k}+\sigma \boldsymbol{u}(\boldsymbol{X}_{t_k},t_k)\Delta t$ with covariance $\sigma^2\Delta t\boldsymbol{I}_d$ given by:
\begin{align}
    \mathbb{P}(\boldsymbol{X}_{t_{k+1}}|\boldsymbol{X}_{t_k})&=\mathcal{N}(\boldsymbol{X}_{t_{k+1}}|\boldsymbol{X}_{t_k}+\sigma\boldsymbol{u}(\boldsymbol{X}_{t_k},t_k)\Delta t)\nonumber\\
    &=\frac{1}{(2\pi\sigma^2\Delta t)^{d/2}}\exp\left(-\frac{\|\boldsymbol{X}_{t_{k+1}}-(\boldsymbol{X}_{t_k}+\sigma \boldsymbol{u}(\boldsymbol{X}_{t_k},t_k)\Delta t)\|^2}{2\sigma^2\Delta t}\right)
\end{align}
with joint probability density of the full discrete process $(\boldsymbol{X}_{t_k})_{k\in \{1, \dots, K\}}=\boldsymbol{X}_{t_1}, \dots, \boldsymbol{X}_{t_K}$ given by:
\begin{small}
\begin{align}
    \mathbb{P}(\boldsymbol{X}_{t_1}, \dots, \boldsymbol{X}_{t_K})&=\prod_{k=1}^K \mathbb{P}(\boldsymbol{X}_{t_{k+1}}|\boldsymbol{X}_{t_k})\nonumber\\
    &=\prod_{k=1}^K \left[\frac{1}{(2\pi\sigma^2\Delta t)^{d/2}}\exp\left(-\frac{\|\boldsymbol{X}_{t_{k+1}}-(\boldsymbol{X}_{t_k}+\sigma \boldsymbol{u}(\boldsymbol{X}_{t_k},t_k)\Delta t)\|^2}{2\sigma^2\Delta t}\right)\right]\nonumber\\
    &=\prod_{k=1}^K \left[\frac{1}{(2\pi\sigma^2\Delta t)^{d/2}}\exp\left(-\frac{\|\boldsymbol{X}_{t_{k+1}}-\boldsymbol{X}_{t_k}-\sigma \boldsymbol{u}(\boldsymbol{X}_{t_k},t_k)\Delta t\|^2}{2\sigma^2\Delta t}\right)\right]
\end{align}
\end{small}
Now, suppose we have a path $(\boldsymbol{X}_{t_k})_{k\in \{1, \dots, K\}}$ generated \textit{under} the Brownian process $\sigma\mathbb{B}$ such that $\boldsymbol{X}_{t_{k+1}}=\boldsymbol{X}_{t_k}+\sigma\Delta \boldsymbol{B}_{t_k}$. Then, we can define the ratio of the path probability under $\mathbb{P}$ with respect to $\sigma\mathbb{B}$ as:
\begin{small}
\begin{align}
    \frac{\mathrm{d}\mathbb{P}}{\mathrm{d}\sigma\mathbb{B}}(\boldsymbol{X}_{t_k})_{k\in \{1, \dots, K\}}&=\frac{\mathbb{P}(\boldsymbol{X}_{t_1}, \dots, \boldsymbol{X}_{t_K})}{\sigma\mathbb{B}(\boldsymbol{X}_{t_1}, \dots , \boldsymbol{X}_{t_K})}\nonumber\\
    &=\frac{\prod_{k=1}^K \left[\frac{1}{(2\pi\sigma^2\Delta t)^{d/2}}\exp\left(-\frac{\|\bluetext{\boldsymbol{X}_{t_{k+1}}-\boldsymbol{X}_{t_k}}-\sigma \boldsymbol{u}(\boldsymbol{X}_{t_k},t_k)\Delta t\|^2}{2\sigma^2\Delta t}\right)\right]}{\prod_{k=1}^K \left[\frac{1}{(2\pi\sigma^2\Delta t)^{d/2}}\exp\left(-\frac{\|\Delta \boldsymbol{B}_{t_k}\|^2}{2\Delta t}\right)\right]}\nonumber\\
    &=\prod_{k=1}^K\exp\left(-\frac{\|\bluetext{\sigma\Delta \boldsymbol{B}_{t_k}}-\sigma \boldsymbol{u}(\boldsymbol{X}_{t_k},t_k)\Delta t\|^2}{2\sigma^2\Delta t}+\frac{\|\Delta \boldsymbol{B}_{t_k}\|^2}{2\Delta t}\right)\nonumber\\
    &=\prod_{k=1}^K\exp\left(-\frac{\|\bluetext{\Delta \boldsymbol{B}_{t_k}}- \boldsymbol{u}(\boldsymbol{X}_{t_k},t_k)\Delta t\|^2}{2\Delta t}+\frac{\|\Delta \boldsymbol{B}_{t_k}\|^2}{2\Delta t}\right)
\end{align}
\end{small}
where the second-last equality follows from the fact that the increment $\boldsymbol{X}_{t_k}\mapsto \boldsymbol{X}_{t_{k+1}}$ was made under the reference process such that $\boldsymbol{X}_{t_{k+1}}=\boldsymbol{X}_{t_k}+\sigma\Delta \boldsymbol{B}_{t_k}$, which allowed us to factor out the $\sigma^2$ and cancel with the denominator. Expanding $\|\Delta \boldsymbol{B}_{t_k}- \boldsymbol{u}(\boldsymbol{X}_{t_k},t_k)\Delta t\|^2=\|\Delta \boldsymbol{B}_{t_k}\|^2-2 \boldsymbol{u}(\boldsymbol{X}_{t_k}, t_k)^\top \Delta \boldsymbol{B}_{t_k}\Delta t+\|\boldsymbol{u}(\boldsymbol{X}_{t_k}, t_k)\|^2\Delta t^2$ gives us:
\begin{small}
\begin{align}
    \frac{\mathrm{d}\mathbb{P}}{\mathrm{d}\sigma\mathbb{B}}(\boldsymbol{X}_{t_k})_{k\in \{1, \dots, K\}}&=\prod_{k=1}^K\exp\left(\frac{-\|\Delta \boldsymbol{B}_{t_k}\|^2+2\boldsymbol{u}(\boldsymbol{X}_{t_k}, t_k)^\top\Delta \boldsymbol{B}_{t_k}\Delta t-\|\boldsymbol{u}(\boldsymbol{X}_{t_k}, t_k)\|^2\Delta t^2+\|\Delta \boldsymbol{B}_{t_k}\|^2}{2\sigma^2\Delta t}\right)\nonumber\\
    &=\exp\left(\sum_{k=1}^K\frac{2\boldsymbol{u}(\boldsymbol{X}_{t_k}, t_k)^\top\Delta \boldsymbol{B}_{t_k}\Delta t-\|\boldsymbol{u}(\boldsymbol{X}_{t_k}, t_k)\|^2\Delta t^2}{2\Delta t}\right)\nonumber\\
    &=\exp\left(\sum_{k=1}^K\boldsymbol{u}(\boldsymbol{X}_{t_k}, t_k)\Delta \boldsymbol{B}_{t_k}-\frac{1}{2}\sum_{k=1}^K\|\boldsymbol{u}(\boldsymbol{X}_{t_k}, t_k)\|^2\Delta t\right)
\end{align}
\end{small}
which depends only on the Brownian increments $\Delta\boldsymbol{B}_{t_k}$ and the drift $\boldsymbol{u}(\boldsymbol{X}_{t_k},t_k)$. Taking the continuous time limit as $\Delta t\to 0$, we have $\sum_k\boldsymbol{u}_{t_k}\Delta \boldsymbol{B}_{t_k}\to \int_0^T\boldsymbol{u}_td\boldsymbol{B}_t$ and $\sum_k\|\boldsymbol{u}_{t_k}\|^2\Delta t\to \int_0^T\|\boldsymbol{u}_t\|^2dt$, resulting in the following form of ratio between path measures:
\begin{align}
    \frac{\mathrm{d}\mathbb{P}}{\mathrm{d}\sigma\mathbb{B}}(\boldsymbol{X}_{0:T})=\exp\left(-\frac{1}{2}\int_0^T\|\boldsymbol{u}(\boldsymbol{X}_t, t)\|^2dt+\int_0^T\boldsymbol{u}(\boldsymbol{X}_t, t)d\boldsymbol{B}_t\right)
\end{align}

This ratio $\frac{\mathrm{d}\mathbb{P}}{\mathrm{d}\sigma\mathbb{B}}(\boldsymbol{X}_{0:T})$ is also the \boldtext{Radon-Nikodym derivative (RND)} between $\mathbb{P}$ and $\sigma\mathbb{B}$, which can be used to transform the likelihood of a path under $\sigma\mathbb{B}$ to its likelihood under $\mathbb{P}$. From this intuition, we can state the formal version of \boldtext{Girsanov's theorem} \citep{girsanov1960transforming, karatzas2014brownian, oksendal2003stochastic}.

\begin{theorem}[Girsanov's Theorem]\label{theorem:girsanov}
    Consider the $d$-dimensional Brownian motion $(\boldsymbol{B}_t)_{t\in [0,T]}$ adapted to the filtration $(\mathcal{F}_t)_{t\in [0,T]}$. Given two path measures $\mathbb{P}, \mathbb{P}'\in \mathcal{P}(\mathbb{R}^d)$, where $\mathbb{P}'\ll\mathbb{P}$, define the density process $(\boldsymbol{Z}_t)_{t\in [0,T]}$ of the ratio:
    \begin{align}
        \boldsymbol{Z}_t:= \mathbb{E}_{\mathbb{P}}\left[\frac{\mathrm{d}\mathbb{P}'}{\mathrm{d}\mathbb{P}}\bigg|\mathcal{F}_t\right]
    \end{align}
    which is the likelihood ratio given the information up to time $t$. Then, there exists a predictable process $(\boldsymbol{\theta}_s)_{s\in[0,T]}$ such that:
    \begin{small}
    \begin{align}
        \boldsymbol{Z}_t=\exp\left(-\frac{1}{2}\int_0^t\|\boldsymbol{\theta}_s\|^2ds+\int_0^t\boldsymbol{\theta}_s^\top d\boldsymbol{B}_s\right), \quad t\in [0,T]\label{eq:girsanov-proof1}
    \end{align}
    \end{small}
    Furthermore, the stochastic process $(\boldsymbol{B}'_t)_{t\in [0,T]}$ defined as:
    \begin{small}
    \begin{align}
        \boldsymbol{B}'_t:=\boldsymbol{B}_t-\int_0^t\boldsymbol{\theta}_sds\tag{Girsanov's Theorem}\label{eq:girsanov-proof2}
    \end{align}
    \end{small}
    is a standard Brownian motion under $\mathbb{P}'$, such that for any discrete time intervals $0=t_0\leq \dots \leq t_k\leq \dots \leq t_K\leq T$ the increments $\Delta \boldsymbol{B}'_{t_k}:=\boldsymbol{B}'_{t_{k+1}}-\boldsymbol{B}'_{t_k}\sim \mathcal{N}(\boldsymbol{0}, (t_{k+1}-t_k)\boldsymbol{I}_d)$ and independent. 
\end{theorem}

Intuitively, Girsanov's theorem provides a way to change the probability measure on path space while leaving the underlying trajectories unchanged. By defining $\boldsymbol{\theta}_t$ such that the drift of the original path probability cancels out after applying the transformation, Girsanov's theorem states that the $\boldsymbol{\theta}_t$-tilted process becomes a standard Brownian motion under the transformed probability measure. To ensure that the resulting path measure obtained from reweighting with Girsanov's theorem is valid, we establish \boldtext{Novikov’s condition}, which states an integrability condition that ensures that the exponential density process is a \textbf{true martingale} satisfying (\ref{eq:martingale-property}) that has constant expectation and thus defines a valid change of measure.

\begin{lemma}[Novikov’s Condition]\label{lemma:novikov-condition}
    Let $(\boldsymbol{B}_t)_{t\in [0,T]}$ be a $d$-dimensional Brownian motion adapted to a filtration $(\mathcal{F}_t)_{t\in [0,T]}$ and let $(\boldsymbol{\theta}_t)_{t\in [0,T]}$ be a $\mathcal{F}_t$-adapted process that is square integrable over every finite time interval, such that  $\int_0^T\|\boldsymbol{\theta}_t\|^2dt< \infty$ almost surely. Then, if the exponential process given by:
    \begin{small}
    \begin{align}
        \boldsymbol{Z}_t=\exp\left(-\frac{1}{2}\int_0^t\|\boldsymbol{\theta}_s\|^2ds+\int_0^t\boldsymbol{\theta}_s^\top d\boldsymbol{B}_s\right) 
    \end{align}
    \end{small}
    satisfies the \textbf{Novikov's condition}:
    \begin{small}
    \begin{align}
        \mathbb{E}\left[\exp\left(\frac{1}{2}\int_0^t\|\boldsymbol{\theta}_s\|^2ds\right) \right]< \infty\tag{Novikov's condition}\label{eq:novikov-condition}
    \end{align}
    \end{small}
    Then, $(\boldsymbol{Z}_t)_{t\in [0,T]}$ is a true martingale satisfying (\ref{eq:martingale-property}) with constant expectation $\mathbb{E}[Z_t]=1$ for $t\in [0,T]$.
\end{lemma}

The purpose of \textbf{Novikov’s Condition} is to ensure $\boldsymbol{Z}_t$ does not lose mass and is properly normalized. In other words, it ensures that the exponential tilt used in Girsanov’s theorem can serve as a \textbf{valid likelihood ratio between path measures} and defines a \textbf{proper change of probability measure on path space}. Leveraging Girsanov's theorem and Novikov’s Condition, we can now define changes in measure and KL divergences in path space as local drift changes in their stochastic differential equations. 

\subsection{Path Measure Radon-Nikodym Derivative and KL Divergence}
\label{subsec:path-measure-rnd-kl}
To understand the SB problem, it is crucial to understand the measurement used to quantify \textbf{distance} between two path measures $\mathbb{Q}, \mathbb{P}\in  \mathcal{P}(C([0,T]; \mathbb{R}^d))$ known as the \boldtext{Kullback-Leibler (KL) divergence} (also referred to as the \textbf{relative entropy} or \textbf{I-divergence}), denoted $\text{KL}(\mathbb{P}\|\mathbb{Q})$ \citep{csiszar1975divergence}. 

\begin{definition}[KL Divergence Between Path Measures]\label{def:kl-path-measure}
    Given two probability measures over the space of paths $\mathbb{P}, \mathbb{P}'\in \mathcal{P}(C([0,T]; \mathbb{R}^d))$, the KL divergence is given by
    \begin{align}
        \text{KL}(\mathbb{P}'\|\mathbb{P})=\mathbb{E}_{\mathbb{P}'}\left[\log \frac{\mathrm{d}\mathbb{P}'}{\mathrm{d}\mathbb{P}}\right]=\int_{C([0,T]; \mathbb{R}^d)}\log \frac{\mathrm{d}\mathbb{P}'}{\mathrm{d}\mathbb{P}}(\boldsymbol{X}_{0:T})\mathrm{d}\mathbb{P}'(\boldsymbol{X}_{0:T})
    \end{align}
    where $\boldsymbol{X}_{0:T}$ is a stochastic process and $\frac{\mathrm{d}\mathbb{P}'}{\mathrm{d}\mathbb{P}}$ is the Radon-Nikodym derivative between $\mathbb{P}'$ and $\mathbb{P}$.
\end{definition}

Since we have defined \textbf{path measures} as solutions to SDEs in Section \ref{subsec:path-measure-ito-processes}, we can now introduce how to write the KL divergence in integral form. First, we derive the \boldtext{Radon-Nikodym (RN) derivative} with respect to the \textbf{drift} and \textbf{diffusion} of two Itô processes.

\begin{figure}
    \centering
    \includegraphics[width=\linewidth]{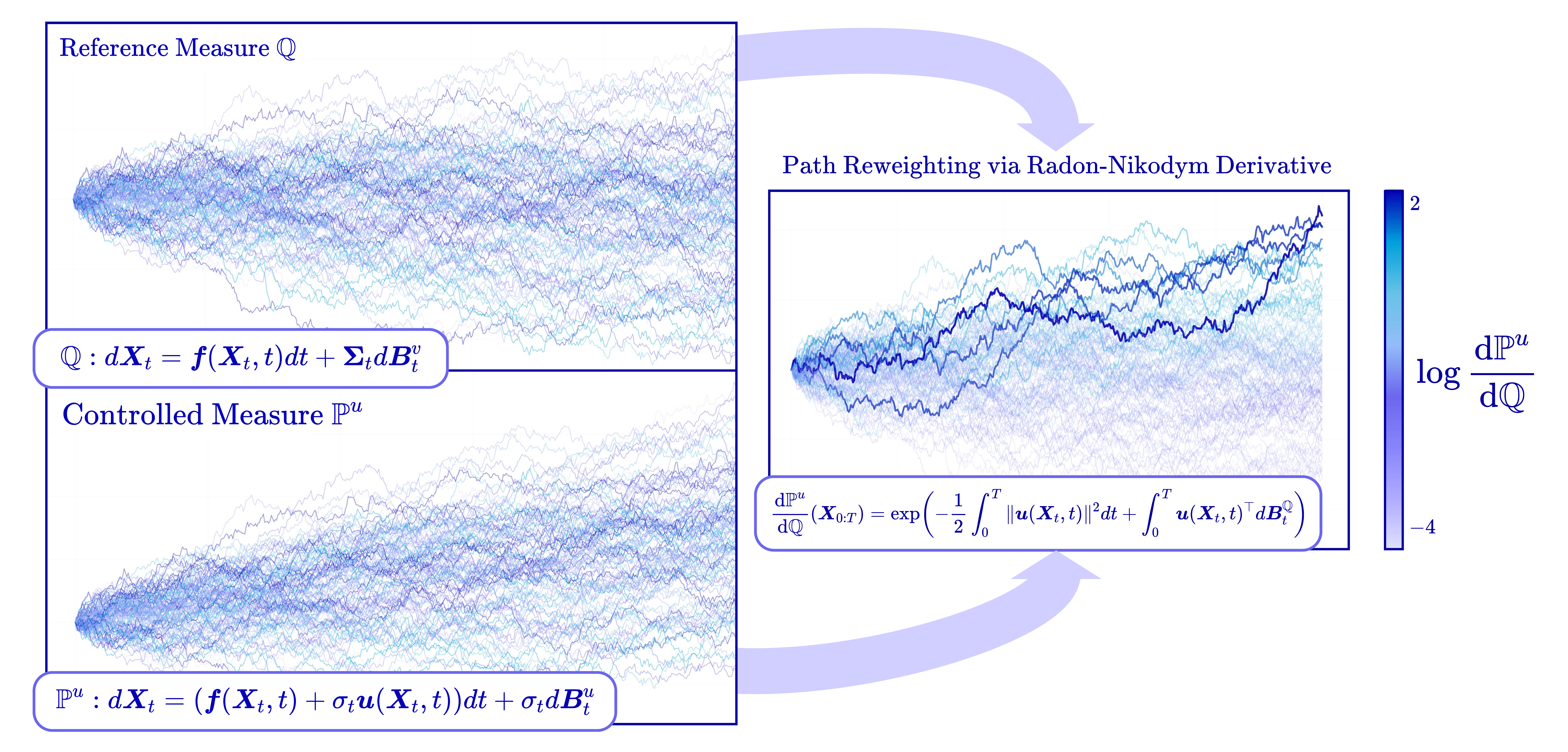}
    \caption{\textbf{Change of Measure Using Radon-Nikodym Derivative.} Sample trajectories of a diffusion process are shown under two probability measures. \textbf{Top left:} trajectories under the reference measure $\mathbb{Q}$, governed by the uncontrolled reference drift $\boldsymbol{f}(\boldsymbol{X}_t,t)$. \textbf{Bottom left:} trajectories under the controlled measure $\mathbb{P}^u$, where the dynamics include an additional control drift $\boldsymbol{u}(\boldsymbol{X}_t,t)$. \textbf{Right:} the same reference trajectories are reweighted according to the Radon–Nikodym derivative $\frac{\mathrm{d}\mathbb{P}^u}{\mathrm{d}\mathbb{Q}}$, which assigns higher likelihood to paths aligned with the control and lower likelihood to paths that oppose it. The color intensity represents the log-likelihood ratio $\log \frac{\mathrm{d}\mathbb{P}^u}{\mathrm{d}\mathbb{Q}}$, illustrating how the controlled dynamics can be interpreted as a change of measure on the same underlying path space.}
    \label{fig:girsanov}
\end{figure}

\begin{theorem}[Radon-Nikodym Derivative of Controlled Itô Path Measures \citep{sarkka2019applied}]\label{theorem:rnd-ito-process}
Consider two controlled Itô path measures $\mathbb{P}^u$ and $\mathbb{P}^{\tilde{u}}$ with control drifts $\boldsymbol{u}$ and $\tilde{\boldsymbol{u}}$, where $\mathbb{P}^{\tilde{u}}\ll \mathbb{P}^u$\footnote{$\mathbb{P}^{\tilde{u}}\ll\mathbb{P}^u$ means that $\mathbb{P}^{\tilde{u}}$ is absolutely continuous with respect to $\mathbb{P}^u$ and any event with zero probability under $\mathbb{P}^u$ also has zero probability under $\mathbb{P}^{\tilde{u}}$}. Assuming the same  reference drift $\boldsymbol{f}(\boldsymbol{x},t)$ and diffusion coefficient $\sigma_t$, they are defined by the SDEs: 
\begin{align}
    \mathbb{P}^u:\quad d\boldsymbol{X}_t&=(\boldsymbol{f}(\boldsymbol{X}_t,t)+\sigma_t\boldsymbol{u}(\boldsymbol{X}_t,t)) dt+\sigma_t d\boldsymbol{B}^u_t, \quad \boldsymbol{X}_0=\boldsymbol{x}\tag{$\boldsymbol{u}$-SDE}\label{eq:rnd-proof-sde1}\\
    \mathbb{P}^{\tilde{u}}:\quad d\tilde{\boldsymbol{X}}_t &=(\boldsymbol{f}(\boldsymbol{X}_t,t)+\sigma_t\tilde{\boldsymbol{u}}(\tilde{\boldsymbol{X}}_t, t))dt+ \sigma_t d\boldsymbol{B}^{\tilde{u}}_t, \quad \tilde{\boldsymbol{X}}_0=\boldsymbol{x}\tag{$\tilde{\boldsymbol{u}}$-SDE}\label{eq:rnd-proof-sde2}
\end{align}
The Radon-Nikodym derivative of the corresponding path measures $\mathbb{P}^u$ and $\mathbb{P}^{\tilde{u}}$ is given by
\begin{small}
\begin{align}
    \frac{\mathrm{d}\mathbb{P}^{\tilde{u}}}{\mathrm{d}\mathbb{P}^u}(\boldsymbol{X}^u_{0:T})&=\exp\bigg(-\frac{1}{2}\int_0^T\|(\tilde{\boldsymbol{u}}-\boldsymbol{u})(\boldsymbol{X}^u_t,t)\|^2dt+\int_0^T\big(\tilde{\boldsymbol{u}}-\boldsymbol{u}\big)(\boldsymbol{X}^u_t,t)^\top d\boldsymbol{B}^u_t\bigg)\tag{Path RND}\label{eq:theorem-path-rnd}
\end{align}
\end{small}
where $\boldsymbol{X}^u_{0:T}$ denotes a stochastic process generated under $\mathbb{P}^u$ and $\boldsymbol{B}^u_t$ denotes the Brownian motion under $\mathbb{P}^u$.
\end{theorem}

\textit{Proof.} The \textbf{key idea} of the Radon-Nikodym derivative is to define how likely a stochastic process $\boldsymbol{X}^u_{0:T}:=(\boldsymbol{X}^u_t)_{t\in [0,T]}$ defined as a standard Brownian motion under $\mathbb{P}^u$ is under the path measure $\mathbb{P}^{\tilde{u}}$. Since a Brownian motion under $\mathbb{P}^u$ and $\mathbb{P}^{\tilde{u}}$ differ only in their control drifts, we define the change in the drifts as:
\begin{align}
    \boldsymbol{\theta}_t:=(\tilde{\boldsymbol{u}}-\boldsymbol{u})(\boldsymbol{X}^u_t,t)\in \mathbb{R}^d
\end{align}
Following (\ref{eq:girsanov-proof2}), we can define an density process $\boldsymbol{Z}_t$ as:
\begin{small}
\begin{align}
    \boldsymbol{Z}_t:=\exp\left(-\frac{1}{2}\int_0^t\|\boldsymbol{\theta}_s\|^2ds+\int_0^t\boldsymbol{\theta}_s^\top d\boldsymbol{B}_s\right)\quad \text{s.t.} \quad \mathbb{E}_{\mathbb{P}^u}\left[\boldsymbol{Z}_T\right]=1
\end{align}
\end{small}
where the condition $\mathbb{E}_{\mathbb{P}^u}\left[\boldsymbol{Z}_T\right]=1$ follows from the assumption that $\boldsymbol{Z}_t$ satisfies (\ref{eq:novikov-condition}) and is a true martingale. From (\ref{eq:girsanov-proof2}), we know that $\boldsymbol{Z}_T$ defines the density ratio of a new measure $\mathbb{P}'$ with respect to $\mathbb{P}^u$ given by:
\begin{align}
    \boldsymbol{Z}_t:= \mathbb{E}_{\mathbb{P}^u}\left[\frac{\mathrm{d}\mathbb{P}'}{\mathrm{d}\mathbb{P}^u}\bigg|\mathcal{F}_t\right]
\end{align}
Then, according to (\ref{eq:girsanov-proof2}), we have that the Brownian motion $\boldsymbol{B}'$ under the new measure $\mathbb{P}'$ is a standard Brownian motion defined by:
\begin{align}
    \boldsymbol{B}'_t=\boldsymbol{B}^u_t-\int_0^t\boldsymbol{\theta}_sds\implies d\boldsymbol{B}^u_t=d\boldsymbol{B}'_t+\boldsymbol{\theta}_tdt
\end{align}
Now, we can rewrite (\ref{eq:rnd-proof-sde1}) under the Brownian motion $\boldsymbol{B}'$ as:
\begin{small}
\begin{align}
    d\boldsymbol{X}^u_t&=(\boldsymbol{f}(\boldsymbol{X}^u_t,t)+\sigma_t\boldsymbol{u}(\boldsymbol{X}^u_t,t)) dt+\sigma_t \bluetext{d\boldsymbol{B}^u_t}\nonumber\\
    &=(\boldsymbol{f}(\boldsymbol{X}^u_t,t)+\sigma_t\boldsymbol{u}(\boldsymbol{X}^u_t,t)) dt+\sigma_t \bluetext{(d\boldsymbol{B}'_t+\boldsymbol{\theta}_tdt)}\nonumber\\
    &=(\boldsymbol{f}(\boldsymbol{X}^u_t,t)+\sigma_t(\boldsymbol{u}(\boldsymbol{X}^u_t,t)\bluetext{+\boldsymbol{\theta}_t}) dt+\sigma_t \bluetext{d\boldsymbol{B}'_t}
\end{align}
\end{small}
Substituting the definition $\boldsymbol{\theta}_t:=(\tilde{\boldsymbol{u}}-\boldsymbol{u})(\boldsymbol{X}^u_t,t)$, we have:
\begin{small}
\begin{align}
    d\boldsymbol{X}^u_t&=(\boldsymbol{f}(\boldsymbol{X}^u_t,t)+\boldsymbol{\Sigma}_t(\boldsymbol{u}(\boldsymbol{X}^u_t,t)\bluetext{+\tilde{\boldsymbol{u}}(\boldsymbol{X}^u_t,t)-\boldsymbol{u}(\boldsymbol{X}^u_t,t)}) dt+\boldsymbol{\Sigma}_t \bluetext{d\boldsymbol{B}^{\tilde{u}}_t}\nonumber\\
    &=(\boldsymbol{f}(\boldsymbol{X}^u_t,t)+\boldsymbol{\Sigma}_t\bluetext{\tilde{\boldsymbol{u}}(\boldsymbol{X}^u_t,t)}) dt+\boldsymbol{\Sigma}_t \bluetext{d\boldsymbol{B}^{\tilde{u}}_t}
\end{align}
\end{small}
which exactly matches (\ref{eq:rnd-proof-sde2}), so we conclude that $\mathbb{P}' =\mathbb{P}^{\tilde{u}}$ and $\boldsymbol{B}' =\boldsymbol{B}^{\tilde{u}}$. Then, we have that $\frac{\mathrm{d}\mathbb{P}^{\tilde{u}}}{\mathrm{d}\mathbb{P}^u}$ is equal to $\boldsymbol{Z}_T$ given by:
\begin{small}
\begin{align}
    \frac{\mathrm{d}\mathbb{P}^{\tilde{u}}}{\mathrm{d}\mathbb{P}^u}(\boldsymbol{X}^u_{0:T})&=\boldsymbol{Z}_T=\exp\left(-\frac{1}{2}\int_0^t\|\bluetext{\boldsymbol{\theta}_s}\|^2ds+\int_0^t\bluetext{\boldsymbol{\theta}_s^\top} d\boldsymbol{B}^u_s\right)\nonumber\\
    &=\exp\left(-\frac{1}{2}\int_0^t\|\bluetext{(\tilde{\boldsymbol{u}}-\boldsymbol{u})(\boldsymbol{X}^u_t,t)}\|^2ds+\int_0^t\bluetext{(\tilde{\boldsymbol{u}}-\boldsymbol{u})(\boldsymbol{X}^u_t,t)^\top} d\boldsymbol{B}^u_s\right)
\end{align}
\end{small}
which is exactly the Radon-Nikodym derivative that reweights stochastic processes under $\mathbb{P}^u$ generated with (\ref{eq:rnd-proof-sde1}) with their probabilities under $\mathbb{P}^{\tilde{u}}$ defined by (\ref{eq:rnd-proof-sde2}).\hfill $\square$

From Definition \ref{def:kl-path-measure}, we defined the KL divergence between path measures as the expectation of the logarithm of the Radon-Nikodym derivative between two Itô processes. Using Theorem \ref{theorem:rnd-ito-process}, we can define the corresponding integral form for the KL divergence.

\begin{corollary}[KL Divergence of Itô Path Measures]\label{corollary:kl-divergence-ito}
    Consider two controlled path measures $\mathbb{P}^{\tilde{u}}$ and $\mathbb{P}^u$, where $\mathbb{P}^{\tilde{u}}\ll\mathbb{P}^u$, with control drifts $\boldsymbol{u}$ and $\tilde{\boldsymbol{u}}$ and SDEs (\ref{eq:rnd-proof-sde1}) and (\ref{eq:rnd-proof-sde2}). The \textbf{KL divergence} of $\mathbb{P}^{\tilde{u}}$ with respect to $\mathbb{P}^u$ is given by:
    \begin{small}
    \begin{align}
        \text{KL}(\mathbb{P}^{\tilde{u}}\|\mathbb{P}^u)&=\mathbb{E}_{\boldsymbol{X}^{\tilde{u}}_{0:T}\sim\mathbb{P}^{\tilde{u}}}\left[\log \frac{\mathrm{d}\mathbb{P}^{\tilde{u}}}{\mathrm{d}\mathbb{P}^u}(\boldsymbol{X}^{\tilde{u}}_{0:T})\right]\nonumber\\
        &=\mathbb{E}_{\boldsymbol{X}^{\tilde{u}}_{0:T}\sim\mathbb{P}^{\tilde{u}}}\left[\frac{1}{2}\int_0^t\|\tilde{\boldsymbol{u}}(\boldsymbol{X}^{\tilde{u}}_t,t)-\boldsymbol{u}(\boldsymbol{X}^{\tilde{u}}_t,t)\|^2ds\right]\tag{Path KL Divergence}\label{eq:path-kl-divergence}
    \end{align}
    \end{small}
    where $\boldsymbol{X}_{0:T}^{\tilde{u}}$ denotes stochastic trajectories sampled under the law of $\mathbb{P}^{\tilde{u}}$. When $\mathbb{P}^u:=\mathbb{Q}$ is the \textbf{reference path measure} with zero control $\boldsymbol{u}\equiv 0$ with SDE $d\boldsymbol{X}_t=\boldsymbol{f}(\boldsymbol{X}_t,t)dt+\sigma_td\boldsymbol{B}_t$, then the KL divergence reduces to:
    \begin{small}
    \begin{align}
        \text{KL}(\mathbb{P}^{\tilde{u}}\|\mathbb{Q})&=\mathbb{E}_{\boldsymbol{X}^{\tilde{u}}_{0:T}\sim\mathbb{P}^{\tilde{u}}}\left[\log \frac{\mathrm{d}\mathbb{P}^{\tilde{u}}}{\mathrm{d}\mathbb{Q}}(\boldsymbol{X}^{\tilde{u}}_{0:T})\right]=\mathbb{E}_{\boldsymbol{X}^{\tilde{u}}_{0:T}\sim\mathbb{P}^{\tilde{u}}}\left[\frac{1}{2}\int_0^t\|\tilde{\boldsymbol{u}}(\boldsymbol{X}^{\tilde{u}}_t,t)\|^2ds\right]
    \end{align}
    \end{small}
\end{corollary}

\textit{Proof. }Starting from our definition of (\ref{eq:theorem-path-rnd}), we can expand the expression for path measure KL divergence from Definition \ref{def:kl-path-measure} as:
\begin{small}
\begin{align}
    \text{KL}(\mathbb{P}^{\tilde{u}}\|\mathbb{P}^u)&=\mathbb{E}_{\boldsymbol{X}^{\tilde{u}}_{0:T}\sim\mathbb{P}^{\tilde{u}}}\left[\log \bluetext{\frac{\mathrm{d}\mathbb{P}^{\tilde{u}}}{\mathrm{d}\mathbb{P}^u}(\boldsymbol{X}^{\tilde{u}}_{0:T})}\right]\nonumber\\
    &=\mathbb{E}_{\boldsymbol{X}^{\tilde{u}}_{0:T}\sim\mathbb{P}^{\tilde{u}}}\left[\bluetext{-\frac{1}{2}\int_0^t\|\bluetext{(\tilde{\boldsymbol{u}}-\boldsymbol{u})(\boldsymbol{X}^{\tilde{u}}_t,t)}\|^2ds+\int_0^t\bluetext{(\tilde{\boldsymbol{u}}-\boldsymbol{u})(\boldsymbol{X}^{\tilde{u}}_t,t)^\top} d\boldsymbol{B}^u_s}\right]\label{eq:kl-path-proof1}
\end{align}
\end{small}
where the superscript in $\boldsymbol{X}^{\tilde{u}}_{0:T}$ is used to explicitly denote that they are stochastic processes generated under $\mathbb{P}^{\tilde{u}}$. Recall that the Itô integral of a standard Brownian motion vanishes under the expectation over its associated path measure. However, since the expectation in (\ref{eq:kl-path-proof1}) is under $\mathbb{P}^{\tilde{u}}$ and the Brownian motion is under $\mathbb{P}^{u}$, we need to rewrite $d\boldsymbol{B}^u_s$ in terms of $d\boldsymbol{B}^{\tilde{u}}_s$ for the Itô integral to vanish. To do this, we apply (\ref{eq:girsanov-proof2}) which states that the Brownian motion $\boldsymbol{B}^{\tilde{u}}$ under the measure $\mathbb{P}^{\tilde{u}}$ is a standard Brownian motion defined by:
\begin{align}
    \boldsymbol{B}^{\tilde{u}}_t=\boldsymbol{B}_t^u-\int_0^t\boldsymbol{\theta}_sds\implies d\boldsymbol{B}_t^u=d\boldsymbol{B}^{\tilde{u}}_t+\boldsymbol{\theta}_tdt
\end{align}
where we defined $(\boldsymbol{\theta}_t)_{t\in [0,T]}$ as in Theorem \ref{theorem:rnd-ito-process} as the difference in control drifts $\boldsymbol{\theta}_t:=(\tilde{\boldsymbol{u}}-\boldsymbol{u})(\boldsymbol{X}^{\tilde{u}}_t,t)$ evaluated along controlled trajectories generated under $\mathbb{P}^{\tilde{u}}$. Therefore, we can rewrite (\ref{eq:kl-path-proof1}) as:
\begin{small}
\begin{align}
    \text{KL}(\mathbb{P}^{\tilde{u}}\|\mathbb{P}^u)&=\mathbb{E}_{\boldsymbol{X}^{\tilde{u}}_{0:T}\sim\mathbb{P}^{\tilde{u}}}\left[-\frac{1}{2}\int_0^t\|\bluetext{(\tilde{\boldsymbol{u}}-\boldsymbol{u})(\boldsymbol{X}^{\tilde{u}}_t,t)}\|^2ds+\int_0^t(\tilde{\boldsymbol{u}}-\boldsymbol{u})(\boldsymbol{X}^{\tilde{u}}_t,t)^\top\bluetext{(d\boldsymbol{B}^{\tilde{u}}_s+(\tilde{\boldsymbol{u}}-\boldsymbol{u})(\boldsymbol{X}^{\tilde{u}}_s,s)ds)}\right]\nonumber\\
    &=\mathbb{E}_{\boldsymbol{X}^{\tilde{u}}_{0:T}\sim\mathbb{P}^{\tilde{u}}}\bigg[-\frac{1}{2}\int_0^t\|\bluetext{(\tilde{\boldsymbol{u}}-\boldsymbol{u})(\boldsymbol{X}^{\tilde{u}}_t,t)}\|^2ds+\underbrace{\int_0^t(\tilde{\boldsymbol{u}}-\boldsymbol{u})(\boldsymbol{X}^{\tilde{u}}_t,t)^\top\bluetext{d\boldsymbol{B}^{\tilde{u}}_s}}_{\text{vanishes under expectation}}+\int_0^t\bluetext{\|(\tilde{\boldsymbol{u}}-\boldsymbol{u})(\boldsymbol{X}^{\tilde{u}}_t,t)\|^2}ds\bigg]\nonumber\\
    &=\boxed{\mathbb{E}_{\boldsymbol{X}^{\tilde{u}}_{0:T}\sim\mathbb{P}^{\tilde{u}}}\bigg[\frac{1}{2}\int_0^t\|\bluetext{(\tilde{\boldsymbol{u}}-\boldsymbol{u})(\boldsymbol{X}^{\tilde{u}}_t,t)}\|^2ds\bigg]}
\end{align}
\end{small}
which is the integral form of the path measure KL divergence. By simply substituting $\boldsymbol{u}\equiv 0$, we get the KL divergence for the case where $\mathbb{P}^u=\mathbb{Q}$ is the reference path measure no control drift:
\begin{small}
\begin{align}
    \text{KL}(\mathbb{P}^{\tilde{u}}\|\mathbb{Q})&=\mathbb{E}_{\boldsymbol{X}^{\tilde{u}}_{0:T}\sim\mathbb{P}^{\tilde{u}}}\left[\log \frac{\mathrm{d}\mathbb{P}^{\tilde{u}}}{\mathrm{d}\mathbb{Q}}(\boldsymbol{X}^{\tilde{u}}_{0:T})\right]=\boxed{\mathbb{E}_{\boldsymbol{X}^{\tilde{u}}_{0:T}\sim\mathbb{P}^{\tilde{u}}}\left[\frac{1}{2}\int_0^t\|\tilde{\boldsymbol{u}}(\boldsymbol{X}^{\tilde{u}}_t,t)\|^2ds\right]}
\end{align}
\end{small}
which is the form of the KL divergence that we will use to define the dynamic SB objective. \hfill $\square$

The Radon–Nikodym derivative between controlled diffusion path measures provides an explicit representation of how changes in drift modify the probability of trajectories (Figure \ref{fig:girsanov}). Using (\ref{eq:girsanov-proof2}), this change of measure produces a \textbf{quadratic control cost} whose expectation reduces to the \textbf{path-space KL divergence} between the two measures. The final integral shows that minimizing the KL divergence between controlled path measures is equivalent to the expectation of the squared difference between the drifts over sampled stochastic paths.

\subsection{Schrödinger Bridge with Arbitrary Reference Dynamics}
\label{subsec:nonlinear-sbp}
Recall from Section \ref{subsec:path-measure-ito-processes} the family of controlled path measures $\mathbb{P}^u\in \mathcal{P}(C([0,T];\mathbb{R}^d))$ which characterize the probability law of stochastic processes $\boldsymbol{X}_{0:T}\in C([0,T];\mathbb{R}^d)$ generated by the controlled SDE:
\begin{align}
    d\boldsymbol{X}_t=(\boldsymbol{f}(\boldsymbol{X}_t,t)+\sigma_t\boldsymbol{u}(\boldsymbol{X}_t,t))dt+ \sigma_td\boldsymbol{B}_t\tag{Controlled SDE}
\end{align}
where $\boldsymbol{f}(\boldsymbol{x},t)$ is the reference drift, $\boldsymbol{u}(\boldsymbol{x},t)$ is the control drift and $\sigma_t$ is the scalar diffusion coefficient. We also defined the \textbf{reference path measure} to be the special case where $\boldsymbol{u}\equiv 0$ and the drift consists only of $\boldsymbol{f}(\boldsymbol{x},t)$.

Within the family of controlled path measures, the \boldtext{dynamic Schrödinger bridge (SB) problem} seeks the control drift $\boldsymbol{u}^\star$ that minimally perturbs the reference path measure while steering the process between prescribed marginal distributions.

More precisely, minimizing the perturbation from the reference path measure is defined as \textbf{minimizing the KL divergence}, which, following Corollary \ref{corollary:kl-divergence-ito}, is equivalent to minimizing the kinetic energy produced from perturbing the reference dynamics with the control drift. Consequently, the dynamic SB can be interpreted as selecting the control drift that minimally perturbs the reference dynamics while satisfying the marginal distribution constraints $\boldsymbol{X}_0\sim \pi_0$ and $\boldsymbol{X}_T\sim \pi_T$.

\begin{definition}[Dynamic Schrödinger Bridge with Arbitrary Reference Dynamics]\label{def:dynamic-sb}
Let $\pi_0, \pi_T\in \mathcal{P}(\mathbb{R}^d)$ denote the prescribed initial and terminal distributions. The \textbf{dynamic Schrödinger bridge problem}, where the reference dynamics are defined by a drift $\boldsymbol{f}$, aims to determine the optimal control drift $\boldsymbol{u}^\star$ that minimizes the KL divergence of the induced path measures subject to the marginal constraints:
\begin{align}
    \inf_{\mathbb{P}^u\in \mathcal{P}(C([0,T];\mathbb{R}^d))}&\text{KL}(\mathbb{P}^u\|\mathbb{Q})= \inf_{\boldsymbol{u}\in \mathcal{U}}\mathbb{E}_{\boldsymbol{X}^u_{0:T}\sim \mathbb{P}^u}\left[\int_0^T\frac{1}{2}\|\boldsymbol{u}(\boldsymbol{X}^u_t,t)\|^2dt\right]\tag{Dynamic SB}\label{eq:dynamic-sb-arbitrary}\\
    &\text{s.t.}\quad \begin{cases}
        d\boldsymbol{X}^u_t=(\boldsymbol{f}(\boldsymbol{x},t)+\sigma_t\boldsymbol{u}(\boldsymbol{x},t))dt+\sigma_td\boldsymbol{B}_t\\
        \boldsymbol{X}^u_0\sim \pi_0, \quad \boldsymbol{X}^u_T\sim \pi_T
    \end{cases}\nonumber
\end{align}
where the KL divergence follows from Corollary \ref{corollary:kl-divergence-ito} and we define:
\begin{enumerate}
    \item [(i)] $\mathcal{U}=\{\boldsymbol{u}\in C^1(\mathbb{R}^d\times [0,T];\mathbb{R}^d)|\exists C> 0, \;\forall (\boldsymbol{x},t)\in \mathbb{R}^d\times [0,T], \;\boldsymbol{u}(\boldsymbol{x},t)\leq C(1+\|\boldsymbol{x}\|)\}$ is the set of all \textbf{feasible control drifts} which contains all smooth control fields that don’t grow faster than linearly, ensuring the controlled stochastic dynamics remain well-behaved \citep{nusken2021solving}\footnote{For the remainder of this guide, we assume optimization is over the set of feasible controls and omit stating it explicitly as a constraint}.
    \item [(ii)] $\mathbb{P}^u\in \mathcal{P}(C([0,T];\mathbb{R}^d))$ as the path measure induced by the controlled SDE and $\boldsymbol{X}^u_{0:T}$ as a stochastic process generated under the controlled SDE with probability law $\mathbb{P}^u$.
\end{enumerate}
Solving (\ref{eq:dynamic-sb-arbitrary}) yields the optimal pair $(\boldsymbol{u}^\star, \mathbb{P}^\star)$ defines the Schrödinger bridge between $\pi_0$ and $\pi_T$ relative to $\boldsymbol{f}$.
\end{definition}

Rather than an expectation over stochastic processes under $\mathbb{P}^u$, we can also write the problem in terms of the time-dependent marginal densities $p_t\in \mathcal{P}(\mathbb{R}^d)$ defined over the space of probability distributions over the state $\mathbb{R}^d$:
\begin{align}
    &\inf_{(\boldsymbol{u}, p_t)}\int_0^T\int_{\mathbb{R}^d}\left[\frac{1}{2}\|\boldsymbol{u}(\boldsymbol{x},t)\|^2\right]p_t(\boldsymbol{x}) d\boldsymbol{x}dt\tag{Density Dynamic SB Problem}\label{eq:nonlinear-sbp-density}\\ &\text{s.t.}\quad \begin{cases}
        \partial_tp_t(\boldsymbol{x})=-\nabla\cdot\big(p_t(\boldsymbol{x})(\boldsymbol{f}(\boldsymbol{x},t) +\sigma_t\boldsymbol{u}(\boldsymbol{x},t))\big)+\frac{\sigma_t^2}{2}\Delta p_t(\boldsymbol{x})\\
        p_0= \pi_0, \quad p_T=\pi_T
    \end{cases}\nonumber
\end{align}
where the marginal distributions satisfy $p_0=\pi_0$ and $p_T=\pi_T$ and the evolution of $p_t$ satisfies the (\ref{eq:controlled-fp-equation}) derived in Section \ref{subsec:fp-equation}.

Since the constrained optimization problem in (\ref{eq:nonlinear-sbp-density}) requires minimizing with respect to both $\boldsymbol{u}(\boldsymbol{x},t)$ and $p_t(\boldsymbol{x})$ which are \textbf{coupled} via the Fokker-Planck PDE constraint, we cannot directly minimize this objective as any variation in $\boldsymbol{u}(\boldsymbol{x},t)$ will implicitly change $p_t(\boldsymbol{x})$. Therefore, we introduce a \boldtext{Lagrange multiplier} $\psi_t$ that turns (\ref{eq:nonlinear-sbp-density}) to an \textbf{unconstrained optimization problem} with the Lagrangian: 
\begin{small}
\begin{align}
    \mathcal{L}(p,\boldsymbol{u},\psi_t)&:=\int_0^T\int_{\mathbb{R}^d}\bigg\{\frac{1}{2}\|\boldsymbol{u}(\boldsymbol{x},t)\|^2 p_t(\boldsymbol{x})\nonumber\\
    &+\psi_t(\boldsymbol{x})\bigg(\partial_tp_t(\boldsymbol{x})+\nabla\cdot\big(p_t(\boldsymbol{x})(\boldsymbol{f}(\boldsymbol{x},t) +\sigma_t\boldsymbol{u}(\boldsymbol{x},t))\big)-\sigma_t^2\Delta p_t(\boldsymbol{x})\bigg)\bigg\}d\boldsymbol{x}dt\label{eq:nonlinear-lagrangian}
\end{align}
\end{small}

This optimization problem yields the pair of density and control drifts $(\boldsymbol{u}^\star, p^\star_t)$ that satisfy the following \boldtext{optimality conditions} \citep{caluya2021wasserstein}.

\begin{proposition}[Optimality Conditions for Dynamic Schrödinger Bridge \citep{caluya2021wasserstein}]\label{prop:optimality-nonlinear-sbp}
    The pair of optimal state PDF and optimal control $(\boldsymbol{u}^\star, p^\star_t)$ that minimize (\ref{eq:nonlinear-lagrangian}) is the solution to the pair of PDEs defined as:
    \begin{align}
        \begin{cases}
            \partial_t\psi_t+\frac{\sigma_t^2}{2}\|\nabla\psi_t\|^2+\langle\nabla\psi_t, \boldsymbol{f}\rangle=-\frac{\sigma^2_t}{2}\Delta \psi_t\\
            \partial_t p_t^\star+\nabla\cdot (p_t^\star(\boldsymbol{f}+\sigma^2_t\nabla \psi_t))=\frac{\sigma^2_t}{2}\Delta p^\star_t
        \end{cases}\quad \text{s.t.}\quad \begin{cases}
            p^\star_0=\pi_0\\
            p^\star_T=\pi_T
        \end{cases}\tag{HJB-FP System}\label{eq:prop-hjb-fpe-system}
    \end{align}
    where $\psi_t(\boldsymbol{x})$ is the Lagrange multiplier. The optimal control $\boldsymbol{u}^\star$ can also be written in terms of $\psi_t$ as:
    \begin{align}
        \forall (\boldsymbol{x},t) \in \mathbb{R}^d\times [0, T], \quad \boldsymbol{u}^\star(\boldsymbol{x},t)=\sigma_t\nabla \psi_t(\boldsymbol{x})\label{eq:optimal-u}
    \end{align}
\end{proposition}

\textit{Proof.} Now, we will break down the proof of this statement, which follows the proof of Proposition 1 in \citep{caluya2021wasserstein}. Expanding the Lagrangian from (\ref{eq:nonlinear-lagrangian}) we have:
\begin{small}
\begin{align}
    &\mathcal{L}(p,\boldsymbol{u},\psi):=\int_0^T\int_{\mathbb{R}^d}\frac{1}{2}\|\boldsymbol{u}(\boldsymbol{x},t)\|^2p_t(\boldsymbol{x})d\boldsymbol{x}dt+\underbrace{\int_0^T\int_{\mathbb{R}^d}\psi_t(\boldsymbol{x})\partial_tp_t(\boldsymbol{x})d\boldsymbol{x}dt}_{(\bigstar)}\nonumber\\
    &+\underbrace{\int_0^T\int_{\mathbb{R}^d}\psi_t (\boldsymbol{x})\nabla\cdot\big(p_t(\boldsymbol{x})(\boldsymbol{f}(\boldsymbol{x},t) +\sigma_t\boldsymbol{u}(\boldsymbol{x},t))\big)d\boldsymbol{x}dt}_{(\blacklozenge)}-\underbrace{\int_0^T\int_{\mathbb{R}^d}\psi_t (\boldsymbol{x})\frac{\sigma_t^2}{2}\Delta p_t(\boldsymbol{x})d\boldsymbol{x}dt}_{(\blacktriangle)}\label{eq:lagrangian-labeled}
\end{align}
\end{small}
Since we want to avoid derivatives with respect to $p^\star$, we move the derivatives onto the Lagrange multiplier $\psi_t(\boldsymbol{x})$ using integration by parts. For the term ($\bigstar$), we have:
\begin{align}
    \int_{\mathbb{R}^d}\int_0^T\underbrace{\psi_t(\boldsymbol{x})}_{u}\underbrace{\partial_tp_t(\boldsymbol{x})}_{dv}dtd\boldsymbol{x}&=\int_{\mathbb{R}^d}\left(\underbrace{[\psi_t(\boldsymbol{x})p_t(\boldsymbol{x})]_{t=0}^T}_{\text{constant}}-\int_0^Tp_t(\boldsymbol{x})\partial_t \psi_t(\boldsymbol{x})dt\right)d\boldsymbol{x}\nonumber\\
    &=-\int_{\mathbb{R}^d}\int_0^Tp_t(\boldsymbol{x})\partial_t\psi_t(\boldsymbol{x})dtd\boldsymbol{x}
\end{align}
where the boundary term reduces to constants $\int\psi_0 (\boldsymbol{x})\pi_0(\boldsymbol{x})dtd\boldsymbol{x}$ and $\int\psi_T (\boldsymbol{x})\pi_T(\boldsymbol{x})dtd\boldsymbol{x}$, which do not depend on the optimization variable $p_t$ and can be dropped from the Lagrangian. For the term ($\blacklozenge$), we can apply integration by parts with the divergence identity $\int\psi_t\nabla\cdot \boldsymbol{v}=-\int \boldsymbol{v}\cdot \nabla\psi_t$ since the boundary condition vanishes at infinity\footnote{The marginal density $p_t$ must integrate to 1 and since $\mathbb{R}^d$ is unbounded, it must vanish as $\|\boldsymbol{x}\|\to \infty$.}.
\begin{small}
\begin{align}
    \int_0^T\int_{\mathbb{R}^d}\psi_t (\boldsymbol{x})\nabla&\cdot\big(p_t(\boldsymbol{x})(\boldsymbol{f}(\boldsymbol{x},t) +\sigma_t\boldsymbol{u}(\boldsymbol{x},t))\big)d\boldsymbol{x}dt\nonumber\\
    &=-\int_0^T\int_{\mathbb{R}^d}p_t(\boldsymbol{x})\nabla \psi_t (\boldsymbol{x})\cdot (\boldsymbol{f}(\boldsymbol{x},t) +\sigma_t\boldsymbol{u}(\boldsymbol{x},t)) d\boldsymbol{x}dt
\end{align}
\end{small}

Finally, we apply integration by parts twice to simplify the term ($\blacktriangle$):
\begin{small}
\begin{align}
    -\int_0^T\int_{\mathbb{R}^d}\psi_t (\boldsymbol{x})&\frac{\sigma_t^2}{2}\Delta p_t(\boldsymbol{x})d\boldsymbol{x}dt=- \int_0^T\int_{\mathbb{R}^d}\psi_t (\boldsymbol{x}) \frac{\sigma_t^2}{2}\sum_{i,j=1}^d\frac{\partial^2}{\partial \boldsymbol{x}_i\partial \boldsymbol{x}_j}p_t(\boldsymbol{x})d\boldsymbol{x}dt\nonumber\\
    &=- \sum_{i,j=1}^d\int_0^T\frac{\sigma_t^2}{2}\int_{\mathbb{R}^d}\underbrace{\psi_t (\boldsymbol{x})}_{u}\underbrace{\frac{\partial}{\partial \boldsymbol{x}_i}\left(\frac{\partial}{\partial \boldsymbol{x}_j}p_t(\boldsymbol{x})\right)}_{dv}d\boldsymbol{x}dt\nonumber\\
    &= \sum_{i,j=1}^d\int_0^T\frac{\sigma_t^2}{2}\int_{\mathbb{R}^d}\underbrace{\frac{\partial}{\partial \boldsymbol{x}_i}\psi_t (\boldsymbol{x})}_{u}\underbrace{\frac{\partial}{\partial \boldsymbol{x}_j}p_t(\boldsymbol{x})}_{dv}d\boldsymbol{x}dt\nonumber\\
    &=- \underbrace{\bluetext{\sum_{i,j=1}^d}\int_0^T\frac{\sigma_t^2}{2}\int_{\mathbb{R}^d}p_t(\boldsymbol{x})\bluetext{\frac{\partial^2}{\partial \boldsymbol{x}_i\partial \boldsymbol{x}_j}\psi_t (\boldsymbol{x})}d\boldsymbol{x}dt}_{\text{sum over $i=j$ gives the Laplacian of }\psi_t}\nonumber\\
    &=-\int_0^T\int_{\mathbb{R}^d}\frac{\sigma_t^2}{2}p_t(\boldsymbol{x})\Delta \psi_t (\boldsymbol{x})d\boldsymbol{x}dt
\end{align}
\end{small}

Now, substituting all terms back into (\ref{eq:lagrangian-labeled}), we get:
\begin{small}
\begin{align}
    &\mathcal{L}(p_t,\boldsymbol{u},\psi)=\int_0^T\int_{\mathbb{R}^d}\frac{1}{2}\|\boldsymbol{u}(\boldsymbol{x},t)\|^2p_t(\boldsymbol{x})d\boldsymbol{x}dt-\int_{\mathbb{R}^d}\int_0^Tp_t(\boldsymbol{x})\partial_t\psi_t(\boldsymbol{x})dtd\boldsymbol{x}\nonumber\\
    &\quad \quad-\int_0^T\int_{\mathbb{R}^d}p_t(\boldsymbol{x})\nabla \psi_t (\boldsymbol{x})\cdot (\boldsymbol{f}(\boldsymbol{x},t) +\sigma_t\boldsymbol{u}(\boldsymbol{x},t)) d\boldsymbol{x}dt-\int_0^T\int_{\mathbb{R}^d}\frac{\sigma_t^2}{2}p_t(\boldsymbol{x})\Delta\psi_t (\boldsymbol{x})d\boldsymbol{x}dt\nonumber\\
    &=\int_0^T\int_{\mathbb{R}^d}\bigg[\frac{1}{2}\|\boldsymbol{u}(\boldsymbol{x},t)\|^2 -\partial_t\psi_t(\boldsymbol{x})-\nabla \psi_t (\boldsymbol{x})\cdot (\boldsymbol{f}(\boldsymbol{x},t) +\sigma_t\boldsymbol{u}(\boldsymbol{x},t))-\frac{\sigma_t^2}{2}\Delta \psi_t (\boldsymbol{x})\bigg]p_t(\boldsymbol{x})d\boldsymbol{x}dt\label{eq:lagrangian-simple}
\end{align}
\end{small}
Isolating the terms that depend on $\boldsymbol{u}(\boldsymbol{x},t)$, we have the minimization problem:
\begin{align}
    \inf_{\boldsymbol{u}}\bigg\{\frac{1}{2}\|\boldsymbol{u}(\boldsymbol{x},t)\|^2 -\bluetext{\underbrace{\nabla \psi_t (\boldsymbol{x})\cdot\sigma_t\boldsymbol{u}(\boldsymbol{x},t)}_{(\sigma_t\boldsymbol{u}(\boldsymbol{x},t))^\top(\nabla \psi_t (\boldsymbol{x}))}}=\frac{1}{2}\|\boldsymbol{u}(\boldsymbol{x},t)\|^2 -\sigma_t\boldsymbol{u}(\boldsymbol{x},t)^\top \nabla \psi_t (\boldsymbol{x})\bigg\}
\end{align}
By completing the square, we get:
\begin{small}
\begin{align}
    \inf_{\boldsymbol{u}}&\bigg\{\underbrace{\frac{1}{2}\|\boldsymbol{u}(\boldsymbol{x},t)\|^2 -\sigma_t\boldsymbol{u}(\boldsymbol{x},t)^\top \nabla \psi_t (\boldsymbol{x})\bluetext{+\frac{\sigma_t^2}{2}\|\nabla \psi_t (\boldsymbol{x})\|^2 }}_{\text{square}}\bluetext{-\frac{\sigma_t^2}{2}\| \nabla \psi_t (\boldsymbol{x})\|^2}\bigg\}\nonumber\\
    &=\inf_{\boldsymbol{u}}\bigg\{\bluetext{\underbrace{\frac{1}{2}\|\boldsymbol{u}(\boldsymbol{x},t)-\sigma_t \nabla \psi_t (\boldsymbol{x})\|^2 }_{\text{minimizing this yields }\boldsymbol{u}^\star}}-\frac{\sigma_t^2}{2}\| \nabla \psi_t (\boldsymbol{x})\|^2\bigg\}
\end{align}
\end{small}
Since the second term doesn't depend on $\boldsymbol{u}(\boldsymbol{x},t)$, minimizing the first term yields the expression (\ref{eq:optimal-u}) for the optimal $\boldsymbol{u}^\star(\boldsymbol{x},t)$:
\begin{align}
    \boldsymbol{u}^\star(\boldsymbol{x},t)=\sigma_t\nabla \psi_t(\boldsymbol{x})
\end{align}
Substituting this expression into the Lagrangian in (\ref{eq:lagrangian-labeled}), we get:
\begin{small}
\begin{align}
    &\mathcal{L}(p,\boldsymbol{u}^\star,\psi)=\int_0^T\int_{\mathbb{R}^d}\bigg\{\frac{1}{2}\|\bluetext{\underbrace{\sigma_t\nabla \psi_t(\boldsymbol{x})}_{\boldsymbol{u}^\star(\boldsymbol{x},t)}}\|^2-\partial_t\psi_t(\boldsymbol{x})-\nabla \psi_t (\boldsymbol{x})\cdot (\boldsymbol{f}(\boldsymbol{x},t) +\sigma_t\bluetext{\underbrace{\sigma_t\nabla \psi_t(\boldsymbol{x})}_{\boldsymbol{u}^\star(\boldsymbol{x},t)}})-\frac{\sigma_t^2}{2}\Delta\psi_t (\boldsymbol{x})\bigg\}p_t(\boldsymbol{x})d\boldsymbol{x}dt\nonumber\\
    &=\int_0^T\int_{\mathbb{R}^d}\bigg\{\bluetext{\frac{\sigma_t^2}{2}\|\nabla \psi_t(\boldsymbol{x})\|^2}-\partial_t\psi_t(\boldsymbol{x})-\langle\nabla \psi_t (\boldsymbol{x}), \boldsymbol{f}(\boldsymbol{x},t)\rangle \bluetext{-\sigma_t^2\|\nabla \psi_t(\boldsymbol{x})\|^2}-\frac{\sigma_t^2}{2}\Delta \psi_t (\boldsymbol{x})\bigg\}p_t(\boldsymbol{x})d\boldsymbol{x}dt\nonumber\\
    &=\int_0^T\int_{\mathbb{R}^d}\bigg\{-\frac{\sigma_t^2}{2}\|\nabla\psi_t(\boldsymbol{x})\|^2 -\partial_t\psi_t(\boldsymbol{x})-\langle\nabla \psi_t (\boldsymbol{x}), \boldsymbol{f}(\boldsymbol{x},t)\rangle-\frac{\sigma_t^2}{2}\Delta \psi_t (\boldsymbol{x})\bigg\}p_t(\boldsymbol{x})d\boldsymbol{x}dt
\end{align}
\end{small}

Since $\boldsymbol{u}^\star$ determines the optimal $p^\star_t$ via the Fokker-Planck constraint, the Lagrangian cannot vary with respect to $p^\star_t$, so the expression within the curly brackets must vanish for optimality to hold over arbitrary $p_t$:
\begin{align}
    -\frac{\sigma_t^2}{2}\|\nabla \psi_t(\boldsymbol{x})\|^2-\partial_t\psi_t(\boldsymbol{x})-\langle\nabla \psi_t (\boldsymbol{x}), \boldsymbol{f}(\boldsymbol{x},t)\rangle-\frac{\sigma_t^2}{2}\Delta \psi_t (\boldsymbol{x})=0
\end{align}
or equivalently:
\begin{align}
    \partial_t\psi_t(\boldsymbol{x})+\frac{\sigma_t^2}{2}\|\nabla \psi_t(\boldsymbol{x})\|^2+\langle\nabla \psi_t (\boldsymbol{x}), \boldsymbol{f}(\boldsymbol{x},t)\rangle=\frac{\sigma_t^2}{2}\Delta \psi_t (\boldsymbol{x})\label{eq:hjb-pde}
\end{align}
which is exactly the \textbf{Hamilton-Jacobi-Bellman (HJB)} equation in (\ref{eq:prop-hjb-fpe-system}). Substituting the optimal expression for $\boldsymbol{u}^\star(\boldsymbol{x},t)$ into the FP equation in (\ref{eq:prop-hjb-fpe-system}), we obtain the FP equation for the optimal density function $p^\star_t$:
\begin{align}
    \partial_tp^\star_t(\boldsymbol{x})=-\nabla\cdot\big(p^\star_t(\boldsymbol{x})(\boldsymbol{f}(\boldsymbol{x},t) +\sigma_t\underbrace{\sigma_t\nabla \psi_t(\boldsymbol{x})}_{\boldsymbol{u}^\star(\boldsymbol{x},t )})\big)+\frac{\sigma_t^2}{2}\Delta p^\star_t(\boldsymbol{x})
\end{align}
which gives the \textbf{Fokker-Planck (FP)} equation (\ref{eq:prop-hjb-fpe-system}). Finally, the boundary conditions $p^\star_0=\pi_0$ and $p^\star_T=\pi_T$ follow directly from (\ref{eq:nonlinear-sbp-density}), which yields the system of non-linear PDEs:
\begin{small}
\begin{align}
    \boxed{\begin{cases}
        \partial_t\psi_t+\frac{\sigma_t^2}{2}\|\nabla\psi_t\|^2+\langle\nabla\psi_t, \boldsymbol{f}\rangle=-\frac{\sigma^2_t}{2}\Delta \psi_t\\
        \partial_t p_t^\star+\nabla\cdot (p_t^\star(\boldsymbol{f}+\sigma^2_t\nabla \psi_t))=\frac{\sigma^2_t}{2}\Delta p^\star_t
    \end{cases}\quad \text{s.t.}\quad \begin{cases}
        p^\star_0=\pi_0\\
        p^\star_T=\pi_T
    \end{cases}}\tag{HJB-FP System}
\end{align}
\end{small}
which define the optimal pair $(\psi_t,p^\star_t)$ and equivalently $(\boldsymbol{u}^\star, p_t^\star)$ that solve the dynamic SB problem. \hfill $\square$

Although the coupled nonlinear PDE system in (\ref{eq:prop-hjb-fpe-system}) fully characterizes the \textbf{optimal control–density pair} $(\boldsymbol{u}^\star, p^\star_t)$, solving this system remains challenging in general, particularly under nonlinear prior dynamics and arbitrary marginal constraints. A key step toward tractability is the Hopf–Cole transform, which converts the nonlinear system in (\ref{eq:prop-hjb-fpe-system}) into an equivalent pair of \textbf{linear PDEs}. 

Before moving on to deriving these linear PDE optimality conditions, let's first briefly review several commonly used reference processes $\mathbb{Q}$ that appear across SB formulations and modern generative modeling frameworks.

\purple[Reference Processes for Schrödinger Bridges]{
We now define a few common reference processes $\mathbb{Q}$, in addition to pure Brownian motion $\sigma\mathbb{B}$, to reveal the connection between various generative modeling paradigms and the dynamic Schrödinger bridge problem.

\paragraph{Ornstein–Uhlenbeck (OU) processes.} The class of \textbf{Ornstein–Uhlenbeck} (OU) processes is a common form of reference process in SB literature \citep{vargas2021solving, shi2023diffusion, de2021diffusion} and takes the form:
\begin{align}
    d\boldsymbol{X}_t=-\beta\boldsymbol{X}_tdt +\sigma_td\boldsymbol{B}_t\tag{Ornstein–Uhlenbeck SDE}
\end{align}
Unlike pure Brownian motion, the OU process introduces a deterministic drift $-\beta\boldsymbol{X}_t$ that pulls the system toward the origin, creating \textbf{mean-reverting} stochastic processes. While Brownian motion diverges in variance, OU processes approach a stationary Gaussian distribution.

\paragraph{Variance Exploding SDEs.} The classic \textit{score-based generative modeling}  framework \citep{huang2021variational, song2019generative, song2020score} adopts the class of \textbf{variance exploding SDEs} (VESDEs) as the reference process, which takes the form:
\begin{align}
    d\boldsymbol{X}_t=\sqrt{d\sigma^2_t/dt}d\boldsymbol{B}_t, \quad \boldsymbol{X}_0\sim \pi_0\tag{Variance Exploding SDE}
\end{align}
where $\sigma_t^2$ is the variance that increases with time. By integrating the SDE over time $[0,t]$, the variance is defined as $\beta_t:=\int_0^t\sigma_s^2ds$, which increases over time. Therefore, the marginal density is given by $\mathcal{N}(\boldsymbol{0}, \beta_t\boldsymbol{I}_d)$.

\paragraph{Variance Preserving SDEs.} Another class of reference processes are the \textbf{variance-preserving SDEs} (VPSDEs) that underlie the widely-adopted \textit{denoising diffusion probabalistic model} (DDPM) framework \citep{ho2020denoising, sohl2015deep, song2020score}, which take the form:
\begin{align}
    d\boldsymbol{X}_t=-\frac{1}{2}\beta_t\boldsymbol{X}_tdt+\sqrt{\beta_t}d\boldsymbol{B}_t, \quad \boldsymbol{X}_0\sim \pi_0 \tag{Variance Preserving SDE}
\end{align}
Unlike VESDEs, the drift and diffusion terms in VPSDEs are carefully balanced so that the total variance of the process remains approximately constant, and the marginal distribution follows a constant-variance Gaussian $\mathcal{N}(\boldsymbol{0}, \boldsymbol{I}_d)$.

Other reference processes have also been explored, including alternative definitions of the coefficient $\sigma_t$ in VESDEs \citep{karras2022elucidating, song2023consistency}, sub-VPSDEs \citep{song2020score}, and fractional Brownian motion \citep{nobis2025fractional}, which we will explore in depth in Section \ref{subsec:fractional-sbp}.
}

Having defined the dynamic SB problem given an arbitrary reference process $\mathbb{Q}$, the system of non-linear PDEs that define the optimal control-density pair $(\boldsymbol{u}^\star, p_t^\star)$, and some examples of reference processes $\mathbb{Q}$ commonly seen in literature, we will move on to the derivation of the Hopf-Cole transform. This transform converts the nonlinear PDE conditions to a system of linear PDE conditions, where the solution is a pair of time-evolving Schrödinger potentials $(\varphi_t, \hat\varphi_t)$ that define the marginal density of the SB, which are analogous to the static Schrödinger potentials $(\varphi, \hat\varphi)$ that define the optimal static SB problem $\pi^\star_{0,T}$ from Section \ref{subsec:static-sbp}.

\subsection{Hopf-Cole Transform}
\label{subsec:hopf-cole-transform}
The Hopf-Cole transform allows us to transform the \textbf{non-linear PDEs coupled over the full trajectory} into a system of \textbf{linear PDEs that are only coupled via their boundary constraints}. This is achieved with a change-of-variables from the \textbf{optimal control–density pair} $(\boldsymbol{u}^\star, p^\star_t)$ to a pair of potential functions $(\varphi_t, \hat\varphi_t)$ that define a \textbf{coupled system of linear PDEs}. To derive this, we first recall the (\ref{eq:prop-hjb-fpe-system}) as:
\begin{align}
        \partial_t\psi_t+\frac{\sigma_t^2}{2}\|\nabla \psi_t\|^2+\langle\nabla \psi_t, \boldsymbol{f}\rangle&=-\frac{\sigma^2_t}{2}\Delta \psi_t\tag{HJB Equation}\label{eq:hjb-system}\\
        \partial_t p_t^\star+\nabla\cdot (p_t^\star(\boldsymbol{f}+\sigma^2_t\nabla \psi_t))&=\frac{\sigma^2_t}{2}\Delta p^\star_t\tag{FP Equation}\label{eq:fpe-system}
\end{align}

To understand the intuition behind the Hopf-Cole transform, we can make the simple observation that the quadratic term $\frac{\sigma_t^2}{2}\|\nabla \psi\|^2$ is the \textit{only} non-linear term in the HJB-FP system. Therefore, the goal of the transform is to answer: \textit{how do we define a change-of-variables that makes this linear?}

That is exactly the goal of the \boldtext{Hopf-Cole Transform}, which defines a change-of-variables to the non-linear HJB-FP system such that they can be solved via a system of \textbf{linear PDEs}, whose optimality conditions align \textit{exactly} to the solution of the non-linear system.

\begin{theorem}[Hopf-Cole Transform]\label{thm:hopf-cole-nonlinear}
    Given an reference process defined by the deterministic drift $\boldsymbol{f}(\boldsymbol{x},t)$, diffusion coefficient $\sigma_t$, and boundary marginal distributions $\pi_0, \pi_T\in \mathcal{P}(\mathbb{R}^d)$, we can apply the following change of variables $(\psi, p_t^\star) \mapsto(\varphi_t, \hat{\varphi}_t)$ defined as:
    \begin{align}
        \psi_t(\boldsymbol{x})=\log \varphi_t(\boldsymbol{x}), \quad p^\star_t(\boldsymbol{x})=\varphi(\boldsymbol{x})\hat\varphi(\boldsymbol{x})\tag{Hopf-Cole Transform}\label{eq:hopf-cole-change-var}
    \end{align}
    which transforms (\ref{eq:hjb-system}) and (\ref{eq:fpe-system}) into a system of \textbf{linear PDEs} for $(\varphi_t, \hat{\varphi}_t)$ given by:
    \begin{align}
        \begin{cases}
            \partial_t\varphi_t+\langle \nabla\varphi_t, \boldsymbol{f}\rangle=-\frac{\sigma_t^2}{2}\Delta \varphi_t\\
            \partial_t\hat{\varphi}_t+\nabla\cdot (\hat{\varphi}_t \boldsymbol{f})=\frac{\sigma_t^2}{2}\Delta \hat{\varphi}_t
        \end{cases}\quad\text{s.t.}\quad
        \begin{cases}
            p^\star_0=\varphi_0\hat{\varphi}_0\\
            p^\star_T=\varphi_T\hat{\varphi}_T
        \end{cases}\tag{Hopf-Cole PDEs}\label{eq:hopf-cole-system}
    \end{align}
    which define the dynamics and terminal conditions of $(\varphi_t,\hat\varphi_t)$. Furthermore, the optimal control can be written as:
    \begin{align}
        \boldsymbol{u}^\star(\boldsymbol{x},t)&=\sigma_t\nabla\log\varphi_t(\boldsymbol{x})\tag{Optimal Control with SB Potential}\label{eq:optimal-control-sb-potential}
    \end{align}
\end{theorem}

\textit{Derivation.} To derive this from first principles, we can start with a well-known property of the Laplacian operation, which is that \textbf{the Laplacian of a logarithm produces squared gradients}. We can leverage this property to write the Laplacian $\Delta \psi_t$ as a squared gradient to cancel out the quadratic term. Specifically, let us define an \textit{ansatz} for the change of variables $\psi\mapsto \varphi$ as:
\begin{align}
    \psi_t(\boldsymbol{x})=C\log\varphi_t(\boldsymbol{x})\iff \varphi_t(\boldsymbol{x})=\exp\left(\frac{\psi_t(\boldsymbol{x})}{C}\right)
\end{align}
where $C$ is some unknown scalar constant. We now express each term containing $\psi_t$ in terms of $\varphi$. 
\begin{enumerate}
    \item The quadratic term becomes:
    \begin{align}
        \frac{\sigma_t^2}{2}\|\nabla \bluetext{\psi_t}\|^2=\frac{\sigma_t^2}{2}\|\nabla \bluetext{(C\log\varphi_t)}\|^2=\frac{\sigma_t^2C^2}{2}\left\|\frac{\nabla \varphi_t}{\varphi_t}\right\|^2=\frac{\sigma_t^2C^2}{2}\frac{\|\nabla \varphi_t\|^2}{\varphi_t^2}
    \end{align}
    \item The Laplacian term becomes:
    \begin{align}
        -\frac{\sigma_t^2}{2}\bluetext{\Delta \psi_t}&=-\frac{\sigma_t^2}{2}\left(\bluetext{\nabla\cdot \nabla (C\log\varphi_t)}\right)=-\frac{\sigma_t^2}{2} C\bluetext{\nabla \cdot \left(\frac{\nabla\varphi_t}{\varphi_t}\right)}=-\frac{\sigma_t^2}{2}C\bluetext{\nabla \cdot \left(\varphi_t^{-1}\nabla \varphi_t\right)}
    \end{align}
    Applying the product rule of divergences $\nabla\cdot (u\boldsymbol{v})=(\nabla u)\cdot \boldsymbol{v}+u(\nabla \cdot \boldsymbol{v})$, we get:
    \begin{align}
        -\frac{\sigma_t^2}{2}\Delta \psi&=-\frac{\sigma_t^2}{2} C\left(\bluetext{\nabla \cdot \left(\varphi^{-1}\nabla \varphi\right)}\right)=-\frac{\sigma_t^2}{2}C\left(\bluetext{\nabla (\varphi^{-1})\cdot \nabla \varphi+\varphi^{-1}(\nabla\cdot \nabla \varphi)}\right)\\
        &=-\frac{\sigma_t^2}{2} C\left(\bluetext{\frac{\Delta \varphi}{\varphi}-\frac{\|\nabla \varphi\|^2}{\varphi^2}}\right)=\frac{\sigma_t^2}{2} C\left(\bluetext{\frac{\|\nabla \varphi\|^2}{\varphi^2}-\frac{\Delta \varphi}{\varphi}}\right)
    \end{align}
\end{enumerate}
To determine the value of $C$ that allows the quadratic term to cancel, we set them equal to each other:
\begin{align}
    \frac{\sigma_t^2C^2}{2}\frac{\|\nabla\varphi\|^2 }{\varphi^2}&=\frac{\sigma_t^2}{2} C\frac{\|\nabla \varphi\|^2 }{\varphi^2}\implies  C=1
\end{align}
Now, we have defined the mapping $\psi_t\mapsto \varphi_t$ as $\psi_t(\boldsymbol{x})=\log\varphi_t(\boldsymbol{x})$ which makes the HJB PDE linear. Substituting $\psi_t(\boldsymbol{x})=\log\varphi_t(\boldsymbol{x})$ into the (\ref{eq:fpe-system}), we get:
\begin{align}
    \partial_t\log\varphi_t+\langle\nabla\log \varphi_t, \boldsymbol{f}\rangle +\frac{\sigma_t^2}{2}\|\nabla \log \varphi_t\|^2&=-\frac{\sigma_t^2}{2}\Delta \log \varphi_t\nonumber\\
    \frac{1}{\varphi_t}\partial_t \varphi_t+\frac{1}{\varphi_t}\langle\nabla \varphi_t, \boldsymbol{f}\rangle +\frac{\sigma_t^2}{2}\frac{\|\nabla \varphi_t\|^2}{\varphi_t^2}&=-\frac{\sigma_t^2}{2}\left(\frac{\Delta \varphi_t}{\varphi_t}-\frac{\|\nabla \varphi_t\|^2}{\varphi_t^2}\right)\nonumber\\
    \frac{1}{\varphi_t}\partial_t \varphi_t+\frac{1}{\varphi_t}\langle\nabla \varphi_t, \boldsymbol{f}\rangle  +\frac{\sigma_t^2}{2}\frac{\|\nabla \varphi_t\|^2}{\varphi_t^2}&=-\frac{\sigma_t^2}{2}\frac{\Delta \varphi_t}{\varphi_t}+\frac{\sigma_t^2}{2}\frac{\|\nabla \varphi_t\|^2}{\varphi_t^2}
\end{align}
Subtracting $\frac{\sigma_t^2}{2}\frac{\|\nabla \varphi\|^2}{\varphi^2}$ from both sides and multiplying by $\varphi_t$, we get:
\begin{align}
    \partial_t \varphi_t+\langle\nabla \varphi_t, \boldsymbol{f}\rangle =-\frac{\sigma_t^2}{2} \Delta \varphi_t
\end{align}
which is exactly the \textbf{first equation} in (\ref{eq:hopf-cole-system}). Now, we need to define a change-of-variables for $p_t$ that decouples $p_t^\star$ and $\nabla\psi_t$ in the (\ref{eq:fpe-system}). Substituting our first mapping $\psi_t(\boldsymbol{x})=\log\varphi_t(\boldsymbol{x})$ into the Fokker-Planck equation, we get:
\begin{align}
    \partial_t p^\star_t+\nabla\cdot (p^\star_t(\boldsymbol{f}+\sigma_t^2\bluetext{\nabla \psi_t}))&=\frac{\sigma_t^2}{2} \Delta p^\star_t\nonumber\\
    \partial_t p^\star_t+\nabla\cdot (p^\star_t(\boldsymbol{f}+\sigma^2_t\bluetext{\nabla _{\boldsymbol{x}} \log\varphi_t)})&=\frac{\sigma_t^2}{2} \Delta p^\star_t
\end{align}
Observing this equation, we see that the term $\nabla \log\varphi_t$ appears as an additional drift, which reweights the trajectories according to the potential $\varphi_t$, but also couples the density $p^\star_t$ with the potential $\varphi_t$. To decouple this interaction, we can divide the density by $\varphi_t$ to get the change of variables $\hat{\varphi}_t(\boldsymbol{x})=\frac{p^\star_t(\boldsymbol{x})}{\varphi_t(\boldsymbol{x})}$ such that the optimal density factorizes as:
\begin{align}
    \hat{\varphi}_t(\boldsymbol{x})=\frac{p^\star_t(\boldsymbol{x})}{\varphi_t(\boldsymbol{x})}\implies p^\star_t(\boldsymbol{x})=\hat{\varphi}_t(\boldsymbol{x})\varphi_t(\boldsymbol{x})=\hat{\varphi}_t(\boldsymbol{x})e^{\psi_t(\boldsymbol{x})}
\end{align}
which separates the dynamics into forward and backward components. Substituting $p^\star_t=\varphi_t\hat{\varphi}_t$ into each term in the (\ref{eq:fpe-system}), we get:
\begin{enumerate}
    \item The time derivative becomes:
    \begin{align}
        \partial_t(\varphi_t\hat{\varphi}_t)=\left(\partial_t \varphi_t\right)\hat{\varphi}_t+\varphi_t \left(\partial_t \hat{\varphi}_t\right)
    \end{align}
    and substituting the expression $\frac{\partial \varphi}{\partial t } =-\langle\nabla \varphi_t, \boldsymbol{f}\rangle-\frac{\sigma_t^2}{2} \Delta \varphi_t$ from the first PDE, we get:
    \begin{small}
    \begin{align}
        \partial_t p^\star_t=\left(\bluetext{-\langle\nabla \varphi_t, \boldsymbol{f}\rangle-\frac{\sigma_t^2}{2} \Delta \varphi_t}\right)\hat{\varphi}_t+\varphi_t \left(\partial_t \hat{\varphi}_t\right)=\bluetext{-\hat\varphi_t\langle\nabla \varphi_t, \boldsymbol{f}\rangle-\frac{\sigma_t^2}{2} \hat\varphi_t\Delta \varphi_t}+\varphi_t \left(\partial_t \hat{\varphi}_t\right)\label{eq:hopf-cole-proof1}
    \end{align}
    \end{small}
    \item The drift term becomes:
    \begin{small}
    \begin{align}
        \nabla\cdot &(\bluetext{p^\star_t}(\boldsymbol{f}+\sigma^2_t\nabla _{\boldsymbol{x}} \log\varphi_t))=\nabla\cdot (\bluetext{\varphi_t\hat\varphi_t}(\boldsymbol{f}+\sigma^2_t\nabla _{\boldsymbol{x}} \log\varphi_t))=\nabla\cdot \bigg(\bluetext{\varphi_t\hat\varphi_t}\bigg(\boldsymbol{f}+\sigma^2_t\frac{\nabla _{\boldsymbol{x}} \varphi_t}{\varphi_t}\bigg)\bigg)\nonumber\\
        &=\nabla\cdot (\bluetext{\varphi_t\hat\varphi_t}\boldsymbol{f})+\nabla\cdot \bigg(\sigma^2_t(\varphi_t\hat\varphi_t)\frac{\nabla _{\boldsymbol{x}} \varphi_t}{\varphi_t}\bigg)=\nabla\cdot (\bluetext{\varphi_t\hat\varphi_t}\boldsymbol{f})+\sigma^2_t\nabla\cdot (\hat\varphi_t\nabla _{\boldsymbol{x}} \varphi_t)\nonumber\\
        &=\langle\nabla (\varphi_t\hat\varphi_t), \boldsymbol{f}\rangle +(\varphi_t\hat\varphi_t)\nabla\cdot \boldsymbol{f} +\sigma^2_t(\nabla \hat{\varphi}_t\cdot \nabla \varphi_t+\hat{\varphi}_t\Delta \varphi_t)\nonumber\\
        &=\hat\varphi_t\langle\nabla \varphi_t, \boldsymbol{f}\rangle +\varphi_t\langle\nabla \hat\varphi_t, \boldsymbol{f}\rangle +(\varphi_t\hat\varphi_t)\nabla\cdot \boldsymbol{f} +\sigma^2_t\nabla \hat{\varphi}_t\cdot \nabla \varphi_t+\sigma^2_t\hat{\varphi}_t\Delta \varphi_t\label{eq:hopf-cole-proof2}
    \end{align}
    \end{small}
    \item The Laplacian term becomes:
    \begin{small}
    \begin{align}
        \frac{\sigma_t^2}{2} \Delta p_t^\star&=\frac{\sigma_t^2}{2} \Delta (\varphi_t\hat{\varphi}_t)=\frac{\sigma_t^2}{2}\nabla\cdot (\hat{\varphi}_t\nabla \varphi_t+\varphi_t\nabla \hat{\varphi}_t)\nonumber\\
        &=\frac{\sigma_t^2}{2}(\nabla \cdot (\hat{\varphi}_t\nabla \varphi_t)+\nabla\cdot(\varphi_t\nabla \hat{\varphi}_t))=\frac{\sigma_t^2}{2}(\nabla \hat{\varphi}_t\cdot \nabla \varphi_t+\hat{\varphi}\Delta\varphi+\nabla \varphi_t\cdot\nabla \hat{\varphi}+\varphi_t\Delta\hat{\varphi}_t)\nonumber\\
        &=\frac{\sigma_t^2}{2} (\hat{\varphi}_t\Delta\varphi_t+2\nabla \hat{\varphi}_t\cdot \nabla \varphi_t+\varphi_t\Delta\hat{\varphi}_t)=\frac{\sigma_t^2}{2} \hat{\varphi}_t\Delta\varphi_t+\sigma_t^2\nabla \hat{\varphi}_t\cdot \nabla \varphi_t+\frac{\sigma_t^2}{2}\varphi_t\Delta\hat{\varphi}_t\label{eq:hopf-cole-proof3}
    \end{align}
    \end{small}
\end{enumerate}
Putting it together and canceling terms, we get:
\begin{small}
\begin{align}
    \partial_t p^\star_t+\nabla\cdot (p_t^\star\nabla \log\varphi_t)&=\frac{\sigma_t^2}{2} \Delta p_t^\star\nonumber\\
    \underbrace{\bluetext{-\hat\varphi_t\langle\nabla \varphi_t, \boldsymbol{f}\rangle-\frac{\sigma_t^2}{2} \hat\varphi_t\Delta \varphi_t}+\varphi_t \left(\partial_t \hat{\varphi}_t\right)}_{(\ref{eq:hopf-cole-proof1})}&+\underbrace{\bluetext{\hat\varphi_t\langle\nabla \varphi_t, \boldsymbol{f}\rangle} +\varphi_t\langle\nabla \hat\varphi_t, \boldsymbol{f}\rangle +(\varphi_t\hat\varphi_t)\nabla\cdot \boldsymbol{f} +\bluetext{\sigma^2_t\nabla \hat{\varphi}_t\cdot \nabla \varphi_t}+\bluetext{\sigma^2_t\hat{\varphi}_t\Delta \varphi_t}}_{(\ref{eq:hopf-cole-proof2})}\nonumber\\
    &=\underbrace{\bluetext{\frac{\sigma_t^2}{2} \hat{\varphi}_t\Delta\varphi_t+\sigma_t^2\nabla \hat{\varphi}_t\cdot \nabla \varphi_t}+\frac{\sigma_t^2}{2}\varphi_t\Delta\hat{\varphi}_t}_{(\ref{eq:hopf-cole-proof3})}\nonumber\\
    \varphi_t \left(\bluetext{\partial_t \hat{\varphi}_t}\right)+\varphi_t\langle\nabla \hat\varphi, \boldsymbol{f}\rangle+(\varphi_t\hat\varphi_t)\nabla\cdot\boldsymbol{f}&=\frac{\sigma_t^2}{2}\varphi_t\Delta\hat{\varphi}_t\nonumber\\
    \partial_t \hat{\varphi}_t+\bluetext{\underbrace{\langle\nabla \hat\varphi, \boldsymbol{f}\rangle+\hat\varphi_t\nabla\cdot\boldsymbol{f}}_{=\nabla \cdot (\boldsymbol{f}\hat\varphi_t)}}&=\frac{\sigma_t^2}{2}\Delta\hat{\varphi}_t\nonumber\\
    \partial_t \hat{\varphi}_t+\nabla \cdot (\hat\varphi_t\boldsymbol{f})&=\frac{\sigma_t^2}{2}\Delta\hat{\varphi}_t
\end{align}
\end{small}
which recovers the \textbf{second equation} in (\ref{eq:hopf-cole-system}). Now, we have derived the Hopf-Cole transform $(\psi, p_t^\star)\mapsto (\varphi_t, \hat{\varphi}_t)$ that solve a system of linear PDE equations:
\begin{align}
    \boxed{\begin{cases}
        \partial_t\varphi_t+\langle \nabla\varphi_t, \boldsymbol{f}\rangle=-\frac{\sigma_t^2}{2}\Delta \varphi_t\\
        \partial_t\hat{\varphi}_t+\nabla\cdot (\hat{\varphi}_t \boldsymbol{f})=\frac{\sigma_t^2}{2}\Delta \hat{\varphi}_t
    \end{cases}
    }\tag{Hopf-Cole PDEs}
\end{align}
which \textit{uniquely} define the optimal control-density pair $(\boldsymbol{u}^\star, p^\star_t)$ of the Schrödinger bridge:
\begin{align}
    \boxed{\boldsymbol{u}^\star(\boldsymbol{x},t)=\sigma_t\nabla\log \varphi_t(\boldsymbol{x}),\quad p^\star_t(\boldsymbol{x})=\varphi_t(\boldsymbol{x})\hat{\varphi}_t(\boldsymbol{x})}\tag{Optimal Control-Density Pair}
\end{align}
yielding the boundary constraints $p_0^\star(\boldsymbol{x})=\pi_0(\boldsymbol{x})= \varphi_0(\boldsymbol{x})\hat\varphi_0(\boldsymbol{x})$ and $p_T^\star(\boldsymbol{x})=\pi_T(\boldsymbol{x})=\varphi_T(\boldsymbol{x}) \hat\varphi_T(\boldsymbol{x})$. \hfill $\square$

Given these optimality constraints, we can derive a system of equations that explicitly define the forward and backward potential functions.

\begin{corollary}[Forward–Backward Schrödinger Potentials]\label{corollary:schrodinger-system}
    Let $(\varphi_t, \hat\varphi_t)$ denote the Schrödinger potentials obtained through the Hopf-Cole transform, which define the solution to the (\ref{eq:dynamic-ot-prob}) as:
    \begin{small}
    \begin{align}
        p_t^\star(\boldsymbol{x})=\varphi_t(\boldsymbol{x})\hat\varphi_t(\boldsymbol{x}), \quad \boldsymbol{u}^\star(\boldsymbol{x},t)=\sigma_t\nabla\log \varphi_t(\boldsymbol{x})
    \end{align}
    \end{small}
    Then, $(\varphi_t, \hat\varphi_t)$ can be represented as the solution to a system of equations with the transition density under the reference path measure $\mathbb{Q}$ given by:
    \begin{small}
    \begin{align}
        \begin{cases}
        \varphi_t(\boldsymbol{x})=\int_{\mathbb{R}^d}\mathbb{Q}_{T|t}(\boldsymbol{y}|\boldsymbol{x})\varphi_T(\boldsymbol{y})d\boldsymbol{y}\\
        \hat{\varphi}_t(\boldsymbol{x})=\int_{\mathbb{R}^d}\mathbb{Q}_{t|0}(\boldsymbol{x}|\boldsymbol{y})\hat{\varphi}_0(\boldsymbol{y})d\boldsymbol{y}
    \end{cases}\quad\text{s.t.}\quad
    \begin{cases}
        \pi_0(\boldsymbol{x})=\varphi_0(\boldsymbol{x})\hat{\varphi}_0(\boldsymbol{x})\\
        \pi_T(\boldsymbol{x})=\varphi_T(\boldsymbol{x})\hat{\varphi}_T(\boldsymbol{x})
    \end{cases}\tag{Schrödinger Potentials}\label{eq:proof-sb-system}
    \end{align}
    \end{small}
    subject to the boundary factorization constraints. 
\end{corollary}

\textit{Proof.} Observe that the system of linear PDEs derived in Theorem \ref{thm:hopf-cole-nonlinear} aligns with the form of the (\ref{eq:kolmogorov-bwd-equation}) defined in Corollary \ref{corollary:kolmogorov-backward}, which we recall is defined for some function $r(\boldsymbol{x},t)$ as:
\begin{small}
\begin{align}
    \partial_t r(\boldsymbol{x},t)+\langle\boldsymbol{f}(\boldsymbol{x},t), \nabla r(\boldsymbol{x},t)\rangle+\frac{\sigma_t^2}{2}\Delta r(\boldsymbol{x},t)=0, \quad r(\boldsymbol{x},T)=\Phi(\boldsymbol{x})\tag{Kolmogorov Backward Equation}
\end{align}
\end{small}
which admits the (\ref{eq:feynman-kac-kolmogorov-bwd}):
\begin{small}
\begin{align}
r(\boldsymbol{x},t):=\mathbb{E}_{\boldsymbol{X}_{t:T}\sim \mathbb{Q}}\left[\Phi(\boldsymbol{X}_T)|\boldsymbol{X}_t=\boldsymbol{x}\right]\tag{Feynman-Kac Representation}
\end{align}
\end{small}
The forward potential $\varphi_t$ satisfies the first linear PDE in (\ref{eq:hopf-cole-system}), which is equivalent to the backward Kolmogorov equation with $r(\boldsymbol{x},t):=\varphi_t(\boldsymbol{x})$ and $\Phi(\boldsymbol{x})=\varphi_T(\boldsymbol{x})$ associated with the reference process $\mathbb{Q}$. Therefore, $\varphi_t$ can be written with the Feynman-Kac representation as:
\begin{small}
\begin{align}
\varphi_t(\boldsymbol{x})=\mathbb{E}_{\boldsymbol{X}_{t:T}\sim \mathbb{Q}}\left[\bluetext{\varphi_T(\boldsymbol{X}_T)}|\boldsymbol{X}_t=\boldsymbol{x}\right]
\end{align}
\end{small}
Equivalently, since the conditional density of $\boldsymbol{X}_T$ given $\boldsymbol{X}_t =\boldsymbol{x}$ is defined by the transition density under the reference process $\mathbb{Q}_{T|t}(\cdot|\boldsymbol{x})=\mathbb{Q}(\boldsymbol{X}_T=\cdot|\boldsymbol{X}_t=\boldsymbol{x})$, we can write the Feynman-Kac representation the \textbf{linear integral operator}:
\begin{small}
\begin{align}
\boxed{\varphi_t(\boldsymbol{x})=\int_{\mathbb{R}^d}\mathbb{Q}_{T|t}(\boldsymbol{x}_T|\boldsymbol{x})\varphi_T(\boldsymbol{x}_T) d\boldsymbol{x}_T}\tag{Forward Potential}\label{eq:forward-potential-integral}
\end{align}
\end{small}
Similarly, the backward potential $\hat\varphi_t$ satisfies the second linear PDE in (\ref{eq:hopf-cole-system}), which is equivalent to the backward Kolmogorov equation with $r(\boldsymbol{x},t):=\hat\varphi_t(\boldsymbol{x})$ and terminal constraint $\Phi(\boldsymbol{x})=\hat\varphi_0(\boldsymbol{x})$. Therefore, $\hat\varphi_t$ can be written with the Feynman-Kac representation as:
\begin{small}
\begin{align}
\boxed{\hat\varphi_t(\boldsymbol{x})=\mathbb{E}_{\boldsymbol{X}_{0:t}\sim \mathbb{Q}}\left[\bluetext{\hat\varphi_0(\boldsymbol{X_0})}|\boldsymbol{X}_t=\boldsymbol{x}\right]=\int_{\mathbb{R}^d}\mathbb{Q}_{t|0}(\boldsymbol{x}|\boldsymbol{x}_0)\hat\varphi_0(\boldsymbol{x}_0) d\boldsymbol{x}_0}\tag{Backward Potential}\label{eq:backward-potential-integral}
\end{align}
\end{small}
which are exactly the equations defined in (\ref{eq:proof-sb-system}).\hfill $\square$

This result shows that the forward potential $\varphi_t$ propogates backward from the terminal constraint $\pi_T= \varphi_T\hat\varphi_T$ via a linear integral operator over the expected distribution of $\varphi_T(\boldsymbol{X}_T)$ and the backward potential propogates forward from the reversed terminal constraint $\pi_0=\varphi_0\hat\varphi_0$ via a linear integral operator in the reverse time over the expected distribution of $\hat\varphi_0(\boldsymbol{X}_0)$.

Recall from our discussion on the static Schrödinger bridge problem in Section \ref{subsec:static-sbp} where we define the pair of SB potentials $(\varphi, \hat\varphi)$ as the solution to the (\ref{eq:schrodinger-system}). Crucially, we can interpret the Schrödinger potentials $(\varphi_t,\hat\varphi_t)$ obtained through the Hopf–Cole transform as the \textbf{dynamic or continuous-time analogues of the Schrödinger potentials}. In the static formulation, the optimal coupling between the marginals $(\pi_0, \pi_T)$ under the reference kernel $\mathcal{K}(\boldsymbol{x},\boldsymbol{y})=e^{-c(\boldsymbol{x},\boldsymbol{y})}$ admits the factorized form:
\begin{small}
\begin{align}
    \pi_{0,T}^\star(\boldsymbol{x},\boldsymbol{y})=e^{\varphi(\boldsymbol{x})+\hat{\varphi}(\boldsymbol{y})-c(\boldsymbol{x},\boldsymbol{y})}= e^{\varphi(\boldsymbol{x})}\mathcal{K}(\boldsymbol{x},\boldsymbol{y})e^{\hat{\varphi}(\boldsymbol{y})}\tag{Solution to Static SB}
\end{align}
\end{small}

The (\ref{eq:dynamic-sb-arbitrary}) problem generalizes this structure from couplings of endpoints to path measures of stochastic processes. In this setting, the time-dependent potentials $(\varphi_t,\hat\varphi_t)$ propagate according to the forward–backward linear PDE system derived above, and their product recovers the optimal marginal density at each time $p_t^\star=\varphi_t\hat\varphi_t$. 

As shown in Theorem \ref{theorem:dual-static-sb}, the static Schrödinger system admits a \textit{unique solution up to an additive constant}. Since the solution to the dynamic SB problem $p^\star_t$ is also unique by the strict convexity of the KL divergence, the product $\varphi_t\hat\varphi_t$ is unique, and the individual potentials $\varphi_t$ and $\hat\varphi_t$ are \textbf{unique up to a multiplicative constant} which leaves their product invariant.

\subsection{Schrödinger Bridges as Entropy-Regularized Dynamic Optimal Transport}
\label{subsec:sb-regularized-dynamic-ot}

One of the most important perspectives on the Schrödinger bridge problem is its close relationship to optimal transport. In Section \ref{sec:static-sb}, we derived the static SB problem directly from the entropic OT problem, with a simple reparameterization of the reference coupling. 

In this section, we make a similar connection to the \textbf{Benamou and Brenier (dynamic) formulation of the OT problem} defined in Section \ref{subsec:dynamic-ot-problem}. We will show that by reparameterizing the control drift in the controlled Fokker–Planck equation, the stochastic dynamics can be expressed as a deterministic continuity equation describing the transport of probability mass by a velocity field. Under this transformation, the dynamic SB objective decomposes into a kinetic transport energy that aligns with the dynamic OT objective, with additional \textbf{entropy-regularization terms} induced by diffusion.

\begin{proposition}[Dynamic Optimal Transport Form of Schrödinger Bridge]
    Consider the dynamic Schrödinger bridge problem written in terms of the marginal density in (\ref{eq:controlled-fp-equation}) as:
    \begin{align}
        &\inf_{(\boldsymbol{u}, p_t)}\left[\int_0^T\int_{\mathbb{R}^d}\frac{1}{2}\|\boldsymbol{u}(\boldsymbol{x},t)\|^2p_t(\boldsymbol{x}) d\boldsymbol{x}dt\right]\tag{Density Dynamic SB Objective}\label{eq:nonlinear-sbp-density-proof}\\ &\text{s.t.}\quad \begin{cases}
            \partial_tp_t(\boldsymbol{x})=-\nabla\cdot\big(p_t(\boldsymbol{x})(\boldsymbol{f}(\boldsymbol{x},t) +\sigma_t\boldsymbol{u}(\boldsymbol{x},t))\big)+\frac{\sigma_t^2}{2}\Delta p_t(\boldsymbol{x})\\
            p_0= \pi_0, \quad p_T=\pi_T
        \end{cases}\nonumber
    \end{align}
    which enforces the marginal constraints $p_0=\pi_0$ and $p_T=\pi_T$ and Fokker-Planck equation constraint. Then, by reparameterizing $\boldsymbol{v}(\boldsymbol{x},t):=\boldsymbol{u}(\boldsymbol{x},t)+\frac{\sigma_t}{2}\nabla \log p_t(\boldsymbol{x})$, the dynamic SB problem takes an \textbf{equivalent form of a dynamic optimal transport problem} given by:
    \begin{small}
    \begin{align}
        \inf_{\boldsymbol{v}}&\mathbb{E}_{p_t}\int_0^T\bigg[\frac{1}{2}\|\boldsymbol{v}(\boldsymbol{x},t)\|^2+\frac{\sigma^2_t}{8}\|\nabla \log p_t(\boldsymbol{x})\|^2-\frac{1}{2}\langle\nabla \log p_t(\boldsymbol{x}), \boldsymbol{f}(\boldsymbol{x},t)\rangle\bigg]dt\tag{Entropy-Regularized Dynamic OT}\label{eq:entropy-regularized-dynamic-ot}\\
        &\begin{cases}
            \partial_tp_t(\boldsymbol{x})=-\nabla \cdot (p_t(\boldsymbol{f}(\boldsymbol{x},t)+\sigma_t\boldsymbol{v}(\boldsymbol{x},t)))\\
            p_0=\pi_0, \quad p_T=\pi_T
        \end{cases}\nonumber
    \end{align}
    \end{small}
    where $\partial_tp_t=-\nabla \cdot (p_t(\boldsymbol{f}_t+\sigma_t\boldsymbol{v}))$ is the continuity equation constraint. 
\end{proposition}

\textit{Proof.} In this proof, our goal is to absorb the stochastic term in the Fokker-Planck constraint into the objective functional, such that the constrained objective is in the form of the (\ref{eq:continuity-equation-remark}), which can be interpreted as a special case of the Fokker-Planck equation (Remark \ref{remark:fokker-planck-continuity}). 

\textbf{Step 1: Rewrite Fokker-Planck Equation as Continuity Equation.} 
We start by rewriting the Fokker-Planck equation by absorbing the Laplacian diffusion term into the divergence term to get the continuity equation with a newly defined drift field $\boldsymbol{v}(\boldsymbol{x},t)$:
\begin{small}
\begin{align}
    \partial_tp_t(\boldsymbol{x})&=-\nabla\cdot\big(p_t(\boldsymbol{x})(\boldsymbol{f}(\boldsymbol{x},t) +\sigma_t\boldsymbol{u}(\boldsymbol{x},t))\big)+\frac{\sigma_t^2}{2}\Delta p_t(\boldsymbol{x})\nonumber\\
    &=-\bluetext{\nabla\cdot}\big(p_t(\boldsymbol{x})(\boldsymbol{f}(\boldsymbol{x},t) +\sigma_t\boldsymbol{u}(\boldsymbol{x},t))\big)+\frac{\sigma_t^2}{2}\bluetext{\nabla\cdot}(\nabla p_t(\boldsymbol{x}))\nonumber\\
    &=-\nabla\cdot\left(p_t(\boldsymbol{x})(\boldsymbol{f}(\boldsymbol{x},t) +\sigma_t\boldsymbol{u}(\boldsymbol{x},t))-\frac{\sigma_t^2}{2}\nabla p_t(\boldsymbol{x})\right)\nonumber\\
    &=-\nabla\cdot\bigg(\bluetext{p_t(\boldsymbol{x})}\bigg(\boldsymbol{f}(\boldsymbol{x},t) +\sigma_t\boldsymbol{u}(\boldsymbol{x},t)-\frac{\sigma_t^2}{2}\bluetext{\nabla \log p_t(\boldsymbol{x})}\bigg)\bigg)\nonumber\\
    &=-\nabla\cdot\bigg(p_t(\boldsymbol{x})\bigg(\boldsymbol{f}(\boldsymbol{x},t) +\sigma_t\bluetext{\underbrace{\left(\boldsymbol{u}(\boldsymbol{x},t)-\frac{\sigma_t}{2}\nabla \log p_t(\boldsymbol{x})\right)}_{=:\boldsymbol{v}(\boldsymbol{x},t)}}\bigg)\bigg)\nonumber\\
    &=-\nabla\cdot(p_t(\boldsymbol{x})(\boldsymbol{f}(\boldsymbol{x},t)+\sigma_t\boldsymbol{v}(\boldsymbol{x},t))\tag{Continuity Equation}\label{eq:entropy-dynamic-ot-proof1}
\end{align}
\end{small}
where we define a new velocity $\boldsymbol{v}(\boldsymbol{x},t):=\boldsymbol{u}(\boldsymbol{x},t)-\frac{\sigma_t}{2}\nabla \log p_t(\boldsymbol{x})$ that satisfies a continuity equation. We rearrange to write $\boldsymbol{u}(\boldsymbol{x},t)$ in terms of the new velocity $\boldsymbol{v}(\boldsymbol{x},t)$ as:
\begin{small}
\begin{align}
    &\boldsymbol{v}(\boldsymbol{x},t)=\boldsymbol{u}(\boldsymbol{x},t)-\frac{\sigma_t}{2}\nabla \log p_t(\boldsymbol{x})\nonumber\\
    &\implies\boldsymbol{u}(\boldsymbol{x},t)=\boldsymbol{v}(\boldsymbol{x},t)+\frac{\sigma_t}{2}\nabla \log p_t(\boldsymbol{x})\label{eq:ot-sb-proof1}
\end{align}
\end{small}
Substituting this expression (\ref{eq:ot-sb-proof1}) for $\boldsymbol{u}(\boldsymbol{x},t)$ into the objective functional for the (\ref{eq:nonlinear-sbp-density-proof}), we get:
\begin{small}
\begin{align}
    \int_0^T\int_{\mathbb{R}^d}&\left[\bluetext{\frac{1}{2}\|\boldsymbol{u}(\boldsymbol{x},t)\|^2}\right]p_t(\boldsymbol{x}) d\boldsymbol{x}dt=\int_0^T\int_{\mathbb{R}^d}\left[\bluetext{\frac{1}{2}\left\|\boldsymbol{v}(\boldsymbol{x},t)+\frac{\sigma_t}{2}\nabla \log p_t(\boldsymbol{x})\right\|^2}\right]p_t(\boldsymbol{x}) d\boldsymbol{x}dt\nonumber\\
    &=\int_0^T\int_{\mathbb{R}^d}\left[\frac{1}{2}\bluetext{\left(\|\boldsymbol{v}(\boldsymbol{x},t)\|^2+2\left\langle \boldsymbol{v}(\boldsymbol{x},t),\frac{\sigma_t}{2}\nabla\log p_t(\boldsymbol{x})\right\rangle+\frac{\sigma^2_t}{4}\|\nabla\log p_t(\boldsymbol{x})\|^2\right)}\right]p_t(\boldsymbol{x}) d\boldsymbol{x}dt\nonumber\\
    &=\int_0^T\int_{\mathbb{R}^d}\bigg[\underbrace{\frac{1}{2}\|\boldsymbol{v}(\boldsymbol{x},t)\|^2}_{\text{kinetic energy}}+\underbrace{\frac{\sigma_t}{2}\left\langle \boldsymbol{v}(\boldsymbol{x},t),\nabla\log p_t(\boldsymbol{x})\right\rangle}_{\text{cross term}}+\underbrace{\frac{\sigma^2_t}{8}\|\nabla\log p_t(\boldsymbol{x})\|^2}_{\text{Fisher information}}\bigg]p_t(\boldsymbol{x}) d\boldsymbol{x}dt\label{eq:ot-sb-proof5}
\end{align}
\end{small}
which is the objective that corresponds to the (\ref{eq:continuity-equation-remark}). Since the kinetic energy $\frac{1}{2}\|\boldsymbol{v}(\boldsymbol{x},t)\|^2$ is convex in $\boldsymbol{v}$, we want to write the objective such that the drift being optimized (i.e., $\boldsymbol{v}$) appears only in the kinetic energy. To do this, we will rewrite the cross term dependent on $\boldsymbol{v}$ in the next step.

\textbf{Step 2: Expand the Cross-Term. }
Observing (\ref{eq:ot-sb-proof5}), the cross term is dependent on both of the variables being optimized $\boldsymbol{v}$ and $p_t$, which yields challenges in practice. Isolating the cross term, we aim to expand it to remove the cross dependency:
\begin{small}
\begin{align}
    \int_0^T\int_{\mathbb{R}^d}\frac{\sigma_t}{2}\left\langle \boldsymbol{v}(\boldsymbol{x},t),\nabla\log p_t(\boldsymbol{x})\right\rangle p_t(\boldsymbol{x}) d\boldsymbol{x}dt\tag{Cross Term}
\end{align}
\end{small}
Intuitively, we can consider the expression $\langle\boldsymbol{v}, \nabla \log p_t\rangle p_t=\boldsymbol{v}\cdot \nabla p_t$ as measuring the alignment between the velocity field $\boldsymbol{v}$ and the gradient of the density. In other words, it answers: \textit{is velocity field pushing probability mass toward regions of high density or away from them?} This has a direct connection to \textbf{entropy}, which measures the concentration of probability density. Let $H(p_t)=\int p_t\log p_t d\boldsymbol{x}$ denote the entropy. Since the objective integrates the cross term over time $t\in [0,T]$, we consider the \textit{change} in entropy between the terminal marginals:
\begin{small}
\begin{align}
    H(p_T)-H(p_0)&=\int_0^T\partial_tH(p_t)dt\nonumber\\
    &=\int_0^T\left(\partial_t\int_{\mathbb{R}^d} p_t\log p_t d\boldsymbol{x}\right)dt\nonumber\\
    &=\int_0^T\int_{\mathbb{R}^d}\bigg(\log p_t\partial_tp_t+p_t\partial_t\log p_t\bigg)d\boldsymbol{x}dt\nonumber\\
    &=\int_0^T\int_{\mathbb{R}^d}(1+\log p_t)\bluetext{\partial_tp_t}d\boldsymbol{x}dt\label{eq:ot-sb-proof2}
\end{align}
\end{small}
Substituting the continuity equation constraint $\partial_tp_t=-\nabla\cdot(p_t(\boldsymbol{f}+\sigma_t\boldsymbol{v}))$, we have:
\begin{small}
\begin{align}
    H(p_T)-H(p_0)
    &=\int_0^T\int_{\mathbb{R}^d}(1+\log p_t)\bluetext{\big(-\nabla\cdot(p_t(\boldsymbol{f}+\sigma_t\boldsymbol{v})\big)}d\boldsymbol{x}dt\label{eq:ot-sb-proof3}
\end{align}
\end{small}
Then, applying the integration of parts identity for divergence\footnote{which states for a scalar function $\phi$ and vector $\boldsymbol{\xi}$, the following holds: $\int\phi(-\nabla\cdot \boldsymbol{\xi})d\boldsymbol{x}=\int\langle\nabla\phi,\boldsymbol{\xi}\rangle d\boldsymbol{x}$}, we can rewrite (\ref{eq:ot-sb-proof3}) as:
\begin{small}
\begin{align}
    H(p_T)-H(p_0)&=\int_0^T\int_{\mathbb{R}^d}\bigg\langle\nabla\log p_t,\bluetext{p_t(\boldsymbol{f}+\sigma_t\boldsymbol{v})}\bigg\rangle d\boldsymbol{x}dt\nonumber\\
    &=\int_0^T\int_{\mathbb{R}^d}\bigg\langle\nabla\log p_t,(\boldsymbol{f}+\sigma_t\boldsymbol{v})\bigg\rangle \bluetext{p_t}d\boldsymbol{x}dt\nonumber\\
    &=\int_0^T\int_{\mathbb{R}^d}\big\langle\nabla\log p_t,\boldsymbol{f}\big\rangle p_td\boldsymbol{x}dt+\int_0^T\int_{\mathbb{R}^d}\sigma_t\big\langle\nabla\log p_t,\boldsymbol{v}\big\rangle p_td\boldsymbol{x}dt\nonumber\\
    \implies & \int_0^T\int_{\mathbb{R}^d}\bluetext{\frac{\sigma_t}{2}}\big\langle\nabla\log p_t,\boldsymbol{v}\big\rangle p_td\boldsymbol{x}dt=\underbrace{\bluetext{\frac{1}{2}}(H(p_T)-H(p_0))}_{\text{constant}} -\int_0^T\int_{\mathbb{R}^d}\bluetext{\frac{1}{2}}\big\langle\nabla\log p_t,\boldsymbol{f}\big\rangle p_t d\boldsymbol{x}dt\label{eq:ot-sb-proof4}
\end{align}
\end{small}
In the (\ref{eq:nonlinear-sbp-density-proof}), the marginal distributions $p_0=\pi_0$ and $p_T=\pi_T$ are fixed, so the entropy difference term $H(p_T)-H(p_0)$ is a constant and can be dropped in the objective. Therefore, substituting (\ref{eq:ot-sb-proof3}) into the objective functional, we have:
\begin{align}
    \boxed{\inf_{(p_t, \boldsymbol{v}, g)}\int_0^T\int_{\mathbb{R}^d}\bigg[\frac{1}{2}\|\boldsymbol{v}(\boldsymbol{x},t)\|^2+\frac{\sigma^2_t}{8}\|\nabla \log p_t(\boldsymbol{x})\|^2\bluetext{-\frac{1}{2}\langle\nabla \log p_t(\boldsymbol{x}), \boldsymbol{f}(\boldsymbol{x},t)\rangle}\bigg]p_t(\boldsymbol{x})d\boldsymbol{x}dt}\label{eq:ot-sb-proof6}
\end{align}
Equivalently, we can write (\ref{eq:ot-sb-proof6}) as an expectation over the marginal density $p_t$ corresponding to the velocity field $\boldsymbol{v}_t$ given by:
\begin{align}
    \inf_{\boldsymbol{v}}&\mathbb{E}_{p_t}\int_0^T\bigg[\frac{1}{2}\|\boldsymbol{v}(\boldsymbol{x},t)\|^2+\frac{\sigma^2_t}{8}\|\nabla \log p_t(\boldsymbol{x})\|^2-\frac{1}{2}\langle\nabla \log p_t(\boldsymbol{x}), \boldsymbol{f}(\boldsymbol{x},t)\rangle\bigg]dt\\
    &\begin{cases}
        \partial_tp_t(\boldsymbol{x})=-\nabla \cdot (p_t(\boldsymbol{f}(\boldsymbol{x},t)+\sigma_t\boldsymbol{v}(\boldsymbol{x},t)))\\
        p_0=\pi_0, \quad p_T=\pi_T
    \end{cases}
\end{align}
which concludes our derivation of the dynamical OT form of the dynamic SB problem. \hfill $\square$

This derivation yields an objective with three distinct terms that capture both the transport and diffusion terms in the original Fokker-Planck constraint. \textbf{Intuitively, each term can be interpreted as follows:}
\begin{enumerate}
\item[(i)] The \textbf{kinetic energy term $\frac{1}{2}\|\boldsymbol{v}(\boldsymbol{x},t)\|^2$} measures the deterministic cost of transporting probability mass along the velocity field $\boldsymbol{v}(\boldsymbol{x},t)$. This term is exactly the kinetic energy appearing in the (\ref{eq:dynamic-ot-prob}), where the optimal velocity minimizes the kinetic energy required to move mass on the deterministic flow between the marginal distributions.
\item[(ii)] The \textbf{Fisher information term $\frac{\sigma^2_t}{4}\|\nabla \log p_t(\boldsymbol{x})\|^2$} measures the sharpness of the marginal distribution. Smooth distributions yield low Fisher information, while sharp and concentrated distributions yield high Fisher information. This term appears from the diffusion term in the Fokker-Planck equation as the diffusion in the SDE acts to smooth out the distribution.
\item[(iii)] The \textbf{cross term or drift interaction $\frac{1}{2}\langle\nabla \log p_t(\boldsymbol{x}), \boldsymbol{f}(\boldsymbol{x},t)\rangle$} measures the interaction between the reference drift $\boldsymbol{f}(\boldsymbol{x},t)$ and the density evolution $\nabla \log p_t$. This term reflects how the prior dynamics influence the evolution of the distribution by either aligning with or opposing the natural directions of increasing probability mass.
\end{enumerate}

This result shows that the (\ref{eq:dynamic-sb-problem}) problem can be explicitly rewritten as an entropy-regularized version of the (\ref{eq:dynamic-ot-prob}), where the optimization over stochastic path measures $\mathbb{P}^u$ reduces to an optimization over deterministic density flows $p_t$ and velocity fields $\boldsymbol{v}(\boldsymbol{x},t)$. In the limit of vanishing diffusion ($\sigma_t\to 0$), the entropy regularization from the Fisher information and cross-term vanish, and the formulation recovers the classic (\ref{eq:dynamic-ot-prob}).

\subsection{Closing Remarks for Section \ref{sec:dynamic-sb}}
In this section, we took the crucial step of lifting the static formulation of the Schrödinger bridge (SB) problem to the space of continuous-time \textbf{path measures}. Starting from the dynamic optimal transport (OT) problem, we show that the static OT problem with a quadratic cost function can be written in an equivalent \textit{dynamic} form, which aims to find the optimal probability flow that smoothly transports mass between the prescribed marginals via a velocity field over a continuous time interval, yielding the straight probability flow between distributions. Given that most real-world systems do not evolve in straight lines between distributions, we extend the entropic OT problem to the space of \textbf{stochastic processes}, where path measures are determined by both a deterministic drift and random fluctuations in the form of Brownian motion, and entropy regularization is performed with a reference stochastic process.

To understand the dynamic SB problem, we first introduce the governing theory of path measures and Itô processes, from Itô's formula, which describes the evolution of functions evaluated on stochastic processes, to the Fokker-Planck and Feynman-Kac equations, which characterize how probability densities and functions evolve over stochastic processes through partial differential equations (PDEs). Just like the static case, we introduce the concept of relative entropy in the space of path measures, which can be expressed explicitly in terms of the difference between the drifts of two stochastic differential equations using Girsanov's theorem. 

Building on this framework, we analyzed the dynamic SB problem by expressing the KL minimization over path measures as a minimization of the \textbf{kinetic energy of a control drift}, corresponding to the minimal perturbation required to steer the reference SDE so that its marginals match the prescribed constraints. We leverage Lagrange multipliers to derive the optimality conditions for the dynamic SB problem, which result in a pair of \textbf{non-linear PDEs} describing the forward and backward dynamics of the optimal solution $(\psi_t, p^\star_t)$. Finally, we introduce the Hopf-Cole transform as a method of linearizing the non-linear PDEs with a simple change-of-variables, revealing that the optimal SB dynamics can be factorized into the product of a pair of forward and backward potentials $p_t^\star=\varphi_t\hat\varphi_t$ which solve a pair of linear PDEs. 

While we have analyzed the dynamic SB problem through the lenses of minimizing the KL divergence between stochastic path measures, it admits an alternative interpretation as an \textbf{optimal control problem}, where the goal is to determine the control drift that minimizes the expected future cost of steering a stochastic system toward a desired terminal distribution. This is precisely the idea of \textbf{stochastic optimal control} (SOC), which leverages Bellman's Principle of Optimality to define an optimal control as the minimizer to the expected \textit{cost-to-go} from an intermediate state to all possible terminal states under an SDE. In the next section, we explore this perspective in detail, first introducing the theory of SOC and its properties before explicitly reformulating the dynamic SB problem as an SOC problem.

\newpage
\section{Schrödinger Bridge Problem as Optimal Control}
\label{sec:sb-optimal-control}
In this section, we reformulate the Schrödinger bridge (SB) problem through the lens of \textbf{stochastic optimal control (SOC)} theory, providing a dynamic and decision-theoretic perspective on entropy-regularized transport. Rather than directly optimizing over path measures, this view interprets the SB problem as learning an optimal control that steers a reference stochastic process between prescribed marginals while minimizing a control cost. 

We begin in Section \ref{subsec:stochastic-optimal-control} by introducing the general SOC framework, including Bellman’s Principle of Optimality and the value function that characterizes optimal policies under terminal constraints. In Section \ref{subsec:sb-soc}, we establish the connection between SB and SOC by showing that the optimal bridge corresponds to an \textbf{optimal control drift} applied to the reference dynamics. Finally, in Section \ref{subsec:soc-objectives}, we develop practical training objectives that enable solving the resulting control problem in high-dimensional settings. Since we work with multiple path measures with different controls, we denote the stochastic process generated with a \textit{specific} control $\boldsymbol{u}$ with the superscripted notation $\boldsymbol{X}^u_{0:T}=(\boldsymbol{X}_t^u)_{t\in [0,T]}$ to clearly denote the path measure $\mathbb{P}^u$ that the process was generated under.

\subsection{Stochastic Optimal Control}
\label{subsec:stochastic-optimal-control}
While the Schrödinger bridge problem aims to optimize the intermediate bridge between fixed endpoints, it is natural to consider an alternative variational perspective where given a particle sampled from the initial distribution $\boldsymbol{X}_0\sim \pi_0$, we want to \textbf{optimize its path such that it reaches a state in the target distribution while minimially deviating from the reference SDE.} This is the key idea behind \boldtext{stochastic optimal control (SOC)} theory \citep{nusken2021solving}, which seeks an \textit{optimal control drift} $\boldsymbol{u}^\star(\boldsymbol{x},t) $ that corrects the particles trajectory such that it takes the path of \textbf{minimal cost} toward the target distribution $\pi_T$. 
\begin{definition}[Stochastic Optimal Control (SOC) Objective]\label{def:soc-objective}
    Given an running cost $c(\boldsymbol{x},t):\mathbb{R}^d\times[0,T]\to \mathbb{R}$ and a terminal cost $\Phi(\boldsymbol{x}):\mathbb{R}^d\to \mathbb{R}$, we consider the following \textbf{stochastic optimal control} (SOC) objective:
    \begin{align}
        &\inf_{\boldsymbol{u}}\mathbb{E}_{\boldsymbol{X}^u_{0:T}\sim \mathbb{P}^u}\left[\int_0^T\left(\frac{1}{2}\|\boldsymbol{u}(\boldsymbol{X}^u_t,t)\|^2+c(\boldsymbol{X}^u_t,t)\right)dt+\Phi(\boldsymbol{X}^u_T)\right]\tag{SOC Objective}\label{eq:soc-objective}\\
        &\text{s.t.}\quad d\boldsymbol{X}^u_t=\left(\boldsymbol{f}(\boldsymbol{X}^u_t,t)+\sigma_t\boldsymbol{u}(\boldsymbol{X}^u_t,t)\right)dt+\sigma_td\boldsymbol{B}_t, \quad \boldsymbol{X}^u_0\sim \pi_0\nonumber
    \end{align}
    where $\boldsymbol{u}(\boldsymbol{x},t)$ is the control drift which produces the path measure $\mathbb{P}^u$, $\boldsymbol{f}(\boldsymbol{x},t)$ is the drift of the reference process $\mathbb{Q}$, $\sigma_t$ is the diffusion coefficient, and $\boldsymbol{B}_t$ is $d$-dimensional Brownain motion.
\end{definition}

Under this objective, we can define the \boldtext{cost functional} $J(\boldsymbol{x},t;\boldsymbol{u})$ being optimized as the \textit{cost-to-go} from any fixed point $(\boldsymbol{x}, t)\in \mathbb{R}^d\times [0,T]$ at time $t$ under the control $\boldsymbol{u}$ as the expected running cost and terminal cost of integrating the controlled SDE from $\boldsymbol{X}^u_t=\boldsymbol{x}$ over $s\in [t,T]$.
\begin{small}
\begin{align}
    J(\boldsymbol{x},t;\boldsymbol{u}):=\mathbb{E}_{\boldsymbol{X}^u_{t:T}\sim \mathbb{P}^u}\left[\int_t^T\left(\frac{1}{2}\|\boldsymbol{u}(\boldsymbol{X}^u_s,s)\|^2+c(\boldsymbol{X}^u_s,s)\right)ds+\Phi(\boldsymbol{X}^u_T)\bigg|\boldsymbol{X}^u_t=\boldsymbol{x}\right]\tag{Cost Functional}\label{eq:soc-cost-functional}
\end{align}
\end{small}
Given the cost-to-go for an arbitrary control $\boldsymbol{u}$, we define the \boldtext{value function} $V_t(\boldsymbol{x}):\mathbb{R}^d\to \mathbb{R}$ as the \textit{optimal cost-to-go} obtained with the optimal control $\boldsymbol{u}^\star$.
\begin{align}
    V_t(\boldsymbol{x}):=J^\star(\boldsymbol{x},t;\boldsymbol{u}^\star):=\inf_{\boldsymbol{u}}J(\boldsymbol{x},t; \boldsymbol{u})\tag{Value Function}\label{eq:soc-value-func}
\end{align}
which solves the \textbf{Hamilton-Jacobi-Bellman (HJB) equation}, similarly to the Lagrange multiplier $\psi_t(\boldsymbol{x})$ from Section \ref{subsec:nonlinear-sbp}. Before defining the HJB for the value function, we first define \boldtext{Bellman's Principle of Optimality}, also referred to as the \textbf{dynamic programming principle}, which provides an intuitive and rigorous foundation for stochastic optimal control theory. 

\begin{definition}[Bellman's Principle of Optimality]\label{def:bellman-principle}
    Let $V_t(\boldsymbol{x})$ denote the value function or optimal cost-to-go of a stochastic control problem starting from state $\boldsymbol{x}$ at time $t$. Then, the optimal control satisfies \textbf{Bellman’s principle of optimality},  which states that for all intermediate time steps $t\leq \tau \leq T$ the optimal cost-to-go is equal to the cost incurred over the interval $[t,\tau]$ and the future cost-to-go $V_\tau(\boldsymbol{X}_\tau)$ starting from $\boldsymbol{X}_\tau$ over time $[\tau,T]$, which yields the following \textbf{expanded expression or the value function}:
    \begin{small}
    \begin{align}
        V_t(\boldsymbol{x})&=\inf_{\boldsymbol{u}}\mathbb{E}_{\boldsymbol{X}^u_{t:T}\sim \mathbb{P}^u}\left[\left(\bluetext{\int_t^{\tau}}+\pinktext{\int_\tau^T}\right)\left(\frac{1}{2}\|\boldsymbol{u}(\boldsymbol{X}^u_s,s)\|^2+c(\boldsymbol{X}^u_s,s)\right)ds+\Phi(\boldsymbol{X}^u_T)\bigg|\boldsymbol{X}^u_t=\boldsymbol{x}\right]\nonumber\\
        &= \inf_{\boldsymbol{u}}\mathbb{E}_{\boldsymbol{X}^u_{t:T}\sim \mathbb{P}^u}\left[\bluetext{\int_t^{\tau}\left(\frac{1}{2}\|\boldsymbol{u}(\boldsymbol{X}^u_s,s)\|^2+c(\boldsymbol{X}^u_s,s)\right)ds}+\pinktext{V_{\tau}(\boldsymbol{X}^u_{\tau})}\bigg|\boldsymbol{X}^u_t=\boldsymbol{x}\right]\tag{Bellman's Principle of Optimality}\label{eq:bellman-principle}
    \end{align}
    \end{small}
\end{definition}

Intuitively, this means that in an optimally controlled process, regardless of the initial state $\boldsymbol{x}$ and changes over $[t,\tau]$, the remaining controlled process from $\boldsymbol{X}_\tau$ follows the optimal control law for the state resulting from the \textit{first decision}. Taking the infinitesimal limit when $\tau\to t$ and $\tau-t\to 0$, this property yields the \boldtext{Hamilton–Jacobi–Bellman (HJB) equations} \citep{bardi1997optimal}, which characterizes the evolution of the value function.

\begin{lemma}[Hamilton-Jacobi-Bellman (HJB) Equations]\label{lemma:soc-hjb-eq}
    Given the infinitesimal generator $\mathcal{A}_t$ of the \textbf{uncontrolled SDE} $d\boldsymbol{X}_t=\boldsymbol{f}(\boldsymbol{X}_t,t)dt+\sigma_td\boldsymbol{B}_t$ that acts on the value function $V_t(\boldsymbol{x})$ as: 
    \begin{align}
        (\mathcal{A}_tV_t)(\boldsymbol{x}) :=\langle \boldsymbol{f}(\boldsymbol{x},t),  \nabla V_t(\boldsymbol{x})\rangle +\frac{\sigma_t^2}{2}\Delta V_t(\boldsymbol{x})
    \end{align}
    Then, $V_t(\boldsymbol{x})$ solves the \textbf{Hamilton-Jacobi-Bellman} equation defined as:
    \begin{align}
        \partial_tV_t(\boldsymbol{x})=-(\mathcal{A}_tV_t)(\boldsymbol{x})+\frac{\sigma_t^2}{2}\|\nabla V_t(\boldsymbol{x})\|^2-c(\boldsymbol{x},t), \quad V_T(\boldsymbol{x})=\Phi(\boldsymbol{x})\tag{Value HJB Equation}\label{eq:soc-hjb-eq}
    \end{align}
\end{lemma}

\textit{Proof.} The proof follows from (\ref{eq:bellman-principle}), where we define $\tau:=t+\Delta t$ as a small time step $\Delta t$ to get the following expanded expression for the value function:
\begin{small}
\begin{align}
    V_t(\boldsymbol{x})&=\inf_{\boldsymbol{u}}\mathbb{E}_{\boldsymbol{X}^u_{t:T}\sim \mathbb{P}^u}\left[\left(\bluetext{\int_t^{t+\Delta t}}+\pinktext{\int_{t+\Delta t}^T}\right)\left(\frac{1}{2}\|\boldsymbol{u}(\boldsymbol{X}^u_s,s)\|^2+c(\boldsymbol{X}^u_s,s)\right)ds+\Phi(\boldsymbol{X}^u_T)\bigg|\boldsymbol{X}^u_t=\boldsymbol{x}\right]\nonumber\\
    &= \inf_{\boldsymbol{u}}\mathbb{E}_{\boldsymbol{X}^u_{t:T}\sim \mathbb{P}^u}\bigg[\bluetext{\int_t^{t+\Delta t}\left(\frac{1}{2}\|\boldsymbol{u}(\boldsymbol{X}^u_s,s)\|^2+c(\boldsymbol{X}^u_s,s)\right)ds}+\underbrace{\pinktext{\int_{t+\Delta t}^T\left(\frac{1}{2}\|\boldsymbol{u}(\boldsymbol{X}^u_s,s)\|^2+c(\boldsymbol{X}^u_s,s)\right)ds+\Phi(\boldsymbol{X}_T)}}_{=:V_{t+\Delta t}(\boldsymbol{X}^u_{t+\Delta t})}\bigg|\boldsymbol{X}^u_t=\boldsymbol{x}\bigg]\nonumber\\
    &= \inf_{\boldsymbol{u}}\mathbb{E}_{\boldsymbol{X}^u_{t:T}\sim \mathbb{P}^u}\left[\bluetext{\int_t^{t+\Delta t}\left(\frac{1}{2}\|\boldsymbol{u}(\boldsymbol{X}^u_s,s)\|^2+c(\boldsymbol{X}^u_s,s)\right)ds}+V_{t+\Delta t}(\boldsymbol{X}^u_{t+\Delta t})\bigg|\boldsymbol{X}^u_t=\boldsymbol{x}\right]\nonumber\\
    &= \inf_{\boldsymbol{u}}\bluetext{\left[\left(\frac{1}{2}\|\boldsymbol{u}(\boldsymbol{X}^u_t,t)\|^2+c(\boldsymbol{X}^u_t,t)\right)\Delta t\right]}+\mathcal{O}(\Delta t^2)+\inf_{\boldsymbol{u}}\pinktext{\mathbb{E}_{\boldsymbol{X}^u_{t:T}\sim \mathbb{P}^u}\left[V_{t+\Delta t}(\boldsymbol{X}^u_{t+\Delta t})\big|\boldsymbol{X}^u_t=\boldsymbol{x}\right]}\label{eq:hjb-proof2}
\end{align}
\end{small}
Applying (\ref{eq:controlled-ito-formula-generator}), the change in the value function over the interval $[t, t+\Delta t]$ expands into:
\begin{small}
\begin{align}
    &V_{t+\Delta t}(\boldsymbol{X}^u_{t+\Delta t})-V_t(\boldsymbol{X}^u_t)=\int_t^{t+\Delta t}\bluetext{\underbrace{dV_s(\boldsymbol{X}^u_s)}_{\text{Itô's formula}}}\nonumber\\
    &\quad\quad\overset{(\ref{thm:ito-formula})}{=} \int_t^{t+\Delta t}(\partial_s+\mathcal{A}^u_s)V_s(\boldsymbol{X}^u_s)ds+\underbrace{\int_0^{t+\Delta t}\sigma_s\nabla V_s(\boldsymbol{X}^u_s)^\top d\boldsymbol{B}_s}_{\text{Itô integral vanishes under expectation}}
\end{align}
\end{small}
Moving $V_t(\boldsymbol{X}^u_t)$ to the right-hand side and applying the conditional expectation $\mathbb{E}[\cdot|\boldsymbol{X}^u_t=\boldsymbol{x}]$ to both sides, we get: 
\begin{small}
\begin{align}
    \mathbb{E}[V_{t+\Delta t}(\boldsymbol{X}^u_{t+\Delta t})|\boldsymbol{X}^u_t=\boldsymbol{x}]&=\mathbb{E}\left[V_t(\boldsymbol{X}^u_t)+\int_t^{t+\Delta t}(\partial_s+\mathcal{A}^u_s)V_s(\boldsymbol{X}^u_s)ds\bigg|\boldsymbol{X}_t=\boldsymbol{x}\right]\nonumber\\
    &=V_t(\boldsymbol{x})+(\partial_t+\mathcal{A}^u_t)V_t(\boldsymbol{x})\Delta t+o(\Delta t)\label{eq:hjb-proof1}
\end{align}
\end{small}
Substituting this expression (\ref{eq:hjb-proof1}) into (\ref{eq:hjb-proof2}), we obtain the equation:
\begin{small}
\begin{align}
    V_t(\boldsymbol{x})&=\inf_{\boldsymbol{u}}\bigg[\bluetext{\bigg(\frac{1}{2}\|\boldsymbol{u}(\boldsymbol{X}^u_t,t)\|^2+c(\boldsymbol{X}^u_t,t)\bigg)\Delta t }+\pinktext{V_t(\boldsymbol{x})+(\partial_t+\mathcal{A}^u_t)V_t(\boldsymbol{x})\Delta t}\bigg]+\mathcal{O}(\Delta t^2)
\end{align}
\end{small}
Now, we can subtract $V_t(\boldsymbol{x})$ from both sides, divide by $\Delta t$ and take the continuous time limit $\Delta t\to 0$ to get the expression:
\begin{small}
\begin{align}
    0&=\inf_{\boldsymbol{u}}\left[\frac{1}{2}\|\boldsymbol{u}(\boldsymbol{x},t)\|^2+c(\boldsymbol{x},t)+(\partial_t+\mathcal{A}^u_t)V_t(\boldsymbol{x})\right]\label{eq:hjb-proof3}
\end{align}
\end{small}
Completing the square for the $\boldsymbol{u}$-dependent terms, we get the form of the minimizer:
\begin{small}
\begin{align}
    \inf_{\boldsymbol{u}}\bigg[\frac{1}{2}\|\boldsymbol{u}(\boldsymbol{x},t)\|^2+&\langle \nabla V_t(\boldsymbol{x}), \sigma_t\boldsymbol{u}(\boldsymbol{x},t)\rangle\bigg]=\inf_{\boldsymbol{u}}\left[\frac{1}{2}\|\boldsymbol{u}(\boldsymbol{x},t)\|^2+\langle \nabla V_t(\boldsymbol{x}), \sigma_t\boldsymbol{u}(\boldsymbol{x},t)\rangle\bluetext{+\frac{\sigma^2_t}{2}\|\nabla V_t(\boldsymbol{x})\|^2-\frac{\sigma^2_t}{2}\|\nabla V_t(\boldsymbol{x})\|^2}\right]\nonumber\\
    &=\inf_{\boldsymbol{u}}\left[\frac{1}{2}\|\boldsymbol{u}(\boldsymbol{x},t)+\sigma_t\nabla V_t(\boldsymbol{x})\|^2-\frac{\sigma^2_t}{2}\|\nabla V_t(\boldsymbol{x})\|^2\right]\nonumber\\
    &\implies \boxed{\boldsymbol{u}^\star(\boldsymbol{x},t)=-\sigma_t\nabla V_t(\boldsymbol{x})}
\end{align}
\end{small}
Plugging the expression for the optimal $\boldsymbol{u}^\star$ into (\ref{eq:hjb-proof3}) and rewriting the controlled generator in terms of the uncontrolled generator with (\ref{eq:generator-control-to-uncontrolled}), we recover the HJB equation: 
\begin{align}
    0&=\frac{1}{2}\|\bluetext{\boldsymbol{u}^\star(\boldsymbol{x},t)}\|^2+c(\boldsymbol{x},t)+\bluetext{(\partial_t+\mathcal{A}^{u^\star}_t)V_t(\boldsymbol{x})}\nonumber\\
    0&=\frac{1}{2}\|\bluetext{\boldsymbol{u}^\star(\boldsymbol{x},t)}\|^2+c(\boldsymbol{x},t)+\bluetext{\partial_tV_t(\boldsymbol{x})+(\mathcal{A}_tV_t)(\boldsymbol{x})+\langle \nabla V_t(\boldsymbol{x}), \sigma_t\bluetext{\boldsymbol{u}^\star(\boldsymbol{x},t)}\rangle}\nonumber\\
    0&=\bluetext{\frac{\sigma_t^2}{2}\|\nabla V_t(\boldsymbol{x})\|^2}+c(\boldsymbol{x},t)+\partial_tV_t(\boldsymbol{x})+(\mathcal{A}_tV_t)(\boldsymbol{x})\bluetext{-\sigma_t^2\|\nabla V_t(\boldsymbol{x})\|^2}\nonumber\\
    \partial_tV_t(\boldsymbol{x})&=-(\mathcal{A}_tV_t)(\boldsymbol{x})+\frac{\sigma_t^2}{2}\|\nabla V_t(\boldsymbol{x})\|^2-c(\boldsymbol{x},t)\tag{Hamilton-Jacobi-Bellman Equation}
\end{align}
which is the HJB equation stated in (\ref{eq:soc-hjb-eq}). Intuitively, this states that the time evolution of the value function $\partial_tV_t(\boldsymbol{x})$ is equal to the instantaneous change in value induced by the \textbf{uncontrolled dynamics} $-(\mathcal{A}_tV_t)(\boldsymbol{x})$, the cost that arises from tilting the dynamics with the optimal control $\frac{\sigma_t^2}{2}\|V_t(\boldsymbol{x})\|^2$, and the immediate running cost $-c(\boldsymbol{x},t)$. \hfill $\square$

Recall from Section \ref{subsec:nonlinear-sbp} that the optimality conditions for (\ref{eq:dynamic-sb-problem}) with arbitrary reference dynamics are defined by the pair $(\psi_t, p^\star_t)$, where $\psi_t$ is the Lagrange multiplier of the constrained optimization problem and $p^\star_t$ is the optimal marginal density evolution over $t\in[0,T]$. Then, we showed that $(\psi_t, p_t^\star)$ that characterize the dynamic SB problem satisfy a system of \textbf{non-linear PDEs} called the (\ref{eq:prop-hjb-fpe-system}) which can then be transformed into a system of \textbf{linear PDEs} (\ref{eq:hopf-cole-system}) via the (\ref{eq:hopf-cole-change-var}). Leveraging this idea, we can also write the SOC problem described by the (\ref{eq:soc-hjb-eq}) into a \textbf{linear PDE} via the \boldtext{Hopf-Cole transform}.

\begin{corollary}[Hopf-Cole Transform for Value Function]\label{cor:hopf-cole-value}
    Let $V_t(\boldsymbol{x})$ be the value function that defines the solution to the (\ref{eq:soc-objective}) and satisfies the (\ref{eq:soc-hjb-eq}). Then, we define the change of variables:
    \begin{small}
    \begin{align}
        V_t(\boldsymbol{x})=-\log \varphi_t(\boldsymbol{x}) \iff \varphi_t(\boldsymbol{x}) =e^{-V_t(\boldsymbol{x})}\label{eq:hopf-cole-value-variable}
    \end{align}
    \end{small}
    where $\varphi_t(\boldsymbol{x})$ satisfies the \textbf{linear PDE} defined as:
    \begin{small}
    \begin{align}
        \partial_t\varphi_t(\boldsymbol{x})+\langle\boldsymbol{f}(\boldsymbol{x},t),\nabla \varphi_t(\boldsymbol{x})\rangle+\frac{\sigma_t^2}{2}\Delta\varphi_t(\boldsymbol{x})-c(\boldsymbol{x},t)\varphi_t(\boldsymbol{x})=0, \quad \varphi_T(\boldsymbol{x})=e^{-\Phi(\boldsymbol{x})}\tag{Hopf-Cole Value PDE}\label{eq:hopf-cole-value-pde}
    \end{align}
    \end{small}
\end{corollary}

\textit{Proof. } First, we recall the (\ref{eq:soc-hjb-eq}) given by:
\begin{small}
\begin{align}
    &\partial_tV_t(\boldsymbol{x})+\bluetext{(\mathcal{A}_tV_t)(\boldsymbol{x})}-\frac{\sigma_t^2}{2}\|\nabla V_t(\boldsymbol{x})\|^2+c(\boldsymbol{x},t)=0, \quad V_T(\boldsymbol{x})=\Phi(\boldsymbol{x})\label{eq:proof-value-hjb}\\
    &\text{where}\quad \bluetext{(\mathcal{A}_tV_t)(\boldsymbol{x})}= \langle\boldsymbol{f}(\boldsymbol{x},t), \nabla V_t(\boldsymbol{x})\rangle+\frac{\sigma_t^2}{2}\Delta V_t(\boldsymbol{x})
\end{align}
\end{small}
Following the analogous steps from Section \ref{subsec:hopf-cole-transform}, we express each term containing $V_t$ in terms of $\varphi$:
\begin{enumerate}
    \item[(i)] \textbf{Time Derivative:} $\partial_tV_t=-\partial_t\log \varphi_t=-\frac{\partial_t\varphi_t}{\varphi_t}$.
    \item[(ii)] \textbf{Gradient Term:} $\nabla V_t=-\nabla \log \varphi_t=-\frac{\nabla \varphi_t}{\varphi_t}\implies \frac{\sigma_t^2}{2}\|\nabla V_t\|^2=\frac{\sigma_t^2}{2}\frac{\|\nabla \varphi_t\|^2}{\varphi_t^2}$.
    \item[(iii)] \textbf{Generator Term:} First, we expand the Laplacian term in $(\mathcal{A}_tV_t)$ as: 
    \begin{small}
    \begin{align}
        \Delta V_t&=\nabla\cdot\nabla (-\log\varphi_t)=-\nabla\cdot\left(\frac{\nabla \varphi_t}{\varphi_t}\right)=-\nabla\cdot(\varphi_t^{-1}\nabla \varphi_t)\nonumber\\
        &=\underbrace{-\nabla\cdot (\varphi^{-1})\cdot \nabla \varphi_t+\varphi_t^{-1}(\nabla \cdot \nabla \varphi_t)}_{\text{product rule of divergences}}=-\frac{\Delta \varphi_t}{\varphi_t}+\frac{\|\nabla \varphi_t\|^2}{\varphi_t^2}
    \end{align}
    \end{small}
    Then, the generator becomes:
    \begin{small}
    \begin{align}
        (\mathcal{A}_tV_t)&=\bigg\langle \boldsymbol{f}, \bluetext{-\frac{\nabla \varphi_t}{\varphi_t}}\bigg\rangle+\frac{\sigma_t^2}{2}\left(\bluetext{-\frac{\Delta \varphi_t}{\varphi_t}+\frac{\|\nabla \varphi_t\|^2}{\varphi_t^2}}\right)=-\frac{\langle\boldsymbol{f}, \nabla \varphi_t\rangle}{\varphi_t}-\frac{\sigma_t^2}{2}\frac{\Delta\varphi_t}{\varphi_t}+\frac{\sigma_t^2}{2}\frac{\|\nabla \varphi_t\|^2}{\varphi_t^2}
    \end{align}
    \end{small}
\end{enumerate}
Substituting each term back into the HJB equation (\ref{eq:proof-value-hjb}) and cancelling terms, we get:
\begin{small}
\begin{align}
    \underbrace{-\frac{\partial_t\varphi_t}{\varphi_t}}_{\partial_tV_t(\boldsymbol{x})}\underbrace{-\frac{\langle\boldsymbol{f}, \nabla \varphi_t\rangle}{\varphi_t}-\frac{\sigma_t^2}{2}\frac{\Delta\varphi_t}{\varphi_t}+\bluetext{\frac{\sigma_t^2}{2}\frac{\|\nabla \varphi_t\|^2}{\varphi_t^2}}}_{(\mathcal{A}_tV_t)(\boldsymbol{x})}-\underbrace{\bluetext{\frac{\sigma_t^2}{2}\frac{\|\nabla \varphi_t\|^2}{\varphi_t^2}}}_{\frac{\sigma_t^2}{2}\|\nabla V_t(\boldsymbol{x})\|^2}+c(\boldsymbol{x},t)&=0\nonumber\\
    -\frac{\partial_t\varphi_t}{\varphi_t}-\frac{\langle\boldsymbol{f}, \nabla \varphi_t\rangle}{\varphi_t}-\frac{\sigma_t^2}{2}\frac{\Delta\varphi_t}{\varphi_t}+c(\boldsymbol{x},t)&=0\nonumber\\
    \partial_t\varphi_t+\langle\boldsymbol{f}, \nabla \varphi_t\rangle+\frac{\sigma_t^2}{2}\Delta\varphi_t-c(\boldsymbol{x},t)\varphi_t&=0\label{eq:proof-hjb-soc-1}
\end{align}
\end{small}
where the final line follows from multiplying both sides by $-\varphi_t$. Given the terminal condition $V_T(\boldsymbol{x})=\Phi(\boldsymbol{x})$ defined in (\ref{eq:proof-value-hjb}), we apply the change of variables to get $\varphi_T(\boldsymbol{x})=e^{-V_T(\boldsymbol{x})}=e^{-\Phi(\boldsymbol{x})}$. Putting this together with (\ref{eq:proof-hjb-soc-1}), we have that $\varphi_t$ satisfies the transformed \textbf{linear PDE} given by:
\begin{small}
\begin{align}
    \boxed{\partial_t\varphi_t(\boldsymbol{x})+\langle\boldsymbol{f}(\boldsymbol{x},t), \nabla \varphi_t(\boldsymbol{x})\rangle+\frac{\sigma_t^2}{2}\Delta\varphi_t(\boldsymbol{x})-c(\boldsymbol{x},t)\varphi_t(\boldsymbol{x})=0, \quad \varphi_T(\boldsymbol{x})=e^{-\Phi(\boldsymbol{x})}}\tag{Hopf-Cole Value PDE}
\end{align}
\end{small}
which concludes our derivation of the Hopf-Cole transform for the value function that solves (\ref{eq:soc-objective}).\hfill $\square$

It is worth noticing that the Hopf-Cole transform that we performed in Section \ref{subsec:hopf-cole-transform} resulted in \textit{two linear PDEs} that describe the evolution of the forward and backward Schrödinger potentials $(\varphi_t,\hat\varphi_t)$, whereas we only derived a single linear PDE for the SOC problem. This is because while the (\ref{eq:dynamic-sb-problem}) has two terminal constraints at both marginals $p_0=\pi_0$ and $p_T=\pi_T$, the (\ref{eq:soc-objective}) only has one terminal constraint for the value function at $V_T(\boldsymbol{x})=\Phi(\boldsymbol{x})$, resulting in a single potential change-of-variables $\varphi_t(\boldsymbol{x})=e^{-V_t(\boldsymbol{x})}$ that satisfies the backward (\ref{eq:hopf-cole-value-pde}) with terminal constraint $\varphi_T(\boldsymbol{x})=e^{-\Phi(\boldsymbol{x})}$. Despite this difference in the two formulations, we make the following remark, which we explore further in Section \ref{subsec:sb-soc}.

\begin{remark}[]
    Although the Schrödinger bridge problem imposes constraints on both the initial and terminal distributions, it can still be reformulated as an SOC problem. The key observation is that the initial distribution can be treated as the \textbf{starting distribution} of the controlled process, while the terminal constraint can be enforced through an appropriate terminal cost that penalizes deviations from the desired terminal marginal.
\end{remark}

We have shown that the value function $V_t$  can be analyzed in the form of a linear PDE. We now present an alternative perspective by analyzing the evolution of $V_t$ along stochastic trajectories generated under the uncontrolled reference measure $\mathbb{Q}$. In particular, we show that the stochastic process $\boldsymbol{X}_{0:T}$ together with the value function evaluated along the trajectory satisfies a system of \textbf{forward-backward stochastic differential equations (FBSDEs)}. In this formulation, the forward equation describes the evolution of the uncontrolled diffusion process, while the backward equation governs the dynamics of the value function along the stochastic path.

\begin{proposition}[Forward-Backward Stochastic Differential Equations]\label{prop:soc-fbsdes}
    The optimal controlled process can be defined with a pair of forward and backward SDEs (FBSDEs) which define the evolution of the uncontrolled process $\boldsymbol{X}_{0:T}$ and the value function $V_t(\boldsymbol{X}_t)$, given by:
    \begin{small}
    \begin{align}
    \begin{cases}
        d\boldsymbol{X}_t=\boldsymbol{f}(\boldsymbol{X}_t,t)dt+\sigma_td\boldsymbol{B}_t\\
        dV_t(\boldsymbol{X}_t)=\left(\frac{\sigma^2_t}{2}\|\nabla V_t(\boldsymbol{X}_t)\|^2-c(\boldsymbol{X}_t,t)\right)dt+\nabla V_t(\boldsymbol{X}^u_t)^\top \sigma_t d\boldsymbol{B}_t\\ V_T(\boldsymbol{X}_T)=\Phi(\boldsymbol{X}_T)
    \end{cases}
    \end{align}
    \end{small}
    In addition, for the controlled process $(\boldsymbol{X}^u_t)_{t\in [0,T]}$, the FBSDEs are given by:
    \begin{small}
    \begin{align}
    \begin{cases}
        d\boldsymbol{X}^u_t=(\boldsymbol{f}(\boldsymbol{X}^u_t,t)+\sigma_t\boldsymbol{u}(\boldsymbol{X}^u_t,t))dt+\sigma_td\boldsymbol{B}_t\\
        dV_t(\boldsymbol{X}^u_t)=\left(\frac{\sigma^2_t}{2}\|\nabla V_t(\boldsymbol{X}^u_t)\|^2-c(\boldsymbol{X}^u_t,t)+\langle \sigma_t\boldsymbol{u}(\boldsymbol{X}^u_t,t), \nabla V_t(\boldsymbol{X}^u_t)\rangle\right)dt+\nabla V_t(\boldsymbol{X}^u_t)^\top \sigma_t d\boldsymbol{B}_t\\
        V_T(\boldsymbol{X}^u_T)=\Phi(\boldsymbol{X}^u_T)
    \end{cases}
    \end{align}
    \end{small}
    where $g:\mathbb{R}^d\to \mathbb{R}$ is the terminal cost and $\sigma_t$ is the diffusion coefficient. 
\end{proposition}

\textit{Proof.} Applying (\ref{eq:uncontrolled-ito-formula-generator}) to the value function, we have:
\begin{small}
\begin{align}
    dV_t(\boldsymbol{X}_t)=\partial_tV_t(\boldsymbol{X}_t)dt+(\mathcal{A}_tV_t)(\boldsymbol{X}_t)dt +\nabla V_t(\boldsymbol{X}_t)^\top \sigma_sd\boldsymbol{B}_t
\end{align}
\end{small}
From (\ref{eq:soc-hjb-eq}), we can substitute $\partial_tV_t(\boldsymbol{X}_t)+(\mathcal{A}_tV_t)(\boldsymbol{X}_t)=\frac{\sigma_t^2}{2}\|\nabla V_t(\boldsymbol{X}_t)\|^2-c(\boldsymbol{X}_t,t)$ to get the backward evolution of the value function over the uncontrolled process as:
\begin{align}
    dV_t(\boldsymbol{X}_t)&=\bigg(\bluetext{\frac{\sigma_t^2}{2}\|\nabla V_t(\boldsymbol{X}_t)\|^2-c(\boldsymbol{X}_t,t)}\bigg)dt + \nabla V_t(\boldsymbol{X}_t)^\top \sigma_td\boldsymbol{B}_t\nonumber\\
    V_T(\boldsymbol{X}_T)&=\Phi(\boldsymbol{X}_T)
\end{align}
Applying (\ref{eq:controlled-ito-formula-generator}) and the same steps as above to the value function evaluated along the controlled process $\boldsymbol{X}^u_{0:T}$, we get:
\begin{small}
\begin{align}
    dV_t(\boldsymbol{X}^u_t)&=\big(\partial_tV_t(\boldsymbol{X}^u_t)+\bluetext{(\mathcal{A}^u_tV_t)(\boldsymbol{X}^u_t)}\big)dt + \sigma_t\nabla V_t(\boldsymbol{X}^u_t)^\top \sigma_td\boldsymbol{B}_t\tag{\ref{eq:controlled-ito-formula-generator}}\\
    &\overset{\ref{eq:generator-control-to-uncontrolled}}{=}\big(\partial_tV_t(\boldsymbol{X}^u_t)+\bluetext{(\mathcal{A}_tV_t)(\boldsymbol{X}^u_t)+\langle \sigma_t\boldsymbol{u}(\boldsymbol{X}_t^u,t),\nabla V_t(\boldsymbol{X}_t^u)\rangle}\big)dt + \nabla V_t(\boldsymbol{X}^u_t)^\top \sigma_td\boldsymbol{B}_t
\end{align}
\end{small}
Substituting the expression for $\partial_tV_t(\boldsymbol{X}_t)+(\mathcal{A}_tV_t)(\boldsymbol{X}_t)$ from (\ref{eq:soc-hjb-eq}), we have:
\begin{small}
\begin{align}
    dV_t(\boldsymbol{X}^u_t)&=\bigg(\bluetext{\frac{\sigma_t^2}{2}\|\nabla V_t(\boldsymbol{X}^u_s)\|^2-c(\boldsymbol{X}^u_t,t)}+\langle \sigma_s\boldsymbol{u}(\boldsymbol{X}^u_t,t), \nabla V_t(\boldsymbol{X}^u_t)\rangle\bigg)ds + \nabla V_t(\boldsymbol{X}^u_t)^\top \sigma_t d\boldsymbol{B}_t\\ V_T(\boldsymbol{X}^u_T)&=\Phi(\boldsymbol{X}_T^u)
\end{align}
\end{small}
which recovers the backward evolution of the value function over the controlled process from the terminal constraint $V_T(\boldsymbol{X}^u_T)=\Phi(\boldsymbol{X}_T^u)$. \hfill $\square$

The FBSDE representation of the value function establishes the evolution of the value function under a stochastic process in the form of a backward SDE. By integrating this backward SDE along the trajectory and taking expectations, we can derive an explicit relationship between the cost functional $J(\boldsymbol{x},t;\boldsymbol{u})$ for an arbitrary control $\boldsymbol{u}$ and the value function $V_t(\boldsymbol{x}):=J^\star(\boldsymbol{x},t;\boldsymbol{u}^\star)$ defined with the \boldtext{optimal control} $\boldsymbol{u}^\star$, which can be written with respect to the value function as shown in the following Proposition.

\begin{proposition}[Optimal Control and Value Function]\label{prop:optimal-control-and-value-func}
    Given a reference path measure $\mathbb{Q}$, the relationship between the \textbf{cost functional}, or cost-to-go, $J(\boldsymbol{x},t;\boldsymbol{u})$ and the \textbf{value function} $V_t(\boldsymbol{x})$ is defined as:
    \begin{small}
    \begin{align}
        J(\boldsymbol{x},t;\boldsymbol{u})=V_t(\boldsymbol{x})+\mathbb{E}_{\boldsymbol{X}^u_{t:T}\sim \mathbb{P}^u}\left[\int_t^T\frac{1}{2}\|\sigma_s\nabla V_s(\boldsymbol{X}^u_s)+\boldsymbol{u}(\boldsymbol{X}^u_s,s)\|^2\bigg|\boldsymbol{X}^u_t=\boldsymbol{x}\right]\tag{Cost Functional}
    \end{align}
    \end{small}
    where $\mathbb{Q}$ is defined by the \textbf{uncontrolled} SDE $d\boldsymbol{X}_t=\boldsymbol{f}(\boldsymbol{X}_t,t)dt +\sigma_td\boldsymbol{B}_t$. Furthermore, the optimal control satisfies:
    \begin{align}
        \boldsymbol{u}^\star(\boldsymbol{x},t)=-\sigma_t\nabla V_t(\boldsymbol{x})\tag{Optimal Control}\label{eq:optimal-control-value-func}
    \end{align}
\end{proposition}
\textit{Proof.} First, we can write the value change over the time interval $s\in [t,T]$ using Itô's lemma from Theorem \ref{thm:ito-formula} for the controlled stochastic process $(\boldsymbol{X}^u_t)_{t\in [0,T]}$ and the value function $V_t(\boldsymbol{x})\in C^{2,1}(\mathbb{R}^d\times [0,T])$ to get the time derivative:
\begin{small}
\begin{align}
    dV_s(\boldsymbol{X}^u_s)=\left[\partial_sV_s(\boldsymbol{X}^u_s)+\langle \boldsymbol{f}(\boldsymbol{X}^u_s,s)+\sigma_s\boldsymbol{u}(\boldsymbol{X}^u_s,s), \nabla V_s(\boldsymbol{X}^u_s)\rangle +\frac{1}{2}\sigma_s^2\Delta V_s(\boldsymbol{X}^u_s)\right]ds + \nabla V_s(\boldsymbol{X}^u_s)^\top \sigma_s d\boldsymbol{B}_s\label{eq:value-func-ito}
\end{align}
\end{small}
Then, we can derive the value difference over the interval $s\in [t,T]$ as:
\begin{small}
\begin{align}
    &V_T(\boldsymbol{X}^u_T)-V_t(\boldsymbol{X}^u_t)=\int_t^T\bluetext{dV_s(\boldsymbol{X}^u_s)}\nonumber\\
    &=\int_t^T\bigg(\bluetext{\partial_sV_s(\boldsymbol{X}^u_s)+\mathcal{A}_sV_s(\boldsymbol{X}^u_s)}+\langle\sigma_s\boldsymbol{u}(\boldsymbol{X}^u_s,s), \nabla V_s(\boldsymbol{X}^u_s)\rangle\bigg)ds +\int _t^T \nabla V_s(\boldsymbol{X}^u_s)^\top \sigma_s d\boldsymbol{B}_s\label{eq:soc-val-proof1}
\end{align}
\end{small}
We can substitute the expression for $\partial_sV_s(\boldsymbol{X}_s)+(\mathcal{A}_sV_s)(\boldsymbol{X}_t)$ from (\ref{eq:soc-hjb-eq}) and complete the square to get:
\begin{small}
\begin{align}
    \partial_sV_s(\boldsymbol{X}^u_s)+&\mathcal{A}_sV_s(\boldsymbol{X}^u_s)+\langle\sigma_s\boldsymbol{u}(\boldsymbol{X}^u_s,s), \nabla V_s(\boldsymbol{X}^u_s)\rangle\nonumber\\
    &=\bluetext{\frac{\sigma_t^2}{2}\|\nabla V_s(\boldsymbol{X}^u_s)\|^2-c(\boldsymbol{X}^u_s,s)}+\langle\sigma_s\boldsymbol{u}(\boldsymbol{X}^u_s,s), \nabla V_s(\boldsymbol{X}^u_s)\rangle\nonumber\\
    &=\frac{\sigma_t^2}{2}\|\nabla V_s(\boldsymbol{X}^u_s)\|^2+\langle\sigma_s\boldsymbol{u}(\boldsymbol{X}^u_s,s), \nabla V_s(\boldsymbol{X}^u_s)\rangle+\bluetext{\frac{1}{2}\|\boldsymbol{u}(\boldsymbol{X}^u_s,s)\|^2-\frac{1}{2}\|\boldsymbol{u}(\boldsymbol{X}^u_s,s)\|^2}-c(\boldsymbol{X}^u_s,s)\nonumber\\
    &=\frac{1}{2}\|\sigma_s\nabla V_s(\boldsymbol{X}^u_s)+\boldsymbol{u}(\boldsymbol{X}^u_s,s)\|^2-\frac{1}{2}\|\boldsymbol{u}(\boldsymbol{X}^u_s,s)\|^2-c(\boldsymbol{X}^u_s,s)\label{eq:soc-val-proof2}
\end{align}
\end{small}
Substituting (\ref{eq:soc-val-proof2}) back into (\ref{eq:soc-val-proof1}) and setting $V_T(\boldsymbol{X}^u_T)=\Phi(\boldsymbol{X}^u_T)$, we can rearrange and take the conditional expectation $\mathbb{E}[\cdot|\boldsymbol{X}^u_t=\boldsymbol{x}]$ to get:
\begin{small}
\begin{align}
    &\Phi(\boldsymbol{X}^u_T)-V_t(\boldsymbol{X}^u_t)=\int_t^T\bigg[\bluetext{\frac{1}{2}\|\sigma_s\nabla V_s(\boldsymbol{X}^u_s)+\boldsymbol{u}(\boldsymbol{X}^u_s,s)\|^2-\frac{1}{2}\|\boldsymbol{u}(\boldsymbol{X}^u_s,s)\|^2-c(\boldsymbol{X}^u_s,s)}\bigg]ds +\int _t^T \nabla V_s(\boldsymbol{X}^u_s)^\top \sigma_s d\boldsymbol{B}_s\nonumber\\
    & =\int_t^T\frac{1}{2}\|\sigma_s\nabla V_s(\boldsymbol{X}^u_s)+\boldsymbol{u}(\boldsymbol{X}^u_s,s)\|^2ds-\underbrace{\int_t^T\left(\frac{1}{2}\|\boldsymbol{u}(\boldsymbol{X}^u_s,s)\|^2-c(\boldsymbol{X}^u_s,s)\right)ds}_{(\bigstar)} +\int _t^T \nabla V_s(\boldsymbol{X}^u_s)^\top \sigma_s d\boldsymbol{B}_s
\end{align}
\end{small}
Observing that both $\Phi(\boldsymbol{X}_T)$ and $(\bigstar)$ appear in the conditional expectation of the cost functional $J(\boldsymbol{x},t;\boldsymbol{u})$, we can rearrange and take expectations $\mathbb{E}[\cdot|\boldsymbol{X}_t=\boldsymbol{x}]$ to get:
\begin{small}
\begin{align}
    (\bigstar)+\Phi(\boldsymbol{X}^u_T)&=V_t(\boldsymbol{X}^u_t)+\int_t^T\frac{1}{2}\|\sigma_s\nabla V_s(\boldsymbol{X}^u_s)+\boldsymbol{u}(\boldsymbol{X}^u_s,s)\|^2+\int _t^T \nabla V_s(\boldsymbol{X}^u_s)^\top \sigma_s d\boldsymbol{B}_s\nonumber\\
    \underbrace{\bluetext{\mathbb{E}[}(\bigstar)+\Phi(\boldsymbol{X}^u_T)\bluetext{\big|\boldsymbol{X}^u_t=\boldsymbol{x}]}}_{=:J(\boldsymbol{x},t;\boldsymbol{u})}&=\bluetext{\mathbb{E}\bigg[}V_t(\boldsymbol{X}^u_t)+\int_t^T\frac{1}{2}\|\sigma_s\nabla V_s(\boldsymbol{X}^u_s)+\boldsymbol{u}(\boldsymbol{X}^u_s,s)\|^2+\underbrace{\int _t^T \nabla V_s(\boldsymbol{X}^u_s)^\top \sigma_s d\boldsymbol{B}_s}_{\text{vanishes under expectation}}\bluetext{\bigg|\boldsymbol{X}^u_t=\boldsymbol{x}\bigg]}\nonumber
\end{align}
\end{small}
Recognizing that the left-hand side is the definition of the \textbf{cost-to-go} function $J(\boldsymbol{x},t;\boldsymbol{u})$ and the expectation of the Itô integral vanishes, we conclude: 
\begin{align}
    \boxed{J(\boldsymbol{x},t;\boldsymbol{u})=V_t(\boldsymbol{x})+\mathbb{E}\left[\int_t^T\frac{1}{2}\|\sigma_s\nabla V_s(\boldsymbol{X}^u_s)+\boldsymbol{u}(\boldsymbol{X}^u_s,s)\|^2\bigg|\boldsymbol{X}^u_t=\boldsymbol{x}\right]}\tag{Cost Functional}
\end{align}
which establishes the relationship between the \textbf{value function} defined with the \textit{optimal control} $\boldsymbol{u}^\star$ and the \textbf{cost-to-go} under the arbitrary control $\boldsymbol{u}$. Since, the value function is equal to the \textit{optimal cost-to-go} by definition, where $V_t(\boldsymbol{x}) :=J^\star(\boldsymbol{x},t; \boldsymbol{u}^\star)$, the exepectation must vanish at optimality. Therefore, we can derive the expression for the optimal control $\boldsymbol{u}^\star(\boldsymbol{x},t)$ from the condition:
\begin{align}
    \forall (\boldsymbol{x},s)\in \mathbb{R}^d\times [0,T], \quad \|\sigma_s\nabla V_s(\boldsymbol{x})+\boldsymbol{u}^\star(\boldsymbol{x},s)\|^2=0\implies \boxed{\boldsymbol{u}^\star(\boldsymbol{x},s)=-\sigma_s\nabla V_s(\boldsymbol{x})}
\end{align}
which recovers the expression in (\ref{eq:optimal-control-value-func}). \hfill $\square$

The value function can alternatively be expressed using expectations over an uncontrolled reference measure $\mathbb{Q}$ \citep{kappen2012optimal, domingo2024stochastic}. This representation, derived using the (\ref{eq:feynman-kac-formula}), will later allow us to connect stochastic optimal control with the Schrödinger bridge formulation based on path measures.

\begin{proposition}[Closed-Form Expression for Value Function (Equation 8 in \citet{domingo2024stochastic})]\label{prop:feynman-kac}
    Given a reference path measure $\mathbb{Q}$, the \textbf{value function} $V_t(\boldsymbol{x})$ can be derived independently of the optimal control $\boldsymbol{u}^\star$ as:
    \begin{align}
        V_t(\boldsymbol{x})=-\log \mathbb{E}_{\boldsymbol{X}_{0:T}\sim \mathbb{Q}}\left[\exp \left(-\int_t^Tc(\boldsymbol{X}_s,s)ds+\Phi(\boldsymbol{X}_T)\right)\bigg|\boldsymbol{X}_t=\boldsymbol{x}\right]
    \end{align}
    where $\mathbb{Q}$ is defined by the \textbf{uncontrolled} SDE $d\boldsymbol{X}_t=\boldsymbol{f}(\boldsymbol{X}_t,t)dt +\sigma_td\boldsymbol{B}_t$.
\end{proposition}

\textit{Proof.} This proof combines the (\ref{eq:soc-hjb-eq}), (\ref{eq:hopf-cole-value-pde}), and the (\ref{eq:feynman-kac-formula}). First, we recall that the value function solves (\ref{eq:soc-hjb-eq}), defined as:
\begin{small}
\begin{align}
    \partial_tV_t+\mathcal{A}_t V_t-\frac{1}{2}\|\sigma_t^\top\nabla V_t\|^2+c =0
\end{align}
\end{small}
which we show in Corollary \ref{cor:hopf-cole-value} be transformed to a (\ref{eq:hopf-cole-value-pde}) via the change in variables $\varphi_t(\boldsymbol{x})=e^{-V_t(\boldsymbol{x})}$ defined as:
\begin{small}
\begin{align}
    \partial_t\varphi_t(\boldsymbol{x})+\langle\boldsymbol{f}(\boldsymbol{x},t), \nabla \varphi_t(\boldsymbol{x})\rangle+\frac{\sigma_t^2}{2}\Delta\varphi_t(\boldsymbol{x})-c(\boldsymbol{x},t)\varphi_t(\boldsymbol{x})&=0, \quad \varphi_T(\boldsymbol{x})=e^{-\Phi(\boldsymbol{x})}
\end{align}
\end{small}
Applying the (\ref{eq:feynman-kac-formula}), we can write $\varphi_t(\boldsymbol{x})$ as an expectation over uncontrolled stochastic processes as:
\begin{small}
\begin{align}
    \varphi_t(\boldsymbol{x})=e^{-V_t(\boldsymbol{x})}&=\mathbb{E}_{\boldsymbol{X}_{0:T}\sim\mathbb{Q}}\left[\exp\left(-\int_t^Tc(\boldsymbol{X}_s,s)ds \right)e^{-\Phi(\boldsymbol{X}_T)}\bigg|\boldsymbol{X}_t=\boldsymbol{x}\right]\nonumber\\
    &=\mathbb{E}_{\boldsymbol{X}_{0:T}\sim\mathbb{Q}}\left[\exp\left(-\int_t^Tc(\boldsymbol{X}_s,s)ds-\Phi(\boldsymbol{X}_T) \right)\bigg|\boldsymbol{X}_t=\boldsymbol{x}\right]
\end{align}
\end{small}
Taking the logarithm on both sides and inverting the sign, we get:
\begin{small}
\begin{align}
    V_t(\boldsymbol{x})=-\log \bluetext{\mathbb{E}_{\boldsymbol{X}_{0:T}\sim\mathbb{Q}}\left[\exp\left(-\int_t^Tc(\boldsymbol{X}_s,s)ds-\Phi(\boldsymbol{X}_T) \right)\bigg|\boldsymbol{X}_t=\boldsymbol{x}\right]}
\end{align}
\end{small}
which defines the closed form expression for the value function as an \textbf{expectation over uncontrolled stochastic processes}. \hfill $\square$

The Feynman-Kac representation above expresses the value function as a log-expectation with respect to the reference path measure $\mathbb{Q}$. While this formulation provides a closed-form characterization of the optimal cost-to-go, it does not yet explicitly describe how the \textbf{optimal controlled process differs from the reference dynamics}. To make this relationship concrete, we interpret SOC as a \textbf{change of measure on path space}, where the optimal path $\mathbb{P}^\star$ is obtained by reweighting the reference process $\mathbb{Q}$ using the \textit{change in the value function}. 

\begin{proposition}[Radon-Nikodym Derivative Between Optimal and Reference Path Measure]\label{prop:soc-optimal-measure}
    The \textbf{Radon-Nikodym Derivative} (RND) between the optimal controlled path measure $\mathbb{P}^\star$ and the reference path measure $\mathbb{Q}$ is defined as:
    \begin{align}
        \frac{\mathrm{d}\mathbb{P}^\star}{\mathrm{d}\mathbb{Q}}(\boldsymbol{X}_{0:T})=e^{-\Phi(\boldsymbol{X}_T)+V_0(\boldsymbol{X}_0)-\int_0^Tc(\boldsymbol{X}_t,t)dt}\tag{Optimal Path RND}\label{eq:optimal-path-rnd}
    \end{align}
    which directly yields the optimal path measure:
    \begin{align}
        \mathbb{P}^\star(\boldsymbol{X}_{0:T})&=\frac{1}{Z}\mathbb{Q}(\boldsymbol{X}_{0:T})e^{-\Phi(\boldsymbol{X}_T)+V_0(\boldsymbol{X}_0)-\int_0^Tc(\boldsymbol{X}_t,t)dt}\tag{Optimal Path Measure}\label{eq:soc-optimal-path-measure}\\
        Z&:= \mathbb{E}_{\mathbb{Q}}\left[e^{-\Phi(\boldsymbol{X}_T)+V_0(\boldsymbol{X}_0)-\int_0^Tc(\boldsymbol{X}_t,t)dt}\right]\tag{Normalization Constant}\label{eq:soc-normalization}
    \end{align}
    where $Z$ is the normalization constant. Furthermore, given that the exponential factor depends only on the endpoints $(\boldsymbol{X}_0, \boldsymbol{X}_T)$, the joint endpoint law of the optimal path measure $\mathbb{P}^\star(\boldsymbol{X}_0, \boldsymbol{X}_T)$ is given by:
    \begin{align}
        \mathbb{P}^\star(\boldsymbol{X}_0, \boldsymbol{X}_T)=\frac{1}{Z}\mathbb{Q}(\boldsymbol{X}_0, \boldsymbol{X}_T)e^{-\Phi(\boldsymbol{X}_T)+V_0(\boldsymbol{X}_0)-\int_0^Tc(\boldsymbol{X}_t,t)dt}\label{eq:soc-optimal-joint-dist}
    \end{align}
\end{proposition}

\textit{Proof.} To prove this, we leverage the form of the RND from Theorem \ref{theorem:rnd-ito-process} where the two path measures are generated via the SDEs:
\begin{align}
    \mathbb{P}^\star:&\quad d\boldsymbol{X}_t=(\boldsymbol{f}(\boldsymbol{X}_t,t)+\sigma_t\boldsymbol{u}^\star(\boldsymbol{X}_t,t)) dt+ \sigma_td\boldsymbol{B}_t\\
    \mathbb{Q}:&\quad d \boldsymbol{X}_t=\boldsymbol{f}(\boldsymbol{X}_t,t)dt+ \sigma_td\boldsymbol{B}_t
\end{align}
which yields the RND:
\begin{small}
\begin{align}
    \frac{\mathrm{d}\mathbb{P}^\star}{\mathrm{d}\mathbb{Q}}(\boldsymbol{X}_{0:T})=\exp\left(-\frac{1}{2}\int_0^T\|\bluetext{\boldsymbol{u}^\star(\boldsymbol{X}_t,t)}\|^2dt+\int_0^T\bluetext{\boldsymbol{u}^\star(\boldsymbol{X}_t,t)}^\top d\boldsymbol{B}_t\right)
\end{align}
\end{small}
Substituting the definition of the optimal control with the value function $\boldsymbol{u}^\star(\boldsymbol{X}_t,t)=-\sigma_t\nabla V_t(\boldsymbol{X}_t)$, we get:
\begin{small}
\begin{align}
    \frac{\mathrm{d}\mathbb{P}^\star}{\mathrm{d}\mathbb{Q}}(\boldsymbol{X}_{0:T})&=\exp\left(-\frac{1}{2}\int_0^T\|\bluetext{-\sigma_t\nabla V_t(\boldsymbol{X}_t)}\|^2dt+\int_0^T\bluetext{-\sigma_t\nabla V_t(\boldsymbol{X}_t)}^\top d\boldsymbol{B}_t\right)\nonumber\\
    &=\exp\left(-\int_0^T\frac{\sigma_t^2}{2}\|\nabla V_t(\boldsymbol{X}_t)\|^2dt-\int_0^T\nabla V_t(\boldsymbol{X}_t)^\top \sigma_td\boldsymbol{B}_t\right)
\end{align}
\end{small}
Recall that (\ref{eq:soc-hjb-eq}) also contains the expression $\frac{\sigma_t^2}{2}\|\nabla V_t(\boldsymbol{X}_t)\|^2$ which can be isolated as:
\begin{small}
\begin{align}
    \partial_tV_t(\boldsymbol{x})=-(\mathcal{A}_tV_t)(\boldsymbol{x})+\frac{\sigma_t^2}{2}\|\nabla V_t(\boldsymbol{x})\|^2-c(\boldsymbol{x},t)\implies\frac{\sigma_t^2}{2}\|\nabla V_t(\boldsymbol{x})\|^2=\partial_tV_t(\boldsymbol{x})+(\mathcal{A}_tV_t)(\boldsymbol{x})+c(\boldsymbol{x},t)
\end{align}
\end{small}
Substituting this into the integral, we have:
\begin{small}
\begin{align}
    \frac{\mathrm{d}\mathbb{P}^\star}{\mathrm{d}\mathbb{Q}}(\boldsymbol{X}_{0:T})&=\exp\left(-\int_0^T\bluetext{\left(\partial_tV_t(\boldsymbol{X}_t)+(\mathcal{A}_tV_t)(\boldsymbol{X}_t)+c(\boldsymbol{X}_t,t)\right)}dt-\int_0^T\nabla V_t(\boldsymbol{X}_t)^\top \sigma_td\boldsymbol{B}_t\right)\nonumber\\
    &=\exp\bigg(-\int_0^T\bluetext{\underbrace{\bigg(\left(\partial_tV_t(\boldsymbol{X}_t)+(\mathcal{A}_tV_t)(\boldsymbol{X}_t)\right)dt-\nabla V_t(\boldsymbol{X}_t)^\top \sigma_td\boldsymbol{B}_t\bigg)}_{dV_t(\boldsymbol{X}_t)}}-\int_0^Tc(\boldsymbol{X}_t,t)dt\bigg)
\end{align}
\end{small}
Since integrating the Itô integral of the value process is equal to the value difference, we can write:
\begin{small}
\begin{align}
    \frac{\mathrm{d}\mathbb{P}^\star}{\mathrm{d}\mathbb{Q}}(\boldsymbol{X}_{0:T})&=\exp\bigg(-\bluetext{\underbrace{\int_0^TdV_t(\boldsymbol{X}_t)}_{(V_T(\boldsymbol{X}_T)-V_0(\boldsymbol{X}_0))}}-\int_0^Tc(\boldsymbol{x},t)dt\bigg)\nonumber\\
    &=\boxed{\exp\bigg(-\bluetext{V_T(\boldsymbol{X}_T)+V_0(\boldsymbol{X}_0)}-\int_0^Tc(\boldsymbol{x},t)dt\bigg)}
\end{align}
\end{small}
Multiplying both sides by the reference path measure and scaling by a normalization factor, we get the expression for the optimal path measure as:
\begin{align}
    \boxed{\mathbb{P}^\star(\boldsymbol{X}_{0:T})=\frac{1}{Z}e^{-V_T(\boldsymbol{X}_T)+V_0(\boldsymbol{X}_0)-\int_0^Tc(\boldsymbol{X}_t,t)dt}\mathbb{Q}(\boldsymbol{X}_{0:T})}
\end{align}
where $Z$ is defined such that the distribution integrates to one:
\begin{align}
    &\int\frac{1}{Z}e^{-V_T(\boldsymbol{X}_T)+V_0(\boldsymbol{X}_0)}\mathbb{Q}(\boldsymbol{X}_{0:T})-\int\int_0^Tc(\boldsymbol{x},t)dt\mathbb{Q}(\boldsymbol{X}_{0:T})=\frac{1}{Z}\mathbb{E}_{\mathbb{Q}}\left[e^{-V_T(\boldsymbol{X}_T)+V_0(\boldsymbol{X}_0)-\int_0^Tc(\boldsymbol{X}_t,t)dt}\right]=1\nonumber\\
    &\implies \boxed{Z:=\mathbb{E}_{\mathbb{Q}}\left[e^{-V_T(\boldsymbol{X}_T)+V_0(\boldsymbol{X}_0)-\int_0^Tc(\boldsymbol{X}_t,t)dt}\right]}
\end{align}
which concludes our derivation of the expression for the optimal RND and optimal path measure. Since the exponential scaling factor depends only on the endpoints $(\boldsymbol{X}_0, \boldsymbol{X}_T)$, we can derive the expression for the endpoint law:  
\begin{align}
    \boxed{\mathbb{P}^\star(\boldsymbol{X}_0, \boldsymbol{X}_T)=\frac{1}{Z}\mathbb{Q}(\boldsymbol{X}_0, \boldsymbol{X}_T)e^{-V_T(\boldsymbol{X}_T)+V_0(\boldsymbol{X}_0)-\int_0^Tc(\boldsymbol{X}_t,t)dt}}
\end{align}
which means that the endpoint law of the optimal path measure is an exponential tilting of the reference path measure by the initial and terminal value functions. Substituting the terminal constraint $V_T(\boldsymbol{X}_T)=\Phi(\boldsymbol{X}_T)$ yields the final expressions in Proposition \ref{prop:soc-optimal-measure}. \hfill $\square$

The expression for (\ref{eq:soc-optimal-path-measure}) reveals the \textbf{key challenge} in solving the SOC problem. That is, the optimal path measure is a \textit{tilted} version of the reference path measure by not only the terminal cost $-V_T(\boldsymbol{X}_T)$ but also the initial value function $V_0(\boldsymbol{X}_0)$, which is defined as an expectation over stochastic paths as defined in Proposition \ref{prop:feynman-kac} and is \textbf{intractable}. This motivates the definition of \boldtext{memoryless reference processes}, which removes the dependency of the optimal path on the initial value function. 

\purple[Memoryless Reference Process Eliminates Initial Value Function Bias]{\label{box:memoryless}
    Consider a reference process $\mathbb{Q}$ where the joint distribution of initial and terminal states is \textbf{independent}, such that the endpoint distribution factorizes: 
    \begin{align}
        \mathbb{Q}(\boldsymbol{X}_0, \boldsymbol{X}_T)=q_0(\boldsymbol{X}_0)q_T(\boldsymbol{X}_T)
    \end{align}
    From Proposition \ref{prop:soc-optimal-measure}, we show that the endpoint distribution of the optimal path measure $\mathbb{P}^\star$ is an exponential tilting of the reference law, given by:
    \begin{align}
        \mathbb{P}^\star(\boldsymbol{X}_0, \boldsymbol{X}_T)=\frac{1}{Z}\mathbb{Q}(\boldsymbol{X}_0, \boldsymbol{X}_T)e^{-\Phi(\boldsymbol{X}_T)+V_0(\boldsymbol{X}_0)}
    \end{align}
    where we consider the common case with no running cost function $c\equiv 0$. Then, the terminal distribution under the optimal process $\mathbb{P}^\star$ satisfies:
    \begin{small}
    \begin{align}
        p^\star_T(\boldsymbol{X}_T)&=\int_{\mathbb{R}^d}\mathbb{P}^\star(\boldsymbol{X}_0, \boldsymbol{X}_T)d\boldsymbol{X}_0=\int_{\mathbb{R}^d}\bluetext{\mathbb{Q}(\boldsymbol{X}_0, \boldsymbol{X}_T)}e^{-\Phi(\boldsymbol{X}_T)+V_0(\boldsymbol{X}_0)}d\boldsymbol{X}_0\nonumber\\
        &=\int_{\mathbb{R}^d}\bluetext{q_0(\boldsymbol{X}_0)q_T(\boldsymbol{X}_T)}e^{-\Phi(\boldsymbol{X}_T)}e^{V_0(\boldsymbol{X}_0)}d\boldsymbol{X}_0=q_T(\boldsymbol{X}_T)e^{-\Phi(\boldsymbol{X}_T)}\bluetext{\underbrace{\int_{\mathbb{R}^d} q_0(\boldsymbol{X}_0)e^{V_0(\boldsymbol{X}_0)}d\boldsymbol{X}_0}_{\text{constant normalization}}}\nonumber\\
        &=\boxed{\bluetext{\frac{1}{Z}}q_T(\boldsymbol{X}_T)e^{-\Phi(\boldsymbol{X}_T)}}
    \end{align}
    \end{small}
    where the initial value function integrates to a constant normalization factor, eliminating the need to compute $V_0(\boldsymbol{X}_0)$ explicitly. Furthermore, we can define $\Phi(\boldsymbol{X}_T)$ such that the terminal marginal is exactly the target distribution $\pi_T(\boldsymbol{X}_T)$ as:
    \begin{small}
    \begin{align}
        \frac{1}{Z}q_T(\boldsymbol{X}_T)e^{-\Phi(\boldsymbol{X}_T)}=\pi_T(\boldsymbol{X}_T)&\implies \frac{q_T(\boldsymbol{X}_T)}{Z\pi_T(\boldsymbol{X}_T)}=e^{\Phi(\boldsymbol{X}_T)}\nonumber\\
        &\implies \Phi(\boldsymbol{X}_T)=\log \frac{q_T(\boldsymbol{X}_T)}{\pi_T(\boldsymbol{X}_T)}-\log Z
    \end{align}
    \end{small}
    which yields a tractable objective that does not require computing $V_0(\boldsymbol{X}_0)$. Examples of memoryless reference processes include:
    \begin{enumerate}
        \item[(i)] Linear reference drift $\boldsymbol{f}:=\boldsymbol{A}_t\boldsymbol{X}_t +\boldsymbol{b}_t$, for some $\boldsymbol{A}_t\in \mathbb{R}^{d\times d}$ and $\boldsymbol{b}_t\in \mathbb{R}^d$, and noise that grows with time such that the initial distribution is a standard zero-mean Gaussian $\boldsymbol{X}_0\sim \mathcal{N}(\boldsymbol{0}, \boldsymbol{I}_d)$. This includes the variance-preserving SDE used in \citet{song2020score}.
        \begin{small}
        \begin{align}
            \mathbb{Q}:\;d\boldsymbol{X}_t=-\frac{1}{2}\beta_t\boldsymbol{X}_tdt +\sqrt{\beta_t}d\boldsymbol{B}_t, \quad \boldsymbol{X}_0\sim \mathcal{N}(\boldsymbol{0}, \boldsymbol{I}_d) \implies \mathbb{Q}(\boldsymbol{X}_0, \boldsymbol{X}_T)\approx q_0(\boldsymbol{X}_0)q_T(\boldsymbol{X}_T)\nonumber
        \end{align}
        \end{small}
        \item[(ii)] Brownian motion reference drift $\boldsymbol{f}:=\boldsymbol{0}$ and a initial distribution that is a Dirac delta $\pi_0:=\delta_0$.
        \begin{small}
        \begin{align}
            \mathbb{Q}:\;d\boldsymbol{X}_t=\sigma_td\boldsymbol{B}_t, \quad \boldsymbol{X}_0=\boldsymbol{0} \implies \mathbb{Q}(\boldsymbol{X}_0, \boldsymbol{X}_T)\approx \delta_0(\boldsymbol{X}_0)q_T(\boldsymbol{X}_T)\nonumber
        \end{align}
        \end{small}
    \end{enumerate}
}

While this shows that we can solve the SOC problem without computing the initial value function $V_0(\boldsymbol{X}_0)$ by explicitly designing the reference process to be \textbf{memoryless}, this construction restricts us to a limited class of relatively uninformative reference dynamics, which may not provide a meaningful prior for a given system. To overcome this limitation, we will reformulate the SOC objective by absorbing the initial value function into the terminal cost, which allows for a broader class of reference processes and more expressive prior dynamics.

\subsection{Schrödinger Bridges with Stochastic Optimal Control}
\label{subsec:sb-soc}
In this section, we will adapt stochastic optimal control (SOC) theory for solving the (\ref{eq:dynamic-sb-problem}) with arbitrary prior dynamics and initial distributions. First, we recall that the Hopf-Cole transform from Theorem \ref{thm:hopf-cole-nonlinear} expresses the optimal control $\boldsymbol{u}^\star$ and probability density $p^\star_t$ using a pair of \boldtext{forward-backward SB potentials} $(\varphi_t, \hat{\varphi}_t)$ as: 
\begin{small}
\begin{align}
    \boldsymbol{u}^\star(\boldsymbol{x},t)=\sigma_t\nabla \log\varphi_t(\boldsymbol{x}), \quad p^\star_t(\boldsymbol{x})=\varphi_t(\boldsymbol{x})\hat{\varphi}_t(\boldsymbol{x})\label{eq:sb-system2}
\end{align}
\end{small}
where $(\varphi_t, \hat{\varphi}_t)$ are defined in (\ref{eq:proof-sb-system}) as:
\begin{small}
\begin{align}
    \begin{cases}
        \varphi_t(\boldsymbol{x})=\int_{\mathbb{R}^d}\mathbb{Q}_{T|t}(\boldsymbol{y}|\boldsymbol{x})\varphi_T(\boldsymbol{y})d\boldsymbol{y}\\
        \hat{\varphi}_t(\boldsymbol{x})=\int_{\mathbb{R}^d}\mathbb{Q}_{t|0}(\boldsymbol{x}|\boldsymbol{y})\hat{\varphi}_T(\boldsymbol{y})d\boldsymbol{y}
    \end{cases}\quad\text{s.t.}\quad
    \begin{cases}
        \pi_0(\boldsymbol{x})=\varphi_0(\boldsymbol{x})\hat{\varphi}_0(\boldsymbol{x})\\
        \pi_T(\boldsymbol{x})=\varphi_T(\boldsymbol{x})\hat{\varphi}_T(\boldsymbol{x})
    \end{cases}\tag{Schrödinger Potentials}\label{eq:sb-system1}
\end{align}
\end{small}
Now, we can rewrite the original SOC problem in (\ref{def:soc-objective}) by observing that the optimal control can be written in \textbf{two equivalent ways}: using the \textbf{value function} $V_t(\boldsymbol{x})$ (Proposition \ref{prop:optimal-control-and-value-func}) \textit{or} the \textbf{Schrödinger potential} $\varphi_t(\boldsymbol{x})$ (\ref{eq:sb-system2}). This allows us to derive the \textbf{relationship between the value function and the SB potential} as:
\begin{align}
    \boldsymbol{u}^\star(\boldsymbol{x},t)=-\sigma_t\nabla V_t(\boldsymbol{x})=\sigma_t\nabla \log\varphi_t(\boldsymbol{x})\implies \bluetext{V_t(\boldsymbol{x})=-\log \varphi_t(\boldsymbol{x})}\tag{SB Value Function}\label{eq:value-sb-potential}
\end{align}
Since the terminal cost is given by the terminal value function $\Phi(\boldsymbol{x})=V_T(\boldsymbol{x})$, we can also write the terminal cost  in terms of the SB potentials as:
\begin{align}
    \pinktext{\Phi(\boldsymbol{x})}&=V_T(\boldsymbol{x})=-\log \varphi_T(\boldsymbol{x})=-\log \frac{\pi_T(\boldsymbol{x})}{\hat{\varphi}_T(\boldsymbol{x})}\pinktext{=\log \frac{\hat{\varphi}_T(\boldsymbol{x})}{\pi_T(\boldsymbol{x})}}\tag{SB Terminal Cost}\label{eq:terminal-cost-sb-potential}
\end{align}
where we substitute $\pi_T(\boldsymbol{x})=\varphi_T(\boldsymbol{x})\hat{\varphi}(\boldsymbol{x})\implies \varphi_T(\boldsymbol{x})=\frac{\pi_T(\boldsymbol{x})}{\hat{\varphi}_t(\boldsymbol{x})}$ from (\ref{eq:sb-system1}). Substituting (\ref{eq:value-sb-potential}) and (\ref{eq:terminal-cost-sb-potential}) into the results from Proposition \ref{prop:soc-optimal-measure}, we can write the optimal path RND and path measure $\mathbb{P}^\star$ that solves the dynamic SB problem as:
\begin{small}
\begin{align}
    \frac{\mathrm{d}\mathbb{P}^\star}{\mathrm{d}\mathbb{Q}}(\boldsymbol{X}_{0:T})&=\exp\left(-\Phi(\boldsymbol{X}_T)+V_0(\boldsymbol{X}_0)\right)=\exp\left(-\log \frac{\hat{\varphi}_T(\boldsymbol{X}_T)}{\pi_T(\boldsymbol{X}_T)}-\log \varphi_0(\boldsymbol{X}_0)\right)\tag{Optimal SB Path RND}\label{eq:optimal-sb-path-rnd}\\
    \mathbb{P}^\star(\boldsymbol{X}_{0:T})&=\frac{1}{Z}\mathbb{Q}(\boldsymbol{X}_{0:T})\exp\left(-\log \frac{\hat{\varphi}_T(\boldsymbol{X}_T)}{\pi_T(\boldsymbol{X}_T)}-\log \varphi_0(\boldsymbol{X}_0)\right)\tag{Optimal SB Path Measure}\label{eq:optimal-sb-path-measure}
\end{align}
\end{small}

Using (\ref{eq:value-sb-potential}) and (\ref{eq:terminal-cost-sb-potential}), we also define the \textbf{alternative Schrödinger bridge form of the SOC problem} which yields the optimal SB solution. 

\begin{definition}[Schrödinger Bridge with Stochastic Optimal Control (SB-SOC)]\label{def:soc-sb-potential}
    Given the Schrödinger potentials $(\varphi_t, \hat{\varphi}_t)$ that satisfy the linear PDEs in (\ref{eq:hopf-cole-system}), the (\ref{eq:soc-objective}) can be expressed in terms of $(\varphi_t, \hat{\varphi}_t)$ as:
    \begin{align}
        &\inf_{\boldsymbol{u}}\mathbb{E}_{\boldsymbol{X}^u_{0:T}\sim \mathbb{P}^u}\left[\int_0^T\frac{1}{2}\|\boldsymbol{u}(\boldsymbol{X}^u_t,t)\|^2dt+\log \frac{\hat{\varphi}_T(\boldsymbol{X}^u_T)}{\pi_T(\boldsymbol{X}^u_T)}\right]\tag{SB-SOC Objective}\label{eq:sb-soc-objective}\\
        &\text{s.t.}\quad d\boldsymbol{X}^u_t=(\boldsymbol{f}(\boldsymbol{X}^u_t,t)+\sigma_t\boldsymbol{u}(\boldsymbol{X}^u_t,t))dt+\sigma_td\boldsymbol{B}_t, \quad \boldsymbol{X}^u_0\sim \pi_0\nonumber
    \end{align}
    where $\boldsymbol{u}(\boldsymbol{x},t)$ is the control drift which produces the path measure $\mathbb{P}^u$, $\boldsymbol{f}(\boldsymbol{x},t)$ is the drift of the reference process $\mathbb{Q}$, $\sigma_t$ is the diffusion coefficient, and $d\boldsymbol{B}_t$ is $d$-dimensional Brownian motion.
\end{definition}

Given the (\ref{eq:sb-soc-objective}), we can define an \textit{cost-to-go} analogously to (\ref{eq:soc-cost-functional}) from any fixed point $(\boldsymbol{x}, t)\in \mathbb{R}^d\times [0,T]$ at time $t$ under the control $\boldsymbol{u}$ as:
\begin{align}
    J(\boldsymbol{x},t;\boldsymbol{u}):=\mathbb{E}_{\boldsymbol{X}^u_{t:T}\sim \mathbb{P}^u}\left[\int_t^T\frac{1}{2}\|\boldsymbol{u}(\boldsymbol{X}^u_s,s)\|^2ds+\log \frac{\hat{\varphi}_T(\boldsymbol{X}^u_T)}{\pi_T(\boldsymbol{X}^u_T)}\bigg|\boldsymbol{X}^u_t=\boldsymbol{x}\right]\label{eq:sb-soc-cost-functional}
\end{align}
We can also show that (\ref{eq:sb-soc-objective}) eliminates the initial value bias and does not require defining a memoryless reference process as discussed in Box \ref{box:memoryless}.

\begin{proposition}[Schrödinger Potential Eliminates the Initial Value Bias]\label{prop:initial-value-bias}
The (\ref{eq:sb-soc-objective}) does not require computing the initial value function $V_0(\boldsymbol{x})$. In particular, under the optimal control $\boldsymbol{u}^\star$, the induced path measure $\mathbb{P}^\star$ automatically satisfies the terminal marginal constraint $p_T^\star(\boldsymbol{x}_T)=\pi_T(\boldsymbol{x}_T)$ independently of the initial value function $V_0(x)$.
\end{proposition}

\textit{Proof.} 
Starting from the SB-SOC representation of the (\ref{eq:optimal-sb-path-measure}), we can write the optimal endpoint law as:
\begin{align}
    \mathbb{P}^\star(\boldsymbol{X}_0, \boldsymbol{X}_T)&=\frac{1}{Z}\mathbb{Q}(\boldsymbol{X}_0, \boldsymbol{X}_T)\exp\left(-\log \frac{\hat{\varphi}_T(\boldsymbol{X}_T)}{\pi_T(\boldsymbol{X}_T)}-\log \varphi_0(\boldsymbol{X}_0)\right)
\end{align}
Integrating over $\boldsymbol{X}_0$, we have that the terminal marginal of the optimal controlled path $p^\star_T$ is equal to:
\begin{align}
    p_T^\star(\boldsymbol{X}_T)&=\int_{\mathbb{R}^d}\mathbb{Q}(\boldsymbol{X}_0, \boldsymbol{X}_T)\exp\left(\bluetext{-\log \frac{\hat{\varphi}_T(\boldsymbol{X}_T)}{\pi_T(\boldsymbol{X}_T)}} \pinktext{-\log \varphi_0(\boldsymbol{X}_0)}\right)d\boldsymbol{X}_0
\end{align}
Rearranging the exponential term, we have:
\begin{align}
p_T^\star(\boldsymbol{X}_T)=\bluetext{\frac{\pi_T(\boldsymbol{X}_T)}{\hat{\varphi}_T(\boldsymbol{X}_T)}}\int_{\mathbb{R}^d}\mathbb{Q}(\boldsymbol{X}_0, \boldsymbol{X}_T)\pinktext{\frac{1}{\varphi_0(\boldsymbol{X}_0)}}d\boldsymbol{X}_0
\end{align}
Using the factorization $\mathbb{Q}(\boldsymbol{X}_0,\boldsymbol{X}_T)=\mathbb{Q}(\boldsymbol{X}_T|\boldsymbol{X}_0)\pi_0(\boldsymbol{X}_0)$ and the boundary condition $\pi_0(\boldsymbol{x})=\varphi_0(\boldsymbol{x})\hat{\varphi}_0(\boldsymbol{x})$, we get:
as follows:
\begin{align}
    p_T^\star(\boldsymbol{X}_T)&=\bluetext{\frac{\pi_T(\boldsymbol{X}_T)}{\hat{\varphi}_T(\boldsymbol{X}_T)}}\int_{\mathbb{R}^d}\mathbb{Q}(\boldsymbol{X}_T|\boldsymbol{X}_0)\pi_0(\boldsymbol{X}_0)\pinktext{\frac{\hat{\varphi}_0(\boldsymbol{X}_0)}{\pi_0(\boldsymbol{X}_0)}}d\boldsymbol{X}_0=\pi_T(\boldsymbol{X}_T)\nonumber\\
    &=\bluetext{\frac{\pi_T(\boldsymbol{X}_T)}{\hat{\varphi}_T(\boldsymbol{X}_T)}}\underbrace{\int_{\mathbb{R}^d}\mathbb{Q}(\boldsymbol{X}_T|\boldsymbol{X}_0)\pinktext{\hat{\varphi}_0(\boldsymbol{X}_0)}d\boldsymbol{X}_0}_{=\hat{\varphi}_T(\boldsymbol{X}_T)}=\pi_T(\boldsymbol{X}_T)
\end{align}
which shows that the optimal path measure $\mathbb{P}^\star$ under the optimal control $\boldsymbol{u}^\star$ that solves (\ref{eq:sb-soc-objective}) exactly matches the terminal distribution $p^\star_T=\pi_T$, without dependence on the initial value $V_0(\boldsymbol{X}_0)$. \hfill $\square$

We can now characterize the full optimal path measure induced by the optimal control $\boldsymbol{u}^\star$. In particular, we will show that the Schrödinger potentials $(\varphi_t,\hat{\varphi}_t)$ provide a clean factorization of the optimal bridge measure with respect to the reference process $\mathbb{Q}$, which reveals
how the endpoint potentials reweight trajectories of the reference dynamics to produce the Schrödinger bridge. Furthermore, the optimal path density $\mathbb{P}^\star(\boldsymbol{X}_{0:T})$, as well as its associated marginal density $p^\star_t$ and joint density $p^\star_{s,t}(\boldsymbol{y}, \boldsymbol{x})$, can be written explicitly in terms of the potentials as stated below.

\begin{proposition}[Optimal Path Density of SB-SOC]\label{prop:optimal-path-sbsoc}
    The Schrödinger bridge path measure that solves (\ref{eq:sb-soc-objective}) can be written with respect to the SB potentials as:
    \begin{small}
    \begin{align}
        \mathbb{P}^\star(\boldsymbol{X}_{0:T})=\frac{1}{Z}\mathbb{Q}(\boldsymbol{X}_{0:T}) \varphi_T(\boldsymbol{X}_T)\frac{\hat{\varphi}_0(\boldsymbol{X}_0)}{\pi_0(\boldsymbol{X}_0)}=\frac{1}{Z}\mathbb{Q}(\boldsymbol{X}_{0:T}|\boldsymbol{X}_0)\varphi_T(\boldsymbol{X}_T)\hat{\varphi}_0(\boldsymbol{X}_0)\tag{SB-SOC Path}\label{eq:sb-soc-path-measure}
    \end{align}
    \end{small}
    and the marginal density at time $t$ can be factorized as:
    \begin{align}
        p^\star_t(\boldsymbol{x})=\hat{\varphi}_t(x)\varphi_t(x)\tag{SB-SOC Density}\label{eq:sb-soc-density}
    \end{align}
    Additionally, for any $s\leq t$, the joint density of $\boldsymbol{X}_s=\boldsymbol{y}$ and $\boldsymbol{X}_t=\boldsymbol{x}$ satisfies:
    \begin{align}
        p^\star_{s,t}(\boldsymbol{y}, \boldsymbol{x})=\mathbb{Q}(\boldsymbol{X}_t=\boldsymbol{x}|\boldsymbol{X}_s=\boldsymbol{y})\hat{\varphi}_s(\boldsymbol{y})\varphi_t(\boldsymbol{x}), \quad s\leq t\tag{SB-SOC Joint Density}\label{eq:sb-soc-joint-density}
    \end{align}
\end{proposition}

\textit{Proof.} Using the alternative definition of the value function, we can also derive the optimal SB path measure $\mathbb{P}^\star$ with respect to the Schrödinger potentials. From Proposition \ref{prop:soc-optimal-measure}, we showed that the optimal path measure can be written as:
\begin{small}
\begin{align}
    \mathbb{P}^\star(\boldsymbol{X}_{0:T})&=\frac{1}{Z}\mathbb{Q}(\boldsymbol{X}_{0:T})\exp\left(-V_T(\boldsymbol{X}_T)+V_0(\boldsymbol{X}_0)-\int_0^Tc(\boldsymbol{X}_t,t)dt\right)
\end{align}
\end{small}
Since the dynamic SB problem is defined as a KL minimization problem, which only produces a quadratic control cost, we have $c\equiv 0$. Therefore, substituting the (\ref{eq:terminal-cost-sb-potential}) and (\ref{eq:value-sb-potential}), we get: 
\begin{small}
\begin{align}
    \mathbb{P}^\star(\boldsymbol{X}_{0:T})&=\frac{1}{Z}\mathbb{Q}(\boldsymbol{X}_{0:T})\exp\left(-\bluetext{V_T(\boldsymbol{X}_T)}+\pinktext{V_0(\boldsymbol{X}_0)}\right)=\frac{1}{Z}\mathbb{Q}(\boldsymbol{X}_{0:T})\exp\left(\bluetext{\log \varphi_T(\boldsymbol{X}_T)}-\pinktext{\log \varphi_0(\boldsymbol{X}_0)}\right)\nonumber\\
    &=\frac{1}{Z}\mathbb{Q}(\boldsymbol{X}_{0:T})\bluetext{ \varphi_T(\boldsymbol{X}_T)}\frac{1}{\pinktext{\varphi_0(\boldsymbol{X}_0)}}=\frac{1}{Z}\mathbb{Q}(\boldsymbol{X}_{0:T})\bluetext{ \varphi_T(\boldsymbol{X}_T)}\pinktext{\frac{\hat{\varphi}_0(\boldsymbol{X}_0)}{\pi_0(\boldsymbol{X}_0)}}\nonumber\\
    &=\frac{1}{Z}\underbrace{\frac{\mathbb{Q}(\boldsymbol{X}_{0:T})}{\pi_0(\boldsymbol{X}_0)}}_{=\mathbb{Q}(\boldsymbol{X}_{0:T}|\boldsymbol{X}_0)} \varphi_T(\boldsymbol{X}_T)\hat{\varphi}_0(\boldsymbol{X}_0)=\frac{1}{Z}\mathbb{Q}(\boldsymbol{X}_{0:T}|\boldsymbol{X}_0)\varphi_T(\boldsymbol{X}_T)\hat{\varphi}_0(\boldsymbol{X}_0)
\end{align}
\end{small}
where we use the identity $\mathbb{Q}(\boldsymbol{X}_{0:T})=\mathbb{Q}(\boldsymbol{X}_{0:T}|\boldsymbol{X}_0)\pi_0(\boldsymbol{X}_0)$. To obtain the marginal density at time $t$ given by $p^\star_t$ we can define the distribution over paths $\boldsymbol{X}_{0:T}$ where $\boldsymbol{X}_t=\boldsymbol{x}$ and integrate over all paths $\boldsymbol{X}_{0:T}$ to get:
\begin{small}
\begin{align}
    p^\star_t(\boldsymbol{x})&=\int\delta(\boldsymbol{X}_t-\boldsymbol{x})\varphi_T(\boldsymbol{X}_T)\hat{\varphi}_0(\boldsymbol{X}_0)\mathbb{Q}(d\boldsymbol{X}_{0:T}|\boldsymbol{X}_0)
\end{align}
\end{small}
Since the integrand is multiplied only by the potentials evaluated at $\boldsymbol{X}_0$ and $\boldsymbol{X}_T$, we integrate out the path coordinates leaving only $\boldsymbol{X}_0$ and $\boldsymbol{X}_T$ and apply the Markov property of the reference path measure to get a simplified expression for the marginal density:
\begin{small}
\begin{align}
    p^\star_t(\boldsymbol{x})&=\int\bluetext{\mathbb{Q}(\boldsymbol{X}_T, \boldsymbol{X}_t=\boldsymbol{x}|\boldsymbol{X}_0)}\varphi_T(\boldsymbol{X}_T)\hat{\varphi}_0(\boldsymbol{X}_0)\bluetext{d\boldsymbol{X}_0d\boldsymbol{X}_T}\nonumber\\
    &=\int\int\bluetext{\mathbb{Q}(\boldsymbol{X}_T| \boldsymbol{X}_t=\boldsymbol{x})\mathbb{Q}(\boldsymbol{X}_t=\boldsymbol{x}|\boldsymbol{X}_0)}\varphi_T(\boldsymbol{X}_T)\hat{\varphi}_0(\boldsymbol{X}_0)\bluetext{d\boldsymbol{X}_0d\boldsymbol{X}_T}\nonumber\\
    &=\underbrace{\left(\int\mathbb{Q}(\boldsymbol{X}_T| \boldsymbol{X}_t=\boldsymbol{x})\varphi_T(\boldsymbol{X}_T)d\boldsymbol{X}_T\right)}_{=\bluetext{\hat{\varphi}_t(x)}}\underbrace{\left(\int\mathbb{Q}(\boldsymbol{X}_t=\boldsymbol{x}|\boldsymbol{X}_0)\hat{\varphi}_0(\boldsymbol{X}_0)d\boldsymbol{X}_0\right)}_{=\pinktext{\varphi_t(x)}}\nonumber\\
    &=\boxed{\bluetext{\hat{\varphi}_t(x)}\pinktext{\varphi_t(x)}}
\end{align}
\end{small}

To derive the joint distribution for arbitrary $s\leq t$, we can include additional conditioning on $\boldsymbol{X}_s=\boldsymbol{y}$ and apply the Markov property of $\mathbb{Q}$ similarly to (\ref{eq:sb-soc-density}) to get:
\begin{small}
\begin{align}
    p^\star_{s,t}(\boldsymbol{y},\boldsymbol{x})&=\int\mathbb{Q}(\bluetext{\boldsymbol{X}_T}, \pinktext{\boldsymbol{X}_t=\boldsymbol{x}}, \greentext{\boldsymbol{X}_s=\boldsymbol{y}}|\boldsymbol{X}_0)\varphi_T(\boldsymbol{X}_T)\hat{\varphi}_0(\boldsymbol{X}_0)d\boldsymbol{X}_0d\boldsymbol{X}_T\nonumber\\
    &=\int\int\bluetext{\mathbb{Q}(\boldsymbol{X}_T| \boldsymbol{X}_t=\boldsymbol{x})}\pinktext{\mathbb{Q}(\boldsymbol{X}_t=\boldsymbol{x}|\boldsymbol{X}_s=\boldsymbol{y})}\greentext{\mathbb{Q}(\boldsymbol{X}_s=\boldsymbol{y}|\boldsymbol{X}_0)}\varphi_T(\boldsymbol{X}_T)\hat{\varphi}_0(\boldsymbol{X}_0)\bluetext{d\boldsymbol{X}_0d\boldsymbol{X}_T}\nonumber\\
    &=\pinktext{\mathbb{Q}(\boldsymbol{X}_t=\boldsymbol{x}|\boldsymbol{X}_s=\boldsymbol{y})}\underbrace{\left(\int\mathbb{Q}(\boldsymbol{X}_T| \boldsymbol{X}_t=\boldsymbol{x})\varphi_T(\boldsymbol{X}_T)d\boldsymbol{X}_T\right)}_{=\bluetext{\hat{\varphi}_t(x)}}\underbrace{\left(\int\mathbb{Q}(\boldsymbol{X}_t=\boldsymbol{x}|\boldsymbol{X}_0)\hat{\varphi}_0(\boldsymbol{X}_0)d\boldsymbol{X}_0\right)}_{=\greentext{\varphi_t(x)}}\nonumber\\
    &=\boxed{\mathbb{Q}(\boldsymbol{X}_t=\boldsymbol{x}|\boldsymbol{X}_s=\boldsymbol{y})\hat{\varphi}_t(x)\varphi_t(x)}
\end{align}
\end{small}
which is the joint probability density of $\boldsymbol{X}_s=\boldsymbol{y}$ at time $s$ and $\boldsymbol{X }_t=\boldsymbol{x}$ at time $t$ under the optimal SB-SOC path measure.\hfill $\square$

Together, these results show that the SOC formulation of the Schrödinger bridge can be fully characterized by the reference dynamics $\mathbb{Q}$ and the terminal Schrödinger potentials $\varphi_0$ and $\hat\varphi_T$, with the intermediate potentials defined via the (\ref{eq:sb-system1}) without the need to explicitly compute the value function. Building on this reformulation, we can now derive \textbf{tractable training objectives} that estimate the optimal control by sampling stochastic paths from proposal SDEs and minimizing divergences between control drift with respect to the reference drift.

\subsection{Objectives for Solving the SOC Problem}
\label{subsec:soc-objectives}
%log variance, relative entropy, cross-entropy
In this section, we introduce three different objective functions that can be used to solve the SOC problem defined in (\ref{eq:soc-objective}): the \textbf{relative-entropy loss} $\mathcal{L}_{\text{RE}}$, the \textbf{log-variance loss} $\mathcal{L}_{\text{LV}}$, and the \textbf{cross-entropy loss} $\mathcal{L}_{\text{CE}}$ \citep{nusken2021solving}. 

Since the SOC objective is inherently a KL divergence between the controlled path measure $\mathbb{P}^u$ generated with $\boldsymbol{u}$ and the optimal path measure $\mathbb{P}^\star:=\mathbb{P}^{\boldsymbol{u}^\star}$ generated with the \textit{optimal control} $\boldsymbol{u}^\star$, it is natural to consider a loss function that minimizes the KL divergence or the \textit{reverse KL divergence} between $\mathbb{P}^u$ and $\mathbb{P}^\star$ defined as:
\begin{align}
    \text{KL}(\mathbb{P}^u\|\mathbb{P}^\star)=\mathbb{E}_{\mathbb{P}^u}\left[\log \frac{\mathrm{d}\mathbb{P}^u}{\mathrm{d}\mathbb{P}^\star}\right], \quad \text{KL}(\mathbb{P}^\star\|\mathbb{P}^u)=\mathbb{E}_{\mathbb{P}^\star}\left[\log \frac{\mathrm{d}\mathbb{P}^\star}{\mathrm{d}\mathbb{P}^u}\right]
\end{align}
which both yield a \textbf{unique minimizer} at $\mathbb{P}^u=\mathbb{P}^\star$ when $\text{KL}(\mathbb{P}^u\|\mathbb{P}^\star)=\text{KL}(\mathbb{P}^\star\|\mathbb{P}^u)=0$. These two divergences are exactly what define the \boldtext{relative-entropy} (RE) and \boldtext{cross-entropy} (CE) losses \citep{nusken2021solving}. In the following definitions, we will define both its \textbf{path measure form} and derive its \textbf{path-integral form} with respect to the control drifts $\boldsymbol{u}$.

\begin{definition}[Relative Entropy (RE) Loss]\label{def:relative-entropy}
    The \textbf{relative entropy} (RE) loss between the controlled path measure $\mathbb{P}^u$ and the optimal path measure $\mathbb{P}^\star$ is defined as the KL divergence:
    \begin{align}
        \mathcal{L}_{\text{RE}}(\mathbb{P}^u, \mathbb{P}^\star):=\text{KL}(\mathbb{P}^u\|\mathbb{P}^\star)=\mathbb{E}_{\mathbb{P}^u}\left[\log \frac{\mathrm{d}\mathbb{P}^u}{\mathrm{d}\mathbb{P}^\star}\right]\tag{Relative Entropy Objective}\label{eq:relative-entropy-path}
    \end{align}
    Let $\boldsymbol{u}(\boldsymbol{x},t)$ denote the control that generates $\mathbb{P}^u$ and let $\boldsymbol{X}^u_{0:T}=(\boldsymbol{X}^u_t)_{t\in [0,T]}$ denote a stochastic process under the (\ref{eq:controlled-sde}). Then, the RE loss takes the path-integral form:
    \begin{small}
    \begin{align}
        \mathcal{L}_{\text{RE}}(\boldsymbol{u}):=\mathbb{E}_{\boldsymbol{X}^u_{0:T}\sim \mathbb{P}^u}\left[\frac{1}{2}\int_0^T\|\boldsymbol{u}(\boldsymbol{X}^u_t,t)\|^2dt+\Phi(\boldsymbol{X}^u_T)+\int_0^Tc(\boldsymbol{X}^u_t,t)dt\right]\tag{RE Loss}\label{eq:relative-entropy-vector}
    \end{align}
    \end{small}
\end{definition}

\textit{Derivation.} To derive this loss, we leverage (\ref{eq:girsanov-proof2}) and our derivation for the Radon-Nikodym derivative between the optimal and reference path measure derived in Proposition \ref{prop:optimal-path-sbsoc}. Since $\mathbb{P}^u$ is the \textit{controlled} version of the reference process $\mathbb{Q}$, we write:
\begin{align}
    \mathbb{E}_{\mathbb{P}^u}\left[\log \frac{\mathrm{d}\mathbb{P}^u}{\mathrm{d}\mathbb{P}^\star}\right]&=\mathbb{E}_{\mathbb{P}^u}\left[\log \frac{\mathrm{d}\mathbb{P}^u}{\mathrm{d}\mathbb{Q}}+\log \frac{\mathrm{d}\mathbb{Q}}{\mathrm{d}\mathbb{P}^\star}\right]=\mathbb{E}_{\mathbb{P}^u}\bigg[\log \underbrace{\frac{\mathrm{d}\mathbb{P}^u}{\mathrm{d}\mathbb{Q}}}_{\text{Girsanov}}-\log \frac{\mathrm{d}\mathbb{P}^\star}{\mathrm{d}\mathbb{Q}}\bigg]
\end{align}
Since $\mathbb{P}^u$ and $\mathbb{Q}$ differ only by the control drift $\sigma\boldsymbol{u}(\boldsymbol{x},t)$, we can apply (\ref{eq:theorem-path-rnd}) to get write the first term $\frac{\mathrm{d}\mathbb{P}^u}{\mathrm{d}\mathbb{Q}}$ as:
\begin{small}
    \begin{align}
        \frac{\mathrm{d}\mathbb{P}^u}{\mathrm{d}\mathbb{Q}}(\boldsymbol{X}_{0:T})=\exp\left(-\frac{1}{2}\int_0^T\|\boldsymbol{u}(\boldsymbol{X}_t,t)\|^2dt+\int_0^T\boldsymbol{u}(\boldsymbol{X}_t,t)^\top d\boldsymbol{B}_t\right)
    \end{align}
\end{small}
Substituting this and (\ref{eq:optimal-path-rnd}), we get the integral form of the relative-entropy objective $\mathcal{L}_{\text{RE}}$ as:
\begin{small}
\begin{align}
    &\mathcal{L}_{\text{RE}}(\boldsymbol{u})=\mathbb{E}_{\boldsymbol{X}^u_{0:T}\sim \mathbb{P}^u}\bigg[ \bluetext{\log \frac{\mathrm{d}\mathbb{P}^u}{\mathrm{d}\mathbb{Q}}(\boldsymbol{X}^u_{0:T})}- \pinktext{\log \frac{\mathrm{d}\mathbb{P}^\star}{\mathrm{d}\mathbb{Q}}(\boldsymbol{X}^u_{0:T})}\bigg]\nonumber\\
    &=\mathbb{E}_{\boldsymbol{X}^u_{0:T}\sim \mathbb{P}^u}\bigg[ \bluetext{-\frac{1}{2}\int_0^T\|\boldsymbol{u}(\boldsymbol{X}^u_t,t)\|^2dt+\int_0^T\boldsymbol{u}(\boldsymbol{X}^u_t,t)^\top d\boldsymbol{B}_t}+\pinktext{V_T(\boldsymbol{X}^u_T)-V_0(\boldsymbol{X}^u_0)+\int_0^Tc(\boldsymbol{X}^u_t,t)dt\bigg)}\bigg]
\end{align}
\end{small}
By definition, we have $V_T(\boldsymbol{X}^u_T)=\Phi(\boldsymbol{X}^u_T)$ and the initial distribution is fixed at $\boldsymbol{X}_0\sim\pi_0$ so $V_0(\boldsymbol{X}_0)$ is constant. Then, we get the final form of $\mathcal{L}_{\text{RE}}(\boldsymbol{u})$ as:
\begin{small}
\begin{align}
    \mathcal{L}_{\text{RE}}(\boldsymbol{u}):=\mathbb{E}_{\boldsymbol{X}^u_{0:T}\sim \mathbb{P}^u}\left[\frac{1}{2}\int_0^T\|\boldsymbol{u}(\boldsymbol{X}^u_t,t)\|^2dt+\Phi(\boldsymbol{X}^u_T)+\int_0^Tc(\boldsymbol{X}^u_t,t)dt\right]
\end{align}
\end{small}
which is exactly (\ref{eq:relative-entropy-vector}) in our definition. \hfill $\square$

While the relative-entropy loss has a \textit{unique minimizer} when $\mathbb{P}^u=\mathbb{P}^\star$, optimizing $\boldsymbol{u}$ using the gradient $\nabla_{u}\mathcal{L}_{\text{RE}}(\boldsymbol{u})$ with respect to $\boldsymbol{u}$ or its parameters $\nabla_\theta\mathcal{L}_{\text{RE}}(\boldsymbol{u}_\theta)$, requires differentiating through the full SDE trajectories $(\boldsymbol{X}^u_t)_{t\in [0,T]}$ due to the expectation over $\mathbb{P}^u$. In practice, if $(\boldsymbol{X}^u_t)_{t\in [0,T]}$ is generated via an SDE solver, like the Euler-Maruyama method, this requires storing the full computational graph at each simulation step, which is memory-intensive. 

To obtain a practical estimator, the \boldtext{REINFORCE trick} allows us to rewrite (\ref{eq:relative-entropy-path}) as an expectation under a stop-gradient sampling measure $\mathbb{P}^{\bar{u}}$, where $\boldsymbol{Q}^{\bar{u}}:=\texttt{stopgrad}(\boldsymbol{Q}^u)$ is the non-gradient-tracking controlled generator \citep{williams1992simple, ranganath2014black, mnih2014neural}.

\begin{definition}[REINFORCE Relative Entropy (RERF) Loss]\label{def:reinforce}
    The \textbf{REINFORCE relative entropy} (RERF) loss is defined as an expectation over $\mathbb{P}^{\bar{u}}$, where $\boldsymbol{Q}^{\bar{u}}:=\texttt{stopgrad}(\boldsymbol{Q}^u)$ is the non-gradient-tracking controlled generator:
    \begin{small}
    \begin{align}
        \mathcal{L}_{\text{RERF}}(\mathbb{P}^u, \mathbb{P}^\star):=\mathbb{E}_{\mathbb{P}^{\bar{u}}}\left[\log \frac{\mathrm{d}\mathbb{P}^u}{\mathrm{d}\mathbb{P}^{\bar{u}}}\left(\log \frac{\mathrm{d}\mathbb{P}^\star}{\mathrm{d}\mathbb{P}^{\bar{u}}}+C\right)\right]\tag{RERF Loss}\label{eq:rerf-loss}
    \end{align}
    \end{small}
    where $C\in \mathbb{R}$ is any constant. Crucially, the gradient aligns with the gradient of the relative-entropy loss, i.e., $\nabla_{\boldsymbol{u}}\text{KL}(\mathbb{P}^u\|\mathbb{P}^u)=\nabla_{\boldsymbol{u}}\mathcal{L}_{\text{RERF}}(\mathbb{P}^u, \mathbb{P}^\star)$.
\end{definition}

While the (\ref{eq:rerf-loss}) provides an unbiased estimator of the relative-entropy gradient that satisfies $\nabla_{\boldsymbol{u}}\text{KL}(\mathbb{P}^u\|\mathbb{P}^u)=\nabla_{\boldsymbol{u}}\mathcal{L}_{\text{RERF}}(\mathbb{P}^u, \mathbb{P}^\star)$, it is important to note that it should be interpreted as a computational surrogate rather than a true loss function, as decreasing its value does not necessarily correspond to a monotonic reduction of the KL divergence itself. Instead, its role is to provide a tractable estimator of the gradient needed to learn the optimal control that induces the target path measure.

An alternative objective that is a true loss function and doesn't require an expectation over $\mathbb{P}^u$ is the \boldtext{cross-entropy (CE) loss}, which is simply the \textit{reverse KL divergence}, where the expectation is over the fixed optimal path measure $\mathbb{P}^\star$ instead of $\mathbb{P}^u$.

\begin{definition}[Cross Entropy (CE) Loss]\label{def:cross-entropy}
    The \textbf{cross entropy} (CE) loss between the controlled path measure $\mathbb{P}^u$ and the optimal path measure $\mathbb{P}^\star$ is defined as the KL divergence:
    \begin{align}
        \mathcal{L}_{\text{CE}}(\mathbb{P}^u, \mathbb{P}^\star):=\text{KL}(\mathbb{P}^\star\|\mathbb{P}^u)=\mathbb{E}_{\mathbb{P}^\star}\left[\log \frac{\mathrm{d}\mathbb{P}^\star}{\mathrm{d}\mathbb{P}^u}\right]\tag{Cross Entropy Objective}\label{eq:cross-entropy-path}
    \end{align}
    Let $\boldsymbol{v}(\boldsymbol{x},t)$ denote an arbitrary \textbf{fixed} control drift that generates $\mathbb{P}^v$ and let $\boldsymbol{X}^v_{0:T}=(\boldsymbol{X}^v_t)_{t\in [0,T]}$ denote the stochastic process under $\mathbb{P}^v$. Then, the CE loss takes the path-integral form:
    \begin{small}
    \begin{align}
        &\mathcal{L}_{\text{CE}}(\boldsymbol{u})=\frac{1}{Z}\mathbb{E}_{\boldsymbol{X}_{0:T}^v\sim\mathbb{P}^v}\bigg[\exp\left(-g(\boldsymbol{X}^v_T)-\int_0^Tc(\boldsymbol{X}^v_t,t)dt-\frac{1}{2}\int_0^T\|\boldsymbol{v}(\boldsymbol{X}^v_t,t)\|^2dt-\int_0^T\boldsymbol{v}(\boldsymbol{X}^v_t,t)^\top d\boldsymbol{B}^v_t\right)\nonumber\\
        &\left(\frac{1}{2}\int_0^T\|\boldsymbol{u}(\boldsymbol{X}^v_t,t)\|^2dt-\int_0^T(\boldsymbol{u}\cdot \boldsymbol{v})(\boldsymbol{X}^v_t,t)dt-\int_0^T\boldsymbol{u}(\boldsymbol{X}^v_t,t)^\top d\boldsymbol{B}^v_t-g(\boldsymbol{X}^v_T)-\int_0^Tc(\boldsymbol{X}^v_t,t)dt\right)\bigg]+C\tag{CE Loss}\label{eq:soc-cross-entropy}
    \end{align}
    \end{small}
\end{definition}

\textit{Derivation.} This proof follows similar steps to the relative-entropy
\begin{small}
\begin{align}
\mathbb{E}_{\mathbb{P}^\star}\left[\log \frac{\mathrm{d}\mathbb{P}^\star}{\mathrm{d}\mathbb{P}^u}\right]=\mathbb{E}_{\mathbb{P}^v}\left[\frac{\mathrm{d}\mathbb{P}^\star}{\mathrm{d}\mathbb{P}^v}\log \frac{\mathrm{d}\mathbb{P}^\star}{\mathrm{d}\mathbb{P}^u}\right]&=\mathbb{E}_{\mathbb{P}^v}\left[\frac{\mathrm{d}\mathbb{P}^\star}{\mathrm{d}\mathbb{P}^v}\log \frac{\mathrm{d}\mathbb{P}^\star}{\mathrm{d}\mathbb{Q}}\frac{\mathrm{d}\mathbb{Q}}{\mathrm{d}\mathbb{P}^u}\right]\nonumber\\
&=\mathbb{E}_{\mathbb{P}^v}\bigg[\underbrace{\frac{\mathrm{d}\mathbb{P}^\star}{\mathrm{d}\mathbb{Q}}\frac{\mathrm{d}\mathbb{Q}}{\mathrm{d}\mathbb{P}^v}}_{(\bigstar)}\underbrace{\left(\log \frac{\mathrm{d}\mathbb{P}^\star}{\mathrm{d}\mathbb{Q}}-\log \frac{\mathrm{d}\mathbb{P}^u}{\mathrm{d}\mathbb{Q}}\right)}_{(\diamond)}\bigg]
\end{align}
\end{small}
Applying Girsanov's theorem and substituting (\ref{eq:optimal-path-rnd}) into ($\diamond$), we get the integral form of the relative-entropy objective $\mathcal{L}_{\text{RE}}$ as:
\begin{small}
\begin{align}
    &\bluetext{\log \frac{\mathrm{d}\mathbb{P}^\star}{\mathrm{d}\mathbb{Q}}(\boldsymbol{X}^v_{0:T})}-\pinktext{\log \frac{\mathrm{d}\mathbb{P}^u}{\mathrm{d}\mathbb{Q}}(\boldsymbol{X}^v_{0:T})}\nonumber\\
    &=\bluetext{-V_T(\boldsymbol{X}^v_T)+V_0(\boldsymbol{X}^v_0)-\int_0^Tc(\boldsymbol{X}^v_t,t)dt}\pinktext{+\frac{1}{2}\int_0^T\|\boldsymbol{u}(\boldsymbol{X}^v_t,t)\|^2dt-\int_0^T\boldsymbol{u}(\boldsymbol{X}^u_t,t)^\top d\boldsymbol{B}^{\mathbb{Q}}_t}\nonumber\\
    &=\underbrace{\frac{1}{2}\int_0^T\|\boldsymbol{u}(\boldsymbol{X}^v_t,t)\|^2dt-\int_0^T\boldsymbol{u}(\boldsymbol{X}^v_t,t)^\top d\boldsymbol{B}_t-g(\boldsymbol{X}^v_T)-\int_0^Tc(\boldsymbol{X}^v_t,t)dt+C}_{(\diamond)}\label{eq:ce-proof1}
\end{align}
\end{small}
Since the Brownian motion of the RND is under the reference path measure $\mathbb{Q}$ in the denominator, we need to transform it to match the path measure under which $\boldsymbol{X}_{0:T}^v$ is generated. By (\ref{eq:girsanov-proof2}), we substitute $d\boldsymbol{B}^{\mathbb{Q}}_t=d\boldsymbol{B}^v_t+\boldsymbol{v}(\boldsymbol{X}_t,t)dt$ which gives us:
\begin{small}
\begin{align}
    &-V_T(\boldsymbol{X}^v_T)+V_0(\boldsymbol{X}^v_0)-\int_0^Tc(\boldsymbol{X}^v_t,t)dt+\frac{1}{2}\int_0^T\|\boldsymbol{u}(\boldsymbol{X}^v_t,t)\|^2dt-\int_0^T\boldsymbol{u}(\boldsymbol{X}^v_t,t)^\top \bluetext{(d\boldsymbol{B}^v_t+\boldsymbol{v}(\boldsymbol{X}^v_t,t)dt)}\nonumber\\
    &=\underbrace{\frac{1}{2}\int_0^T\|\boldsymbol{u}(\boldsymbol{X}^v_t,t)\|^2dt\bluetext{-\int_0^T(\boldsymbol{u}\cdot \boldsymbol{v})(\boldsymbol{X}^v_t,t)dt-\int_0^T\boldsymbol{u}(\boldsymbol{X}^v_t,t)^\top d\boldsymbol{B}^v_t}-g(\boldsymbol{X}^v_T)-\int_0^Tc(\boldsymbol{X}^v_t,t)dt+C}_{(\diamond)}
\end{align}
\end{small}
where we set $V_T(\boldsymbol{X}^v_T)=g(\boldsymbol{X}^v_T)$ and denote $V_0(\boldsymbol{X}^v_0)$ as a constant. For $(\bigstar)$, we can derive the (\ref{eq:theorem-path-rnd}) of $\mathbb{Q}$ with respect to $\mathbb{P}^v$, where the difference between the control drifts is $-\boldsymbol{v}(\boldsymbol{X}^v_t,t)$, and substitute (\ref{eq:optimal-path-rnd}) to get:
\begin{small}
\begin{align}
    &\frac{\mathrm{d}\mathbb{P}^\star}{\mathrm{d}\mathbb{Q}}(\boldsymbol{X}^v_{0:T})\frac{\mathrm{d}\mathbb{Q}}{\mathrm{d}\mathbb{P}^v}(\boldsymbol{X}^v_{0:T})\nonumber\\
    &=\bluetext{\frac{1}{Z}\exp\left(\bluetext{-V_T(\boldsymbol{X}^v_T)+V_0(\boldsymbol{X}^v_0)-\int_0^Tc(\boldsymbol{X}^v_t,t)dt}\right)}\pinktext{\exp\left(-\frac{1}{2}\int_0^T\|\boldsymbol{v}(\boldsymbol{X}^v_t,t)\|^2dt-\int_0^T\boldsymbol{v}(\boldsymbol{X}^v_t,t)^\top d\boldsymbol{B}^v_t\right)}\nonumber\\
    &=\underbrace{\frac{1}{Z}\exp\left(-g(\boldsymbol{X}^v_T)-\int_0^Tc(\boldsymbol{X}^v_t,t)dt-\frac{1}{2}\int_0^T\|\boldsymbol{v}(\boldsymbol{X}^v_t,t)\|^2dt-\int_0^T\boldsymbol{v}(\boldsymbol{X}^v_t,t)^\top d\boldsymbol{B}^v_t\right)}_{(\bigstar)}\label{eq:ce-proof2}
\end{align}
\end{small}
Substituting the expanded forms for $(\bigstar)$ and $(\diamond)$, we get:
\begin{small}
\begin{align}
    \mathcal{L}_{\text{CE}}(\boldsymbol{u})&=\frac{1}{Z}\mathbb{E}_{\boldsymbol{X}_{0:T}^v\sim\mathbb{P}^v}\bigg[\underbrace{\exp\left(-g(\boldsymbol{X}^v_T)-\int_0^Tc(\boldsymbol{X}^v_t,t)dt-\frac{1}{2}\int_0^T\|\boldsymbol{v}(\boldsymbol{X}^v_t,t)\|^2dt-\int_0^T\boldsymbol{v}(\boldsymbol{X}^v_t,t)^\top d\boldsymbol{B}^v_t\right)}_{(\bigstar)}\nonumber\\
    &\underbrace{\left(\frac{1}{2}\int_0^T\|\boldsymbol{u}(\boldsymbol{X}^v_t,t)\|^2dt-\int_0^T(\boldsymbol{u}\cdot \boldsymbol{v})(\boldsymbol{X}^v_t,t)dt-\int_0^T\boldsymbol{u}(\boldsymbol{X}^v_t,t)^\top d\boldsymbol{B}^v_t-g(\boldsymbol{X}^v_T)-\int_0^Tc(\boldsymbol{X}^v_t,t)dt\right)}_{(\diamond)}\bigg]+C
\end{align}
\end{small}
which recovers the (\ref{eq:soc-cross-entropy}) from our definition. \hfill $\square$

Since the paths $\boldsymbol{X}_{0:T}^v$ are generated under the fixed controlled process $\mathbb{P}^v$ instead of the path that is being optimized $\mathbb{P}^u$, it is considered an \boldtext{off-policy objective}. This admits a much more computationally tractable objective as the gradient $\nabla_{\boldsymbol{u}}\mathcal{L}_{\text{CE}}(\boldsymbol{u})$ no longer depends on the SDE trajectories, and computing the objective does not require differentiating through or maintaining the computational graph of the SDE solver. Typically, the off-policy control $\boldsymbol{v}$ is defined as the non-gradient-tracked control being optimized $\boldsymbol{v}:=\texttt{stopgrad}(\boldsymbol{u})$, which generates paths from the controlled SDE without maintaining the computational graph used to generate it. We also note that while both $\mathcal{L}_{\text{RE}}$ and $\mathcal{L}_{\text{CE}}$ are \textit{uniquely minimized} at $\mathbb{P}^u = \mathbb{P}^\star$, the $\mathcal{L}_{\text{CE}}$ is \textbf{convex in $\mathbb{P}^u$}, which yields a more favorable optimization landscape. 

\begin{remark}[Convexity of the Cross-Entropy Objective]\label{remark:ce-convex}
The cross-entropy loss $\mathcal{L}_{\mathrm{CE}}(\mathbb{P}^u,\mathbb{P}^\star) = \mathrm{KL}(\mathbb{P}^\star \| \mathbb{P}^u)$ is \textbf{convex in the controlled path measure $\mathbb{P}^u$}. This follows from the fact that $\mathbb{P}^u\mapsto  \log \mathbb{P}^u$ is convex when the reference distribution $\mathbb{P}^\star$ is fixed. Therefore, minimizing the CE objective corresponds to a convex optimization problem in the space of path measures, and the global minimum is achieved when $\mathbb{P}^u = \mathbb{P}^\star$. 
\end{remark}

Now that we have defined two objectives corresponding to the KL divergence between path measures, we consider an alternative class of objectives that aims to minimize the \boldtext{variance} between the controlled and optimal path measures. To build the intuition behind the variance-based losses, we recall that a stochastic process $\boldsymbol{X}_{0:T}^u\sim\mathbb{P}^u$ generated under a controlled process can be \textit{reweighted} by the Radon-Nikodym derivative such that the law matches the optimal path measure $\mathbb{P}^\star$:
\begin{small}
\begin{align}
    \underbrace{\mathbb{P}^\star(\boldsymbol{X}^u_{0:T})}_{\text{optimal measure}}=\underbrace{\frac{\mathrm{d}\mathbb{P}^\star}{\mathrm{d}\mathbb{P}^u}(\boldsymbol{X}^u_{0:T})}_{\text{importance weight}}\underbrace{\mathbb{P}^u(\boldsymbol{X}^u_{0:T})}_{\text{paths under }\boldsymbol{u}}
\end{align}
\end{small}
However, when $\mathbb{P}^u$ deviates far from the optimal measure $\mathbb{P}^\star$, the importance weights will vary significantly for different paths $\boldsymbol{X}_{0:T}^u$, with some weights being huge and others small. This means that the \boldtext{variance of the importance weight} can be a measure of the \textit{similarity between $\mathbb{P}^u$ and $\mathbb{P}^\star$}. Crucially, the variance is only zero when the importance weight is a constant that is not dependent on the path $\boldsymbol{X}_{0:T}^u$. Since the RND is constant only when the two measures are equal $\mathbb{P}^u=\mathbb{P}^\star$ up to a normalization, we have:
\begin{align}
    \text{Var}_{\boldsymbol{X}_{0:T}^u\sim\mathbb{P}^u}\left(\frac{\mathrm{d}\mathbb{P}^\star}{\mathrm{d}\mathbb{P}^u}(\boldsymbol{X}_{0:T}^u)\right)=0\iff \frac{\mathrm{d}\mathbb{P}^\star}{\mathrm{d}\mathbb{P}^u}(\boldsymbol{X}_{0:T}^u)=\text{const.}\iff\mathbb{P}^u=\mathbb{P}^\star
\end{align}
Similar to the cross-entropy objective, we want to prevent taking gradients under paths generated directly from the controlled path measure that we are optimizing $\mathbb{P}^u$, so we define an arbitrary, fixed control $\boldsymbol{v}$ and sample paths $\boldsymbol{X}_{0:T}^v\sim \mathbb{P}^v$. Using these paths \textit{off-policy paths}, the importance weights are still minimized when the RND is constant. Using this idea, we define the \boldtext{variance loss} and \boldtext{log-variance loss}.

\begin{definition}[Variance and Log-Variance Losses]\label{def:soc-log-variance}
    The \textbf{variance} and \textbf{log-variance} (LV) loss between the controlled path measure $\mathbb{P}^u$ and the optimal path measure $\mathbb{P}^\star$ is defined as the KL divergence:
    \begin{align}
        \mathcal{L}_{\text{Var}_{\mathbb{P}^v}}(\mathbb{P}^u, \mathbb{P}^\star):=\text{Var}_{\mathbb{P}^v}\left( \frac{\mathrm{d}\mathbb{P}^\star}{\mathrm{d}\mathbb{P}^u}\right), \quad\mathcal{L}^{\text{log}}_{\text{Var}}(\mathbb{P}^u, \mathbb{P}^\star):=\text{Var}_{\mathbb{P}^v}\left(\log \frac{\mathrm{d}\mathbb{P}^\star}{\mathrm{d}\mathbb{P}^u}\right)
    \end{align}
    Let $\boldsymbol{u}(\boldsymbol{x},t)$ denote the control being optimized and $\boldsymbol{v}(\boldsymbol{x},t)$ denote an arbitrary fixed control that generates the stochastic paths $(\boldsymbol{X}_t^v)_{t\in [0,T]}$. Then, the variance and log-variance losses take the path-integral form:
    \begin{small}
    \begin{align}
        \mathcal{L}_{\text{Var}_{\mathbb{P}^v}}(\boldsymbol{u}):=\frac{1}{Z^2}\text{Var}_{\mathbb{P}^v}\left(e^{\mathcal{F}_{u,v}-\Phi(\boldsymbol{X}_T^v)}\right), \quad\mathcal{L}^{\text{log}}_{\text{Var}}(\boldsymbol{u}):=\text{Var}_{\mathbb{P}^v}\left(\mathcal{F}_{u,v}-\Phi(\boldsymbol{X}_T^v)\right)\tag{Variance Losses}\label{eq:variance-losses}
    \end{align}
    \end{small}
    where $\mathcal{F}_{u,v}$ is given by the expression:
    \begin{small}
    \begin{align}
        \mathcal{F}_{u,v}=\frac{1}{2}\int_0^T\|\boldsymbol{u}(\boldsymbol{X}_t^v,t)\|^2dt-\int_0^T(\boldsymbol{u}\cdot \boldsymbol{v})(\boldsymbol{X}^v_t,t)dt-\int_0^T\boldsymbol{u}(\boldsymbol{X}_t^v,t)^\top d \boldsymbol{B}_t-\int_0^Tc(\boldsymbol{X}_t^v ,t)dt
    \end{align}
    \end{small}
\end{definition}

\textit{Derivation.} We can break down the variance losses as follows:
\begin{small}
\begin{align}
    \text{Var}_{\mathbb{P}^v}\left( \frac{\mathrm{d}\mathbb{P}^\star}{\mathrm{d}\mathbb{P}^u}\right)&=\text{Var}_{\mathbb{P}^v}\bigg( \frac{\mathrm{d}\mathbb{P}^\star}{\mathrm{d}\mathbb{Q}}\frac{\mathrm{d}\mathbb{Q}}{\mathrm{d}\mathbb{{P}}^u}\bigg)
    , \quad\text{Var}_{\mathbb{P}^v}\left( \log \frac{\mathrm{d}\mathbb{P}^\star}{\mathrm{d}\mathbb{P}^u}\right)=\text{Var}_{\mathbb{P}^v}\bigg(\log  \frac{\mathrm{d}\mathbb{P}^\star}{\mathrm{d}\mathbb{Q}}-\log \frac{\mathrm{d}\mathbb{P}^u}{\mathrm{d}\mathbb{{Q}}}\bigg)
\end{align}
\end{small}
Expanding these terms similar form as (\ref{eq:ce-proof1}), where $d\boldsymbol{B}^{\mathbb{Q}}_t=d\boldsymbol{B}^v_t+\boldsymbol{v}(\boldsymbol{X}_t^v,t)dt$, we get:
\begin{small}
\begin{align}
    &\text{Var}_{\boldsymbol{X}_{0:T}^v\sim\mathbb{P}^v}\left( \frac{\mathrm{d}\mathbb{P}^\star}{\mathrm{d}\mathbb{P}^u}(\boldsymbol{X}_{0:T}^v)\right)=\text{Var}_{\boldsymbol{X}_{0:T}^v\sim \mathbb{P}^v}\bigg( \frac{\mathrm{d}\mathbb{P}^\star}{\mathrm{d}\mathbb{Q}}\frac{\mathrm{d}\mathbb{Q}}{\mathrm{d}\mathbb{{P}}^u}(\boldsymbol{X}_{0:T}^v)\bigg)\nonumber\\
    &=\bluetext{\frac{1}{Z^2}}\text{Var}_{\boldsymbol{X}_{0:T}^v\sim\mathbb{P}^v}\bigg(\bluetext{\exp\bigg(\underbrace{\frac{1}{2}\int_0^T\|\boldsymbol{u}(\boldsymbol{X}^v_t,t)\|^2dt-\int_0^T(\boldsymbol{u}\cdot\boldsymbol{v})(\boldsymbol{X}^v_t,t)dt-\int_0^T\boldsymbol{u}(\boldsymbol{X}^v_t,t)^\top d\boldsymbol{B}^v_t}_{=:\mathcal{F}_{u,v}}-g(\boldsymbol{X}^v_T)-\int_0^Tc(\boldsymbol{X}^v_t,t)dt\bigg)}\bigg)\nonumber\\
    &\text{Var}_{\boldsymbol{X}_{0:T}^v\sim\mathbb{P}^v}\left( \log\frac{\mathrm{d}\mathbb{P}^\star}{\mathrm{d}\mathbb{P}^u}(\boldsymbol{X}_{0:T}^v) \right)=\text{Var}_{\boldsymbol{X}_{0:T}^v\sim \mathbb{P}^v}\bigg(\log  \frac{\mathrm{d}\mathbb{P}^\star}{\mathrm{d}\mathbb{Q}}(\boldsymbol{X}_{0:T}^v)-\log \frac{\mathrm{d}\mathbb{P}^u}{\mathrm{d}\mathbb{{Q}}}(\boldsymbol{X}_{0:T}^v)\bigg)\nonumber\\
    &=\text{Var}_{\boldsymbol{X}_{0:T}^v\sim\mathbb{P}^v}\bigg( \bluetext{\underbrace{\frac{1}{2}\int_0^T\|\boldsymbol{u}(\boldsymbol{X}^v_t,t)\|^2dt-\int_0^T(\boldsymbol{u}\cdot\boldsymbol{v})(\boldsymbol{X}^v_t,t)dt-\int_0^T\boldsymbol{u}(\boldsymbol{X}^v_t,t)^\top d\boldsymbol{B}^v_t}_{=:\mathcal{F}_{u,v}}}-g(\boldsymbol{X}^v_T)-\int_0^Tc(\boldsymbol{X}^v_t,t)dt+C\bigg)\nonumber
\end{align}
\end{small}
Then, we can define $\mathcal{F}_{u,v}:=\frac{1}{2}\int_0^T\|\boldsymbol{u}(\boldsymbol{X}^v_t,t)\|^2dt-\int_0^T(\boldsymbol{u}\cdot\boldsymbol{v}) (\boldsymbol{X}^v_t,t)dt-\int_0^T\boldsymbol{u}(\boldsymbol{X}^v_t,t)^\top d\boldsymbol{B}^v_t$ and rewrite the variance losses as:
\begin{small}
\begin{align}
    \mathcal{L}_{\text{Var}_{\mathbb{P}^v}}(\boldsymbol{u}):=\frac{1}{Z^2}\text{Var}_{\mathbb{P}^v}\left(e^{\bluetext{\mathcal{F}_{u,v}}-\Phi(\boldsymbol{X}_T^v)}\right), \quad\mathcal{L}^{\text{log}}_{\text{Var}_{\mathbb{P}^v}}(\boldsymbol{u}):=\text{Var}_{\mathbb{P}^v}\left(\bluetext{\mathcal{F}_{u,v}}-\Phi(\boldsymbol{X}_T^v)\right)
\end{align}
\end{small}
which recovers the (\ref{eq:variance-losses}) from our definition.\hfill $\square$

Comparing the (\ref{eq:soc-cross-entropy}) and (\ref{eq:variance-losses}), we observe that they both sample paths from an arbitrary fixed path measure $\mathbb{P}^v$ with control $\boldsymbol{v}$. However, the (\ref{eq:soc-cross-entropy}) does not change with different choices of $\boldsymbol{v}$, since the dependence on $\mathbb{P}^v$ \textit{cancels out} when scaling with the RND $\mathrm{d}\mathbb{P}^\star/\mathrm{d}\mathbb{P}^v$. In contrast, the variance-based objectives change with respect to different definitions of $\boldsymbol{v}$, since the distribution under which the variance is calculated $\mathbb{P}^v$, changes with different $\boldsymbol{v}$.

\begin{remark}[Variance Loss Depends on Sampling Distribution]
    The variance-based objectives in (\ref{eq:variance-losses}) depend on the definition of $\boldsymbol{v}$ used to sample $\boldsymbol{X}_{0:T}^v\sim \mathbb{P}^v$. However, it holds that the variance is minimized exactly when the controlled measure $\mathbb{P}^u$ matches the optimal measure $\mathbb{P}^\star$:
    \begin{small}
    \begin{align}
        \forall \boldsymbol{u}\in \mathcal{U}, \quad \mathcal{L}_{\text{Var}_{\mathbb{P}^v}}(\boldsymbol{u})=0 \iff\boldsymbol{u}=\boldsymbol{u}^\star\quad\text{and}\quad \mathcal{L}^{\text{log}}_{\text{Var}_{\mathbb{P}^v}}(\boldsymbol{u})=0 \iff\boldsymbol{u}=\boldsymbol{u}^\star
    \end{align}
    \end{small}
    This is easy to see given that the RND between the optimal path measure and itself must be equal to one, i.e.,  $\mathrm{d}\mathbb{P}^\star/\mathrm{d}\mathbb{P}^\star\equiv 1$, regardless of the path on which it is evaluated, and the variance of a constant is zero.
\end{remark}

We have introduced three variations of the SOC objective, which are all derived from the RND between path measures. The \textbf{key insight} is that all three objectives can be \textbf{estimated from paths generated from a tractable SDE}, enabling practical learning of the optimal bridge dynamics. Although the objectives differ in their sampling laws, training stability, and convergence guarantees, they are all minimized when the controlled dynamics recover the optimal bridge $\mathbb{P}^u=\mathbb{P}^\star$.

\subsection{Closing Remarks for Section \ref{sec:sb-optimal-control}}
In this section, we introduced the stochastic optimal control (SOC) problem, which defines a path-space variational objective that minimizes a running cost and terminal cost function generated by a \textit{controlled SDE}. Leveraging dynamic programming theory, we introduce the \textbf{value function} $V_t$, which solves the Hamilton-Jacobi-Bellman equation similarly to the Lagrangian $\psi_t$ defined in Section \ref{subsec:nonlinear-sbp} and connect it to the optimal control drift $\boldsymbol{u}^\star$ and optimal path measure Radon-Nikodym derivative.

A \textbf{key observation} is that the quadratic control cost in the SOC objective is equivalent to the KL divergence between the controlled and reference Itô processes from Corollary \ref{corollary:kl-divergence-ito}. This connection allows us to reformulate the dynamic Schrödinger bridge problem as an SOC objective in which the terminal cost is expressed as a log-ratio involving the backward Schrödinger potential $\hat\varphi_T$ and the terminal marginal constraint $\pi_T$. This reformulation avoids the need for explicit couplings between $\pi_0$ and $\pi_T$, and we show that the backward potential $\hat\varphi_T$ effectively absorbs the dependence on the initial value $V_0(\boldsymbol{X}_0)$ appearing in classical SOC formulations, making the framework applicable to arbitrary initial distributions and reference dynamics.

Building on this formulation, we introduced several tractable training objectives that are uniquely minimized by the optimal control drift $\boldsymbol{u}^\star$ and its corresponding path measure $\mathbb{P}^\star$. A \textbf{key step} in deriving these objectives is using Girsanov’s theorem to express the Radon–Nikodym derivative between the controlled and reference path measures directly in terms of the control drift, yielding closed-form expressions and a practical way to evaluate and optimize path-space divergences using trajectories simulated from the controlled SDE.

Now that we have explored several complementary formulations of the SB problem, from the static formulation to dynamic formulations with connections to SOC, we next turn to the practical construction of Schrödinger bridges. While the SOC objectives introduced in Section \ref{subsec:soc-objectives} already provide a glimpse of how such optimization can be carried out, the next section will present several concrete algorithms and computational strategies for building stochastic bridges and Schrödinger bridges in practice.

\newpage
\section{Building Schrödinger Bridges}
\label{sec:building-stochastic-bridge}
While the Schrödinger bridge problem provides a variational characterization of the optimal stochastic transport between two marginals, this formulation alone does not immediately reveal how such bridges can be constructed in practice. In this section, we move from the abstract formulation to explore several complementary mechanisms for building stochastic bridges between prescribed endpoint distributions.

We begin by interpreting Schrödinger bridges as a mixture of conditional bridges (Section \ref{subsect:mixture-bridges}), which provides an intuitive pathwise construction. We then study time reversal (Section \ref{subsec:time-reversal}) and the resulting forward–backward SDE representation (Section \ref{subsec:forward-backward-sde}), revealing how drift corrections arise from the score of the evolving distribution. Next, we present Doob’s $h$-transform (Section \ref{subsec:doob-transform}) as a change-of-measure construction that enforces endpoint constraints, followed by an interpretation in terms of Markovian and reciprocal projections (Section \ref{subsec:markov-reciprocal-proj}) as entropy-minimizing projections in path space. Finally, we connect these ideas to the stochastic interpolant framework (Section \ref{subsec:stochastic-interpolants}), which provides practical parameterizations of bridges used in modern generative modeling. 

Throughout this section, we will use a slight abuse of notation and define the \textit{transition density} from $\boldsymbol{X}_t$ at time $t$ to $\boldsymbol{X}_\tau$ at a later time $\tau \geq t$ under a path measure as $\mathbb{P}_{\tau |t}(\boldsymbol{x}_\tau|\boldsymbol{x}_t)$ or equivalently $\mathbb{P}_{\tau |t}(\boldsymbol{X}_\tau=\boldsymbol{x}_\tau|\boldsymbol{X}_t=\boldsymbol{x}_t)$. For instance, the transition density from $\boldsymbol{x}$ at time $t$ to $\boldsymbol{x}_T$ at time $T$ under the reference measure $\mathbb{Q}$ is denoted $\mathbb{Q}_{T|t}(\boldsymbol{x}_T|\boldsymbol{x})=\mathbb{Q}(\boldsymbol{X}_T=\boldsymbol{x}_T|\boldsymbol{X}_t=\boldsymbol{x})$.

\subsection{Mixture of Conditional Bridges}
\label{subsect:mixture-bridges}
A useful way to understand the structure of Schrödinger bridges is through endpoint conditioning. Rather than viewing the bridge as a single global stochastic process, we can interpret it as a \textbf{mixture of conditional bridges} under the reference process $\mathbb{Q}$ connecting specific endpoint pairs $(\boldsymbol{x}_0, \boldsymbol{x}_T)$ drawn from a coupling $\pi_{0,T}$. Each conditional bridge describes the distribution of paths under the reference dynamics conditioned on fixed endpoints, which recovers the optimal Schrödinger bridge when conditioned on the \textbf{optimal coupling} $\pi^\star_{0,T}$ that solves the static Schrödinger bridge problem. 

\begin{proposition}[Mixture of Endpoint-Conditioned Bridges]\label{prop:mixture-of-bridges}
    Consider the dynamic SB problem with reference process $\mathbb{Q}$ and marginal constraints $\pi_0,\pi_T\in \mathcal{P}(\mathbb{R}^d)$ defined as:
    \begin{small}
    \begin{align}
        \mathbb{P}^\star=\underset{\mathbb{P}^u}{\arg\min}\left\{\text{KL}\left(\mathbb{P}^u\|\mathbb{Q}\right):\mathbb{P}^u_0=\pi_0, \mathbb{P}^u_T=\pi_T\right\}
    \end{align}
    \end{small}
    Then, the unique minimizer $\mathbb{P}^\star$ can be written as a \textbf{mixture of endpoint-conditioned bridges} $\mathbb{Q}(\cdot|\boldsymbol{x}_0, \boldsymbol{x}_T)$:
    \begin{small}
    \begin{align}
        \mathbb{P}^\star(\boldsymbol{X}_{0:T})=\int\mathbb{Q}(\boldsymbol{X}_{0:T}|\boldsymbol{x}_0, \boldsymbol{x}_T)\pi^\star_{0,T}(d\boldsymbol{x}_0,d\boldsymbol{x}_T)\tag{Mixture of Bridges}\label{eq:mixture-of-bridge}
    \end{align}
    \end{small}
    where $\pi^\star_{0,T}$ is the optimal coupling that solves the \textbf{static Schrödinger bridge problem} with the reference coupling defined as the endpoint law of the reference process $\mathbb{Q}_{0,T}$ given by:
    \begin{small}
    \begin{align}
        \pi^\star_{0,T}=\underset{\pi_{0,T}\in \Pi(\pi_0, \pi_T)}{\arg\min}\text{KL}(\pi_{0,T}\|\mathbb{Q}_{0,T})
    \end{align}
    \end{small}
\end{proposition}

\textit{Proof.} By the law of total probability, we can decompose the reference path measure $\mathbb{Q}$ as:
\begin{small}
\begin{align}
    \mathbb{Q}(\boldsymbol{X}_{0:T})=\mathbb{Q}(\boldsymbol{X}_{0:T}|\boldsymbol{X}_0=\boldsymbol{x}_0, \boldsymbol{X}_T=\boldsymbol{x}_T)\mathbb{Q}_{0,T}(\boldsymbol{x}_0, \boldsymbol{x}_T)
\end{align}
\end{small}
By definition, the conditional law of the reference process $\mathbb{Q}(\cdot|\boldsymbol{x}_0, \boldsymbol{x}_T)$ is the bridge connecting $\boldsymbol{x}_0$ and $\boldsymbol{x}_T$. Similarly, we can decompose any candidate path measure $\mathbb{P}^u$ with endpoint law $\mathbb{P}_{0,T}\equiv\pi_{0,T}$ as:
\begin{small}
\begin{align}
    \mathbb{P}(\boldsymbol{X}_{0:T})=\mathbb{P}(\boldsymbol{X}_{0:T}|\boldsymbol{X}_0=\boldsymbol{x}_0, \boldsymbol{X}_T=\boldsymbol{x}_T)\pi_{0,T}(\boldsymbol{x}_0, \boldsymbol{x}_T)
\end{align}
\end{small}
Applying the (\ref{eq:chain-rule-kl}), we can split the dynamic SB objective into the KL divergence of the endpoint couplings and conditional KL divergence between the endpoint-conditioned path distributions:
\begin{small}
\begin{align}
    \text{KL}(\mathbb{P}^u\|\mathbb{Q})=\underbrace{\text{KL}(\pi_{0,T}\|\mathbb{Q}_{0,T})}_{(\bigstar)}+\bluetext{\underbrace{\mathbb{E}_{(\boldsymbol{x}_0, \boldsymbol{x}_T)\sim \pi_{0,T}}\left[\text{KL}\big(\mathbb{P}(\cdot|\boldsymbol{x}_0, \boldsymbol{x}_T)\|\mathbb{Q}(\cdot|\boldsymbol{x}_0, \boldsymbol{x}_T)\big)\right]}_{(\diamond)}}
\end{align}
\end{small}
Clearly, for any fixed endpoint law $\pi_{0,T}$ the minimizer of ($\diamond$) is achieved when $\mathbb{P}(\cdot|\boldsymbol{x}_0, \boldsymbol{x}_T)=\mathbb{Q}(\cdot|\boldsymbol{x}_0, \boldsymbol{x}_T)$ and the KL is zero. Therefore, \textbf{solving the dynamic SB over the restricted space of path measures where $\mathbb{P}(\cdot|\boldsymbol{x}_0, \boldsymbol{x}_T)=\mathbb{Q}(\cdot|\boldsymbol{x}_0, \boldsymbol{x}_T)$ reduces to solving the static SB problem} which yields the optimal endpoint law:
\begin{small}
\begin{align}
    \pi^\star_{0,T}=\underset{\pi_{0,T}\in \Pi(\pi_0, \pi_T)}{\arg\min}\text{KL}(\pi_{0,T}\|\mathbb{Q}_{0,T})
\end{align}
\end{small}
Since we have shown that the optimal $\mathbb{P}^\star$ satisfies $\mathbb{P}^\star(\cdot|\boldsymbol{x}_0, \boldsymbol{x}_T)=\mathbb{Q}(\cdot|\boldsymbol{x}_0, \boldsymbol{x}_T)$ and $\mathbb{P}^\star_{0,T}=\pi^\star_{0,T}$, we can conclude that the minimizer is given by:
\begin{small}
\begin{align}
    \boxed{\mathbb{P}^\star(\boldsymbol{X}_{0:T})=\int\mathbb{Q}(\boldsymbol{X}_{0:T}|\boldsymbol{x}_0, \boldsymbol{x}_T)\pi^\star_{0,T}(d\boldsymbol{x}_0,d\boldsymbol{x}_T)}
\end{align}
\end{small}
which is \textit{unique} following proof of uniqueness of the static SB problem in Proposition \ref{prop:solution-static-sb}. \hfill $\square$

We note that this result is a generalization of the result from \citet{follmer2006random}, which states that the solution to the dynamic SB problem with a pure Brownian reference process is given by the \textbf{mixture of endpoint-conditioned Brownian bridges} weighted by the \textbf{entropic OT plan} with quadratic transport cost.

\begin{corollary}[Mixture of Brownian Bridges (\citet{follmer2006random})]\label{corollary:mixture-brownian}
    Consider the dynamic SB problem with Brownian reference process $\mathbb{Q}:d\boldsymbol{X}_t=\sigma_td\boldsymbol{B}_t$ and marginal constraints $\pi_0,\pi_T\in \mathcal{P}(\mathbb{R}^d)$. Then, the unique minimizer $\mathbb{P}^\star$ can be written as a \textbf{mixture of endpoint-conditioned Brownian bridges} $\mathbb{Q}(\cdot|\boldsymbol{x}_0, \boldsymbol{x}_T)$ given by:
    \begin{small}
    \begin{align}
        \mathbb{P}^\star(\boldsymbol{X}_{0:T})=\int\mathbb{Q}(\boldsymbol{X}_{0:T}|\boldsymbol{x}_0, \boldsymbol{x}_T)\pi^\star_{0,T}(d\boldsymbol{x}_0,d\boldsymbol{x}_T)\tag{Mixture of Brownian Bridges}
    \end{align}
    \end{small}
    where $\pi^\star_{0,T}$ solves the (\ref{eq:entropic-ot-problem}) with quadratic state cost $c(\boldsymbol{x}, \boldsymbol{y}):= \|\boldsymbol{x}-\boldsymbol{y}\|^2 $.
\end{corollary}

This factorized definition of the (\ref{eq:dynamic-sb-problem}) can be used to define a tractable objective based on \boldtext{conditional stochastic optimal control}, which samples pairs from the optimal marginal law $\pi^\star_{0,T}$ and optimizes the interpolating controlled dynamics such that they minimize the KL divergence from the reference drift.

\begin{proposition}[Conditional Stochastic Optimal Control (Proposition 2 in \citet{liu2023generalized})]\label{prop:condsoc-mixture}
    Consider the controlled path measure $\mathbb{P}^u$, where the marginal density $p_t$ can be factorized as $p_t(\boldsymbol{x})=\int p_t(\boldsymbol{x}|\boldsymbol{x}_0, \boldsymbol{x}_T)\pi_{0,T}(d\boldsymbol{x}_0, d\boldsymbol{x}_T)$. Given the optimal endpoint distribution $\pi^\star_{0,T}$ that solves the \textbf{static SB problem} with reference distribution $\mathbb{Q}_{0,T}$, the \textbf{dynamic SB problem} objective decomposes into a \textbf{mixture of conditional stochastic optimal control problems} with endpoints sampled from $\pi^\star_{0,T}$:
    \begin{align}
        \inf_{\boldsymbol{u}}&\mathbb{E}_{\pi^\star_{0,T}(d\boldsymbol{x}_0,d\boldsymbol{x}_T)}\int_0^T\mathbb{E}_{p_t(\cdot|\boldsymbol{x}_0, \boldsymbol{x}_T)}\left[\frac{1}{2}\|\boldsymbol{u}(\boldsymbol{X}_t,t|\boldsymbol{x}_0, \boldsymbol{x}_T)\|^2 \right]dt\tag{Conditional SOC Objective}\label{eq:condsoc-objective}\\
        &\text{s.t.}\quad d\boldsymbol{X}_t=(\boldsymbol{f}(\boldsymbol{X}_t,t)+\sigma_t\boldsymbol{u}(\boldsymbol{X}_t,t|\boldsymbol{x}_0, \boldsymbol{x}_T))dt+\sigma_td\boldsymbol{B}_t\nonumber
    \end{align}
\end{proposition}

\textit{Proof.} We recall the dynamic SB problem with the Fokker-Planck constraint as:
\begin{small}
\begin{align}
    \inf_{\boldsymbol{u}}&\int_0^T\mathbb{E}_{p_t}\left[\frac{1}{2}\|\boldsymbol{u}(\boldsymbol{X}_t,t)\|^2 \right]dt\quad \text{s.t.}\quad \begin{cases}
        \partial_tp_t=-\nabla\cdot \left((\boldsymbol{f}(\boldsymbol{x},t)+\sigma_t\boldsymbol{u}(\boldsymbol{x},t))p_t(\boldsymbol{x})\right)+\frac{\sigma^2_t}{2}\Delta p_t(\boldsymbol{x})\\
        p_0=\pi_0, \quad p_T=\pi_T
    \end{cases}
\end{align}
\end{small}

\textbf{Step 1: Decompose the Minimization Objective. }
Let $\pi^\star_{0,T}$ denote the optimal endpoint coupling solving the static SB problem with reference endpoint law $\mathbb{Q}_{0,T}$. Then, the marginal density $p_t$ can be factorized as:
\begin{small}
\begin{align}
    p^\star_t(\boldsymbol{x})=\int p_t(\boldsymbol{x}|\boldsymbol{x}_0, \boldsymbol{x}_T)\pi^\star_{0,T}(d\boldsymbol{x}_0, d\boldsymbol{x}_T)\label{eq:condsoc-proof1}
\end{align}
\end{small}
Under mild regularity assumptions allowing Leibniz's rule, Fubini's theorem, and differentiation under the integral sign\footnote{We assume standard regularity conditions ensuring that expectations and time integrals can be interchanged and that differentiation under the integral sign is valid. Such conditions are typically satisfied for diffusion processes with smooth drift and nondegenerate diffusion coefficients.}, we can separate the optimal endpoint law $\pi^\star_{0,T}$ out of the integral, condition on the endpoints, and decompose the objective as:
\begin{small}
\begin{align}
    \int_0^T\mathbb{E}_{p_t}\left[\frac{1}{2}\|\boldsymbol{u}(\boldsymbol{X}_t,t)\|^2 \right]dt&=\int_0^T\bluetext{\mathbb{E}_{\pi^\star_{0,T}}}\left[\mathbb{E}_{p_t(\cdot|\bluetext{\boldsymbol{x}_0, \boldsymbol{x}_T})}\left[\frac{1}{2}\|\boldsymbol{u}(\boldsymbol{X}_t,t|\bluetext{\boldsymbol{x}_0, \boldsymbol{x}_T})\|^2 \right]\right]dt\tag{law of total expectation}\\
    &=\mathbb{E}_{\pi^\star_{0,T}}\left[\bluetext{\int_0^T}\mathbb{E}_{p_t(\cdot|\boldsymbol{x}_0, \boldsymbol{x}_T)}\left[\frac{1}{2}\|\boldsymbol{u}(\boldsymbol{X}_t,t|\boldsymbol{x}_0, \boldsymbol{x}_T)\|^2 \right]\right]dt\tag{Fubini's theorem}
\end{align}
\end{small}
which recovers the conditional SOC objective in (\ref{eq:condsoc-objective}).

\textbf{Step 2: Derive the Conditional Fokker-Planck Constraint. }
We still need to show that the conditional dynamics $p_t(\cdot|\boldsymbol{x}_0, \boldsymbol{x}_T)$ satisfy the conditional Fokker-Planck constraint. To do this, we substitute the factorization for $p_t^\star$ in (\ref{eq:condsoc-proof1}) into each term of the Fokker-Planck constraint defined as:
\begin{small}
\begin{align}
    \underbrace{\partial_tp_t(\boldsymbol{x})}_{(\bigstar)}=\underbrace{-\nabla\cdot \left((\boldsymbol{f}(\boldsymbol{x},t)+\sigma_t\boldsymbol{u}(\boldsymbol{x},t))p_t(\boldsymbol{x})\right)}_{(\diamond)}+\underbrace{\frac{\sigma^2_t}{2}\Delta p_t(\boldsymbol{x})}_{(\blacklozenge)}\label{eq:condsoc-proof2}
\end{align}
\end{small}
Starting off, the time derivative $(\bigstar)$ can be written as an expectation over the optimal coupling as:
\begin{small}
\begin{align}
    \partial_tp_t(\boldsymbol{x})&=\partial_t\bluetext{\int p_t(\boldsymbol{x}|\boldsymbol{x}_0, \boldsymbol{x}_T)\pi^\star_{0,T}(d\boldsymbol{x}_0, d\boldsymbol{x}_T)}\nonumber\\
    &= \int\left[\bluetext{\partial_t p_t(\boldsymbol{x}|\boldsymbol{x}_0, \boldsymbol{x}_T)}\right]\pi^\star_{0,T}(d\boldsymbol{x}_0, d\boldsymbol{x}_T)=\mathbb{E}_{\pi^\star_{0,T}}\left[\partial_tp_t(\boldsymbol{x}|\boldsymbol{x}_0, \boldsymbol{x}_T)\right]
\end{align}
\end{small}
where the integral can be separated out as $\pi_{0,T}^\star$ is not dependent on time. Next, we rearrange the divergence term $(\diamond)$ as:
\begin{small}
\begin{align}
    \nabla\cdot \left((\boldsymbol{f}(\boldsymbol{x},t)+\sigma_t\boldsymbol{u}(\boldsymbol{x},t))p_t(\boldsymbol{x})\right)&=\nabla\cdot \left((\boldsymbol{f}(\boldsymbol{x},t)+\sigma_t\boldsymbol{u}(\boldsymbol{x},t))\int p_t(\boldsymbol{x}|\boldsymbol{x}_0, \boldsymbol{x}_T)\pi_{0,T}^\star(d\boldsymbol{x}_0, d\boldsymbol{x}_T)\right)\nonumber\\
    &=\nabla\cdot \left(\int(\boldsymbol{f}(\boldsymbol{x},t)+\sigma_t\boldsymbol{u}(\boldsymbol{x},t)) p_t(\boldsymbol{x}|\boldsymbol{x}_0, \boldsymbol{x}_T)\pi_{0,T}^\star(d\boldsymbol{x}_0, d\boldsymbol{x}_T)\right)\tag{linearity of integration}\\
    &=\int\bluetext{\nabla\cdot \left((\boldsymbol{f}(\boldsymbol{x},t)+\sigma_t\boldsymbol{u}(\boldsymbol{x},t)) p_t(\boldsymbol{x}|\boldsymbol{x}_0, \boldsymbol{x}_T)\right)}\pi_{0,T}^\star(d\boldsymbol{x}_0, d\boldsymbol{x}_T)\tag{Leibniz rule}\\
    &=\mathbb{E}_{\pi_{0,T}^\star}\bigg[\bluetext{\nabla\cdot \left((\boldsymbol{f}(\boldsymbol{x},t)+\sigma_t\boldsymbol{u}(\boldsymbol{x},t)) p_t(\boldsymbol{x}|\boldsymbol{x}_0, \boldsymbol{x}_T)\right)}\bigg]
\end{align}
\end{small}
Finally, we write the Laplacian term $(\blacklozenge)$ as expectation over the optimal coupling given by:
\begin{small}
\begin{align}
    \Delta p_t(\boldsymbol{x})&=\nabla\cdot \nabla \bluetext{p_t(\boldsymbol{x})}=\nabla\cdot \nabla \left(\bluetext{\int p_t(\boldsymbol{x}|\boldsymbol{x}_0, \boldsymbol{x}_T)\pi_{0,T}^\star(d\boldsymbol{x}_0, d\boldsymbol{x}_T)}\right)\nonumber\\
    &=\int\bluetext{\left(\nabla\cdot \nabla p_t(\boldsymbol{x}|\boldsymbol{x}_0, \boldsymbol{x}_T)\right)}\pi_{0,T}^\star(d\boldsymbol{x}_0, d\boldsymbol{x}_T)\tag{Leibniz rule}\\
    &=\mathbb{E}_{\pi_{0,T}^\star}\left[\bluetext{\nabla\cdot \nabla p_t(\boldsymbol{x}|\boldsymbol{x}_0, \boldsymbol{x}_T)}\right]=\mathbb{E}_{\pi_{0,T}^\star}\left[\bluetext{\Delta p_t(\boldsymbol{x}|\boldsymbol{x}_0, \boldsymbol{x}_T)}\right]
\end{align}
\end{small}

Substituting each of the conditional expressions into (\ref{eq:condsoc-proof2}), we get:
\begin{small}
\begin{align}
   & \underbrace{\mathbb{E}_{\pi^\star_{0,T}}\left[\partial_tp_t(\boldsymbol{x}|\boldsymbol{x}_0, \boldsymbol{x}_T)\right]}_{(\bigstar)}=\underbrace{-\mathbb{E}_{\pi_{0,T}^\star}\bigg[\nabla\cdot \left((\boldsymbol{f}(\boldsymbol{x},t)+\sigma_t\boldsymbol{u}(\boldsymbol{x},t)) p_t(\boldsymbol{x}|\boldsymbol{x}_0, \boldsymbol{x}_T)\right)\bigg]}_{(\diamond)}+\underbrace{\frac{\sigma_t^2}{2}\mathbb{E}_{\pi_{0,T}^\star}\left[\Delta p_t(\boldsymbol{x}|\boldsymbol{x}_0, \boldsymbol{x}_T)\right]}_{(\blacklozenge)}\nonumber\\
    \implies & \mathbb{E}_{\pi^\star_{0,T}}\left[\partial_tp_t(\boldsymbol{x}|\boldsymbol{x}_0, \boldsymbol{x}_T)+\nabla\cdot \left((\boldsymbol{f}(\boldsymbol{x},t)+\sigma_t\boldsymbol{u}(\boldsymbol{x},t)) p_t(\boldsymbol{x}|\boldsymbol{x}_0, \boldsymbol{x}_T)\right)-\frac{\sigma_t^2}{2}\Delta p_t(\boldsymbol{x}|\boldsymbol{x}_0, \boldsymbol{x}_T)\right]=0\nonumber\\
    \implies & \boxed{\partial_tp_t(\boldsymbol{x}|\boldsymbol{x}_0, \boldsymbol{x}_T)+\nabla\cdot \left((\boldsymbol{f}(\boldsymbol{x},t)+\sigma_t\boldsymbol{u}(\boldsymbol{x},t)) p_t(\boldsymbol{x}|\boldsymbol{x}_0, \boldsymbol{x}_T)\right)-\frac{\sigma_t^2}{2}\Delta p_t(\boldsymbol{x}|\boldsymbol{x}_0, \boldsymbol{x}_T)=0}\tag{Conditional FP Equation}\label{eq:conditional-fp-equation}
\end{align}
\end{small}
where the last implication follows from the fact that for the expectation to equal zero pointwise for all $\boldsymbol{x}$, then the terms inside the expectation must equal zero for $\pi_{0,T}^\star$-almost every $(\boldsymbol{x}_0, \boldsymbol{x}_T)$. The final line is exactly the (\ref{eq:controlled-fp-equation}) conditioned on the endpoints $(\boldsymbol{x}_0, \boldsymbol{x}_T)$.

Since we have shown that the conditional density $p_t(\cdot|\boldsymbol{x}_0, \boldsymbol{x}_T)$ satsifies the (\ref{eq:conditional-fp-equation}), we can conclude that it is a valid density evolution with endpoint constraints $p_0(\cdot|\boldsymbol{x}_0, \boldsymbol{x}_T)=\mathbb{E}_{\pi^\star_{0,T}}[\delta_{\boldsymbol{x}_0}]$ and $p_T(\cdot|\boldsymbol{x}_0, \boldsymbol{x}_T)=\mathbb{E}_{\pi^\star_{0,T}}[\delta_{\boldsymbol{x}_T}]$. By equivalence between the conditional Fokker-Planck equation and the controlled SDE generated by $\boldsymbol{u}(\boldsymbol{X}_t,t|\boldsymbol{x}_0, \boldsymbol{x}_T)$, we can also derive the corresponding SDE representation:
\begin{small}
\begin{align}
    \boxed{d\boldsymbol{X}_t =(\boldsymbol{f}(\boldsymbol{X}_t,t)+\sigma_t\boldsymbol{u}(\boldsymbol{X}_t,t|\boldsymbol{x}_0, \boldsymbol{x}_T))dt+\sigma_td\boldsymbol{B}_t,  \quad \boldsymbol{X}_0=\boldsymbol{x}_0, \quad\boldsymbol{X}_T=\boldsymbol{x}_T}  
\end{align}
\end{small}
which shows that once the endpoints are sampled from the optimal coupling $\pi_{0,T}^\star$, the (\ref{eq:dynamic-sb-problem}) decomposes into a mixture of conditional SOC problems. \hfill $\square$

This mixture conditional SOC problem decomposes the path space KL minimization to the optimal measure (\ref{eq:mixture-of-bridge}) from Proposition \ref{prop:mixture-of-bridges} into a static SB problem and a family of conditional bridge problems. This formulation avoids optimizing directly over full path measures and allows us to solve the dynamic SB problem through a \textbf{tractable two-stage procedure}: first, estimate the optimal static coupling $(\boldsymbol{x}_0, \boldsymbol{x}_T) \sim \pi_{0,T}^\star$, then learn the corresponding conditional stochastic control law $\boldsymbol{u}(\boldsymbol{X}_t,t|\boldsymbol{x}_0, \boldsymbol{x}_T)$. We will leverage this idea in Section \ref{subsec:score-and-flow} to efficiently learn parameterized Schrödinger bridges with score and flow matching. 

\subsection{Time Reversal}
\label{subsec:time-reversal}
A useful approach for constructing stochastic bridges is to analyze the time reversal of a stochastic process, where we leverage the \textbf{key idea} that conditioning on a terminal constraint induces a \textit{modified reverse-time dynamics} whose drift differs from the original forward drift. Rather than directly enforcing the endpoint condition, we reinterpret the forward process as a backward process and derive the forward Markovian dynamics that reproduce the same marginal density evolution under time reversal, revealing how information about the terminal distribution propagates backward through the SDE dynamics.

\begin{figure}
    \centering
    \includegraphics[width=\linewidth]{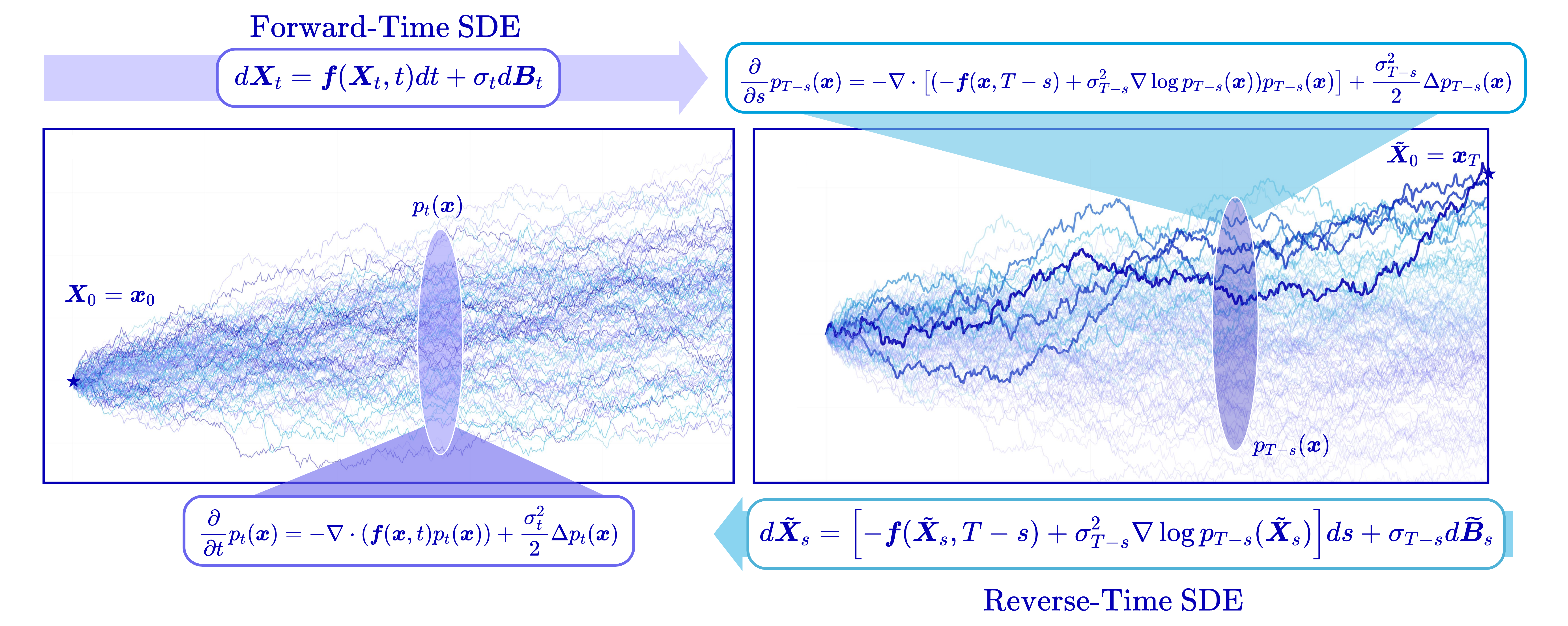}
    \caption{\textbf{Time-Reversal of Stochastic Differential Equations.} \textbf{Left:} A forward diffusion process over $t\in [0,T]$ evolves from an initial state $\boldsymbol{X}_0=\boldsymbol{x}_0$ under the forward-time SDE $d\boldsymbol{X}_t$, producing marginal densities $p_t(\boldsymbol{x})$ that evolve via the Fokker-Planck equation. \textbf{Right:} The corresponding reverse-time stochastic process generates trajectories that traverse the same sequence of marginal densities in the reversed time coordinate $s=T-t\in [0,T]$. Its drift includes an additional score correction $\nabla \log p_{T-s}(\boldsymbol{x})$ that accounts for the probability flow of the forward dynamics. This reverse SDE allows trajectories to be simulated starting from the terminal distribution such that they reconstruct the initial distribution.}
    \label{fig:time-reversal}
\end{figure}

Consider a \boldtext{forward-time stochastic process} defined on the time-horizon $t\in [0,T]$ generated from the SDE:
\begin{align}
    d\boldsymbol{X}_t=\boldsymbol{f}(\boldsymbol{X}_t, t)dt+\sigma_td\boldsymbol{B}_t, \quad \boldsymbol{X}_0=\boldsymbol{x}_0\label{eq:forward-time}
\end{align}
One method of building a stochastic bridge that evolves via (\ref{eq:forward-time}) but is \textbf{conditioned to reach a target state} $\boldsymbol{X}_T=\boldsymbol{x}_T$ is to reverse the time coordinate of the original forward process such that it can be conditioned on the target state and then, derive a \boldtext{reverse stochastic process} that matches density evolution of the forward process with time-reversal.

\begin{proposition}[Time Reversal Formula]
    Consider a forward-time SDE of the form:
    \begin{small}
    \begin{align}
        d\boldsymbol{X}_t=\boldsymbol{f}(\boldsymbol{X}_t,t) dt+ \sigma_td\boldsymbol{B}_t
    \end{align}
    \end{small}
    where $\boldsymbol{f}(\boldsymbol{x},t)$ is the drift and $\sigma_t$ is the scalar diffusion coefficient. Let $p_t(\boldsymbol{x})$ denote the marginal density of $\boldsymbol{X}_t$ at time $t$. Then, the time-reversed process $\tilde{\boldsymbol{X}}_s:=\boldsymbol{X}_{T-s}$ follows the SDE:
    \begin{small}
    \begin{align}
        d\tilde{\boldsymbol{X}}_s=\left[-\boldsymbol{f}(\tilde{\boldsymbol{X}}_s,T-s)+\sigma_{T-s}^2\nabla\log p(\tilde{\boldsymbol{X}}_s,T-s)\right]ds+\sigma_{T-s}d\widetilde{\boldsymbol{B}}_s
    \end{align}
    \end{small}
    where $\tilde{\boldsymbol{B}}_s$ is the Brownian motion with respect to the reverse-time filtration $\tilde{\mathcal{F}}=\sigma(\tilde{\boldsymbol{X}}_\tau:0\leq \tau \leq s)$. Together, the pair of forward and reverse-time SDEs are given by:  
    \begin{small}
    \begin{align}
        \begin{cases}
            d\boldsymbol{X}_t=\boldsymbol{f}(\boldsymbol{X}_t,t) dt+ \sigma_td\boldsymbol{B}_t\\
            d\tilde{\boldsymbol{X}}_s=\left[-\boldsymbol{f}(\tilde{\boldsymbol{X}}_s,T-s)+\sigma_{T-s}^2\nabla\log p(\tilde{\boldsymbol{X}}_s,T-s)\right]ds+\sigma_{T-s}d\widetilde{\boldsymbol{B}}_s
        \end{cases}\tag{Time Reversal Formula}\label{eq:time-rev-forward-back-sde}
    \end{align}
    \end{small}
\end{proposition}
\textit{Proof.} Intuitively, defining an SDE that simulates stochastic processes on the reversed time coordinate such that the marginals match those of the forward time stochastic processes requires first \textbf{reparameterizing the time coordinate of the forward process} and then \textbf{deriving a backward drift that \textit{propogates} particles on a reversed time coordinate} such that it matches the corresponding marginal on the reparameterized forward process. To this end, we break the proof into two steps.

\textbf{Step 1: Define the Forward Process with Time-Reversal.} 
This is a simple reparameterization of the time coordinate such that it follows $s=T-t$ that starts at $t=0$ and $s=T$ and terminates at $t=T$ and $s=0$. We denote this process as $\overleftarrow{\boldsymbol{X}}_s=\boldsymbol{X}_{T-t}$ This reparameterization allows us to define a initial condition $\overleftarrow{\boldsymbol{X}}_0=\boldsymbol{x}_T$ and the density $\tilde{p}_s(\boldsymbol{x})$ such that $q(\boldsymbol{x},0)=\pi_T$. With the same drift and diffusion terms, we define the time-reversed process with the SDE:
\begin{align}
    d\overleftarrow{\boldsymbol{X}}_s&=\underbrace{\boldsymbol{f}(\overleftarrow{\boldsymbol{X}}_s, T-s)}_{:=\overleftarrow{\boldsymbol{f}}(\boldsymbol{X}_s,s)}ds+\sigma_{T-s}d\boldsymbol{B}_s, \quad \overleftarrow{\boldsymbol{X}}_s=\boldsymbol{x}_T
\end{align}
The \textbf{marginal density} $p_t(\boldsymbol{x})$ evolves via the Fokker-Planck equation derived in Section \ref{subsec:fp-equation} as:
\begin{align}
    \partial_tp(\boldsymbol{x},t)&=-\nabla\cdot (\boldsymbol{f}(\boldsymbol{x},t) p(\boldsymbol{x},t))+\frac{\sigma_t^2}{2}\Delta p(\boldsymbol{x},t)
\end{align}
To write this as an evolution over the time coordinate $s=T-t$, we perform a change of variables as:
\begin{align}
    \frac{\partial}{\partial s}\underbrace{\frac{\partial s}{\partial t}}_{=-1}\bluetext{\underbrace{p(\boldsymbol{x},T-s)}_{\tilde{p}_s(\boldsymbol{x})}}&=-\nabla\cdot (\boldsymbol{f}(\boldsymbol{x},T-s) \bluetext{\underbrace{p(\boldsymbol{x},T-s)}_{\tilde{p}_s(\boldsymbol{x})}})+\frac{\sigma_{T-s}^2}{2}\Delta \bluetext{\underbrace{p(\boldsymbol{x},T-s)}_{\tilde{p}_s(\boldsymbol{x})}}\nonumber\\
    -\frac{\partial}{\partial s}\tilde{p}_s(\boldsymbol{x})&=-\nabla\cdot (\boldsymbol{f}(\boldsymbol{x},T-s) \tilde{p}_s(\boldsymbol{x}))+\frac{\sigma_{T-s}^2}{2}\Delta \tilde{p}_s(\boldsymbol{x})
\end{align}
Multiplying both sides by $-1$, we get that the evolution of time-reversed density $\tilde{p}_s(\boldsymbol{x})$ follows the PDE:
\begin{align}
    \frac{\partial}{\partial s}\tilde{p}_s(\boldsymbol{x})&=\nabla\cdot (\boldsymbol{f}(\boldsymbol{x},T-s) \tilde{p}_s(\boldsymbol{x}))-\frac{\sigma_{T-s}^2}{2}\Delta \tilde{p}_s(\boldsymbol{x})\label{eq:time-rev-eq2}
\end{align}

\textbf{Step 2: Derive the Fokker-Planck Equation of the Reverse Stochastic Process.} 
Although reversing the time coordinate defines the density evolution conditioned on a target distribution, it does \textit{not} tell us the drift and diffusion to follow when \textbf{simulating paths backward in time} since the drift and diffusion are still defined in the forward direction.  

We define $(\tilde{\boldsymbol{X}}_s)_{s\in [0,T]}$ as the Itô process that evolves backward in time. Let $\boldsymbol{b}(\tilde{\boldsymbol{X}}_s, s)$ be a placeholder for the unknown drift. Since $(\tilde{\boldsymbol{X}}_s)_{s\in [0,T]}$ must be adapted to the backward filtration $(\mathcal{F}_s)_{s\in [0,T]}$ such that it depends only on the past and present given the backward direction, we denote this new $\mathcal{F}_s$-adapted Brownian motion as $d\widetilde{\boldsymbol{B}}_s$. To define the diffusion coefficient of the reverse process, we denote $\tilde{\sigma}_s\equiv\sigma_{T-s}$. Therefore, we can write the SDE of the backward stochastic process as: 
\begin{align}
    d\tilde{\boldsymbol{X}}_s=\boldsymbol{b}(\tilde{\boldsymbol{X}}_s,s)ds+\tilde{\sigma}_sd\widetilde{\boldsymbol{B}}_s
\end{align}
and the corresponding Fokker-Planck equation as:
\begin{align}
    \frac{\partial}{\partial s}\tilde{p}_s(\boldsymbol{x})&=-\nabla\cdot(\boldsymbol{b}(\boldsymbol{x},s)\tilde{p}_s(\boldsymbol{x}))+\frac{\tilde{\sigma}_s^2}{2}\Delta \tilde{p}_s(\boldsymbol{x})\label{eq:time-rev-eq1}
\end{align}
Setting (\ref{eq:time-rev-eq1}) and (\ref{eq:time-rev-eq2}) equal to each other, we have:
\begin{small}
\begin{align}
    -\nabla\cdot(\boldsymbol{b}(\boldsymbol{x},s)\tilde{p}_s(\boldsymbol{x}))+\frac{\tilde{\sigma}_s^2}{2}\Delta \tilde{p}_s(\boldsymbol{x}) &=\nabla\cdot (\boldsymbol{f}(\boldsymbol{x},T-s) \tilde{p}_s(\boldsymbol{x}))-\frac{\tilde{\sigma}_s^2}{2}\Delta \tilde{p}_s(\boldsymbol{x})\nonumber\\
    -\nabla\cdot(\boldsymbol{b}(\boldsymbol{x},s)\tilde{p}_s(\boldsymbol{x}))-\nabla\cdot (\boldsymbol{f}(\boldsymbol{x},T-s) \tilde{p}_s(\boldsymbol{x}))&=\tilde{\sigma}_s^2\Delta \tilde{p}_s(\boldsymbol{x})\nonumber\\
    \nabla\cdot \bigg(\big[\boldsymbol{b}(\boldsymbol{x},s)+\boldsymbol{f}(\boldsymbol{x},T-s)\big] \tilde{p}_s(\boldsymbol{x})\bigg)&=\tilde{\sigma}_s^2\Delta \tilde{p}_s(\boldsymbol{x})
\end{align}
\end{small}
By definition, the Laplacian is the divergence of the gradient $\Delta\equiv\nabla\cdot\nabla$, so we can write:
\begin{align}
    \nabla\cdot \bigg(\big[\boldsymbol{b}(\boldsymbol{x},s)+\boldsymbol{f}(\boldsymbol{x},T-s)\big] \tilde{p}_s(\boldsymbol{x})\bigg)&=\nabla\cdot \left(\tilde{\sigma}_s^2\nabla \tilde{p}_s(\boldsymbol{x})\right)
\end{align}
Since the terms within the divergences are unknown and we aim to determine an expression for the backward drift $\boldsymbol{b}(\boldsymbol{x},s)$ with respect to the forward drift and diffusion, we apply a similar technique to the one used to derive the Fokker-Planck equation in Section \ref{subsec:fp-equation} by integrating both sides with an \textbf{arbitrary test function} $\phi(\boldsymbol{x})$.
\begin{align}
    \int_{\mathbb{R}^d}\nabla\cdot \bigg(\big[\boldsymbol{b}(\boldsymbol{x},s)+\boldsymbol{f}(\boldsymbol{x},T-s)\big] \tilde{p}_s(\boldsymbol{x})\bigg)\phi(\boldsymbol{x})d\boldsymbol{x}&=\int_{\mathbb{R}^d}\nabla\cdot \left(\tilde{\sigma}_s^2\nabla \tilde{p}_s(\boldsymbol{x})\right)\phi(\boldsymbol{x})d\boldsymbol{x}
\end{align}
Applying the integration by parts identity $\int_{\mathbb{R}^d} \nabla\cdot f(\boldsymbol{x})\phi(\boldsymbol{x})d\boldsymbol{x}=-\int_{\mathbb{R}^d}f(\boldsymbol{x})\cdot\nabla\phi(\boldsymbol{x})d\boldsymbol{x}$\footnote{which assumes $f$ is continuously differentiable and $\phi$ is smooth with sufficient decay at infinity such that the boundary term from integration by parts vanishes.} to both sides, we have: 
\begin{align}
    \int_{\mathbb{R}^d} \bluetext{\underbrace{\bigg(\big[\boldsymbol{b}(\boldsymbol{x},s)+\boldsymbol{f}(\boldsymbol{x},T-s)\big] \tilde{p}_s(\boldsymbol{x})\bigg)}_{(\bigstar)}}\cdot\nabla\phi(\boldsymbol{x})d\boldsymbol{x}&=\int_{\mathbb{R}^d}\pinktext{\underbrace{\left(\tilde{\sigma}_s^2\nabla \tilde{p}_s(\boldsymbol{x})\right)}_{(\diamond)}}\cdot \nabla\phi(\boldsymbol{x})d\boldsymbol{x}
\end{align}

Since we defined $\phi(\boldsymbol{x})$ to be any arbitrary test function, $(\bigstar)$ and $(\diamond)$ must be equal to each other for all $\boldsymbol{x}$, and thus we can write: 
\begin{align}
    \big[\boldsymbol{b}(\boldsymbol{x},s)+\boldsymbol{f}(\boldsymbol{x},T-s)\big] \tilde{p}_s(\boldsymbol{x})=\tilde{\sigma}_s^2\nabla \tilde{p}_s(\boldsymbol{x})
\end{align}

Rearranging and dividing both sides by $\tilde{p}_s(\boldsymbol{x})\geq 0$, we have:
\begin{align}
    \boldsymbol{b}(\boldsymbol{x},s)\tilde{p}_s(\boldsymbol{x}) &=-\boldsymbol{f}(\boldsymbol{x},T-s)\tilde{p}_s(\boldsymbol{x})+\tilde{\sigma}_s^2\nabla \tilde{p}_s(\boldsymbol{x})\nonumber\\
    \boldsymbol{b}(\boldsymbol{x},s) &=-\boldsymbol{f}(\boldsymbol{x},T-s)+\tilde{\sigma}_s^2\bluetext{\underbrace{\frac{\nabla \tilde{p}_s(\boldsymbol{x})}{\tilde{p}_s(\boldsymbol{x})}}_{\text{score function}}}
\end{align}
\noindent Since $\frac{\nabla \tilde{p}_s(\boldsymbol{x})}{\tilde{p}_s(\boldsymbol{x})}=\nabla\log \tilde{p}_s(\boldsymbol{x})$ which is the \boldtext{score function} of the density $\tilde{p}_s(\boldsymbol{x})$, we can write the final \textbf{backward drift} as:
\begin{align}
    \boldsymbol{b}(\boldsymbol{x},s) &=-\boldsymbol{f}(\boldsymbol{x},T-s)+\tilde{\sigma}_s^2\nabla\log \tilde{p}_s(\boldsymbol{x})
\end{align}
which can also be expressed in terms of the forward density $\tilde{p}_s(\boldsymbol{x})=p(\boldsymbol{x},T-s)$ and forward diffusion $\tilde{\sigma}_s=\sigma_{T-s}$ as:
\begin{align}
    \boldsymbol{b}(\boldsymbol{x},s) &=-\boldsymbol{f}(\boldsymbol{x},T-s)+\sigma_{T-s}^2\nabla\log p(\boldsymbol{x},T-s)
\end{align}

Substituting $\boldsymbol{b}(\boldsymbol{x},s)$ into (\ref{eq:time-rev-eq1}), we get the Fokker-Planck equation satisfied by the backward stochastic process (or \textit{bridge}) as:
\begin{align}
    \frac{\partial}{\partial s}p_{T-s}(\boldsymbol{x})&=-\nabla\cdot\big[\bluetext{(-\boldsymbol{f}(\boldsymbol{x},T-s)+\sigma_{T-s}^2\nabla\log p_{T-s}(\boldsymbol{x}))}p_{T-s}(\boldsymbol{x})\big]+\frac{\sigma_{T-s}^2}{2}\Delta p_{T-s}(\boldsymbol{x})
\end{align}
and the corresponding SDE given by:
\begin{small}
\begin{align}
    d\tilde{\boldsymbol{X}}_s=\left[-\boldsymbol{f}(\tilde{\boldsymbol{X}}_s,T-s)+\sigma_{T-s}^2\nabla\log p(\tilde{\boldsymbol{X}}_s,T-s)\right]ds+\sigma_{T-s}d\widetilde{\boldsymbol{B}}_s\nonumber
\end{align}
\end{small}
which completes the pair of \boldtext{forward and reverse-time SDEs}:
\begin{small}
\begin{align}
    \begin{cases}
        d\boldsymbol{X}_t=\boldsymbol{f}(\boldsymbol{X}_t, t)dt+\sigma_td\boldsymbol{B}_t\\
        d\tilde{\boldsymbol{X}}_s=\left[-\boldsymbol{f}(\tilde{\boldsymbol{X}}_s,T-s)+\sigma_{T-s}^2\nabla\log p(\tilde{\boldsymbol{X}}_s,T-s)\right]ds+\sigma_{T-s}d\widetilde{\boldsymbol{B}}_s
    \end{cases}
\end{align}
\end{small}
and we conclude our proof. \hfill $\square$

Using (\ref{eq:time-rev-forward-back-sde}), denoising diffusion can be viewed as a special case of the Schrödinger bridge problem where the reference process is a variance-explooding SDE with zero drift. 

\purple[Denoising Diffusion as a Special Case of the Schrödinger Bridge]{
Here, we will show that Schrödinger bridges are a \textbf{generalization} of score-based diffusion models introduced in \citep{song2019generative, song2020improved, song2020score, huang2021variational}. Specifically, score-based models are a special case where the forward-time stochastic process is the variance exploding SDE with zero-drift:
\begin{align}
    d\boldsymbol{X}_t=\sigma_td\boldsymbol{B}_t, \quad \boldsymbol{X}_0=\boldsymbol{x}_0\sim p_{\text{data}}
\end{align}
where the variance of the noise accumulates over time following $\beta_t:=\int _0^t\sigma_s^2ds$. Applying the time reversal described above, we can derive the SDE of the backward stochastic process $\tilde{\boldsymbol{X}}_s=\boldsymbol{X}_{T-s}$ as: 
\begin{small}
\begin{align}
    d\tilde{\boldsymbol{X}}_s&=\bigg[\underbrace{-\boldsymbol{f}(\boldsymbol{x},T-s)}_{=0 \text{ (no drift)}}+\sigma_{T-s}^2\nabla\log \tilde{p}_s(\boldsymbol{x})\bigg]ds+\sigma_{T-s}d\widetilde{\boldsymbol{B}}_s\nonumber\\
    &=\left[\sigma_{T-s}^2\nabla\log \tilde{p}_s(\boldsymbol{x})\right]ds+\sigma_{T-s}d\widetilde{\boldsymbol{B}}_s\label{eq:diffusion-reverse-deriv}
\end{align}
where the initial distribution is a Gaussian with large variance $\tilde{\boldsymbol{X}}_0\sim \mathcal{N}(\boldsymbol{x}_0, \beta_T)$. Since the variance of the forward process is $\beta_t:=\int _0^t\sigma_s^2ds$, we can derive the variance of the backward process as the total variance subtracted by the variance reduced in the backward process up to time $s$ given by $\tilde{\beta}_s=\beta_T-\beta_s$. Therefore, the density of the backward process can be written explicitly as a Gaussian with mean $\boldsymbol{x}$ and variance $\tilde{\beta}_s$, denoted $\tilde{p}_s(\boldsymbol{x})=\mathcal{N}(\boldsymbol{x}_0,\beta_T-\beta_s)$. Then, the score function can be rewritten as:
\begin{align}
    \tilde{p}_s(\boldsymbol{x})&=\frac{1}{(2\pi(\beta_T-\beta_s))^{d/2}}\exp\left(-\frac{\|\boldsymbol{x}-\boldsymbol{x}_0\|^2}{2(\beta_T-\beta_s)}\right)\nonumber\\
    \log \tilde{p}_s(\boldsymbol{x})&=-\frac{d}{2}\log(2\pi(\beta_T-\beta_s))-\frac{1}{2(\beta_T-\beta_s)}\|\boldsymbol{x}-\boldsymbol{x}_0\|^2\nonumber\\
    \nabla \log \tilde{p}_s(\boldsymbol{x})&=-\frac{1}{2(\beta_T-\beta_s)}\nabla \left(\|\boldsymbol{x}-\boldsymbol{x}_0\|^2\right)\nonumber\\
    &=-\frac{1}{2(\beta_T-\beta_s)}\cdot2(\boldsymbol{x}-\boldsymbol{x}_0)\nonumber\\
    &=-\frac{\boldsymbol{x}-\boldsymbol{x}_0}{\beta_T-\beta_s}
\end{align}
Substituting this expression for the score into the backward SDE, we get the SDE of a \textbf{Brownian bridge}:
\begin{align}
    d\tilde{\boldsymbol{X}}_s&=\bigg[\sigma_{T-s}^2\frac{\boldsymbol{x}_0-\tilde{\boldsymbol{X}}_s}{\beta_T-\beta_s}\bigg]ds+\sigma_{T-s}d\widetilde{\boldsymbol{B}}_s
\end{align}
\end{small}
\noindent In practice, we do not know clean samples $\boldsymbol{x}_0\sim p_{\text{data}}$ during the backward process, so instead of initializing the distribution from $\tilde{\boldsymbol{X}}_0\sim \mathcal{N}(\boldsymbol{x}_0, \beta_T\boldsymbol{I}_d)$, we approximate it with $\tilde{\boldsymbol{X}}_0\sim \mathcal{N}(\boldsymbol{0}, \beta_T\boldsymbol{I}_d)$, which works when $\beta_T$ is large relative to the data variance. 
}

We can also define the time-reversal formula for a \textit{controlled} SDE by simply replacing the drift $\boldsymbol{f}(\boldsymbol{X}_t,t)$ with a controlled drift $(\boldsymbol{f}(\boldsymbol{X}_t,t)+ \sigma_t\boldsymbol{u}(\boldsymbol{X}_t,t))$ and following the same derivation. 

\begin{corollary}[Time Reversal for Controlled SDEs]\label{corollary:time-reversal-control}
    Consider a forward-time controlled SDE of the form:
    \begin{small}
    \begin{align}
        d\boldsymbol{X}_t=(\boldsymbol{f}(\boldsymbol{X}_t,t)+\sigma_t\boldsymbol{u}(\boldsymbol{X}_t,t)) dt+ \sigma_td\boldsymbol{B}_t
    \end{align}
    \end{small}
    where $\boldsymbol{u}(\boldsymbol{x},t)$ is the control drift and $\sigma_t$ is the scalar diffusion coefficient. Let $p_t(\boldsymbol{x})$ denote the marginal density of $\boldsymbol{X}_t$ at time $t$. Then, the time-reversed process $\tilde{\boldsymbol{X}}_s:=\boldsymbol{X}_{T-s}$ follows the SDE:
    \begin{small}
    \begin{align}
        d\tilde{\boldsymbol{X}}_s=\left[-\boldsymbol{f}(\tilde{\boldsymbol{X}}_s,T-s)-\sigma_{T-s}\boldsymbol{u}(\tilde{\boldsymbol{X}}_s,T-s)+\sigma_{T-s}^2\nabla\log p(\tilde{\boldsymbol{X}}_s,T-s)\right]ds+\sigma_{T-s}d\widetilde{\boldsymbol{B}}_s
    \end{align}
    \end{small}
    and the pair of forward and reverse-time SDEs is given by:  
    \begin{small}
    \begin{align}
        \begin{cases}
            d\boldsymbol{X}_t=(\boldsymbol{f}(\boldsymbol{X}_t,t)+\sigma_t\boldsymbol{u}(\boldsymbol{X}_t,t)) dt+ \sigma_td\boldsymbol{B}_t\\
            d\tilde{\boldsymbol{X}}_s=\left[-\boldsymbol{f}(\tilde{\boldsymbol{X}}_s,T-s)-\sigma_{T-s}\boldsymbol{u}(\tilde{\boldsymbol{X}}_s,T-s)+\sigma_{T-s}^2\nabla\log p(\tilde{\boldsymbol{X}}_s,T-s)\right]ds+\sigma_{T-s}d\widetilde{\boldsymbol{B}}_s
        \end{cases}\label{eq:controlled-time-reversal-formula}
    \end{align}
    \end{small}
\end{corollary}

The \textbf{key takeaway} from this section is that reversing a forward-time stochastic process is not as simple as inverting the time coordinate of the trajectories. While the forward process propagates densities $p_t$ according to its drift $\boldsymbol{f}(\boldsymbol{X}_t,t)$ and diffusion coefficients, running the process backward requires deriving a \textbf{new stochastic process} whose drift correctly propagates the marginal densities in the reverse time direction such that it reconstructs the same marginal densities as the forward process $s=T-t\in [0,T]$. 

The final (\ref{eq:time-rev-forward-back-sde}) reveals that the reverse-time drift includes the negative forward drift $-\boldsymbol{f}(\boldsymbol{X}_s,T-s)$ and an \textbf{additional correction proportional to the score function} $\nabla \log p_{T-s}(\boldsymbol{x})$, which compensates for the spreading of probability density from the forward diffusion. This insight is fundamental to modern generative modeling frameworks, such as score-based diffusion models and Schrödinger bridge methods, where learning the score function enables simulation of reverse-time dynamics and thus the generation of samples from complex target distributions.

\subsection{Forward-Backward Stochastic Differential Equations}
\label{subsec:forward-backward-sde}
\begin{figure}
    \centering
    \includegraphics[width=\linewidth]{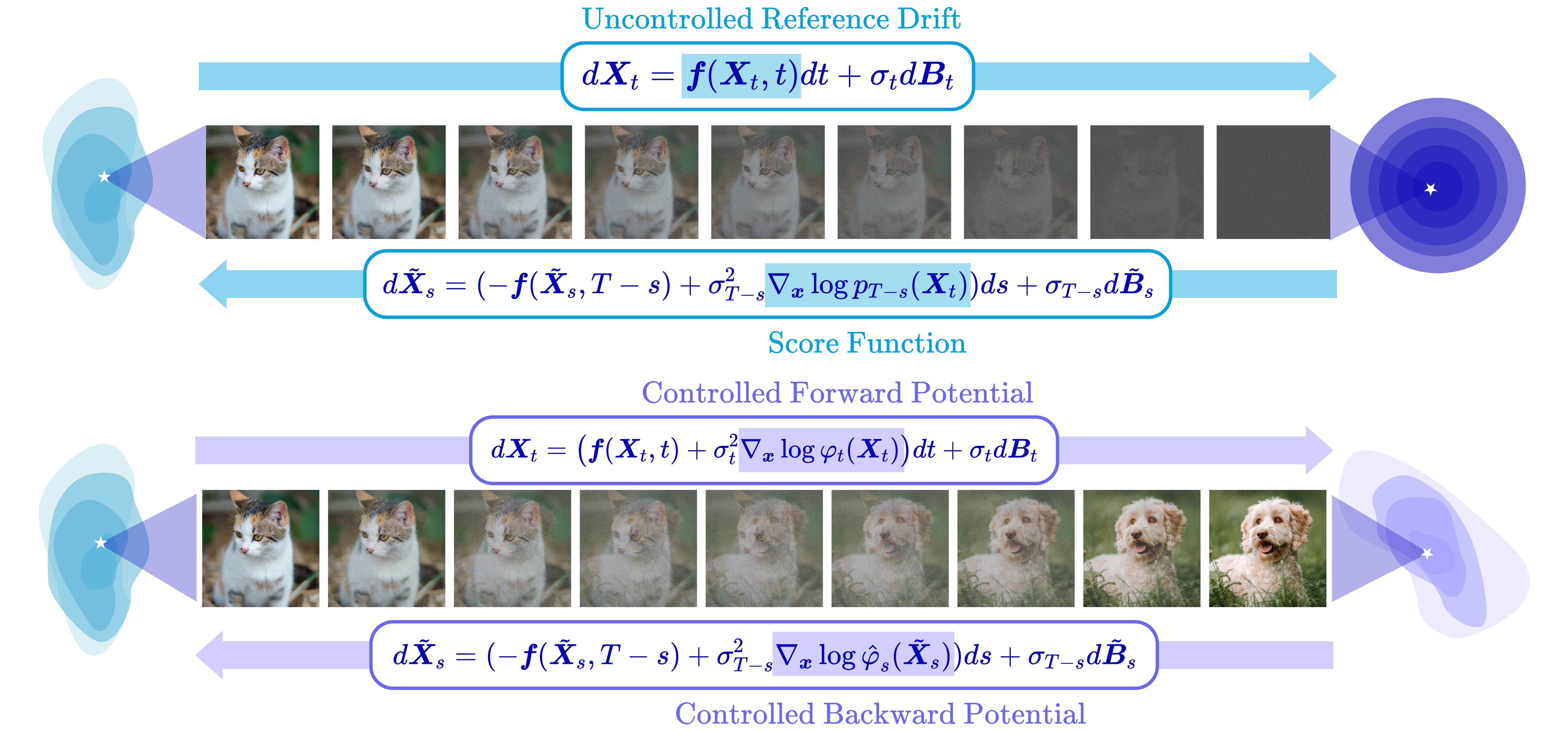}
    \caption{\textbf{Time Reversal vs. Forward-Backward SDE.} \textbf{Top}: Reversing an uncontrolled forward diffusion requires correcting the drift using the score function $\nabla \log p_t(\boldsymbol{x})$, which depends on the evolving marginal density and can lead to unstable or uncontrolled trajectories. \textbf{Bottom}: In the more general Schrödinger bridge setting, the forward process evolves via a control drift $\boldsymbol{u}(\boldsymbol{x},t):=\sigma_t\nabla_{\boldsymbol{x}}\log\varphi_t(\boldsymbol{x})$ which contains the forward Schrödinger potential $\varphi_t$. This controlled forward process induces a corresponding backward SDE with the control $\sigma_t\nabla_{\boldsymbol{x}}\log\hat\varphi_t(\boldsymbol{x})$, containing the backward Schrödinger potential $\hat\varphi_t$. This enables accurate transport between two structured distributions with optimal bridge dynamics.}
    \label{fig:bridge-vs-score}
\end{figure}

A \textbf{key limitation} of the standard time-reversal formula from Section \ref{subsec:time-reversal} is that it assumes an \textit{uncontrolled} forward SDE, where the drift $\boldsymbol{f}(\boldsymbol{x},t)$ is deterministic and can be easily simulated in either direction, such as denoising diffusion, where the forward process is a known variance-exploding SDE. In the general Schrödinger bridge setting, however, the goal is to interpolate between two \textit{structured distributions}, and the optimal forward drift that transports $\pi_0$ to $\pi_T$ is itself \textit{unknown} and must be solved for as part of the optimization problem. As a result, (\ref{eq:time-rev-forward-back-sde}) is no longer sufficient on its own, and we must instead derive a coupled pair of \textbf{forward-backward SDEs} that characterizes both the forward and backward controlled dynamics.

Recall the (\ref{eq:dynamic-sb-problem}), where the optimal control and optimal marginal density $(\boldsymbol{u}^\star, p_t^\star)$ are defined by the non-linear (\ref{eq:prop-hjb-fpe-system}) as: 
\begin{align}
    \begin{cases}
        \partial_t\psi_t+\frac{\sigma_t^2}{2}\|\nabla\psi_t\|^2 +\langle\nabla \psi_t, \boldsymbol{f}\rangle=-\sigma_t^2\Delta \psi_t\\
        \partial_t p^\star_t+\nabla\cdot (p^\star_t(\boldsymbol{f}+\sigma_t^2\nabla \psi))=\sigma_t^2\Delta p^\star_t
    \end{cases}\quad \text{s.t.}\quad \begin{cases}
        p^\star_0(\boldsymbol{x})=\pi_0\\
        p^\star_T(\boldsymbol{x})=\pi_T
    \end{cases}
\end{align}

While the Hopf-Cole transform described in Section \ref{subsec:hopf-cole-transform} transforms the non-linear PDEs into linear PDEs, solving them remains challenging. In Section \ref{subsec:primer-sgm}, we show that score-based generative modeling can be derived from the time reversal of a forward SDE described in Section \ref{subsec:time-reversal}. This naturally leads to the question: \textit{how can we turn the PDEs defining the solution to the dynamic SB problem into SDEs that can be solved with likelihood training?}

To answer this, we can leverage the theory of \boldtext{forward-backward SDEs} \citep{exarchos2018stochastic}. This theory extends the Feynman-Kac theory from Section \ref{subsec:fp-equation}, which represents solutions of certain linear parabolic PDEs as expectations over trajectories of a forward stochastic process. Forward–backward SDE theory generalizes this idea by introducing a \textbf{coupled system consisting of a forward SDE that generates trajectories and a backward stochastic process that evolves along those trajectories} and encodes the solution of the PDE.

\purple[Forward-Backward SDE Theory]{
Consider a function $\psi_t(\boldsymbol{x}):\mathbb{R}^d\times [0,T] \to \mathbb{R}\in C^{2,1}(\mathbb{R}^d\times [0,T])$ and a \textbf{parabolic PDE} of the form: 
\begin{align}
    &\partial_t \psi_t+\frac{\sigma_t^2}{2}\Delta \psi_t+\langle \boldsymbol{f}, \nabla \psi_t\rangle+\boldsymbol{h}(\boldsymbol{x},t, \psi_t, \sigma_t\nabla \psi_t)=0, \quad \psi_T(\boldsymbol{x})=\Phi(\boldsymbol{x})\label{eq:fbsde-theory-1}
\end{align}
where $\psi_T(\boldsymbol{x})=\Phi(\boldsymbol{x})$ is the terminal constraint. Suppose the state $\boldsymbol{X}_t=\boldsymbol{x}$ is a stochastic process $\boldsymbol{X}_{0:T}$ that evolves via a forward SDE defined as:
\begin{align}
    d\boldsymbol{X}_t=\boldsymbol{f}(\boldsymbol{X}_t,t)dt+\sigma_td\boldsymbol{B}_t, \quad \boldsymbol{X}_0=\boldsymbol{x}_0\label{eq:fbsde-theory-2}
\end{align}
Then, we can define the stochastic process $\boldsymbol{Y}_t:=\psi_t(\boldsymbol{X}_t)$ and $\boldsymbol{Z}_t:=\sigma_t \nabla \psi_t(\boldsymbol{X}_t)$ which evolves via the \textbf{backward SDE} defined as: 
\begin{align}
    d\boldsymbol{Y}_t=-\boldsymbol{h}(\boldsymbol{X}_t, t, \boldsymbol{Y}_t, \boldsymbol{Z}_t)dt+\boldsymbol{Z}_t^\top d\boldsymbol{B}_t, \quad \boldsymbol{Y}_T=\Phi(\boldsymbol{X}_T)\label{eq:fbsde-theory-3}
\end{align}
such that the solution $\psi_t$ to (\ref{eq:fbsde-theory-1}) is equivalent to the solution to the forward-backward SDEs defined in (\ref{eq:fbsde-theory-2}) and (\ref{eq:fbsde-theory-3}). 
}

While this gives us a way to convert a single non-linear PDE into a pair of FBSDEs, the \textbf{dynamic SB problem with arbitrary reference dynamics} is defined by a \textbf{pair of coupled PDEs} given in Section \ref{subsec:hopf-cole-transform} by the Hopf-Cole transform. Therefore, we must derive a \textbf{system of forward-backward SDEs that describes the evolution of both potentials}.

\purple[Forward-Backward SDEs for Dynamic Schrödinger Bridge]{\label{box:fbsde-non-linear}
In Section \ref{subsec:hopf-cole-transform}, we transformed the non-linear (\ref{eq:prop-hjb-fpe-system}) with the \textbf{Hopf-Cole transform} to get the system of \textbf{linear PDEs} that defines the solution to the Schrödinger bridge problem with arbitrary reference dynamics:
\begin{align}
    \begin{cases}
        \partial_t\varphi_t+\langle \nabla\varphi_t, \boldsymbol{f}\rangle=-\frac{\sigma_t^2}{2}\Delta \varphi_t\\
        \partial_t\hat{\varphi}_t+\nabla\cdot (\hat{\varphi}_t \boldsymbol{f})=\frac{\sigma_t^2}{2}\Delta \hat{\varphi}_t
    \end{cases}\quad\text{s.t.}\quad
    \begin{cases}
        p^\star_0=\varphi_0\hat{\varphi}_0\\
        p^\star_T=\varphi_T\hat{\varphi}_T
    \end{cases}\label{eq:nonlinear-fbsde-thm-1}
\end{align}
whre $(\varphi_t, \hat\varphi_t)$ are the pair of \textbf{Schrödinger potentials} that \textit{uniquely} characterize the optimal control and optimal state PDF $(\boldsymbol{u}^\star, p_t^\star)$ given by:
\begin{align}
    \boldsymbol{u}^\star(\boldsymbol{x},t)&=\sigma_t\nabla\log\varphi_t(\boldsymbol{x})\\
    p^\star_t(\boldsymbol{x})&=\varphi_t(\boldsymbol{x})\hat{\varphi}_t(\boldsymbol{x})
\end{align}
Given the \textbf{stochastic process} $\boldsymbol{X}_{0:T}$, we can define the random variables associated with the forward and backward potentials as:
\begin{align}
    &\boldsymbol{Y}_t\equiv \boldsymbol{Y}_t(\boldsymbol{X}_t,t)=\log \varphi_t(\boldsymbol{X}_t), \quad &\boldsymbol{Z}_t\equiv\boldsymbol{Z}_t(\boldsymbol{X}_t,t)=\sigma_t\nabla\log \varphi_t(\boldsymbol{X}_t)\\
    &\widehat{\boldsymbol{Y}}_t\equiv \widehat{\boldsymbol{Y}}_t(\boldsymbol{X}_t,t)=\log \hat{\varphi}_t(\boldsymbol{X}_t), \quad &\widehat{\boldsymbol{Z}}_t\equiv\widehat{\boldsymbol{Z}}_t(\boldsymbol{X}_t,t)=\sigma_t\nabla\log \hat{\varphi}_t(\boldsymbol{X}_t)
\end{align}
Then, the forward-backward SDEs that define the evolution of these random variables are given by:
\begin{small}
\begin{align}
\begin{cases}
    d\boldsymbol{X}_t=\left(\boldsymbol{f}(\boldsymbol{X}_t,t)+ \sigma_t \boldsymbol{Z}_t\right)dt+\sigma_t d\boldsymbol{B}_t\\
    d\boldsymbol{Y}_t=\frac{1}{2}\|\boldsymbol{Z}_t\|^2dt+\boldsymbol{Z}_t^\top d \boldsymbol{B}_t\\
    d\widehat{\boldsymbol{Y}}_t=\big[\nabla\cdot (\sigma_t\widehat{\boldsymbol{Z}}_t-\boldsymbol{f})+\frac{1}{2}\|\widehat{\boldsymbol{Z}}_t\|^2+ \boldsymbol{Z}_t^\top \widehat{\boldsymbol{Z}}_t\big]dt+\widehat{\boldsymbol{Z}}_t^\top d \boldsymbol{B}_t
\end{cases}\text{s.t.}\begin{cases}
    \boldsymbol{X}_0=\boldsymbol{x}_0\\
    \boldsymbol{Y}_T+\widehat{\boldsymbol{Y}}_T=\log \pi_T(\boldsymbol{X}_T)\\
    \boldsymbol{Y}_t+\widehat{\boldsymbol{Y}}_t=\log p^\star_t(\boldsymbol{X}_t)\\
    \boldsymbol{u}^\star(\boldsymbol{X}_t,t)=\boldsymbol{Z}_t
\end{cases}\label{eq:nonlinear-fbsde-thm-2}
\end{align}
\end{small}

where the solution to the FBSDEs in (\ref{eq:nonlinear-fbsde-thm-2}) is the solution to the linear PDEs in (\ref{eq:nonlinear-fbsde-thm-1}).
}

\textit{Derivation. } 
\textbf{Step 1: Derive the Stochastic Processes that Encode Optimality Conditions.} 
Recall from Theorem \ref{thm:hopf-cole-nonlinear} that the dynamic SB problem has the control drift $\boldsymbol{u}^\star(\boldsymbol{x},t)=\sigma_t\nabla \log \varphi_t(\boldsymbol{x})$ which defines the forward SDE:
\begin{align}
    d\boldsymbol{X}_t=\left[\boldsymbol{f}(\boldsymbol{X}_t,t)+ \sigma_t^2\nabla \log \varphi_t(\boldsymbol{X}_t)\right]dt+\sigma_t d\boldsymbol{B}_t
\end{align}
Furthermore, the optimal marginal density $p^\star_t(\boldsymbol{x})$ is defined by both the forward and backward potentials $p^\star_t(\boldsymbol{x})=\varphi_t(\boldsymbol{x}) \hat{\varphi}_t(\boldsymbol{x})$, which can be transformed to a logarithm as:
\begin{align}
    \log p^\star_t(\boldsymbol{x})=\log \varphi_t(\boldsymbol{x})+\log \hat{\varphi}_t(\boldsymbol{x})
\end{align}
Therefore, both the optimal control drift and marginal density can be expressed as the logarithm of the SB potentials $(\varphi_t, \hat\varphi_t)$. The evolution of $\log \varphi_t$ and $\log\hat\varphi_t$ along the stochastic trajectories $\boldsymbol{X}_{0:T}$ can be described by a pair of stochastic processes $(\boldsymbol{Y}_t)_{t\in [0,T]}$ and $(\widehat{\boldsymbol{Y}}_t)_{t\in [0,T]}$ defined as:
\begin{align}
\begin{cases}
    \boldsymbol{Y}_t:=\log \varphi_t(\boldsymbol{X}_t)\\
    \widehat{\boldsymbol{Y}}_t:=\log \hat{\varphi}_t(\boldsymbol{X}_t)
\end{cases}\quad \text{s.t.}\quad \log p^\star_t(\boldsymbol{X}_t)=\boldsymbol{Y}_t+\widehat{\boldsymbol{Y}}_t
\end{align}
Unlike score-based generative modeling, the score function $\nabla\log q_t(\boldsymbol{x})$ is replaced with the gradient of the log potentials $\nabla\log \varphi_t(\boldsymbol{x})$ and $\nabla\log \hat\varphi_t(\boldsymbol{x})$. Since the gradient of a random variable defined by a stochastic process is also a random variable, we define the auxillary stochastic processes $(\boldsymbol{Z}_t)_{t\in [0,T]}$ and $(\widehat{\boldsymbol{Z}}_t)_{t\in [0,T]}$ which track the \textit{gradient} of the log potentials:
\begin{align}
    \begin{cases}
        \boldsymbol{Z}_t:=\sigma_t\nabla \log \varphi_t(\boldsymbol{X}_t)\\
        \widehat{\boldsymbol{Z}}_t:=\sigma_t\nabla \log \hat{\varphi}_t(\boldsymbol{X}_t)
    \end{cases}\quad \text{s.t.}\quad \boldsymbol{u}^\star(\boldsymbol{X}_t,t)=\boldsymbol{Z}_t
\end{align}
which transforms the forward SDE into:
\begin{align}
    d\boldsymbol{X}_t=\left[\boldsymbol{f}(\boldsymbol{X}_t,t)+ \sigma_t \boldsymbol{Z}_t\right]dt+\sigma_t d\boldsymbol{B}_t
\end{align}
Now that we have defined the time-indexed random variables $(\boldsymbol{Y}_t, \widehat{\boldsymbol{Y}}_t)$ and $(\boldsymbol{Z}_t, \widehat{\boldsymbol{Z}}_t)$ that encode the optimality constraints, we can derive the corresponding forward-backward SDEs.

\textbf{Step 2: Derive the Forward-Backward SDEs.} 
To determine the time-evolution of $\boldsymbol{Y}_t$ and $\widehat{\boldsymbol{Y}}_t$, we apply (\ref{eq:ito-2}). For simplicity, we will drop explicit $(\boldsymbol{X}_t,t)$ dependence and denote $\varphi_t\equiv\varphi_t(\boldsymbol{X}_t)$ and $\boldsymbol{f}\equiv\boldsymbol{f}(\boldsymbol{X}_t,t)$ throughout the derivation. Starting with $\boldsymbol{Y}_t=\log \varphi_t(\boldsymbol{X}_t)$, we apply Itô's formula to get the SDE:
\begin{small}
\begin{align}
    d\log \varphi_t&=\bigg[\partial_t \log \varphi_t+\bluetext{\langle \boldsymbol{f}+\sigma^2_t\nabla \log \varphi_t,  \nabla \log \varphi_t\rangle}+\frac{\sigma^2_t}{2}\Delta \log\varphi_t\bigg]dt+\nabla \log\varphi_t^\top \sigma_t d \boldsymbol{B}_t\nonumber\\
    &=\bigg[\underbrace{\partial_t \log \varphi_t}_{\text{given by  PDE}}+\bluetext{\langle \boldsymbol{f},\nabla \log \varphi_t\rangle+\sigma^2_t\|\nabla \log \varphi_t\|^2 }+\frac{\sigma^2_t}{2}\Delta \log\varphi_t\bigg]dt+\nabla \log\varphi_t^\top \sigma_t d \boldsymbol{B}_t\label{eq:fwd-potential-sde}
\end{align}
\end{small}
From the (\ref{eq:hopf-cole-system}), we have that the potential function $\varphi_t$ evolves via the linear PDE given by:
\begin{align}
    \partial_t \varphi_t=-\langle \nabla \varphi_t, \boldsymbol{f}\rangle-\frac{\sigma^2_t}{2}\Delta\varphi_t
\end{align}
Using the chain rule, we have $\partial_t\log \varphi_t=\frac{\partial \log \varphi_t}{\partial \varphi_t}\partial_t \varphi_t$, so the PDE gives us the following form of the time evolution for $\log\varphi_t$:
\begin{small}
\begin{align}
    \partial_t\log \varphi_t&=\frac{\partial \log \varphi_t}{\partial \varphi_t}\partial_t \varphi_t=\frac{1}{\varphi_t}\left(\bluetext{-\langle \nabla \varphi_t, \boldsymbol{f}\rangle-\frac{\sigma^2_t}{2}\Delta\varphi_t}\right)\nonumber\\
    &=-\bigg\langle\underbrace{\frac{\nabla \varphi_t}{\varphi}}_{=\nabla \log \varphi_t}, \boldsymbol{f}\bigg\rangle-\frac{\sigma^2_t}{2}\frac{1}{\varphi_t}\Delta \varphi_t=\bluetext{-\left\langle\nabla\log \varphi_t, \boldsymbol{f}\right\rangle -\frac{\sigma^2_t}{2}\Delta \log \varphi_t}
\end{align}
\end{small}
Substituting the expression for $\partial_t \log \varphi_t$ into the SDE (\ref{eq:fwd-potential-sde}), we get:
\begin{small}
\begin{align}
    d\log \varphi_t&=\bigg[\bluetext{\underbrace{\partial_t\log \varphi_t}_{\text{given by  PDE}}}+\langle \boldsymbol{f},\nabla \log \varphi_t\rangle+\sigma^2_t\|\nabla \log \varphi_t\|^2 +\frac{\sigma^2_t}{2}\Delta \log\varphi_t\bigg]dt+(\sigma_t\nabla \log\varphi_t)^\top d \boldsymbol{B}_t\nonumber\\
    &=\bigg[\bluetext{-\left\langle\nabla\log \varphi_t, \boldsymbol{f}\right\rangle -\frac{\sigma^2_t}{2}\Delta \log \varphi_t}+\pinktext{\langle \boldsymbol{f},\nabla \log \varphi_t\rangle}+\sigma^2_t\|\nabla \log \varphi_t\|^2 +\pinktext{\frac{\sigma^2_t}{2}\Delta \log\varphi_t}\bigg]dt+(\sigma_t\nabla \log\varphi_t)^\top  d \boldsymbol{B}_t\nonumber\\
    &=\|\underbrace{\sigma^2_t\nabla \log \varphi_t}_{=:\boldsymbol{Z}_t}\|^2 dt+\underbrace{(\sigma_t\nabla \log\varphi_t)^\top}_{=:\boldsymbol{Z}_t^\top} d \boldsymbol{B}_t\nonumber\\
    &=\frac{1}{2}\|\boldsymbol{Z}_t\|^2dt+\boldsymbol{Z}_t^\top d \boldsymbol{B}_t
\end{align}
\end{small}
which is exactly the SDE for $\boldsymbol{Y}_t$:
\begin{align}
    \boxed{d\boldsymbol{Y}_t=\frac{1}{2}\|\boldsymbol{Z}_t\|^2dt+\boldsymbol{Z}_t^\top d \boldsymbol{B}_t}
\end{align}
Now, we apply Itô's formula to $\widehat{\boldsymbol{Y}}_t=\log \hat{\varphi}_t(\boldsymbol{X}_t)$ to get:
\begin{small}
\begin{align}
    &d\log \hat{\varphi}_t(\boldsymbol{X}_t,t)=\bigg[\partial_t \log \hat{\varphi}_t+(\boldsymbol{f}+\sigma_t^2\nabla \log \varphi_t)^\top \nabla \log \hat{\varphi}_t+\frac{\sigma_t^2}{2}\Delta \log\hat{\varphi}_t\bigg]dt+(\sigma_t\nabla \log\hat{\varphi}_t)^\top d \boldsymbol{B}_t\nonumber\\
    &=\bigg[\underbrace{\partial_t \log \hat{\varphi}_t}_{\text{given by  PDE}}+\langle \boldsymbol{f}, \nabla \log \hat{\varphi}_t\rangle + \langle \sigma_t^2 \nabla \log \varphi_t, \nabla \log \hat\varphi_t\rangle +\frac{\sigma_t^2}{2}\Delta \log\hat{\varphi}_t\bigg]dt+(\sigma_t\nabla \log\hat{\varphi}_t)^\top d \boldsymbol{B}_t\label{eq:bwd-potential-sde}
\end{align}
\end{small}

Similarly to $\varphi_t$, the (\ref{eq:hopf-cole-system}) defines the time evolution of $\hat{\varphi}$ via the following linear PDE:
\begin{align}
    \partial_t \hat{\varphi}_t=-\nabla \cdot (\hat{\varphi}_t\boldsymbol{f})+\frac{\sigma_t^2}{2}\Delta \hat{\varphi}_t
\end{align}
Applying the chain rule, we have:
\begin{align}
    \partial_t \log \hat{\varphi}_t&=\frac{\partial \log \hat{\varphi}_t}{\partial \hat{\varphi}_t}\partial_t \hat{\varphi}_t=\frac{1}{\hat{\varphi}_t}\left(-\nabla \cdot (\hat{\varphi}_t\boldsymbol{f})+\frac{\sigma_t^2}{2}\Delta \hat{\varphi}_t\right)\nonumber\\
    &=-\underbrace{\frac{1}{\hat{\varphi}_t}(\nabla \hat{\varphi}_t}_{\nabla \log\hat{\varphi}_t}\cdot \boldsymbol{f})-\frac{1}{\hat{\varphi}_t}\hat{\varphi}_t\nabla\cdot \boldsymbol{f}+\frac{\sigma^2}{2}\frac{\Delta \hat{\varphi}_t}{\hat\varphi_t}=-\langle \nabla \log\hat{\varphi}_t, \boldsymbol{f}\rangle-\nabla\cdot \boldsymbol{f}+\frac{\sigma_t^2}{2}\frac{\Delta \hat{\varphi}_t}{\hat\varphi_t}
\end{align}
Substituting this expression into the SDE (\ref{eq:bwd-potential-sde}), we can cancel like terms to get:
\begin{small}
\begin{align}
    d\log \hat{\varphi}_t(\boldsymbol{X}_t)&=\bigg[\bluetext{-\langle \nabla\log\hat{\varphi}_t, \boldsymbol{f}\rangle-\nabla\cdot \boldsymbol{f}+\frac{\sigma_t^2}{2}\frac{\Delta \hat{\varphi}_t}{\hat\varphi_t}}+\langle \boldsymbol{f}, \nabla \log \hat{\varphi}_t\rangle \nonumber\\
    &\quad \quad + \langle \sigma_t^2 \nabla \log \varphi_t, \nabla \log \hat\varphi_t\rangle +\frac{\sigma_t^2}{2}\Delta \log\hat{\varphi}_t\bigg]dt+(\sigma_t\nabla \log\hat{\varphi}_t)^\top d \boldsymbol{B}_t\nonumber\\
    &=\bigg[-\nabla\cdot \boldsymbol{f}+\frac{\sigma_t^2}{2}\frac{\Delta \hat{\varphi}_t}{\hat\varphi_t}+ \langle \sigma_t^2 \nabla \log \varphi_t, \nabla \log \hat\varphi_t\rangle +\frac{\sigma_t^2}{2}\pinktext{\Delta \log \hat\varphi_t}\bigg]dt+(\sigma_t\nabla\log\hat{\varphi}_t)^\top d \boldsymbol{B}_t\label{eq:bwd-sde-intermediate}
\end{align}
\end{small}
We can expand the Laplacian using the product rule:
\begin{small}
\begin{align}
    \Delta \log \hat\varphi_t&= \nabla \cdot (\hat\varphi_t^{-1}\nabla \hat\varphi_t)=\bluetext{\nabla (\hat\varphi_t^{-1})}^\top\nabla \hat\varphi_t+\hat\varphi_t^{-1}\Delta\hat\varphi_t\nonumber\\
    &= \bluetext{(-\hat\varphi_t^{-2}\nabla \hat\varphi_t)}^\top\nabla \hat\varphi_t+ \hat\varphi_t^{-1}\Delta \hat\varphi_t=-\frac{\|\nabla \hat\varphi_t\|^2}{\varphi^2_t}+\frac{\Delta \hat\varphi_t}{\hat\varphi_t}\nonumber\\
    &= -\|\nabla \log \hat\varphi_t\|^2+\frac{\Delta\hat\varphi_t}{\hat\varphi_t}\\
    \implies &\frac{\Delta \hat\varphi_t}{\hat\varphi_t}=\Delta \log\hat\varphi_t+\|\nabla \log \hat\varphi_t\|^2
\end{align}
\end{small}
Substituting the expression for $\frac{\Delta \hat{\varphi}_t}{\hat\varphi_t}$ back into (\ref{eq:bwd-sde-intermediate}), we get:
\begin{small}
\begin{align}
    d\log \hat{\varphi}_t(\boldsymbol{X}_t)&=\bigg[-\nabla\cdot \boldsymbol{f}+\frac{\sigma_t^2}{2}\left(\bluetext{\Delta \log\hat\varphi_t+\|\nabla \log \hat\varphi_t\|^2}\right)+ \langle \sigma_t^2 \nabla \log \varphi_t, \nabla \log \hat\varphi_t\rangle +\pinktext{\frac{\sigma_t^2}{2}\Delta \log \hat\varphi_t}\bigg]dt+(\sigma_t\nabla\log\hat{\varphi}_t)^\top d \boldsymbol{B}_t\nonumber\\
    &=\bigg[\underbrace{-\nabla\cdot \boldsymbol{f}+\bluetext{\sigma_t^2\Delta \log\hat\varphi_t}}_{\nabla\cdot(\sigma_t^2\nabla \log \hat\varphi_t-\boldsymbol{f})}+\bluetext{\frac{\sigma_t^2}{2}\|\nabla \log \hat\varphi_t\|^2}+ \langle \sigma_t^2 \nabla \log \varphi_t, \nabla \log \hat\varphi_t\rangle\bigg]dt+(\sigma_t\nabla\log\hat{\varphi}_t)^\top d \boldsymbol{B}_t\nonumber\\
    &=\bigg[\nabla\cdot(\pinktext{\underbrace{\sigma_t^2\nabla \log \hat\varphi_t}_{=:\sigma_t\widehat{\boldsymbol{Z}}_t}}-\boldsymbol{f})+\frac{1}{2}\|\pinktext{\underbrace{\sigma_t\nabla \log \hat\varphi_t}_{=:\widehat{\boldsymbol{Z}}_t}}\|^2+ \langle\underbrace{\sigma_t \nabla \log \varphi_t}_{=:\boldsymbol{Z}_t}, \pinktext{\underbrace{\sigma_t\nabla \log \hat\varphi_t}_{=:\widehat{\boldsymbol{Z}}_t}}\rangle\bigg]dt+\pinktext{\underbrace{(\sigma_t\nabla\log\hat{\varphi}_t)^\top}_{=:\widehat{\boldsymbol{Z}}_t^\top}} d \boldsymbol{B}_t\nonumber\\
    &=\bigg[\nabla\cdot(\pinktext{\sigma_t\widehat{\boldsymbol{Z}}_t}-\boldsymbol{f})+\frac{1}{2}\|\pinktext{\widehat{\boldsymbol{Z}}_t}\|^2+  \boldsymbol{Z}_t^\top \pinktext{\widehat{\boldsymbol{Z}}_t}\bigg]dt+\pinktext{\widehat{\boldsymbol{Z}}_t^\top} d\boldsymbol{B}_t
\end{align}
\end{small}
which gives us the SDE for $\widehat{\boldsymbol{Y}}_t$:
\begin{align}
    \boxed{d\widehat{\boldsymbol{Y}}_t=\bigg[\nabla\cdot (\sigma_t\widehat{\boldsymbol{Z}}_t-\boldsymbol{f})+\frac{1}{2}\|\widehat{\boldsymbol{Z}}_t\|^2+ \boldsymbol{Z}_t^\top \widehat{\boldsymbol{Z}}_t\bigg]dt+\widehat{\boldsymbol{Z}}_t^\top d \boldsymbol{B}_t}
\end{align}
Finally, by combining the three SDEs, we have shown that the solution to the dynamic SB problem is equivalent to the solution to the system of three FBSDEs.
\begin{align}
    \boxed{\begin{cases}
        d\boldsymbol{X}_t=\left(\boldsymbol{f}(\boldsymbol{X}_t,t)+ \sigma_t \boldsymbol{Z}_t\right)dt+\sigma_t d\boldsymbol{B}_t\\
        d\tilde{\boldsymbol{X}}_s=(-\boldsymbol{f}(\tilde{\boldsymbol{X}}_s,T-s)+\sigma_{T-s}\widehat{\boldsymbol{Z}}_s)ds+\sigma_{T-s}d\tilde{\boldsymbol{B}}_s\\
        d\boldsymbol{Y}_t=\frac{1}{2}\|\boldsymbol{Z}_t\|^2dt+\boldsymbol{Z}_t^\top d \boldsymbol{B}_t\\
        d\widehat{\boldsymbol{Y}}_t=\big[\nabla\cdot (\sigma_t\widehat{\boldsymbol{Z}}_t-\boldsymbol{f})+\frac{1}{2}\|\widehat{\boldsymbol{Z}}_t\|^2+ \boldsymbol{Z}_t^\top \widehat{\boldsymbol{Z}}_t\big]dt+\widehat{\boldsymbol{Z}}_t^\top d \boldsymbol{B}_t
    \end{cases}}\tag{Forward-Backward SDEs}
\end{align}
which describe the evolution of the Schrödinger bridge potentials $(\varphi_t,\hat\varphi_t)$ that characterize the optimal control drift $\boldsymbol{u}^\star(\boldsymbol{X}_t,t)=\boldsymbol{Z}_t$ and marginal density $\log p^\star_t(\boldsymbol{X}_t)=\boldsymbol{Y}_t+\widehat{\boldsymbol{Y}}_t$ as SDEs that can be simulated forward and backward in time.\hfill $\square$

We now connect the FBSDE representation to the time-reversal representation and show that the Schrödinger bridge potentials yield a \textit{generalization} of the time-reversal strategy to settings where the forward SDE has an additional \textbf{non-deterministic control drift} $\sigma_t\boldsymbol{u}(\boldsymbol{x},t)$.

\begin{remark}[Forward-Backward SDE Theory Generalizes the Time-Reversal Formula]\label{remark:fbsde-generalize}
    The forward–backward SDE formulation provides a natural generalization of the classical time-reversal formula for uncontrolled forward SDEs. In the special case where the forward dynamics contain no control drift (i.e., $\boldsymbol{Z}_t\equiv \boldsymbol{0}$), the forward SDE reduces to the reference diffusion, and the backward control $\widehat{\boldsymbol{Z}}_t$ becomes proportional to the \textbf{score} of the marginal density $p^\star_t$:
    \begin{small}
    \begin{align}
        &\boldsymbol{Y}_t=\log \varphi_t(\boldsymbol{X}_t) \equiv 0\nonumber\\
        \implies &\log p^\star_t (\boldsymbol{X}_t)=0+\widehat{\boldsymbol{Y}}_t=\log \hat\varphi_t(\boldsymbol{X}_t)\nonumber\\
        \implies &\widehat{\boldsymbol{Z}}_t=\sigma_t\nabla \log \hat\varphi_t(\boldsymbol{X}_t)=\sigma_t\nabla \log p_t^\star(\boldsymbol{X}_t)
    \end{align}
    \end{small}
    which yields a reverse-time SDE that is exactly the time-reversal SDE derived in (\ref{eq:time-rev-forward-back-sde}):
    \begin{small}
    \begin{align}
        d\tilde{\boldsymbol{X}}_s=\left[-\boldsymbol{f}(\tilde{\boldsymbol{X}}_s,T-s)+\sigma_{T-s}^2\nabla\log p(\tilde{\boldsymbol{X}}_s,T-s)\right]ds+\sigma_{T-s}d\widetilde{\boldsymbol{B}}_s
    \end{align}
    \end{small}
    that characterize the generation process of \textbf{score-based diffusion models}. Therefore, we can interpret forward-backward SDEs as a generalization of the time-reversal formula and the dynamic SB problem as a generalization of score-based diffusion.
\end{remark}

In Section \ref{subsec:likelihood-training}, we will turn this FBSDE construction into a tractable likelihood-based training objective. However, before introducing any training formulations, we continue our exploration of building Schrödinger bridges. Next, we discuss endpoint conditioning via Doob's $h$-transform, which provides additional insight into the structure of Schrödinger bridges.

\subsection{Doob's $h$-Transform}
\label{subsec:doob-transform}

The \textbf{central goal} when solving the Schrödinger bridge problem is to derive a controlled stochastic process by minimally shaping a reference stochastic process that originates an initial distribution $\pi_0$ by \textit{conditioning} it on a terminal distribution $\pi_T$. One way to achieve this conditioned process is using \boldtext{Doob's $h$-Transform} \citep{rogers2000diffusions, sarkka2019applied}, which introduces a \textit{tilting function}, known as the $h$-function, such that multiplying the transition density of the reference stochastic process by the $h$-function and re-normalizing produces the transition density of the optimal Schrödinger bridge.

\begin{proposition}[Doob's $h$-Transform]\label{prop:doob-h}
    Under a reference stochastic process $\mathbb{Q}$, let $\mathbb{Q}(\boldsymbol{X}_{\tau}=\boldsymbol{y}|\boldsymbol{X}_t=\boldsymbol{x})$ denote the transition kernel from $\boldsymbol{x}$ at $t$ to $\boldsymbol{y}$ at time $\tau \geq t$. Define a function $h(\boldsymbol{x},t):\mathbb{R}^d\times [0,T] \to \mathbb{R}$ that satisfies the following space-time Markov consistency property:
    \begin{align}
        h(\boldsymbol{x},t)=\mathbb{E}_{\mathbb{Q}}[h(\boldsymbol{X}_\tau,\tau)|\boldsymbol{X}_t=\boldsymbol{x}]=\int_{\mathbb{R}^d}\mathbb{Q}(\boldsymbol{X}_{\tau}=\boldsymbol{y}|\boldsymbol{X}_t=\boldsymbol{x})h(\boldsymbol{y},\tau )d\boldsymbol{y}\tag{$h$-Function}\label{eq:doob-h-function}
    \end{align}
    Then, we define the stochastic process $\mathbb{P}^h$ by \textbf{tilting}  the reference process $\mathbb{Q}$ as:
    \begin{align}
        \mathbb{P}^h(\boldsymbol{X}_{\tau }=\boldsymbol{y}|\boldsymbol{X}_t=\boldsymbol{x})=\mathbb{Q}(\boldsymbol{X}_{\tau}=\boldsymbol{y}|\boldsymbol{X}_t=\boldsymbol{x})\frac{h(\boldsymbol{y},\tau )}{h(\boldsymbol{x},t)}\tag{Doob's $h$-Transform}\label{eq:doob-h-eq1}
    \end{align}
    where $\mathbb{P}^h(\boldsymbol{X}_{\tau }=\boldsymbol{y}|\boldsymbol{X}_t=\boldsymbol{x})$ is the tilted transition kernel of $\mathbb{P}^h$. Then, $\mathbb{P}^h$ is Markov and has the associated SDE:
    \begin{align}
        d\boldsymbol{X}_t=\left[\boldsymbol{f}(\boldsymbol{x},t)+\sigma_t^2\nabla\log h (\boldsymbol{x},t)\right]dt+\sigma_td\boldsymbol{B}_t\tag{Doob's $h$-Transform SDE}\label{eq:doob-h-sde}
    \end{align}
\end{proposition}

\textit{Proof.} First, we show that the tilted transition density integrates to one:
\begin{align}
    \int_{\mathbb{R}^d}\mathbb{P}^h(\boldsymbol{X}_{\tau }=\boldsymbol{y}|\boldsymbol{X}_t=\boldsymbol{x})d\boldsymbol{y}&=\int_{\mathbb{R}^d}\mathbb{Q}(\boldsymbol{X}_{\tau }= \boldsymbol{y}|\boldsymbol{X}_t=\boldsymbol{x})\frac{h(\boldsymbol{y},\tau  )}{h(\boldsymbol{x},t)}d\boldsymbol{y}\nonumber\\
    &=\frac{1}{h(\boldsymbol{x},t)}\underbrace{\int_{\mathbb{R}^d}\mathbb{Q}(\boldsymbol{X}_{\tau }=\boldsymbol{y}|\boldsymbol{X}_t=\boldsymbol{x})h(\boldsymbol{y},\tau  )d\boldsymbol{y}}_{=:h(\boldsymbol{x},t)}\bluetext{=1}
\end{align}
which proves that $\mathbb{P}^h$ is a valid Markov process which depends only on $\boldsymbol{X}_t$. Next, we will derive the corresponding SDE that generates $\mathbb{P}^h$ by defining a smooth test function $\phi(\boldsymbol{x},t):\mathbb{R}^d\times [0,T]\to \mathbb{R}\in C^{2,1}(\mathbb{R}^d \times [0,T])$ and derive the expression for the generator $\mathcal{A}_t$ and using Itô's formula to recover the expression for the drift field $\boldsymbol{v}(\boldsymbol{x},t)$. By definition, the generator of a stochastic process is given by:
\begin{align}
    (\mathcal{A}^h_t\phi)(\boldsymbol{x})&:=\lim_{\Delta t \to 0}\left\{\frac{\bluetext{\mathbb{E}^h[\phi(\boldsymbol{X}_{t+\Delta t},t+\Delta t)|\boldsymbol{X}_t=\boldsymbol{x}]}-\phi(\boldsymbol{x})}{\Delta t}\right\}\label{eq:doob-h-eq2}
\end{align}
From (\ref{eq:doob-h-eq1}), we have that the expectation of the test function evaluated on the tilted measure can be written as:
\begin{align}
    \mathbb{E}^h[\phi(\boldsymbol{X}_{t+\Delta t},t+\Delta t )|\boldsymbol{X}_t=\boldsymbol{x}]&:=\frac{1}{h(\boldsymbol{x},t)}\mathbb{E}[\phi(\boldsymbol{X}_{t+\Delta t}, t+\Delta t )h(\boldsymbol{X}_{t+\Delta t}, t+\Delta t)|\boldsymbol{X}_t=\boldsymbol{x}]
\end{align}
which can be substituted back into (\ref{eq:doob-h-eq2}) to get:
\begin{align}
    (\mathcal{A}^h_t\phi)(\boldsymbol{X}_t)&=\lim_{\Delta t \to 0}\left\{\frac{{\mathbb{E}[\phi(\boldsymbol{X}_{t+\Delta t}, t+\Delta t 
    )\bluetext{h(\boldsymbol{X}_{t+\Delta t},t+\Delta t)}|\boldsymbol{X}_t=\boldsymbol{x}]}-\phi(\boldsymbol{x})\bluetext{h(\boldsymbol{x},t)}}{\bluetext{h(\boldsymbol{x},t)}\Delta t}\right\}\nonumber\\
    &=\frac{1}{h(\boldsymbol{x},t)}\underbrace{\lim_{\Delta t \to 0}\left\{\frac{{\mathbb{E}[\phi(\boldsymbol{X}_{t+\Delta t}, t+\Delta t)h(\boldsymbol{X}_{t+\Delta t},t+\Delta t)}|\boldsymbol{X}_t=\boldsymbol{x}]-\phi(\boldsymbol{x})h(\boldsymbol{x},t)}{\Delta t}\right\}}_{=:\bluetext{\mathcal{A}_t(\phi(\boldsymbol{x})h(\boldsymbol{x},t))}}\nonumber\\
    &=\frac{1}{h(\boldsymbol{x},t)}\bluetext{\mathcal{A}_t(\phi(\boldsymbol{x})h(\boldsymbol{x},t))}
\end{align}
where $\mathcal{A}_t$ is the generator of the untilted reference process defined in (\ref{eq:ito-uncontrolled-generator}). Now, using the expanded form of the generator derived in (\ref{eq:ito-uncontrolled-generator}) and denoting $\phi\equiv\phi(\boldsymbol{x},t)$ and $h\equiv h(\boldsymbol{x},t)$ for simplicity, we have:
\begin{small}
\begin{align}
    \mathcal{A}^h_t\phi&=\frac{1}{h}\left[\bluetext{\partial_t(\phi h )}+\pinktext{\langle \boldsymbol{f}, \nabla (\phi h)\rangle}+\greentext{\frac{\sigma_t^2}{2}\Delta(\phi h)}\right]\nonumber\\
    &=\frac{1}{h}\left[\bluetext{\phi\partial_th} +\pinktext{\langle \boldsymbol{f}, h\nabla \phi\rangle+\langle \boldsymbol{f},\phi\nabla h\rangle} +\greentext{\frac{\sigma_t^2}{2} \nabla\cdot (h\nabla \phi +\phi\nabla h)}\right]\nonumber\\
    &=\frac{1}{h}\left[\bluetext{\phi\partial_th} +\pinktext{h\langle \boldsymbol{f}, \nabla \phi\rangle+\phi\langle \boldsymbol{f},\nabla h\rangle} +\greentext{\frac{\sigma_t^2}{2}(h\Delta \phi +\phi\Delta h+2\nabla \phi\cdot \nabla h) }\right]\nonumber\\
\end{align}
\end{small}
Since, we defined the $h$-function as a conditional expectation $h(\boldsymbol{x},t)=\mathbb{E}_{\mathbb{Q}}[h(\boldsymbol{X}_\tau,\tau)|\boldsymbol{X}_t=\boldsymbol{x}]$, it satisfies the (\ref{eq:martingale-property}), and the expression for the unconditional generator vanishes:
\begin{small}
\begin{align}
    \mathcal{A}^h_t\phi&=\frac{1}{h}\phi\underbrace{\left[\partial_th +\langle \boldsymbol{f},\nabla h\rangle +\frac{\sigma_t^2}{2}\Delta h\right]}_{=\mathcal{A}_th=0\text{ ($h$ is a Martingale)}} +\frac{1}{h}\left[h\langle \boldsymbol{f}, \nabla \phi\rangle+\frac{\sigma_t^2}{2}(h\Delta \phi+2\nabla \phi\cdot \nabla h)\right]\nonumber\\
    &=\langle \boldsymbol{f}, \nabla \phi\rangle+\frac{\sigma_t^2}{2}\Delta \phi +\sigma_t^2\nabla \phi\cdot \bluetext{\underbrace{\frac{\nabla h}{h}}_{\nabla \log h}}\nonumber\\
    &=\langle \bluetext{\boldsymbol{f}}, \nabla \phi\rangle+\frac{\sigma_t^2}{2}\Delta \phi +\langle\nabla \phi, \bluetext{\sigma_t^2\nabla \log h}\rangle\nonumber\\
    &=\langle \bluetext{\underbrace{\boldsymbol{f}+\sigma_t^2\nabla \log h}_{\boldsymbol{v}(\boldsymbol{x},t)}}, \nabla \phi\rangle+\frac{\sigma_t^2}{2}\Delta \phi 
\end{align}
\end{small}
which means the control drift of the Doob $h$-transformed process is defined as $\boldsymbol{v}(\boldsymbol{x},t) :=\boldsymbol{f}(\boldsymbol{x},t)+\sigma_t^2\nabla \log h(\boldsymbol{x},t)$ and the corresponding SDE is:
\begin{align}
    \boxed{d\boldsymbol{X}_t=\left[\bluetext{\boldsymbol{f}(\boldsymbol{x},t)+\sigma_t^2\nabla \log h (\boldsymbol{x},t)}\right]dt+\sigma_td\boldsymbol{B}_t}\tag{Doob's $h$-Transform SDE}
\end{align}
and we conclude our proof. \hfill $\square$

The previous derivation explicitly shows how reweighting a reference diffusion with a positive space–time function $h(\boldsymbol{x},t)$ modifies the infinitesimal generator of the process. The following Corollary summarizes this result and provides an \textbf{expression for the generator of the reweighted path measure} in compact form.

\begin{corollary}[Generator of Reweighted Path Measure]\label{lemma:generator-reweighted}
    Let $\mathbb{Q}$ be a reference path measure with infinitesimal generator $\mathcal{A}_t$ and $h(\boldsymbol{x},t):\mathbb{R}^d\times[0,T]\to \mathbb{R}$ be the $h$-function defined in (\ref{eq:doob-h-function}). Then, the path measure \textbf{reweighted by $h$} is the $h$-transform of $\mathbb{Q}$ with generator defined as:
    \begin{small}
    \begin{align}
        \mathcal{A}_t^h\phi(\boldsymbol{x}):=\frac{\mathcal{A}_t(\phi(\boldsymbol{x}) h(\boldsymbol{x},t))-\phi(\boldsymbol{x})\mathcal{A}_th(\boldsymbol{x},t)}{h(\boldsymbol{x},t)}=\mathcal{A}_t\phi(\boldsymbol{x})+\langle \sigma_t^2\nabla \log h(\boldsymbol{x},t), \nabla \phi(\boldsymbol{x})\rangle
    \end{align}
    \end{small}
    
\end{corollary}

Therefore, Doob's $h$-transform provides a precise mechanism for incorporating endpoint information into the dynamics of a reference process. This mechanism allows us to define an $h$-function that exactly recovers the Markov dynamics of the optimal Schrödinger bridge.

\begin{corollary}[Schrödinger Bridge as Doob's $h$-Transform]\label{eq:sb-doob-h-transform}
    Given the Schrödinger potentials $(\varphi_t,\hat\varphi_t)$ that generate the solution to the (\ref{eq:dynamic-sb-problem}), we can define the $h$-function as:
    \begin{align}
        h(\boldsymbol{x},t):=\varphi_t(\boldsymbol{x})=\mathbb{E}_{\mathbb{Q}}[\varphi_T(\boldsymbol{X}_T) |\boldsymbol{X}_t=\boldsymbol{x}]=\int_{\mathbb{R}^d}\mathbb{Q}(\boldsymbol{X}_T=\boldsymbol{y}|\boldsymbol{X}_t=\boldsymbol{x})\varphi_T(\boldsymbol{X}_T)d\boldsymbol{y}\tag{$h$-Function for SB}
    \end{align}
    Then, the optimal Schrödinger bridge path measure $\mathbb{P}^\star$ is the Doob's $h$-transform of $\mathbb{Q}$, where for any $0\leq t\leq \tau\leq T$, we have:
    \begin{align}
        \mathbb{P}^\star(\boldsymbol{X}_{\tau }=\boldsymbol{y}|\boldsymbol{X}_t=\boldsymbol{x})=\mathbb{Q}(\boldsymbol{X}_{\tau}=\boldsymbol{y}|\boldsymbol{X}_t=\boldsymbol{x})\frac{h(\boldsymbol{y},\tau )}{h(\boldsymbol{x},t)}\tag{Doob's $h$-Transform for SB}\label{eq:doob-h-sb-eq}
    \end{align}
    Equivalently, $\mathbb{P}^\star$ is Markov and has the associated SDE:
    \begin{align}
        d\boldsymbol{X}_t=\left[\boldsymbol{f}(\boldsymbol{x},t)+\sigma_t^2\nabla\log h (\boldsymbol{x},t)\right]dt+\sigma_td\boldsymbol{B}_t\tag{Doob's $h$-Transform SDE}
    \end{align}
\end{corollary}

\textit{Proof Sketch.} From the derivation of the (\ref{eq:proof-sb-system}), the optimal path measure satisfies:
\begin{small}
\begin{align}
    \frac{\mathrm{d}\mathbb{P}^\star}{\mathrm{d}\mathbb{Q}}(\boldsymbol{X}_{0:T})=\hat\varphi_0(\boldsymbol{X}_0)\varphi_T(\boldsymbol{X}_T)
\end{align}
\end{small}
Conditioning on $\boldsymbol{X}_t=\boldsymbol{x}$ and using the Markov property of $\mathbb{Q}$, we observe that the future evolution from time $t$ to $\tau$ depends \textit{only on the terminal reweighting} through:
\begin{small}
\begin{align}
    \varphi_t(\boldsymbol{x})=\mathbb{E}_{\mathbb{Q}}[\varphi_T(\boldsymbol{X}_T)|\boldsymbol{X}_t=\boldsymbol{x}]
\end{align}
\end{small}
Therefore, the conditional transition kernel under $\mathbb{P}^\star$ is obtained by tilting the reference kernel by the ratio $\frac{\varphi_\tau(\boldsymbol{y})}{\varphi_t(\boldsymbol{x})}$, which is exactly Doob's $h$-transform. \hfill $\square$

This perspective shows that the Schrödinger bridge corresponds to a precise reweighting the reference process $\mathbb{Q}$ through a harmonic function $h(\boldsymbol{x},t):= \varphi_t(\boldsymbol{x})$ which modifies the forward drift by $\sigma_t^2\nabla\log h(\boldsymbol{x},t)$.  The resulting process therefore evolves according to the controlled drift $\boldsymbol{f}(\boldsymbol{x},t) + \sigma_t^2 \nabla \log h(\boldsymbol{x},t)$, which can be interpreted as the \textbf{minimal modification of the reference dynamics required to enforce the desired endpoint constraints}.

\subsection{Markovian and Reciprocal Projections}
\label{subsec:markov-reciprocal-proj}
In many settings, the Schrödinger bridge yields a path measure whose dependencies span the entire trajectory, making direct simulation difficult. This raises a natural question: \textit{Can we approximate a general bridge by a Markov process that is easier to simulate, while remaining as close as possible in relative entropy?}

The \boldtext{Markovian projection} provides a principled answer to this question. Given an arbitrary path measure, we project it onto the space of Markov measures $\mathcal{M}$ by minimizing KL divergence over processes whose future states depend \textit{only on the present state}. This produces the closest Markov approximation to the original bridge. Formally, we define the \boldtext{space of Markov measures} as:
\begin{align}
    \mathcal{M}:=\big\{\mathbb{M}\in \mathcal{P}(C([0,T]; \mathbb{R}^d))\;|\;\forall 0\leq s<t\leq T, \;\mathbb{E}_{\mathbb{M}}[f(\boldsymbol{X}_t)|\boldsymbol{X}_s]=\mathbb{E}_{\mathbb{M}}[f(\boldsymbol{X}_t)|\mathcal{F}_s]\big\}
\end{align}
where $\mathcal{F}_s$ denotes the filtration generated by the process up to time $s$ (Definition \ref{def:filtration-adapted-process}).

In the Schrödinger bridge setting, we seek to construct a stochastic bridge between empirical endpoint distributions $\pi_0$ and $\pi_T$ that is close in relative entropy to a reference measure $\mathbb{Q}$ defined by the SDE: 
\begin{small}
\begin{align}
    \mathbb{Q}&:d\boldsymbol{X}_t=\boldsymbol{f}(\boldsymbol{X}_t,t)dt+\sigma_td\boldsymbol{B}_t\label{eq:single-bridge-ref-sde}\tag{Reference SDE}
\end{align}
\end{small}
Given a coupling $\pi_{0,T}\in \mathcal{P}(\mathbb{R}^d\times\mathbb{R}^d)$ between the endpoint distributions, we will consider the path measure generated by \boldtext{mixture of endpoint-conditioned bridges} defined as: 
\begin{small}
\begin{align}
    \Pi=\pi_{0,T}\mathbb{Q}_{\cdot|0,T}\in \mathcal{P}(C([0,T];\mathbb{R}^d))\tag{Mixture of Bridges}
\end{align}
\end{small}
which can be interpreted as constructing a bridge by first sampling endpoints $(\boldsymbol{x}_0, \boldsymbol{x}_T) \sim \Pi_{0,T}$ and then generating the intermediate bridge dynamics according to the reference path measure conditioned on the endpoints. For fixed $(\boldsymbol{x}_0, \boldsymbol{x}_T)$, the corresponding \textbf{conditional bridge dynamics} under the reference measure $\mathbb{Q}$ are given by the (\ref{eq:doob-h-sde}) defined as:
\begin{align}
    \mathbb{Q}_{\cdot |0,T}(\cdot |\boldsymbol{x}_0, \boldsymbol{x}_T)&: d\boldsymbol{X}_t=\left[\boldsymbol{f}(\boldsymbol{X}_t,t)+\sigma^2_t\nabla \log \mathbb{Q}_{T|t}(\boldsymbol{x}_T|\boldsymbol{X}_t)\right]dt+\sigma_td\boldsymbol{B}_t\label{eq:single-bridge-sde}\tag{Bridge SDE}
\end{align}
where $\mathbb{Q}_{T|t}(\boldsymbol{x}_T|\boldsymbol{X}_t)$ is the probability of reaching $\boldsymbol{x}_T$ from the current state $\boldsymbol{X}_t$ via the SDE of $\mathbb{Q}$. Since the drift in (\ref{eq:single-bridge-sde}) is \textbf{conditioned on the future endpoint} $\boldsymbol{x}_T$, it is not Markov with respect to $\boldsymbol{X}_t$. Therefore, simulating the SDE requires evaluating the transition density $\mathbb{Q}_{T|t}(\boldsymbol{x}_T|\boldsymbol{X}_t)$ at every step, which in general is computationally intractable.

To obtain a tractable process, we instead construct the \boldtext{Markovian projection of the bridge measure $\Pi$}, which is the Markov process whose drift depends only on the current state $\boldsymbol{X}_t$ and that minimizes the KL divergence to the original bridge measure.

\begin{proposition}[Markovian Projection (Proposition 2 in \citet{shi2023diffusion})]\label{prop:markov-proj}
    Consider a \textbf{mixture of bridges} $\Pi=\Pi_{0,T}\mathbb{Q}_{\cdot|0,T}$ that bridges distributions $\pi_0$ and $\pi_T$ via the endpoint law $\Pi_{0,T}$, where each conditional bridge is defined by (\ref{eq:single-bridge-sde}) generated from the reference measure. Then, the \textbf{Markovian projection} of $\Pi$ is denoted:
    \begin{align}
        \mathbb{M}^\star:=\text{proj}_{\mathcal{M}}(\Pi)\in \mathcal{M}\tag{Markovian Projection}\label{eq:markovian-projection}
    \end{align}
    with the associated SDE:
    \begin{align}
        \mathbb{M}^\star: \;&d\boldsymbol{X}_t=\left[\boldsymbol{f}(\boldsymbol{X}_t,t)+\sigma_t\boldsymbol{u}^\star(\boldsymbol{X}_t,t)\right]dt+\sigma_td\boldsymbol{B}_t\tag{Markovian Projection SDE}\label{eq:markov-proj-sde}\\
        &\text{s.t. }\quad \boldsymbol{u}^\star(\boldsymbol{x},t)=\sigma_t\mathbb{E}_{\Pi_{T|t}}\left[\nabla \log \mathbb{Q}_{T|t}(\boldsymbol{X}_T|\boldsymbol{X}_t)|\boldsymbol{X}_t=\boldsymbol{x}\right]\tag{Markovian Projection Drift}\label{eq:markov-proj-optimal-drif}
    \end{align}
   where $\sigma_t> 0$.  Then, $\mathbb{M}^\star$ satisfies the following properties:
   \begin{enumerate}
       \item It is the Markov measure $\mathbb{M}^\star\in \mathcal{M}$ that \textbf{minimizes the reverse KL divergence} with the mixture of bridges $\Pi$ such that:
       \begin{small}
        \begin{align}
           \mathbb{M}^\star&=\underset{\mathbb{M}}{\arg\min}\{\text{KL}\left(\Pi|\mathbb{M}\right):\mathbb{M}\in \mathcal{M}\}\\
           \text{KL}\left(\Pi|\mathbb{M}\right)&=\frac{1}{2}\int_0^T\mathbb{E}_{\Pi_{0,t}}\bigg[\big\|\sigma_t\mathbb{E}_{\Pi_{T|0,t}}\big[\nabla  \log \mathbb{Q}_{T|t}(\boldsymbol{X}_T|\boldsymbol{X}_t)|\boldsymbol{X}_0, \boldsymbol{X}_T\big]-\boldsymbol{u}^\star(\boldsymbol{X}_t,t)\big\|^2\bigg]
       \end{align}
       \end{small}
       \item It preserves the time marginals of $\Pi_t$ for all $t\in [0,T]$ such that:
       \begin{align}
           \forall t\in [0,T], \quad \mathbb{M}^\star_t=\Pi_t
       \end{align}
   \end{enumerate}
\end{proposition}

\textit{Proof.} We break down the proof in three steps. 

\textbf{Step 1: Derive the Optimal Control Drift of the Markovian Projection.} 
Recall that the bridge path measure can be obtained by reweighting the reference measure $\mathbb{Q}$ by the Radon–Nikodym derivative of the endpoint coupling $\Pi_{0,T}$ with respect to the reference endpoint law $\mathbb{Q}_{0,T}$:
\begin{small}
\begin{align}
    \pi_{0,T}(\boldsymbol{x}_0, \boldsymbol{x}_T):=\frac{\mathrm{d}\Pi_{0,T}}{\mathrm{d}\mathbb{Q}_{0,T}}(\boldsymbol{x}_0, \boldsymbol{x}_T)\implies \Pi(\boldsymbol{X}_{0:T})=\pi_{0,T}(\boldsymbol{X}_0, \boldsymbol{X}_T)\mathbb{Q}(\boldsymbol{X}_{0:T})
\end{align}
\end{small}
Conditioning on the initial state $\boldsymbol{X}_0=\boldsymbol{x}_0$, the corresponding \textbf{conditional bridge law} becomes:
\begin{small}
\begin{align}
    \Pi(\boldsymbol{X}_{0:T}  |\boldsymbol{X}_0=\boldsymbol{x}_0)=h(\boldsymbol{X}_T,T)\mathbb{Q}(\boldsymbol{X}_{0:T} |\boldsymbol{X}_0=\boldsymbol{x}_0), \quad \text{where}\quad h(\boldsymbol{X}_T,T):= \frac{\mathrm{d}\Pi(\boldsymbol{X}_T|\boldsymbol{X}_0=\boldsymbol{x}_0)}{\mathrm{d}\mathbb{Q}(\boldsymbol{X}_T|\boldsymbol{X}_0=\boldsymbol{x}_0)}\label{eq:markov-proj-proof2}
\end{align}
\end{small}
This means that given an initial state $\boldsymbol{x}_0$, the bridge law is obtained by reweighting the reference diffusion by a terminal weight. This is exactly the (\ref{eq:doob-h-function}) from (\ref{eq:doob-h-eq1}) for time $t\to T$ with $h$-function defined as:
\begin{small}
\begin{align}
    h(\boldsymbol{x},t)=\int_{\mathbb{R}^d}\mathbb{Q}_{T|t}(\boldsymbol{X}_T=\boldsymbol{x}_T|\boldsymbol{X}_t=\boldsymbol{x})h(\boldsymbol{X}_T,T)d\boldsymbol{x}_T=\mathbb{E}_{\mathbb{Q}}[h(\boldsymbol{X}_T,T)|\boldsymbol{X}_t=\boldsymbol{x},\boldsymbol{X}_0=\boldsymbol{x}_0]\tag{Doob's $h$-transform}\label{eq:markov-proj-proof1}
\end{align}
\end{small}
which is the \textbf{conditional expectation of the endpoint weight under the reference process}. Corollary \ref{lemma:generator-reweighted} establishes the fact that reweighting paths from $\mathbb{Q}$ by the $h$-function changes the generator by adding the drift $\langle \sigma_t^2\nabla\log h(\boldsymbol{x},t),  \nabla \boldsymbol{f}(\boldsymbol{x},t)\rangle$ which is still Markov. Therefore, the bridge measure $\Pi^{\cdot|0=\boldsymbol{x}_0}$ conditioned on $\boldsymbol{X}_0=\boldsymbol{x}_0$ follows the SDE:
\begin{align}
    d\boldsymbol{X}_t=\left[\boldsymbol{f}(\boldsymbol{X}_t,t)+\sigma_t^2\nabla _{\boldsymbol{x}}\log h(\boldsymbol{X}_t,t)\right]dt +\sigma_td\boldsymbol{B}_t\label{eq:markov-eq-4}
\end{align}
To derive the form of the correction term $\nabla \log h(\boldsymbol{X}_t,t)$ in terms of $\mathbb{Q}$, we can differentiate (\ref{eq:markov-proj-proof1}) with respect to $\boldsymbol{x}$ to get:
\begin{small}
\begin{align}
    \bluetext{\nabla }h(\boldsymbol{x},t)&=\bluetext{\nabla }\int_{\mathbb{R}^d}\mathbb{Q}_{T|t}(\boldsymbol{X}_T=\boldsymbol{x}_T|\boldsymbol{X}_t=\boldsymbol{x})h(\boldsymbol{X}_T,T)d\boldsymbol{x}_T\nonumber\\
    &=\int_{\mathbb{R}^d}h(\boldsymbol{X}_T,T)\bluetext{\nabla }\mathbb{Q}_{T|t}(\boldsymbol{X}_T=\boldsymbol{x}_T|\boldsymbol{X}_t=\boldsymbol{x})d\boldsymbol{x}_T
\end{align}
\end{small}
To get the expression for the gradient of the logarithm, we divide both sides by $h(\boldsymbol{x},t)$ to get:
\begin{small}
\begin{align}
    \bluetext{\nabla \log h(\boldsymbol{x},t)}&=\int_{\mathbb{R}^d}\frac{h(\boldsymbol{X}_T,T)}{\bluetext{h(\boldsymbol{x},t)}}\nabla \mathbb{Q}_{T|t}(\boldsymbol{X}_T=\boldsymbol{x}_T|\boldsymbol{X}_t=\boldsymbol{x})d\boldsymbol{x}_T
\end{align}
\end{small}
Since we want to obtain a \textit{score-like} function for the reference process, we can rewrite $\nabla \mathbb{Q}_{T|t}=\mathbb{Q}_{T|t}\nabla \log \mathbb{Q}_{T|t}$ then substitute the identity $h(\boldsymbol{x},T)=\frac{\mathrm{d} \Pi_{T|0}}{d \mathbb{Q}_{T|0}}(\boldsymbol{x}|\boldsymbol{x}_0)$ from (\ref{eq:markov-proj-proof2}) to get:
\begin{small}
\begin{align}
    \bluetext{\nabla \log h(\boldsymbol{x},t)}&=\int_{\mathbb{R}^d}\frac{\pinktext{h(\boldsymbol{X}_T,T)}\bluetext{\mathbb{Q}_{T|t}(\boldsymbol{x}_T|\boldsymbol{x})}}{\bluetext{h(\boldsymbol{x},t)}}\nabla \log\mathbb{Q}_{T|t}(\boldsymbol{x}_T|\boldsymbol{x})d\boldsymbol{x}_T\nonumber\\
    &=\int_{\mathbb{R}^d}\frac{\pinktext{\Pi_{T|0}(\boldsymbol{x}_T|\boldsymbol{x}_0)}\mathbb{Q}_{T|t}(\boldsymbol{x}_T|\boldsymbol{x})}{\pinktext{\mathbb{Q}_{T|0}(\boldsymbol{x}_T|\boldsymbol{x}_0)}h(\boldsymbol{x},t)}\nabla \log\mathbb{Q}_{T|t}(\boldsymbol{x}_T|\boldsymbol{x})d\boldsymbol{x}_T\label{eq:markov-eq-1}
\end{align}
\end{small}

The ratio $\frac{\mathbb{Q}_{T|t}(\boldsymbol{x}_T|\boldsymbol{x})}{\mathbb{Q}_{T|0}(\boldsymbol{x}_T|\boldsymbol{x}_0)}$ between reference conditionals can be rewritten using the \textbf{Markov property of the reference measure} which states that the distribution of $\boldsymbol{x}, \boldsymbol{x}_T$ given $\boldsymbol{x}_0$ is equal, regardless of the order of sampling $0\to T\to t$ or $0\to t\to T$:
\begin{small}
\begin{align}
    \mathbb{Q}_{T|0}(\boldsymbol{x}_T|\boldsymbol{x}_0)\mathbb{Q}_{t|0,T}(\boldsymbol{x}|\boldsymbol{x}_0,\boldsymbol{x}_T)&=\mathbb{Q}_{t|0}(\boldsymbol{x}|\boldsymbol{x}_0)\mathbb{Q}_{T|t}(\boldsymbol{x}_T|\boldsymbol{x})\implies \bluetext{\frac{\mathbb{Q}_{T|t}(\boldsymbol{x}_T|\boldsymbol{x})}{\mathbb{Q}_{T|0}(\boldsymbol{x}_T|\boldsymbol{x}_0)}}=\frac{\mathbb{Q}_{t|0,T}(\boldsymbol{x}|\boldsymbol{x}_0,\boldsymbol{x}_T)}{\mathbb{Q}_{t|0}(\boldsymbol{x}|\boldsymbol{x}_0)}
\end{align}
\end{small}
Substituting the expression into (\ref{eq:markov-eq-1}), we get:
\begin{small}
\begin{align}
    \nabla \log h(\boldsymbol{x},t)&=\int_{\mathbb{R}^d}\frac{\bluetext{\Pi_{T|0}(\boldsymbol{x}_T|\boldsymbol{x}_0)\mathbb{Q}_{t|0,T}(\boldsymbol{x}|\boldsymbol{x}_0,\boldsymbol{x}_T)}}{\pinktext{\mathbb{Q}_{t|0}(\boldsymbol{x}|\boldsymbol{x}_0)h(\boldsymbol{x},t)}}\nabla\log \mathbb{Q}_{T|t}(\boldsymbol{x}_T|\boldsymbol{x})d\boldsymbol{x}_T\label{eq:markov-eq-3}
\end{align}
\end{small}

Now, we observe that the \textbf{numerator} $\Pi_{T|0}(\boldsymbol{x}_T|\boldsymbol{x}_0)\mathbb{Q}_{t|0,T}(\boldsymbol{x}|\boldsymbol{x}_0,\boldsymbol{x}_T)$ is equivalent to the joint law of $(\boldsymbol{x}, \boldsymbol{x}_T)$ given $\boldsymbol{X}_0=\boldsymbol{x}_0$:
\begin{small}
\begin{align}
    \Pi_{T|0}(\boldsymbol{x}_T|\boldsymbol{x}_0)\mathbb{Q}_{t|0,T}(\boldsymbol{x}|\boldsymbol{x}_0,\boldsymbol{x}_T)=\Pi_{t,T|0}(\boldsymbol{x}, \boldsymbol{x}_T|\boldsymbol{x}_0)\tag{Numerator of (\ref{eq:markov-eq-3})}
\end{align}
\end{small}
Furthermore, marginalizing the conditional bridge law $\Pi(\boldsymbol{X}_{0:T}  |\boldsymbol{X}_0=\boldsymbol{x}_0)=h(\boldsymbol{X}_T,T)\mathbb{Q}(\boldsymbol{X}_{0:T} |\boldsymbol{X}_0=\boldsymbol{x}_0)$ in (\ref{eq:markov-proj-proof2}) to time $t$ gives us the following expression for the \textbf{denominator}:
\begin{small}
\begin{align}
    \mathbb{Q}_{t|0}(\boldsymbol{x}|\boldsymbol{x}_0)h(\boldsymbol{x},t)=\Pi_{t|0}(\boldsymbol{x}|\boldsymbol{x}_0)\tag{Denominator of (\ref{eq:markov-eq-3})}
\end{align}
\end{small}
by the definition of (\ref{eq:markov-proj-proof1}) as $h(\boldsymbol{x},t)=\mathbb{E}_{\mathbb{Q}}[h(\boldsymbol{X}_T,T)|\boldsymbol{X}_t=\boldsymbol{x},\boldsymbol{X}_0=\boldsymbol{x}_0]$. Plugging these expressions into (\ref{eq:markov-eq-3}) yields:
\begin{align}
    \nabla \log h(\boldsymbol{x},t)&=\int_{\mathbb{R}^d}\bluetext{\underbrace{\frac{\Pi_{t,T|0}(\boldsymbol{x}, \boldsymbol{x}_T|\boldsymbol{x}_0)}{\Pi_{t|0}(\boldsymbol{x}|\boldsymbol{x}_0)}}_{=\Pi_{T|0,t}(\boldsymbol{x}_T|\boldsymbol{x}, \boldsymbol{x}_0}}\nabla \log \mathbb{Q}_{T|t}(\boldsymbol{x}_T|\boldsymbol{x})d\boldsymbol{x}_T\tag{Bayes' rule}\\
    &=\underbrace{\int_{\mathbb{R}^d}\nabla \log \mathbb{Q}_{T|t}(\boldsymbol{x}_T|\boldsymbol{x})\mathrm{d}\Pi_{T|0,t}(\boldsymbol{x}_T|\boldsymbol{x}, \boldsymbol{x}_0)}_{\text{expectation over $\Pi$ given $\boldsymbol{X}_0=\boldsymbol{x}_0$ and $\boldsymbol{X}_t=\boldsymbol{x}$}}\nonumber\\
    &=\mathbb{E}_{\Pi_{T|t,0}}\left[\nabla \log \mathbb{Q}_{T|t}(\boldsymbol{x}_T|\boldsymbol{x})\big|\boldsymbol{X}_0=\boldsymbol{x}_0, \boldsymbol{X}_t=\boldsymbol{x}\right]
\end{align}
which can be used to express the correction drift in (\ref{eq:markov-eq-4}) as a conditional expectation of endpoint-conditioned drifts. Substituting this expression into the SDE for $\Pi_{t|0}$ defined in (\ref{eq:markov-eq-4}), we get an SDE with \textbf{non-Markov drift}:
\begin{align}
    d\boldsymbol{X}_t=\left[\boldsymbol{f}(\boldsymbol{X}_t,t)+\sigma_t^2\mathbb{E}_{\Pi_{T|t,0}}\left[\nabla \log \mathbb{Q}_{T|t}(\boldsymbol{X}_T|\boldsymbol{X}_t)\big|\boldsymbol{X}_t, \boldsymbol{X}_0\right]\right]dt +\sigma_td\boldsymbol{B}_t
\end{align}

\textbf{Step 2: Show that Markovian Projection is the Minimizer of Reverse KL Divergence. }
Let $\mathbb{M}$ be an arbitrary Markov path measure that evolves via the SDE $d\boldsymbol{X}_t=[\boldsymbol{f}(\boldsymbol{X}_t,t)+\boldsymbol{u}(\boldsymbol{X}_t,t)]dt+\sigma_td\boldsymbol{B}_t$. Then, we can write the KL divergence between $\Pi$ and $\mathbb{M}$ with the same diffusion coefficient $\sigma_t$ and using Corollary \ref{corollary:kl-divergence-ito} as the squared difference between their drifts:
\begin{align}
    \text{KL}(\Pi\|\mathbb{M})=\frac{1}{2}\int_0^T\mathbb{E}_{\Pi_{0,t}}\left[\|\sigma_t\mathbb{E}_{\Pi_{T|0,t}}\left[\nabla \log \mathbb{Q}_{T|t}(\boldsymbol{X}_T|\boldsymbol{X}_t)\big|\boldsymbol{X}_t, \boldsymbol{X}_0\right]-\boldsymbol{u}(\boldsymbol{X}_t,t)\|^2\right]dt
\end{align}

Minimizing this KL yields the optimal $\boldsymbol{u}^\star(\boldsymbol{x}, t)$ that best approximates the non-Markovian drift $\sigma_t\mathbb{E}_{\Pi_{T|t,0}}\left[\nabla\log \mathbb{Q}_{T|t}(\boldsymbol{x}_T|\boldsymbol{x})\big|\boldsymbol{X}_0, \boldsymbol{X}_t\right]$ using only $\boldsymbol{X}_t$ across all time steps $t$. For any random variable $\boldsymbol{Z}$, the minimizer of the expected squared loss with respect to a function is equal to the expectation of the random variable given only the input to the function:
\begin{small}
\begin{align}
    \underset{\boldsymbol{b}(\boldsymbol{X}_t)}{\arg\min}\mathbb{E}[\|\boldsymbol{Z}-\boldsymbol{b}(\boldsymbol{X}_t)\|^2]=\mathbb{E}[\boldsymbol{Z}|\boldsymbol{X}_t]
\end{align}
\end{small}
Therefore, the minimizer is the optimal Markovian drift given by:
\begin{align}
    \boldsymbol{u}^\star(\boldsymbol{x},t)=\sigma_t\mathbb{E}_{\Pi_{T|t}}\left[\nabla\log \mathbb{Q}_{T|t}(\boldsymbol{X}_T|\boldsymbol{X}_t)\big|\boldsymbol{X}_t=\boldsymbol{x}\right]
\end{align}
which concludes the proof of the first condition \textbf{(i)} in Proposition \ref{prop:markov-proj}. 

\textbf{Step 3: Prove Equality of the Marginal Distributions. }
To prove the second condition \textbf{(ii)} which states that $\mathbb{M}^\star_t=\Pi_t$ for all $t\in [0,T]$, we first establish that $\mathbb{M}^\star_t$ and $\Pi_t$ solve the same Fokker-Planck equation given by:
\begin{align}
    \partial_t p_t(\boldsymbol{X}_t,t)=-\nabla\cdot \left((\boldsymbol{f}(\boldsymbol{X}_t,t)+\boldsymbol{u}^\star(\boldsymbol{X}_t,t))p_t(\boldsymbol{X}_t)\right)+\frac{\sigma_t^2}{2}\Delta p_t(\boldsymbol{X}_t)\tag{Fokker-Planck Equation}\label{eq:markov-proj-proof-fp}
\end{align}
Since the divergence extracts only the drift that depends on $\boldsymbol{X}_t$, and the drift for $\Pi$ contains only an additional dependence on $\boldsymbol{X}_0$, the \textit{effective drift} in the (\ref{eq:markov-proj-proof-fp}) is equal for $\Pi_t$ and $\mathbb{M}^\star_t$. Given that both diffusion coefficients are also equal, we conclude that they both satisfy the same (\ref{eq:markov-proj-proof-fp}). Following the uniqueness of the solution to the Fokker-Planck equation given fixed initial conditions \footnote{For proof, see \citep{bogachev2021uniqueness}}, we conclude that the time marginals are equal $\mathbb{M}^\star_t=\Pi_t$. \hfill $\square$

While the Markovian projection provides a way of simulating endpoint-conditioned bridges with only dependence on the current state, it generally fails to preserve the \boldtext{bridge measure} of $\mathbb{Q}_{\cdot|0,T}\equiv \mathbb{Q}(\cdot|\boldsymbol{x}_0, \boldsymbol{x}_T)$, which is the distribution over bridge paths conditioned on a pair of endpoints $(\boldsymbol{x}_0, \boldsymbol{x}_T)\sim \Pi_{0,T}$. To define a measure that exactly matches the bridge of the endpoint-conditioned reference measure $\mathbb{Q}_{\cdot|0,T}$, we define the \textbf{reciprocal projection} which projects any path measure to the \textbf{reciprocal class} $\mathcal{R}(\mathbb{Q})$ of $\mathbb{Q}$. To understand the reciprocal projection, we first define a \boldtext{reciprocal process} and a unique property that will become useful in later sections.

\begin{proposition}[Reciprocal Processes (Lemma 1.4 in \citet{jamison1974reciprocal})]\label{eq:prop:reciprocal-process}
    A stochastic process $(\boldsymbol{X}_t)_{t\in[0,T]}$ on a measurable state space $\mathcal{X}$ is considered a \textbf{reciprocal process} if for any $0\le s<t<\tau\le T$ and any bounded measurable function $\phi(\boldsymbol{x}):\mathcal{X}\to\mathbb R$, the \textbf{reciprocal property} holds:
    \begin{align}
    \mathbb E[\phi(\boldsymbol{X}_t)| \boldsymbol{X}_{0:s},\boldsymbol{X}_{\tau :T}]= \mathbb E[\phi(\boldsymbol{X}_t)| \boldsymbol{X}_s,\boldsymbol{X}_\tau ]\tag{Reciprocal Property}\label{eq:reciprocal-property}
    \end{align}
    which means that the interior of any interval $(s,\tau)$ is \textit{conditionally independent} of the rest of the trajectory given the two boundary states $(\boldsymbol{X}_s,\boldsymbol{X}_\tau)$. While reciprocal processes are generally not Markov, fixing either boundary state ($\boldsymbol{X}_s$ or $\boldsymbol{X}_\tau)$ to a constant yields a Markov process. 
\end{proposition}

\textit{Proof.} To prove the Markov property of reciprocal processes given a fixed constant endpoint, we first observe that conditioning on a fixed constant does not change the conditional expectation. Suppose first that $\boldsymbol{X}_\tau=\bar{\boldsymbol{x}}$ almost surely. Let $s \le t_1 < \cdots < t_k < t < \tau$ and $\phi(\boldsymbol{x}):\mathcal X\to\mathbb R$ be any bounded and measurable test function. Since $\boldsymbol{X}_\tau$ is a constant, conditioning on $\boldsymbol{X}_\tau$ does not change the conditional expectation:
\begin{align}
    \mathbb E[\phi(\boldsymbol{X}_t)| \boldsymbol{X}_{t_1},\dots,\boldsymbol{X}_{t_k}] = \mathbb E[\phi(\boldsymbol{X}_t)| \boldsymbol{X}_{t_1},\dots,\boldsymbol{X}_{t_k},\bluetext{\boldsymbol{X}_\tau}].
\end{align}
By the \textbf{reciprocal property}, once the two boundary states $(\boldsymbol{X}_{t_k},\boldsymbol{X}_\tau)$ are given, the interior of the interval $(t_k,\tau)$ is conditionally independent of the past before $t_n$. Therefore, we have:
\begin{align}
    \mathbb E[\phi(\boldsymbol{X}_t)| \boldsymbol{X}_{t_1},\dots,\boldsymbol{X}_{t_k},\boldsymbol{X}_\tau] = \mathbb E[\phi(\boldsymbol{X}_t)| \boldsymbol{X}_{t_k},\boldsymbol{X}_\tau].
\end{align}
Since $\boldsymbol{X}_\tau=\bar{\boldsymbol{x}}$ is constant, this reduces to:
\begin{align}
    \mathbb E[\phi(\boldsymbol{X}_t)| \boldsymbol{X}_{t_k},\boldsymbol{X}_\tau]= \mathbb E[\phi(\boldsymbol{X}_t)| \boldsymbol{X}_{t_k}].
\end{align}
Therefore, we conclude that the future state $\boldsymbol{X}_t$ can be determined with only knowledge of the current state $\boldsymbol{X}_{t_k}$ and independent of the past trajectory prior to $t_k$:
\begin{align}
    \mathbb E[\phi(\boldsymbol{X}_t)| \boldsymbol{X}_{t_1},\dots,\boldsymbol{X}_{t_k}] = \mathbb E[\phi(\boldsymbol{X}_t)| \boldsymbol{X}_{t_k}],
\end{align}
which is exactly the \textbf{Markov property}. If instead $\boldsymbol{X}_s$ is fixed, the same conclusion follows by time reversal, since both the reciprocal and Markov properties are preserved under reversing time. \hfill $\square$

Since we have shown that \textbf{reciprocal processes} are characterized by their \textbf{conditional independence structure}, we will now apply this idea to level of path measures. In particular, reciprocal processes naturally arise as mixtures of Markov bridges sharing the same conditional dynamics between endpoints. This observation leads to the notion of the \boldtext{reciprocal class} $\mathcal{R}(\mathbb{Q})$ of a reference path measure $\mathbb{Q}$. 

Intuitively, all path measures in the reciprocal class $\mathcal{R}(\mathbb{Q})$ generate trajectories with the \textbf{same bridge dynamics} as $\mathbb{Q}$, differing only in how probability mass is assigned to the endpoint pairs $(\boldsymbol{x}_0,\boldsymbol{x}_T)$. Concretely, this class consists of all path measures obtained by reweighting the endpoint distribution of $\mathbb{Q}$ while preserving its conditional bridge law. 

\begin{definition}[Reciprocal Class]\label{def:reciprocal-class}
The \textbf{reciprocal class} of $\mathbb{Q}$ is the collection of all path measures that share the same \textit{conditional bridge distribution} given fixed endpoints $(\boldsymbol{x}_0, \boldsymbol{x}_T)\sim \Pi_{0,T}$ but may differ in their endpoint law, which defines how the endpoints are weighted. Formally, we define the reciprocal class $\mathcal{R}(\mathbb{Q})$ as:
\begin{align}
    \mathcal{R}(\mathbb{Q}):=\{\Pi\in \mathcal{P}(C([0,T] ; \mathbb{R}^d)) :\Pi=\Pi_{0,T}\mathbb{Q}_{\cdot|0,T}\}
\end{align}
where $\Pi_{0,T}\in \mathcal{P}(\mathbb{R}^d, \mathbb{R}^d)$ denotes an arbitrary endpoint coupling and $\mathbb{Q}_{\cdot|0,T}$ is the endpoint-conditioned bridge distribution of $\mathbb{Q}$. Equivalently, a path measure belongs in the reciprocal class $\Pi\in \mathcal{R}(\mathbb{Q})$ if it admits the \textbf{mixture-of-bridges representation}:
\begin{align}
    \Pi(\cdot)=\int_{\mathbb{R}^d\times \mathbb{R}^d}\mathbb{Q}_{\cdot|0,T}(\cdot|\boldsymbol{x}_0, \boldsymbol{x}_T)\mathrm{d}\Pi_{0,T}(\boldsymbol{x}_0, \boldsymbol{x}_T)
\end{align}
which can be written equivalently as $\Pi=\Pi_{0,T}\mathbb{Q}_{\cdot|0,T}$.
\end{definition}

Having characterized the reciprocal class $\mathcal{R}(\mathbb{Q})$ as the family of path measures that share the same bridge dynamics as the reference process $\mathbb{Q}$, we now consider how to \textbf{approximate an arbitrary path measure} $\mathbb{P}$ by an element of the reciprocal class. A natural approach is to choose the measure in $\mathcal{R}(\mathbb{Q})$ that is closest to $\mathbb{P}$ in relative entropy. This is exactly the goal of the \boldtext{reciprocal projection}, which we show is the \textit{unique minimizer} of the KL divergence from $\mathbb{P}$.

\begin{proposition}[Reciprocal Projection (Proposition 4 in \citet{shi2023diffusion})]\label{prop:reciprocal-proj}
Let $\mathbb{Q}$ be a reference path measure and $\mathbb{P}$ be an arbitrary path measure that we wish to project onto the reciprocal class of $\mathbb{Q}$. The \textbf{reciprocal projection} of $\mathbb{P}$ onto the reciprocal class $\mathcal{R}(\mathbb{Q})$ is defined as:
\begin{align}
    \Pi^\star:=\text{proj}_{\mathcal{R}(\mathbb{Q})}(\mathbb{P})\in \mathcal{R}(\mathbb{Q})\tag{Reciprocal Projection}\label{eq:reciprocal-projection}
\end{align}
Then, $\Pi^\star$ is the element of the reciprocal class which \textbf{minimizes the KL divergence} from $\mathbb{P}$:
\begin{align}
    \Pi^\star=\underset{\Pi\in \mathcal{R}(\mathbb{Q})}{\arg\min}\text{KL}(\mathbb{P}\|\Pi)
\end{align}
Furthermore, the reciprocal projection admits the \textbf{mixture-of-bridges representation}:
\begin{align}
    \Pi^\star(\boldsymbol{X}_{0:T})=\int_{\mathbb{R}^d\times \mathbb{R}^d}\mathbb{Q}_{\cdot|0,T}(\boldsymbol{X}_{0:T}|\boldsymbol{x}_0, \boldsymbol{x}_T)\mathrm{d}\mathbb{P}_{0,T}(\boldsymbol{x}_0, \boldsymbol{x}_T) \iff \Pi^\star=\mathbb{P}_{0,T}\mathbb{Q}_{\cdot|0,T}\label{eq:reciprocal-projection-2}
\end{align}
where $\mathbb{P}_{0,T}$ denotes the endpoint distribution of $\mathbb{P}$ and $\mathbb{Q}_{\cdot|0,T}(\cdot|\boldsymbol{x}_0, \boldsymbol{x}_T)$ is the reference bridge conditioned on the endpoints. 
\end{proposition}

\textit{Proof.} We start by applying the (\ref{eq:chain-rule-kl}) from Lemma \ref{lemma:chain-rule-kl} to decompose $\text{KL}(\mathbb{P}\|\Pi)$ into the sum of the \textbf{divergence between endpoints} and the \textbf{divergence in bridge measure}:
\begin{align}
    \text{KL}(\mathbb{P}\|\Pi)&=\underbrace{\text{KL}(\mathbb{P}_{0,T}\|\Pi_{0,T})}_{\text{endpoint divergence}}+\underbrace{\mathbb{E}_{(\boldsymbol{X}_0, \boldsymbol{X}_T)\sim\mathbb{P}_{0,T}}\left[\text{KL}(\mathbb{P}_{\cdot|0,T}(\cdot|\boldsymbol{X}_0, \boldsymbol{X}_T)\|\Pi_{\cdot|0,T}(\cdot|\boldsymbol{X}_0, \boldsymbol{X}_T))\right]}_{\text{bridge divergence}}
\end{align}

Since all elements of the reciprocal class $\mathcal{R}(\mathbb{Q})$ have the same bridge as $\mathbb{Q}$ such that $\Pi_{\cdot|0,T}(\cdot|\boldsymbol{x}_0, \boldsymbol{x}_T)=\mathbb{Q}_{\cdot|0,T}(\cdot|\boldsymbol{x}_0, \boldsymbol{x}_T)$, we can write:
\begin{align}
    \text{KL}(\mathbb{P}\|\Pi)&=\text{KL}(\mathbb{P}_{0,T}\|\Pi_{0,T})+\underbrace{\mathbb{E}_{(\boldsymbol{X}_0, \boldsymbol{X}_T)\sim \mathbb{P}_{0,T}}\left[\text{KL}(\mathbb{P}_{\cdot|0,T}(\cdot|\boldsymbol{X}_0, \boldsymbol{X}_T)\|{\color{MyGreen}\mathbb{Q}_{\cdot|0,T}(\cdot|\boldsymbol{X}_0, \boldsymbol{X}_T)})\right]}_{\text{not dependent on $\Pi$}}
\end{align}

Since the bridge divergence is no longer dependent on the measure $\Pi$ which we are optimizing, the only term left to minimize is the endpoint divergence $\text{KL}(\mathbb{P}_{0,T}\|\Pi_{0,T})$, which is minimized \textit{uniquely} when:
\begin{align}
    \Pi^\star_{0,T}=\mathbb{P}_{0,T}
\end{align}
which follows the \textbf{strict convexity of the KL divergence}. Given that $\Pi^\star\in \mathcal{R}(\mathbb{Q})$ and $\Pi^\star_{0,T}=\mathbb{P}_{0,T}$, we can write:
\begin{align}
    \Pi^\star(\cdot)=\int_{\mathbb{R}^d\times \mathbb{R}^d}\mathbb{Q}_{\cdot|0,T}(\cdot|\boldsymbol{x}_0, \boldsymbol{x}_T)\mathrm{d}\mathbb{P}_{0,T}(\boldsymbol{x}_0, \boldsymbol{x}_T)
\end{align}
which can equivalently be written as $\Pi^\star=\mathbb{P}_{0,T}\mathbb{Q}_{\cdot|0,T}$ and is exactly our definition of the (\ref{eq:reciprocal-projection}). \hfill $\square$

Now that we have shown that any path measure can be projected onto the reciprocal class $\mathcal{R}(\mathbb{Q})$ of a bridge measure $\mathbb{Q}$ such that the bridge is preserved, we will show that there exists a \textbf{unique Markov measure in the reciprocal class} $\mathcal{R}(\mathbb{Q})$ which is also the solution to the (\ref{def:dynamic-sb-problem}). 

\begin{proposition}[Solution to Dynamic Schrödinger Bridge]
The \textbf{Markov measure in the reciprocal class} of $\mathbb{Q}$, i.e., $\mathbb{M}\in \mathcal{R}(\mathbb{Q})$, that satisfies $\mathbb{M}_0=\pi_0$ and $\mathbb{M}_T=\pi_T$ is the unique solution to the Schrödinger bridge $\mathbb{M}=\mathbb{P}^\star$.
\end{proposition}

\textit{Proof.} By definition of the reciprocal class $\mathcal{R}(\mathbb{Q})$, $\mathbb{M}$ and $\mathbb{Q}$ and share the exact same bridge such that $\mathbb{M}_{t|0,T}=\mathbb{Q}_{t|0,T}$ for all $t\in [0,T]$. Thus, the Radon-Nikodym derivative $\frac{\mathrm{d}\mathbb{M}}{\mathrm{d}\mathbb{Q}}(\boldsymbol{X}_{0:T})$ depends only on the endpoints $(\boldsymbol{X}_0, \boldsymbol{X}_T)$ and can be written as:
\begin{align}
    \frac{\mathrm{d}\mathbb{M}}{\mathrm{d}\mathbb{Q}}(\boldsymbol{X}_{0:T})=\xi(\boldsymbol{X}_0, \boldsymbol{X}_T)
\end{align}
where $\xi:\mathbb{R}^d\times\mathbb{R}^d\to \mathbb{R}$ is a measurable function. Since we define $\mathbb{P}$ to be Markov and the reference measure $\mathbb{Q}$ is Markov by construction, conditioning on some intermediate state $\boldsymbol{X}_t$ yields the factorization:
\begin{small}
\begin{align}
    \frac{\mathrm{d}\mathbb{M}(\boldsymbol{X}_{0:T}|\boldsymbol{X}_t)}{\mathrm{d}\mathbb{Q}(\boldsymbol{X}_{0:T}|\boldsymbol{X}_t)}=\frac{\mathrm{d}\mathbb{M}(\boldsymbol{X}_{0:t}, \boldsymbol{X}_{t:T}|\boldsymbol{X}_t)}{\mathrm{d}\mathbb{Q}(\boldsymbol{X}_{0:t}, \boldsymbol{X}_{t:T}|\boldsymbol{X}_t)}=\frac{\mathrm{d}\mathbb{M}(\boldsymbol{X}_{0:t}|\boldsymbol{X}_t)}{\mathrm{d}\mathbb{Q}(\boldsymbol{X}_{0:t}|\boldsymbol{X}_t)}\frac{\mathrm{d}\mathbb{M}(\boldsymbol{X}_{t:T}|\boldsymbol{X}_t)}{\mathrm{d}\mathbb{Q}(\boldsymbol{X}_{t:T}|\boldsymbol{X}_t)}
\end{align}
\end{small}
where the past and future given $\boldsymbol{X}_t$ are \textit{independent}. Given that the RND depends only on the endpoints via the measurable function $\xi(\boldsymbol{X}_0, \boldsymbol{X}_T)$, there exist two measurable functions $a, b:\mathbb{R}^d\to \mathbb{R}$ that depend only on each endpoint, respectively, such that:
\begin{small}
\begin{align}
    \frac{\mathrm{d}\mathbb{M}}{\mathrm{d}\mathbb{Q}}(\boldsymbol{X}_{0:T})=\xi(\boldsymbol{X}_0, \boldsymbol{X}_T)=a(\boldsymbol{X}_0)b(\boldsymbol{X}_T)
\end{align}
\end{small}
Recall from Section \ref{sec:static-sb} that the optimal endpoint law $\pi^\star_{0,T}$ that solves the static SB problem factorizes into Schrödinger potentials $(\varphi, \hat\varphi)$ that are \textit{unique} up to a constant. Therefore, we conclude:
\begin{small}
\begin{align}
    \frac{\mathrm{d}\mathbb{M}}{d\mathbb{Q}}(\boldsymbol{X}_{0:T})=\frac{d\pi^\star_{0,T}}{dq}(\boldsymbol{X}_0, \boldsymbol{X}_T)=e^{\varphi(\boldsymbol{X}_0)}e^{\hat\varphi(\boldsymbol{X}_T)}
\end{align}
\end{small}
and we have shown that the Markov measure in the reciprocal class $\mathbb{M}\in\mathcal{R}(\mathbb{Q})$ is \textit{unique} and equals the Schrödinger bridge $\mathbb{M}=\mathbb{P}^\star$. \hfill $\square$

Building on the ideas of Markovian and reciprocal projections, we can define the \textbf{Iterative Markovian Fitting} (IMF) scheme \citep{shi2023diffusion} which constructs the optimal SB via alternating Markovian and reciprocal projections of an SDE which becomes the foundation of the practical DSBM algorithm which fits a parameterized velocity field by matching the control drift.

\begin{figure}
    \centering
    \includegraphics[width=\linewidth]{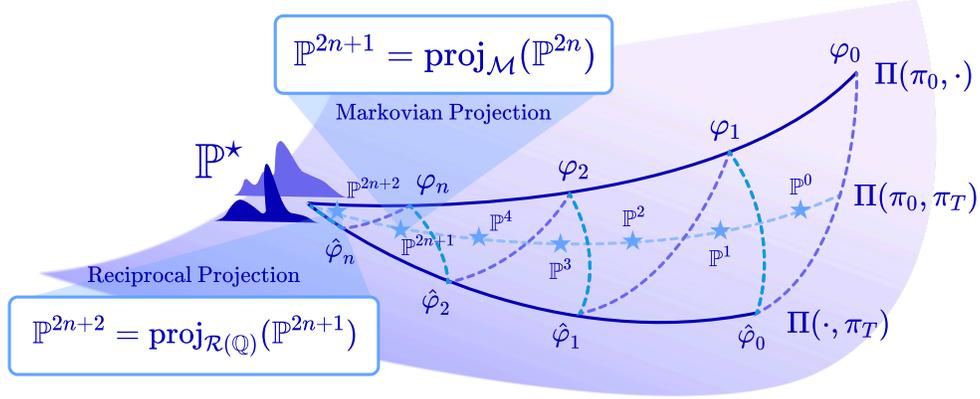}
    \caption{\textbf{Iterative Markovian Fitting (IMF) vs. Iterative Proportional Fitting (IPF).} While IPF alternates projections onto the marginal constraint sets $\Pi(\pi_0, \cdot)$ and $\Pi(\cdot , \pi_T)$, intermediate iterations generally do not preserve the endpoint coupling. In contrast, IMF performs Markovian and reciprocal projections that maintain both marginal constraints $\mathbb{P}^n\in \Pi(\pi_0, \pi_T)$ at every iteration, progressively refining the path measure until convergence to the optimal bridge measure $\mathbb{P}^\star$.}
    \label{fig:imf}
\end{figure}

\boldtext{Iterative Markovian Fitting} (IMF) generates a sequence of path measures $(\mathbb{P}^n)_{n\in \mathbb{N}}$ with \textbf{alternating Markovian projections and reciprocal projections} defined as:
\begin{align}
    \mathbb{P}^{2n+1}=\text{proj}_{\mathcal{M}}(\mathbb{P}^{2n}), \quad \mathbb{P}^{2n+2}=\text{proj}_{\mathcal{R}(\mathbb{Q})}(\mathbb{P}^{2n+1})
\end{align}
where the first path measure in the sequence $\mathbb{P}^0$ is the path measure in the reciprocal class of the reference measure $\mathbb{P}^0\in \mathcal{R}(\mathbb{Q})$ that satisfies the marginal constraints $\mathbb{P}^0_0=\pi_0$ and $\mathbb{P}^0_T=\pi_T$. The IMF procedure is grounded in three \textbf{key theoretical results} which are summarized as follows:

\begin{enumerate}
    \item [(i)] \textbf{Pythagorean Identity (Lemma \ref{lemma:imf-pythagorean-thm}):} The KL divergence between a mixture of bridges $\Pi\in \mathcal{R}(\mathbb{Q})$ in the recprocal class of $\mathbb{Q}$ and an arbitrary Markov measure $\mathbb{M}\in \mathcal{M}$ satisfies the Pythagorean identity:
    \begin{align}
        \text{KL}(\Pi\|\mathbb{M})&=\underbrace{\text{KL}(\Pi|\bluetext{\text{proj}_{\mathcal{M}}(\Pi)})}_{\text{distance to Markovian projection}}+\underbrace{\text{KL}(\bluetext{\text{proj}_{\mathcal{M}}(\Pi)}\|\mathbb{M})}_{\text{remaining distance to target}}
    \end{align}
    Similarly, the KL divergence between a path measure $\mathbb{P}$ and its reciprocal projection $\text{proj}_{\mathcal{R}(\mathbb{Q})}(\mathbb{P})$ satisfies the Pythagorean identity defined as: 
    \begin{align}
        \text{KL}(\mathbb{P}\|\Pi)&=\underbrace{\text{KL}(\mathbb{P}\|\bluetext{\text{proj}_{\mathcal{R}(\mathbb{Q})}}(\mathbb{P}))}_{\text{distance to reciprocal projection}}+\underbrace{\text{KL}(\bluetext{\text{proj}_{\mathcal{R}(\mathbb{Q})}(\mathbb{P})}\|\Pi)}_{\text{remaining distance to target}}
    \end{align}
    \item[(ii)] \textbf{Monotone Improvement (Proposition \ref{prop:imf-monotone-decrease}):} The distance to the optimal SB measure $\mathbb{P}^\star$ is \textbf{strictly decreasing} with each iteration of IMF, such that:
    \begin{align}
        \text{KL}\left(\mathbb{P}^{n+1}\|\mathbb{P}^\star\right)\leq \text{KL}\left(\mathbb{P}^n\|\mathbb{P}^\star\right), \quad \lim_{n\to +\infty}\text{KL}\left(\mathbb{P}^{n}\|\mathbb{P}^{n+1}\right)=0
    \end{align}
    \item[(iii)] \textbf{Convergence Guarantee (Theroem \ref{thm:imf-convergence}):} The sequence of path measures generated via IMF $(\mathbb{P}^n)_{n\in \mathbb{N}}$ converges to a unique fixed point $\mathbb{P}^\star$ that is exactly the SB measure $\mathbb{P}^\star$:
    \begin{align}
        \lim_{n\to +\infty}\text{KL}\left(\mathbb{P}^{n}\|\mathbb{P}^\star\right)=0 \quad \text{where} \quad \mathbb{P}^\star=\mathbb{P}^\star
    \end{align}
\end{enumerate}

To understand why each of these statements holds and build our intuition on the structure of Markovian and reciprocal projections, we will prove each of them and conclude with the convergence of the IMF procedure.

\begin{lemma}[Pythagorean Theorem of Markovian and Reciprocal Projections (Lemma 6 in \citet{shi2023diffusion})]\label{lemma:imf-pythagorean-thm}
    Let $\Pi\in \mathcal{R}(\mathbb{Q})$ be a bridge measure in the reciprocal class of $\mathbb{Q}$ and $\text{proj}_{\mathcal{M}}(\Pi)$ be the Markovian projection of $\Pi$. Given some arbitrary Markov measure that has finite KL divergence with $\Pi$ and $\text{proj}_{\mathcal{M}}(\Pi)$, the following identity holds:
    \begin{small}
    \begin{align}
        \text{KL}(\Pi\|\mathbb{M})=\text{KL}(\Pi\|\text{proj}_{\mathcal{M}}(\Pi))+\text{KL}(\text{proj}_{\mathcal{M}}(\Pi)\|\mathbb{M})\tag{Markovian Projection Identity}\label{eq:markovian-proj-identity}
    \end{align}
    \end{small}
    which means that the KL divergence between $\Pi$ and $\mathbb{M}$ is the sum of the KL incurred from projecting $\Pi$ to the Markov class and the KL incurred from projecting $\text{proj}_{\mathcal{M}}(\Pi)$ to the Markov measure $\mathbb{M}$. Similarly, for any arbitrary path measure $\mathbb{P}\in \mathcal{P}(C([0,T];\mathbb{R}^d))$ projected onto the reciprocal class $\text{proj}_{\mathcal{R}(\mathbb{Q})}(\mathbb{P})$, the KL divergence with $\Pi\in \mathcal{R}(\mathbb{Q})$ can be decomposed as:
    \begin{small}
    \begin{align}
        \text{KL}(\mathbb{P}\|\Pi)=\text{KL}(\mathbb{P}\|\text{proj}_{\mathcal{R}(\mathbb{Q})}(\mathbb{P}))+\text{KL}(\text{proj}_{\mathcal{R}(\mathbb{Q})}(\mathbb{P})\|\Pi)\tag{Reciprocal Projection Identity}\label{eq:reciprocal-proj-identity}
    \end{align}
    \end{small}
\end{lemma}

\textit{Proof.} \textbf{Step 1: Derive the Markovian Projection Identity.}
Intuitively, this step aims to derive the KL divergence to the Markov measure, which can be interpreted as the \textbf{information lost} when going from conditioning on $(\boldsymbol{X}_0, \boldsymbol{X}_t)$ in $\Pi$ to conditioning only on $\boldsymbol{X}_t$ in $\mathbb{M}$. To break this down, we define the \textbf{key component} that appears in both the drift of the Markovian projection and the mixture of bridges $\Pi$ as:
\begin{align}
    \boldsymbol{G}_t:=\nabla \log \mathbb{Q}_{T|t}(\boldsymbol{X}_T|\boldsymbol{X}_t)
\end{align}
Then, rewriting (\ref{eq:markov-proj-optimal-drif}) and the drift of the reciprocal process with additional conditioning on $\boldsymbol{X}_0$ in terms of $\boldsymbol{G}_t$, we have:
\begin{small}
\begin{align}
    \underbrace{\boldsymbol{A}_t:= \mathbb{E}_{\Pi}\left[\boldsymbol{G}_t|\boldsymbol{X}_0,\boldsymbol{X}_t\right]}_{\text{drift of reciprocal measure }\Pi}, \quad \underbrace{\boldsymbol{B}_t:= \mathbb{E}_{\Pi}\left[\boldsymbol{G}_t|\boldsymbol{X}_t\right]}_{\text{drift of Markovian measure }\mathbb{M}}
\end{align}
\end{small}
Using these definitions, we can rewrite the Markovian projection identity in terms of the integral differences between drifts as derived in Section \ref{subsec:path-measure-rnd-kl}. First, recall the identity:
\begin{align}
    \underbrace{\text{KL}(\Pi\|\mathbb{M})}_{(\bigstar)}=\underbrace{\text{KL}(\Pi\|\text{proj}_{\mathcal{M}}(\Pi))}_{(\diamond)}+\underbrace{\text{KL}(\text{proj}_{\mathcal{M}}(\Pi)\|\mathbb{M})}_{(\blacklozenge)}\label{eq:pythag-proof6}
\end{align}
For $(\bigstar)$ defined as the KL divergence between $\Pi\in \mathcal{R}(\mathbb{Q})$ with control drift $\sigma_t\boldsymbol{A}_t$ and the Markov measure $\mathbb{M}$ with control drift $\boldsymbol{u}$, we can write as the integral difference:
\begin{small}
\begin{align}
    \text{KL}(\Pi\|\mathbb{M})&=\frac{1}{2}\int_0^T\mathbb{E}_{\Pi_{0,t}}\left[\|\boldsymbol{u}(\boldsymbol{X}_t,t)-\sigma_t\boldsymbol{A}_t\|^2\right]dt
\end{align}
\end{small}
For $(\diamond)$, the Markovian projection $\text{proj}_{\mathcal{M}}(\Pi)$ has control drift $\sigma_t\boldsymbol{B}_t$ which removes conditioning on $\boldsymbol{X}_0$ and yields following KL divergence with $\Pi$:
\begin{small}
\begin{align}
    \text{KL}(\Pi\|\text{proj}_{\mathcal{M}}(\Pi))&=\frac{1}{2}\int_0^T\mathbb{E}_{\Pi_{0,t}}\left[\|\sigma_t\boldsymbol{A}_t-\sigma_t\boldsymbol{B}_t\|^2\right]dt
\end{align}
\end{small}
Finally, for $(\blacklozenge)$, the KL divergence between the Markovian projection $\text{proj}_{\mathcal{M}}(\Pi)$ has control drift $\sigma_t\boldsymbol{B}_t$ and the Markov measure $\mathbb{M}$ with drift $\boldsymbol{u}(\boldsymbol{X}_t,t)$ is given by:
\begin{small}
\begin{align}
    \text{KL}(\text{proj}_{\mathcal{M}}(\Pi)\|\mathbb{M})&=\frac{1}{2}\int_0^T\mathbb{E}_{\Pi_{0,t}}\left[\|\boldsymbol{u}(\boldsymbol{X}_t,t)-\sigma_t\boldsymbol{B}_t\|^2\right]dt
\end{align}
\end{small}
Now, summing $(\diamond)$ and $(\blacklozenge)$ from the left-hand side of (\ref{eq:pythag-proof6}) and observing that projecting onto the subspace spanned by $\boldsymbol{B}_t$ gives $\langle\boldsymbol{A}_t,\boldsymbol{B}_t\rangle =\|\boldsymbol{B}_t\|^2$ and $\boldsymbol{B}_t=\mathbb{E}_{\Pi}[\boldsymbol{A}_t|\boldsymbol{X}_t]$, we have:
\begin{small}
\begin{align}
    &\underbrace{\text{KL}(\Pi\|\text{proj}_{\mathcal{M}}(\Pi))}_{(\diamond)}+\underbrace{\text{KL}(\text{proj}_{\mathcal{M}}(\Pi)\|\mathbb{M})}_{(\blacklozenge)}=\frac{1}{2}\int_0^T\mathbb{E}_{\Pi_{0,t}}\left[\|\sigma_t\boldsymbol{A}_t-\sigma_t\boldsymbol{B}_t\|^2\right]dt+\frac{1}{2}\int_0^T\mathbb{E}_{\Pi_{0,t}}\left[\|\boldsymbol{u}(\boldsymbol{X}_t,t)-\sigma_t\boldsymbol{B}_t\|^2\right]dt\nonumber\\
    &=\frac{1}{2}\int \mathbb{E}_{\Pi_{0,t}}\bigg[\sigma_t^2\|\boldsymbol{A}_t\|^2-\underbrace{2\sigma_t^2\langle \boldsymbol{A}_t,\boldsymbol{B}_t\rangle}_{=2\sigma_t^2\|\boldsymbol{B}_t\|^2}+\sigma_t^2\|\boldsymbol{B}_t\|^2+\|\boldsymbol{u}(\boldsymbol{X}_t,t)\|^2-2\sigma_t\langle \boldsymbol{u}(\boldsymbol{X}_t,t),\boldsymbol{B}_t\rangle +\sigma_t^2\|\boldsymbol{B}_t\|^2\bigg]\nonumber\\
    &=\frac{1}{2}\int \mathbb{E}_{\Pi_{0,t}}\bigg[\sigma_t^2\|\boldsymbol{A}_t\|^2\bluetext{-2\sigma_t^2\|\boldsymbol{B}_t\|^2}+\bluetext{\sigma_t^2\|\boldsymbol{B}_t\|^2}+\|\boldsymbol{u}(\boldsymbol{X}_t,t)\|^2-2\sigma_t\langle \boldsymbol{u}(\boldsymbol{X}_t,t),\boldsymbol{B}_t\rangle +\bluetext{\sigma_t^2\|\boldsymbol{B}_t\|^2}\bigg]\nonumber\\
    &=\frac{1}{2}\int \mathbb{E}_{\Pi_{0,t}}\bigg[\sigma_t^2\|\boldsymbol{A}_t\|^2+\|\boldsymbol{u}(\boldsymbol{X}_t,t)\|^2-2\sigma_t\bluetext{\underbrace{\langle \boldsymbol{u}(\boldsymbol{X}_t,t),\boldsymbol{B}_t\rangle}_{=\langle \boldsymbol{u}(\boldsymbol{X}_t,t),\boldsymbol{A}_t\rangle }}\bigg]\nonumber\\
    &=\frac{1}{2}\int \mathbb{E}_{\Pi_{0,t}}\bigg[\sigma_t^2\|\boldsymbol{A}_t\|^2+\|\boldsymbol{u}(\boldsymbol{X}_t,t)\|^2-2\sigma_t\langle \boldsymbol{u}(\boldsymbol{X}_t,t),\boldsymbol{A}_t\rangle\bigg]\tag{given $\mathbb{E}_{\Pi_t}[\langle \boldsymbol{u},\boldsymbol{B}_t\rangle]=\mathbb{E}_{\Pi_{0,t}}[\langle \boldsymbol{u},\boldsymbol{A}_t\rangle]$}\\
    &=\frac{1}{2}\int_0^T\mathbb{E}_{\Pi_{0,t}}\left[\|\boldsymbol{u}(\boldsymbol{X}_t,t)-\sigma_t\boldsymbol{A}_t\|^2\right]dt=\boxed{\text{KL}(\Pi\|\mathbb{M})}
\end{align}
\end{small}
where the final equality exactly yields the KL divergence between $\text{KL}(\Pi\|\mathbb{M})$, which proves (\ref{eq:markovian-proj-identity}). Next, we prove the (\ref{eq:reciprocal-proj-identity}) with fewer steps. 

\textbf{Step 2: Derive the Reciprocal Projection Identity. }
Let $\Pi^\star= \text{proj}_{\mathcal{R}(\mathbb{Q})}(\mathbb{P})$ denote the reciprocal projection of $\mathbb{P}$. By the definition of reciprocal projection, $\Pi^\star$ has the same endpoint law as $\mathbb{P}$ and the same conditional bridge as $\mathbb{Q}$, defined as:
\begin{small}
\begin{align}
    \Pi^\star=\mathbb{Q}_{\cdot|0,T}\mathbb{P}_{0,T}
\end{align}
\end{small}
Applying the (\ref{eq:chain-rule-kl}), we have:
\begin{small}
\begin{align}
    \text{KL}(\mathbb{P}\|\Pi)&=\mathbb{E}_{\mathbb{P}}\left[\log \frac{\mathrm{d}\mathbb{P}}{\mathrm{d}\Pi}\right]=\mathbb{E}_{\mathbb{P}}\left[\log \frac{\mathrm{d}\mathbb{P}}{\mathrm{d}\Pi^\star}+\log \frac{\mathrm{d}\Pi^\star}{\mathrm{d}\Pi}\right]\nonumber\\
    &=\mathbb{E}_{\mathbb{P}}\left[\log \frac{\mathrm{d}\mathbb{P}}{\mathrm{d}\Pi^\star}\right]+\mathbb{E}_{\mathbb{P}}\left[\log \frac{\mathrm{d}\Pi^\star}{\mathrm{d}\Pi}\right]=\text{KL}(\mathbb{P}\|\Pi^\star)+ \int \log \frac{\mathrm{d}\Pi^\star}{\mathrm{d}\Pi}\mathrm{d}\mathbb{P}\label{eq:pythag-proof1}
\end{align}
\end{small}
Now, we leverage the key property of measures in the same reciprocal class $\mathcal{R}(\mathbb{Q})$ that share the same \textit{conditional bridge} as $\mathbb{Q}$. Therefore, the log RND depends only on the endpoints $(\boldsymbol{X}_0, \boldsymbol{X}_T)$. In addition, the original path measure $\mathbb{P}$ and its reciprocal projection $\Pi^\star$ share the same \textit{endpoint law}, so we can write (\ref{eq:pythag-proof1}) as:
\begin{small}
\begin{align}
    \text{KL}(\mathbb{P}\|\Pi)&=\text{KL}(\mathbb{P}\|\Pi^\star)+ \int \underbrace{\log \frac{\mathrm{d}\Pi^\star}{\mathrm{d}\Pi}(\boldsymbol{X}_{0:T})}_{\text{depends on $(\boldsymbol{X}_0, \boldsymbol{X}_T)$ only}}\mathrm{d}\mathbb{P}(\boldsymbol{X}_{0:T})=\text{KL}(\mathbb{P}\|\Pi^\star)+ \int \log \frac{\mathrm{d}\Pi^\star_{0,T}}{\mathrm{d}\Pi_{0,T}}(\boldsymbol{X}_0, \boldsymbol{X}_T)\bluetext{\underbrace{\mathrm{d}\mathbb{P}(\boldsymbol{X}_0, \boldsymbol{X}_T)}_{=\mathrm{d}\Pi^\star_{0,T}}} \nonumber\\
    &=\text{KL}(\mathbb{P}\|\Pi^\star)+ \underbrace{\int \log \frac{\mathrm{d}\Pi^\star_{0,T}}{\mathrm{d}\Pi_{0,T}}(\boldsymbol{X}_0, \boldsymbol{X}_T)\mathrm{d}\Pi^\star_{0,T}(\boldsymbol{X}_0,\boldsymbol{X}_T)}_{=\text{KL}(\Pi^\star\|\Pi)}=\text{KL}(\mathbb{P}\|\Pi^\star)+\text{KL}(\Pi^\star\|\Pi)
\end{align}
\end{small}
which recovers the (\ref{eq:reciprocal-proj-identity}) by subsituting back $\Pi^\star =\text{proj}_{\mathcal{R}(\mathbb{Q})}(\mathbb{P})$.\hfill $\square$

Using Lemma \ref{lemma:imf-pythagorean-thm}, we can prove that iterating between Markov and reciprocal projections yields a monotonically decreasing KL divergence to the Schrödinger bridge measure, which will be the foundation for our final convergence proof.

\begin{proposition}[Monotonically Decreasing KL Divergence (Proposition 7 in \citet{shi2023diffusion})]\label{prop:imf-monotone-decrease}
    Given a sequence of Markov and reciprocal projections $(\mathbb{P}^n)_{n \in \mathbb{N}}$ and the Schrödinger bridge path measure $\mathbb{P}^\star$, the reverse KL divergences between $\mathbb{P}^n$ and $\mathbb{P}^\star$ decreases monotonically:
    \begin{small}
    \begin{align}
        \forall n \in \mathbb{N}, \quad \text{KL}(\mathbb{P}^{n+1}\|\mathbb{P}^\star)\leq \text{KL}(\mathbb{P}^{n}\|\mathbb{P}^\star)\leq \infty
    \end{align}
    \end{small}
    and in the limit $n\to \infty$, the KL divergence between subsequent projections converges to zero:
    \begin{align}
        \lim_{n \to \infty}\text{KL}(\mathbb{P}^{n}\|\mathbb{P}^{n+1})= 0
    \end{align}
\end{proposition}

\textit{Proof.} Recall that the Schrödinger bridge is the unique path measure that is Markov $\mathbb{P}^\star\in \mathcal{M}$ and in the reciprocal class $\mathbb{P}^\star\in \mathcal{R}(\mathbb{Q})$. Therefore, both (\ref{eq:markovian-proj-identity}) and (\ref{eq:reciprocal-proj-identity}) from Lemma \ref{lemma:imf-pythagorean-thm} hold with respect to $\mathbb{P}^\star$. This means that for all $\mathbb{P}^n$, the following identity holds:
\begin{small}
\begin{align}
    \text{KL}(\mathbb{P}^n\|\mathbb{P}^\star)=\underbrace{\text{KL}(\mathbb{P}^{n}\|\mathbb{P}^{n+1})}_{\geq 0}+ \text{KL}(\mathbb{P}^{n+1}\|\mathbb{P}^\star)\implies \boxed{\text{KL}(\mathbb{P}^n\|\mathbb{P}^\star)\geq \text{KL}(\mathbb{P}^{n+1}\|\mathbb{P}^\star)}\label{eq:pythag-proof2}
\end{align}
\end{small}
where $\mathbb{P}^{n+1}$ is either the \textbf{Markovian projection of $\mathbb{P}^n$} if $\mathbb{P}^n$ was generated from a reciprocal projection or the \textbf{reciprocal projection of $\mathbb{P}^{n+1}$} if $\mathbb{P}^n$ was generated from a Markovian projection. This proves the first part of the proposition. 

\textbf{Next, we show that at the limit as $n\to \infty$, the KL between iterations converges to zero}. To do this, we can compute the accumulated KL divergence over $N$ iterations using the same identity from (\ref{eq:pythag-proof2}). To do this, we apply the identity for each iteration $n=0, \dots, N$ to get:
\begin{align}
    n=0&:\text{KL}(\mathbb{P}^0\|\mathbb{P}^\star)=\text{KL}(\mathbb{P}^{0}\|\mathbb{P}^{1})+ \bluetext{\text{KL}(\mathbb{P}^{1}\|\mathbb{P}^\star)}\nonumber\\
    n=1&:\bluetext{\text{KL}(\mathbb{P}^1\|\mathbb{P}^\star)}=\text{KL}(\mathbb{P}^{1}\|\mathbb{P}^{2})+ \bluetext{\text{KL}(\mathbb{P}^{2}\|\mathbb{P}^\star)}\nonumber\\
    \vdots\nonumber\\
    n=N&:\bluetext{\text{KL}(\mathbb{P}^N\|\mathbb{P}^\star)}=\text{KL}(\mathbb{P}^{N}\|\mathbb{P}^{N+1})+ \text{KL}(\mathbb{P}^{N+1}\|\mathbb{P}^\star)
\end{align}
Observing that the intermediate KL divergences with $\mathbb{P}^\star$ (indicated in \bluetext{blue}) cancel after summing over $n=0, \dots , N$ (telescoping identity), we can write:
\begin{small}
\begin{align}
    \underbrace{\text{KL}(\mathbb{P}^0\|\mathbb{P}^\star)}_{\text{fixed }\leq \infty}=\sum_{n=0}^N\text{KL}(\mathbb{P}^{n}\|\mathbb{P}^{n+1})+ \text{KL}(\mathbb{P}^{N+1}\|\mathbb{P}^\star)\implies \sum_{n=0}^N\text{KL}(\mathbb{P}^{n}\|\mathbb{P}^{n+1})\leq \infty\label{eq:pythag-proof3}
\end{align}
\end{small}
where we observe that the KL divergence between the initial path measure $\mathbb{P}^0$ and $\mathbb{P}^\star$ is fixed and bounded. Since (\ref{eq:pythag-proof3}) holds as we increase $N$ to infinity, we have shown that the series of non-negative KL divergences is bounded $\text{KL}(\mathbb{P}^{n}\|\mathbb{P}^{n+1})$ is finite and bounded, the additive terms must converge to zero:
\begin{align}
   \forall n\in \mathbb{N}, \quad \sum_{n=0}^N\text{KL}(\mathbb{P}^{n}\|\mathbb{P}^{n+1})\leq \infty\implies \boxed{\lim_{n \to \infty}\text{KL}(\mathbb{P}^{n}\|\mathbb{P}^{n+1})= 0}
\end{align}
and we conclude our proof. \hfill $\square$

This leads to the final result, which ensures that an IMF sequence converges to the optimal SB path measure $\mathbb{P}^\star$.

\begin{proposition}[Iterative Markovian Fitting Converges to the Unique Schrödinger Bridge]\label{thm:imf-convergence}
    The sequence of path measures $(\mathbb{P}^n)_{n\in \mathbb{N}}$ generated from alternating Markovian and reciprocal projections of the IMF algorithm has a unique fixed point $\mathbb{P}^\star$ which equals the Schrödinger bridge. Furthermore, in the limit $n\to \infty$, the KL divergence converges to the fixed point:
    \begin{align}
        \lim_{n \to \infty}\text{KL}(\mathbb{P}^n\|\mathbb{P}^\star) =  0
    \end{align}
\end{proposition}

\textit{Proof Sketch.} By Proposition \ref{prop:imf-monotone-decrease}, we know that each path measure in the sequence $(\mathbb{P}^{n})_{n \in \mathbb{N}}$ decreases the KL divergence to $\mathbb{P}^\star$, so the sequence remains trapped in a compact region in path space that is \textbf{bounded below by zero}, since KL divergence is non-negative. Therefore, both the sequence of Markovian projections and reciprocal projections converge to their optimal fixed points $\mathbb{M}^\star\in \mathcal{M}$ and $\Pi^\star\in \mathcal{R}(\mathbb{Q})$. From Proposition \ref{prop:imf-monotone-decrease}, we also have that the the KL divergence between each iteration converges to zero, which implies that the fixed points of the Markovian and reciprocal projections coincide exactly\footnote{Note that $\mathbb{M}^\star$ and $\Pi^\star$ are used to denote the Markovian and reciprocal projections for any path measure, but they only coincide at the unique fixed point when they are equal to the Schrödinger bridge.}:
\begin{align}
    \lim_{n \to \infty}\text{KL}(\mathbb{P}^{n}\|\mathbb{P}^{n+1})= 0\implies \text{KL}(\mathbb{M}^\star\|\Pi) = 0\implies \boxed{\mathbb{M}^\star=\Pi^\star=\mathbb{P}^\star}
\end{align}
which means that the shared limit of the Markov and reciprocal projections is both Markov and in the reciprocal class $\mathcal{R}(\mathbb{Q})$, and therefore, must be the Schrödinger bridge $\mathbb{P}^\star$. Given that both subsequences converge to $\mathbb{P}^\star$, we have that the full sequence $(\mathbb{P}^{n})_{n \in \mathbb{N}}$ also converges to $\mathbb{P}^\star$ and the KL divergence converges to zero.\footnote{for more rigorous proof, see \citep{shi2023diffusion}} \hfill $\square$ 

The relationship between Markov and reciprocal projections reveals a \textbf{key structural property} of the Schrödinger bridge. The (\ref{eq:markovian-projection}) enforces the Markov property by selecting the closest Markov process in relative entropy to some bridge measure, and the (\ref{eq:reciprocal-projection}) adjusts the path measure so that the endpoint marginals match the prescribed distributions, while preserving the bridge structure inherited from the reference process. 

Crucially, the Schrödinger bridge $\mathbb{P}^\star$ lies exactly at the \textbf{equilibrium} of these two constraints, as it is the unique path measure that simultaneously satisfies the endpoint conditions and remains the closest Markov measure to the reference bridge dynamics. We further show that alternating between performing Markovian and reciprocal projections yields a unique fixed point that coincides with $\mathbb{P}^\star$, which is exactly what defines the Iterative Markovian Fitting (IMF) procedure. We will revisit the IMF procedure in Section \ref{subsec:diffusionsbm}, where we apply this procedure in the context of generative modeling. 

\subsection{Stochastic Interpolants to Schrödinger Bridges}
\label{subsec:stochastic-interpolants}

The \boldtext{stochastic interpolants} framework \citep{albergo2025stochastic} can be used to construct the Schrödinger bridge solution. First, we will provide some background on the framework, which will naturally lead to its extension in solving the SB problem. 

\begin{definition}[Stochastic Interpolant \citep{albergo2025stochastic}]\label{def:stochastic-interpolants}
    Let $\pi_0, \pi_T\in \mathcal{P}(\mathbb{R}^d)$ be two probability densities on the state space. The \textbf{stochastic interpolant} between $\pi_0$ and $\pi_T$ is a stochastic process $\boldsymbol{X}_{0:T}$ of the form:
    \begin{align}
        \boldsymbol{x}_t=I(\boldsymbol{x}_0, \boldsymbol{x}_T,t)+\gamma(t)\boldsymbol{z}, \quad t\in [0,T]
    \end{align}
    where the following are satisfied:
    \begin{enumerate}
        \item [(i)] The map $I\in C^2((C^2(\mathbb{R}^d\times \mathbb{R}^d))^d, [0,T])$ has boundary conditions $I(\boldsymbol{x}_0, \boldsymbol{x}_T,0)=\boldsymbol{x}_0$ and $I(\boldsymbol{x}_0, \boldsymbol{x}_T,T)=\boldsymbol{x}_T$ and controlled time variation:
        \begin{small}
        \begin{align}
            \exists C_1< \infty\quad \text{s.t.}\quad|\partial_tI(\boldsymbol{x}_0, \boldsymbol{x}_T,t)|\leq C_1|\boldsymbol{x}_T-\boldsymbol{x}_0|, \quad \forall (\boldsymbol{x}_0, \boldsymbol{x}_T,t)\in[0,T]\times\mathbb{R}^d\times \mathbb{R}^d
        \end{align}
        \end{small}
        \item[(ii)] The scalar noise function $\gamma: [0,T]\to \mathbb{R}$ satisfies $\gamma(0)=\gamma(1)=0$ and $\gamma(t)> 0$ for all $t\in (0,T)$. In addition, $\gamma^2\in C^2([0,T])$.
        \item[(iii)] The pair $(\boldsymbol{x}_0, \boldsymbol{x}_T)$ are sampled from a probability measure $\pi_{0,T}$ whose marginals are $\pi_0$ and $\pi_T$ defined as:
        \begin{align}
            \pi_{0,T}(d\boldsymbol{x}_0, \mathbb{R}^d)=\pi_0(\boldsymbol{x}_0)d\boldsymbol{x}_0, \quad\pi_{0,T}(\mathbb{R}^d, d\boldsymbol{x}_T)=\pi_T(\boldsymbol{x}_T)d\boldsymbol{x}_T
        \end{align}
        \item[(iv)] The Gaussian random variable $\boldsymbol{z}\sim \mathcal{N}(\boldsymbol{0}, \boldsymbol{I}_d)$ is independent of $(\boldsymbol{x}_0, \boldsymbol{x}_T)$.
    \end{enumerate}
\end{definition}

Intuitively, the stochastic interpolant is a general method of connecting samples from two marginal distributions with a \textbf{deterministic path} $I(\boldsymbol{x}_0, \boldsymbol{x}_T,t)$ that is perturbed by a time-dependent Gaussian diffusion $\gamma(t)\boldsymbol{z}$ along the interior of the time interval. By definition, the noise $\gamma(t)\boldsymbol{z}$ vanishes at $t=0$ and $t=T$, ensuring that the process matches exactly the terminal marginals $\pi_0$ and $\pi_T$.

\begin{theorem}[Properties of Stochastic Interpolants (Theorem 2.6 in \citet{albergo2025stochastic})]
    The stochastic interpolant $\boldsymbol{x}_t=I(\boldsymbol{x}_0, \boldsymbol{x}_T,t)$ satisfies the following properties:
    \begin{align}
        \partial_tp_t+\nabla\cdot(p_t\boldsymbol{v})=0\label{eq:si-continuity}
    \end{align}
    where the velocity is defined as the expectation of the time derivative:
    \begin{align}
        \boldsymbol{v}(\boldsymbol{x},t)=\mathbb{E}_{p_t}[\dot{\boldsymbol{x}}_t|\boldsymbol{x}_t=\boldsymbol{x}]=\mathbb{E}[\partial_tI(\boldsymbol{x}_0, \boldsymbol{x}_T,t)+\dot{\gamma}(t)\boldsymbol{z}|\boldsymbol{x}_t=\boldsymbol{x}]
    \end{align}
    which is bounded on the domain of the density function $p_t(\boldsymbol{x})$ given by:
    \begin{align}
        \forall t\in [0,T]:\int_{\mathbb{R}^d}\|\boldsymbol{v}( \boldsymbol{x},t)\|^2p_t(\boldsymbol{x})d\boldsymbol{x}< \infty
    \end{align}
\end{theorem}

\textit{Proof.} We start with the definition of the Fourier transform $\mathcal{F}(\boldsymbol{\omega},t)[p_t]$ of the density $p_t(\boldsymbol{x})$ for the random variable $\boldsymbol{x}_t=I(\boldsymbol{x}_0, \boldsymbol{x}_T,t)+\gamma(t)\boldsymbol{z}$, given by:
\begin{align}
    \mathcal{F}(\boldsymbol{\omega},t)[p_t]=\int_{\mathbb{R}^d}e^{i\boldsymbol{\omega}\cdot \boldsymbol{x}_t}p_t(\boldsymbol{x})=\mathbb{E}\left[e^{i\boldsymbol{\omega}\cdot \boldsymbol{x}_t}\right]=\mathbb{E}\left[e^{i\boldsymbol{\omega}\cdot (I(\boldsymbol{x}_0, \boldsymbol{x}_T,t) +\gamma(t)\boldsymbol{z})}\right]
\end{align}
where $\boldsymbol{\omega}\in \mathbb{R}^d$ is the $d$-dimensional Fourier frequency variable. To show that $p_t(\boldsymbol{x})$ satisfies the continuity equation in (\ref{eq:si-continuity}), we can take the time derivative of its Fourier transform $\mathcal{F}(\boldsymbol{\omega},t)[p_t]$ to get:
\begin{small}
\begin{align}
    \partial_t\mathcal{F}(\boldsymbol{\omega},t)[p_t]&=\bluetext{\partial_t}\mathbb{E}\left[e^{i\boldsymbol{\omega}\cdot (I(\boldsymbol{x}_0, \boldsymbol{x}_T,t) +\gamma(t)\boldsymbol{z})}\right]=\mathbb{E}\left[\bluetext{\partial_t}e^{i\boldsymbol{\omega}\cdot (I(\boldsymbol{x}_0, \boldsymbol{x}_T,t) +\gamma(t)\boldsymbol{z})}\right]\nonumber\\
    &=\mathbb{E}\left[e^{i\boldsymbol{\omega}\cdot \boldsymbol{x}_t}\partial_t\big(\pinktext{i\boldsymbol{\omega}}\cdot (I(\boldsymbol{x}_0, \boldsymbol{x}_T,t) +\gamma(t)\boldsymbol{z})\big)\right]=\pinktext{i\boldsymbol{\omega}}\cdot \underbrace{\mathbb{E}\left[e^{i\boldsymbol{\omega}\cdot \boldsymbol{x}_t}(\partial_tI(\boldsymbol{x}_0, \boldsymbol{x}_T,t)+\dot{\gamma}(t)\boldsymbol{z})\right]}_{(\bigstar)}
\end{align}
\end{small}
Now, we can apply the law of total expectation $\mathbb{E}[X]=\mathbb{E}[\mathbb{E}[X|Y]]$ to write $(\bigstar)$ with respect to a conditional expectation on $\boldsymbol{X}_t=\boldsymbol{x}$:
\begin{small}
\begin{align}
    \partial_t\mathcal{F}(\boldsymbol{\omega},t)[p_t]&=\pinktext{i\boldsymbol{\omega}}\cdot\mathbb{E}\left[e^{i\boldsymbol{\omega}\cdot \boldsymbol{x}_t}(\partial_tI(\boldsymbol{x}_0, \boldsymbol{x}_T,t)+\dot{\gamma}(t)\boldsymbol{z})\right]\nonumber\\
    &=\pinktext{i\boldsymbol{\omega}}\cdot\mathbb{E}\left[e^{i\boldsymbol{\omega}\cdot \boldsymbol{x}}\bluetext{\mathbb{E}\big[(\partial_tI(\boldsymbol{x}_0, \boldsymbol{x}_T,t)+\dot{\gamma}(t)\boldsymbol{z})|\boldsymbol{X}_t=\boldsymbol{x}\big]}\right]\nonumber\\
    &=\pinktext{i\boldsymbol{\omega}}\cdot\int_{\mathbb{R}^d} e^{i\boldsymbol{\omega}\cdot \boldsymbol{x}}\bluetext{\underbrace{\mathbb{E}\big[(\partial_tI(\boldsymbol{x}_0, \boldsymbol{x}_T,t)+\dot{\gamma}(t)\boldsymbol{z})|\boldsymbol{X}_t=\boldsymbol{x}\big]}_{=:\boldsymbol{v}(\boldsymbol{x},t)}}p_t(\boldsymbol{x})d\boldsymbol{x}\nonumber\\
    &=\pinktext{i\boldsymbol{\omega}}\cdot\int_{\mathbb{R}^d} e^{i\boldsymbol{\omega}\cdot \boldsymbol{x}}\bluetext{\boldsymbol{v}(\boldsymbol{x},t)}p_t(\boldsymbol{x})d\boldsymbol{x}
\end{align}
\end{small}
Next, we show that the Fourier transform of $-\nabla \cdot (\boldsymbol{v}p_t)$ is exactly $i\boldsymbol{\omega}\mathcal{F}(\boldsymbol{\omega},t)[\boldsymbol{v}p_t]$. Given that $\nabla\cdot (\boldsymbol{v}p_t)=\sum_{j=1}^d\partial_{\boldsymbol{x}_j}(\boldsymbol{v}p_t)_j$, we have:
\begin{small}
\begin{align}
    \mathcal{F}(\boldsymbol{\omega},t)[-\nabla \cdot (\boldsymbol{v}p_t)]&=-\sum_{j=1}^d\int_{\mathbb{R}^d}\underbrace{e^{i\boldsymbol{\omega}\cdot\boldsymbol{x}}}_{u}\underbrace{\partial_{\boldsymbol{x}_j}(\boldsymbol{v}p_t)_j}_{dv}d\boldsymbol{x}\nonumber\\
    &=\underbrace{\left[e^{i\boldsymbol{\omega}\cdot\boldsymbol{x}}(\boldsymbol{v}p_t)\right]_{-\infty}^\infty}_{=0}+\sum_{j=1}^d\int_{\mathbb{R}^d}(\boldsymbol{v}p_t)_j\partial_{\boldsymbol{x}_j}\left(e^{i\boldsymbol{\omega}\cdot \boldsymbol{x}}\right)d\boldsymbol{x}\nonumber\\
    &=\sum_{j=1}^d\int_{\mathbb{R}^d}(\boldsymbol{v}p_t)_j\partial_{\boldsymbol{x}_j}\left(e^{i\sum_{\ell=1}^d\boldsymbol{\omega}_\ell \boldsymbol{x}_\ell}\right)d\boldsymbol{x}\nonumber\\
    &=\sum_{j=1}^d\int_{\mathbb{R}^d}(\boldsymbol{v}p_t)_j\left(i\boldsymbol{\omega}_je^{i\boldsymbol{\omega}\cdot\boldsymbol{x}}\right)d\boldsymbol{x}\nonumber\\
    &=\sum_{j=1}^di\boldsymbol{\omega}_j\int_{\mathbb{R}^d}e^{i\boldsymbol{\omega}\cdot\boldsymbol{x}}(\boldsymbol{v}p_t)_jd\boldsymbol{x}\nonumber\\
    &=i\boldsymbol{\omega}\cdot \int_{\mathbb{R}^d}e^{i\boldsymbol{\omega}\cdot\boldsymbol{x}}\boldsymbol{v}p_td\boldsymbol{x}
\end{align}
\end{small}
which proves that $\boldsymbol{v}$ and $p_t$ satisfies the continuity equation in the real space:
\begin{align}
    \partial_t\mathcal{F}(\boldsymbol{\omega},t)[p_t]&=\mathcal{F}(\boldsymbol{\omega},t)[-\nabla \cdot (\boldsymbol{v}p_t)]\implies \boxed{\partial_tp_t=-\nabla\cdot (\boldsymbol{v}p_t)}
\end{align}
and we conclude our proof. \hfill $\square$

Having established the fundamental properties of stochastic interpolants, we now observe that they induce a family of time-evolving densities $(p_t)_{t\in[0,T]}$ governed by the continuity equation (\ref{eq:si-continuity}). This equation characterizes deterministic mass transport under a velocity field $\boldsymbol{v}$, and forms the core dynamical constraint underlying optimal transport. In particular, stochastic interpolants provide a constructive way to define admissible trajectories that transport probability mass between endpoint distributions.

However, while stochastic interpolants describe valid transport dynamics, they do not yet specify which trajectory is \textit{optimal}. The SB problem resolves this ambiguity by selecting, among all admissible paths satisfying the same marginal constraints, the one that is closest to a reference stochastic process in the sense of minimizing path-space KL divergence. As we now show, this optimality can be expressed through stochastic control, where the velocity field $\boldsymbol{v}$ is parameterized via a controlled drift that minimizes a quadratic control cost while matching the prescribed marginals.

We begin by recalling the (\ref{eq:dynamic-sb-problem}), which aims to determine the optimal control drift and density evolution $(\boldsymbol{u}^\star, p_t^\star)$ that solve the minimization problem: 
\begin{small}
\begin{align}
    &\inf_{(\boldsymbol{u}, p_t)}\left[\int_0^T\int_{\mathbb{R}^d}\frac{1}{2}\|\boldsymbol{u}(\boldsymbol{x},t)\|^2p_t(\boldsymbol{x})d\boldsymbol{x}dt \right]\quad\text{s.t.}\quad \begin{cases}
        \partial_tp_t=-\nabla\cdot (p_t\boldsymbol{u})+ \epsilon\Delta p_t\\
        p_0=\pi_0, \quad p_T=\pi_T
    \end{cases}
    \label{eq:si-sb-problem}
\end{align}
\end{small}
which can be rewritten as solving (\ref{eq:prop-hjb-fpe-system}) given by:
\begin{align}
    \begin{cases}
        \partial_t\psi_t+\frac{1}{2} \|\nabla\psi_t\|^2=-\epsilon\Delta \psi_t\\
        \partial_t p_t^\star+\nabla\cdot (p_t^\star\boldsymbol{u}^\star)=\epsilon\Delta p^\star_t
    \end{cases}\quad \text{s.t.}\quad \begin{cases}
        p^\star_0=\pi_0\\
        p^\star_T=\pi_T
    \end{cases}\label{eq:si-hjb-fpe}
\end{align}
where we simplify the setting by setting $\boldsymbol{f}\equiv 0$ and $\epsilon =\frac{\sigma^2_t}{2}$.

Before we construct the stochastic interpolant that solves the SB problem, we will define an invertible map that generates the optimal SB density $p_t^\star$ that solves (\ref{eq:si-sb-problem}).
\begin{definition}[Invertible Map]\label{def:interp-invertable-map}
    Define an invertible map $M: \mathbb{R}^d\times[0,T]\to \mathbb{R}^d$ where $M, M^{-1}\in C^1([0,T], (C^d(\mathbb{R}^d))^d)$ such that:
    \begin{align}
        p^\star_t(\boldsymbol{x})=M(\cdot, t)_{\#}\mathcal{N}(\boldsymbol{0}, \boldsymbol{I}_d)
    \end{align}
    In other words, given a Gaussian random variable $\boldsymbol{z}\sim \mathcal{N}(\boldsymbol{0}, \boldsymbol{I}_d)$, we have $\boldsymbol{x}_t=M(\boldsymbol{z},t)\sim p^\star_t$.
\end{definition}
Given the existence of the invertible map, we can now derive the \textbf{stochastic interpolant} that solves the SB problem.

\begin{lemma}[Stochastic Interpolant Form of SB Solution (Lemma 3.12 in \citet{albergo2025stochastic}]\label{lemma:stochastic-interpolant-sb}
    Given the existence of an invertible map $M$ defined in Definition \ref{def:interp-invertable-map}, the optimal density $p_t^\star$ that solves the dynamic SB problem can be written as a \textbf{stochastic interpolant} of the form:
    \begin{align}
        \boldsymbol{x}_t=M\big(\bluetext{\alpha(t)M^{-1}(\boldsymbol{x}_0, 0 ) +\beta(t)M^{-1}(\boldsymbol{x}_T, T)},t\big)+\gamma(t)\label{eq:si-lemma-eq1}
    \end{align}
    where $\alpha^2(t)+\beta^2(t)+\gamma^2(t)=1$. This corresponds to defining the interpolant function from Definition \ref{def:stochastic-interpolants} as $I(\boldsymbol{x}_0, \boldsymbol{x}_T,t)=M(\alpha(t)M^{-1}(\boldsymbol{x}_0,0)+\beta(t)M^{-1}(\boldsymbol{x}_T,T))$.
\end{lemma}

\textit{Proof.} Intuitively, this Lemma states that the density $p_t^\star$ that solves the HJB-FP system defined in (\ref{eq:si-hjb-fpe}) is exactly the distribution of the random variable obtained defined in (\ref{eq:si-lemma-eq1}).

By definition of the map $M(\cdot, t)$ which \textit{transports} a standard Gaussian to the target density $p_t^\star$ at time $t$, so the \textbf{inverse map} $M^{-1}(\cdot, t)$ must transport the target density $p_t^\star$ back to a standard Gaussian, such that:
\begin{align}
    \boldsymbol{x}_0\sim \pi_0, \quad &M^{-1}(\boldsymbol{x}_0, 0)\sim \mathcal{N}(\boldsymbol{0}, \boldsymbol{I}_d)\\
    \boldsymbol{x}_T\sim \pi_T, \quad &M^{-1}(\boldsymbol{x}_T,T)\sim \mathcal{N}(\boldsymbol{0}, \boldsymbol{I}_d)
\end{align}

Since $\boldsymbol{z}\sim \mathcal{N}(\boldsymbol{0}, \boldsymbol{I}_d)$ is also sampled from a standard Gaussian and $\boldsymbol{x}_0, \boldsymbol{x}_T, \boldsymbol{z}$ are drawn independently in the stochastic interpolant construction, we have that the linear combination of independent standard Gaussians given by:
\begin{align}
    \alpha(t)M^{-1}(\boldsymbol{x}_0, 0)+\beta(t)M^{-1}(\boldsymbol{x}_T,T)+\gamma(t)\boldsymbol{z}\sim \bluetext{\mathcal{N}(\boldsymbol{0}, \alpha^2(t)\boldsymbol{I}_d+\beta^2(t)\boldsymbol{I}_d+\gamma^2(t)\boldsymbol{I}_d)}=\mathcal{N}(\boldsymbol{0},\boldsymbol{I}_d)\label{eq:si-lemma-linear}
\end{align}
is a Gaussian with zero-mean and covariance $(\alpha^2(t)+\beta^2(t)+\gamma^2(t))\boldsymbol{I}_d$. Since we defined $\alpha^2(t)+\beta^2(t)+\gamma^2(t)=1$, then this is just a standard Gaussian $\mathcal{N}(\boldsymbol{0},\boldsymbol{I}_d)$. Therefore, we can apply the definition of the map $M(\cdot, t)$ on the linear combination (\ref{eq:si-lemma-linear}) which yields a random variable from the target distribution $p_t^\star$:
\begin{align}
    \boldsymbol{x}_t=M\big(\bluetext{\alpha(t)M^{-1}(\boldsymbol{x}_0, 0 ) +\beta(t)M^{-1}(\boldsymbol{x}_T, T)},t\big)+\gamma(t)\boldsymbol{z}\sim \bluetext{p_t^\star}
\end{align}
which concludes the proof. \hfill $\square$ 

Lemma \ref{lemma:stochastic-interpolant-sb} shows that the density $p_t^\star$ that solves the HJB-FP system can be obtained as the distribution of a random variable constructed by interpolation of samples from the terminal marginals $\boldsymbol{x}_0\sim \pi_0$ and $\boldsymbol{x}_T\sim \pi_T$ in Gaussian space, which establishes the form of the optimal interpolant $I^\star$ that generates the Schrödinger bridge marginals. Now, we can derive an objective whose minimizer recovers this optimal interpolant $I^\star$. 

\begin{proposition}[Solving SB problem with Stochastic Interpolants (Theorem 3.13 in \citet{albergo2025stochastic}]\label{eq:sbp-stochastic-interpoalnts}
    Define a scalar function $\gamma(t):[0,T]\to[0,1)$ which returns zero at the terminal time points (i.e., $\gamma(0)=\gamma(T)=0$), returns non-zero at intermediate times (i.e., $\forall t\in (0,T), \quad \gamma(t)>0$), and satisfies $\gamma\in C^2((0,T))$ and $\gamma^2\in C^1([0,T] )$. 
    
    Then, given independent $\boldsymbol{x}_0, \boldsymbol{x}_T, \boldsymbol{z}$ solving the max-min problem over $\hat{I}\in C^1([0,T], (C^1(\mathbb{R}^d\times \mathbb{R}^d))^d)$ and $\hat{\boldsymbol{u}}\in C^0([0,T], (C^1(\mathbb{R}^d))^d)$ given by:
    \begin{small}
    \begin{align}
        \underset{\hat{I}}{\max}\underset{\hat{\boldsymbol{u}}}{\min}\int_0^T&\mathbb{E}\bigg[\frac{1}{2}\|\hat{\boldsymbol{u}}(\boldsymbol{x},t)\|^2-\left(\partial_t\hat{I}(\boldsymbol{x}_0, \boldsymbol{x}_T,t)+\left(\dot{\gamma}(t)-\epsilon\gamma^{-1}(t)\right)\boldsymbol{z}\right)\cdot \hat{\boldsymbol{u}}(\boldsymbol{x},t)\bigg]dt\label{eq:max-min-orig}\\
        &\text{s.t.}\quad \begin{cases}
            \boldsymbol{x}_t=\hat{I}(t, \boldsymbol{x}_0, \boldsymbol{x}_T)+\gamma(t)\boldsymbol{z}\\
            \boldsymbol{x}_0\sim \pi_0, \quad \boldsymbol{x}_T\sim \pi_T, \quad \boldsymbol{z}\sim \mathcal{N}(\boldsymbol{0}, \boldsymbol{I}_d)
        \end{cases}
    \end{align}
    \end{small}
    where, given the existence of the invertible map $M$, all optimal $(I^\star, \boldsymbol{u}^\star)$ produces the stochastic interpolant $\boldsymbol{x}_t=I^\star(\boldsymbol{x}_0, \boldsymbol{x}_T,t)+\gamma(t)\boldsymbol{z}$ with marginals $p_t^\star$ that satisfy the continuity equation of the form:
    \begin{align}
        \partial_tp_t=-\nabla \cdot(\hat{p}_t\hat{\boldsymbol{v}})
    \end{align}
    where $\hat{\boldsymbol{v}}$ is the effective velocity field that accounts for the Gaussian perturbation induced by $\gamma(t)\boldsymbol{z}$. 
\end{proposition}

\textit{Proof.} 
To prove this, we begin by defining the effective velocity $\hat{\boldsymbol{v}}$ of the stochastic interpolant and rewriting the objective as an optimization over $\hat{\boldsymbol{v}}$. Then, we write the constrained objective using Lagrange multipliers and solve for the optimal solution. First, we rewrite the Fokker-Planck equation into a continuity constraint that must be satisfied by the effective velocity:
\begin{align}
    \partial_tp_t=-\nabla \cdot(\hat{p}_t\boldsymbol{v})+\epsilon\Delta \hat{p}_t&\implies \partial_tp_t=-\nabla \cdot(\hat{p}_t\boldsymbol{v})+\epsilon\nabla \cdot\nabla  \hat{p}_t\nonumber\\
    &\implies\boxed{\partial_tp_t=-\nabla \cdot(\hat{p}_t\underbrace{(\boldsymbol{v}-\bluetext{\epsilon \nabla\log  \hat p_t})}_{=:\hat{\boldsymbol{v}}})}
\end{align}
To define the effective velocity that accounts for the contribution of the Gaussian perturbation, we need to incorporate $\epsilon \nabla\log  \hat p_t$. Since the stochastic interpolant is defined as $\boldsymbol{x}_t=I(\boldsymbol{x}_0, \boldsymbol{x}_T,t)+\gamma(t)\boldsymbol{z}$, the conditional distribution of $\boldsymbol{x}_t$ given $(\boldsymbol{x}_0, \boldsymbol{x}_T)$ is Gaussian with variance $\gamma^2(t)$. A standard identity for Gaussian variables implies that
\begin{align}
    \mathbb{E}[\boldsymbol{z} | \boldsymbol{X}_t = \boldsymbol{x}] = -\gamma(t) \nabla \log \hat p_t(\boldsymbol{x}),
\end{align}
where $\hat p_t$ is the density of $\boldsymbol{x}_t$. Therefore, we have
\begin{align}
    -\epsilon \gamma^{-1}(t)\mathbb{E}[\boldsymbol{z} | \boldsymbol{X}_t = \boldsymbol{x}]&= \epsilon \nabla \log \hat p_t(\boldsymbol{x})
\end{align}
Now, we can define the average effective velocity as:
\begin{align}
    \hat{\boldsymbol{v}}(\boldsymbol{x},t)&:=\mathbb{E}[\partial_t\hat{I}(\boldsymbol{x}_0, \boldsymbol{x}_T,t)+(\dot{\gamma}(t)\bluetext{-\epsilon\gamma^{-1}(t)})\boldsymbol{z}|\boldsymbol{X}_t=\boldsymbol{x}]
\end{align}
where $-\epsilon\gamma^{-1}(t)$ \textit{corrects for the Gaussian perturabtion}. Then, rewriting the expectation in the original objective as the integral, we have:
\begin{small}
\begin{align}
    \underset{\hat{I}}{\max}\underset{\hat{\boldsymbol{u}}}{\min}\int_0^T&\mathbb{E}\bigg[\frac{1}{2}\|\hat{\boldsymbol{u}}(\hat{\boldsymbol{x}}_t,t)\|^2-\left(\partial_t\hat{I}(\boldsymbol{x}_0, \boldsymbol{x}_T,t)+\left(\dot{\gamma}(t)-\epsilon\gamma^{-1}(t)\right)\boldsymbol{z}\right)\cdot \hat{\boldsymbol{u}}(\hat{\boldsymbol{x}}_t,t)\bigg]dt\nonumber\\
    &=\underset{\hat{p}, \boldsymbol{v}}{\max}\underset{\hat{\boldsymbol{u}}}{\min}\int_0^T\int_{\mathbb{R}^d}\bigg[\frac{1}{2}\|\hat{\boldsymbol{u}}(\boldsymbol{x},t)\|^2-\bluetext{\mathbb{E}\left[\partial_t\hat{I}(\boldsymbol{x}_0, \boldsymbol{x}_T,t)+\left(\dot{\gamma}(t)-\epsilon\gamma^{-1}(t)\right)\boldsymbol{z}\big|\boldsymbol{X}_t=\boldsymbol{x}\right]}\cdot \hat{\boldsymbol{u}}(\boldsymbol{x},t)\bigg]\hat{p}_t(\boldsymbol{x})d\boldsymbol{x}dt\nonumber\\
    &=\underset{\hat{p}, \boldsymbol{v}}{\max}\underset{\hat{\boldsymbol{u}}}{\min}\int_0^T\int_{\mathbb{R}^d}\bigg[\frac{1}{2}\|\hat{\boldsymbol{u}}(\boldsymbol{x},t)\|^2\hat{p}_t(\boldsymbol{x})-\bluetext{\hat{\boldsymbol{v}}(\boldsymbol{x},t)}\cdot \hat{\boldsymbol{u}}(\boldsymbol{x},t)\bigg]\hat{p}_t(\boldsymbol{x})d\boldsymbol{x}dt\label{eq:si-constrained-obj1}\\
    &\text{s.t.}\quad \partial_t\hat{p}_t+\nabla\cdot (\boldsymbol{v}\hat{p}_t)=\epsilon \Delta\hat{p}_t, \quad \hat{p}_0=\pi_0, \quad \hat{p}_ T=\pi_T\nonumber
\end{align}
\end{small}
Rewriting (\ref{eq:si-constrained-obj1}) as an unconstrained objective with Lagrange multipliers $\psi_t(\boldsymbol{x}), \eta_0(\boldsymbol{x})$ and $\eta_T(\boldsymbol{x})$, we have:
\begin{align}
    \underset{\hat{p}, \hat{\boldsymbol{v}}}{\max}\underset{\hat{\boldsymbol{u}}}{\min}&\bigg(\int_0^T\int_{\mathbb{R}^d}\bigg[\frac{1}{2}\|\hat{\boldsymbol{u}}(\boldsymbol{x},t)\|^2\hat{p}_t(\boldsymbol{x})-\bluetext{\hat{\boldsymbol{v}}(\boldsymbol{x},t)}\cdot \hat{\boldsymbol{u}}(\boldsymbol{x},t)\bigg]\hat{p}_t(\boldsymbol{x})d\boldsymbol{x}dt\nonumber\\
    &-\int_0^T\int_{\mathbb{R}^d}\psi_t(\boldsymbol{x})\left(\partial_t\hat{p}_t(\boldsymbol{x})+\nabla\cdot (\hat{\boldsymbol{v}}(\boldsymbol{x},t)\hat{p}_t(\boldsymbol{x}))\right)d\boldsymbol{x}dt\nonumber\\
    &+\int_{\mathbb{R}^d}\eta_0(\boldsymbol{x})\left(\hat{p}_0(\boldsymbol{x})-\pi_0(\boldsymbol{x})\right)dt+\int_{\mathbb{R}^d}\eta_T(\boldsymbol{x})\left(\hat{p}_T(\boldsymbol{x})-\pi_T(\boldsymbol{x})\right)dt\bigg)
\end{align}
The optimal $(p_t^\star, \boldsymbol{u}^\star)$ that minimizes the Lagrangian pointwise satisfies the following optimality conditions:
\begin{enumerate}
    \item[(i)] Varying with respect to $\psi_t(\boldsymbol{x})$ enforces:
    \begin{small}
    \begin{align}
        \bluetext{\partial_t\hat p_t+\nabla \cdot (\hat{\boldsymbol{v}} \hat p_t)=0}
    \end{align}
    \end{small}
    \item[(ii)] Varying with respect to $\hat{\boldsymbol{u}}$ enforces:
    \begin{small}
    \begin{align}
        \frac{\partial}{\partial \boldsymbol{u}} \left(\frac{1}{2}\|\hat{\boldsymbol{u}}\|^2p_t-\hat{\boldsymbol{u}}\cdot (\hat{\boldsymbol{v}} \hat p_t)\right)=\hat{\boldsymbol{u}} \hat p_t-(\hat{\boldsymbol{v}} \hat p_t)=0\implies \bluetext{\boldsymbol{u}^\star=\hat{\boldsymbol{v}}}
    \end{align}
    \end{small}
    \item[(iii)] Since $(\hat{\boldsymbol{v}} \hat p_t)$ appears in $-\hat{\boldsymbol{u}}\cdot (\hat{\boldsymbol{v}}^\star \hat p_t)$ and $\psi_t(\boldsymbol{x})\nabla\cdot (\hat{\boldsymbol{v}} \hat p_t)$, we can integrate by parts to get:
    \begin{small}
    \begin{align}
        -\int\hat{\boldsymbol{u}}\cdot (\hat{\boldsymbol{v}} \hat p_t)-\int\psi_t\nabla\cdot (\hat{\boldsymbol{v}} \hat p_t)&=-\int\hat{\boldsymbol{u}}\cdot (\hat{\boldsymbol{v}} \hat p_t)-\underbrace{[\psi_t (\boldsymbol{x})(\hat{\boldsymbol{v}} \hat p_t)]_{-\infty}^\infty}_{=0}+\int\nabla \psi_t\cdot (\hat{\boldsymbol{v}} \hat p_t)\nonumber\\
        &=-\int\hat{\boldsymbol{u}}\cdot (\hat{\boldsymbol{v}} \hat p_t)+\int\nabla \psi_t\cdot (\hat{\boldsymbol{v}} \hat p_t)\nonumber\\
        &=\int(\nabla \psi_t-\hat{\boldsymbol{u}})\cdot (\hat{\boldsymbol{v}} \hat p_t)\nonumber\\
        &\implies \bluetext{\nabla \psi_t-\boldsymbol{u}^\star=0}
    \end{align}
    \end{small}
    \item[(iv)] Isolating the terms that depend on $\hat p_t$ (excluding those that involve $\hat{\boldsymbol{v}} \hat p_t$, which we accounted for previously) and integrating by parts to factor out $\hat p_t$, we have:
    \begin{small}
    \begin{align}
        \int_0^T\int_{\mathbb{R}^d}\left[\frac{1}{2}\|\hat{\boldsymbol{u}}\|^2\hat p_t-\psi_t\partial_t\hat p_t\right]d\boldsymbol{x}dt&=\int_0^T\int_{\mathbb{R}^d}\left[\frac{1}{2}\|\hat{\boldsymbol{u}}\|^2\hat p_t+(\partial_t\psi_t)\hat p_t\right]d\boldsymbol{x}dt\nonumber\\
        &=\int_0^T\int_{\mathbb{R}^d}\bluetext{\underbrace{\left[\frac{1}{2}\|\hat{\boldsymbol{u}}\|^2+(\partial_t\psi_t)\right]}_{\text{vanishes pointwise at minimum}}}\hat p_td\boldsymbol{x}dt\nonumber\\
        &\implies  \bluetext{\partial_t\psi_t+\frac{1}{2}\|\boldsymbol{u}^\star\|^2=0}
    \end{align}
    \end{small}
    \item[(v)] Varying with respect to $\eta_0(\boldsymbol{x})$ and $\eta_T(\boldsymbol{x})$ yields $p_0^\star=\pi_0$ and $p^\star_T=\pi_T$.
\end{enumerate}
Putting it all together, (ii) and (iii) enforce $\boldsymbol{u}^\star=\nabla \psi_t$, which can be substituted into (i) to get the following system:
\begin{align}
    \begin{cases}
        \partial_t\psi_t+\frac{1}{2}\|\nabla\psi_t\|^2=0\\
        \partial_tp^\star_t+\nabla \cdot(p_t^\star\nabla \psi_t)=0
    \end{cases}\quad \textbf{s.t.}\quad\begin{cases}
        p_0=\pi_0\\
        p_T=\pi_T
    \end{cases}
\end{align}
which is exactly the HJB-FP system satisfied by the solution to the SB problem with given in (\ref{eq:si-hjb-fpe}) with vanishing diffusion, since the Gaussian diffusion is absorbed into the marginal density $p_t$ through the effective velocity $\hat{\boldsymbol{v}}$. Given the existance of the invertible map from Definition \ref{def:interp-invertable-map} that yields the optimal density $p^\star_t$, we have that the pair $(I^\star, \boldsymbol{u}^\star)$ that solves the original max-min objective in (\ref{eq:max-min-orig}) exists which we have shown yields the optimal density $\boldsymbol{x}_t\sim p^\star_t$ for $\boldsymbol{x}_t=I^\star(\boldsymbol{x}_0, \boldsymbol{x}_T,t)+\gamma(t)\boldsymbol{z}$.\hfill $\square$

This result shows that stochastic interpolants provide a alternative representation of the Schrödinger bridge dynamics $(\boldsymbol{u}^\star,p^\star_t)$, where the optimal control and velocity fields coincide and are given by the gradient of the Lagrange multiplier $\nabla\psi_t$, yielding the coupled HJB-FP system that define the optimality conditions of the SB solution. Therefore, sampling from the stochastic interpolant $\boldsymbol{x}_t=I^\star(\boldsymbol{x}_0, \boldsymbol{x}_T,t)+\gamma(t)\boldsymbol{z}$, where $\boldsymbol{z}\sim \mathcal{N}(\boldsymbol{0},\boldsymbol{I}_d)$ generates trajectories whose marginals match the marginal density flow $p_t^\star$ of the SB path measure $\mathbb{P}^\star$.

\subsection{Closing Remarks for Section \ref{sec:building-stochastic-bridge}}
This section explored the theoretical foundations for building a Schrödinger bridge using several approaches. While each approach originates from a different mathematical viewpoint, they ultimately converge to a unified form of a Markov control drift that minimally corrects the uncontrolled reference dynamics such that they reconstruct the prescribed marginal distributions.

The \textbf{key takeaway} of this section is that stochastic bridges are not arbitrary conditioned processes, but rather minimal-entropy corrections of a reference diffusion that preserve the bridge structure while introducing the smallest possible dynamical adjustment. This adjustment consistently appears as a gradient of a logarithmic potential, providing a unifying perspective to the structure of the Schrödinger bridge.

The different constructions introduced in this section can therefore be understood as alternative ways of identifying the optimal control drift that modifies the reference dynamics:
\begin{enumerate}
    \item [(i)] \textbf{Mixture of Conditional Bridges (Section \ref{subsect:mixture-bridges}):} This section shows that the dynamic SB $\mathbb{P}^\star$ can be expressed as a mixture of endpoint-conditioned stochastic bridges given samples from the optimal endpoint law $(\boldsymbol{x}_0, \boldsymbol{x}_T) \sim \pi^\star_{0,T}$, which can be constructed with a conditional drift $\boldsymbol{u}(\boldsymbol{x},t;\boldsymbol{x}_0, \boldsymbol{x}_T)$. This decomposition separates the SB problem into estimating the optimal static coupling and learning the conditional bridge dynamics.
    \item [(ii)] \textbf{Time-reversal formula (Section \ref{subsec:time-reversal}):} This section shows that the time-reversal formula yields a backward correction term $\nabla \log p_{T-s}(\boldsymbol{x})$ that corresponds to the \textit{score function} in score-based generative modeling. This formulation models the endpoint-conditioned bridge for \textit{uncontrolled} forward process. 
    \item [(iii)] \textbf{Forward-backward SDEs (Section \ref{subsec:forward-backward-sde}):} This section analyzes how the theory of forward–backward theory generalizes the time-reversal formula for a forward process containing a non-deterministic control drift $\nabla \log \varphi_t(\boldsymbol{x})$ which yields a backward control drift $\nabla \log \hat\varphi_t(\boldsymbol{x})$ that evolves via a system of forward-backward SDEs.
    \item[(iv)] \textbf{Doob's $h$-transform (Section \ref{subsec:doob-transform}):} This section defines an $h$-function $h(\boldsymbol{x},t)$ that \textit{reweights path transitions} by its potential at time $\tau$. The reweighted path measure $\mathbb{P}^h$ is defined by an SDE with a correction term $\nabla \log h(\boldsymbol{x},t)$. By defining $h(\boldsymbol{x},t):= \mathbb{E}_{\mathbb{Q}}[\varphi_T(\boldsymbol{X}_T)|\boldsymbol{X}_t=\boldsymbol{x}]$, we show that the tilted path measure recovers the Schrödinger bridge $\mathbb{P}^\star$.
    \item[(v)] \textbf{Markov and reciprocal projections (Section \ref{subsec:markov-reciprocal-proj}):} This section shows that the optimal Schrödinger bridge $\mathbb{P}^\star$ can be interpreted as the equilibrium point between projections onto the space of Markov path measures and the reciprocal class $\mathcal{R}(\mathbb{Q})$ containing mixtures of bridges under the reference path $\mathbb{Q}$. In this formulation, the optimal drift is expressed as an expectation over the target-conditioned path measure $\mathbb{E}_{\Pi_{T|t}}[\nabla  \log \mathbb{Q}_{T|t}(\boldsymbol{X}_T|\boldsymbol{X}_t) |\boldsymbol{X}_t]$. 
    \item[(vi)] \textbf{Stochastic interpolants (Section \ref{subsec:stochastic-interpolants}):} The stochastic interpolant framework represents the bridge by expressing the intermediate state as $\boldsymbol{x}_t=I^\star(\boldsymbol{x}_0, \boldsymbol{x}_T,t) +\gamma(t)\boldsymbol{z}$, where $I^\star$ is the optimal interpolant between samples $\boldsymbol{x}_0\sim \pi_0$ and $\boldsymbol{x}_T\sim \pi_T$ and $\boldsymbol{z}\sim \mathcal{N}(\boldsymbol{0}, \boldsymbol{I}_d)$ is Gaussian noise. We show that the induced velocity field satisfies the same optimality conditions that characterize the Schrödinger bridge dynamics.
\end{enumerate}

While this section provides a principled framework for constructing stochastic bridges between prescribed endpoint distributions that solve the (\ref{eq:dynamic-sb-problem}), recent advances in generative modeling have motivated a variety of \textbf{specialized Schrödinger bridge formulations} tailored to different modeling assumptions and problem settings. These variants extend the original framework in several directions, ranging from alternative reference processes to mean-field interactions, unbalanced mass transport, and multi-marginal and multi-modal constraints. In the next section, we analyze several of these problem variations and discuss how they modify our dynamic SB problem formulation while preserving its \textbf{core principle}: constructing stochastic dynamics that minimally deviate from a reference process while matching prescribed marginal distributions.

\newpage
\section{Variations of the Schrödinger Bridge Problem}
\label{sec:variations-of-sb}
In previous sections, we have established the foundational theories and intuition behind the classical static and dynamic Schrödinger bridge problem and have shown how to derive stochastic bridges from scratch using various techniques. Now, we are ready to describe diverse variations of the SB problem that have been introduced in conjunction to novel generative modeling techniques, each of which are specialized for different settings and tasks. 

Specifically, we analyze the Gaussian SB problem (Section \ref{subsec:gaussian-sb}), the generalized SB problem (Section \ref{subsec:generalized-sb}), the multi-marginal SB problem (Section \ref{subsec:multi-marginal-sb}), the unbalanced SB problem (Section \ref{subsec:unbalanced-sbp}), the branched SB problem (Section \ref{subsec:branched-sbp}), and finally the fractional SB problem (Section \ref{subsec:fractional-sbp}). For each problem variation, we provide the intuition and formal definition of the problem, the necessary theoretical background, and relevant proofs and derivations that are useful for interpreting the problem and its optimal solution.

\subsection{Gaussian Schrödinger Bridge Problem}
\label{subsec:gaussian-sb}

While the (\ref{eq:dynamic-sb-problem}) defined in Section \ref{sec:dynamic-sb} does not admit a close form solution in general, in the special case where the marginal distributions $\pi_0, \pi_T$ are Gaussian distributions defined as $\pi_0\sim \mathcal{N}(\boldsymbol{\mu}_0, \boldsymbol{\Sigma}_0), \pi_T\sim\mathcal{N}(\boldsymbol{\mu}_T, \boldsymbol{\Sigma}_T)$, the SB solution can be solved in closed form \citep{bunne2023schrodinger}. This special case of the dynamic SB problem is called the Gaussian Schrödinger Bridge (SB) problem \citep{bunne2023schrodinger, mallasto2022entropy}. Beforing defining the Gaussian SB problem, we start by defining the \textbf{Gaussian formulation of the entropic OT problem} \citep{bojilov2016matching, janati2020entropic, mallasto2022entropy}, which will become crucial for our later derivation of the closed-form solution in the Gaussian SB setting. 

\begin{lemma}[Static Entropy-Regularized Gaussian Optimal Transport]\label{lemma:static-entropy-gaussian-ot}
Let $\pi_0 =\mathcal{N}(\boldsymbol{\mu}, \boldsymbol{\Sigma})$ and $\pi_T = \mathcal{N}(\boldsymbol{\mu}, \boldsymbol{\Sigma})$ be Gaussian probability measures on $\mathbb{R}^d$ where $\boldsymbol{\mu}, \boldsymbol{\mu}'\in \mathbb{R}^d$ are the mean vectors and $\boldsymbol{\Sigma}$. Consider the entropy-regularized optimal transport problem
\begin{align}
    \min_{\pi_{0,T} \in \Pi(\pi_0, \pi_T)}\int \|\boldsymbol{x}_T - \boldsymbol{x}_0\|^2  d\pi(\boldsymbol{x}_0, \boldsymbol{x}_T)+2\sigma^2 \text{KL}\left(\pi_{0,T} \| \pi_0 \otimes \pi_T\right)
\end{align}
where $\sigma \ge 0$ and $\Pi(\pi_0, \pi_T)$ denotes the set of couplings with marginals $\pi_0$ and $\pi_T$. Then, the unique optimal coupling $\pi^\star_{0,T}$ is Gaussian and satisfies
\begin{align}
    \pi^\star_{0,T} \sim \mathcal{N}\!\left(
\begin{bmatrix}
\boldsymbol{\mu}_0 \\
\boldsymbol{\mu}_T
\end{bmatrix}, \begin{bmatrix}
\boldsymbol{\Sigma}_0 & \boldsymbol{C}_\sigma \\
\boldsymbol{C}_\sigma^\top & \boldsymbol{\Sigma}_T\end{bmatrix}
\right)
\end{align}
where 
\begin{align}
\boldsymbol{C}_\sigma := \frac{1}{2}\left( \boldsymbol{\Sigma}_0^{1/2} \boldsymbol{D}_\sigma \boldsymbol{\Sigma}_0^{-1/2} - \sigma^2 \boldsymbol{I}_d \right),\quad \text{where}\quad \boldsymbol{D}_\sigma &:= \left(4 \boldsymbol{\Sigma}_0^{1/2} \boldsymbol{\Sigma}_T \boldsymbol{\Sigma}_0^{1/2} + \sigma^4 \boldsymbol{I}_d \right)^{1/2}
\end{align}
In particular, when $\sigma = 0$, the solution reduces to the classical Gaussian optimal transport coupling with quadratic transport cost.
\end{lemma}

\textit{Intuition.} While we omit the full proof\footnote{See \citet{janati2020entropic} for proof.}, we observe that when transporting between two Gaussian distributions under entropy regularization, the \textbf{optimal coupling remains Gaussian} and is fully characterized by how the two variables are correlated. Since the marginals are fixed to be $\pi_0$ and $\pi_T$, the only degree of freedom is the cross-covariance $\boldsymbol{C}_\sigma$, which determines how samples are paired. 

The objective balances two competing effects: the quadratic cost $\|\boldsymbol{x}_T - \boldsymbol{x}_0\|^2$ encourages pairs $(\boldsymbol{x}_0,\boldsymbol{x}_T)$ to be as close as possible, whereas the entropy regularization term $\text{KL}\left(\pi_{0,T} \| \pi_0 \otimes \pi_T\right)$ penalizes deviations from independence, pushing the coupling toward the product measure where $\boldsymbol{x}_0$ and $\boldsymbol{x}_T$ are uncorrelated. Therefore, the optimal cross-covariance $\boldsymbol{C}_\sigma$ can be interpreted as the optimal trade-off between these two objectives.

Given this result, we are ready to define the \boldtext{Gaussian Schrödinger Bridge} (SB) problem and derive its closed form solution.

\begin{definition}[Gaussian Schrödinger Bridge Problem]\label{def:gaussian-sbp}
Let $\pi_0 =\mathcal{N}(\boldsymbol{\mu}, \boldsymbol{\Sigma})$ and $\pi_T = \mathcal{N}(\boldsymbol{\mu}, \boldsymbol{\Sigma})$ be Gaussian probability measures on $\mathbb{R}^d$ where $\boldsymbol{\mu}, \boldsymbol{\mu}'\in \mathbb{R}^d$ are the mean vectors and $\boldsymbol{\Sigma}, \boldsymbol{\Sigma}'\in \mathbb{R}^{d\times d}$ are the covariances, and let $\mathbb{Q}$ be a reference path measure. The \textbf{Gaussian Schrödinger bridge} seeks the path measure that matches the Gaussian marginals while minimizing the relative entropy with respect to $\mathbb{Q}$:
\begin{small}
\begin{align}
    \mathbb{P}^\star=\underset{\mathbb{P}\in \mathcal{P}(C([0,T]; \mathbb{R}^d))}{\arg\min}\big\{\text{KL}(\mathbb{P}\|\mathbb{Q}):\pi_0 =\mathcal{N}(\boldsymbol{\mu}, \boldsymbol{\Sigma}),\pi_T=\mathcal{N}(\boldsymbol{\mu}', \boldsymbol{\Sigma}')\big\}\tag{Gaussian SB Problem}\label{eq:gaussian-sb-problem}
\end{align}
\end{small}
which can also be written in the form of a (\ref{eq:entropy-regularized-dynamic-ot})\footnote{to align notation with \citep{bunne2023schrodinger}, we denote the full drift $\boldsymbol{f}_\mathcal{N}(\boldsymbol{x},t)$ without scaling by $\sigma_t$, which results the additional factor of $\frac{\sigma_t^2}{4}$ in the Fisher information term.} as:
\begin{small}
\begin{align}
    \inf_{(p_t, \boldsymbol{v})}\int_0^T\mathbb{E}_{p_t}\left[\frac{1}{2}\|\boldsymbol{v}(\boldsymbol{x},t)\|^2+\frac{\sigma^4_t}{8}\|\nabla\log p_t(\boldsymbol{x})\|^2\right]dt\quad \text{s.t.}\quad \begin{cases}
        \partial_tp_t=-\nabla \cdot(p_t\boldsymbol{v})\\
        p_0=\mathcal{N}(\boldsymbol{\mu},\boldsymbol{\Sigma}), \quad p_T=\mathcal{N}(\boldsymbol{\mu}',\boldsymbol{\Sigma}')
    \end{cases}
\end{align}
\end{small}
where we set $\boldsymbol{f}\equiv 0$.
\end{definition}

To obtain a tractable characterization of this problem, we exploit the special structure of Gaussian measures, which are uniquely defined by their mean and covariance. When the marginals remain Gaussian along the interpolation, the evolution of the process is fully characterized by the trajectories of its \textbf{mean} $\boldsymbol{\mu}_t\in \mathbb{R}^d$ and \textbf{covariance} $\boldsymbol{\Sigma }_t\in \mathbb{R}^{d\times d}$. In this setting, the Gaussian Schrödinger bridge can be interpreted as minimizing the energy of the change in covariance matrices as they move along the manifold of symmetric positive definite matrices $\boldsymbol{\Sigma}\in\mathbb{S}_{++}^d$, whose \textbf{tangent space} is the space of symmetric matrices, whose \textbf{tangent space} is the space of symmetric matrices:

\begin{align}
    \mathcal{T}_{\boldsymbol{\Sigma}}\mathbb{S}^d_{++}:=\{\boldsymbol{U}\in \mathbb{R}^{d\times d}: \boldsymbol{U}^\top =\boldsymbol{U}\}
\end{align}

The natural geometry governing optimal transport between Gaussian covariances is called the \boldtext{Bures-Wasserstein manifold} \citep{takatsu2010wasserstein, bhatia2019bures}. This Riemannian manifold defines the metric structure in the space of covariance matrices induced by the Wasserstein distance between Gaussian distributions, which characterizes optimality of transport between covariances. The geometry of this manifold can be described using the \boldtext{Lyapunov operator}, which we define below.

\begin{definition}[Lyapunov Operator and Bures-Wasserstein Manifold]\label{def:bures-wasserstein}
    Given a covariance matrix $\boldsymbol{\Sigma}\in \mathbb{S}^d_{++}$ and tangent matrix $\boldsymbol{U}\in \mathcal{T}_{\boldsymbol{\Sigma}}\mathbb{S}^d_{++}$, the Lyapunov operator $\mathcal{L}_{\boldsymbol{\Sigma}}[\boldsymbol{U}]: \mathcal{T}_{\boldsymbol{\Sigma}}\mathbb{S}^d_{++}\to \mathbb{S}^d_{++}$ is the operator that returns the unique symmetric matrix  $\boldsymbol{A}$ that solves:
    \begin{align}
        \boldsymbol{\Sigma}\boldsymbol{A}+\boldsymbol{A}\boldsymbol{\Sigma}=\boldsymbol{U}
    \end{align}
    which defines the Riemannian metric of the Bures-Wasserstein manifold given by:
    \begin{align}
        \langle \boldsymbol{U}, \boldsymbol{V}\rangle_{\boldsymbol{\Sigma}}:=\frac{1}{2}\text{Tr}(\mathcal{L}_{\boldsymbol{\Sigma}}[\boldsymbol{U}]\boldsymbol{V}), \quad \boldsymbol{U}, \boldsymbol{V}\in \mathcal{T}_{\boldsymbol{\Sigma}}\mathbb{S}^d_{++}\label{eq:bw-metric}
    \end{align}
    where $\boldsymbol{U}, \boldsymbol{V}\in \mathcal{T}_{\boldsymbol{\Sigma}}\mathbb{S}^d_{++}$ are two tangent vectors in the tangent space of the covariance matrices. 
\end{definition}

Using the Lyapunov operator, the (\ref{eq:gaussian-sb-problem}) can be reformulated as an \textbf{action minimization problem} on the Bures-Wasserstein manifold of covariance matrices. This geometric formulation reveals that the optimal covariance trajectory evolves along curves minimizing a kinetic energy functional augmented by a diffusion correction term.

\begin{figure}
    \centering
    \includegraphics[width=\linewidth]{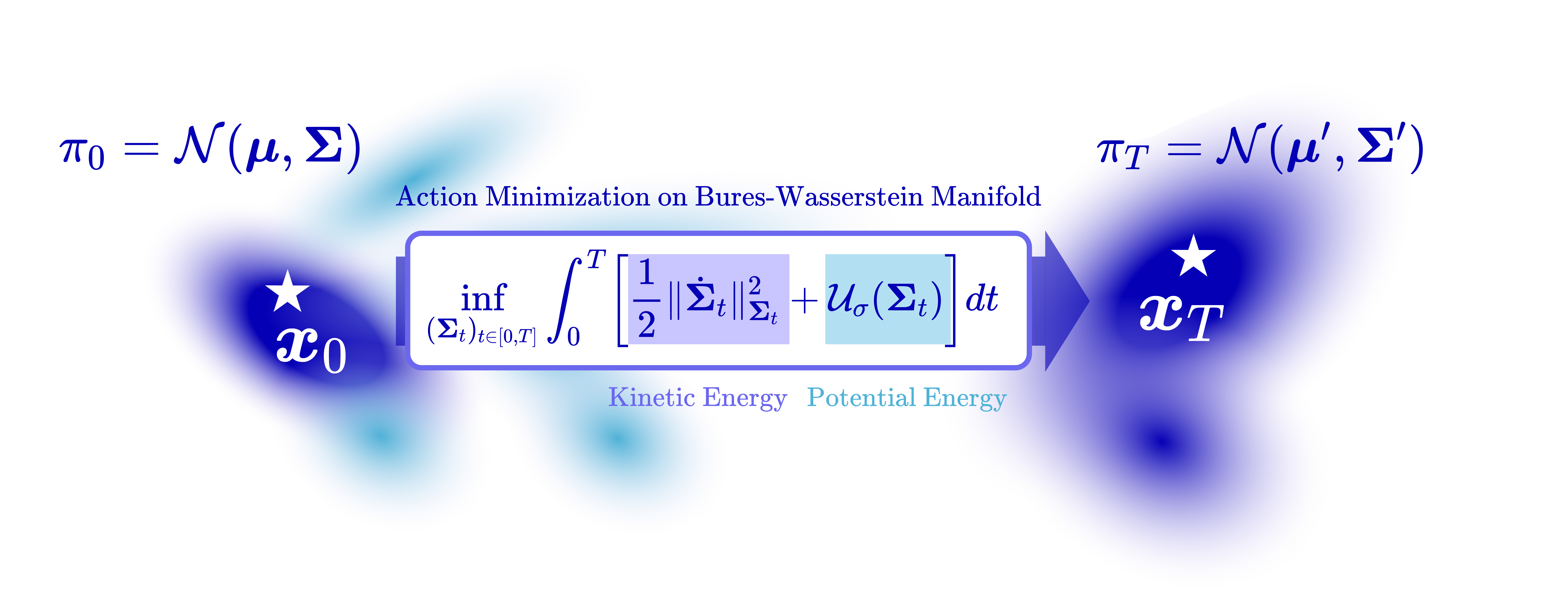}
    \caption{\textbf{Gaussian Schrödinger Bridge Problem.}  The Gaussian SB between initial and terminal Gaussian distributions $\pi_0=\mathcal{N}(\mu,\Sigma)$ and $\pi_T=\mathcal{N}(\mu',\Sigma')$ reduces to solving an action minimization problem on the Bures--Wasserstein manifold of covariance matrices, where the covariance trajectory $(\Sigma_t)_{t\in[0,T]}$ balances a kinetic energy term describing the rate of covariance change and a potential energy term induced by the diffusion of the reference process.}
    \label{fig:gaussian}
\end{figure}

\begin{proposition}[Gaussian Schrödinger Bridge Problem as Action Minimization (Theorem 2 of \citep{bunne2023schrodinger})]
    The solution to the (\ref{eq:gaussian-sb-problem}) where $\mathbb{Q}$ is defined as pure Brownian motion $\mathbb{Q}:d\boldsymbol{X}_t=\sigma_td\boldsymbol{B}_t$ is equivalent to the solution of the \textbf{action minimization} problem on the Bures-Wasserstein manifold, defined as: 
    \begin{align}
        \inf_{(\boldsymbol{\Sigma}_t)_{t\in [0,T]}}\int_0^T\left[\frac{1}{2}\|\dot{\boldsymbol{\Sigma}}_t\|^2_{\boldsymbol{\Sigma}_t}+\mathcal{U}_\sigma(\boldsymbol{\Sigma}_t)\right]dt\quad\text{s.t.}\quad \begin{cases}
            \boldsymbol{\Sigma}_0=\boldsymbol{\Sigma}\\
            \boldsymbol{\Sigma}_T=\boldsymbol{\Sigma}'
        \end{cases}\label{eq:gaussian-action-functional}
    \end{align}
    where $\mathcal{U}_\sigma(\boldsymbol{\Sigma}_t):= \frac{\sigma_t^4}{8}\text{Tr}\left(\boldsymbol{\Sigma}_t^{-1}\right)$ is the potential energy that captures the entropic contribution of the diffusion term and $\|\dot{\boldsymbol{\Sigma}}_t\|^2_{\boldsymbol{\Sigma}_t}=\langle \dot{\boldsymbol{\Sigma}}_t, \dot{\boldsymbol{\Sigma}}_t\rangle_{\boldsymbol{\Sigma}}$ is the Riemmannian metric defined in (\ref{eq:bw-metric}). Furthermore, the solution satisfies the \textbf{Euler-Lagrange equation} in Bures-Wasserstein geometry:
    \begin{align}
    \begin{cases}
        \nabla_{\dot{\boldsymbol{\Sigma}}_t}\dot{\boldsymbol{\Sigma}}_t=-\text{grad}\left(-\frac{\sigma_t^4}{8}\text{Tr}\boldsymbol{\Sigma}_t^{-1}\right)\\
        \boldsymbol{\Sigma}_0=\boldsymbol{\Sigma}, \quad \boldsymbol{\Sigma}_T=\boldsymbol{\Sigma}'
    \end{cases}\tag{Euler-Lagrange Equation}\label{eq:euler-lagrange-bures}
    \end{align}
    where $\nabla_{\dot{\boldsymbol{\Sigma}}_t}\dot{\boldsymbol{\Sigma}}_t$ is the acceleration defined by the Riemannian gradient in the Bures-Wasserstein geometry.
\end{proposition}

\textit{Proof Sketch.} While we don't prove all required lemmas rigorously, we break down the key components of this proof in three steps\footnote{for rigorous proof, see \citep{bunne2023schrodinger} Appendix B.2}.

\textbf{Step 1: Define the Lyapunov Operator. }
A \textbf{key component} of this proof is that the Lyapunov operator $\mathcal{L}_{\boldsymbol{\Sigma}_t}[\dot{\boldsymbol{\Sigma}}_t]$ of the \textit{velocity of the covariance matrix along the curve $\boldsymbol{\Sigma}_t$} takes the form:
\begin{small}
\begin{align}
    \mathcal{L}_{\boldsymbol{\Sigma}_t}[\dot{\boldsymbol{\Sigma}}_t]=\bluetext{\tilde{\boldsymbol{S}}_t}^\top\boldsymbol{\Sigma}_t^{-1}
\end{align}
\end{small}
where $\tilde{\boldsymbol{S}}_t\in \mathbb{R}^{d\times d}$ can be interpreted as the linear velocity field that transports the Gaussian distribution defined as:
\begin{small}
\begin{align}
    \tilde{\boldsymbol{S}}_t:=t\boldsymbol{\Sigma}'+(T-t)\boldsymbol{C}_\sigma-(T-t)\boldsymbol{\Sigma}-t\boldsymbol{C}_\sigma^\top+\frac{\sigma_t}{2}(T-2t)\boldsymbol{I}_d
\end{align}
\end{small}
Recall from Definition \ref{def:bures-wasserstein} that the Lyapunov operator $\mathcal{L}_{\boldsymbol{\Sigma}_t}[\dot{\boldsymbol{\Sigma}}_t]$ is the symmetric matrix $\boldsymbol{A}$ that solves $\boldsymbol{\Sigma}_t\boldsymbol{A}+\boldsymbol{A}\boldsymbol{\Sigma}_t=\dot{\boldsymbol{\Sigma}}_t$ and defines the Riemannian metric $\|\dot{\boldsymbol{\Sigma}}_t\|^2_{\boldsymbol{\Sigma}_t}=\text{Tr}(\boldsymbol{A}\boldsymbol{\Sigma}_t\boldsymbol{A})$. Therefore, kinetic energy term appearing in the action functional can be written as:
\begin{small}
\begin{align}
    \frac{1}{2}\|\dot{\boldsymbol{\Sigma}}_t\|^2_{\boldsymbol{\Sigma}_t}=\frac{1}{2}\text{Tr}\left(\bluetext{\mathcal{L}_{\boldsymbol{\Sigma}_t}[\dot{\boldsymbol{\Sigma}}_t]}\boldsymbol{\Sigma}_t\bluetext{\mathcal{L}_{\boldsymbol{\Sigma}_t}[\dot{\boldsymbol{\Sigma}}_t]}\right)=\frac{1}{2}\text{Tr}\big(\bluetext{\tilde{\boldsymbol{S}}_t^\top}\underbrace{\bluetext{\boldsymbol{\Sigma}_t^{-1}}\boldsymbol{\Sigma}_t}_{\boldsymbol{I}}\bluetext{\boldsymbol{\Sigma}_t^{-1}}\tilde{\boldsymbol{S}}_t\big)=\frac{1}{2}\text{Tr}\big(\tilde{\boldsymbol{S}}_t^\top\bluetext{\boldsymbol{\Sigma}_t^{-1}}\tilde{\boldsymbol{S}}_t\big)
\end{align}
\end{small}
Substituting this expression for the kinetic energy into the integrand of the action functional in (\ref{eq:gaussian-action-functional}) yields:
\begin{small}
\begin{align}
    \frac{1}{2}\|\dot{\boldsymbol{\Sigma}}_t\|^2_{\boldsymbol{\Sigma}_t}+\mathcal{U}_\sigma(\boldsymbol{\Sigma}_t)=\frac{1}{2}\text{Tr}\big(\tilde{\boldsymbol{S}}_t^\top\bluetext{\boldsymbol{\Sigma}_t^{-1}}\tilde{\boldsymbol{S}}_t\big)+\frac{\sigma_t^4}{8}\text{Tr}\left(\boldsymbol{\Sigma}_t^{-1}\right)=\text{Tr}\left(\frac{1}{2}\tilde{\boldsymbol{S}}_t^\top\boldsymbol{\Sigma}_t^{-1}\tilde{\boldsymbol{S}}_t+\frac{\sigma_t^4}{8}\boldsymbol{\Sigma}_t^{-1}\right)\label{eq:integrand-action-func}
\end{align}
\end{small}
which yields an expression to the action functional written entirely in terms of the matrix $\tilde{\boldsymbol{S}}_t$ and the inverse covariance $\boldsymbol{\Sigma}_t^{-1}$.

\textbf{Step 2: Use Continuity Equation to Derive Optimal Velocity. }
To finish the proof, we aim to show that the action minimization problem in (\ref{eq:gaussian-action-functional}) is equivalent to the (\ref{eq:gaussian-sb-problem}) with $\mathbb{Q}:=\sigma_t\mathbb{B}$. To do this, we consider the special form of the (\ref{eq:gaussian-sb-problem}) with zero-mean marginals\footnote{since the mean follows the straight-line interpolation between the terminal marginals, we focus our attention on the covariance dynamics.}:
\begin{small}
\begin{align}
    \inf_{(p_t, \boldsymbol{v})}\int_0^T\mathbb{E}_{p_t}\left[\frac{1}{2}\|\boldsymbol{v}(\boldsymbol{x},t)\|^2+\frac{\sigma^4}{8}\|\nabla\log p_t(\boldsymbol{x})\|^2\right]dt\quad \text{s.t.}\quad \begin{cases}
        \partial_tp_t=-\nabla \cdot(p_t\boldsymbol{v})\\
        p_0=\mathcal{N}(\boldsymbol{0},\boldsymbol{\Sigma}), \quad p_T=\mathcal{N}(\boldsymbol{0},\boldsymbol{\Sigma}')
    \end{cases}\label{eq:gaus-alt-objective}
\end{align}
\end{small}
Then, the solution yields a Gaussian density $p_t(\boldsymbol{x})=\mathcal{N}(\boldsymbol{0}, \boldsymbol{\Sigma}_t)$ with zero-mean and covariance $\boldsymbol{\Sigma}_t$ defined as:
\begin{small}
\begin{align}
    p_t(\boldsymbol{x})=(2\pi)^{-\frac{1}{d}}(\text{det}\boldsymbol{\Sigma}_t)^{-\frac{1}{2}}\exp\left(-\frac{1}{2}\boldsymbol{x}^\top\boldsymbol{\Sigma}_t^{-1}\boldsymbol{x}\right)
\end{align}
\end{small}
To derive the form of the optimal velocity $\boldsymbol{v}$, we can expand both sides of the continuity constraint given by:
\begin{small}
\begin{align}
    \underbrace{\partial_tp_t}_{(\bigstar)}=\underbrace{-\nabla \cdot(p_t\boldsymbol{v})}_{(\diamond)}\label{eq:gaussian-continutiy-proof}
\end{align}
\end{small}
For $(\bigstar)$, we differentiate the Gaussian form of $p_t(\boldsymbol{x})$ with respect to time using the Jacobi formula\footnote{which states for any matrix valued function $\boldsymbol{A}_t:\mathbb{R}_{\geq 0}\to \mathbb{R}^{d\times d}$, the time derivative of its determinant is equal to $\frac{d}{dt}\text{det}(\boldsymbol{A}_t)=\text{det}(\boldsymbol{A}_t)\cdot \text{Tr}(\boldsymbol{A}_t^{-1})\cdot\frac{d}{dt}\boldsymbol{A}_t$}, we get:
\begin{small}
\begin{align}
    \partial_t p_t(\boldsymbol{x})= p_t(\boldsymbol{x}) \left(\frac{1}{2}\boldsymbol{x}^\top\boldsymbol{\Sigma}^{-1}_t\dot{\boldsymbol{\Sigma}}_t\boldsymbol{\Sigma}_t^{-1}\boldsymbol{x}-\frac{1}{2}\text{Tr}(\boldsymbol{\Sigma}_t^{-1}\dot{\boldsymbol{\Sigma}}_t)\right)\tag{$\bigstar$}
\end{align}
\end{small}
For $(\diamond)$, we use the chain rule of divergence and the gradient of the log-density $\nabla \log p_t(\boldsymbol{x})=-\boldsymbol{\Sigma}_t^{-1}\boldsymbol{x}$ to get:
\begin{small}
\begin{align}
    \nabla \cdot(p_t\boldsymbol{v})&=\langle \bluetext{\nabla p_t}, \boldsymbol{v}\rangle+p_t(\nabla \cdot \boldsymbol{v})=\langle \bluetext{p_t\nabla \log p_t}, \boldsymbol{v}\rangle+p_t(\nabla \cdot \boldsymbol{v})\nonumber\\
    &=\langle \bluetext{p_t(-\boldsymbol{\Sigma}_t^{-1}\boldsymbol{x})}, \boldsymbol{v}\rangle+p_t(\nabla \cdot \boldsymbol{v})=p_t\left(\langle -\boldsymbol{\Sigma}_t^{-1}\boldsymbol{x}, \boldsymbol{v}\rangle+\nabla \cdot \boldsymbol{v}\right)\tag{$\diamond$}
\end{align}
\end{small}
Plugging $(\bigstar)$ and $(\diamond)$ back into (\ref{eq:gaussian-continutiy-proof}), we have: 
\begin{small}
\begin{align}
    &p_t(\boldsymbol{x}) \left(\frac{1}{2}\boldsymbol{x}^\top\boldsymbol{\Sigma}^{-1}_t\dot{\boldsymbol{\Sigma}}_t\boldsymbol{\Sigma}_t^{-1}\boldsymbol{x}-\frac{1}{2}\text{Tr}(\boldsymbol{\Sigma}_t^{-1}\dot{\boldsymbol{\Sigma}}_t)\right)=p_t(\boldsymbol{x})\left(\langle \boldsymbol{\Sigma}_t^{-1}\boldsymbol{x}, \boldsymbol{v}(\boldsymbol{x},t)\rangle-\nabla \cdot \boldsymbol{v}(\boldsymbol{x},t)\right)\nonumber\\
    &\implies \begin{cases}
        \nabla \cdot \boldsymbol{v}(\boldsymbol{x},t)=\frac{1}{2}\text{Tr}(\boldsymbol{\Sigma}_t^{-1}\dot{\boldsymbol{\Sigma}}_t)\\
        \langle \boldsymbol{\Sigma}_t^{-1}\boldsymbol{x}, \boldsymbol{v}(\boldsymbol{x},t)\rangle=\frac{1}{2}\langle \boldsymbol{\Sigma}^{-1}_t\boldsymbol{x}, \dot{\boldsymbol{\Sigma}}_t\boldsymbol{\Sigma}_t^{-1}\boldsymbol{x}\rangle
    \end{cases}
\end{align}
\end{small}
which means that the optimal velocity field must be linear in $\boldsymbol{x}$ with the form:
\begin{small}
\begin{align}
    \boldsymbol{v}(\boldsymbol{x},t)=\tilde{\boldsymbol{S}}_t\boldsymbol{\Sigma}^{-1}_t\boldsymbol{x}\label{eq:gaussian-proof-velocity-field}
\end{align}
\end{small}

\textbf{Step 3: Prove Equivalence of the Action Minimization and Gaussian SB Problem. }
Using the linear form of the velocity field in (\ref{eq:gaussian-proof-velocity-field}), the expectations appearing in the objective (\ref{eq:gaus-alt-objective}) can now be computed explicitly:
\begin{small}
\begin{align}
    \frac{1}{2}\underbrace{\mathbb{E}[\|\boldsymbol{v}(\boldsymbol{x},t)\|^2]}_{(\bigstar)}+\frac{\sigma^4}{8}\underbrace{\mathbb{E}[\|\nabla\log p_t(\boldsymbol{x})\|^2]}_{(\diamond)}\label{eq:gaus-proof-eq1}
\end{align}
\end{small}
Since $\boldsymbol{x}\sim \mathcal{N}(\boldsymbol{0},\boldsymbol{\Sigma}_t)$, we have $\mathbb{E}[\boldsymbol{x}\boldsymbol{x}^\top ] =\boldsymbol{\Sigma}_t$. Using this and the velocity field from (\ref{eq:gaussian-proof-velocity-field}), the kinetic energy term $(\bigstar)$ can be written as:
\begin{small}
\begin{align}
    \mathbb{E}[\|\boldsymbol{v}(\boldsymbol{x},t)\|^2]&=\mathbb{E}\left[\text{Tr}\left(\tilde{\boldsymbol{S}}^\top_t\boldsymbol{\Sigma}^{-1}_t\boldsymbol{x}\boldsymbol{x}^\top\boldsymbol{\Sigma}^{-1}_t\tilde{\boldsymbol{S}}_t\right)\right]=\text{Tr}\bigg(\tilde{\boldsymbol{S}}^\top_t\boldsymbol{\Sigma}^{-1}_t\bluetext{\underbrace{\mathbb{E}\left[\boldsymbol{x}\boldsymbol{x}^\top\right]}_{\boldsymbol{\Sigma}_t}}\boldsymbol{\Sigma}^{-1}_t\tilde{\boldsymbol{S}}_t\bigg)\nonumber\\
    &=\text{Tr}\bigg(\tilde{\boldsymbol{S}}^\top_t\underbrace{\boldsymbol{\Sigma}^{-1}_t\bluetext{\boldsymbol{\Sigma}_t}}_{=\boldsymbol{I}}\boldsymbol{\Sigma}^{-1}_t\tilde{\boldsymbol{S}}_t\bigg) = \text{Tr}\big(\tilde{\boldsymbol{S}}^\top_t\boldsymbol{\Sigma}^{-1}_t\tilde{\boldsymbol{S}}_t\big)
\end{align}
\end{small}
Similarly, using $\nabla \log p_t(\boldsymbol{x})=-\boldsymbol{\Sigma}_t^{-1}\boldsymbol{x}$ and $\mathbb{E}[\boldsymbol{x}\boldsymbol{x}^\top ] =\boldsymbol{\Sigma}_t\implies \mathbb{E}[\boldsymbol{x}^\top \boldsymbol{A}\boldsymbol{x} ] =\text{Tr}(\boldsymbol{A}\boldsymbol{\Sigma}_t)$, the score term $(\diamond)$ reduces to:
\begin{small}
\begin{align}
    \mathbb{E}\left[\|\nabla\log p_t(\boldsymbol{x}) \|^2\right]&=\mathbb{E}\left[\|\bluetext{\boldsymbol{x}^\top\boldsymbol{\Sigma}_t^{-2}\boldsymbol{x}}\|^2\right]=\text{Tr}(\boldsymbol{\Sigma}_t^{-2}\boldsymbol{\Sigma}_t)=\text{Tr}(\boldsymbol{\Sigma}_t^{-1})
\end{align}
\end{small}
Plugging $(\bigstar)$ and $(\diamond)$ back into (\ref{eq:gaus-proof-eq1}), we get:
\begin{small}
\begin{align}
    \frac{1}{2}\underbrace{\text{Tr}\big(\tilde{\boldsymbol{S}}^\top_t\boldsymbol{\Sigma}^{-1}_t\tilde{\boldsymbol{S}}_t\big)}_{(\bigstar)}+\frac{\sigma^4}{8}\underbrace{\text{Tr}(\boldsymbol{\Sigma}_t^{-1})}_{(\diamond)}=\text{Tr}\left(\frac{1}{2}\tilde{\boldsymbol{S}}^\top_t\boldsymbol{\Sigma}^{-1}_t\tilde{\boldsymbol{S}}_t+\frac{\sigma_t^4}{8}\boldsymbol{\Sigma}_t^{-1}\right)
\end{align}
\end{small}
which exactly aligns with the form of the action function derived in (\ref{eq:integrand-action-func}), and we have shown that solving the action minimization problem is equivalent to solving the (\ref{eq:gaussian-sb-problem}).\hfill $\square$

This result shows that the infinite-dimensional \ref{eq:gaussian-sb-problem}) reduces to a finite-dimensional action functional defined entirely on the trajectory of covariance matrices. Intuitively, we have shown that the optimal evolution of the covariance follows a curve on the manifold of symmetric positive definite matrices $\mathbb{S}_{++}^d$ where the first term measures the \textbf{kinetic energy of the covariance transport in the Bures–Wasserstein geometry}, and the second term acts as a potential induced by the diffusion term. The Schrödinger bridge, therefore, corresponds to the curve that minimizes this action while matching the endpoint covariances $\boldsymbol{\Sigma}_0=\boldsymbol{\Sigma}$ and $\boldsymbol{\Sigma}_T=\boldsymbol{\Sigma}'$. Next, we show that the solution for both problems yields a \textbf{closed-form solution} which can be written explicitly as an SDE.

\begin{theorem}[Closed-Form Solution to Gaussian SB problem (Theorem 3 in \citet{bunne2023schrodinger})]\label{thm:closed-form-gaussiansbp}
    The solution to the Gaussian SB problem $\mathbb{P}^\star$ with linear reference measure $\mathbb{Q}$ defined with the SDE $d\boldsymbol{X}^{\mathbb{Q}}_t=(c_t\boldsymbol{X}_t+\boldsymbol{\alpha}_t)dt+\sigma_td\boldsymbol{B}_t$ is itself a Gaussian Markov stochastic process $\boldsymbol{X}_{0:T}$ where the intermediate marginals are Gaussians $p_t=\mathcal{N}(\boldsymbol{\mu}_t, \boldsymbol{\Sigma}_t)$, with mean and covariance defined by:
    \begin{align}
        &\begin{cases}
            \boldsymbol{\mu}^\star_t&=\bar{r}_t\boldsymbol{\mu}_0+r_t\boldsymbol{\mu}_T+\boldsymbol{\zeta}(t)-r_t\boldsymbol{\zeta}(T)\\
        \boldsymbol{\Sigma}^\star_t&=\bar{r}_t^2\boldsymbol{\Sigma}_0+r_t^2\boldsymbol{\Sigma}_T+r_t\bar{r}_t(\boldsymbol{C}_{\sigma_\star}+\boldsymbol{C}_{\sigma^\star}^\top)+\kappa(t,t)(1-\rho_t)\boldsymbol{I}
        \end{cases}\label{eq:gaussian-sb-meancov}\\
        &\text{s.t.}\quad\begin{cases}
            r_t:=\frac{\kappa(t,T)}{\kappa(T,T)}, \quad r_t:=\tau_t-r_t\tau_T, \quad \sigma_\star:=\sqrt{\tau_T^{-1}\kappa(T,T)}\\
            \boldsymbol{\zeta}(t):=\tau_t\int_0^t\tau_s^{-1}\boldsymbol{\alpha}_sds, \quad \rho_t:=\frac{\int_0^t\tau_s^{-2}\sigma_s^2ds}{\int_0^T\tau_s^{-2}\sigma_s^2ds}
        \end{cases}\nonumber
    \end{align}
    The time evolution of $\boldsymbol{X}_t$ follows a closed-form SDE:
    \begin{small}
    \begin{align}
        d\boldsymbol{X}_t=\boldsymbol{f}_{\mathcal{N}}(\boldsymbol{X}_t,t)dt+\sigma_td\boldsymbol{B}_t, \quad \text{s.t.}\quad \begin{cases}
            \boldsymbol{f}_{\mathcal{N}}(\boldsymbol{x},t):=\boldsymbol{S}_t^\top\boldsymbol{\Sigma}_t^{-1}(\boldsymbol{x}-\boldsymbol{\mu}_t)+\dot{\boldsymbol{\mu}}_t\\
            \boldsymbol{P}_t:=\dot{r}_t(r_t\boldsymbol{\Sigma}_T+\bar{r}_t\boldsymbol{C}_{\sigma_\star})\\
            \boldsymbol{Q}_t:=-\dot{\bar{r}}_t(\bar{r}_t\boldsymbol{\Sigma}_0+r_t\boldsymbol{C}_{\sigma_\star})\\
            \boldsymbol{S}_t:=\boldsymbol{P}_t-\boldsymbol{Q}_t^\top+(c_t\kappa(t,t)(1-\rho_t)-\sigma_t^2\rho_t)\boldsymbol{I}
        \end{cases}
    \end{align}
    \end{small}
    where the matrix $\boldsymbol{S}_t^\top \boldsymbol{\Sigma}_t^{-1}$ is symmetric.
\end{theorem}

\textit{Proof.} Following \citet{bunne2023schrodinger}, we break down the proof in several steps. 

\textbf{Step 1: Solving the Static Gaussian SB problem.} 
First, we will solve the \textbf{static Gaussian SB} defined as the minimization problem:
\begin{align}
    \min_{\mathbb{P}_{0,T}}\bigg\{\text{KL}(\mathbb{P}_{0,T}\|\mathbb{Q}_{0,T}):p_0=\mathcal{N}(\boldsymbol{\mu}_0, \boldsymbol{\Sigma}_0), p_T=\mathcal{N}(\boldsymbol{\mu}_T, \boldsymbol{\Sigma}_T)\bigg\}
\end{align}
where $\mathbb{Q}_{0,T}$ is the endpoint law of the reference Gaussian process. By definition of linear SDEs of the form $d\boldsymbol{X}_t=(c_t\boldsymbol{X}_t+\boldsymbol{\alpha}_t)dt+\sigma_td\boldsymbol{B}_t$, the intermediate state $\boldsymbol{X}_t$ can be written as \citep{platen2010numerical}:
\begin{align}
    \boldsymbol{X}_t=\tau_t\left(\boldsymbol{X}_0+\int_0^t\tau_s^{-1}\boldsymbol{\alpha}_sds+\int_0^t\tau_s^{-1}\sigma_td\boldsymbol{B}_t\right), \quad \tau_t:=\exp\left(\int_0^tc_sds\right)\label{eq:linear-sde-sol}
\end{align}
Conditioned on an initial state $\boldsymbol{X}_0\sim \mathcal{N}(\boldsymbol{\mu}_0, \boldsymbol{\Sigma}_0)$, the mean and covariance of $\boldsymbol{X}_t$ takes the form:
\begin{align}
    \boldsymbol{\eta}(t)&:=\mathbb{E}[\boldsymbol{X}_t|\boldsymbol{X}_0]=\tau_t\left(\boldsymbol{X}_0+\int_0^t\tau_s^{-1}\boldsymbol{\alpha}_sds\right)\label{eq:guassian-sde-mean}\\
    \boldsymbol{\kappa}(t, t')&:=\mathbb{E}\left[(\boldsymbol{X}_t-\boldsymbol{\eta}(t))(\boldsymbol{X}_t-\boldsymbol{\eta}(t'))^{\top}\big|\boldsymbol{X}_0\right]= \left(\tau_t\tau_{t'}\int_0^t\tau_s^{-2}\sigma_s^2ds\right)\boldsymbol{I}\label{eq:guassian-sde-cov}
\end{align}
Therefore, under the linear Gaussian measure $\mathbb{Q}$, the probability of the pair $(\boldsymbol{X}_0, \boldsymbol{X}_T)$ is a Gaussian defined by:
\begin{align}
    \mathbb{Q}(\boldsymbol{X}_T=\boldsymbol{x}_T|\boldsymbol{X}_0=\boldsymbol{x}_0)&=\mathcal{N}(\boldsymbol{\eta}(T), \boldsymbol{\kappa}(T,T)\boldsymbol{I})\nonumber\\
    &=(2\pi)^{-\frac{d}{2}}\text{det}(\boldsymbol{\kappa}(T,T)\boldsymbol{I})^{-\frac{1}{2}}\exp\left(-\frac{1}{2}(\boldsymbol{x}_T-\boldsymbol{\eta}(T))(\kappa(T,T)\boldsymbol{I})^{-1}(\boldsymbol{x}_T-\boldsymbol{\eta}(T))\right)\nonumber\\
    &=(2\pi)^{-\frac{d}{2}}\text{det}(\boldsymbol{\kappa}(T,T)\boldsymbol{I})^{-\frac{1}{2}}\exp\left(-\frac{1}{2\boldsymbol{\kappa}(T,T)}\|\boldsymbol{x}_T-\tau_T\boldsymbol{x}_0-\boldsymbol{\zeta}(T)\|^2\right)\label{eq:q-coupling}
\end{align}
where the third equality follows from setting $\eta(T):=\tau_T\boldsymbol{x}_0+\tau_T\int_0^T\tau_s^{-1}\boldsymbol{\alpha}_sds=\tau_T\boldsymbol{x}_0+\boldsymbol{\zeta}(T)$ and the fact that the covariance $\kappa(T,T)\boldsymbol{I}$ is isotropic. Now, we can expand the KL divergence and substitute (\ref{eq:q-coupling}) for $\mathbb{Q}_{0,T}$ to get:
\begin{small}
\begin{align}
    &\text{KL}(\mathbb{P}_{0,T}\|\mathbb{Q}_{0,T})=\int_{\mathbb{R}^d\times \mathbb{R}^d}\log \frac{\mathrm{d}\mathbb{P}_{0,T}}{\mathrm{d}\mathbb{Q}_{0,T}}\mathrm{d}\mathbb{P}_{0,T}\nonumber\\
    &= \int_{\mathbb{R}^d\times \mathbb{R}^d}(\log\mathrm{d}\mathbb{P}_{0,T}) \mathrm{d}\mathbb{P}_{0,T}-\int_{\mathbb{R}^d\times \mathbb{R}^d}(\log\mathrm{d}\mathbb{Q}_{0,T}) \mathrm{d}\mathbb{P}_{0,T}\nonumber\\
    &=\int_{\mathbb{R}^d\times \mathbb{R}^d}(\log\mathrm{d}\mathbb{P}_{0,T}) \mathrm{d}\mathbb{P}_{0,T}-\int_{\mathbb{R}^d\times \mathbb{R}^d}\log\left((2\pi)^{-\frac{d}{2}}\text{det}(\kappa(T,T)\boldsymbol{I})^{-\frac{1}{2}}\exp\left(-\frac{1}{2\kappa(T,T)}\|\boldsymbol{x}_T-\tau_T\boldsymbol{x}_0-\boldsymbol{\zeta}(T)\|^2\right)\right) \mathrm{d}\mathbb{P}_{0,T}(\boldsymbol{x}_0, \boldsymbol{x}_T)\nonumber\\
    &=\int_{\mathbb{R}^d\times \mathbb{R}^d}(\log\mathrm{d}\mathbb{P}_{0,T}) \mathrm{d}\mathbb{P}_{0,T}-\int_{\mathbb{R}^d\times \mathbb{R}^d}\left(-\frac{d}{2}\log (2\pi)-\frac{1}{2}\text{det}(\kappa(T,T)\boldsymbol{I})-\frac{1}{2\kappa(T,T)}\|\boldsymbol{x}_T-\tau_T\boldsymbol{x}_0-\boldsymbol{\zeta}(T)\|^2\right) \mathrm{d}\mathbb{P}_{0,T}(\boldsymbol{x}_0, \boldsymbol{x}_T)\nonumber\\
    &=\int_{\mathbb{R}^d\times \mathbb{R}^d}(\log\mathrm{d}\mathbb{P}_{0,T}) \mathrm{d}\mathbb{P}_{0,T}+\underbrace{\frac{d}{2}\log (2\pi)+\frac{1}{2}\text{det}(\kappa(T,T)\boldsymbol{I})}_{\text{constant w.r.t. }\mathbb{P}_{0,T}}+\frac{1}{2\kappa(T,T)}\int_{\mathbb{R}^d\times \mathbb{R}^d}\|\boldsymbol{x}_T-\tau_T\boldsymbol{x}_0-\boldsymbol{\zeta}(T)\|^2\mathrm{d}\mathbb{P}_{0,T}(\boldsymbol{x}_0, \boldsymbol{x}_T)\nonumber\\
    &=\int_{\mathbb{R}^d\times \mathbb{R}^d}(\log\mathrm{d}\mathbb{P}_{0,T}) \mathrm{d}\mathbb{P}_{0,T}+\frac{1}{2\kappa(T,T)}\int_{\mathbb{R}^d\times \mathbb{R}^d}\|\boldsymbol{x}_T-\tau_T\boldsymbol{x}_0-\boldsymbol{\zeta}(T)\|^2\mathrm{d}\mathbb{P}_{0,T}(\boldsymbol{x}_0, \boldsymbol{x}_T)+\text{const.}\label{eq:gaussian-kl-objective}
\end{align}
\end{small}
Now, we aim to simplify this objective such that rather than matching $\boldsymbol{X}_T$ to $\tau_T\boldsymbol{X}_0+\boldsymbol{\zeta}(T)$ over $\mathbb{P}_{0,T}$, where $\boldsymbol{X}_0, \boldsymbol{X}_T$ are sampled from two disjoint marginals, we can directly match the initial marginal transformed to time $T$ to the target marginal at time $T$. Concretely, we define a transformed marginal $\tilde{\mathbb{P}}_0$ defined by applying a change-of-variables:
\begin{align}
    \tilde{\boldsymbol{X}}_0=\tau_T\boldsymbol{X}_0+\boldsymbol{\zeta}(T)\sim \mathcal{N}(\tilde{\boldsymbol{\mu}}_0, \tilde{\boldsymbol{\Sigma}}_0)\quad \text{s.t.}\quad \begin{cases}
        \tilde{\boldsymbol{\mu}}_0=\tau_T\boldsymbol{\mu}_0+\boldsymbol{\zeta}(T)\\
        \tilde{\boldsymbol{\Sigma}}_0=\tau_T^2\boldsymbol{\Sigma}_0
    \end{cases}\label{eq:change-of-var-gsb}
\end{align}
which yields the joint distribution $\tilde{\mathbb{P}}_{0,T}$ with marginals $\tilde{\boldsymbol{X}}_0\sim\mathcal{N}(\tilde{\boldsymbol{\mu}}_0, \tilde{\boldsymbol{\Sigma}}_0)$ and $\boldsymbol{X}_T\sim \mathcal{N}(\boldsymbol{\mu}_T, \boldsymbol{\Sigma}_T)$. Since the change of variables is invertible and the map $\mathbb{P}_{0,T}\mapsto\tilde{\mathbb{P}}_{0,T}$ is bijective, optimizing (\ref{eq:gaussian-kl-objective}) over $\mathbb{P}_{0,T}$ is equivalent to optimizing over $\tilde{\mathbb{P}}_{0,T}$.

Under $\tilde{\mathbb{P}}_{0,T}$, the quadratic cost becomes $\|\boldsymbol{x}_T-\tau_T\boldsymbol{x}_0-\boldsymbol{\zeta}(T)\|^2=\|\boldsymbol{x}_T-\tilde{\boldsymbol{x}}_0\|^2$ and the entropy term $(\log\mathrm{d}\mathbb{P}_{0,T}) \mathrm{d}\mathbb{P}_{0,T}$ is shifted by a constant that does not depend on $\mathbb{P}_{0,T}$, yielding the KL divergence:
\begin{align}
    \text{KL}(\tilde{\mathbb{P}}_{0,T}\|\mathbb{Q}_{0,T})=\int_{\mathbb{R}^d\times \mathbb{R}^d}(\log\mathrm{d}\tilde{\mathbb{P}}_{0,T}) \mathrm{d}\tilde{\mathbb{P}}_{0,T}+\frac{1}{2\kappa(T,T)}\int_{\mathbb{R}^d\times \mathbb{R}^d}\|\boldsymbol{x}_T-\tilde{\boldsymbol{x}}_0\|^2\mathrm{d}\tilde{\mathbb{P}}_{0,T}(\boldsymbol{x}_0, \boldsymbol{x}_T)+C
\end{align}
where $C$ is some constant. Since multiplying the objective by a constant $\kappa(T,T)> 0$ does not change the minimizer, the optimization problem reduces to a standard entropic OT problem, defined as: 
\begin{align}
    \min_{\mathbb{P}_{0,T}} \text{KL}(\mathbb{P}_{0,T}\|\mathbb{Q}_{0,T})&=\min_{\tilde{\mathbb{P}}_{0,T}} \text{KL}(\tilde{\mathbb{P}}_{0,T}\|\mathbb{Q}_{0,T})\nonumber\\
    &=\min_{\tilde{\mathbb{P}}_{0,T}}\left\{\int_{\mathbb{R}^d\times \mathbb{R}^d}\frac{1}{2}\|\boldsymbol{x}_T-\tilde{\boldsymbol{x}}_0\|^2\mathrm{d}\tilde{\mathbb{P}}_{0,T}+\kappa(T,T)\int_{\mathbb{R}^d\times \mathbb{R}^d}(\log\mathrm{d}\tilde{\mathbb{P}}_{0,T}) \mathrm{d}\tilde{\mathbb{P}}_{0,T}\right\}\label{eq:entropic-ot-proof}
\end{align}
where $\kappa(T,T)$ is now the entropic regularization weight $\tilde{\sigma}^2:=\kappa(T,T)$. From Lemma \ref{lemma:static-entropy-gaussian-ot}, the solution of (\ref{eq:entropic-ot-proof}) is given by the joint Gaussian:
\begin{align}
    \tilde{\mathbb{P}}^\star_{0,T}=\mathcal{N}\left(\begin{bmatrix}
        \boldsymbol{\mu}_0\\ \boldsymbol{\mu}_T
    \end{bmatrix}, \begin{bmatrix}
        \tilde{\boldsymbol{\Sigma}}_0 & \tilde{\boldsymbol{C}}_{\tilde{\sigma}}\\
        \tilde{\boldsymbol{C}}_{\tilde{\sigma}}^\top & \tilde{\boldsymbol{\Sigma}}_T
    \end{bmatrix}\right), \quad \text{s.t.}\quad \begin{cases}
        \tilde{\sigma}=\sqrt{\kappa(T,T)}\\
        \tilde{\boldsymbol{C}}_{\tilde{\sigma}}=\frac{1}{2}\left(\tilde{\boldsymbol{\Sigma}}_0^{\frac{1}{2}}\tilde{\boldsymbol{D}}_{\tilde{\sigma}}\tilde{\boldsymbol{\Sigma}}_0^{-\frac{1}{2}}-\tilde{\sigma}^2\boldsymbol{I}\right)\\
        \tilde{\boldsymbol{D}}_{\tilde{\sigma}}=\left(4\tilde{\boldsymbol{\Sigma}}_0^{\frac{1}{2}}\boldsymbol{\Sigma}_T\tilde{\boldsymbol{\Sigma}}_0^{\frac{1}{2}}-\tilde{\sigma}^4\boldsymbol{I}\right)^{\frac{1}{2}}
    \end{cases}
\end{align}
Applying the inverse transform $\boldsymbol{X}_0=\tau_T^{-1}(\tilde{\boldsymbol{X}}_0-\boldsymbol{\zeta}(T))$ and the identity $\text{Cov}(\boldsymbol{X}_0, \boldsymbol{X}_T)=\text{Cov}(\tau_T^{-1}\tilde{\boldsymbol{X}}_0, \boldsymbol{X}_T)=\tau_T^{-1}\text{Cov}(\tilde{\boldsymbol{X}}_0, \boldsymbol{X}_T)$, we recover the optimal joint Gaussian $\mathbb{P}_{0,T}$ that solves the static Gaussian SB problem: 
\begin{align}
    \pi_{0,T}^\star=\mathcal{N}\left(\begin{bmatrix}
        \boldsymbol{\mu}_0\\ \boldsymbol{\mu}_T
    \end{bmatrix}, \begin{bmatrix}
        \boldsymbol{\Sigma}_0 & \tau_T^{-1}\tilde{\boldsymbol{C}}_{\tilde{\sigma}}\\
        \tau_T^{-1}\tilde{\boldsymbol{C}}_{\tilde{\sigma}}^\top & \boldsymbol{\Sigma}_T
    \end{bmatrix}\right)=\mathcal{N}\left(\begin{bmatrix}
        \boldsymbol{\mu}_0\\ \boldsymbol{\mu}_T
    \end{bmatrix}, \begin{bmatrix}
        \boldsymbol{\Sigma}_0 & \boldsymbol{C}_{\sigma^\star}\\
       \boldsymbol{C}_{\sigma^\star}^\top & \boldsymbol{\Sigma}_T
    \end{bmatrix}\right)\label{eq:optimal-joint-gaussian}
\end{align}
where $\boldsymbol{C}_{\sigma^\star}:=\tau_T^{-1}\tilde{\boldsymbol{C}}_{\tilde{\sigma}}$. From our definition $\tilde{\boldsymbol{\Sigma}}_0=\tau_T^2\boldsymbol{\Sigma}_0$ in (\ref{eq:change-of-var-gsb}) and $\tilde{\sigma}^2:=\kappa(T,T)$, we expand $\tilde{\boldsymbol{D}}_{\tilde{\sigma}}$ as
\begin{small}
\begin{align}
    \tilde{\boldsymbol{D}}_{\tilde{\sigma}}=\left(4(\tau_T^2\boldsymbol{\Sigma}_0)^{\frac{1}{2}}\boldsymbol{\Sigma}_T(\tau_T^2\boldsymbol{\Sigma}_0)^{\frac{1}{2}}-\kappa(T,T)^2\boldsymbol{I}\right)^{\frac{1}{2}}&=\left(4\tau_T^2\boldsymbol{\Sigma}_0^{\frac{1}{2}}\boldsymbol{\Sigma}_T\boldsymbol{\Sigma}_0^{\frac{1}{2}}-\kappa(T,T)^2\boldsymbol{I}\right)^{\frac{1}{2}}\nonumber\\
    &=\tau_T\bigg(\underbrace{4\boldsymbol{\Sigma}_0^{\frac{1}{2}}\boldsymbol{\Sigma}_T\boldsymbol{\Sigma}_0^{\frac{1}{2}}-\frac{\kappa(T,T)^2}{\tau_T^2}\boldsymbol{I}}_{\boldsymbol{D}_{\sigma_\star}}\bigg)^{\frac{1}{2}}
\end{align}
\end{small}
which means $\sigma_\star^4:=\frac{\kappa(T,T)^2}{\tau_T^2}$ and $\sigma_\star^2=\frac{\kappa(T,T)}{\tau_T}$, proving the definition of $\sigma_\star$ in Theorem \ref{thm:closed-form-gaussiansbp}.

\textbf{Step 2: Deriving the Endpoint-Conditioned Gaussian Bridge. }
In this step, we aim to derive a closed form solution for the distribution of $\boldsymbol{X}_t$ \textit{conditioned} on the endpoint pair $(\boldsymbol{X}_0, \boldsymbol{X}_T)$, where the stochastic process follows a linear SDE defined in (\ref{eq:linear-sde-sol}). Conditioned on the initial state $\boldsymbol{X}_0$, we use the mean $\boldsymbol{\eta}(t)$ defined in (\ref{eq:guassian-sde-mean}) and covariance $\boldsymbol{\kappa}(t,t')$ defined in (\ref{eq:guassian-sde-cov}) to write the joint Gaussian distribution of $\boldsymbol{X}_t, \boldsymbol{X}_T|\boldsymbol{X}_0$ as:
\begin{small}
\begin{align}
    \boldsymbol{X}_t, \boldsymbol{X}_T|\boldsymbol{X}_0\sim \mathcal{N}\left(\begin{bmatrix}
        \boldsymbol{\mu}_0\\ \boldsymbol{\mu}_T
    \end{bmatrix}, \begin{bmatrix}
        \boldsymbol{\Sigma}_{00}& \boldsymbol{\Sigma}_{01}\\
       \boldsymbol{\Sigma}_{10}& \boldsymbol{\Sigma}_{11}
    \end{bmatrix}\right)=\mathcal{N}\left(\begin{bmatrix}
        \boldsymbol{\eta}(0)\\ \boldsymbol{\eta}(T)
    \end{bmatrix}, \begin{bmatrix}
        \boldsymbol{\kappa}(t,t)\boldsymbol{I}_d& \boldsymbol{\kappa}(t,T)\boldsymbol{I}_d\\
       \boldsymbol{\kappa}(t,T)\boldsymbol{I}_d & \boldsymbol{\kappa}(T,T)\boldsymbol{I}_d
    \end{bmatrix}\right)\label{eq:gaus-proof-final-2}
\end{align}
\end{small}
To compute the conditional distribution $\boldsymbol{X}_t|\boldsymbol{X}_T$ from the joint Gaussian $(\boldsymbol{X}_t, \boldsymbol{X}_T)$, we use the fact that any conditional distribution of a joint Gaussian random vector remains Gaussian, with mean and covariance given by:
\begin{small}
\begin{align}
    \boldsymbol{X}_t|\boldsymbol{X}_0, \boldsymbol{X}_T\sim \mathcal{N}(\tilde{\boldsymbol{\mu}}, \tilde{\boldsymbol{\Sigma}}), \quad \text{s.t.}\quad \begin{cases}
        \tilde{\boldsymbol{\mu}}=\boldsymbol{\mu}_0+\boldsymbol{\Sigma}_{01}\boldsymbol{\Sigma}_{11}^{-1}(\boldsymbol{x}-\boldsymbol{\mu}_1)\\
        \tilde{\boldsymbol{\Sigma}}=\boldsymbol{\Sigma}_{00}-\boldsymbol{\Sigma}_{01}\boldsymbol{\Sigma}_{11}^{-1}\boldsymbol{\Sigma}_{10}
    \end{cases}\label{eq:conditional-gaussian}
\end{align}
\end{small}
Therefore, from (\ref{eq:gaus-proof-final-2}), we have:
\begin{align}
    \mathbb{E}[\boldsymbol{X}_t|\boldsymbol{X}_0, \boldsymbol{X}_T]&=\boldsymbol{\eta}(t)+\frac{\boldsymbol{\kappa}(t,T)}{\boldsymbol{\kappa}(T,T)}(\boldsymbol{X}_T-\boldsymbol{\eta}(T))\nonumber\\
    &\overset{(\ref{eq:guassian-sde-mean})}{=}\tau_t\boldsymbol{X}_0+\boldsymbol{\zeta}(t)+\frac{\boldsymbol{\kappa}(t,T)}{\boldsymbol{\kappa}(T,T)}(\boldsymbol{X}_T-\tau_T\boldsymbol{X}_0-\boldsymbol{\zeta}(T))\nonumber\\
    &=\underbrace{\left(\tau_t-\frac{\boldsymbol{\kappa}(t,T)}{\boldsymbol{\kappa}(T,T)}\tau_T\right)}_{=:\bar{r}_t}\boldsymbol{X}_0+\underbrace{\frac{\boldsymbol{\kappa}(t,T)}{\boldsymbol{\kappa}(T,T)}}_{=:r_t}\boldsymbol{X}_T+\boldsymbol{\zeta}(t)-\underbrace{\frac{\boldsymbol{\kappa}(t,T)}{\boldsymbol{\kappa}(T,T)}}_{=:r_t}\boldsymbol{\zeta}(T)\nonumber\\
    &=\bar{r}_t\boldsymbol{X}_0+r_t\boldsymbol{X}_T+\boldsymbol{\zeta}(t)-r_t\boldsymbol{\zeta}(T)
\end{align}
which means the expected bridge at $\boldsymbol{X}_t$ is a weighted interpolation of the endpoints, with $\bar{r}_t$ defining the weight of the starting point $\boldsymbol{X}_0$ and $r_t$ defining the weight of the end point, with a deterministic correction term $\boldsymbol{\zeta}(t)-r_t\boldsymbol{\zeta}(T)$. Similarly, we can derive the cross-covariance for any $t' \geq t$ as:
\begin{small}
\begin{align}
    \mathbb{E}\left[\left(\boldsymbol{X}_t-\mathbb{E}[\boldsymbol{X}_t|\boldsymbol{X}_0, \boldsymbol{X}_T]\right)\left(\boldsymbol{X}_{t'}-\mathbb{E}[\boldsymbol{X}_{t'}|\boldsymbol{X}_0, \boldsymbol{X}_T]\right)^\top\big|\boldsymbol{X}_0, \boldsymbol{X}_T\right]=\left(\boldsymbol{\kappa}(t,t')-\frac{\boldsymbol{\kappa}(t,T)\boldsymbol{\kappa}(t',T)}{\boldsymbol{\kappa}(T,T)}\right)\boldsymbol{I}_d\label{eq:gaussian-cov-process}
\end{align}
\end{small}
which follows from the covariance identity $\text{Cov}(\boldsymbol{X}|\boldsymbol{Y})=\text{Cov}(\boldsymbol{X})-\text{Cov}(\boldsymbol{X},\boldsymbol{Y})\text{Cov}(\boldsymbol{Y})^{-1}\text{Cov}(\boldsymbol{Y},\boldsymbol{X})$. Since the mean and covariance uniquely define a Gaussian process, the law of the conditional bridge is equivalent to the expected mean of the bridge with some random fluctuations defined by a zero-mean Gaussian random variable $\boldsymbol{\xi}_t$ and covariance process (\ref{eq:gaussian-cov-process}):
\begin{align}
    \boldsymbol{X}_t|\boldsymbol{X}_0, \boldsymbol{X}_T&\overset{\text{law}}{=}\underbrace{\mathbb{E}\left[\boldsymbol{X}_t|\boldsymbol{X}_0, \boldsymbol{X}_T\right]}_{\text{deterministic given }\boldsymbol{X}_0, \boldsymbol{X}_T}+\underbrace{\boldsymbol{\xi}_t}_{\text{random fluctuation}}\nonumber\\
    &=\bar{r}_t\boldsymbol{X}_0+r_t\boldsymbol{X}_T+\boldsymbol{\zeta}(t)-r_t\boldsymbol{\zeta}(T)+\boldsymbol{\xi}_t
\end{align}
which proves that the Gaussian process conditioned on endpoints is simply a deterministic interpolation with independent residual noise $\boldsymbol{\xi}_t$.

\textbf{Step 3: Constructing the Gaussian SB from Endpoint Conditioned Bridges. }
Now, we can derive the solution to the Gaussian SB problem (\ref{def:gaussian-sbp}) as a mixture of endpoint conditioned briges, where the endpoints are sampled from the optimal joint distribution $(\boldsymbol{X}_0, \boldsymbol{X}_T)\sim \mathbb{P}_{0,T}^\star$ for $\pi_{0,T}^\star$ defined in (\ref{eq:optimal-joint-gaussian}) as the solution to the static Gaussian SB. First, we recall the (\ref{eq:chain-rule-kl}) which states:
\begin{align}
    \text{KL}(\mathbb{P}\|\mathbb{Q})=\text{KL}(\mathbb{P}_{0,T}\|\mathbb{Q}_{0,T})+\mathbb{E}_{\mathbb{P}_{0,T}}\left[\text{KL}(\mathbb{P}_{\cdot|0,T}(\cdot|\boldsymbol{X}_0, \boldsymbol{X}_T)\|\mathbb{Q}_{\cdot|0,T}(\cdot|\boldsymbol{X}_0, \boldsymbol{X}_T))\right]
\end{align}
Therefore, the Gaussian SB can be constructed by first sampling $(\boldsymbol{X}_0, \boldsymbol{X}_T)\sim \mathbb{P}_{0,T}^\star$ via (\ref{eq:optimal-joint-gaussian}) which yields the law:
\begin{small}
\begin{align}
    &\boldsymbol{X}_t\overset{\text{law}}{=}\bar{r}_t\boldsymbol{X}_0+r_t\boldsymbol{X}_T+\boldsymbol{\zeta}(t)-r_t\boldsymbol{\zeta}(T)+\boldsymbol{\xi}_t\label{eq:xt-law-gaussian}
\end{align}
\end{small}
where the covariance of the bridge measure is given by (\ref{eq:guassian-sde-cov}) with $t'=t$:
\begin{small}
\begin{align}
    \text{Cov}(\boldsymbol{\xi}_t)\overset{(\ref{eq:gaussian-cov-process})}{=}\bigg(\boldsymbol{\kappa}(t,t)-\underbrace{\frac{\boldsymbol{\kappa}(t,T)^2}{\boldsymbol{\kappa}(T,T)}}_{r_t\boldsymbol{\kappa}(t,T)}\bigg)\boldsymbol{I}_d=\boldsymbol{\kappa}(t,t)\bigg(1-\underbrace{\frac{\boldsymbol{\kappa}(t,T)^2}{\boldsymbol{\kappa}(T,T)\boldsymbol{\kappa}(t,t)}}_{=:\rho_t}\bigg)\boldsymbol{I}=:\boldsymbol{\kappa}(t,t)(1-\rho_t)\boldsymbol{I}\label{eq:cov-noise-gaussian}
\end{align}
\end{small}
Applying the linearity of expectation, we derive the mean $\boldsymbol{\mu}_t^\star$ of the Gaussian SB as:
\begin{align}
    \boldsymbol{\mu}_t^\star=\mathbb{E}_{\mathbb{P}_{0,T}^\star}[\boldsymbol{X}_t]&=\bar{r}_t\mathbb{E}_{\mathbb{P}_{0,T}^\star}[\boldsymbol{X}_0]+r_t\mathbb{E}_{\mathbb{P}_{0,T}^\star}[\boldsymbol{X}_T]+\boldsymbol{\zeta}(t)-r_t\boldsymbol{\zeta}(T)+\mathbb{E}_{\mathbb{P}_{0,T}^\star}[\boldsymbol{\xi}_t]\nonumber\\
    &=\bar{r}_t\boldsymbol{\mu}_0+r_t\boldsymbol{\mu}_T+\boldsymbol{\zeta}(t)-r_t\boldsymbol{\zeta}(T)
\end{align}
Applying the covariance identity\footnote{$\text{Cov}(a\boldsymbol{A}+b\boldsymbol{B})=a^2\text{Cov}(\boldsymbol{A}) +b^2\text{Cov}(\boldsymbol{B})+ab\text{Cov}(\boldsymbol{A}, \boldsymbol{B})+ab\text{Cov}(\boldsymbol{B}, \boldsymbol{A})$}, we derive the covariance $\boldsymbol{\Sigma}^\star_t$ of the Gaussian SB as:
\begin{align}
    \boldsymbol{\Sigma}^\star_t=\text{Cov}(\boldsymbol{X}_t)&=\text{Cov}\left( \bar{r}_t\boldsymbol{X}_0+r_t\boldsymbol{X}_T+\boldsymbol{\zeta}(t)-r_t\boldsymbol{\zeta}(T)+\boldsymbol{\xi}_t\right)\nonumber\\
    &=\text{Cov}\left( \bar{r}_t\boldsymbol{X}_0+r_t\boldsymbol{X}_T\right)+\text{Cov}\left(\boldsymbol{\xi}_t\right)\\
    &\overset{(\ref{eq:cov-noise-gaussian})}{=}\bar{r}_t^2\boldsymbol{\Sigma}_0+r_t^2\boldsymbol{\Sigma}_T+r_t\bar{r}_t\boldsymbol{C}_{\sigma_\star}+r_t\bar{r}_t\boldsymbol{C}_{\sigma_\star}^\top+\boldsymbol{\kappa}(t,t)(1-\rho_t)\boldsymbol{I}\nonumber\\
    &=\bar{r}_t^2\boldsymbol{\Sigma}_0+r_t^2\boldsymbol{\Sigma}_T+r_t\bar{r}_t(\boldsymbol{C}_{\sigma_\star}+\boldsymbol{C}_{\sigma_\star}^\top)+\boldsymbol{\kappa}(t,t)(1-\rho_t)\boldsymbol{I}
\end{align}
which concludes the first part of the proof that states the mean and covariance of the optimal marginal distributions of the Gaussian SB, given by:
\begin{align}
    \boxed{\begin{cases}
        \boldsymbol{\mu}^\star_t&=\bar{r}_t\boldsymbol{\mu}_0+r_t\boldsymbol{\mu}_T+\boldsymbol{\zeta}(t)-r_t\boldsymbol{\zeta}(T)\\
    \boldsymbol{\Sigma}^\star_t&=\bar{r}_t^2\boldsymbol{\Sigma}_0+r_t^2\boldsymbol{\Sigma}_T+r_t\bar{r}_t(\boldsymbol{C}_{\sigma_\star}+\boldsymbol{C}_{\sigma^\star}^\top)+\kappa(t,t)(1-\rho_t)\boldsymbol{I}_d
    \end{cases}}
\end{align}
Since the endpoints are defined by a joint Gaussian $\pi_{0,T}^\star$ and the bridge is a linear interpolation of the endpoints with added Gaussian noise $\boldsymbol{\xi}_t$, we conclude that the optimal SB measure is indeed Gaussian. Furthermore, applying the ideas from Section \ref{subsec:markov-reciprocal-proj}, we have that the solution to the Gaussian SB problem is the unique measure $\mathbb{P}^\star$ that is both Markov and in the reciprocal class $\mathcal{R}(\mathbb{Q})$. 

\textbf{Step 4: Deriving the SDE of the Gaussian Bridge. }
To simulate the solution to the Gaussian SB in practice, it is useful to derive the \textbf{associated SDE} which defines the forward propagation of the bridge given an intermediate state. 

Let $\phi: \mathbb{R}^d\times [0,T]\to \mathbb{R}$ be a smooth test function on the SB process $\boldsymbol{X}_{0:T}$. Then, the core of this derivation is to show that $\phi(\boldsymbol{X}_t)$ is an Itô process satisfying Itô's formula (\ref{thm:ito-formula}) with a \textbf{generator} $\mathcal{A}_t$ of the form:
\begin{align}
    \mathcal{A}_t\phi(\boldsymbol{x},t):=\partial_t\phi(\boldsymbol{x},t)+\boldsymbol{f}_{\mathcal{N}}(\boldsymbol{x},t)^\top\nabla \phi(\boldsymbol{x},t)+\frac{\sigma_t^2}{2}\Delta \phi(\boldsymbol{x},t)\label{eq:gaussian-sb-generator}
\end{align}

for a control drift $\boldsymbol{f}_{\mathcal{N}}:\mathbb{R}^d\times[0,T]\to \mathbb{R}$ that defines the SDE of $\boldsymbol{X}_t$ given by $d\boldsymbol{X}_t=\boldsymbol{f}_{\mathcal{N}}(\boldsymbol{X}_t,t)dt+\sigma_td\boldsymbol{B}_t$. To do this, we must match the time derivative of $\phi(\boldsymbol{X}_t,t)$ with (\ref{eq:gaussian-sb-generator}) by taking the continuous time limit:
\begin{align}
    \lim_{\Delta t\to 0}\left\{\frac{\mathbb{E}[\phi(\boldsymbol{X}_{t+\Delta t}, t+\Delta t)|\boldsymbol{X}_t=\boldsymbol{x}]-\phi(\boldsymbol{x},t)}{\Delta t}\right\}
\end{align}
The first step is to derive $\mathbb{E}[\phi(\boldsymbol{X}_{t+\Delta t}, t+\Delta t)|\boldsymbol{X}_t=\boldsymbol{x}]$ and take the infinitesimal time limit $\Delta t\to 0$. Since the conditional expectation is a Gaussian, we aim to derive its \textbf{mean} $\boldsymbol{\mu}_{t+\Delta t}:=\mathbb{E}[\boldsymbol{X}_{t+\Delta t}|\boldsymbol{X}_t=\boldsymbol{x}]=\boldsymbol{\mu}_t+\boldsymbol{\Sigma}_{t, t+\Delta t}\boldsymbol{\Sigma}_t^{-1}(\boldsymbol{x}-\boldsymbol{\mu}_t)$ (where $\boldsymbol{\Sigma}_{t, t+\Delta t}$ denotes the cross-covariance) and \textbf{covariance} $\boldsymbol{\Sigma}_{t+\Delta t}:=\text{Cov}(\boldsymbol{X}_{t+\Delta t}|\boldsymbol{X}_t=\boldsymbol{x})$ using the \textbf{first-order approximation}. First, we compute the time derivative $\dot{\boldsymbol{\Sigma}}^\star_t$ as:
\begin{small}
\begin{align}
    \dot{\boldsymbol{\Sigma}}^\star_t:=\partial_t\boldsymbol{\Sigma}^\star_t&=\partial_t\left(\bar{r}_t^2\boldsymbol{\Sigma}_0+r_t^2\boldsymbol{\Sigma}_T+r_t\bar{r}_t(\boldsymbol{C}_{\sigma_\star}+\boldsymbol{C}_{\sigma_\star}^\top)+\left(\boldsymbol{\kappa}(t,t)-\frac{\boldsymbol{\kappa}(t,T)^2}{\boldsymbol{\kappa}(T,T)}\right)\boldsymbol{I}\right)\nonumber\\
    &=2\bar{r}_t\dot{\bar{r}}_t\boldsymbol{\Sigma}_0+2r_t\dot{r}_t\boldsymbol{\Sigma}_T+(\dot{r}_t\bar{r}_t+r_t\dot{\bar{r}}_t)(\boldsymbol{C}_{\sigma_\star}+\boldsymbol{C}_{\sigma_\star}^\top)+\bigg(\partial_t\boldsymbol{\kappa}(t,t)-2\boldsymbol{\kappa}(t,T)\underbrace{\partial_t\frac{\boldsymbol{\kappa}(t,T)}{\boldsymbol{\kappa}(T,T)}}_{=\dot{r}_t}\bigg)\boldsymbol{I}\nonumber\\
    &=\underbrace{\dot{r}_t\left(r_t\boldsymbol{\Sigma}_T+\bar{r}_t\boldsymbol{C}_{\sigma_\star}\right)}_{\boldsymbol{P}_t}+\underbrace{\dot{r}_t\left(r_t\boldsymbol{\Sigma}_T+\bar{r}_t\boldsymbol{C}_{\sigma_\star}^\top\right)}_{\boldsymbol{P}_t^\top}\nonumber\\
    &+\underbrace{\dot{\bar{r}}_t\left(\bar{r}_t\boldsymbol{\Sigma}_0+r_t\boldsymbol{C}_{\sigma_\star}\right)}_{-\boldsymbol{Q}_t}+\underbrace{\dot{\bar{r}}_t\left(\bar{r}_t\boldsymbol{\Sigma}_0+r_t\boldsymbol{C}_{\sigma_\star}^\top\right)}_{-\boldsymbol{Q}_t^\top}+\bigg(\partial_t\boldsymbol{\kappa}(t,t)-2\dot{r}_t\boldsymbol{\kappa}(t,T)\bigg)\boldsymbol{I}\nonumber\\
    &=\left(\boldsymbol{P}_t+\boldsymbol{P}^\top_t\right)-\left(\boldsymbol{Q}_t+\boldsymbol{Q}^\top_t\right)+\bigg(\partial_t\boldsymbol{\kappa}(t,t)-2\dot{r}_t\boldsymbol{\kappa}(t,T)\bigg)\boldsymbol{I}\label{eq:gaus-dot-cov-def}
\end{align}
\end{small}
Next, we derive the cross-covariance $\boldsymbol{\Sigma}_{t, t+\Delta t}$ using the definition in (\ref{eq:gaussian-cov-process}) given by $\boldsymbol{\Sigma}_{t, t+\Delta t}:=\mathbb{E}\left[(\boldsymbol{X}_t-\boldsymbol{\mu}_t)(\boldsymbol{X}_{t+\Delta t}-\boldsymbol{\mu}_{t+\Delta t})^\top\right]$ where we make the following decomposition:
\begin{align}
    \boldsymbol{X}_t-\boldsymbol{\mu}_t&\overset{(\ref{eq:xt-law-gaussian})}{=}\bar{r}_t(\boldsymbol{X}_0-\boldsymbol{\mu}_0)+r_t(\boldsymbol{X}_T-\boldsymbol{\mu}_T)+\boldsymbol{\xi}_t\\
    \boldsymbol{X}_{t+\Delta t}-\boldsymbol{\mu}_{t+\Delta t}&\overset{(\ref{eq:xt-law-gaussian})}{=}\bar{r}_{t+\Delta t}(\boldsymbol{X}_0-\boldsymbol{\mu}_0)+r_{t+\Delta t}(\boldsymbol{X}_T-\boldsymbol{\mu}_T)+\boldsymbol{\xi}_{t+\Delta t}
\end{align}
Setting $\boldsymbol{A}:=\boldsymbol{X}_0-\boldsymbol{\mu}_0$, $\boldsymbol{B}:=\boldsymbol{X}_T-\boldsymbol{\mu}_T$, we can write the cross covariance as:
\begin{small}
\begin{align}
    \boldsymbol{\Sigma}_{t, t+\Delta t}&=\mathbb{E}\left[(\bar{r}_t\boldsymbol{A}+r_t\boldsymbol{B}+\boldsymbol{\xi}_t)( \bar{r}_{t+\Delta t}\boldsymbol{A}+r_{t+\Delta t}\boldsymbol{B}+\boldsymbol{\xi}_{t+\Delta t})^\top\right]\nonumber\\
    &=\bar{r}_t\bar{r}_{t+\Delta t}\underbrace{\mathbb{E}[\boldsymbol{A}\boldsymbol{A}^\top]}_{=:\boldsymbol{\Sigma}_0}+\bar{r}_tr_{t+\Delta t}\underbrace{\mathbb{E}[\boldsymbol{A}\boldsymbol{B}^\top]}_{=:\boldsymbol{C}_{\sigma_\star}}+\bar{r}_t\underbrace{\mathbb{E}[\boldsymbol{A}\boldsymbol{\xi}_{t+\Delta t}^\top]}_{=0}+r_t\bar{r}_{t+\Delta t}\underbrace{\mathbb{E}[\boldsymbol{B}\boldsymbol{A}^\top]}_{=:\boldsymbol{C}^\top_{\sigma_\star}}\nonumber\\
    &+r_tr_{t+\Delta t}\underbrace{\mathbb{E}[\boldsymbol{B}\boldsymbol{B}^\top]}_{\boldsymbol{\Sigma}_T}+r_t\underbrace{\mathbb{E}[\boldsymbol{B}\boldsymbol{\xi}_t^\top]}_{=0}+\bar{r}_{t+\Delta t}\underbrace{\mathbb{E}[\boldsymbol{\xi}_t\boldsymbol{A}^\top]}_{=0}+r_{t+\Delta t}\underbrace{\mathbb{E}[\boldsymbol{\xi}_t\boldsymbol{B}^\top]}_{=0}+\mathbb{E}[\boldsymbol{\xi}_t\boldsymbol{\xi}_{t+\Delta t}^\top]\nonumber\\
    &\overset{(\ref{eq:gaussian-cov-process})}{=}\bar{r}_t\bar{r}_{t+\Delta t}\boldsymbol{\Sigma}_0+\bar{r}_tr_{t+\Delta t}\boldsymbol{C}_{\sigma_\star}+r_t\bar{r}_{t+\Delta t}\boldsymbol{C}^\top_{\sigma_\star}+r_tr_{t+\Delta t}\boldsymbol{\Sigma}_T+\left(\boldsymbol{\kappa}(t,t+\Delta t)-\frac{\boldsymbol{\kappa}(t,T)\boldsymbol{\kappa}(t+\Delta t,T)}{\boldsymbol{\kappa}(T,T)}\right)\boldsymbol{I}\nonumber\\
    &=\bar{r}_t\bar{r}_{t+\Delta t}\boldsymbol{\Sigma}_0+\bar{r}_tr_{t+\Delta t}\boldsymbol{C}_{\sigma_\star}+r_t\bar{r}_{t+\Delta t}\boldsymbol{C}^\top_{\sigma_\star}+r_tr_{t+\Delta t}\boldsymbol{\Sigma}_T+(\boldsymbol{\kappa}(t,t+\Delta t)-r_{t+\Delta t}\boldsymbol{\kappa}(t,T))\boldsymbol{I}
\end{align}
\end{small}
where all covariances with only a single noise term $\boldsymbol{\xi}_t$ vanish since the noise is \textit{independent} of $\boldsymbol{X}_0, \boldsymbol{X}_T$. Since $\boldsymbol{\Sigma}_t$ has the same structure as $\boldsymbol{\Sigma}_{t+\Delta t}$, we can write the cross covariance as $\boldsymbol{\Sigma}_{t, t+\Delta t}=\boldsymbol{\Sigma}_t+\Delta \boldsymbol{\Sigma}$, where $\Delta\boldsymbol{\Sigma}$ is the change in covariance from $t\to t+\Delta t$, which gives the following expression: 
\begin{align}
    \boldsymbol{\Sigma}_{t, t+\Delta t}=&\boldsymbol{\Sigma}_t+\bar{r}_t(\bar{r}_{t+\Delta t}\bluetext{-\bar{r}_t})\boldsymbol{\Sigma}_0+\bar{r}_t(r_{t+\Delta t}\bluetext{-r_t})\boldsymbol{C}_{\sigma_\star}+r_t(\bar{r}_{t+\Delta t}\bluetext{-\bar{r}_{t}})\boldsymbol{C}^\top_{\sigma_\star}+r_t(r_{t+\Delta t}\bluetext{-r_t})\boldsymbol{\Sigma}_T\nonumber\\
    &+(\boldsymbol{\kappa}(t,t+\Delta t)\bluetext{-\boldsymbol{\kappa}(t,t)}-r_{t+\Delta t}\boldsymbol{\kappa}(t,T)\bluetext{+r_t\boldsymbol{\kappa}(t,T)})\boldsymbol{I}\nonumber\\
    =&\boldsymbol{\Sigma}_t+(\bar{r}_{t+\Delta t}-\bar{r}_t)\left(\bar{r}_t\boldsymbol{\Sigma}_0+r_t\boldsymbol{C}_{\sigma_\star}^\top\right)+(r_{t+\Delta t}-r_t)\left(\bar{r}_t\boldsymbol{C}_{\sigma_\star}+r_t\boldsymbol{\Sigma}_T\right)\nonumber\\
    &+(\boldsymbol{\kappa}(t,t+\Delta t)-\boldsymbol{\kappa}(t,t)-r_{t+\Delta t}\boldsymbol{\kappa}(t,T)+r_t\boldsymbol{\kappa}(t,T))\boldsymbol{I}\nonumber\\
    =&\boldsymbol{\Sigma}_t+\frac{\bar{r}_{t+\Delta t}-\bar{r}_t}{\dot{r}_t}\boldsymbol{P}_t-\frac{r_{t+\Delta t}-r_t}{\dot{\bar{r}}_t}\boldsymbol{Q}^\top_t+(\boldsymbol{\kappa}(t,t+\Delta t)\bluetext{\boldsymbol{-\kappa}(t,t)}-r_{t+\Delta t}\boldsymbol{\kappa}(t,T)\bluetext{+r_t\boldsymbol{\kappa}(t,T)})\boldsymbol{I}
\end{align}
Applying the \textbf{first-order approximation} which preserves all terms scaled by $h$ and denoting the higher-order terms with $o(\Delta t)$, we have:
\begin{small}
\begin{align}
    &\boldsymbol{\Sigma}_{t, t+\Delta t}=\boldsymbol{\Sigma}_t+\Delta t\underbrace{\left\{\boldsymbol{P}_t-\boldsymbol{Q}_t^\top+\left[\bluetext{\frac{\partial\boldsymbol{\kappa}}{dt'}(t,t)}-\pinktext{\dot{r}_t}\boldsymbol{\kappa}(t,T)\right]\boldsymbol{I}\right\}}_{=:\boldsymbol{S}_t}+o(\Delta t)\label{eq:gaus-S-def2}\\
    &=\boldsymbol{\Sigma}_t+\Delta t\left\{\boldsymbol{P}_t-\boldsymbol{Q}_t^\top+\left[\bluetext{\partial_t\left(\tau_t\tau_{t'}\int_0^t\tau_s^{-2}\sigma_s^2ds\right)\bigg\vert_{t'=t}}-\pinktext{\frac{1}{\boldsymbol{\kappa}(T,T)}\partial_t\left(\tau_t\tau_T\int_0^t\tau_s^{-2}\sigma_s^2ds\right)}\boldsymbol{\kappa}(t,T)\right]\boldsymbol{I}\right\}+o(\Delta t)\nonumber\\
    &=\boldsymbol{\Sigma}_t+\Delta t\bigg\{\boldsymbol{P}_t-\boldsymbol{Q}_t^\top+\bigg[\bluetext{\underbrace{\tau_t\dot{\tau}_t\int_0^t\tau_s^{-2}\sigma_s^2ds}_{c_t\boldsymbol{\kappa}(t,t)}}-\pinktext{\frac{1}{\boldsymbol{\kappa}(T,T)}\bigg(\underbrace{\dot{\tau}_t\tau_T\int_0^t\tau_s^{-2}\sigma_s^2ds}_{c_t\boldsymbol{\kappa}(t,T)}+\tau_t\tau_T\underbrace{\partial_t\int_0^t\tau_s^{-2}\sigma_s^2ds}_{=\tau_t^{-2}\sigma_t^2\text{ (Leibniz rule)}}\bigg)}\boldsymbol{\kappa}(t,T)\bigg]\boldsymbol{I}\bigg\}+o(\Delta t)\nonumber\\
    &=\boldsymbol{\Sigma}_t+\Delta t\bigg\{\boldsymbol{P}_t-\boldsymbol{Q}_t^\top+\bigg[\bluetext{c_t\boldsymbol{\kappa}(t,t)}-\pinktext{\bigg(\frac{c_t\boldsymbol{\kappa}(t,T)}{\boldsymbol{\kappa}(T,T)}+\frac{\tau_T\sigma_t^2}{\tau_t\boldsymbol{\kappa}(T,T)}}\bigg)\boldsymbol{\kappa}(t,T)\bigg]\boldsymbol{I}\bigg\}+o(\Delta t)\nonumber\\
    &=\boldsymbol{\Sigma}_t+\Delta t\bigg\{\boldsymbol{P}_t-\boldsymbol{Q}_t^\top+\bigg[\bluetext{c_t\boldsymbol{\kappa}(t,t)}-\pinktext{\underbrace{\frac{c_t\boldsymbol{\kappa}(t,T)^2}{\boldsymbol{\kappa}(T,T)}}_{c_t\rho_t\boldsymbol{\kappa}(t,t)}+\sigma_t^2\underbrace{\frac{\tau_T\boldsymbol{\kappa}(t,T)}{\tau_t\boldsymbol{\kappa}(T,T)}}_{=:\rho_t}}\bigg]\boldsymbol{I}\bigg\}+o(\Delta t)\nonumber\\
    &=\boldsymbol{\Sigma}_t+\Delta t\greentext{\underbrace{\big\{\boldsymbol{P}_t-\boldsymbol{Q}_t^\top+\big[c_t\boldsymbol{\kappa}(t,t)(1-\rho_t)+\sigma_t^2\rho_t\big]\boldsymbol{I}\big\}}_{=:\boldsymbol{S}_t}}+o(\Delta t)\nonumber\\
    &=\boldsymbol{\Sigma}_t+\Delta t\greentext{\boldsymbol{S}_t}+o(\Delta t)\label{eq:gaussian-cross-cov}
\end{align}
\end{small}
which gives us the expression for the cross covariance $\boldsymbol{\Sigma}_{t, t+\Delta t}=\boldsymbol{\Sigma}_t+\Delta t\boldsymbol{S}_t+o(\Delta t)$. Now, we can expand the mean $\tilde{\boldsymbol{\mu}}_{t+\Delta t}$ and covariance $\tilde{\boldsymbol{\Sigma}}_{t+\Delta t}$ of the conditional process $\boldsymbol{X}_{t+\Delta t}|\boldsymbol{X}_t$ using the same logic as (\ref{eq:conditional-gaussian}) to get:
\begin{align}
    \tilde{\boldsymbol{\mu}}_{t+\Delta t}&=\boldsymbol{\mu}_{t+\Delta t}+\boldsymbol{\Sigma}^{\top}_{t, t+\Delta t}\boldsymbol{\Sigma}_t^{-1}(\boldsymbol{x}-\boldsymbol{\mu}_t)\\
    \tilde{\boldsymbol{\Sigma}}_{t+\Delta t}&=\boldsymbol{\Sigma}_{t+\Delta t}+\boldsymbol{\Sigma}^{\top}_{t, t+\Delta t}\boldsymbol{\Sigma}_t^{-1}\boldsymbol{\Sigma}_{t, t+\Delta t}
\end{align}
Starting with the mean $\tilde{\boldsymbol{\mu}}_{t+\Delta t}$, we substitute (\ref{eq:gaussian-cross-cov}) to get:
\begin{align}
    \tilde{\boldsymbol{\mu}}_{t+\Delta t}&=\pinktext{\boldsymbol{\mu}_{t+\Delta t}}+\bluetext{(\boldsymbol{\Sigma}_t+\Delta t\boldsymbol{S}_t)}^{\top}\boldsymbol{\Sigma}_t^{-1}(\boldsymbol{x}-\boldsymbol{\mu}_t)\bluetext{+o(\Delta t)}\nonumber\\
    &=\pinktext{\boldsymbol{\mu}_t+\Delta t\dot{\boldsymbol{\mu}}_t}+\bluetext{(\underbrace{\boldsymbol{\Sigma}_t^\top\boldsymbol{\Sigma}_t^{-1}}_{=\boldsymbol{I}}+h\boldsymbol{S}_t^{\top}\boldsymbol{\Sigma}_t^{-1})}(\boldsymbol{x}-\boldsymbol{\mu}_t)\bluetext{+o(\Delta t)}\nonumber\\
    &=\boldsymbol{\mu}_t+\Delta t\dot{\boldsymbol{\mu}}_t+\boldsymbol{x}-\boldsymbol{\mu}_t+(h\boldsymbol{S}_t^{\top}\boldsymbol{\Sigma}_t^{-1})(\boldsymbol{x}-\boldsymbol{\mu}_t)+o(\Delta t)\nonumber\\
    &=\boxed{\boldsymbol{x}+h(\boldsymbol{S}_t^{\top}\boldsymbol{\Sigma}_t^{-1}(\boldsymbol{x}-\boldsymbol{\mu}_t)+\dot{\boldsymbol{\mu}}_t)+o(\Delta t)}\label{eq:gaussian-increment-mean}
\end{align}
Then, substituting (\ref{eq:gaussian-cross-cov}) into the covariance, we get:
\begin{align}
    \tilde{\boldsymbol{\Sigma}}_{t+\Delta t}&=\pinktext{\boldsymbol{\Sigma}_{t+\Delta t}}-\bluetext{(\boldsymbol{\Sigma}_t+\Delta t\boldsymbol{S}_t)^\top}\boldsymbol{\Sigma}_t^{-1}\bluetext{(\boldsymbol{\Sigma}_t+\Delta t\boldsymbol{S}_t) + o(\Delta t)}\nonumber\\
    &=\pinktext{\boldsymbol{\Sigma}_t+\Delta t\dot{\boldsymbol{\Sigma}}_t}-\bluetext{(\boldsymbol{\Sigma}_t+\Delta t\boldsymbol{S}_t)^\top}\boldsymbol{\Sigma}_t^{-1}\bluetext{(\boldsymbol{\Sigma}_t+\Delta t\boldsymbol{S}_t) + o(\Delta t)}\nonumber\\
    &=\pinktext{\boldsymbol{\Sigma}_t+\Delta t\dot{\boldsymbol{\Sigma}}_t}-\bluetext{(\boldsymbol{I}+h\boldsymbol{S}_t^\top\boldsymbol{\Sigma}_t^{-1})}\bluetext{(\boldsymbol{\Sigma}_t+\Delta t\boldsymbol{S}_t) + o(\Delta t)}\nonumber\\
    &=\pinktext{\boldsymbol{\Sigma}_t+\Delta t\dot{\boldsymbol{\Sigma}}_t}-\bluetext{(\boldsymbol{\Sigma}_t+\Delta t\boldsymbol{S}_t+\Delta t\boldsymbol{S}_t^\top+\underbrace{h^2\boldsymbol{S}_t^\top\boldsymbol{\Sigma}_t^{-1}\boldsymbol{S}_t}_{o(\Delta t)})} + o(\Delta t)\nonumber\\
    &=h(\dot{\boldsymbol{\Sigma}}_t-\boldsymbol{S}_t-\boldsymbol{S}_t^\top) + o(\Delta t)
\end{align}
and substituting the definition for $\boldsymbol{S}_t$ from (\ref{eq:gaus-S-def2}) and $\dot{\boldsymbol{\Sigma}}_t$ from (\ref{eq:gaus-dot-cov-def}), we get:
\begin{align}
    \tilde{\boldsymbol{\Sigma}}_{t+\Delta t}&\overset{(\ref{eq:gaus-S-def2}, \ref{eq:gaus-dot-cov-def})}{=}h\bigg[\bluetext{\left(\boldsymbol{P}_t+\boldsymbol{P}^\top_t\right)-\left(\boldsymbol{Q}_t+\boldsymbol{Q}^\top_t\right)+\bigg(\partial_t\boldsymbol{\kappa}(t,t)-2\dot{r}_t\boldsymbol{\kappa}(t,T)\bigg)\boldsymbol{I}}\nonumber\\
    &-\pinktext{\boldsymbol{P}_t-\boldsymbol{Q}_t^\top+\left[\frac{\partial\boldsymbol{\kappa}}{dt'}(t,t)-\pinktext{\dot{r}_t}\boldsymbol{\kappa}(t,T)\right]\boldsymbol{I}}-\greentext{\bigg(\boldsymbol{P}_t-\boldsymbol{Q}_t^\top+\left[\frac{\partial\boldsymbol{\kappa}}{dt'}(t,t)-\dot{r}_t\boldsymbol{\kappa}(t,T)\right]\boldsymbol{I}\bigg)^\top}\bigg] + o(\Delta t)\nonumber\\
    &=h\bigg(\bluetext{\partial_t\boldsymbol{\kappa}(t,t)}-\pinktext{2\frac{\partial\boldsymbol{\kappa}}{dt'}(t,t)}\bigg)\boldsymbol{I}+o(\Delta t)
\end{align}
Now, substituting (\ref{eq:guassian-sde-cov}) and decomposing $\frac{\partial\boldsymbol{\kappa}}{dt'}(t,t)-\dot{r}_t\boldsymbol{\kappa}(t,T)$ using the same procedure as (\ref{eq:gaus-S-def2}-\ref{eq:gaussian-cross-cov}), we have:
\begin{align}
    \tilde{\boldsymbol{\Sigma}}_{t+\Delta t}&=h\bigg(\bluetext{\partial_t\left(\tau_t^2\int_0^t\tau_s^{-2}\sigma_s^2ds\right)}-\pinktext{2\tau_t\dot{\tau}\int_0^t\tau_s^{-2}\sigma_s^2ds}\bigg)\boldsymbol{I}+o(\Delta t)\nonumber\\
    &=h\bigg(\bluetext{2\dot{\tau}_t\tau_t\int_0^t\tau_s^{-2}\sigma_t^2ds+\sigma_t^2}-\pinktext{2\tau_t\dot{\tau}\int_0^t\tau_s^{-2}\sigma_s^2ds}\bigg)\boldsymbol{I}+o(\Delta t)\nonumber\\
    &=\boxed{h\sigma_t^2\boldsymbol{I}+o(\Delta t)}\label{eq:gaus-proof-final}
\end{align}
Finally, we are ready to derive the form of $\mathbb{E}[\phi(\boldsymbol{X}_{t+\Delta t},t+\Delta t)|\boldsymbol{X}_t=\boldsymbol{x}]$ as the expected value at $\boldsymbol{X}_{t+\Delta t}$. Since $\boldsymbol{X}_{t+\Delta t}|\boldsymbol{X}_t=\boldsymbol{x}$ is Gaussian with mean $\tilde{\boldsymbol{\mu}}_{t+\Delta t}$ and covariance $\tilde{\boldsymbol{\Sigma}}_{t+\Delta t}$, we can write:
\begin{small}
\begin{align}
    \boldsymbol{X}_{t+\Delta t}=\tilde{\boldsymbol{\mu}}_{t+\Delta t}+\tilde{\boldsymbol{\Sigma}}_{t+\Delta t}^{1/2}\boldsymbol{z}, \quad \boldsymbol{z}\sim \mathcal{N}(\boldsymbol{0}, \boldsymbol{I}_d)
\end{align}
\end{small}
Applying the second-order Taylor expansion of $\phi(\tilde{\boldsymbol{x}}+\tilde{\boldsymbol{\mu}}_{t+\Delta t} ,t+\Delta t)$ around $\tilde{\boldsymbol{x}}=0$ yields:
\begin{small}
\begin{align}
    \phi(\tilde{\boldsymbol{x}}+\tilde{\boldsymbol{\mu}}_{t+\Delta t} ,t+\Delta t)= \phi(\tilde{\boldsymbol{\mu}}_{t+\Delta t} ,t+\Delta t)+\nabla\phi^\top\tilde{\boldsymbol{x}}+\frac{1}{2}\tilde{\boldsymbol{x}}^\top \nabla^2\phi\tilde{\boldsymbol{x}}+o(\|\tilde{\boldsymbol{x}}\|^2)
\end{align}
\end{small}
Given the Gaussian identities $\mathbb{E}[\boldsymbol{z}]=\boldsymbol{0}$ and $\mathbb{E}[\boldsymbol{z}\boldsymbol{z}^\top]=\boldsymbol{I}_d$, it follows that $\mathbb{E}[\tilde{\boldsymbol{x}}]=\boldsymbol{0}$ and $\mathbb{E}[\tilde{\boldsymbol{x}}\tilde{\boldsymbol{x}}^\top]=\tilde{\boldsymbol{\Sigma}}_{t+\Delta t}$. Using these identities and taking the conditional expectation, we have:
\begin{small}
\begin{align}
    \mathbb{E}[\phi(\boldsymbol{X}_{t+\Delta t},t+\Delta t)|\boldsymbol{X}_t=\boldsymbol{x}]&=\pinktext{\phi(\tilde{\boldsymbol{\mu}}_{t+\Delta t}, t+\Delta t)}+\greentext{\frac{1}{2}\text{Tr}\left(\tilde{\boldsymbol{\Sigma}}_{t+\Delta t}\nabla^2 \phi(\tilde{\boldsymbol{\mu}}_{t+\Delta t}, t+\Delta t)\right)}+o(\Delta t)\nonumber\\
    &\overset{(\ref{eq:gaus-proof-final})}{=}\pinktext{\phi(\tilde{\boldsymbol{\mu}}_{t+\Delta t}, t+\Delta t)}+\greentext{\frac{h\sigma_t^2}{2}\Delta \phi(\tilde{\boldsymbol{\mu}}_{t+\Delta t}, t+\Delta t)}+o(\Delta t)\label{eq:gaus-exp-1}
\end{align}
\end{small}
where the second line follows from $\tilde{\boldsymbol{\Sigma}}_{t+\Delta t}=h\sigma_t^2\boldsymbol{I}+o(\Delta t)$ from (\ref{eq:gaus-proof-final}). From (\ref{eq:gaussian-increment-mean}), the mean expands to $\tilde{\boldsymbol{\mu}}_{t+\Delta t}=\boldsymbol{x}+h(\boldsymbol{S}_t^{\top}\boldsymbol{\Sigma}_t^{-1}(\boldsymbol{x}-\boldsymbol{\mu}_t)+\dot{\boldsymbol{\mu}}_t)+o(\Delta t)=\boldsymbol{x}+\mathcal{O}(\Delta t)$, where $h$ is some infinitesimal time increment, we can apply the Taylor expansion around $\boldsymbol{x}$ to get:
\begin{align}
    \phi(\tilde{\boldsymbol{\mu}}_{t+\Delta t}, t+\Delta t)&=\phi(\boldsymbol{x},t+\Delta t)+h\big\langle \nabla\phi(\boldsymbol{x},t+\Delta t),  \boldsymbol{S}_t^{\top}\boldsymbol{\Sigma}_t^{-1}(\boldsymbol{x}-\boldsymbol{\mu}_t)+ \dot{\boldsymbol{\mu}}_t\big\rangle +o(\Delta t)\label{eq:gaus-taylor-exp-phi}
\end{align}
Then, substituting the expression (\ref{eq:gaus-taylor-exp-phi}) into (\ref{eq:gaus-exp-1}), we have:
\begin{align}
    &\mathbb{E}[\phi(\boldsymbol{X}_{t+\Delta t},t+\Delta t)|\boldsymbol{X}_t=\boldsymbol{x}]=\pinktext{\phi(\tilde{\boldsymbol{\mu}}_{t+\Delta t}, t+\Delta t)}+\greentext{\frac{h\sigma_t^2}{2}\Delta \phi(\tilde{\boldsymbol{\mu}}_{t+\Delta t}, t+\Delta t)}+o(\Delta t)\nonumber\\
    &=\pinktext{\phi(\boldsymbol{x},t+\Delta t)+h\big\langle \nabla\phi(\boldsymbol{x},t+\Delta t),  \boldsymbol{S}_t^{\top}\boldsymbol{\Sigma}_t^{-1}(\boldsymbol{x}-\boldsymbol{\mu}_t)+ \dot{\boldsymbol{\mu}}_t\big\rangle }+\greentext{\frac{h\sigma_t^2}{2}\Delta \phi(\boldsymbol{x}, t+\Delta t)}+ o(\Delta t)\nonumber\\
    &=\phi(\boldsymbol{x},t+\Delta t)+h\left(\frac{\sigma^2_t}{2}\Delta\phi(\boldsymbol{x},t+\Delta t)+\big\langle \nabla\phi(\boldsymbol{x},t+\Delta t),  \boldsymbol{S}_t^{\top}\boldsymbol{\Sigma}_t^{-1}(\boldsymbol{x}-\boldsymbol{\mu}_t)+ \dot{\boldsymbol{\mu}}_t\big\rangle \right)+o(\Delta t)
\end{align}
Dividing both sides by $h$ and taking the limit as $h \to 0$, we recover the final form of the \textbf{generator} $\mathcal{A}_t$ for $\phi(\boldsymbol{x},t)$ as:
\begin{align}
    \mathcal{A}_t\phi(\boldsymbol{x},t)&=\lim_{\Delta t\to 0}\left\{\frac{\mathbb{E}[\phi(\boldsymbol{X}_{t+\Delta t}, t+\Delta t)|\boldsymbol{X}_t=\boldsymbol{x}]-\phi(\boldsymbol{x},t)}{h}\right\}\nonumber\\
    &=\lim_{\Delta t\to 0}\bigg\{\pinktext{\frac{\phi(\boldsymbol{x},t+\Delta t)-\phi(\boldsymbol{x},t)}{h}}+\greentext{\frac{\sigma^2_t}{2}\Delta\phi(\boldsymbol{x},t+\Delta t)}\nonumber\\
    &\quad \quad+\greentext{\big\langle \nabla\phi(\boldsymbol{x},t+\Delta t),  \boldsymbol{S}_t^{\top}\boldsymbol{\Sigma}_t^{-1}(\boldsymbol{x}-\boldsymbol{\mu}_t)+ \dot{\boldsymbol{\mu}}_t\big\rangle}+\mathcal{O}(\Delta t)\bigg\}\nonumber\\
    &=\partial_t\phi(\boldsymbol{x},t)+\frac{\sigma^2_t}{2}\Delta\phi(\boldsymbol{x},t+\Delta t)+\big\langle \nabla\phi(\boldsymbol{x},t+\Delta t),  \bluetext{\boldsymbol{S}_t^{\top}\boldsymbol{\Sigma}_t^{-1}(\boldsymbol{x}-\boldsymbol{\mu}_t)+ \dot{\boldsymbol{\mu}}_t}\big\rangle
\end{align}
where the control drift takes the form:
\begin{align}
    \boxed{\boldsymbol{f}_{\mathcal{N}}(\boldsymbol{x},t):=\underbrace{\boldsymbol{S}_t^{\top}\boldsymbol{\Sigma}_t^{-1}(\boldsymbol{x}-\boldsymbol{\mu}_t)}_{\text{shape-correcting term}}+ \underbrace{\dot{\boldsymbol{\mu}}_t}_{\text{drift of mean}}}\tag{Gaussian SB Drift}
\end{align}
which can be decomposed into the mean drift of the Gaussian and a shape-correcting term that consists of the deviation from the mean $(\boldsymbol{x}-\boldsymbol{\mu}_t)$, how expensive it is under the covariance $\boldsymbol{\Sigma}_t^{-1}$, and how strongly to correct it $\boldsymbol{S}_t^\top$. 

As we show in Proposition \ref{prop:optimality-nonlinear-sbp}, the optimal control drift of the non-linear SB problem can be written as a gradient field as $\boldsymbol{f}_{\mathcal{N}}(\boldsymbol{x},t)=\nabla \psi_t(\boldsymbol{x})$ of the Lagrange multiplier $\psi_t(\boldsymbol{x})$ such that the Jacobian of $\boldsymbol{f}_{\mathcal{N}}$ is symmetric or $\partial_i\boldsymbol{f}_j=\partial_j\boldsymbol{f}_i$. Since we have:
\begin{align}
    \nabla \boldsymbol{f}_{\mathcal{N}}(\boldsymbol{x},t)=\nabla (\boldsymbol{S}_t^{\top}\boldsymbol{\Sigma}_t^{-1}(\boldsymbol{x}-\boldsymbol{\mu}_t)+ \dot{\boldsymbol{\mu}}_t)=\boldsymbol{S}_t^{\top}\boldsymbol{\Sigma}_t^{-1}
\end{align}
this implies that $\boldsymbol{S}_t^{\top}\boldsymbol{\Sigma}_t^{-1}$ is symmetric and concludes our proof. \hfill $\square$

From this result, we establish that, when the reference dynamics are linear–Gaussian, the (\ref{eq:gaussian-sb-problem}) admits an \textbf{exact closed-form solution that remains within the class of Gaussian Markov processes}, which is fully characterized by finite-dimensional evolutions of the mean and covariance and by an affine drift field that admits a gradient-field structure. This reveals the \textbf{core insight} that entropy-regularized transport between Gaussian marginals preserves the Gaussian structure of the distribution and induces a \textbf{potential-driven flow} that conserves distributional structure. This Gaussian case serves as a solvable model that provides a foundation for the upcoming sections on more specialized variations of the SB problem.

\subsection{Generalized Schrödinger Bridge Problem}
\label{subsec:generalized-sb}
\textit{Prerequisite: Section \ref{subsec:nonlinear-sbp}, \ref{subsec:hopf-cole-transform}, \ref{subsec:forward-backward-sde}}

\begin{figure}
    \centering
    \includegraphics[width=\linewidth]{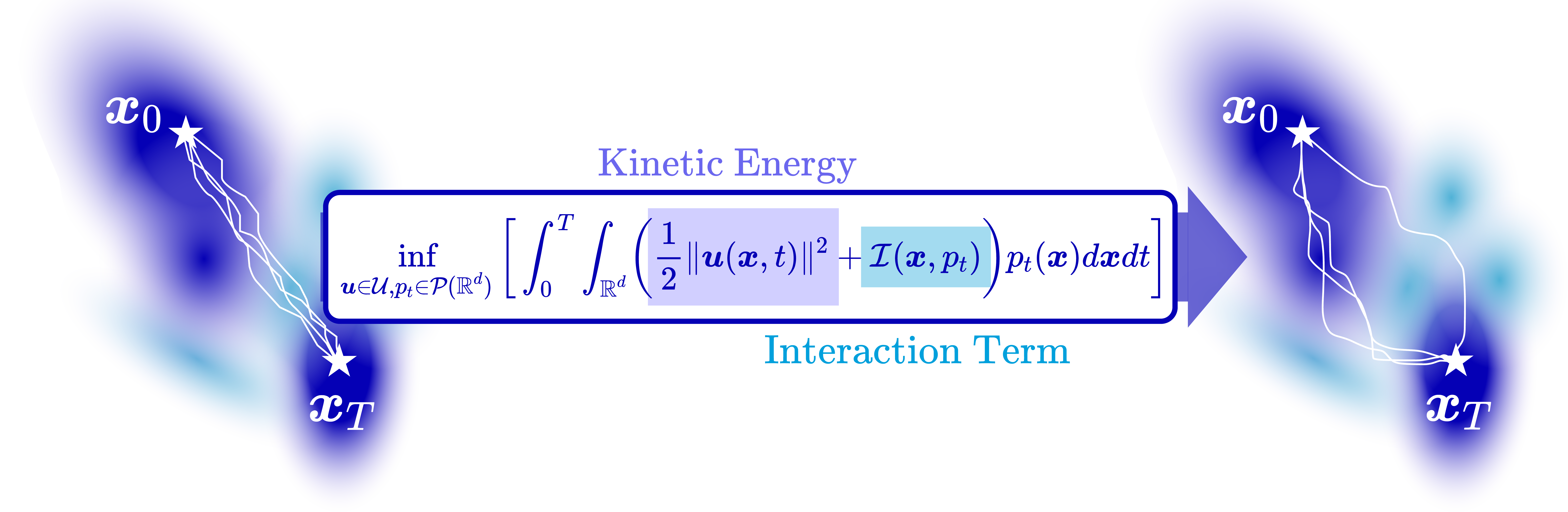}
    \caption{\textbf{Generalized Schrödinger Bridge Problem.} The generalized SB problem extends the dynamic SB problem by introducing a mean-field interaction cost $\mathcal{I}(x,p_t)$ that depends on the evolving marginal distribution. The resulting objective balances kinetic energy $\frac{1}{2}\|\boldsymbol{u}(\boldsymbol{x},t)\|^2$ with an interaction energy term integrated against the state density $p_t(x)$. The optimal dynamics transport particles between the prescribed marginals $(\pi_0,\pi_T)$ while accounting for collective interactions encoded by $\mathcal{I}(x,p_t)$, leading to a coupled Hamilton-Jacobi-Bellman and Fokker-Planck system that characterizes the optimal bridge.}
    \label{fig:generalized-sb}
\end{figure}

Up to this point, we have considered particles as acting independently via an optimal path along the Schrödinger bridge such that the total distribution over many particles matches the endpoint marginals. However, in many settings, particles evolve via stochastic trajectories that depend not only on their individual state and control but also \textbf{interactions with the population distribution} $p_t$. Solving for the optimal evolution of particles is referred to as solving a \boldtext{Mean-Field Game}, since the particles are influenced by the average dynamics of the population.

At equilibrium, each particle evolves via optimal control given the density of the particles $p_t$, and the population $p_t$ is generated from these optimally controlled particles. To determine the optimal dynamics, we leverage the definition of the \textbf{value function} $\psi_t(\boldsymbol{x}): \mathbb{R}^d\times [0,T]\to \mathbb{R}$ which defines the minimum cost of transporting $\boldsymbol{x}$ at time $t$. The gradient of the function $\nabla \psi_t(\boldsymbol{x})\in \mathbb{R}^d$ defines a potential field that adjusts the drift of the reference process so that the evolving distribution satisfies the endpoint constraints. 

The value function $\psi_t$ and marginal density $p_t$ are coupled via a \boldtext{Hamiltonian function} $\mathcal{H}(\boldsymbol{x}, \nabla \psi_t, p_t):\mathbb{R}^d\times \mathbb{R}\times \mathcal{P}(\mathbb{R}^d)\to \mathbb{R}$, which describes the dynamics of the interacting particles and an \boldtext{interaction function} $\mathcal{I}(\boldsymbol{x},p_t):\mathbb{R}^d\times \mathcal{P}(\mathbb{R}^d)\to \mathbb{R}$ that can be defined depending on the task. Given these functions, the pair of optimal value function and optimal state PDF $(\psi_t, p^\star_t)$ solves the following pair of PDEs:
\begin{align}
    \begin{cases}
        \partial_t \psi_t+\mathcal{H}(\boldsymbol{x},\nabla \psi_t, p^\star_t)+\frac{\sigma^2_t}{2}\Delta \psi_t=\mathcal{I}(\boldsymbol{x},p_t^\star)  &\psi_T(\boldsymbol{x})=\Phi(\boldsymbol{x}) \\
        \partial_t p^\star_t+\nabla \cdot (p_t^\star \nabla_{\nabla \psi_t} \mathcal{H}(\boldsymbol{x},\nabla \psi_t, p^\star_t))-\frac{\sigma^2_t}{2}\Delta p^\star_t=0  &p_0=\pi_0, p_T=\pi_T
    \end{cases}\label{eq:meanfieldgame-pdes}
\end{align}
where $\Phi(\boldsymbol{x}):\mathbb{R}^d\to \mathbb{R}$ is a terminal cost that can be defined to enforce a target distribution in the case of Schrödinger bridges. Given a solution pair $(\psi_t,p^\star_t)$, the dynamics of each particle evolve via the SDE defined by:
\begin{align}
    d\boldsymbol{X}_t =-\nabla_{\nabla \psi_t(\boldsymbol{X}_t)} \mathcal{H}(\boldsymbol{X}_t, \nabla \psi_t(\boldsymbol{X}_t),p_t)dt+\sigma_td\boldsymbol{B}_t, \quad \boldsymbol{X}_0\sim \pi_0\label{eq:meanfieldgames-sde}
\end{align}
which satisfies the marginals defined by $p_t$ as the number of particles goes to infinity. 

Now that we have defined the Mean-Field Game problem, we can observe that the coupled PDEs in (\ref{eq:meanfieldgame-pdes}) resemble the (\ref{eq:prop-hjb-fpe-system}) defined for the dynamic SB problem in Proposition \ref{prop:optimality-nonlinear-sbp} with the additional \boldtext{mean field interaction} $\mathcal{I}(\boldsymbol{x},p_t):\mathbb{R}^d\times\mathcal{P}(\mathbb{R}^d)\to \mathbb{R}$. Therefore, we can leverage this connection between mean-field games and dynamic SB to define the \boldtext{Generalized Schrödinger Bridge Problem} \citep{chen2015optimal, chen2023density,liu2022deep, liu2023generalized}.

\begin{definition}[Generalized Schrödinger Bridge Problem]\label{def:generalized-sbp}
    Given an interaction cost $\mathcal{I}(\boldsymbol{x},p_t):\mathbb{R}^d\times\mathcal{P}(\mathbb{R}^d)\to \mathbb{R}$, reference drift $\boldsymbol{f}(\boldsymbol{x},t)$, diffusion coefficient $\sigma_t$, and terminal marginal constraints $\pi_0, \pi_T\in \mathcal{P}(\mathbb{R}^d)$, the \textbf{generalized SB problem} can be written as:
    \begin{align}
        &\inf_{\boldsymbol{u}}\mathbb{E}_{\boldsymbol{X}_{0,T}\sim \mathbb{P}^u}\bigg[\int_0^T\left(\frac{1}{2}\|\boldsymbol{u}(\boldsymbol{X}_t,t)\|^2+\mathcal{I}(\boldsymbol{X}_t,p_t,t)\right)dt\bigg]\tag{Generalized SB Problem}\label{eq:generalized-sbp}\\
        &\text{s.t.}\quad\begin{cases}
            d\boldsymbol{X}_t=(\boldsymbol{f}(\boldsymbol{X}_t,t)+\sigma_t\boldsymbol{u}(\boldsymbol{X}_t,t))dt+\sigma_td\boldsymbol{B}_t\\
            \boldsymbol{X}_0\sim \pi_0, \quad \boldsymbol{X}_T\sim\pi_T
        \end{cases}
    \end{align}
    which can also be written as the density-space objective:
    \begin{align}
        &\inf_{(\boldsymbol{u}, p_t)}\bigg[\int_0^T\int_{\mathbb{R}^d}\left(\frac{1}{2}\|\boldsymbol{u}(\boldsymbol{x},t)\|^2+\mathcal{I}(\boldsymbol{x},p_t,t)\right)p_t(\boldsymbol{x})d\boldsymbol{x}dt\bigg]\\
        &\text{s.t.}\quad\begin{cases}
            \partial_tp_t(\boldsymbol{x}) = -\nabla\cdot\big(p_t(\boldsymbol{x})(\boldsymbol{f}(\boldsymbol{x},t)+\sigma_t\boldsymbol{u}(\boldsymbol{x},t))\big)+\frac{\sigma_t^2}{2} \Delta p_t\\
            p_0=\pi_0, \quad p_T=\pi_T
        \end{cases}
    \end{align}
\end{definition}

Following the derivation of the nonlinear HJB-FP system in Section \ref{subsec:nonlinear-sbp}, which defines the optimal pair $(\psi_t, p^\star_t)$ and the Hopf-Cole transform in Section \ref{subsec:hopf-cole-transform}, which defines the optimal potentials $(\varphi_t,\hat\varphi_t)$, we can derive the optimality conditions of the \textbf{generalized SB problem}.

\begin{definition}[Optimality Conditions of Generalized Schrödinger Bridge Problem]\label{def:optimality-generalized-sbp}
    Defining the Hamiltonian as:
    \begin{align}
        \mathcal{H}(\boldsymbol{x}, \nabla \psi_t, p_t):=\frac{1}{2}\|\sigma_t\nabla \psi_t\|^2-\nabla \psi_t^{\top}\boldsymbol{f}(\boldsymbol{x}, p_t, t)
    \end{align}
    We can write the optimality conditions $(\psi_t, p^\star_t)$ as the pair of coupled non-linear HJB-FP system with the \textbf{interaction term} $\mathcal{I}(\boldsymbol{x},p_t^\star):\mathbb{R}^d\times \mathcal{P}(\mathbb{R}^d)\to \mathbb{R}$ given by:
    \begin{small}
    \begin{align}
        \begin{cases}
        \partial_t \psi_t+\frac{\sigma_t^2}{2}\|\nabla \psi_t\|^2+\langle\nabla \psi_t, \boldsymbol{f}\rangle +\frac{\sigma_t^2}{2}\Delta \psi_t=\mathcal{I}(\boldsymbol{x},p^\star_t)\\
        \partial_t p^\star_t+\nabla \cdot (p_t^\star (\boldsymbol{f}+\sigma_t^2\nabla \psi_t))-\frac{\sigma_t^2}{2}\Delta p_t^\star=0
    \end{cases}\quad \text{s.t.}\quad 
    \begin{cases}
        p_0 =\pi_0\\
        p_T =\pi_T
    \end{cases}\label{eq:generalized-sbp-pdes}
    \end{align}
    \end{small}
    To transform the system of non-linear PDEs to linear PDEs, we can apply the \textbf{Hopf-Cole transform} discussed in Section \ref{subsec:hopf-cole-transform}, which defines the following change-of-variables:
    \begin{align}
        \psi_t(\boldsymbol{x})=\log \varphi_t(\boldsymbol{x}),\quad p^\star_t(\boldsymbol{x})=\varphi_t(\boldsymbol{x}) \hat\varphi_t(\boldsymbol{x})
    \end{align}
    which satisfy the pair of \textbf{linear PDEs}:
    \begin{align}
        \begin{cases}
            \partial_t \varphi_t=-\langle\nabla \varphi_t, \boldsymbol{f}\rangle-\frac{\sigma_t^2}{2}\Delta \varphi_t+\mathcal{I}\varphi_t\\
            \partial_t \hat{\varphi}_t=-\nabla \cdot (\hat{\varphi}_t \boldsymbol{f})+\frac{\sigma_t^2}{2}\Delta \hat{\varphi}_t-\mathcal{I}\hat\varphi_t
        \end{cases}\quad \text{s.t.}\quad
        \begin{cases}
            \pi_0=\varphi_0\hat\varphi_0\\
            \pi_T=\varphi_T\hat\varphi_T
        \end{cases}\label{eq:gsbm-hopf-cole}
    \end{align}
    which differs from the linear PDEs from (\ref{eq:hopf-cole-system}) only by the interaction term $\mathcal{I}(\boldsymbol{x},p_t^\star)$.
\end{definition}

In Section \ref{subsec:forward-backward-sde}, we established a pair of \textbf{forward-backward SDEs} that define the evolution of the dynamic Schrödinger bridge problem, which describes the evolution of the Schrödinger potentials $(\varphi_t, \hat\varphi_t)$ from the system of linear PDEs derived via the Hopf-Cole transform (Section \ref{subsec:hopf-cole-transform}). Given the modified Hopf-Cole linear PDEs for the generalized SB problem, we must define a new set of FBSDEs that integrates the mean-field interaction term $\mathcal{I}$.

\begin{proposition}[Generalized SB Forward-Backward SDEs (Theorem 2 in \citet{liu2022deep})]\label{prop:generalized-fbsdes}
    Given the Schrödinger potentials $(\varphi_t, \hat\varphi_t)$ that solve the Hopf-Cole linear PDEs from (\ref{eq:gsbm-hopf-cole}) and a stochastic process $\boldsymbol{X}_{0:T}$ that satisfies the forward-time SDE, we define additional stochastic processes as:
    \begin{small}
    \begin{align}
        &\boldsymbol{Y}_t=\boldsymbol{Y}(\boldsymbol{X}_t, t):=\log \varphi_t(\boldsymbol{X}_t), &\boldsymbol{Z}_t=\boldsymbol{Z}(\boldsymbol{X}_t,t):=\sigma_t\nabla \log \varphi_t(\boldsymbol{X}_t)\\
        &\widehat{\boldsymbol{Y}}_t= \widehat{\boldsymbol{Y}}(\boldsymbol{X}_t,t):=\log \hat\varphi_t(\boldsymbol{X}_t), &\widehat{\boldsymbol{Z}}_t=\widehat{\boldsymbol{Z}}(\boldsymbol{X}_t,t) :=\sigma_t\nabla \log \hat\varphi(\boldsymbol{X}_t)
    \end{align}
    \end{small}
    Then, the forward time evolution $t\in [0,T]$ of $\boldsymbol{X}_{0:T}$, $(\boldsymbol{Y}_t)_{t\in [0,T]}$ and $(\widehat{\boldsymbol{Y}}_t)_{t\in [0,T]}$ are characterized by the FBSDEs:
    \begin{align}
        \begin{cases}
            d\boldsymbol{X}_t=(\boldsymbol{f}(\boldsymbol{X}_t,p^\star_t)+\sigma_t\bluetext{\boldsymbol{Z}_t})dt+ \sigma_td\boldsymbol{B}_t\\
            d\boldsymbol{Y}_t=\left(\frac{1}{2}\|\bluetext{\boldsymbol{Z}_t}\|^2+\pinktext{\mathcal{I}(\boldsymbol{X}_t,p^\star_t)}\right)dt+\bluetext{\boldsymbol{Z}_t}^\top d\boldsymbol{B}_t\\
            d\widehat{\boldsymbol{Y}}_t=\left(\frac{1}{2}\|\bluetext{\widehat{\boldsymbol{Z}}_t}\|^2+\nabla\cdot (\sigma_t\bluetext{\widehat{\boldsymbol{Z}}_t}-\boldsymbol{f}(\boldsymbol{X}_t,p^\star_t))-\bluetext{\widehat{\boldsymbol{Z}}_t}^\top\bluetext{\boldsymbol{Z}_t}-\pinktext{\mathcal{I}(\boldsymbol{X}_t,p^\star_t)}\right)dt+\bluetext{\widehat{\boldsymbol{Z}}_t}^\top d\boldsymbol{B}_t\\
        \end{cases}
    \end{align}
    To condition the dynamics on the target distribution $\pi_T$, on the time-reversed coordinate $s:=T-t \in [0,T]$, we define the time-reversed SDE $(\tilde{\boldsymbol{X}}_s)_{s\in [0,T]}$ and the corresponding FBSDEs for $(\boldsymbol{Y}_s)_{s\in [0,T]}$ and $(\widehat{\boldsymbol{Y}}_s)_{s\in [0,T]}$ as:
    \begin{align}
        \begin{cases}
            d\tilde{\boldsymbol{X}}_s=(-\boldsymbol{f}(\tilde{\boldsymbol{X}}_s,p^\star_s)+\sigma_s\bluetext{\widehat{\boldsymbol{Z}}_s})ds+ \sigma_td\boldsymbol{B}_s\\
            d\boldsymbol{Y}_s=\left(\frac{1}{2}\|\bluetext{\boldsymbol{Z}_s}\|^2+\nabla\cdot (\sigma_t\bluetext{\boldsymbol{Z}_s}+\boldsymbol{f}(\tilde{\boldsymbol{X}}_s,p^\star_s))-\bluetext{\boldsymbol{Z}_s}^\top\bluetext{\widehat{\boldsymbol{Z}}_s}-\pinktext{\mathcal{I}(\tilde{\boldsymbol{X}}_s,p^\star_s)}\right)dt+\bluetext{\boldsymbol{Z}_s}^\top d\boldsymbol{B}_s\\
            d\widehat{\boldsymbol{Y}}_s=\left(\frac{1}{2}\|\bluetext{\widehat{\boldsymbol{Z}}_s}\|^2+\pinktext{\mathcal{I}(\boldsymbol{X}_s,p_s^\star)}\right)ds+\bluetext{\widehat{\boldsymbol{Z}}_s}^\top d\boldsymbol{B}_s\\
        \end{cases}
    \end{align}
    Given the SB optimality condition $p^\star_t=\varphi_t\hat\varphi_t$, we define the \textbf{interaction term} $\mathcal{I}:\mathbb{R}^d\times\mathcal{P}(\mathbb{R}^d)\to \mathbb{R}$ as:
    \begin{align}
        &\mathcal{I}(\boldsymbol{X}_t, p^\star_t)=\mathcal{I}(\boldsymbol{X}_t, \varphi_t\hat\varphi_t), \quad \boldsymbol{f}(\boldsymbol{X}_t,p^\star_t)=\boldsymbol{f}(\boldsymbol{X}_t,\varphi_t\hat\varphi_t)\\
        &\mathcal{I}(\tilde{\boldsymbol{X}}_s, p^\star_s)=\mathcal{I}(\tilde{\boldsymbol{X}}_s, \varphi_s\hat\varphi_s), \quad \boldsymbol{f}(\tilde{\boldsymbol{X}}_s,p^\star_s)=\boldsymbol{f}(\tilde{\boldsymbol{X}}_s,\varphi_s\hat\varphi_s)
    \end{align}
\end{proposition}

\textit{Proof Sketch.} This proof follows closely from the derivation in Section \ref{subsec:forward-backward-sde}, where we apply Itô's formula to get $d\log \varphi_t$ and $d\log\hat\varphi_t$, but instead of substituting the Hopf-Cole transform for the standard dynamic SB problem for $\partial_t \varphi_t$ and $\partial_t\hat\varphi_t$, we substitute the Hopf-Cole transform with the interaction term $\mathcal{I}$ given in (\ref{eq:gsbm-hopf-cole}). The reverse-time FBSDEs follow the same steps after reformulating the HJB-FP PDEs with the reversed time coordinate $s:=T-t$ and applying the Hopf-Cole transform. \hfill $\square$

Compared to the dynamic SB FBSDEs derived in Section \ref{subsec:forward-backward-sde}, the only structural modification appears through the interaction term $\mathcal{I}(\boldsymbol{x},p_t^\star)$, which introduces a mean-field dependence into both the forward and backward SDEs. The forward–backward structure remains similar to the dynamic SB FB-SDEs, but the system now encodes collective effects through the optimal marginal flow defined as $p^\star_t=\varphi_t\hat\varphi_t$.

\begin{remark}[Generalization of Dynamic SB]\label{remark:generalized-sb}
    When the interaction term vanishes for all $(\boldsymbol{x},t)\in \mathbb{R}^d \times[0,T]$, i.e., $\mathcal{I}\equiv 0$, and the reference drift is independent of the marginal, i.e., $\boldsymbol{f}(\boldsymbol{x},p_t,t)= \boldsymbol{f}(\boldsymbol{x},t)$, the \textbf{generalized SB problem} reduces to (\ref{eq:dynamic-sb-problem}) and can be interpreted as a generalization of dynamic SB to McKean-Vlasov settings with mean-field interactions. 
\end{remark}

In summary, the generalized Schrödinger bridge extends the classical dynamic SB formulation by introducing an interaction term $\mathcal{I}(\boldsymbol{x},p_t)$, that allows the dynamics to depend on the evolving marginal distribution $p_t$. Crucially, we show that the extension of the dynamic SB problem to settings with mean-field interactions \textit{preserves the fundamental structure of the dynamic SB problem} with the only difference being an additive interaction term. Therefore, many frameworks used to solve the dynamic SB problem can be adapted to solve the generalized SB problem.

\subsection{Multi-Marginal Schrödinger Bridge Problem}
\label{subsec:multi-marginal-sb}

The standard dynamic SB problem aims to determine the optimal bridge that maps particles from the initial distribution $\pi_0$ to the terminal distribution $\pi_T$ while minimizing the KL divergence to the reference measure. A natural extension of this problem is to consider \textbf{multiple marginal constraints} at multiple points along the time horizon, which can be applied to construct feasible trajectories between observed snapshots over coarse time intervals. This variation of the SB problem is considered the \boldtext{Multi-Marginal Schrödinger Bridge Problem} \citep{chen2019multi, chen2023deep, theodoropoulos2025momentum}.

\begin{figure}
    \centering
    \includegraphics[width=\linewidth]{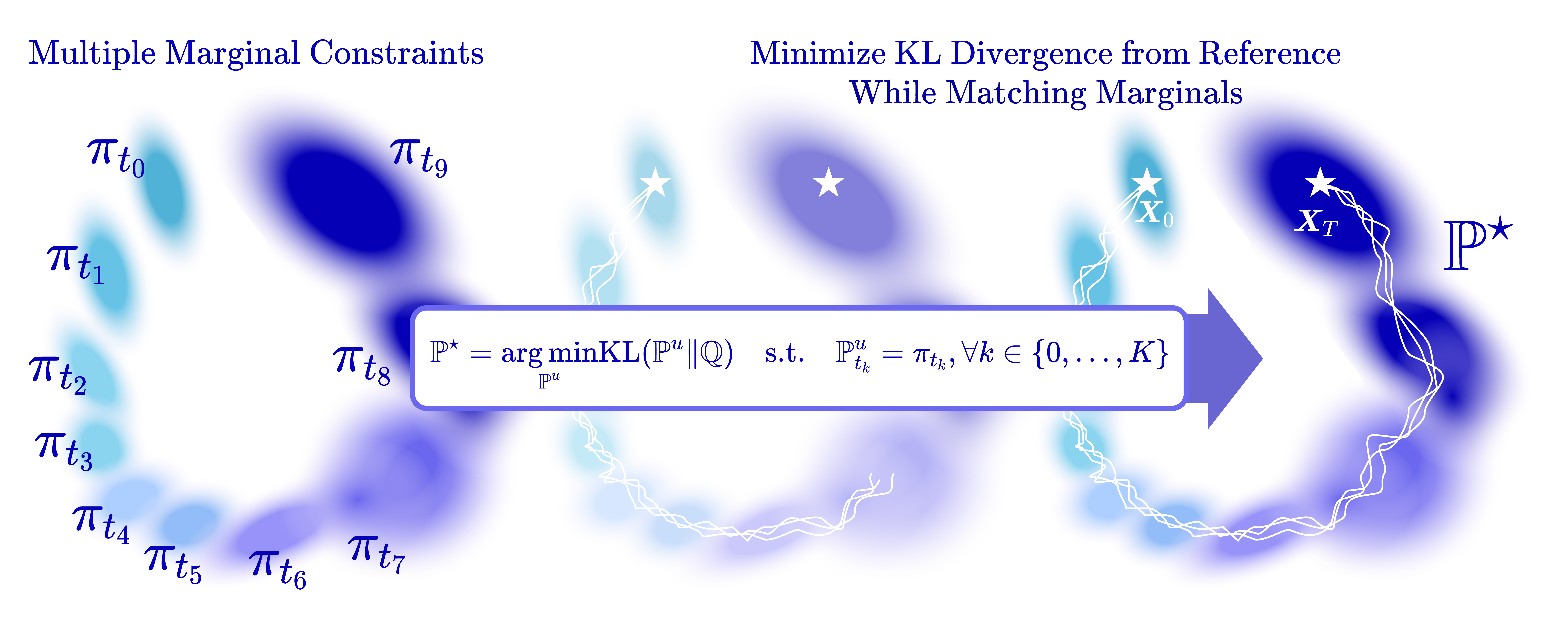}
    \caption{\textbf{Multi-Marginal Schrödinger Bridge Problem.} The multi-marginal Schrödinger bridge constructs an optimal stochastic process whose marginal distributions match a sequence of prescribed intermediate densities $\{\pi_{t_k}\}_{k=0}^K$. Starting from a reference stochastic process $\mathbb{Q}$, the optimal path measure $\mathbb{P}^\star$ is obtained by minimizing the path-space KL divergence while matching the marginal constraints. }
    \label{fig:multimarginal}
\end{figure}

\begin{definition}[Standard Multi-Marginal Schrödinger Bridge Problem]
    Given a uncontrolled reference measure $\mathbb{Q}$ multiple marginal constraints $\{\pi_{t_k}\in \mathcal{P}(\mathbb{R}^d)\}_{k=1}^K$ at sequential time points $0=t_0< \dots< t_k < \dots < t_K=T$, the standard \textbf{multi-marginal Schrödinger bridge problem} aims to determine a control $\boldsymbol{u}(\boldsymbol{x},t)$ that minimizes:
    \begin{small}
    \begin{align}
        \inf_{\boldsymbol{u}}&\mathbb{E}_{\boldsymbol{X}_{0:T}\sim \mathbb{P}^u}\left[\int_0^T\frac{1}{2}\|\boldsymbol{u}(\boldsymbol{X}_t,t)\|^2dt\right]\tag{Multi-Marginal SB Problem}\label{eq:multi-marginal-sbp}\\
        &\text{s.t.}\quad \begin{cases}
            d\boldsymbol{X}_t=(\boldsymbol{f}(\boldsymbol{X}_t,t)+\sigma_t\boldsymbol{u}(\boldsymbol{X}_t,t))dt+\sigma_td\boldsymbol{B}_t\\
            \boldsymbol{X}_{t_k}\sim \pi_{t_k}, \forall k\in \{0, \dots, K\}
        \end{cases}
    \end{align}
    \end{small}
    where the optimal $\boldsymbol{u}^\star$ generates the path measure $\mathbb{P^\star}$ of minimal relative entropy with respect $\mathbb{Q}$ among all controlled measures matching the prescribed marginals. This can equivalently be written as the density-space objective:
    \begin{align}
        \inf_{\boldsymbol{u}}&\mathbb{E}_{\boldsymbol{X}_{0:T}\sim \mathbb{P}^u}\left[\int_0^T\int_{\mathbb{R}^d}\frac{1}{2}\|\boldsymbol{u}(\boldsymbol{x},t)\|^2p_t(\boldsymbol{x})d\boldsymbol{x}dt\right]\\
        &\text{s.t.}\quad \begin{cases}
            \partial_tp_t(\boldsymbol{x}) = -\nabla\cdot\big(p_t(\boldsymbol{x})(\boldsymbol{f}(\boldsymbol{x},t)+\sigma_t\boldsymbol{u}(\boldsymbol{x},t))\big)+\frac{\sigma_t^2}{2} \Delta p_t\\
            p_{t_k}= \pi_{t_k}, \forall k\in \{0, \dots, K\}
        \end{cases}
    \end{align}
\end{definition}

Since intermediate marginal state distributions can have associated velocity information that determines how it propagates to the subsequent distribution such that $\pi_{t_k}(\boldsymbol{x}, \boldsymbol{v})\in \mathcal{P}(\mathbb{R}^{2d})$, we re-formulate the multi-marginal SB problem in \textbf{phase space}, known as the \boldtext{Momentum Multi-Marginal Schrödinger Bridge} problem \citep{theodoropoulos2025momentum, chen2023deep}, where the state is given by a vector $(\boldsymbol{x}, \boldsymbol{v})\in \mathbb{R}^{2d}$. In this setting, the reference dynamics that follow a pair of second-order SDEs:
\begin{align}
    \begin{cases}
        d\boldsymbol{x}_t=\boldsymbol{v}_tdt\\
        d\boldsymbol{v}_t=\boldsymbol{a}_tdt+ \sigma_td\boldsymbol{B}_t
    \end{cases}\label{eq:second-order-sde}
\end{align}
where stochasticity is introduced into the second-order SDE. We define the marginal distribution generated by the SDEs in (\ref{eq:second-order-sde}) over the phase space as $p_t\in \mathcal{P}(\mathbb{R}^{2d})$, the distribution of the state as $\mu(\boldsymbol{x},t)=\int_{\mathbb{R}^d}p_t(\boldsymbol{x},\boldsymbol{v})d\boldsymbol{v}$, and the distribution over velocity as $\xi(\boldsymbol{v},t)=\int_{\mathbb{R}^d}p_t(\boldsymbol{x},\boldsymbol{v})d\boldsymbol{x}$.

\begin{definition}[Momentum Multi-Marginal Schrödinger Bridge Problem]
    Given multiple marginal distributions over the phase space $\pi_{t_0}, \dots , \pi_{t_K}\in \mathcal{P}(\mathbb{R}^{2d})$ defined at sequential time points $t_0, \dots, t_K\in [0,T]$ on the time horizon, the \textbf{multi-marginal SB problem} aims to determine the optimal acceleration $\boldsymbol{a}_t^\star\equiv\boldsymbol{a}^\star_t(\boldsymbol{x}, \boldsymbol{v},t)\in \mathbb{R}^{2d}$ that solves the SOC problem: 
    \begin{small}
    \begin{align}
        \boldsymbol{a}^\star_t=\underset{\boldsymbol{a}_t}{\arg\min}\mathbb{E}_{p_t}\left[\int_0^T\frac{1}{2}\|\boldsymbol{a}_t\|^2 dt\right]\quad \text{s.t.}\quad \begin{cases}
            d\boldsymbol{x}_t=\boldsymbol{v}_tdt\\
            d\boldsymbol{v}_t=\boldsymbol{a}_tdt+ \sigma_td\boldsymbol{B}_t\\
            (\boldsymbol{x}_{t_k}, \boldsymbol{v}_{t_k})\sim \pi_{t_k}, \forall k\in \{0, \dots, K\}
        \end{cases}
    \end{align}
    \end{small}
    Equivalently, writing the problem in terms of the distribution $p_t$ and the Fokker-Planck equation from Section \ref{subsec:fp-equation}, we have:
    \begin{align}
    \boldsymbol{a}^\star_t&=\underset{\boldsymbol{a}_t}{\arg\min}\mathbb{E}_{p_t}\left[\int_0^T\int_{\mathbb{R}^d}\int_{\mathbb{R}^d}\frac{1}{2}\|\boldsymbol{a}_t\|^2p_td\boldsymbol{x}d\boldsymbol{v}dt\right]\\ &\text{s.t.}\quad \begin{cases}
            \partial_tp_t(\boldsymbol{x},\boldsymbol{v})+\boldsymbol{v}\cdot \nabla p_t+\nabla_{\boldsymbol{v}}\cdot (\boldsymbol{a}_tp_t)-\frac{\sigma_t^2}{2}\Delta_{\boldsymbol{v}}p_t=0\\
            \mu(\boldsymbol{x}, t_k)=\int_{\mathbb{R}^d}\pi_{t_k}(\boldsymbol{x}, \boldsymbol{v})d\boldsymbol{v}, \forall k\in \{0, \dots, K\}
        \end{cases}\label{eq:mmsb-fpe}
    \end{align}
\end{definition}

The diffusion term in the FP equation makes the dynamics \textit{irreversible} since diffusion increases entropy over a single time direction. To absorb the diffusion term into a deterministic drift, we can expand the velocity term $\frac{\sigma_t^2}{2}\Delta_{\boldsymbol{v}}p_t$ as:
\begin{align}
    \Delta_{\boldsymbol{v}}p_t=\nabla_{\boldsymbol{v}} \cdot (\nabla _{\boldsymbol{v}}p_t)=\nabla_{\boldsymbol{v}}\cdot (p_t\nabla_{\boldsymbol{v}} \log p_t)
\end{align}
Substituting this into the FP equation in (\ref{eq:mmsb-fpe}), we get:
\begin{align}
    \partial_tp_t(\boldsymbol{x},\boldsymbol{v})+\boldsymbol{v}\cdot \nabla p_t+\nabla_{\boldsymbol{v}}\cdot \left(\boldsymbol{a}_tp_t-\frac{\sigma_t^2}{2}\bluetext{p_t\nabla_{\boldsymbol{v}}\log p_t}\right)&=0\nonumber\\
    \partial_tp_t(\boldsymbol{x},\boldsymbol{v})+\boldsymbol{v}\cdot \nabla p_t+\nabla_{\boldsymbol{v}}\cdot \bluetext{\underbrace{\left(\boldsymbol{a}_t-\frac{\sigma_t^2}{2}\nabla_{\boldsymbol{v}}\log p_t\right)}_{=:\hat{\boldsymbol{a}}_t}}p_t&=0
\end{align}
Then, we can define $\hat{\boldsymbol{a}}_t:=\boldsymbol{a}_t-\frac{\sigma_t^2}{2}\nabla_{\boldsymbol{v}}\log p_t$ to get:
\begin{align}
    \partial_tp_t(\boldsymbol{x},\boldsymbol{v})+\boldsymbol{v}\cdot \nabla p_t+\nabla_{\boldsymbol{v}}\cdot (\hat{\boldsymbol{a}}_tp_t)=0\label{eq:continuity-ahat}
\end{align}
To write the cost functional in terms of $\hat{\boldsymbol{a}}$, we substitute $\boldsymbol{a}_t=\hat{\boldsymbol{a}}_t+\frac{\sigma_t^2}{2}\nabla_{\boldsymbol{v}}\log p_t$ in the squared cost to get:
\begin{align}
    \frac{1}{2}\|\boldsymbol{a}\|^2 =\frac{1}{2}\left\|\hat{\boldsymbol{a}}+\frac{\sigma_t^2}{2}\nabla_{\boldsymbol{v}}\log p_t\right\|^2=\frac{1}{2}\|\hat{\boldsymbol{a}}\|^2 +\frac{\sigma_t^2}{8}\|\nabla_{\boldsymbol{v}}\log p_t\|^2+\frac{1}{2}\langle \hat{\boldsymbol{a}}, \nabla_{\boldsymbol{v}}\log p_t\rangle
\end{align}
and integrating the expanded form over time and phase space, we have:
\begin{align}
    \underbrace{\int_0^T\int\frac{1}{2}\|\hat{\boldsymbol{a}}\|^2 p_td\boldsymbol{x}d\boldsymbol{v}dt}_{\text{transport cost}}+\underbrace{\int_0^T\int\frac{\sigma_t^2}{8}\|\nabla_{\boldsymbol{v}}\log p_t\|^2p_td\boldsymbol{x}d\boldsymbol{v}dt}_{\text{velocity Fisher information}}+\underbrace{\int_0^T\int\frac{1}{2}\langle \hat{\boldsymbol{a}}, \nabla_{\boldsymbol{v}}\log p_t\rangle p_td\boldsymbol{x}d\boldsymbol{v}dt}_{\text{cross term}}\label{eq:expanded-squared-a}
\end{align}
where the first term is the transport cost. The second term is the velocity \textbf{Fisher information} (squared norm of the score function), which measures the sensitivity of the distribution under infinitesimal changes to the velocity. Since the diffusion only acts on the velocity coordinate in (\ref{eq:second-order-sde}), this term acts as an uncertainty regularization after absorbing the diffusion into the deterministic drift. Finally, the cross term can be expanded as follows:
\begin{align}
    \int_0^T\int \frac{1}{2}\langle \hat{\boldsymbol{a}}, \nabla_{\boldsymbol{v}}\log p_t\rangle p_td\boldsymbol{x}d\boldsymbol{v}dt&=\int_0^T\int\frac{1}{2}\hat{\boldsymbol{a}}\cdot (p_t\nabla_{\boldsymbol{v}}\log p_t)d\boldsymbol{x}d\boldsymbol{v}dt\nonumber\\
    &=\int_0^T\int\frac{1}{2}\hat{\boldsymbol{a}}\cdot \nabla_{\boldsymbol{v}}p_td\boldsymbol{x}d\boldsymbol{v}dt
\end{align}

We now observe that this cross term is the alignment of the transport velocity $\hat{\boldsymbol{a}}$ with the density gradient $\nabla_{\boldsymbol{v}}p_t$ in velocity space. Using the identity $\nabla_{\boldsymbol{v}} \cdot (p_t\hat{\boldsymbol{a}})=\hat{\boldsymbol{a}}\cdot \nabla_{\boldsymbol{v}}p_t + p_t(\nabla_{\boldsymbol{v}} \cdot \hat{\boldsymbol{a}})$ and applying integration by parts in velocity space under vanishing boundary conditions\footnote{Assumes that $p_t(\boldsymbol{x}, \boldsymbol{v})\to 0$ as $\|\boldsymbol{v}\|\to \infty$ is satisfied.}, we get:
\begin{align}
    \int_0^T\int_{\mathbb{R}^d}\frac{1}{2}(\hat{\boldsymbol{a}}\cdot \nabla_{\boldsymbol{v}}p_t)d\boldsymbol{v}d\boldsymbol{x}dt&=\bluetext{\int_0^T\int\frac{1}{2}(\nabla_{\boldsymbol{v}} \cdot (p_t\hat{\boldsymbol{a}}))d\boldsymbol{v}d\boldsymbol{x}dt}\underbrace{-\int_0^T\int \frac{1}{2}p_t(\nabla_{\boldsymbol{v}} \cdot \hat{\boldsymbol{a}})d\boldsymbol{v}d\boldsymbol{x}dt}_{=0\text{ (divergence integrates to 0)}}
\end{align}
where the divergence $\nabla_{\boldsymbol{v}}\cdot \hat{\boldsymbol{a}}$ integrated over $\mathbb{R}^d$ is 0 since probability mass is conserved, and since there are no boundaries on velocity, the flux must be 0. Now, substituting the continuity equation from (\ref{eq:continuity-ahat}), we get:
\begin{align}
     \int_0^T\int\frac{1}{2}(\nabla_{\boldsymbol{v}} \cdot (p_t\hat{\boldsymbol{a}}))d\boldsymbol{v}d\boldsymbol{x}dt&=-\int_0^T\int\frac{1}{2}\partial_tp_t(\boldsymbol{x}, \boldsymbol{v})d\boldsymbol{v}d\boldsymbol{x}dt-\underbrace{\int_0^T\int\frac{1}{2}\boldsymbol{v}\cdot \nabla p_td\boldsymbol{v}d\boldsymbol{x}dt}_{= 0}\nonumber\\
     &=-\int_0^T\int\frac{1}{2}\partial_tp_t\log p_td\boldsymbol{v}d\boldsymbol{x}dt\nonumber\\
     &=-\int_0^T\frac{1}{2}\partial_t\int p_t\log p_td\boldsymbol{v}d\boldsymbol{x}dt
\end{align}
Defining $F(t):=\int p_t\log p_td\boldsymbol{v}d\boldsymbol{x}$ gives $\partial_tF(t)=\partial_t\int p_t\log p_td\boldsymbol{v}d\boldsymbol{x}$, so we can apply the Fundamental Theorem of Calculus which states $\int_0^TF'(t)dt=F(T) -F(0)$ to get:
\begin{align}
    \int_0^T\int \frac{1}{2}\langle \hat{\boldsymbol{a}}, \nabla_{\boldsymbol{v}}\log p_t\rangle p_td\boldsymbol{x}d\boldsymbol{v}dt&=\int \frac{1}{2}\left(p_T\log p_T-p_0\log p_0\right)d\boldsymbol{v}d\boldsymbol{x}
\end{align}
Finally, substituting this back into (\ref{eq:expanded-squared-a}), we get the new form of the cost function with respect to $\hat{\boldsymbol{a}}_t$:
\begin{align}
    \underset{\boldsymbol{a}_t}{\arg\min}\int_0^T\int \left[\frac{1}{2}\|\boldsymbol{a}_t\|^2 p_t+\frac{\sigma^2_t}{8}\|\nabla_{\boldsymbol{v}}\log p_t\|^2p_t\right]d\boldsymbol{x}d\boldsymbol{v}dt+\underbrace{\int \frac{1}{2}\left(p_T\log p_T-p_0\log p_0\right)d\boldsymbol{x}d\boldsymbol{v}}_{\text{constant}}
\end{align}
In the SBM setting, the phase space marginals are constrained to $p_0=\pi_0$ and $p_T=\pi_T$, so we can drop the second term of the minimization, and write the multi-marginal SB problem as: 
\begin{align}
    \boldsymbol{a}^\star_t&=\underset{\boldsymbol{a}_t}{\arg\min}\int_0^T\int \left[\frac{1}{2}\|\boldsymbol{a}_t\|^2 p_t+\frac{\sigma^2_t}{8}\|\nabla_{\boldsymbol{v}}\log p_t\|^2p_t\right]d\boldsymbol{x}d\boldsymbol{v}dt\\ &\text{s.t.}\quad \begin{cases}
            \partial_tp_t(\boldsymbol{x},\boldsymbol{v})+\boldsymbol{v}\cdot \nabla p_t+\nabla_{\boldsymbol{v}}\cdot (\hat{\boldsymbol{a}}_tp_t)=0\\
            \mu(\boldsymbol{x}, t_k)=\int_{\mathbb{R}^d}\pi_{t_k}(\boldsymbol{x}, \boldsymbol{v})d\boldsymbol{v}, \quad \forall k\in \{0, \dots, K\}
        \end{cases}\label{eq:mmsb-fpe-final}
\end{align}
This derivation shows that the multi-marginal Schrödinger bridge problem in phase space can be reformulated as an entropy-regularized dynamic optimal transport problem over joint position–velocity distributions. The quadratic acceleration cost $\frac{1}{2}\|\boldsymbol{a}_t\|^2 $ governs how trajectories bend in phase space, penalizing deviations from inertial motion and enforcing smooth transitions that interpolate through the prescribed intermediate marginals $\{\pi_{t_k}\}_{k=1}^K$. In addition, the Fisher information term $\frac{\sigma_t^2}{8}\|\nabla_{\boldsymbol{v}}\log p_t\|^2$ acts specifically in the velocity coordinate, regularizing the uncertainty and dispersion of velocities induced by stochasticity.

Just like how the static and dynamic Schrödinger bridge problems are an entropically regularized analogue of static and dynamic optimal transport, the multi-marginal SB problem can be interpreted as the stochastic, entropically regularized analogue of \boldtext{measure-valued splines}. This perspective will help us understand the behavior of multi-marginal SB in the zero-noise limit. 

\purple[Connection to Measure-Valued Splines]{
The \boldtext{variational spline problem} aims to select the smoothest curve $(\boldsymbol{x}(t))_{t\in [0,T]}$ that interpolates between a set of sequential points $\bar{\boldsymbol{x}}_{t_0}, \dots, \bar{\boldsymbol{x}}_{t_K}\in \mathbb{R}^d$ at times $0=t_0<\dots<t_k<\dots < t_K=T$ by minimizing:
\begin{small}
\begin{align}
    \min_{(\boldsymbol{x}_t)_{t\in [0,T]}}\int_0^T\|\ddot{\boldsymbol{x}}_t\|^2 dt \quad \text{s.t.}\quad \boldsymbol{x}_{t_k}=\bar{\boldsymbol{x}}_{t_k},\forall k \in \{0, \dots, K\}
\end{align}
\end{small}
where the minimization is taken over all twice-differentiable curves satisfying the interpolation constraints. The objective penalizes squared acceleration $\|\ddot{\boldsymbol{x}}_t\|^2 $, so the resulting trajectory is the smoothest curve connecting the prescribed points in the sense of minimizing the \textbf{total bending energy}.

This principle can be generalized from deterministic trajectories to evolving probability distributions by replacing the single curve $\boldsymbol{x}(t)$ with a time-dependent probability density $p_t(\boldsymbol{x},\boldsymbol{v})$ defined over the phase space of particle positions and velocities $(\boldsymbol{x}, \boldsymbol{v})\in \mathbb{R}^{2d}$. Then, the distributional analogue of the variational spline problem, known as the \boldtext{measure-valued spline problem}, becomes:
\begin{small}
\begin{align}
    \inf_{\boldsymbol{a},p_t}&\left\{\int_0^T\int_{\mathbb{R}^d}\int_{\mathbb{R}^d}\|\boldsymbol{a}(\boldsymbol{x},\boldsymbol{v},t)\|^2 p_t(\boldsymbol{x},\boldsymbol{v})d\boldsymbol{x}d\boldsymbol{v}dt\right\}\tag{Measure-Valued Spline Problem}\\
    &\text{s.t.}\quad \begin{cases}
        d\boldsymbol{x}_t=\boldsymbol{v}dt, \quad d\boldsymbol{v}_t=\boldsymbol{a}_tdt\\
        \partial_tp_t+\langle \boldsymbol{v}, \nabla p_t\rangle +\nabla_{\boldsymbol{v}}\cdot (\boldsymbol{a}p_t)=0\\
        \int_{\mathbb{R}^d} p_t(\boldsymbol{x},\boldsymbol{v})d\boldsymbol{v}=\pi_{t_k}, \forall k\in \{0, \dots, K \}
    \end{cases}\nonumber
\end{align}
\end{small}
where $\partial_tp_t+\langle \boldsymbol{v}, \nabla p_t\rangle +\nabla_{\boldsymbol{v}}\cdot (\boldsymbol{a}p_t)=0$ is the phase-space continuity equation and $\int_{\mathbb{R}^d} p_t(\boldsymbol{x},\boldsymbol{v})d\boldsymbol{v}=\pi_{t_k}$ enforces the position marginals. Intuitively, this problem yields the evolution of a probability distribution that passes through prescribed marginal distributions while minimizing the average squared acceleration of the particles.

Notice that this problem is exactly the (\ref{eq:multi-marginal-sbp}) except with zero diffusion $\sigma_t\equiv 0$, which reduces the Fokker-Planck equation with the Laplacian term $\frac{\sigma_t^2}{2}\Delta_{\boldsymbol{v}}p_t\equiv 0$ into the classic phase-space continuity equation. To this end, the multi-marginal SB problem can be interpreted as a stochastic, entropy-regularized analogue of the measure-valued spline.
}

This formulation extends the dynamic Schrödinger bridge problem along two key directions: incorporating multiple intermediate marginal constraints and lifting the dynamics to phase space, where marginals can encode both position and velocity information. In the zero-noise limit, the problem recovers a deterministic measure-valued spline problem, revealing how SB generalizes smooth trajectory interpolation between distributions with stochastic uncertainty. 

\subsection{Unbalanced Schrödinger Bridge Problem}
\label{subsec:unbalanced-sbp}
Since the Schrödinger bridge problems originate from the optimal mass transport (OMT) problem, where the probability mass across the full time horizon is conserved, and no mass is lost. In many applications, such as cell dynamics, probability mass is not necessarily conserved, and particles undergo growth and death, producing \textbf{unbalanced marginal distributions}. However, this presents the \textit{key problem} of determining the way in which particles should transport and vanish along intermediate time points that minimizes the deviation from some reference dynamics while reconstructing the unbalanced marginals.

To account for the difference in mass between terminal marginals, we can relax the mass conservation constraint in the standard regularized OT problem by introducing a growth rate $g(\boldsymbol{x},t): \mathbb{R}^d\times[0,T]\to \mathbb{R}$ into the minimization objective to get the \boldtext{dynamic unbalanced optimal transport} problem \citep{chizat2018interpolating, chizat2018unbalanced}.

\begin{definition}[Dynamic Unbalanced Optimal Transport Problem]
Let $\pi_0,\pi_T\in \mathcal{M}_+(\mathbb{R}^d)$ be non-negative measures that may have different total mass. The \textbf{dynamic unbalanced optimal transport problem} seeks a time-dependent density $p_t$, transport velocity $\boldsymbol{v}(\boldsymbol{x},t):\mathbb{R}^d\times[0,T] \to \mathbb{R}^d$, and growth rate $g(\boldsymbol{x},t):\mathbb{R}^d\times[0,T] \to \mathbb{R}$ that solve the minimization:
\begin{align}
    &\inf_{p_t, \boldsymbol{v}, g}\int_0^T\int_{\mathbb{R}^d}\left\{\frac{1}{2}\|\boldsymbol{v}(\boldsymbol{x},t)\| ^2+\alpha\Psi(g(\boldsymbol{x},t))\right\}p_t(\boldsymbol{x})d\boldsymbol{x}dt\\
    &\text{s.t.}\quad \begin{cases}
        \partial_t p_t(\boldsymbol{x})=-\nabla\cdot (p_t(\boldsymbol{x})\boldsymbol{v}(\boldsymbol{x},t))+g(\boldsymbol{x},t)p_t(\boldsymbol{x})\\
        p_0=\pi_0, \quad p_T=\pi_T
    \end{cases}
\end{align}
where $\Psi:\mathbb{R}\to \mathbb{R}_{\geq 0}$ is a non-negative function that penalizes changes in mass and $\alpha> 0$ is a hyperparameter that controls the penalty.
\end{definition}
Notice that the \textbf{unbalanced continuity equation} constraint in the dynamic unbalanced OT problem $\partial_t p_t=-\nabla\cdot (p_t\boldsymbol{v})+gp_t$ contains an additional $gp_t$ term compared to the standard continuity equation. This term models local mass creation and destruction, where the current density $p_t(\boldsymbol{x})$ at $\boldsymbol{x}$ grows or decays proportionally to $g(\boldsymbol{x},t)$. Therefore, $gp_t$ accounts for the instantaneous change in mass density due to growth or decay, allowing the total mass to vary over time.

The growth penalty $\Psi(g)$, commonly defined as the quadratic growth $\Psi(g(\boldsymbol{x},t)):= |g(\boldsymbol{x},t)|^2$, regularizes the amount of mass creation or destruction along the transport path. By penalizing large growth rates, the optimization balances mass transport and mass variation, yielding the \textbf{minimal-cost combination of transport and local mass change} required to match the terminal marginals.

\begin{figure}
    \centering
    \includegraphics[width=\linewidth]{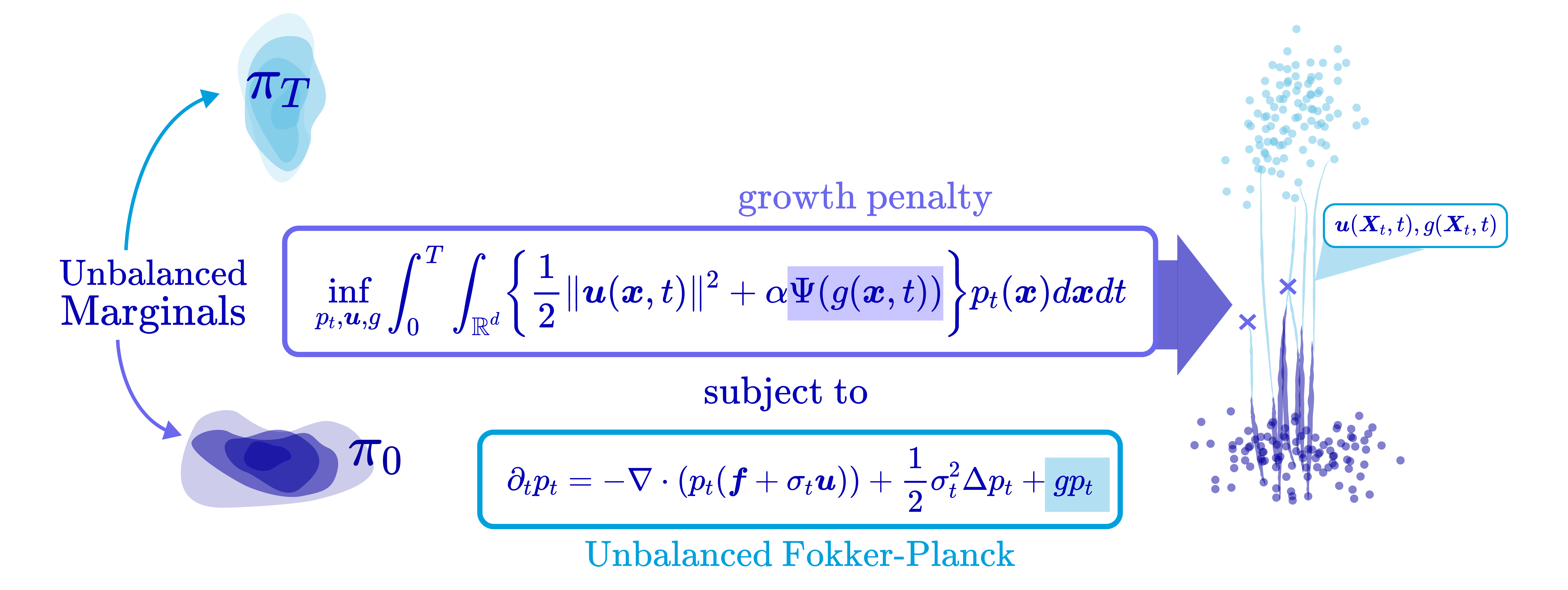}
    \caption{\textbf{Unbalanced Schrödinger Bridge Problem.} The unbalanced SB problem extends the classical formulation to settings where the endpoint marginals $\pi_0$ and $\pi_T$ have different total mass. In addition to the control $\boldsymbol{u}(\boldsymbol{x},t)$ that transports probability, the dynamics contain a growth term $g(\boldsymbol{x},t)$ which allows mass to be created or removed along trajectories, which contributes a growth penalty $\Psi(g)$ in the minimization objective. The dynamics of the density evolve according to the unbalanced Fokker–Planck equation, which augments the standard continuity equation with the mass-change term $gp_t$.}
    \label{fig:unbalanced}
\end{figure}

Extending this formulation to SDEs of the form $d\boldsymbol{X}_t=(\boldsymbol{f}(\boldsymbol{X}_t,t)+ \sigma_t\boldsymbol{u}(\boldsymbol{X}_t,t)) dt+\sigma_td\boldsymbol{B}_t$ yields the \boldtext{unbalanced Schrödinger bridge} problem, which seeks the \textbf{most likely stochastic process} between two unbalanced marginals \citep{zhang2024learning, chen2022most, pariset2023unbalanced}.

\begin{definition}[Unbalanced Schrödinger Bridge Problem]\label{def:regularized-unbalanced-sbp}
    Given a pair of unbalanced marginals $\pi_0, \pi_T\in \mathcal{P}(\mathbb{R}^d)$, a reference path measure $\mathbb{Q}$ with drift $\boldsymbol{f}(\boldsymbol{x},t)$ and a function that penalizes the growth rate $\Psi: \mathbb{R}\to \mathbb{R}_{\geq 0}$, the \textbf{unbalanced Schrödinger bridge (SB) problem} aims to determine the optimal tuple density evolution $p^\star_t$, control drift $\boldsymbol{u}^\star$, and growth rate $g^\star$ that solve the minimization problem:
    \begin{align}
        &\inf_{p_t, \boldsymbol{u}, g}\int_0^T\int_{\mathbb{R}^d}\left[\frac{1}{2}\|\boldsymbol{u}(\boldsymbol{x},t)\|^2+\alpha\Psi(g(\boldsymbol{x},t))\right]p_t(\boldsymbol{x})d\boldsymbol{x}dt\tag{Unbalanced SB Problem}\label{eq:unbalanced-sbp}
    \end{align}
    subject to the marginal constraints $p_0=\pi_0$ and $p_T=\pi_T$ and an \textbf{unbalanced Fokker-Planck constraint} defined as:
    \begin{align}
        \partial_tp_t(\boldsymbol{x})=-\nabla\cdot (p_t(\boldsymbol{x})(\boldsymbol{f}(\boldsymbol{x},t)+\sigma_t\boldsymbol{u}(\boldsymbol{x},t)))+\frac{\sigma^2_t}{2} \Delta p_t(\boldsymbol{x})+g(\boldsymbol{x},t)p_t(\boldsymbol{x})\tag{Unbalanced FP Equation}\label{eq:unbalanced-fpe}
    \end{align}
    where $g(\boldsymbol{x},t)p_t(\boldsymbol{x})$ relaxes the mass conservation constraint of the standard Fokker-Planck equation to allow the mass to change proportionally to the growth rate.
\end{definition}

Since the Schrödinger bridge formulation has three optimization parameters, the density $p_t$, the control drift $\boldsymbol{u}$, \textit{and the growth rate} $g$ which are all coupled in the (\ref{eq:unbalanced-fpe}), optimizing a stochastic flow yields a difficult optimization problem. By reparameterizing the unbalanced Fokker-Planck constraint into a \textit{unbalanced continuity equation} constraint, we can derive an alternative \textbf{entropy-regularized dynamic optimal transport form} of the (\ref{eq:unbalanced-sbp}), which was introduced to model stochastic unbalanced dynamical systems \citep{baradat2021regularized, buze2023entropic, chen2022most, janati2020entropic}, and has recently been used to develop scalable computational methods for simulating high-dimensional real-world data \citep{zhang2024learning}.

\begin{proposition}[Entropy-Regularized Unbalanced Dynamic Optimal Transport Problem]\label{prop:regularized-unbalanced-sbp}
    The (\ref{eq:unbalanced-sbp}) from Definition \ref{def:regularized-unbalanced-sbp} can be written as:
    \begin{small}
    \begin{align}
        &\inf_{(p_t, \boldsymbol{v}, g)}\int_0^T\int_{\mathbb{R}^d}\bigg[\frac{1}{2}\|\boldsymbol{v}\|^2+\frac{\sigma_t^2}{8}\|\nabla \log  p_t\|^2\bluetext{-\frac{1}{2}\big\langle\nabla\log p_t,\boldsymbol{f}\big\rangle -\frac{1}{2}(1+\log p_t)g}+\alpha\Psi(g)\bigg]p_td\boldsymbol{x}dt\tag{Regularized Unbalanced OT }
    \end{align}
    \end{small}
    subject to the same marginal constraints $p_0=\pi_0$ and $p_T=\pi_T$ and the \textbf{unbalanced continuity equation constraint} defined as:
    \begin{align}
        \partial_tp_t(\boldsymbol{x})=-\nabla\cdot(p_t(\boldsymbol{x})(\boldsymbol{f}(\boldsymbol{x},t)+\sigma_t\boldsymbol{v}(\boldsymbol{x},t))+g(\boldsymbol{x},t)p_t(\boldsymbol{x})
    \end{align}
    where $g(\boldsymbol{x},t)p_t(\boldsymbol{x})$ relaxes the mass conservation constraint of the standard (\ref{eq:continuity-equation-remark})\footnote{Note that the regularized unbalanced optimal transport objective introduced in \citep{zhang2024learning} is slightly different than the one stated here as we parameterize the control drift rather than the full velocity field, which results in an additional term containing the reference drift $\boldsymbol{f}$ and different coefficient scaling.}.
\end{proposition}

\textit{Proof.} Starting with the (\ref{eq:unbalanced-fpe}), we follow similar steps as in (\ref{eq:entropy-dynamic-ot-proof1}) to derive a reparameterized form of the control drift that satisfies a \textit{continuity equation constraint} with the additional growth term $g(\boldsymbol{x},t)p_t(\boldsymbol{x})$ as follows:
\begin{align}
    \partial_tp_t(\boldsymbol{x})&=-\nabla\cdot\big(p_t(\boldsymbol{x})(\boldsymbol{f}(\boldsymbol{x},t) +\bluetext{\sigma_t\boldsymbol{u}(\boldsymbol{x},t)})\big)+\bluetext{\frac{\sigma_t^2}{2}\Delta p_t(\boldsymbol{x})}+\pinktext{g(\boldsymbol{x},t)p_t(\boldsymbol{x})}\nonumber\\
    &=-\nabla\cdot\bigg(p_t(\boldsymbol{x})\bigg(\boldsymbol{f}(\boldsymbol{x},t) +\sigma_t\bluetext{\underbrace{\left(\boldsymbol{u}(\boldsymbol{x},t)-\frac{\sigma_t}{2}\nabla \log p_t(\boldsymbol{x})\right)}_{=:\boldsymbol{v}(\boldsymbol{x},t)}}\bigg)\bigg)+\pinktext{g(\boldsymbol{x},t)p_t(\boldsymbol{x})}\nonumber\\
    &=-\nabla\cdot(p_t(\boldsymbol{x})(\boldsymbol{f}(\boldsymbol{x},t)+\sigma_t\boldsymbol{v}(\boldsymbol{x},t))+\pinktext{g(\boldsymbol{x},t)p_t(\boldsymbol{x})}\tag{Unbalanced Continuity Equation}
\end{align}
which is the unbalanced form of the Fokker-Planck constraint in (\ref{prop:regularized-unbalanced-sbp}). Defining $\boldsymbol{v}(\boldsymbol{x},t):=\boldsymbol{u}(\boldsymbol{x},t)-\frac{\sigma_t}{2}\nabla \log p_t(\boldsymbol{x})$ and rearranging to get the change-in-variables $\boldsymbol{u}(\boldsymbol{x},t)=\boldsymbol{v}(\boldsymbol{x},t)+\frac{\sigma_t}{2}\nabla \log  p_t(\boldsymbol{x})$, we write the objective from Definition \ref{def:regularized-unbalanced-sbp} as:\label{connection back to ot formulation}
\begin{small}
\begin{align}
    &\inf_{(p_t, \boldsymbol{v}, g)}\int_0^T\int_{\mathbb{R}^d}\left[\frac{1}{2}\|\boldsymbol{u}(\boldsymbol{x},t)\|^2+\alpha\Psi(g(\boldsymbol{x},t))\right]p_t(\boldsymbol{x})d\boldsymbol{x}dt\nonumber\\
    &=\inf_{(p_t, \boldsymbol{v}, g)}\int_0^T\int_{\mathbb{R}^d}\left[\frac{1}{2}\bigg\|\boldsymbol{v}(\boldsymbol{x},t)+\sigma_t\nabla \log  p_t(\boldsymbol{x})\bigg\|^2+\alpha\Psi(g(\boldsymbol{x},t))\right]p_t(\boldsymbol{x})d\boldsymbol{x}dt\nonumber\\
    &=\inf_{(p_t, \boldsymbol{v}, g)}\int_0^T\int_{\mathbb{R}^d}\left[\frac{1}{2}\left(\|\boldsymbol{v}(\boldsymbol{x},t)\|+\sigma_t\langle \boldsymbol{v}(\boldsymbol{x},t),\nabla \log  p_t(\boldsymbol{x})\rangle + \frac{\sigma_t^2}{4}\|\nabla \log  p_t(\boldsymbol{x})\|^2\right)+\alpha\Psi(g(\boldsymbol{x},t))\right]p_t(\boldsymbol{x})d\boldsymbol{x}dt\nonumber\\
    &=\inf_{(p_t, \boldsymbol{v}, g)}\int_0^T\int_{\mathbb{R}^d}\bigg\{\underbrace{\frac{1}{2}\|\boldsymbol{v}(\boldsymbol{x},t)\|^2 }_{\text{kinetic energy}}+\underbrace{\frac{\sigma_t}{2}\left\langle\boldsymbol{v}(\boldsymbol{x},t), \nabla \log p_t(\boldsymbol{x})\right \rangle}_{\text{cross term}}+\underbrace{\frac{\sigma_t^2}{8}\|\nabla \log  p_t(\boldsymbol{x})\|^2}_{\text{Fisher information}} +\underbrace{\alpha\Psi(g(\boldsymbol{x},t))}_{\text{growth penalty}}\bigg\}p_t(\boldsymbol{x})d\boldsymbol{x}dt\label{eq:unbalanced-proof-2}
\end{align}
\end{small}
which follows the same steps as the derivation in Section \ref{subsec:sb-regularized-dynamic-ot}. The expansion of the cross term deviates from the derivation in Section \ref{subsec:sb-regularized-dynamic-ot} as it incorporates the additional growth term in the continuity equation substitution:
\begin{small}
\begin{align}
    H(p_T)-H(p_0)&\overset{(\ref{eq:ot-sb-proof2})}{=}\int_0^T\int_{\mathbb{R}^d}(1+\log p_t)\bluetext{\partial_tp_t}d\boldsymbol{x}dt
\end{align}
\end{small}
Substituting the \textbf{unbalanced continuity equation} constraint $\partial_tp_t=-\nabla\cdot(p_t(\boldsymbol{f}+\sigma_t\boldsymbol{v}))+gp_t$, we have:
\begin{small}
\begin{align}
    H(p_T)-H(p_0)
    &=\int_0^T\int_{\mathbb{R}^d}(1+\log p_t)\bluetext{\big(-\nabla\cdot(p_t(\boldsymbol{f}+\sigma_t\boldsymbol{v})+gp_t\big)}d\boldsymbol{x}dt\nonumber\\
    &=\int_0^T\int_{\mathbb{R}^d}(1+\log p_t)\bluetext{\big(-\nabla\cdot(p_t(\boldsymbol{f}+\sigma_t\boldsymbol{v})\big)}d\boldsymbol{x}dt+\int_0^T\int_{\mathbb{R}^d}(1+\log p_t)\bluetext{(gp_t)}d\boldsymbol{x}dt
\end{align}
\end{small}
Then, applying the integration of parts identity on the divergence and following the steps in (\ref{eq:ot-sb-proof3}), we get:
\begin{small}
\begin{align}
    H(p_T)&-H(p_0)=\int_0^T\int_{\mathbb{R}^d}\bigg\langle\nabla\log p_t,\bluetext{p_t(\boldsymbol{f}+\sigma_t\boldsymbol{v})}\bigg\rangle d\boldsymbol{x}dt+\int_0^T\int_{\mathbb{R}^d}(1+\log p_t)(gp_t)d\boldsymbol{x}dt\nonumber\\
    &=\int_0^T\int_{\mathbb{R}^d}\big\langle\nabla\log p_t,\boldsymbol{f}\big\rangle p_td\boldsymbol{x}dt+\int_0^T\int_{\mathbb{R}^d}\sigma_t\big\langle\nabla\log p_t,\boldsymbol{v}\big\rangle p_td\boldsymbol{x}dt+\int_0^T\int_{\mathbb{R}^d}(1+\log p_t)(gp_t)d\boldsymbol{x}dt
\end{align}
\end{small}
Dividing both sides by two and rearranging to isolate the cross term in (\ref{eq:unbalanced-proof-2}), we have:
\begin{small}
\begin{align}
     \int_0^T\int_{\mathbb{R}^d}\frac{\sigma_t}{2}\big\langle\nabla\log p_t,\boldsymbol{v}\big\rangle &p_td\boldsymbol{x}dt=\frac{1}{2}(H(p_T)-H(p_0)) +\int_0^T\int_{\mathbb{R}^d}\frac{1}{2}\bigg[-\big\langle\nabla\log p_t,\boldsymbol{f}\big\rangle \bluetext{p_t} -(1+\log p_t)(g\bluetext{p_t})\bigg]d\boldsymbol{x}dt\nonumber\\
     &=\underbrace{\frac{1}{2}(H(p_T)-H(p_0))}_{\text{constant}} +\int_0^T\int_{\mathbb{R}^d}\bigg[-\frac{1}{2}\big\langle\nabla\log p_t,\boldsymbol{f}\big\rangle -\frac{1}{2}(1+\log p_t)g\bigg]\bluetext{p_t}d\boldsymbol{x}dt
\end{align}
\end{small}
Since the entropy difference is fixed given the marginal constraints $p_0=\pi_0$ and $p_T=\pi_T$, the entropy difference term can be dropped, and we get: 
\begin{align}
    \boxed{\inf_{(p_t, \boldsymbol{u}, g)}\int_0^T\int_{\mathbb{R}^d}\bigg[\frac{1}{2}\|\boldsymbol{v}\|^2+\frac{\sigma_t^2}{8}\|\nabla \log  p_t\|^2\bluetext{-\frac{1}{2}\big\langle\nabla\log p_t,\boldsymbol{f}\big\rangle -\frac{1}{2}(1+\log p_t)g}+\alpha\Psi(g)\bigg]p_td\boldsymbol{x}dt}
\end{align}
subject to the marginal constraints $p_0=\pi_0$, $p_T=\pi_T$, and the (\ref{eq:unbalanced-fpe}), which reformulates the unbalanced SB problem as a \textbf{entropy-regularized dynamic OT problem}. \hfill $\square$

The reformulation removes the coupling between $\boldsymbol{v}$ and $\nabla \log p_t$ in the cross interaction term $\langle\boldsymbol{v}(\boldsymbol{x},t), \sigma_t^2\nabla \log p_t(\boldsymbol{x}) \rangle$, leading to a more numerically stable and computationally tractable objective \citep{zhang2024learning}. We can also observe that since the terminal marginals are fixed, the entropy difference is constant, and the objective depends solely on the path-dependent cost. This results in a problem with three \textbf{key components}: the quadratic cost of the velocity field $\frac{1}{2}\|\boldsymbol{v}\|^2$, the Fisher information term that penalizes sharp changes in density, and the contribution of the growth rate determined by the growth penalty $\Psi(g)$ and the entropy change caused by creating or destroying mass through $(1+\log p_t)$.

\subsection{Branched Schrödinger Bridge Problem}
\label{subsec:branched-sbp}
Orthogonal to the problem of matching multiple subsequent marginals along the temporal evolution of probability mass is the problem of matching a complex terminal marginal distribution $\pi_T$ with \textbf{multiple modes} $\pi_T=\pi_{T, 1}\oplus \dots \oplus \pi_{T,K}$. In practice, fitting a standard Schrödinger bridge to accurately transport density from an initial distribution $\pi_0$ to a multi-modal distribution $\pi_T$ suffers from several challenges.

\begin{figure}
    \centering
    \includegraphics[width=\linewidth]{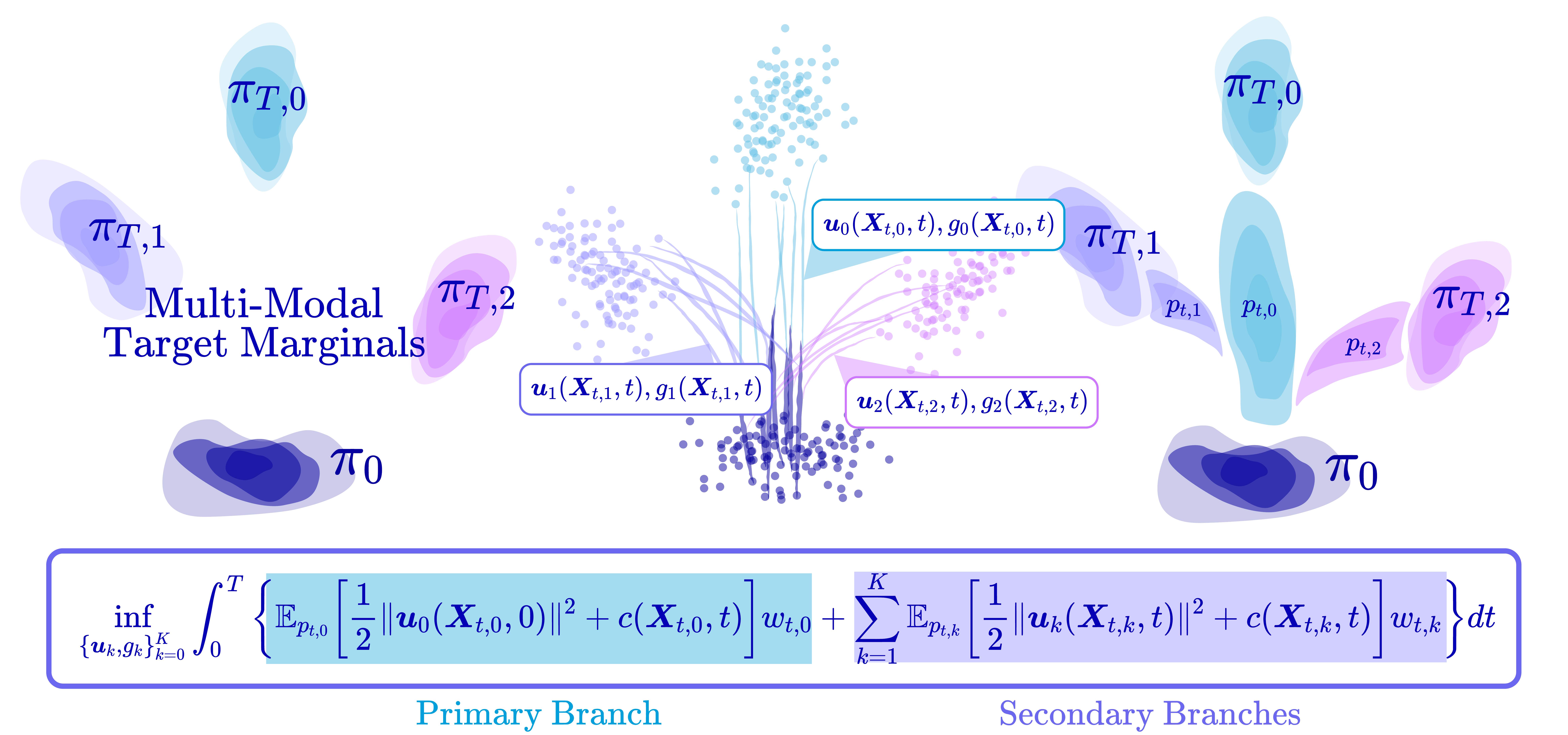}
    \caption{\textbf{Branched Schrödinger Bridge Problem.} The branched Schrödinger bridge generalizes the classical formulation to transport an initial distribution $\pi_0$ toward multiple target marginals $\{\pi_{T,k}\}_{k=1}^K$. A primary trajectory evolves from the source distribution and probabilistically branches into secondary branched paths, each governed by its own control and growth dynamics to reach a different terminal distribution. The objective minimizes the weighted kinetic energy of the controls across all branches, producing a stochastic branching flow that connects a single source to multiple target modes.}
    \label{fig:branched}
\end{figure}

One of these challenges is \textbf{mode collapse}, in which the probability density is concentrated in only one or a few terminal modes, thereby missing the full distribution. To overcome this, one may increase the number of particles simulated from the initial distribution to improve the likelihood of discovering each terminal mode; however, this increases computational cost and does not guarantee accurate reconstruction of all modes and \textit{their relative density weights}. Furthermore, Schrödinger bridges not only sample from the terminal distribution but also recover the energy-minimizing temporal bridge between the distributions. For multimodal target distributions, this optimal bridge can inform us about \textit{branching times} and \textit{mass redistribution}; however, it remains challenging to simulate effectively with standard SB methods. This is the motivation behind the \boldtext{Branched Schrödinger Bridge Problem} \citep{tang2026branchsbm}, which enables simulation of branching and mass redistribution over a stochastic bridge to reconstruct multi-modal target distributions.

Since branching can be interpreted as the unbalanced flow of probability mass from a primary branch to multiple diverging trajectories, it can be formalized as the \textit{sum} of Unbalanced Schrödinger bridges, where mass is progressively depleted from the primary branch and fed into the diverging paths to multiple terminal modes.

\begin{definition}[Branched Schrödinger Bridge Problem]\label{def:branched-sbp}
    Consider an initial distribution $\pi_0$ and a terminal distribution with $K$ distinct modes $\pi_T:=\pi_{T, 1}\oplus \dots\oplus \pi_{T, K}$ on some potential energy landscape defined by a state cost $c(\boldsymbol{x},t)$. Denoting the control drift of each branch as $\{\boldsymbol{u}_{ k}(\boldsymbol{x},t)\}_{k=0}^K$, the growth rate of each branch as $\{g_{k}(\boldsymbol{x},t)\}_{k=0}^K$, and the accumulated weight of each branch as $\{w_{t, k}\}_{k=1}^K$, the Branched SB problem seeks the optimal set of control and growth rates $\{\boldsymbol{u}^\star_{k},g_{k}^\star\}_{k=0}^K$ that solve the following minimization:
    \begin{small}
    \begin{align}
        &\inf_{\{\boldsymbol{u}_{k},g_{k}\}_{k=0}^K}\int_0^T\bigg\{\mathbb{E}_{p_{t, 0}}\left[\frac{1}{2}\|\boldsymbol{u}_{0}(\boldsymbol{X}_{t, 0},t)\|^2+c(\boldsymbol{X}_{t, 0},t)\right]w_{t, 0}+\\
        &\quad\quad \sum_{k=1}^K\mathbb{E}_{p_{t, k}}\left[\frac{1}{2}\|\boldsymbol{u}_{k}(\boldsymbol{X}_{t, k},t)\|^2+c(\boldsymbol{X}_{t,k},t)\right]w_{t, k}\bigg\}dt\tag{Branched SB Problem}\label{eq:branched-sbp}\\
        &\quad \quad \text{s.t.}\quad \begin{cases}
            d\boldsymbol{X}_{t, k}=(\boldsymbol{f}(\boldsymbol{X}_{t,k},t)+\sigma_t\boldsymbol{u}_{k}(\boldsymbol{X}_{t,k},t))dt+\sigma_td\boldsymbol{B}_t\\
            \boldsymbol{X}_0\sim \pi_0, \quad \boldsymbol{X}_{T, k}\sim \pi_{T, k}\\
            w_{0, k}=\delta_{k=0}, \quad w_{T,k}=w^\star_{T,k}
        \end{cases}\nonumber
    \end{align}
    \end{small}
    where the weight of the primary branch is given by $w_{t, 0}=1+\int_0^tg_0(\boldsymbol{X}_{s,0},s)ds$ and the weights of the $K$ secondary branches is given by $w_{t, k}=\int_0^tg_k(\boldsymbol{X}_{s,k},s)ds$.
\end{definition}

When considering a setting where the total mass across all branches is conserved, we can add the constraint $\sum_{k=0}^Kw_{t, k}=1$ for all $t\in [0,T]$ which enforces that the the growth rates sum to zero, i.e., $g_{0}(\boldsymbol{X}_{t, 0},t)+\sum_{k=1}^Kg_{k}(\boldsymbol{X}_{t, k},t)=0$. In this setting, \textit{all} the mass that is lost from the primary branch, such that growth rate is negative $g_{0}<0$, is redistributed to the secondary branches, such that at least one branch has positive growth $\exists k, \; g_{k}> 0$. To formulate a tractable form of the (\ref{eq:branched-sbp}), we can reframe the problem as a \textbf{conditional stochastic optimal control} (CondSOC) problem that can be tractably solved given a finite set of samples from an empirical initial distribution $\pi_0$ and a multi-modal terminal distribution $\pi_T$.

\begin{proposition}[Branched Conditional Stochastic Optimal Control (Proposition 2 in \citet{tang2026branchsbm})]
    Define the endpoint conditioned density of each branch as $p_{t, k}(\boldsymbol{X}_{t, k}):=\mathbb{E}_{\pi_{0,T,k}}[p_{t,k}(\boldsymbol{X}_{t,k}|\boldsymbol{x}_0, \boldsymbol{x}_{T,k})]$, where $\pi_{0,T, k}$ is the joint coupling between the initial distribution $\pi_0$ and the $k$th mode of the terminal distribution $\pi_{T,k}$. The set of optimal control drifts and growth terms $\{\boldsymbol{u}_{k}, g_{k}\}_{k=0}^K$ that solve the Branched SB problem defined in Definition \ref{def:branched-sbp} can be obtained by minimizing the sum of Unbalanced Conditional Stochastic Optimal Control problems defined as:
    \begin{align}
        \inf_{\{\boldsymbol{u}_{k},g_{k}\}_{k=0}^K}&\mathbb{E}_{(\boldsymbol{x}_0, \boldsymbol{x}_{T, 0})\sim \pi_{0,T,0}}\int_0^T\bigg\{\mathbb{E}_{p_{t|0,T, 0}}\left[\frac{1}{2}\|\boldsymbol{u}_{0}(\boldsymbol{X}_{t,0},t)\|^2+c(\boldsymbol{X}_{t,0},t)\right]w_{t,0}\bigg\}dt\\
        &+\sum_{k=1}^K\mathbb{E}_{(\boldsymbol{x}_0, \boldsymbol{x}_{T,k})\sim \pi_{0,T,k}}\int_0^T \bigg\{\mathbb{E}_{p_{t|0,T, k}}\left[\frac{1}{2}\|\boldsymbol{u}_{k}(\boldsymbol{X}_{t, k},t)\|^2+c(\boldsymbol{X}_{t,k},t)\right]w_{t, k}\bigg\}dt\nonumber\\
        &\text{s.t.}\quad\begin{cases}
            d\boldsymbol{X}_{t,k}=(\boldsymbol{f}(\boldsymbol{X}_{t,k},t)+\sigma_t\boldsymbol{u}_{k}(\boldsymbol{X}_{t,k},t))dt +\sigma_td\boldsymbol{B}_t\\
            \boldsymbol{X}_0=\boldsymbol{x}_0, \quad \boldsymbol{X}_{T, k}=\boldsymbol{x}_{T, k}\\
            w_{0, k}=\delta_{k=0}, \quad w_{T, k}=w^\star_{T,k}
        \end{cases}\nonumber
    \end{align}
    where the weight of the primary branch is given by $w_{t, 0}=1+\int_0^tg_0(\boldsymbol{X}_{s,0},s)ds$ and the weights of the $K$ secondary branches is given by $w_{t, k}=\int_0^tg_k(\boldsymbol{X}_{s,k},s)ds$.
\end{proposition}

\textit{Proof.} To prove that the branched CondSOC problem solves the branched SB problem in Definition \ref{def:branched-sbp}, we start by defining each branch $k$ as solving its own Unbalanced SB problem, defined as:
\begin{align}
    &\inf_{\boldsymbol{u}_{k}, g_{k}}\int_0^T\bigg\{\mathbb{E}_{p_{t,k}}\left[\frac{1}{2}\|\boldsymbol{u}_{k}(\boldsymbol{X}_{t,k})\|^2+c(\boldsymbol{X}_{t,k},t)\right]\left(w_{0,k}+\int_0^tg_{s,k}(\boldsymbol{X}_{s,k}) ds\right)\bigg\}dt\\
    &\text{s.t.}\begin{cases}
        \partial_t p_{t,k}=-\nabla\cdot(p_{t,k}(\boldsymbol{f}+\sigma_t\boldsymbol{u}_{k}))+\frac{\sigma_t^2}{2}\Delta p_{t,k}+g_{k}p_{t,k}\\
        p_0=\pi_0, \quad p_{T,k}=\pi_{T,k}
    \end{cases}
\end{align}
Now, it suffices to show that the sum of unbalanced CondSOC problems satisfies the \textbf{global Fokker-Planck equation} of the density over all branches, defined as: 
\begin{align}
    \partial_tp_t(\boldsymbol{X}_t)=-\nabla\cdot(p_t(\boldsymbol{X}_t)(\boldsymbol{f}+\sigma_t\boldsymbol{u}_t)(\boldsymbol{X}_t,t))+\frac{\sigma_t^2}{2}\Delta p_t(\boldsymbol{X}_t)\label{eq:fpe-full}
\end{align}
where the probability density $p_t$ is defined as the weighted sum of the density at each branch at time $t$ given by $p_t(\boldsymbol{X}_t):=\sum_{k=0}^Kw_{t,k}p_{t,k}(\boldsymbol{X}_t)$. To derive the Fokker-Planck equation of $p_t$, we start by differentiating with respect to $t$ and applying the chain rule to get:
\begin{small}
\begin{align}
    \partial_tp_t&=\partial_t\left[\sum_{k=0}^Kw_{t,k}p_{t,k}\right]=\sum_{k=0}^K\bigg[w_{t,k}\underbrace{\left(\partial_tp_{t,k}\right)}_{\text{branched FP}}+\underbrace{\left(\partial_tw_{t,k}\right)}_{:=g_{k}(\boldsymbol{X}_t)}p_{t,k}\bigg]\nonumber\\
    &=\sum_{k=0}^K\bigg[w_{t,k}\left(-\nabla\cdot ((\boldsymbol{f}+\sigma_t\boldsymbol{u}_{k})p_{t,k})+\frac{\sigma_t^2}{2}\Delta p_{t,k}\right)+g_{k}p_{t,k}\bigg]\nonumber\\
    &=\sum_{k=0}^K\bigg[-w_{t,k}\nabla\cdot((\boldsymbol{f}+\sigma_t\boldsymbol{u}_{k})p_{t,k})+w_{t,k}\frac{\sigma_t^2}{2}\Delta p_{t,k}+g_{k}p_{t,k}\bigg]\nonumber\\
    &=\underbrace{\sum_{k=0}^K(-w_{t,k}\nabla \cdot((\boldsymbol{f}+\sigma_t\boldsymbol{u}_{k})p_{t,k}))}_{\text{divergence}}+\underbrace{\sum_{k=0}^Kw_{t,k}\frac{\sigma_t^2}{2}\Delta p_{t,k}}_{\text{diffusion}}+\underbrace{\sum_{k=0}^K g_{k}p_{t,k}}_{\text{growth}}\label{eq:fpe-three-terms}
\end{align}
\end{small}
First, we rewrite the divergence term into the form of the full Fokker-Planck equation (\ref{eq:fpe-full}) by applying the linearity of the divergence operator:
\begin{align}
    \sum_{k=0}^K\big(-w_{t,k}\nabla \cdot((\boldsymbol{f}+\sigma_t\boldsymbol{u}_k)p_{t,k})\big)&=-\nabla\cdot \left(\sum_{k=0}^K w_{t,k}p_{t,k}(\boldsymbol{f}+\sigma_t\boldsymbol{u}_k)\right)\nonumber\\
    &=-\nabla \cdot \bigg(\boldsymbol{f}\sum_{k=0}^K w_{t,k}p_{t,k}+\sigma_t\sum_{k=0}^K w_{t,k}\boldsymbol{u}_{k}p_{t,k}\bigg)\nonumber\\
    &=-\nabla \cdot \bigg(\boldsymbol{f}\underbrace{\frac{1}{p_t}\sum_{k=0}^K w_{t,k}p_{t,k}}_{=1}+\sigma_t\underbrace{\frac{1}{p_t}\sum_{k=0}^K w_{t,k}\boldsymbol{u}_{k}p_{t,k}}_{:=\boldsymbol{u}}\bigg)p_t\nonumber\\
    &=\boxed{-\nabla\cdot \left((\boldsymbol{f}+\sigma_t\boldsymbol{u})p_t\right)} 
\end{align}
where the final equality follows from multiplying and dividing by $p_t$. By defining $\boldsymbol{u}:=\frac{1}{p_t}\sum_{k=0}^K w_{t,k}\boldsymbol{u}_{k}p_{t,k}$, we have that the total control drift is the mass-weighted average of the control drift for each branch. This is theoretically grounded, as under the global context, the control drift $\boldsymbol{u}_{k}(\boldsymbol{X}_t,t)$ of a particle $\boldsymbol{X}_t=\boldsymbol{x}$ along a single branch $k$ should be scaled by its probability of being in branch $k$, given by $p_{t,k}(\boldsymbol{X}_t)$, the weight of the particle itself $w_{t,k}(\boldsymbol{X}_t)$, normalized by the total probability of the particle over all branches $p_t(\boldsymbol{X}_t)$.

Next, we rewrite the diffusion term as the diffusion of the full branched Fokker-Planck equation by applying the linearity of the Laplacian operator:
\begin{align}
    \sum_{k=0}^Kw_{t,k}\frac{\sigma_t^2}{2}\Delta p_{t,k}=\frac{\sigma_t^2}{2}\Delta \underbrace{\left(\sum_{k=0}^Kw_{t,k}p_{t,k}\right)}_{:=p_t}=\boxed{\frac{\sigma_t^2}{2}\Delta p_t}
\end{align}
Finally, for the growth term in (\ref{eq:fpe-three-terms}), we observe that it is simply the weighted sum of the growth over each branch $\sum_{k=0}^K g_{k}p_{t,k}$ and doesn't alter the direction or motion of the particle along the brnached fields in the global context. Therefore, all three terms in (\ref{eq:fpe-three-terms}) satisfy the global Fokker-Planck equation, and we finish the proof. \hfill $\square$

This derivation provides insight into how branched Schrödinger bridges behave under the global dynamics. We observe that even when each branch evolves via its own control and growth field, the overall system evolves as a single stochastic process whose probability density is a weighted superposition of all branches. From the perspective of the global Fokker-Planck constraint, branching does not introduce additional forces or discontinuities in the particle motion. Instead, it induces a mixture of control drifts, where the effective drift at any state is the probability-weighted average of the branch-specific controls.

This solves the challenge of mode collapse because each mode is generated with its own control drift, constrained by a terminal weight to ensure that the correct probability mass and distribution are reconstructed. Furthermore, the potential energy function $c(\boldsymbol{x},t)$ that governs the system dynamics are minimized at the \textit{optimal branched mass redistribution}, such that the optimal control and growth fields $\{\boldsymbol{u}^\star_{k}, g_{k}\}_{k=0}^K$ that solve the Branched SB problem in Definition \ref{def:branched-sbp} yields the branching trajectories and their relative weights such that they minimize the total energy required to reconstruct the terminal distribution. 

\subsection{Fractional Schrödinger Bridge Problem}
\label{subsec:fractional-sbp}

Up to this point, we have considered only stochastic differential equations (SDEs) with standard Brownian motion (BM) processes that are Markov, where each increment of the SDE is independent of all previous steps. This formulation is a specific design choice that ensures the tractability of the bridge solution and corresponding drift. While this choice is often a sufficient approximation, it ignores the effects of long-range temporal dependencies inherent in complex real-world systems. 

\boldtext{Fractional Brownian motion} (fBM) is a generalization of standard BM to non-memoryless processes, where each increment is \textit{dependent} on previous increments. This dependence is characterized by the \boldtext{Hurst index} ($H$), which determines the \textit{roughness} or \textit{pathwise regularity} of the dynamics and the \textit{magnitude of long-range dependencies}. A Hurst index of $H=0.5$ recovers the standard BM.

\begin{definition}[Fractional Brownian Motion \citep{levy1963random}]
    Given standard $d$-dimensional Brownian motion (BM) $(\boldsymbol{B}_t\in \mathbb{R}^d)_{t\in [0,T]}$, \textbf{fractional BM (fBM)} is a centered Gaussian process defined by integrating over the history of the BM:
    \begin{align}
        \boldsymbol{B}_t^H:=\frac{1}{\Gamma\left(H+\frac{1}{2}\right)}\int_0^t(t-s)^{H-\frac{1}{2}}d\boldsymbol{B}_s, \quad t\geq 0
    \end{align}
    where $\Gamma$ is the Gamma function and $H\in (0,1)$ is the Hurst index.
\end{definition}
We observe that the kernel $(t-s)^{H-\frac{1}{2}}$ injects time-dependent memory to the BM by weighting recent states ($s\approx t$) heavily and weighting distant states lightly. For $H> \frac{1}{2}$, the increments are \textbf{positively correlated}, resulting in smoother trajectories where each subsequent increment is more likely to be close to the previous increment. For $H<\frac{1}{2}$, the increments are \textbf{negatively correlated}, resulting in rougher trajectories where each subsequent increment aims to \textit{undo itself} or revert back to the mean. When $H=\frac{1}{2}$, the kernel reduces to $(t-s)^{H-\frac{1}{2}}=1$ which recovers the standard BM process.

Given the dependence on previous states, fBM is not Markov, resulting in an intractable drift for simulation. To overcome this, Markov approximations of fBM have been introduced \citep{harms2019affine, daems2023variational}, which approximate fBM by a weighted sum of \boldtext{Ornstein–Uhlenbeck} (OU) processes, which are a class of Markov BM processes with a restoring force pulling it back toward the mean. 

\begin{definition}[Markov Approximation of Fractional Brownian Motion \citep{harms2019affine, daems2023variational}]
    Given a $d$-dimensional fractional Brownian motion (fBM) $(\boldsymbol{B}_t^H)_{t\in [0,T]}$, we define a \textbf{Markov-approximate fBM} (MA-fBM) $(\widehat{\boldsymbol{B}}_t^H)_{t\in [0,T]}$ as:
    \begin{align}
        \widehat{\boldsymbol{B}}_t^H:=\sum_{k=1}^K\omega_k\boldsymbol{Y}_t^k, \quad\omega_1, \dots, \omega_K\in\mathbb{R}
    \end{align}
    where $\boldsymbol{Y}_t^k$ denote $K$ Ornstein–Uhlenbeck (OU) processes of the form:
    \begin{align}
        \boldsymbol{Y}_t^k:=\int_0^te^{-\gamma_k(t-s)}d\boldsymbol{B}_s, \quad d\boldsymbol{Y}_t^k=-\gamma_k\boldsymbol{Y}_t^kdt+d\boldsymbol{B}_t, \quad k=1, \dots, K\label{eq:def-Yt}
    \end{align}
    where  $\{\gamma_k\}_{k=1}^K$ are the mean reversion coefficients that determine how strongly the dynamics are pulled back to the mean. To approximate the fBM dynamics, $\gamma_k:=r^{k-n}$ for $r> 1$ and $n=\frac{K+1}{2}$, which produces a log-uniformly spread time grid of fast-reverting and slow-reverting OU processes.
\end{definition}

Intuitively, each OU process contributes one exponential memory scale, where short memory modes are captured by strong mean-reversion OU processes (large $\gamma_k$) and long memory modes are captured by weak mean-reversion OU processes (small $\gamma_k$). To determine the \textbf{optimal finite set} of coefficients $\{\omega_k\}_{k=1}^K$ for MA-fBM, one can directly minimize the \textit{expected squared error} between the true fBM and the OU approximation over time:
\begin{align}
    (\omega_1, \dots, \omega_K)=\underset{\omega_1, \dots, \omega_K}{\arg\min}\left\{\int_0^T\mathbb{E}_{\mathbb{P}}\left[\left(\boldsymbol{B}_t^H-\widehat{\boldsymbol{B}}_t^H\right)^2\right]dt\right\}
\end{align}
which yields a $L^2(\mathbb{P})$ optimal approximation\footnote{while this does not yield strong pathwise convergence like in \citep{harms2019strong}, it is more computationally tractable}. 

Since the fBM process is determined by the scaled fBM process, which we denote as  $\widehat{\boldsymbol{X}}_t:=\sqrt{\varepsilon}\widehat{\boldsymbol{B}}_t^H$, and the $K$ OU processes $\boldsymbol{Y}_t:=(\boldsymbol{Y}_t^1, \dots, \boldsymbol{Y}_t^K)$ that approximate $\widehat{\boldsymbol{B}}_t$, we define the full reference process $\mathbb{Q}$ as $\boldsymbol{Z}:=(\sqrt{\varepsilon}\widehat{\boldsymbol{B}}_t, \boldsymbol{Y}_t)$, which follows the SDE:
\begin{align}
    \mathbb{Q}:\quad d\boldsymbol{Z}_t=\boldsymbol{F}\boldsymbol{Z}_tdt+\sigma_td\boldsymbol{B}_t\label{eq:frac-ref-sde}
\end{align}
where $\boldsymbol{F}\in \mathbb{R}^{d( K+1)\times d(K+1)}$ is a block matrix encoding the linear coupling between the state $\boldsymbol{X}_t$ and the auxiliary OU processes $\boldsymbol{Y}_t:=(\boldsymbol{Y}_t^1, \dots, \boldsymbol{Y}_t^K)$ and $\sigma_t$ is the diffusion coefficient. 

However, we observe that the process $\widehat{\boldsymbol{X}}_t$ alone is not Markov, as there is no way to predict the next state $\widehat{\boldsymbol{X}}_{t+h}$ from $\widehat{\boldsymbol{X}}_t$ without knowledge of the OU processes, therefore only the full reference process $\boldsymbol{Z}_t:=(\widehat{\boldsymbol{X}}_t, \boldsymbol{Y}_t^1, \dots, \boldsymbol{Y}_t^K)$ is Markov. Using $\mathbb{Q}$ as the reference Markov process, we can derive the form of the Markov fractional Brownian bridge, where we leverage Doob's $h$-transfrom described in Section \ref{subsec:doob-transform} to condition $\boldsymbol{Z}_t$ on the endpoint marginals.

\begin{proposition}[Markov Approximation of Fractional Brownian Bridge \citep{nobis2025fractional}]\label{prop:markov-approx-fractional-bridge}
    Let $\mathbb{Q}$ be the reference path measure induced by the Markov augmentation $\boldsymbol{Z}_{t|0,T}$ of a scaled fractional Brownian motion, with reference dynamics:
    \begin{align}
        d\boldsymbol{Z}_{t|0,T}=\boldsymbol{F}\boldsymbol{Z}_{t|0,T}dt+\sigma_td\boldsymbol{B}_t
    \end{align}
    Fixing endpoints $\boldsymbol{X}_0=\boldsymbol{x}_0$ and $\boldsymbol{X}_T=\boldsymbol{x}_T$, then the corresponding bridge measure  obtained by conditioning $\mathbb{Q}$ on these endpoint constraints is defined by the Doob $h$-transform of $\mathbb{Q}$ with $h(\boldsymbol{z},t):=\mathbb{S}_{T|t}(\boldsymbol{x}_T|\boldsymbol{Z}_{t|0,T}=\boldsymbol{z})$, and the Markov approximation of the fractional Brownian bridge satisfies the SDE:
    \begin{align}
        d\boldsymbol{Z}_{t|0,T}=(\boldsymbol{F}\boldsymbol{Z}_{t|0,T}+\sigma_t^2\nabla_{\boldsymbol{z}} \log \mathbb{S}_{T|t}(\boldsymbol{x}_T|\boldsymbol{Z}_{t|0,T}=\boldsymbol{z}))dt+\sigma_td\boldsymbol{B}_t
    \end{align}
\end{proposition}

\textit{Proof.} To prove this, we leverage Doob's $h$-Transform described in Section \ref{subsec:doob-transform}, which defines a endpoint conditioned function $h(\boldsymbol{z},t)$. Here, we define $h:\mathbb{R}^{d(K+1)}\times[0,T]\to [0,1]$ as the endpoint probabilty of a state $\boldsymbol{X}_T=\boldsymbol{x}_T$ under the augmented path measure $\mathbb{S}$, given by:
\begin{align}
    h(\boldsymbol{z},t):=\mathbb{S}_{T|t}(\boldsymbol{X}_T=\boldsymbol{x}_T|\boldsymbol{Z}_t=\boldsymbol{z})\equiv\mathbb{S}_{T|t}(\boldsymbol{x}_T|\boldsymbol{z})
\end{align}
where $\mathbb{S}_{T|t}(\cdot|\boldsymbol{z})$ is the conditional density on $\boldsymbol{X}_T$ given $\boldsymbol{z}$. 

Now, we aim to show that $\mathbb{S}_{T|t}(\boldsymbol{x}_T|\boldsymbol{z})$ satisfies the properties of the $h$-function defined in Section \ref{subsec:doob-transform}, which allows us to directly write the SDE associated with the fractional Brownian bridge. First, we check that it satisfies the (\ref{eq:martingale-property}) that defines a valid tilting function:
\begin{align}
    h(\boldsymbol{z},t)&=\int_{\mathbb{R}^d}\mathbb{S}_{t+\Delta t|t}(\boldsymbol{Z}_{t+\Delta t}=\tilde{\boldsymbol{z}}|\boldsymbol{Z}_t=\boldsymbol{z})h(\tilde{\boldsymbol{z}},t+\Delta t)d\tilde{\boldsymbol{z}}\label{eq:frac-space-time-reg}
\end{align}
To derive an expression for $\pinktext{\mathbb{S}_{t+\Delta t|t}(\tilde{\boldsymbol{z}}|\boldsymbol{z})}$, we use Bayes rule to decompose $\mathbb{S}_{t+\Delta t|t, T}(\tilde{\boldsymbol{z}}|\boldsymbol{z}, \boldsymbol{x}_T)$ and apply the Markov property $\mathbb{S}_{T|t, t+\Delta t}(\boldsymbol{x}_T|\boldsymbol{z}, \tilde{\boldsymbol{z}})=\mathbb{S}_{T|t+\Delta t}(\boldsymbol{x}_T|\tilde{\boldsymbol{z}})$ to get:
\begin{align}
    \mathbb{S}_{t+\Delta t|t, T}(\tilde{\boldsymbol{z}}|\boldsymbol{z}, \boldsymbol{x}_T)=\frac{\bluetext{\mathbb{S}_{T|t, t+\Delta t}(\boldsymbol{x}_T|\boldsymbol{z}, \tilde{\boldsymbol{z}})}\pinktext{\mathbb{S}_{t+\Delta t|t}(\tilde{\boldsymbol{z}}|\boldsymbol{z })}}{\mathbb{S}_{T|t}(\boldsymbol{x}_T|\boldsymbol{z})}&=\frac{\bluetext{\mathbb{S}_{T|t+\Delta t}(\boldsymbol{x}_T|\tilde{\boldsymbol{z}})}\pinktext{\mathbb{S}_{t+\Delta t|t}(\tilde{\boldsymbol{z}}|\boldsymbol{z})}}{\mathbb{S}_{T|t}(\boldsymbol{x}_T|\boldsymbol{z})}\nonumber\\
    \implies\pinktext{\mathbb{S}_{t+\Delta t|t}(\tilde{\boldsymbol{z}}|\boldsymbol{z})}=\frac{\mathbb{S}_{t+\Delta t|t, T}(\tilde{\boldsymbol{z}}|\boldsymbol{z}, \boldsymbol{x}_T)\mathbb{S}_{T|t}(\boldsymbol{x}_T|\boldsymbol{z})}{\mathbb{S}_{T|t+\Delta t}(\boldsymbol{x}_T|\tilde{\boldsymbol{z}})}&=\frac{\mathbb{S}_{t+\Delta t|t, T}(\tilde{\boldsymbol{z}}|\boldsymbol{z}, \boldsymbol{x}_T)\greentext{h(\boldsymbol{z},t)}}{\greentext{h(\tilde{\boldsymbol{z}},t+\Delta t)}}\label{eq:frac-S-cond}
\end{align}
where we substitute the definition of $h(\boldsymbol{z},t)$ and $h(\tilde{\boldsymbol{z}},t+\Delta t)$ in the final equality. Now, substituting (\ref{eq:frac-S-cond}) into (\ref{eq:frac-space-time-reg}), we get:
\begin{align}
    h(\boldsymbol{z},t)&=\int_{\mathbb{R}^d}\bluetext{\mathbb{S}_{t+\Delta t|t}(\boldsymbol{Z}_{t+\Delta t}=\tilde{\boldsymbol{z}}|\boldsymbol{Z}_t=\boldsymbol{z})}h(\tilde{\boldsymbol{z}},t+\Delta t)d\tilde{\boldsymbol{z}}\nonumber\\
    &=\int_{\mathbb{R}^d}\bluetext{\frac{\mathbb{S}_{t+\Delta t|t, T}(\tilde{\boldsymbol{z}}|\boldsymbol{z}, \boldsymbol{x}_T)h(\boldsymbol{z},t)}{h(\tilde{\boldsymbol{z}},t+\Delta t)}}h(\tilde{\boldsymbol{z}},t+\Delta t)d\tilde{\boldsymbol{z}}\nonumber\\
    &=h(\boldsymbol{z},t)\underbrace{\int_{\mathbb{R}^d}\mathbb{S}_{t+\Delta t|t, T}(\tilde{\boldsymbol{z}}|\boldsymbol{z}, \boldsymbol{x}_T)d\tilde{\boldsymbol{z}}}_{=1}=\boxed{h(\boldsymbol{z},t)}
\end{align}
which means that $h(\boldsymbol{z},t):=\mathbb{S}_{T|t}(\boldsymbol{x}_T|\boldsymbol{z})$ is a valid tilting function and by Proposition \ref{prop:doob-h} and the reference SDE (\ref{eq:frac-ref-sde}), the SDE of the fractional Brownian bridge can be written as:
\begin{align}
    d\boldsymbol{Z}_{t|0,T}=(\boldsymbol{F}\boldsymbol{Z}_{t|0,T}+\bluetext{\sigma_t^2\nabla_{\boldsymbol{z}} \log \mathbb{S}_{T|t}(\boldsymbol{x}_T|\boldsymbol{Z}_{t|0,T}=\boldsymbol{z})})dt+\sigma_td\boldsymbol{B}_t\label{eq:frac-brownian-bridge}
\end{align}
which concludes the proof of the fractional Brownian bridge. \hfill $\square$.

We can derive the explicit form of $\nabla_{\boldsymbol{z}} \log \mathbb{S}_{T|t}(\boldsymbol{x}_T|\boldsymbol{Z}_{t|0,T}=\boldsymbol{z})$, which turns out to be Gaussian, with the following Lemma.

\begin{lemma}[Gaussian Form of Fractional Brownian Bridge]\label{lemma:frac-gaus-bridge}
    The fractional Brownian bridge has a Gaussian transition density $\mathbb{S}_{t+\Delta t|t}(\cdot|\boldsymbol{z})$. Conditioned on the terminal state $\widehat{\boldsymbol{X}}_T=\boldsymbol{x}_T$, the gradient of the log density takes the form:
    \begin{align}
        \nabla_{\boldsymbol{z}}\log \mathbb{S}_{T|t}(\boldsymbol{x}_T|\boldsymbol{z})=\left[\nabla_{\boldsymbol{z}}\log \mathbb{S}^1_{T|t}(\boldsymbol{x}_T|\boldsymbol{z}), \dots, \nabla_{\boldsymbol{z}}\log \mathbb{S}^d_{T|t}(\boldsymbol{x}_T|\boldsymbol{z})\right]
    \end{align}
    where for each $i\in \{1, \dots, d\}$, we have:
    \begin{align}
        \nabla_{\boldsymbol{z}}\log\mathbb{S}^i_{T|t}(\boldsymbol{x}_T|\boldsymbol{z})=[1, \omega_1\zeta_1(t,T), \dots, \omega_K\zeta_K(t, T)]^\top\frac{\boldsymbol{x}^i_T-\boldsymbol{\mu}^i_{T|t}(\boldsymbol{z})}{\sigma^2_{T|t}}
    \end{align}
    where the conditional mean $\boldsymbol{\mu}_{T|t}$ and covariance $\sigma^2_{T|t}$ are given by:
    \begin{small}
    \begin{align}
        \boldsymbol{\mu}_{T|t}(\boldsymbol{z})&=\boldsymbol{x}+\sum_{k=1}^K\omega_k\boldsymbol{y}_k\zeta_k(t,T),\quad \sigma_{T|t}^2(\boldsymbol{z})=\varepsilon\sum_{k=1}^K\sum_{\ell=1}^K\frac{\omega_k\omega_\ell}{\gamma_k+\gamma_\ell}\left(1-e^{-(1-t)(\gamma_k+\gamma_\ell)}\right)
    \end{align}
    \end{small}
    where $\boldsymbol{z}=(\boldsymbol{y}_1, \dots, \boldsymbol{y}_K)$ are the states of the OU processes.
\end{lemma}

\textit{Proof.} First, we recall the definition of the Markov-approximated fractional BM process given by the state $\boldsymbol{X}_t$ and the OU random variables $(\boldsymbol{Y}_t^1, \dots, \boldsymbol{Y}_t^K)$ defined as:
\begin{align}
    \widehat{\boldsymbol{X}}_t:=\sqrt{\varepsilon}\sum_{k=1}^K\omega\boldsymbol{Y}^k_t, \quad d\boldsymbol{Y}_t^k=-\gamma_k\boldsymbol{Y}_t^kdt+d\boldsymbol{B}_t
\end{align}
which combines to give the expression for the time evolution $d\widehat{\boldsymbol{X}}_t$ as:
\begin{small}
\begin{align}
    d\widehat{\boldsymbol{X}}_t=\sqrt{\varepsilon}\sum_{k=1}^K\omega_k \bluetext{d\boldsymbol{Y}^k_t}&=\sqrt{\varepsilon}\sum_{k=1}^K\omega_k \bluetext{(-\gamma_k\boldsymbol{Y}_t^kdt+d\boldsymbol{B}_t)}=\bluetext{-\sqrt{\varepsilon}\sum_{k=1}^K\omega_k \gamma_k\boldsymbol{Y}_t^kdt +\sqrt{\varepsilon}\sum_{k=1}^K\omega_kd\boldsymbol{B}_t}
\end{align}
\end{small}
Now, we want to derive an expression for the next state $\widehat{\boldsymbol{X}}_{t+\Delta t}$ by taking the integral $\int_{0}^{t+\Delta t}$ to get:
\begin{small}
\begin{align}
    \widehat{\boldsymbol{X}}_{t+\Delta t}&:=\int_0^{t+\Delta t }\left(\bluetext{-\sqrt{\varepsilon}\sum_{k=1}^K\omega_k \gamma_k\boldsymbol{Y}_r^kdr+\sqrt{\varepsilon}\sum_{k=1}^K\omega_kd\boldsymbol{B}_r}\right)\nonumber\\
    &=-\sqrt{\varepsilon}\sum_{k=1}^K\omega_k \gamma_k\int_0^{t+\Delta t }\boldsymbol{Y}_r^kdr +\sqrt{\varepsilon}\sum_{k=1}^K\omega_k\boldsymbol{B}_{t+\Delta t}
\end{align}
\end{small}
Splitting the integral into $\int_0^{t+\Delta t}= \int_0^t+\int_t^{t+\Delta t}$ and the Brownian increment $\boldsymbol{B}_{t+\Delta t}=\boldsymbol{B}_t+(\boldsymbol{B}_{t+\Delta t}-\boldsymbol{B}_t)$, we have:
\begin{small}
\begin{align}
    \widehat{\boldsymbol{X}}_{t+\Delta t}&=\sqrt{\varepsilon}\bluetext{\underbrace{\left(-\sum_{k=1}^K\omega_k \gamma_k\int_0^t\boldsymbol{Y}_r^kdr +\sum_{k=1}^K\omega_k\boldsymbol{B}_t\right)}_{=\widehat{\boldsymbol{B}}_t^H}}-\sqrt{\varepsilon}\sum_{k=1}^K\omega_k \gamma_k\int_t^{t+\Delta t}\pinktext{\boldsymbol{Y}_r^k}dr +\sqrt{\varepsilon}\sum_{k=1}^K\omega_k(\boldsymbol{B}_{t+\Delta t}-\boldsymbol{B}_t)\label{eq:frac-proof-1}
\end{align}
\end{small}
Expanding $\boldsymbol{Y}_r^k$ using the definition in (\ref{eq:def-Yt}), we have:
\begin{small}
\begin{align}
    \boldsymbol{Y}_r^k&=\int_0^re^{-\gamma_k(r-s)}d\boldsymbol{B}_s=\int_0^te^{-\gamma_k(r-s)}d\boldsymbol{B}_s+\int_t^re^{-\gamma_k(r-s)}d\boldsymbol{B}_s\nonumber\\
    &=\int_0^te^{-\gamma_k(r-t)}e^{-\gamma_k(t-s)}d\boldsymbol{B}_s+\int_t^re^{-\gamma_k(r-s)}d\boldsymbol{B}_s=e^{-\gamma_k(r-t)}\underbrace{\int_0^te^{-\gamma_k(t-s)}d\boldsymbol{B}_s}_{=:\boldsymbol{Y}_t^k}+\int_t^re^{-\gamma_k(r-s)}d\boldsymbol{B}_s\nonumber\\
    &=e^{-\gamma_k(r-t)}\boldsymbol{Y}_t^k+\int_t^re^{-\gamma_k(r-s)}d\boldsymbol{B}_s
\end{align}
\end{small}
Substituting this back into (\ref{eq:frac-proof-1}) and applying the Stochastic Fubini Theorem \citep{harms2019affine}, we get:
\begin{small}
\begin{align}
    &\widehat{\boldsymbol{X}}_{t+\Delta t}=\sqrt{\varepsilon}\widehat{\boldsymbol{B}}_t+\sqrt{\varepsilon}\sum_{k=1}^K\omega_k \left[-\gamma_k\int_t^{t+\Delta t}\pinktext{\left(e^{-\gamma_k(r-t)}\boldsymbol{Y}_t^k+\int_t^re^{-\gamma_k(r-s)}d\boldsymbol{B}_s\right)}dr +(\boldsymbol{B}_{t+\Delta t}-\boldsymbol{B}_t)\right]\nonumber\\
    &=\underbrace{\sqrt{\varepsilon}\widehat{\boldsymbol{B}}_t}_{=:\widehat{\boldsymbol{X}}_t}+\sqrt{\varepsilon}\sum_{k=1}^K\omega_k \bigg[-\gamma_k\bigg(\bluetext{\boldsymbol{Y}_t^k\underbrace{\int_t^{t+\Delta t}e^{-\gamma_k(r-t)}dr}_{=\int_0^{\Delta t}e^{-\gamma_ks}ds=\gamma_k^{-1}(1-e^{-\gamma_k\Delta t})}}+\pinktext{\underbrace{\int_t^{t+\Delta t}\int_t^re^{-\gamma_k(r-s)}d\boldsymbol{B}_sdr}_{\int_t^{t+\Delta t}\int_t^r(\cdot)d\boldsymbol{B}_sdr=\int _t^{t+\Delta t}\int_s^{t+\Delta t}(\cdot)drd\boldsymbol{B}_s}}\bigg) +(\boldsymbol{B}_{t+\Delta t}-\boldsymbol{B}_t)\bigg]\nonumber\\
    &=\widehat{\boldsymbol{X}}_t+\sqrt{\varepsilon}\sum_{k=1}^K\omega_k \bigg[\bluetext{-(1-e^{-\gamma_k\Delta t})\boldsymbol{Y}_t^k}-\gamma_k\pinktext{\int_t^{t+\Delta t}\int_s^{t+\Delta t}e^{-\gamma_k(r-s)}drd\boldsymbol{B}_s} +\int_t^{t+\Delta t }d\boldsymbol{B}_s\bigg]\nonumber\\
    &=\widehat{\boldsymbol{X}}_t+\sqrt{\varepsilon}\sum_{k=1}^K\omega_k \bigg[\bluetext{(e^{-\gamma_k\Delta t}-1)\boldsymbol{Y}_t^k}+\gamma_k\pinktext{\int_t^{t+\Delta t}\bigg(-\underbrace{\int_s^{t+\Delta t}e^{-\gamma_k(r-s)}dr}_{=1-e^{-\gamma_k(t+\Delta t-s)}} +1\bigg)d\boldsymbol{B}_s} \bigg]\nonumber\\
    &=\widehat{\boldsymbol{X}}_t+\sqrt{\varepsilon}\sum_{k=1}^K\omega_k \bigg[(e^{-\gamma_k\Delta t}-1)\boldsymbol{Y}_t^k+\gamma_k\pinktext{\int_t^{t+\Delta t}e^{-\gamma_k(t+\Delta t-s)}d\boldsymbol{B}_s} \bigg]\nonumber\\
    &=\widehat{\boldsymbol{X}}_t+\sum_{k=1}^K\omega_k\boldsymbol{Y}_t^k \bluetext{\underbrace{\sqrt{\varepsilon}(e^{-\gamma_k\Delta t}-1)}_{=:\zeta(t,t+\Delta t)}}+\sqrt{\varepsilon}\sum_{k=1}^K\omega_k\gamma_k\pinktext{\int_t^{t+\Delta t}e^{-\gamma_k(t+\Delta t-s)}d\boldsymbol{B}_s} \nonumber\\
    &=\widehat{\boldsymbol{X}}_t+\sum_{k=1}^K\omega_k\boldsymbol{Y}_t^k \bluetext{\zeta(t,t+\Delta t)}+\sqrt{\varepsilon}\sum_{k=1}^K\omega_k\gamma_k\pinktext{\int_t^{t+\Delta t}e^{-\gamma_k(t+\Delta t-s)}d\boldsymbol{B}_s} \label{eq:frac-proof-2}
\end{align}
\end{small}
To derive the conditional mean $\boldsymbol{\mu}_{T|t}(\boldsymbol{z})$ and covariance $\sigma_{T|t}^2$, we set $\Delta t=T-t$ in (\ref{eq:frac-proof-2}) to get:
\begin{align}
    \widehat{\boldsymbol{X}}_T=\widehat{\boldsymbol{X}}_t+\sum_{k=1}^K\omega_k\boldsymbol{Y}_t^k \bluetext{\zeta(t,T)}+\sqrt{\varepsilon}\sum_{k=1}^K\omega_k\gamma_k\pinktext{\int_t^Te^{-\gamma_k(T-s)}d\boldsymbol{B}_s}
\end{align}
Defining $\boldsymbol{Z}_t:=(\widehat{\boldsymbol{X}}_t, \boldsymbol{Y}_t^1, \dots, \boldsymbol{Y}^K_t)$, the conditional mean given the realization $\boldsymbol{z}=(\boldsymbol{x},\boldsymbol{y}_1, \dots, \boldsymbol{y}_K)$ is given by:
\begin{align}
    \boldsymbol{\mu}_{T|t}(\boldsymbol{z})&:=\mathbb{E}[\widehat{\boldsymbol{X}}_t|\boldsymbol{Z}_t=\boldsymbol{z}]=\boldsymbol{x}+\sum_{k=1}^K\omega_k\boldsymbol{y}_k\zeta(t,T)\label{eq:frac-proof-mean}
\end{align}
where the stochastic integral vanishes under the conditional expectation. To compute the conditional variance $\sigma_{T|t}^2:=\text{Var}(\widehat{\boldsymbol{X}}_T|\boldsymbol{Z}_t)$, we let $I_k=\int_t^Te^{-\gamma_k(T-s)}ds$ and expand:
\begin{small}
\begin{align}
    \sigma_{T|t}^2&:=\text{Var}(\widehat{\boldsymbol{X}}_T|\boldsymbol{Z}_t)=\text{Var}\left(\sqrt{\varepsilon}\sum_{k=1}^K\omega_k\gamma_kI_k\right)=\varepsilon\text{Var}\left(\sum_{k=1}^K\omega_k\gamma_kI_k\right)=\varepsilon\sum_{k=1}^K\sum_{\ell=1}^K\omega_k\omega_\ell\text{Cov}\left(I_k, I_\ell\right)
\end{align}
\end{small}
Using \textbf{Itô's isometry}, which states that the covariance of Itô integrals combines to:
\begin{align}
    \text{Cov}\left(I_k, I_\ell\right)&=\text{Cov}\left(\int_t^T\bluetext{e^{-\gamma_k(T-s)}}ds, \int_t^T\pinktext{e^{-\gamma_\ell(T-s)}}ds\right)=\int_t^T\bluetext{e^{-\gamma_k(T-s)}}\pinktext{e^{-\gamma_\ell(T-s)}}ds\nonumber\\
    &=\int_t^T\greentext{e^{-(\gamma_k+\gamma_\ell)(T-s)}}ds=\int_0^{T-t}\greentext{e^{-(\gamma_k+\gamma_\ell)r}}dr=\frac{1-e^{-(\gamma_k+\gamma_\ell)(T-t)}}{\gamma_k+\gamma_\ell}
\end{align}
We can derive in the final expression for the conditional covariance:
\begin{align}
    \sigma_{T|t}^2&=\varepsilon\sum_{k=1}^K\sum_{\ell=1}^K\omega_k\omega_\ell\frac{1-e^{-(\gamma_k+\gamma_\ell)(T-t)}}{\gamma_k+\gamma_\ell}=\varepsilon\sum_{k, \ell=1}^K\frac{\omega_k\omega_\ell}{\gamma_k+\gamma_{\ell}}\left(1-e^{-(\gamma_k+\gamma_\ell)(T-t)}\right)\label{eq:frac-proof-cov}
\end{align}
Therefore, we have shown that $\widehat{\boldsymbol{X}}_T|(\boldsymbol{Z}_t=\boldsymbol{z})\sim \mathbb{S}_{T|t}(\boldsymbol{x}_T|\boldsymbol{z})=\mathcal{N}(\boldsymbol{\mu}_{T|t}(\boldsymbol{z}), \sigma_{T|t}^2)$. We compute the gradient for the log density $\nabla_{\boldsymbol{z}}\log\mathbb{S}^i_{T|t}(\boldsymbol{x}_T|\boldsymbol{z})$ for the $i$th dimension of $\boldsymbol{x}_T$ as:
\begin{small}
\begin{align}
    \nabla_{\boldsymbol{z}}\log\mathbb{S}^i_{T|t}(\boldsymbol{x}_T|\boldsymbol{z})&=\nabla_{\boldsymbol{z}}\log \left(\frac{1}{\sqrt{2\pi\sigma^2_{T|t}}}\exp\left(-\frac{(\boldsymbol{x}^i_T-\boldsymbol{\mu}^i_{T|t}(\boldsymbol{z}))^2}{2\sigma^2_{T|t}}\right)\right)\nonumber\\
    &=\bluetext{\nabla_{\boldsymbol{z}}} \left(-\frac{(\boldsymbol{x}^i_T-\bluetext{\boldsymbol{\mu}^i_{T|t}(\boldsymbol{z})})^2}{2\sigma^2_{T|t}}+C\right)=\frac{\boldsymbol{x}^i_T-\boldsymbol{\mu}^i_{T|t}(\boldsymbol{z})}{\sigma_{T|t}^2}\nabla_{\boldsymbol{z}}\boldsymbol{\mu}^i_{T|t}(\boldsymbol{z})
\end{align}
\end{small}
where the last equality follows from applying the chain rule since only $\boldsymbol{\mu}_{T|t}(\boldsymbol{z})$ depends on $\boldsymbol{z}$. Given $\boldsymbol{z}:=(\boldsymbol{x},\boldsymbol{y}_1, \dots, \boldsymbol{y}_K)$ and our definition for $\boldsymbol{\mu}_{T|t}(\boldsymbol{z})$ in (\ref{eq:frac-proof-mean}), the gradient expands into:
\begin{small}
\begin{align}
    \nabla_{\boldsymbol{z}}\log\mathbb{S}^i_{T|t}(\boldsymbol{x}_T|\boldsymbol{z})&=\frac{\boldsymbol{x}^i_T-\boldsymbol{\mu}_{T|t}(\boldsymbol{z})}{\sigma_{T|t}^2}\nabla_{\boldsymbol{z}}\boldsymbol{\mu}^i_{T|t}(\boldsymbol{z})=\frac{\boldsymbol{x}^i_T-\boldsymbol{\mu}^i_{T|t}(\boldsymbol{z})}{\sigma_{T|t}^2}\bluetext{\left[\frac{\partial \boldsymbol{\mu}^i_{T|t}}{\partial \boldsymbol{x}^i}, \frac{\partial \boldsymbol{\mu}^i_{T|t}}{\partial \boldsymbol{y}^i_1}, \dots, \frac{\partial \boldsymbol{\mu}^i_{T|t}}{\partial \boldsymbol{y}^i_K}\right]^\top}\nonumber\\
    &=\boxed{\bluetext{\left[1,\omega_1\zeta_1(t,T),\dots, \omega_K\zeta_K(t,T) \right]^\top}\frac{\boldsymbol{x}^i_T-\boldsymbol{\mu}^i_{T|t}(\boldsymbol{z})}{\sigma_{T|t}^2}}
\end{align}
\end{small}
which is exactly the form for the log gradient of the $h$-function defined in the Lemma.\hfill $\square$

Now, we can substitute this into the SDE of the fractional Brownian bridge from (\ref{eq:frac-brownian-bridge}) to get its complete form:
\begin{align}
    &d\boldsymbol{Z}_{t|0,T}=(\boldsymbol{F}\boldsymbol{Z}_{t|0,T}+\sigma_t^2\bluetext{\boldsymbol{u}(\boldsymbol{Z}_{t|0,T},t)})dt+\sigma_td\boldsymbol{B}_t\nonumber\\
    &\text{s.t.}\quad \begin{cases}
        \boldsymbol{u}(\boldsymbol{Z}_{t|0,T},t)=[u_1(\boldsymbol{Z}_{t|0,T},t), \dots, u_d(\boldsymbol{Z}_{t|0,T},t)]\\
        u_i(\boldsymbol{Z}_{t|0,T},t)=\nabla_{\boldsymbol{z}}\log\mathbb{S}^i_{T|t}(\boldsymbol{x}_T|\boldsymbol{z})=\left[1,\omega_1\zeta_1(t,T),\dots, \omega_K\zeta_K(t,T) \right]^\top\frac{\boldsymbol{x}^i_T-\boldsymbol{\mu}^i_{T|t}(\boldsymbol{z})}{\sigma_{T|t}^2}
    \end{cases} 
\end{align}

By leveraging the finite-dimensional Markov lift of the MA-fBM approximation, we have derived the explicit conditional law of $\widehat{\boldsymbol{X}}_T$ given $\boldsymbol{Z}_t$, which remains Gaussian with affine mean and time-dependent variance and whose $h$-function can be derived in closed form. The non-Markovian nature of the marginal process $\boldsymbol{X}_t$ provides a principled way to incorporate long-range temporal correlations into generative stochastic dynamics through a finite-dimensional Markov approximation of MA-fBM. This framework therefore, extends classical Schrödinger bridge problems beyond memoryless dynamics and offers a mathematically tractable route to learning physically realistic long-horizon dependencies.

\subsection{Closing Remarks for Section \ref{sec:variations-of-sb}}

In this section, we expanded the theory of the dynamic Schrödinger bridge problem to a broader array of constraints and problem settings. We start with the \textbf{Gassian SB problem}, which we show admits a closed-form solution that is in the class of Gaussian Markov processes. We then introduced the \textbf{generalized SB problem}, which extends the classical formulation to systems with mean-field interactions where the dynamics of individual particles depend on the evolving population distribution.

Next, we explored several important extensions of the SB framework that arise in more complex settings. These include the \textbf{multi-marginal SB problem}, which incorporates multiple intermediate marginal constraints; the \textbf{unbalanced SB problem}, which allows for the creation or destruction of mass along the transport trajectory; and the \textbf{branched SB problem}, which captures scenarios where stochastic trajectories diverge toward multiple terminal modes. Finally, we considered an alternative class of stochastic processes driven by fractional Brownian motion, leading to the formulation of the \textbf{fractional SB problem}.

These extensions illustrate the incredible flexibility of the Schrödinger bridge framework in describing complex stochastic systems across a wide range of settings. Having introduced the theoretical foundations of both the static and dynamic formulations of the SB problem, in addition to their extensions to diverse constraints and dynamics, we are now prepared to dive into modern \textbf{generative modeling frameworks} that leverage Schrödinger bridge theory to construct scalable algorithms for high-dimensional data.

\newpage 
\section{Generative Modeling with Schrödinger Bridges}
\label{sec:generative-modeling}

In this section, we develop the connection between Schrödinger bridge theory and modern generative modeling frameworks, showing how generative modeling can be formulated as the problem of learning controlled stochastic dynamics that interpolate between an initial and a target distribution while minimizing relative entropy with respect to a reference process.

We begin with a brief primer on score-based generative modeling (Section \ref{subsec:primer-sgm}), highlighting its formulation in terms of forward and reverse-time stochastic processes. We then extend this perspective by jointly learning forward and backward controlled drifts through likelihood maximization over coupled forward–backward SDEs (Section \ref{subsec:likelihood-training}). As an alternative paradigm, we introduce diffusion Schrödinger bridge matching, which constructs generative models via path-space reciprocal and Markov projections with parameterized drifts (Section \ref{subsec:diffusionsbm}). Finally, we present two simulation-free approaches for learning Schrödinger bridges: score and flow matching (Section \ref{subsec:score-and-flow}) and adjoint matching (Section \ref{subsec:adjoint-matching}).

Throughout this section, we show how Schrödinger bridges provide a unifying framework that connects likelihood-based training, path-space KL minimization, and score and flow matching frameworks into a single coherent theory.

\subsection{A Primer on Score-Based Generative Modeling}
\label{subsec:primer-sgm}
From Section \ref{subsec:time-reversal}, we have shown that Schrödinger bridges are a generalization of diffusion models where the prior distribution is a simple Gaussian prior. Recall the backward SDE (\ref{eq:time-rev-forward-back-sde}) corresponding to the forward SDE with variance-exploding drift given by:
\begin{align}
    d\boldsymbol{X}_t&=\boldsymbol{f}(\boldsymbol{X}_t, t)dt+\sigma_td\boldsymbol{B}_t, \quad &\boldsymbol{X}_0\sim \pi_0:=p_{\text{data}}\\
    d\tilde{\boldsymbol{X}}_s&=\left[-\boldsymbol{f}(\tilde{\boldsymbol{X}}_s,T-s)+\sigma_{T-s}^2\nabla\log \tilde{p}_s(\tilde{\boldsymbol{X}}_s)\right]ds+\sigma_{T-s}d\widetilde{\boldsymbol{B}}_s, \quad &\tilde{\boldsymbol{X}}_0\sim \pi _T:=p_{\text{prior}}
\end{align}
where $\tilde{p}_s(\boldsymbol{x})$ is the density generated by the diffusion SDE at time $s$ and $\nabla\log \tilde{p}_s(\boldsymbol{x})$ is a gradient drift that pushes the density towards areas of high likelihood given the noisy data distribution at time $s$. This process guides the diffusion process to samples from the true data distribution $\tilde{p}_T(\boldsymbol{x})=p_0(\boldsymbol{x})$. The expression $\nabla \log \tilde{p}_s(\boldsymbol{x})$ is known as the \textbf{score function}.

\boldtext{Score-based generative modeling} aims to train a generative model that samples from the data distribution $\pi_0:=p_{\text{data}}$ by parameterizing the score function with a neural network with parameters $\theta$ known as the \textbf{score-based model} $\boldsymbol{s}_\theta(\boldsymbol{x},s)\approx \nabla \log \tilde{p}_s(\boldsymbol{x})$ which estimates the score function over the state space $\boldsymbol{x}\in \mathbb{R}^d$ and time coordinate $s\in [0,T]$. 

Given the score-based model $\boldsymbol{s}_\theta(\boldsymbol{x},s)$, we can define the estimated backward SDE as:
\begin{align}
    d\tilde{\boldsymbol{X}}_s=\left[-\boldsymbol{f}(\tilde{\boldsymbol{X}}_s,T-s)+\sigma_{T-s}^2\boldsymbol{s}_\theta(\tilde{\boldsymbol{X}}_s, s)\right]ds+\sigma_{T-s}d\widetilde{\boldsymbol{B}}_s, \quad \tilde{\boldsymbol{X}}_0\sim p_{\text{prior}}\label{eq:score-sde}
\end{align}

The distribution of \textit{clean} samples generated from simulating many samples $\tilde{\boldsymbol{X}}_0\sim p_{\text{prior}}$ via the SDE (\ref{eq:score-sde}) over time $s\in [0,T]$ approximates the data distribution $p_{\text{data}}$.

\begin{figure}
    \centering
    \includegraphics[width=\linewidth]{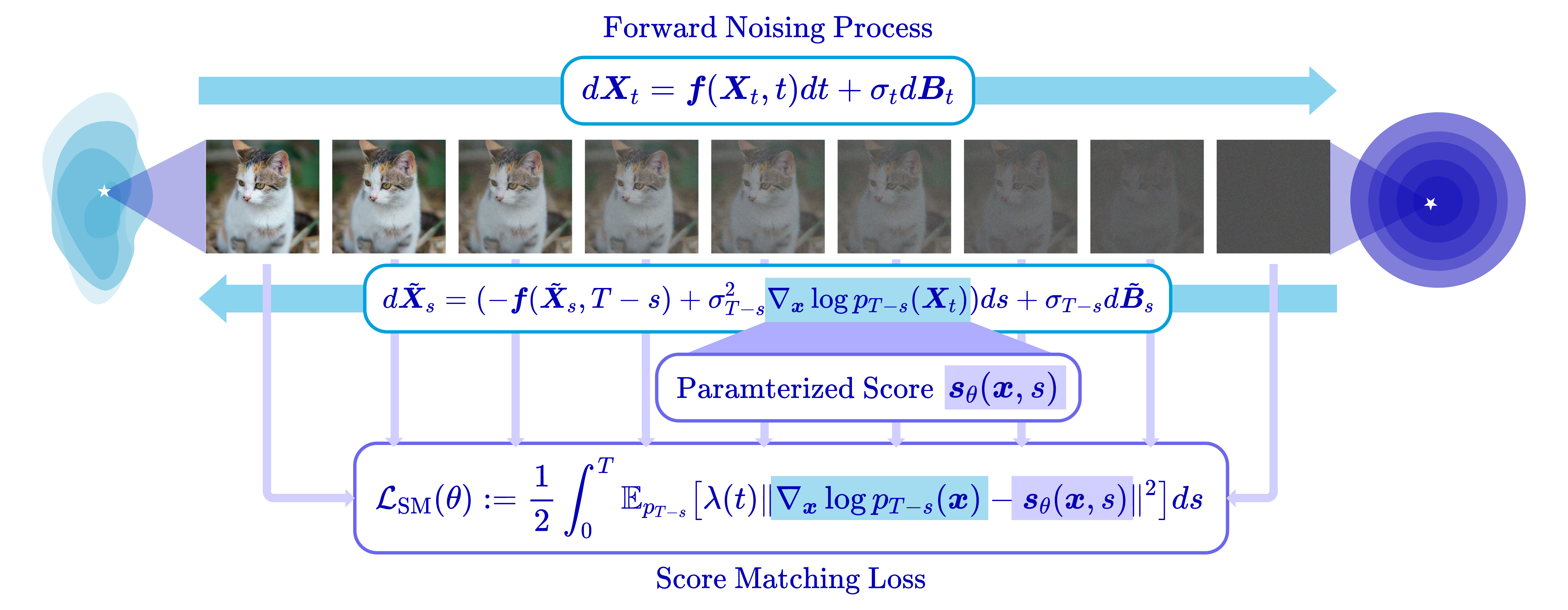}
    \caption{\textbf{Score-Based Generative Models.} Score-based generative models learn the score function $\nabla \log p_t(\boldsymbol{x})$ of a diffusion process that progressively perturbs data with noise through a forward SDE. A neural network $\boldsymbol{s}_\theta(\boldsymbol{x},t)$ is trained using the score matching loss to approximate this time-dependent score. Once learned, the reverse-time SDE uses the estimated score to iteratively denoise samples, transforming noise into data.}
    \label{fig:score-matching}
\end{figure}

A simple objective that is minimized exactly when the score-based model matches the true score function is the \boldtext{score matching loss} $\mathcal{L}_{\text{SM}}$ \citep{hyvarinen2005estimation, song2021maximum} defined as the weighted squared difference integrated over time:
\begin{align}
    \mathcal{L}_{\text{SM}}(\theta):=\frac{1}{2}\int_0^T\mathbb{E}_{\tilde{p}_s}\big[\lambda(t)\|\nabla \log \tilde{p}_s(\boldsymbol{x})-\boldsymbol{s}_\theta(\boldsymbol{x},s)\|^2\big]ds\tag{Score Maching Loss}\label{eq:score-matching-loss}
\end{align}
For this objective to be tractable, we require a closed-form expression for $\nabla \log \tilde{p}_s(\boldsymbol{x})$. For generative modeling of a clean data distribution, the forward stochastic process or \textit{noise injection process} can be defined with the conditional distribution $p_t(\tilde{\boldsymbol{x}}_t|\boldsymbol{x}_0):=\mathcal{N}(\tilde{\boldsymbol{x}}_t;\boldsymbol{x}_0, \sigma^2_t\boldsymbol{I}_d)$ where $\boldsymbol{x}_0\sim p_{\text{data}}$, which progressively smoothes the data distribution with larger variance $\sigma_t>\sigma_s$ for $t> s$ \citep{song2019generative}. This yields the tractable score function:
\begin{small}
\begin{align}
    \nabla\log q_t(\tilde{\boldsymbol{x}}_t|\boldsymbol{x}_0)=\nabla\log \bigg[\frac{1}{\sqrt{2\pi\sigma_t^2}}\exp\left(-\frac{\|\tilde{\boldsymbol{x}}_t-\boldsymbol{x}_0\|^2}{2\sigma_t^2}\right)\bigg]=\nabla\bigg(-\frac{\|\tilde{\boldsymbol{x}}_t-\boldsymbol{x}_0\|^2}{2\sigma_t^2}+C\bigg)=\boxed{\frac{\boldsymbol{x}_0-\tilde{\boldsymbol{x}}_t}{\sigma_t^2}}\label{eq:gaussian-score-func}
\end{align}
\end{small}
which can be optimized with the score matching objective:
\begin{align}
    \mathcal{L}_{\text{SM}}(\theta):=\frac{1}{2}\int_0^T\mathbb{E}_{p_t(\tilde{\boldsymbol{x}}_t|\boldsymbol{x}_0), p_{\text{data}}(\boldsymbol{x}_0)}\left[\left\|\frac{\boldsymbol{x}_0-\tilde{\boldsymbol{x}}_t}{\sigma_t^2}-\boldsymbol{s}_\theta(\boldsymbol{x},t)\right\|^2\right]dt\label{eq:linear-sm-loss}
\end{align}

While score matching provides a simulation-free way to learn the reverse-time drift through estimation of the score function, it does so under the assumption that the generative process has a simple prior distribution, typically a standard Gaussian like in (\ref{eq:gaussian-score-func}), and the forward diffusion dynamics are chosen to be linear like in (\ref{eq:linear-sm-loss}) such that it is analytically tractable without simulation. This design constraint restricts the class of admissible dynamics to perturbations of a simple reference process which transforms between noise and data. To extend this idea to model transport between \textit{structured distributions} with an unknown forward control drift, we introduce a likelihood-based training framework for the forward-backward SDEs that characterize the Schrödinger bridge.

\subsection{Likelihood Training of Forward-Backward SDEs}
\label{subsec:likelihood-training}
\textit{Prerequisite: Section \ref{subsec:forward-backward-sde}}

Just like how likelihood training provides a theoretically-grounded training objective of estimating the score function in score-based generative modeling, we can derive a lower bound for the log-likelihood of the \boldtext{forward-backward stochastic differential equations} (SDEs) defined in Section \ref{subsec:forward-backward-sde}, which yields a tractable training objective that aims to maximize the lower bound \citep{chen2021likelihood}.

\begin{figure}
    \centering
    \includegraphics[width=\linewidth]{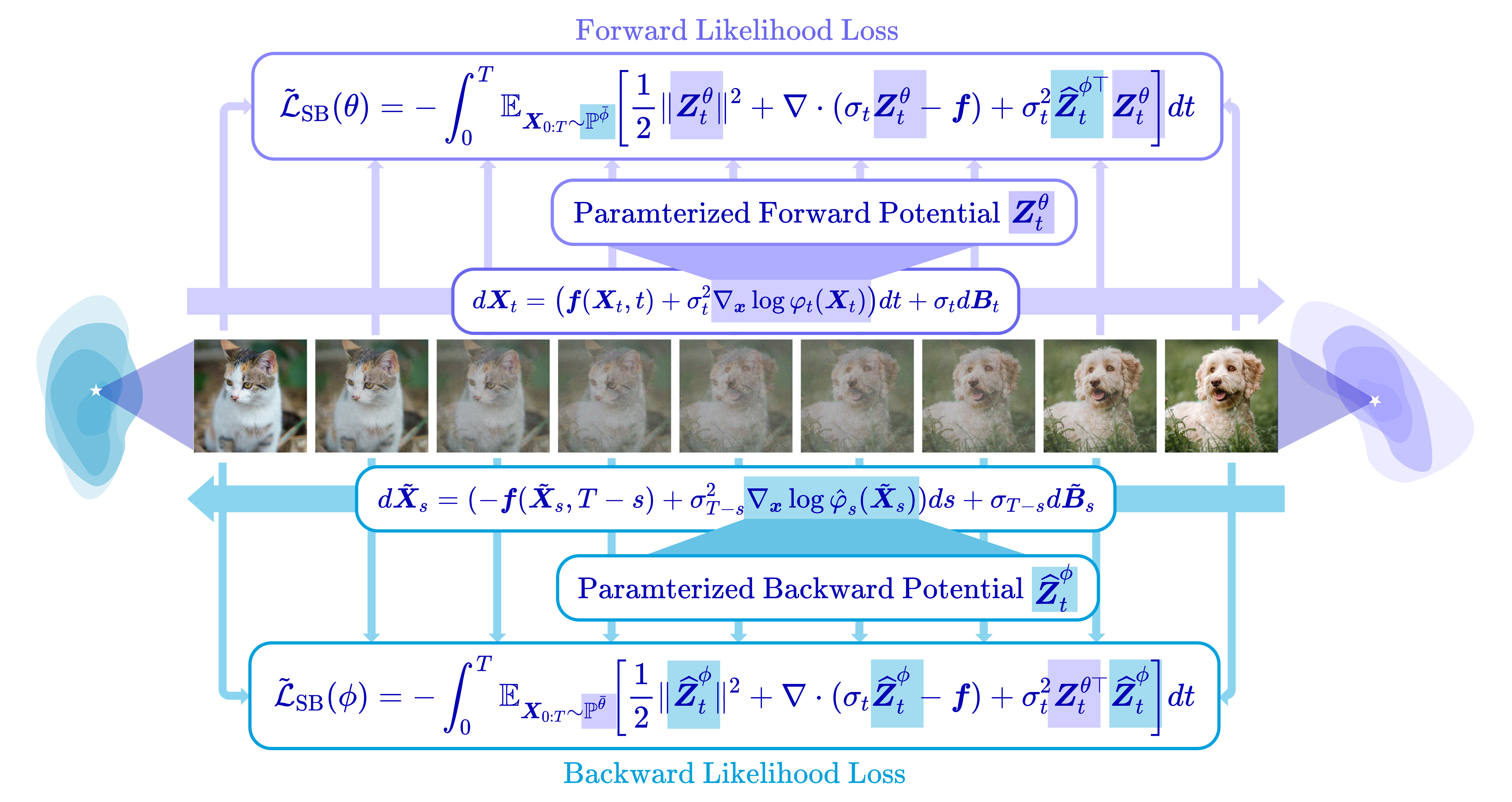}
    \caption{\textbf{Likelihood Training of Forward and Backward Schrödinger Potentials.} The Schrödinger bridge can be learned by parameterizing the log gradient of the forward and backward potentials $\boldsymbol{Z}^\theta_t=\nabla \log \varphi_t(\boldsymbol{X}_t)$ and $\widehat{\boldsymbol{Z}}_t^\phi=\nabla \log \hat\varphi_t(\boldsymbol{X}_t)$, which define the forward and backward SDEs between endpoint distributions. $\theta$ and $\phi$ are trained by maximizing path likelihoods through forward and backward objective functions, each evaluated under trajectories generated by the corresponding controlled SDE. Together, the two likelihood losses couple the forward and reverse dynamics, enabling the model to learn a consistent Schrödinger bridge between the marginals.}
    \label{fig:likelihood-training}
\end{figure}

Since the definition of $\boldsymbol{Y}_t=\log \varphi_t(\boldsymbol{X}_t)$ and $\widehat{\boldsymbol{Y}}_t=\log \hat{\varphi}_t(\boldsymbol{X}_t)$ from Section \ref{subsec:forward-backward-sde} jointly determine the log density of the SB solution $\boldsymbol{Y}_t+\widehat{\boldsymbol{Y}}_t=\log p^\star_t(\boldsymbol{X}_t)$, we can write the likelihood under the FBSDEs as the estimated value of $\boldsymbol{Y}_0+\widehat{\boldsymbol{Y}}_0=\log p_0^\star(\boldsymbol{X}_0)$ given a data point $\boldsymbol{X}_0=\boldsymbol{x}_0$.
\begin{align}
    \log p^\star_0(\boldsymbol{x}_0)&=\mathbb{E}\left[\log p_0^\star(\boldsymbol{X}_0)|\boldsymbol{X}_0=\boldsymbol{x}_0\right]=\mathbb{E}\left[\boldsymbol{Y}_0+\widehat{\boldsymbol{Y}}_0\;\big|\;\boldsymbol{X}_0=\boldsymbol{x}_0\right]
\end{align}
First, we recall the set of three coupled FBSDEs that define the solution to the non-linear SB problem as: 
\begin{align}
\begin{cases}
    d\boldsymbol{X}_t=\left(\boldsymbol{f}(\boldsymbol{X}_t,t)+ \sigma_t \boldsymbol{Z}_t\right)dt+\sigma_t d\boldsymbol{B}_t\\
    d\boldsymbol{Y}_t=\frac{1}{2}\|\boldsymbol{Z}_t\|^2dt+\boldsymbol{Z}_t^\top d \boldsymbol{B}_t\\
    d\widehat{\boldsymbol{Y}}_t=\left(\nabla\cdot (\sigma_t\widehat{\boldsymbol{Z}}_t-\boldsymbol{f})+\frac{1}{2}\|\widehat{\boldsymbol{Z}}_t\|^2+ \sigma_t^2\boldsymbol{Z}_t^\top \widehat{\boldsymbol{Z}}_t\right)dt+\widehat{\boldsymbol{Z}}_t^\top d \boldsymbol{B}_t
\end{cases}
\end{align}
Since the $\boldsymbol{Y}_t$ and $\widehat{\boldsymbol{Y}}_t$ evolve via \textbf{backward SDEs} with boundary constraints $\boldsymbol{Y}_T+\widehat{\boldsymbol{Y}}_T=\log \pi_T(\boldsymbol{X}_T)$, we can write $\boldsymbol{Y}_0$ and $\widehat{\boldsymbol{Y}}_0$ as the terminal condition integrated backward in time via the SDEs defined in (\ref{eq:nonlinear-fbsde-thm-2}) such that:
\begin{small}
\begin{align}
    \boldsymbol{Y}_T-\boldsymbol{Y}_0&=\int_0^T \frac{1}{2}\|\boldsymbol{Z}_t\|^2dt+\int_0^T\boldsymbol{Z}_t^\top d \boldsymbol{B}_t\label{eq:likelihood-derivation-1}\\
    \widehat{\boldsymbol{Y}}_T-\widehat{\boldsymbol{Y}}_0&=\int_0^T\left[\nabla\cdot (\sigma_t\widehat{\boldsymbol{Z}}_t-\boldsymbol{f})+\frac{1}{2}\|\widehat{\boldsymbol{Z}}_t\|^2+ \sigma_t^2\boldsymbol{Z}_t^\top \widehat{\boldsymbol{Z}}_t\right]dt+\int_0^T\widehat{\boldsymbol{Z}}_t^\top d \boldsymbol{B}_t
\end{align}
\end{small}
which we rearrange to get:
\begin{small}
\begin{align}
    \boldsymbol{Y}_0&=\boldsymbol{Y}_T-\int_0^T \frac{1}{2}\|\boldsymbol{Z}_t\|^2dt-\int_0^T\boldsymbol{Z}_t^\top d \boldsymbol{B}_t\\
    \widehat{\boldsymbol{Y}}_0&=\widehat{\boldsymbol{Y}}_T-\int_0^T\left[\nabla\cdot (\sigma_t\widehat{\boldsymbol{Z}}_t-\boldsymbol{f})+\frac{1}{2}\|\widehat{\boldsymbol{Z}}_t\|^2+ \sigma_t^2\boldsymbol{Z}_t^\top \widehat{\boldsymbol{Z}}_t\right]dt-\int_0^T\widehat{\boldsymbol{Z}}_t^\top d \boldsymbol{B}_t
\end{align}
\end{small}
Substituting this into the log-likelihood, we get:
\begin{small}
\begin{align}
    \log p^\star_t(\boldsymbol{x}_0)=\mathbb{E}&\left[\boldsymbol{Y}_0+\widehat{\boldsymbol{Y}}_0\;\big|\;\boldsymbol{X}_0=\boldsymbol{x}_0\right]\nonumber\\
    =\mathbb{E}&\bigg[\boldsymbol{Y}_T-\int_0^T \frac{1}{2}\|\boldsymbol{Z}_t\|^2dt-\int_0^T\boldsymbol{Z}_t^\top d \boldsymbol{B}_t\nonumber\\
    &+\widehat{\boldsymbol{Y}}_T-\int_0^T\left[\nabla\cdot (\sigma_t\widehat{\boldsymbol{Z}}_t-\boldsymbol{f})+\frac{1}{2}\|\widehat{\boldsymbol{Z}}_t\|^2+ \sigma_t^2\boldsymbol{Z}_t^\top \widehat{\boldsymbol{Z}}_t\right]dt-\int_0^T\widehat{\boldsymbol{Z}}_t^\top d \boldsymbol{B}_t\;\bigg|\;\boldsymbol{X}_0=\boldsymbol{x}_0\bigg]
\end{align}
\end{small}
Combining similar terms, we get:
\begin{small}
\begin{align}
    \log p^\star(\boldsymbol{x}_0,0)=\mathbb{E}&\left[\boldsymbol{Y}_T+\widehat{\boldsymbol{Y}}_T\big|\boldsymbol{X}_0=\boldsymbol{x}_0\right]\nonumber\\
    &-\int_0^T\mathbb{E}\left[\frac{1}{2}\|\boldsymbol{Z}_t\|^2+\nabla\cdot (\sigma_t\widehat{\boldsymbol{Z}}_t-\boldsymbol{f})+\frac{1}{2}\|\widehat{\boldsymbol{Z}}_t\|^2+ \sigma_t^2\boldsymbol{Z}_t^\top \widehat{\boldsymbol{Z}}_t\;\bigg|\;\boldsymbol{X}_0=\boldsymbol{x}_0\right]dt\nonumber\\
    &-\underbrace{\mathbb{E}\left[\int_0^T\boldsymbol{Z}_t^\top d \boldsymbol{B}_t+\int_0^T\widehat{\boldsymbol{Z}}_t^\top d \boldsymbol{B}_t\;\bigg|\;\boldsymbol{X}_0=\boldsymbol{x}_0\right]}_{=0\text{ (Itô integral has zero-expectation)}}
\end{align}
\end{small}
Since the Itô integral has zero-expectation and $\boldsymbol{Y}_T+\widehat{\boldsymbol{Y}}_T=\log p_T(\boldsymbol{X}_T)$, we are left with:
\begin{align}
    \log p^\star_0(\boldsymbol{x}_0)=\mathbb{E}&\underbrace{\left[\log p^\star_(\boldsymbol{X}_T)\right]}_{\text{(i)}}-\int_0^T\mathbb{E}\bigg[\underbrace{\frac{1}{2}\|\boldsymbol{Z}_t\|^2}_{\text{(ii)}}+\underbrace{\frac{1}{2}\|\widehat{\boldsymbol{Z}}_t\|^2}_{\text{(iii)}}+\underbrace{\nabla\cdot (\sigma_t\widehat{\boldsymbol{Z}}_t-\boldsymbol{f})}_{\text{(iv)}}+ \underbrace{\sigma_t^2\boldsymbol{Z}_t^\top \widehat{\boldsymbol{Z}}_t}_{\text{(v)}}\bigg]dt\label{eq:fbsde-lower-bound-true}
\end{align}
which is the log-likelihood of the data point that we aim to maximize. Each term can be interpreted as follows:
\begin{enumerate}
    \item The terminal distribution matching reward that is maximized with the distribution at time $T$ generated by the forward bridge matches $p_T$.
    \item The energy cost of steering samples via the forward potential drift $\boldsymbol{Z}_t=\sigma_t\nabla\log \varphi_t(\boldsymbol{X}_t)$, which is minimized when the log-likelihood is maximized.
    \item The energy cost of steering samples via the backward potential drift $\widehat{\boldsymbol{Z}}_t=\sigma_t\nabla \log \hat{\varphi}_t(\boldsymbol{X}_t)$, which is minimized when the log-likelihood is maximized.
    \item This divergence term ensures that the drift is consistent with the time-evolving density $p^\star_t(\boldsymbol{X}_t)$. We can observe this by expanding the divergence term $\nabla\cdot (\sigma_t\widehat{\boldsymbol{Z}}_t)$ using the identity $\nabla\cdot (u\boldsymbol{v})=(\nabla u)\cdot \boldsymbol{v}+u \nabla\cdot \boldsymbol{v}$ rearranged to $u \nabla\cdot \boldsymbol{v}=\nabla\cdot (u\boldsymbol{v})-(\nabla u)\cdot \boldsymbol{v}$:
    \begin{small}
    \begin{align}
        \mathbb{E}_{p^\star_t}\left[\nabla\cdot (\sigma_t\widehat{\boldsymbol{Z}}_t)\right]&=\int_{\mathbb{R}^d}\big(\sigma_t\nabla\cdot \widehat{\boldsymbol{Z}}_t\big)p^\star_t d\boldsymbol{X}_t=\underbrace{\int_{\mathbb{R}^d}\nabla \cdot (\sigma_t p^\star_t\widehat{\boldsymbol{Z}}_t)d\boldsymbol{X}_t}_{=0}-\int_{\mathbb{R}^d}\widehat{\boldsymbol{Z}}_t\cdot \nabla p^\star_td\boldsymbol{X}_t\nonumber\\
        &=-\int_{\mathbb{R}^d}\widehat{\boldsymbol{Z}}_t\cdot \nabla p^\star_td\boldsymbol{X}_t
    \end{align}
    \end{small}
    \item The consistency of the forward-backward potentials. Since we want the drifts to generate exactly symmetric opposite bridges and the dot product is minimized at $-1$ when the vectors are exactly opposite, this term is minimized when the log-likelihood is maximized.
\end{enumerate}

Since $\boldsymbol{Z}_t$ and $\widehat{\boldsymbol{Z}}_t$ determine the optimal drift in the forward and backward directions based on the SB potential, we can train a generative SB model by parameterizing them with neural networks $\boldsymbol{Z}^\theta_t\approx \boldsymbol{Z}_t$ and $\widehat{\boldsymbol{Z}}^\phi_t\approx\widehat{\boldsymbol{Z}}_t$ with parameters $\theta$ and $\phi$. Defining the SB loss with (\ref{eq:fbsde-lower-bound-true}), we get a theoretically-grounded \textbf{maximization objective} that lower bounds the true log-likelihood $\log p_0^\star(\boldsymbol{x}_0)\geq \mathcal{L}_{\text{SB}}$ defined as:
\begin{align}
    \mathcal{L}_{\text{SB}}(\theta, \phi)=\mathbb{E}&\left[\log p_T^\star(\boldsymbol{X}_T)\right]-\int_0^T\mathbb{E}\bigg[\frac{1}{2}\|\boldsymbol{Z}^\theta_t\|^2+\frac{1}{2}\|\widehat{\boldsymbol{Z}}^\phi_t\|^2+\nabla\cdot (\sigma_t\widehat{\boldsymbol{Z}}^\phi_t-\boldsymbol{f})+ \sigma_t^2\boldsymbol{Z}^{\theta\top}_t \widehat{\boldsymbol{Z}}^\phi_t\bigg]dt\label{eq:fbsde-lower-bound-pred}
\end{align}

Jointly training both $\boldsymbol{Z}^\theta_t$ and $\widehat{\boldsymbol{Z}}^\phi_t$ with this objective is carried out by \textbf{(i)} simulating the forward trajectory of the SDE.$\boldsymbol{X}_{0:T}$ using $\boldsymbol{Z}_t^\theta$ and $\widehat{\boldsymbol{Z}}_t^\phi$, \textbf{(ii)} computing the maximum likelihood objective $\mathcal{L}_{\text{SB}}$ with (\ref{eq:fbsde-lower-bound-pred}), and \textbf{(iii)} backpropagating through the SDE solver for every time step with respect to both $\theta$ and $\phi$, which requires maintaining the full computational graph of the SDE. 

While this training scheme works for low-dimensional data, it becomes computationally infeasible for high-dimensional data like images. To overcome this, we can store SDE trajectories in a replay buffer as sequences of static states $\boldsymbol{X}_t$ while discarding the computational graph containing the gradient path. Crucially, this \textbf{breaks the dependency of the trajectories with the current model parameters}, but it enables us to reuse the same SDEs over multiple gradient updates. Updating one model, like $\boldsymbol{Z}_t^\theta$, with respect to $\nabla_\theta\mathcal{L}_{\text{SB}}$ would change the SDE $\boldsymbol{X}_t$, so updating $\widehat{\boldsymbol{Z}}_t^\phi$ simultaneously with the original SDE would break the symmetry of the optimization problem. 

Instead, we can leverage the symmetric property of Schrödinger bridges:
\begin{enumerate}
    \item Given the true \textit{forward} potential drift $\boldsymbol{Z}_t=\sigma_t\nabla \log\varphi_t$, the optimal \textit{backward} potential drift $\widehat{\boldsymbol{Z}}_t^\phi$ can be learned to match the forward trajectories. 
    \item Given the true \textit{backward} potential drift $\widehat{\boldsymbol{Z}}_t=\sigma_t\nabla \log\hat{\varphi}_t$, the optimal \textit{forward} potential drift $\boldsymbol{Z}_t^\theta$ can be learned to match the backward trajectories.
\end{enumerate}

To train the backward potential drift given trajectories generated with the frozen forward model $\boldsymbol{Z}^{\bar{\theta}}_t$, where $\bar{\theta}:=\texttt{stopgrad}(\theta)$, we can use the same likelihood maximization objective defined in (\ref{eq:fbsde-lower-bound-pred}) but dropping all terms that are not dependent on $\phi$:
\begin{align}
    &\tilde{\mathcal{L}}_{\text{SB}}(\phi)=-\int_0^T\mathbb{E}_{\boldsymbol{X}_{0:T}\sim \mathbb{P}^{\bar{\theta}}}\bigg[\frac{1}{2}\|\widehat{\boldsymbol{Z}}^\phi_t\|^2+\nabla\cdot (\sigma_t\widehat{\boldsymbol{Z}}^\phi_t-\boldsymbol{f})+ \sigma_t^2\boldsymbol{Z}^{\theta\top}_t \widehat{\boldsymbol{Z}}^\phi_t\bigg]dt\label{eq:fbsde-lower-bound-fwd}\\
    &\text{s.t.}\quad \mathbb{P}^{\bar{\theta}}: d\boldsymbol{X}_t=(\boldsymbol{f}+\sigma_t^2\nabla \log \varphi_t(\boldsymbol{X}_t))dt+\sigma_t d\boldsymbol{B}_t, \quad \boldsymbol{X}_0\sim \pi_0
\end{align}

Given that the SB is symmetric in either direction, the objective for training the forward potential drift given trajectories generated with the frozen backward model $\widehat{\boldsymbol{Z}}^{\bar{\phi}}_t$, where $\bar{\phi}:=\texttt{stopgrad}(\phi)$, is the same as (\ref{eq:fbsde-lower-bound-fwd}) but flipping $\boldsymbol{Z}_t^\theta$ and $\widehat{\boldsymbol{Z}}^\phi_t$ given by:
\begin{align}
    &\tilde{\mathcal{L}}_{\text{SB}}(\theta)=-\int_0^T\mathbb{E}_{\boldsymbol{X}_{0:T}\sim \mathbb{P}^{\bar{\phi}}}\bigg[\frac{1}{2}\|\boldsymbol{Z}^\theta_t\|^2+\nabla\cdot (\sigma_t\boldsymbol{Z}^\theta_t-\boldsymbol{f})+ \sigma_t^2\widehat{\boldsymbol{Z}}^{\phi\top}_t \boldsymbol{Z}^\theta_t\bigg]dt\label{eq:fbsde-lower-bound-bwd}\\
    &\text{s.t.}\quad \mathbb{P}^{\bar{\phi}}: d\boldsymbol{X}_t=(\boldsymbol{f}-\sigma_t^2\nabla \log \hat{\varphi}_t(\boldsymbol{X}_t))dt+\sigma_t d\boldsymbol{B}_t, \quad \boldsymbol{X}_T\sim \pi_T
\end{align}

\textit{Derivation sketch.} This symmetric objective can be derived by defining a reversed time coordinate $s:=T-t$ and defining the prior non-linear process as the reverse-time analog of the forward prior dynamics with SDE $d\boldsymbol{X}_s=-\boldsymbol{f}(\boldsymbol{X}_s,s)ds+\sigma_t d\boldsymbol{B}_s$. Then, redefining the Hopf-Cole linear PDE constraints and forward-backward SDEs with negative drift $-\boldsymbol{f}$ and following a similar derivation from (\ref{eq:likelihood-derivation-1}) to (\ref{eq:fbsde-lower-bound-true}) for the log-likelihood $\log p^\star_0(\boldsymbol{x}_T)=\mathbb{E}[\boldsymbol{Y}_0+\widehat{\boldsymbol{Y}}_0|\boldsymbol{X}_T=\boldsymbol{x}_T]$ of a sample $\boldsymbol{x}_T$ from the \textit{prior} distribution or $\pi_T$. \hfill $\square$

This section shows that the optimal forward and backward control drifts defined by the Schrödinger potentials can be learned through mazximizing a lower bound on the log-likelihood of reconstructing both of the marginal constraints. These objectives lead to a symmetric alternating optimization procedure that mirrors the structure of Sinkhorn's algorithm from Section \ref{subsec:sinkhorn-algorithm}. Next, we move on to an alternative perspective, considering the optimization process as performing iterative Markovian and reciprocal projections rather than maximizing likelihoods.

\subsection{Diffusion Schrödinger Bridge Matching}
\label{subsec:diffusionsbm}
\textit{Prerequisite: Section \ref{subsec:markov-reciprocal-proj}}

Building on the Iterative Markovian Fitting (IMF) procedure from Section \ref{subsec:markov-reciprocal-proj}, we now describe the \boldtext{Diffusion Schrödinger Bridge Matching} (DSBM) algorithm \citep{shi2023diffusion}. This algorithm unifies ideas from denoising diffusion \citep{song2020score, ho2020denoising} and flow matching \citep{lipman2022flow} to solve the SB problem with arbitrary marginal distributions by parameterizing the Markov drift optimized to match the Schrödinger bridge drift through Markovian and reciprocal projections. 

While the IMF procedure provides a theoretically grounded procedure for constructing the Schrödinger bridge through alternating KL projections in path space, its formulation remains abstract and infinite-dimensional. To make these ideas computationally tractable in high-dimensional settings, we can leverage the explicit SDE representation of the Markovian projection, defined as:
\begin{small}
\begin{align}
    &d\boldsymbol{X}_t=\big(\boldsymbol{f}(\boldsymbol{X}_t,t)+\sigma_t^2\mathbb{E}_{\Pi_{T|t}}\left[\nabla  \log\mathbb{Q}_{T|t}(\boldsymbol{X}_T|\boldsymbol{X}_t)|\boldsymbol{X}_t\right]\big)dt+\sigma_td\boldsymbol{B}_t\tag{Forward Markovian Projection SDE}\label{eq:fwd-markov-proj-sde}
\end{align}
\end{small}
and parameterize the forward-time Markov control drift $\sigma_t\mathbb{E}_{\Pi_{T|t}}\left[\nabla  \log\mathbb{Q}_{T|t}(\boldsymbol{X}_T|\boldsymbol{X}_t)|\boldsymbol{X}_t\right]$ with $\boldsymbol{u}_\theta(\boldsymbol{x},t)$ such that it converges to the optimal drift $\boldsymbol{u}^\star$ through the sequence of IMF iterations. 

Although the Markovian projection preserves the bridge measure $\mathbb{M}^\star_t=\Pi_t$ in theory (Proposition \ref{prop:markov-proj}), parameterizing only the forward-time SDE results in errors in practice, where the terminal marginal $p_T$ generated from simulating (\ref{eq:fwd-markov-proj-sde}) may not exactly match the true marginal constraint $\pi_T$. Therefore, to avoid error accumulation during the IMF sequence, we also parameterize the \boldtext{reverse-time Markovian projection}, which we define below.

\begin{proposition}[Forward and Reverse Time Markovian Projections (Proposition 9 in \citet{shi2023diffusion}]
    Given a mixture of bridges $\Pi=\Pi_{0,T}\mathbb{Q}_{\cdot |0,T}$ in the reciprocal class $\Pi\in \mathcal{R}(\mathbb{Q})$ of the reference measure $\mathbb{Q}$ generated by the SDE $d\boldsymbol{X}_t=\boldsymbol{f}(\boldsymbol{X}_t,t)dt+\sigma_td\boldsymbol{B}_t$, the Markovian projection $\mathbb{M}:=\text{proj}_{\mathcal{M}}(\Pi)$ can be written as both forward and reverse time SDEs defined as:
    \begin{small}
    \begin{align}
        &d\boldsymbol{X}_t=\big(\boldsymbol{f}(\boldsymbol{X}_t,t)+\sigma_t^2\mathbb{E}_{\Pi_{T|t}}\left[\nabla  \log\mathbb{Q}_{T|t}(\boldsymbol{X}_T|\boldsymbol{X}_t)|\boldsymbol{X}_t\right]\big)dt+\sigma_td\boldsymbol{B}_t\\
        &d\tilde{\boldsymbol{X}}_s=\big(-\boldsymbol{f}(\boldsymbol{X}_s,T-s)+\sigma_{T-s}^2\mathbb{E}_{\Pi_{0|T-s}}\big[\nabla  \log\mathbb{Q}_{T-s|0}(\tilde{\boldsymbol{X}}_s|\tilde{\boldsymbol{X}}_T)|\tilde{\boldsymbol{X}}_s\big]\big)dt+\sigma_{T-s}d\tilde{\boldsymbol{B}}_s
    \end{align}
    \end{small}
    with initial conditions $\boldsymbol{X}_0\sim \Pi_0$ and $\tilde{\boldsymbol{X}}_0\sim \Pi_T$, respectively.
\end{proposition}

\textit{Proof.} The proof of this proposition follows directly from the definition of the Markovian projection in (\ref{prop:markov-proj}) and applying the (\ref{eq:time-rev-forward-back-sde}) described in Section \ref{subsec:time-reversal}.\hfill $\square$

Parameterizing both the forward-time Markovian projection with drift $\boldsymbol{u}_\theta(\boldsymbol{x},t)$ and the reverse-time Markovian projection with drift $\boldsymbol{u}_\phi(\boldsymbol{x},t)$, we outline the Diffusion Schrödinger Bridge Matching (DSBM) algorithm \citep{shi2023diffusion} as follows.

\purple[Diffusion Schrödinger Bridge Matching]{\label{box:diffusionsbm}
DSBM generates a sequence of Markov projections $(\mathbb{M}^n)_{n\in \mathbb{N}}$ and reciprocal projections $(\Pi^n)_{n\in \mathbb{N}}$ initialized at $\Pi^0:=\pi_{0,T}\mathbb{Q}_{\cdot|0,T}$ by alternating between the following steps:
\begin{enumerate}
    \item [(1a)] Solve the forward-time Markovian projection $\mathbb{M}^{2n+1}:=\text{proj}_{\mathcal{M}}(\Pi^{2n})$ by updating a parameterized drift $\boldsymbol{u}_\theta$ to minimize $\mathcal{L}_{\text{DSBM}}(\theta) :=\text{KL}(\Pi^{2n}\|\mathbb{M}^\theta)$.
    \item[(1b)] Define the reciprocal projection as $\Pi^{2n+1}:=\mathbb{M}^{2n+1}\mathbb{Q}_{\cdot|0,T}$
    \item[(2a)] Solve the backward-time Markovian projection $\mathbb{M}^{2n+2}:=\text{proj}_{\mathcal{M}}(\Pi^{2n+1})$ by updating a parameterized drift $\boldsymbol{u}_\phi$ to minimize $\mathcal{L}_{\text{DSBM}}(\phi) :=\text{KL}(\Pi^{2n+1}\|\mathbb{M}^\phi)$.
    \item[(2b)] Define the reciprocal projection as $\Pi^{2n+2}:=\mathbb{M}^{2n+2}\mathbb{Q}_{\cdot|0,T}$
\end{enumerate}
}

To learn the forward and reverse time Markovian projections in Steps \textbf{(1a)} and \textbf{(2a)}, we can minimize loss functions defined as the KL divergence\footnote{In \citet{shi2023diffusion}, the parameterized control drift is not scaled by the diffusion coefficient, which yields an extra $\sigma_t^2$ in the denominator. Since we define the control as $\sigma_t\boldsymbol{u}_\theta$ in the SDEs, the extra $\sigma_t^2$ vanishes.} as derived in Section \ref{subsec:path-measure-rnd-kl}:
\begin{small}
\begin{align}
    \mathcal{L}_{\text{DSBM}}(\theta)&=\int_0^T\mathbb{E}_{\Pi_{t,T}}\bigg[\left\|\sigma_t\nabla\log \mathbb{Q}_{T|t}(\boldsymbol{X}_T|\boldsymbol{X}_t)-\boldsymbol{u}_\theta(\boldsymbol{X}_t,t)\right\|^2\bigg]dt\tag{Forward DSBM Loss}\label{eq:dsbm-loss-theta}\\
    \mathcal{L}_{\text{DSBM}}(\phi)&=\int_0^T\mathbb{E}_{\Pi_{t,0}}\bigg[\left\|\sigma_t\nabla  \log \mathbb{Q}_{t|0}(\boldsymbol{X}_t|\boldsymbol{X}_0)-\boldsymbol{u}_\phi(\boldsymbol{X}_t,t)\right\|^2\bigg]dt\tag{Reverse DSBM Loss}\label{eq:dsbm-loss-phi}
\end{align}
\end{small}
Given sufficient expressivity of $\theta, \phi$, it is easy to see that optimizing the above losses for all $(\boldsymbol{x},t)$ exactly yields the control drift of the Markov projection $\text{proj}_{\mathcal{M}}(\Pi)$:
\begin{align}
    \boldsymbol{u}_{\theta^\star}(\boldsymbol{x},t)&=\sigma_t\mathbb{E}_{\Pi_{T|t}}\left[\nabla  \log \mathbb{Q}_{T|t}(\boldsymbol{X}_T|\boldsymbol{X}_t)|\boldsymbol{X}_t=\boldsymbol{x}\right]\\
    \boldsymbol{u}_{\phi^\star}(\boldsymbol{x},t)&=\sigma_t\mathbb{E}_{\Pi_{0|t}}\left[\nabla  \log \mathbb{Q}_{t|0}(\boldsymbol{X}_t|\boldsymbol{X}_0)|\boldsymbol{X}_t=\boldsymbol{x}\right]
\end{align}
The corresponding reciprocal projections performed in \textbf{(1b)} and \textbf{(2b)} are obtained by first simulating trajectories $\boldsymbol{X}_{0:T}$ either with the forward SDE using $\boldsymbol{u}_\theta$ or in the reverse SDE using $\boldsymbol{u}_\phi$ to obtain samples from the endpoint law $(\boldsymbol{x}_0, \boldsymbol{x}_T) \sim \mathbb{M}_{0,T}$, and then sampling from the conditional bridge $\boldsymbol{X}_t\sim\mathbb{Q}(\cdot|\boldsymbol{x}_0, \boldsymbol{x}_T)$ of the reference process $\mathbb{Q}$.

\begin{figure}
    \centering
    \includegraphics[width=\linewidth]{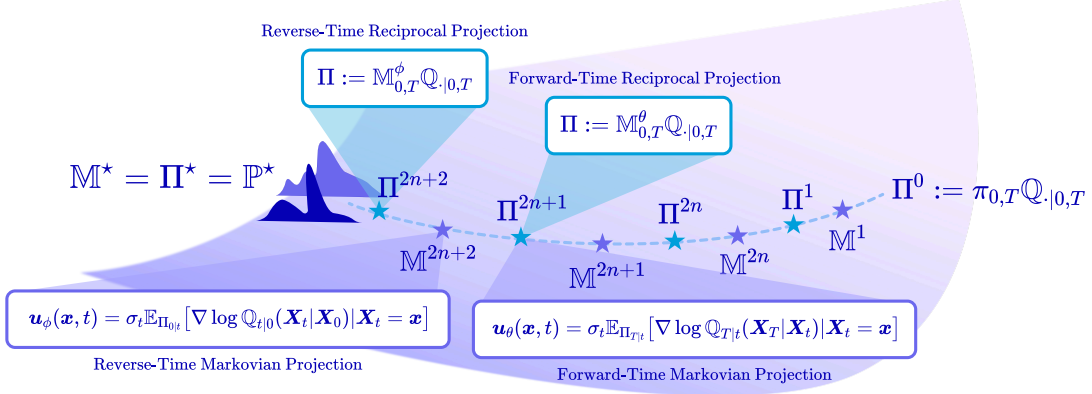}
    \caption{\textbf{Diffusion Schrödinger Bridge Matching.} Illustration of the diffusion Schrödinger bridge matching (DSBM) procedure with alternating Markovian and reciprocal projections. Starting from an initial reciprocal measure $\Pi^0=\pi_{0,T}\mathbb{Q}_{\cdot|0,T}$, the algorithm alternates between \textbf{(i)} forward-time Markovian projections $\mathbb{M}^{2n+1}$ which updates the parameterized forward drift $\boldsymbol{u}_\theta$ to match $\Pi^{2n}$ and \textbf{(ii)} reverse-time Markovian projections $\mathbb{M}^{2n+2}$ which updates the parameterized reverse drift $\boldsymbol{u}_\phi$ to match $\Pi^{2n+1}$. Each Markovian projection is followed by a reciprocal projection that constructs a mixture of reference bridges conditioned on the updated endpoint laws, denoted $\Pi^{2n+1}$ and $\Pi^{2n+2}$. This alternating procedure produces a sequence of path measures that progressively reduce the KL divergence to the optimal bridge, converging to the Schrödinger bridge solution which occurs when the Markov and reciprocal projections reach equilibrium $\mathbb{M}^\star=\Pi^\star=\mathbb{P}^\star$.}
    \label{fig:diffusion-sbm}
\end{figure}

Optimizing these losses through Algorithm \ref{box:diffusionsbm} exactly performs the Iterative Markovian Fitting (IMF) procedure from Section \ref{subsec:markov-reciprocal-proj} through parameterized control drifts of SDEs. Therefore, by Theorem \ref{thm:imf-convergence}, we have that the unique fixed point of the diffusion SBM algorithm yields the optimal Schrödinger bridge $\mathbb{P}^\star$. Rather than separating the forward and reverse Markovian projection steps, \citet{shi2023diffusion} shows that they can be performed simultaneously, which mimics the true IMF procedure where each projection corresponds to the exact Markov and reciprocal projections. 

\purple[Joint Training of Forward and Reverse Markovian Projections]{
Since both objectives can be computed from sampling a pair $(\boldsymbol{x}_0, \boldsymbol{x}_T)\sim \Pi_{0,T}$ from the joint distribution and sampling the intermediate state $\boldsymbol{X}_t\sim \mathbb{Q}_{t|0,T}(\cdot|\boldsymbol{x}_0, \boldsymbol{x}_T)$ from the bridge measure, we can optimize both $\theta$ and $\phi$ \textit{jointly}. After each iteration $n$, the updated parameters can be used to define the forward Markov process $\mathbb{M}^{n+1}_f$ and the backward Markov process $\mathbb{M}^{n+1}_b$ with the following SDEs:
\begin{align}
    \mathbb{M}^{n+1}_f&:d\boldsymbol{X}_t=[\boldsymbol{f}(\boldsymbol{X}_t,t)+\sigma_t\bluetext{\boldsymbol{u}_{\theta}(\boldsymbol{X}_t,t)}]dt+\sigma_td\boldsymbol{B}_t, & \boldsymbol{X}_0\sim \pi_0\label{eq:dsbm-forward-markov}\\
    \mathbb{M}^{n+1}_b&:d\tilde{\boldsymbol{X}}_s=[-\boldsymbol{f}(\tilde{\boldsymbol{X}},T-s)+\sigma_{T-s}\pinktext{\boldsymbol{u}_{\phi}(\tilde{\boldsymbol{X}}_s,T-s)}]ds+\sigma_{T-s}d\boldsymbol{B}_s&\tilde{\boldsymbol{X}}_0\sim \pi_T\label{eq:dsbm-backward-markov}
\end{align}
At equilibrium, \textbf{the forward and backward SDEs should match}, which means reversing $\mathbb{M}_f^n$ yields $\mathbb{M}_b^n$ and reversing $\mathbb{M}_b^n$ yields the $\mathbb{M}^n_f$. To enforce this during training, we can compute the time-reversal of the backward SDE (\ref{eq:dsbm-backward-markov}) using the (\ref{eq:time-rev-forward-back-sde}) to get:
\begin{align}
    d\boldsymbol{X}_t=[\boldsymbol{f}(\boldsymbol{X}_t,t)+\sigma_t(\bluetext{\underbrace{-\boldsymbol{u}_{\phi}(\boldsymbol{X}_t,t)+\sigma_t\nabla \log \Pi_t^{2n}(\boldsymbol{X}_t) }_{\text{should match }\boldsymbol{u}_\theta(\boldsymbol{X}_t,t)}})]dt+\sigma_td\boldsymbol{B}_t, \quad \boldsymbol{X}_0\sim \pi_0\label{eq:dsbm-reverse-bwd}
\end{align}
To ensure the control drifts in (\ref{eq:dsbm-reverse-bwd}) and (\ref{eq:dsbm-forward-markov}) are aligned, we leverage a consistency loss defined as:
\begin{small}
\begin{align}
    &\mathcal{L}_{\text{cons}}(\theta, \phi)=\int_0^T\mathbb{E}_{\Pi_t^{2n}}\bigg[\|\boldsymbol{u}_\theta(\boldsymbol{X}_t,t)+\boldsymbol{u}_\phi(\boldsymbol{X}_t,t)-\sigma_t\bluetext{\nabla  \log \Pi^{2n}_t(\boldsymbol{X}_t)}\|^2\bigg]dt\tag{Consistency Loss}\\
    &=\int_0^T\mathbb{E}_{\Pi_t^{2n}}\bigg[\left\|\boldsymbol{u}_\theta(\boldsymbol{X}_t,t)+\boldsymbol{u}_\phi(\boldsymbol{X}_t,t)-\sigma_t\bluetext{\left(\mathbb{E}_{\Pi_{T|t}^{2n}}\left[\nabla \log \mathbb{Q}_{T|t}|\boldsymbol{x}\right]+\mathbb{E}_{\Pi_{0|t}^{2n}}\left[\nabla \log \mathbb{Q}_{t|0}|\boldsymbol{x}\right]\right)}\right\|^2\bigg]dt\nonumber
\end{align}
\end{small}
where we rewrite the score function $\nabla \log \Pi^{2n}_t(\boldsymbol{X}_t)$ with the known conditional densities $\mathbb{Q}_{T|t}$ and $\mathbb{Q}_{0|t}$ using the identity:
\begin{align}
    \nabla  \log \Pi^{2n}_t(\boldsymbol{x})=\mathbb{E}_{\Pi_{T|t}^{2n}}\left[\nabla\log \mathbb{Q}_{T|t}|\boldsymbol{X}_t=\boldsymbol{x}\right]+\mathbb{E}_{\Pi_{0|t}^{2n}}\left[\nabla\log \mathbb{Q}_{t|0}|\boldsymbol{X}_t=\boldsymbol{x}\right]
\end{align}
which defines the marginal score at time $t$ as the sum of the conditional Gaussian scores at the endpoints. Therefore, the total \textbf{joint training loss} can be defined as:
\begin{align}
    \mathcal{L}_{\text{DSBM}}(\theta, \phi)=\mathcal{L}_{\text{DSBM}}(\theta)+\mathcal{L}_{\text{DSBM}}(\phi)+\lambda\mathcal{L}_{\text{cons}}(\theta, \phi)
\end{align}
where $\lambda>0$ is a positive weight that defines the strength of the consistency loss.
}
This formulation provides a tractable approach for performing path-space projections using parameterized drifts that can be simulated at inference time via an SDE solver. In practice, the Markov projection requires learning both forward and backward drifts to satisfy the marginal constraints, despite the underlying equivalence of the bridge measures. However, this approach remains computationally intensive, as training still relies on simulating full stochastic trajectories due to the absence of a closed-form sampling procedure for the optimal bridge. Motivated by this limitation, we now consider an alternative perspective in which the intermediate states admit tractable expression, enabling simulation-free matching objectives. 

\subsection{Simulation-Free Score and Flow Matching}
\label{subsec:score-and-flow}
\textit{Prerequisite: Section \ref{subsect:mixture-bridges}, \ref{subsec:time-reversal}}

To overcome the restriction to the Gaussian prior distribution of score-based generative modeling described in Section \ref{subsec:primer-sgm}, we highlight the simulation-free score and flow matching ([SF]$^2$M; \citet{tong2023simulation}) framework which extends score matching to arbitrary prior distributions. Crucially, we will show that given the \textbf{optimal entropic OT coupling} (Section \ref{eq:static-sb}), [SF]$^2$M solves the (\ref{eq:dynamic-sb-problem}) through a \textbf{endpoint-conditioned objective} with the same gradient as the unconditional objective.

\begin{figure}
    \centering
    \includegraphics[width=\linewidth]{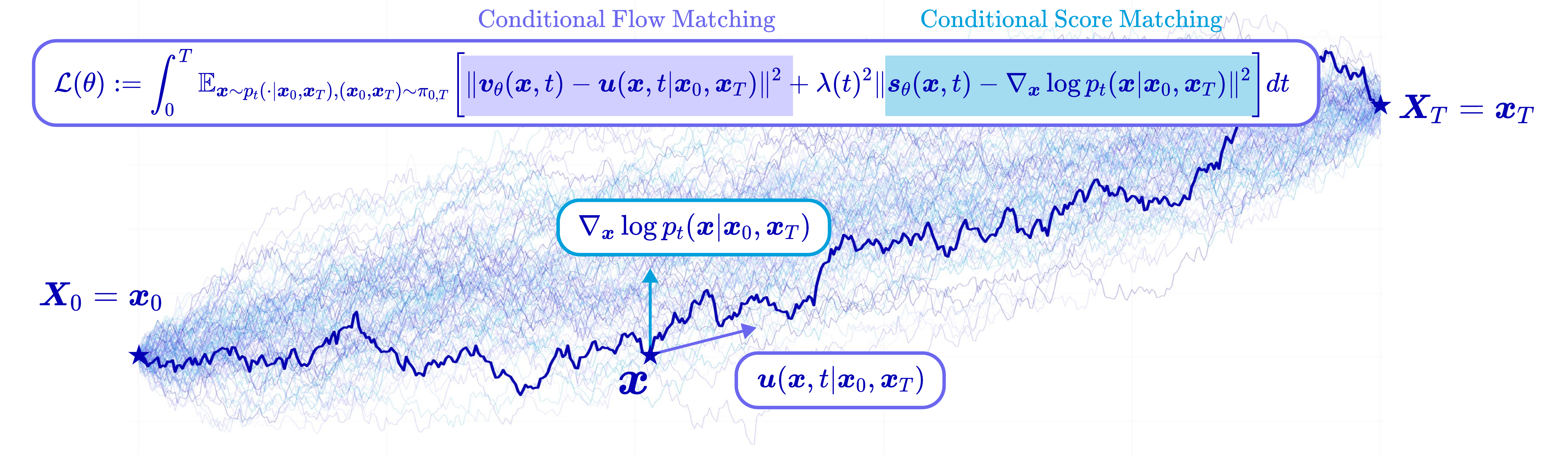}
    \caption{\textbf{Conditional Score and Flow Matching.} Conditional score and flow matching learn the dynamics of stochastic bridges by conditioning trajectories on endpoint pairs $(\boldsymbol{x}_0, \boldsymbol{x}_T)\sim \pi_{0,T}$. The model jointly estimates the conditional control drift $\boldsymbol{u}(\boldsymbol{x},t|\boldsymbol{x}_0, \boldsymbol{x}_T)$ and the conditional score $\nabla \log p_t(\boldsymbol{x}|\boldsymbol{x}_0, \boldsymbol{x}_T)$, which describe how trajectories evolve between the endpoints. By minimizing the combined objective, the learned dynamics produce the mixture of Brownian bridges which is equal to the Schrödinger bridge $\mathbb{P}^\star$ given the entropic OT coupling $\pi_{0,T}^\star$.}
    \label{fig:score-and-flow}
\end{figure}

This framework considers a data-driven SB problem, where we have empirical samples from both the marginal distributions $\boldsymbol{x}_0\sim \pi_0$ and $\boldsymbol{x}_T\sim \pi_T$. In this setting, we can solve the entropic optimal transport (OT) problem between empirical samples to determine the optimal coupling $\pi^\star_{0,T}$, from which the solution to the dynamic SB problem with Brownian reference process $\mathbb{Q}$ is defined simply as a \textbf{mixture of Brownian bridges weighted by the optimal static coupling $\pi^\star_{0,T}$} as proven in Proposition \ref{prop:condsoc-mixture} and Corollary \ref{corollary:mixture-brownian}. 

Recall from Section \ref{subsec:time-reversal}, where we derived the (\ref{eq:controlled-time-reversal-formula}) as:
\begin{align}
\begin{cases}
        d\boldsymbol{X}_t=\sigma_t\boldsymbol{u}(\boldsymbol{X}_t,t)dt+\sigma_td\boldsymbol{B}_t\\
        d\tilde{\boldsymbol{X}}_s=-\sigma_t\boldsymbol{u}(\tilde{\boldsymbol{X}}_s,T-s)+\sigma_{T-s}^2\nabla\log p(\tilde{\boldsymbol{X}}_s,T-s)ds+\sigma_{T-s}d\widetilde{\boldsymbol{B}}_s
    \end{cases}
\end{align}
where we define the reference drift as pure Brownian motion $\boldsymbol{f}:=0$. Crucially, the score function $\nabla \log p_t(\boldsymbol{X}_t)$ appears in the reverse-time drift as a \textit{correction} term that compensates for the entropy-producing forward diffusion, ensuring that the reversed dynamics reproduce the correct marginal distributions. This decomposition of the control and score function motivates a \textbf{combined objective} that learns a parameterized control drift $\boldsymbol{u}_\theta$, which learns the \textit{reverse control} $\boldsymbol{u}_\theta(\boldsymbol{x},t)\approx\boldsymbol{u}(\boldsymbol{x},t)$\footnote{since the control is non-deterministic, we will just consider the reverse control as $\boldsymbol{u}(\boldsymbol{x},t)$ and omit the negative sign for simplicity.},  and the score function $\nabla \log p_t(\boldsymbol{x})$.

\begin{definition}[Unconditional Score and Flow Matching Objective]\label{def:uncond-score-flow}
    The \textbf{unconditional score and flow matching objective} aims to match a parameterized control field $\boldsymbol{u}_\theta(\boldsymbol{x},t):\mathbb{R}^d\times[0,T] \to \mathbb{R}^d$ and score function $\boldsymbol{s}_\theta(\boldsymbol{x},t):\mathbb{R}^d\times[0,T]\to \mathbb{R}^d$ to the true velocity and score function defining the solution to (\ref{eq:dynamic-sb-problem}) by minimizing:
    \begin{small}
    \begin{align}
        \mathcal{L}_{\text{U[SF]}^2\text{M}}(\theta):=\int_0^T\mathbb{E}_{p_t}\bigg[\underbrace{\|\boldsymbol{u}_\theta(\boldsymbol{x},t)-\boldsymbol{u}(\boldsymbol{x},t)\|^2}_{\text{flow matching loss}}+\underbrace{\lambda(t)^2\|\boldsymbol{s}_\theta(\boldsymbol{x},t)-\nabla \log p^\star_t(\boldsymbol{x})\|^2}_{\text{score matching loss}}\bigg]dt \tag{Unconditional Objective}\label{eq:uncond-score-flow}
    \end{align}
    \end{small}
    where $p^\star_t$ is the optimal marginal density of the dynamic SB, and $\lambda(t):[0,T] \to \mathbb{R}$ is some positive weight.
\end{definition}

While this objective is theoretically sound, both $\boldsymbol{u}(\boldsymbol{x},t)$ and $\nabla \log p_t(\boldsymbol{x})$ are undefined or intractable for general target distributions. In this setting, we assume access to explicit samples from both marginal distributions $\pi_0$ and $\pi_T$ and define the tractable control drift and score for a Brownian bridge between a predefined coupling $(\boldsymbol{x}_0, \boldsymbol{x}_T) \sim \pi_{0,T}$ given by:
\begin{small}
\begin{align}
    \boldsymbol{u}(\boldsymbol{x},t|\boldsymbol{x}_0, \boldsymbol{x}_T)&=\frac{1-2t}{t(1-t)}(\boldsymbol{x}-(t\boldsymbol{x}_1+(1-t)\boldsymbol{x}_0))+(\boldsymbol{x}_1-\boldsymbol{x}_0)\nonumber\\
    \nabla \log p_t(\boldsymbol{x}| \boldsymbol{x}_0, \boldsymbol{x}_T)&=\frac{t\boldsymbol{x}_1+(1-t)\boldsymbol{x}_0-\boldsymbol{x}}{\sigma_t^2t(1-t)}
\end{align}
\end{small}

Using this definition of the endpoint-conditioned velocity and score function, we can define the conditional velocity and score over the empirical distribution $\pi_0$ by taking an expectation:
\begin{small}
\begin{align}
    \boldsymbol{u}(\boldsymbol{x},t)&=\mathbb{E}_{\pi_{0,T}}\left[\frac{\boldsymbol{u}(\boldsymbol{x},t|\boldsymbol{x}_0, \boldsymbol{x}_T)p_t(\boldsymbol{x}|\boldsymbol{x}_0, \boldsymbol{x}_T)}{p_t(\boldsymbol{x})}\right]\\
    \nabla \log p_t(\boldsymbol{x})&=\mathbb{E}_{\pi_{0,T }}\left[\frac{p_t(\boldsymbol{x}|\boldsymbol{x}_0, \boldsymbol{x}_T)}{p_t(\boldsymbol{x})}\nabla \log p_t(\boldsymbol{x}| \boldsymbol{x}_0, \boldsymbol{x}_T)\right]
\end{align}
\end{small}

These tractable definitions for the \textit{conditional} control drift and score function motivate the definition of the \boldtext{conditional score and flow matching objective} \citep{tong2023simulation}, which we show yields the same gradients as the unconditional objective.

\begin{proposition}[Conditional Score and Flow Matching Objective (Theorem 3.2 in \citet{tong2023simulation})]\label{def:score-flow-objective}
Consider the \textbf{conditional score and flow matching objective} which aims to match a parameterized control drift $\boldsymbol{v}_\theta(\boldsymbol{x},t):\mathbb{R}^d\times[0,T]\to \mathbb{R}^d$ and score function $\boldsymbol{s}_\theta(\boldsymbol{x},t):\mathbb{R}^d\times[0,T]\to \mathbb{R}^d$ to a distribution of velocity and score functions conditioned on the endpoint $\boldsymbol{z} \sim \pi_T$ by minimizing:
\begin{small}
\begin{align}
    \mathcal{L}_{\text{[SF]}^2\text{M}}(\theta):=\int_0^T\mathbb{E}_{p_{t|0,T}, \pi_{0,T}}\bigg[\underbrace{\|\boldsymbol{v}_\theta(\boldsymbol{x},t)-\boldsymbol{u}(\boldsymbol{x},t|\boldsymbol{x}_0,\boldsymbol{x}_T)\|^2}_{\text{conditional flow matching loss}}+\underbrace{\lambda(t)^2\|\boldsymbol{s}_\theta(\boldsymbol{x},t)-\nabla \log p_t(\boldsymbol{x}|\boldsymbol{x}_0,\boldsymbol{x}_T)\|^2}_{\text{conditional score matching loss}}\bigg]dt \tag{Conditional Objective}\label{eq:sf2m-objective}
\end{align}
\end{small}
where the expectation is taken over samples from the endpoint law $(\boldsymbol{x}_0,\boldsymbol{x}_T)\sim \pi_{0,T}$ and samples from the conditional distribution at time $t\in [0,T]$ given target endpoints $\boldsymbol{x}\sim p_{t|0,T}(\cdot|\boldsymbol{x}_0,\boldsymbol{x}_T)$. Then, we have that the gradients of the conditional objective match the gradients of (\ref{eq:uncond-score-flow}) such that $\nabla_{\theta}\mathcal{L}_{\text{[SF]}^2\text{M}}(\theta)=\nabla_{\theta}\mathcal{L}_{\text{U[SF]}^2\text{M}}(\theta)$.
\end{proposition}

\textit{Proof.} The goal of this proof is to show the equivalence between the gradients of the conditional and unconditional expectations:
\begin{align}
    \nabla_\theta\mathbb{E}_{p_{t|0,T}, \pi_{0,T}}\left[\|\boldsymbol{w}_\theta(\boldsymbol{x},t)-\boldsymbol{w}(\boldsymbol{x},t|\boldsymbol{x}_0,\boldsymbol{x}_T)\|^2\right]=\nabla_\theta\mathbb{E}_{p_t}\left[\|\boldsymbol{w}_\theta(\boldsymbol{x},t)-\boldsymbol{w}(\boldsymbol{x},t)\|^2\right]
\end{align}
which can be applied for both the conditional flow matching loss with $\boldsymbol{w}(\boldsymbol{x},t|\boldsymbol{x}_0,\boldsymbol{x}_T):= \boldsymbol{u}(\boldsymbol{x},t|\boldsymbol{x}_0,\boldsymbol{x}_T)$ and the conditional score matching loss $\boldsymbol{w}(\boldsymbol{x},t|\boldsymbol{x}_0,\boldsymbol{x}_T):= \nabla \log p_t(\boldsymbol{x}|\boldsymbol{x}_0,\boldsymbol{x}_T)$. Expanding the squared loss in the conditional objective, we have:
\begin{small}
\begin{align}
    \|\boldsymbol{w}_\theta(\boldsymbol{x},t)-\boldsymbol{w}(\boldsymbol{x},t|\boldsymbol{x}_0,\boldsymbol{x}_T)\|^2=\underbrace{\|\boldsymbol{w}_\theta(\boldsymbol{x},t)\|^2}_{\text{independent of }\boldsymbol{x}_0,\boldsymbol{x}_T}\bluetext{-2\langle \boldsymbol{w}_\theta(\boldsymbol{x},t), \boldsymbol{w}(\boldsymbol{x},t|\boldsymbol{x}_0,\boldsymbol{x}_T)\rangle}+\underbrace{\|\boldsymbol{w}(\boldsymbol{x},t|\boldsymbol{x}_0,\boldsymbol{x}_T)\|^2}_{\text{independent of }\theta}
\end{align}
\end{small}
where we observe that the first term is independent of the conditional pair $(\boldsymbol{x}_0,\boldsymbol{x}_T)$, which means that it is clearly equivalent to the unconditional gradient. The last term is independent of $\theta$ which has a gradient of zero with respect to $\theta$. Therefore, we can write the difference between the unconditional and conditional expectations as:
\begin{small}
\begin{align}
    \nabla_\theta\mathbb{E}_{p_{t|0,T},  \pi_{0,T}}\left[\|\boldsymbol{w}_\theta(\boldsymbol{x},t)-\boldsymbol{w}(\boldsymbol{x},t|\boldsymbol{x}_0,\boldsymbol{x}_T)\|^2\right]&=\nabla_\theta\mathbb{E}_{ p_t}\left[\|\boldsymbol{w}_\theta(\boldsymbol{x},t)-\boldsymbol{w}(\boldsymbol{x},t)\|^2\right]\nonumber\\
    \nabla_\theta\mathbb{E}_{p_{t|0,T},  \pi_{0,T}}\left[\bluetext{-2\langle \boldsymbol{w}_\theta(\boldsymbol{x},t), \boldsymbol{w}(\boldsymbol{x},t|\boldsymbol{x}_0,\boldsymbol{x}_T)\rangle}\right]&=\nabla_\theta\mathbb{E}_{p_t}\left[\bluetext{-2\langle \boldsymbol{w}_\theta(\boldsymbol{x},t),\boldsymbol{w}(\boldsymbol{x},t)\rangle} \right]\nonumber\\
    \nabla_\theta\mathbb{E}_{p_{t|0,T},  \pi_{0,T}}\left[\langle \boldsymbol{w}_\theta(\boldsymbol{x},t), \boldsymbol{w}(\boldsymbol{x},t|\boldsymbol{x}_0,\boldsymbol{x}_T)\rangle\right]&=\nabla_\theta\mathbb{E}_{p_t}\left[\langle \boldsymbol{w}_\theta(\boldsymbol{x},t),\boldsymbol{w}(\boldsymbol{x},t)\rangle \right]
\end{align}
\end{small}
Now, we aim to show that the expectations are equivalent. Starting from the 
\begin{small}
\begin{align}
    \nabla_\theta\mathbb{E}_{p_t}\left[\langle \boldsymbol{w}_\theta(\boldsymbol{x},t),\boldsymbol{w}(\boldsymbol{x},t)\rangle \right]&=\int_{\mathbb{R}^d}\langle \boldsymbol{w}_\theta(\boldsymbol{x},t),\bluetext{\boldsymbol{w}(\boldsymbol{x},t)}\rangle p_t(\boldsymbol{x})d\boldsymbol{x}\nonumber\\
    &=\int_{\mathbb{R}^d}\left\langle \boldsymbol{w}_\theta(\boldsymbol{x},t),\bluetext{\int_{\mathbb{R}^d}\frac{p_t(\boldsymbol{x}|\boldsymbol{x}_0,\boldsymbol{x}_T)}{p_t(\boldsymbol{x})}\boldsymbol{w}(\boldsymbol{x},t|\boldsymbol{x}_0,\boldsymbol{x}_T)\pi_{0,T}(\boldsymbol{x}_0,\boldsymbol{x}_T)d\boldsymbol{x}_0d\boldsymbol{x}_T}\right\rangle p_t(\boldsymbol{x})d\boldsymbol{x}\nonumber\\
    &\overset{(\bigstar)}{=}\bluetext{\int_{\mathbb{R}^d}}\int_{\mathbb{R}^d}\left\langle \boldsymbol{w}_\theta(\boldsymbol{x},t),\boldsymbol{w}(\boldsymbol{x},t|\boldsymbol{x}_0,\boldsymbol{x}_T)\right\rangle \bluetext{\frac{p_t(\boldsymbol{x}|\boldsymbol{x}_0,\boldsymbol{x}_T)}{p_t(\boldsymbol{x})}\pi_T(\boldsymbol{x}_0,\boldsymbol{x}_T)}p_t(\boldsymbol{x})d\boldsymbol{x}\bluetext{d\boldsymbol{x}_0d\boldsymbol{x}_T}\nonumber\\
    &=\bluetext{\int_{\mathbb{R}^d}}\int_{\mathbb{R}^d}\left\langle \boldsymbol{w}_\theta(\boldsymbol{x},t),\boldsymbol{w}(\boldsymbol{x},t|\boldsymbol{x}_0,\boldsymbol{x}_T)\right\rangle \bluetext{p_t(\boldsymbol{x}|\boldsymbol{x}_0,\boldsymbol{x}_T)\pi_{0,T}(\boldsymbol{x}_0,\boldsymbol{x}_T)}d\boldsymbol{x}\bluetext{d\boldsymbol{x}_0d\boldsymbol{x}_T}\nonumber\\
    &=\boxed{\nabla_\theta\mathbb{E}_{\bluetext{p_{t|0,T}, \pi_{0,T}}}\left[\langle \boldsymbol{w}_\theta(\boldsymbol{x},t),\bluetext{\boldsymbol{w}(\boldsymbol{x},t|\boldsymbol{x}_0,\boldsymbol{x}_T)}\rangle \right]}
\end{align}
\end{small}
where $(\bigstar)$ follows from factoring out the scalar values and applying Fubini's theorem to change the order of integration, resulting in the equivalence between the marginal and conditional objectives. \hfill $\square$

This result is well established in flow matching literature \citep{lipman2022flow, tong2023improving} as a way of training parameterized flows that approximate an intractable marginal distribution using empirical samples from the target data distribution. However, this marginal distribution does not yet solve the Schrödinger bridge problem, as the coupled distribution $\pi_{0,T}$ from which $(\boldsymbol{x}_0, \boldsymbol{x}_T)$ is sampled does not necessarily align with the entropic OT coupling. To establish how the conditional score and flow matching objective can be used to solve the (\ref{eq:dynamic-sb-problem}), we establish the following proposition.

\begin{proposition}[Score and Flow Matching Recovers the Schrödinger Bridge (Proposition 3.4 in \citet{tong2023simulation})]
    Let $\mathbb{P}^\star$ denote the path measure that solves the dynamic SB with marginal constraints $\pi_0, \pi_T\in \mathcal{P}(\mathbb{R}^d)$ and pure Brownian motion reference process $\sigma\mathbb{B}$. Consider the optimal endpoint law $\pi_{0,T}^\star$ that solves the (\ref{eq:entropic-ot-problem}) with quadratic transport cost $c(\boldsymbol{x},\boldsymbol{y}):= \|\boldsymbol{x}-\boldsymbol{y}\|^2 $:
    \begin{small}
    \begin{align}
        \pi^\star_{0,T}=\underset{\pi_{0,T}\in \Pi(\pi_0, \pi_T)}{\arg\min}\left\{\int_{\mathbb{R}^d\times \mathbb{R}^d}\|\boldsymbol{x}-\boldsymbol{y}\|^2 d\pi_{0,T}(\boldsymbol{x},\boldsymbol{y})+ 2\sigma^2_t\text{KL} (\pi_{0,T}\|\pi_0\otimes \pi_T)\right\}
    \end{align}
    \end{small}
    If the parameterized score function $\boldsymbol{s}^\star_\theta(\boldsymbol{x},t)$ and control drift $\boldsymbol{v}^\star_\theta(\boldsymbol{x},t)$ globally minimize the (\ref{eq:sf2m-objective}) under the coupling $(\boldsymbol{x}_0, \boldsymbol{x}_T)\sim \pi_{0,T}^\star$, then the resulting stochastic process is given by the SDE:
    \begin{small}
    \begin{align}
       d\boldsymbol{X}_t=\left[\boldsymbol{v}_\theta^\star(\boldsymbol{X}_t,t)+\sigma_t^2\boldsymbol{s}_\theta^\star(\boldsymbol{X}_t,t)\right]dt+\sigma_td\boldsymbol{B}_t, \quad \boldsymbol{X}_0 \sim \pi_0
    \end{align}
    \end{small}
    and generates the Schrödinger bridge path measure $\mathbb{P}^\star$.
\end{proposition}

\textit{Proof Sketch.} The \textbf{key idea} is that the (\ref{eq:sf2m-objective}) is minimized pointwise for all $(\boldsymbol{x},t)\in \mathbb{R}^d\times [0,T]$ when:
\begin{small}
\begin{align}
    \boldsymbol{v}_\theta(\boldsymbol{x},t)=\mathbb{E}_{(\boldsymbol{x}_0, \boldsymbol{x}_T) \sim p_{0,T|t}}\left[\boldsymbol{u}(\boldsymbol{x},t|\boldsymbol{x}_0,\boldsymbol{x}_T)\right], \quad \boldsymbol{s}_\theta(\boldsymbol{x},t)=\mathbb{E}_{(\boldsymbol{x}_0, \boldsymbol{x}_T) \sim p_{0,T|t}}\left[\nabla \log p_t(\boldsymbol{x}|\boldsymbol{x}_0,\boldsymbol{x}_T)\right]
\end{align}
\end{small}
where $p_{0,T|t}$ is the posterior distribution over the endpoint law $\pi_{0,T}$ given an intermediate state $\boldsymbol{x}$ which can be expressed using Bayes' rule as:
\begin{align}
    p_{0,T|t}(\boldsymbol{x}_0, \boldsymbol{x}_T|\boldsymbol{x})=\frac{p_{t|0,T}(\boldsymbol{x}|\boldsymbol{x}_0, \boldsymbol{x}_T) \pi_{0,T}(\boldsymbol{x}_0,\boldsymbol{x}_T)}{p_t(\boldsymbol{x})}
\end{align}
This minimizer is exactly the probability flow drift and score function of the \textbf{mixture of Brownian bridges}. From Section \ref{subsect:mixture-bridges} Corollary \ref{corollary:mixture-brownian}, we showed if $\pi^\star_{0,T}$ is chosen to be the \textbf{entropic OT plan} with quadratic transport cost, then this bridge mixture is precisely the Schrödinger bridge, so the learned SDE recovers the Schrödinger bridge $\mathbb{P}^\star$. \hfill $\square$

By leveraging a two-stage framework that first determines the optimal static SB coupling and learning the conditional velocity and score functions, score and flow matching provide a scalable, simulation-free framework for learning Schrödinger bridges. However, a key limitation of this approach is that it requires explicit samples from both the source and target distributions to construct the conditional objectives and endpoint couplings. In many practical settings, such as those that are only known up to an unnormalized density or energy function, explicit samples may not be readily available. This motivates our discussion of alternative approaches that do not rely on paired or explicit samples from the target distribution.

\subsection{Schrödinger Bridge with Adjoint Matching}
\label{subsec:adjoint-matching}
\textit{Prerequisite: Section \ref{sec:sb-optimal-control}}

In this section, we will explore how the \textbf{adjoint matching} framework \citep{domingo2024adjoint} has been applied to efficiently solve the Schrödinger bridge problem as described in \textit{Adjoint Schrödinger Bridge Sampler} \citep{liu2025adjoint}. \boldtext{Adjoint matching} (AM; \citet{domingo2024adjoint}) is a generative modeling framework that efficiently solves the stochastic optimal control (SOC) problem, which has been extended to various applications, including fine-tuning \citep{domingo2024adjoint} and sampling  \citep{havens2025adjoint, park2025functional}. The \boldtext{adjoint variable} in the context of the Schrödinger bridge problem refers to the gradient of the value function $V_t(\boldsymbol{x})$ which defines the optimal control drift: 
\begin{align}
    \boldsymbol{u}^\star(\boldsymbol{x},t)=\sigma_t\nabla \psi_t(\boldsymbol{x})=-\sigma_t\nabla V_t(\boldsymbol{x})=-\sigma_t\nabla J^\star(\boldsymbol{x},t;u)
\end{align}
Standard methods for solving for the adjoint variable by directly differentiating through the (\ref{eq:soc-objective}) \citep{han2016deep} or directly match the target $\nabla J^\star(\boldsymbol{x},t;u^\star)$ with importance weighted matching objective \citep{rubinstein2004cross, zhang2014applications, domingo2024stochastic}, however, adjoint matching introduces a computationally favorable and fundamentally different approach. 

\begin{definition}[Adjoint State]\label{def:adjoint-state}
    The \textbf{adjoint state}, denoted $\boldsymbol{a}:C([t,T], \mathbb{R}^d)\times[0,T]\to \mathbb{R}^d$, of a stochastic optimal control (SOC) problem is defined by taking the gradient of the SOC objective defined in (\ref{eq:soc-objective}) to get: 
    \begin{align}
        \boldsymbol{a}(\boldsymbol{X}_{t:T},t)=\nabla \left(\int_t^T\left(\frac{1}{2}\|\boldsymbol{u}(\boldsymbol{X}_s,s)\|^2+c(\boldsymbol{X}_s,s)\right)ds+\Phi(\boldsymbol{X}_T)\right)
    \end{align}
    which yields the gradient field of $J(\boldsymbol{x},t;\boldsymbol{u})$ in expectation:
    \begin{align}
        \nabla J(\boldsymbol{x},t;\boldsymbol{u})=\mathbb{E}_{\boldsymbol{X}_{t:T}\sim \mathbb{P}^u}\left[\boldsymbol{a}(\boldsymbol{X}_{t:T},t)|\boldsymbol{X}_t=\boldsymbol{x}\right]\label{eq:adjoint-state-2}
    \end{align}
    The adjoint state can be solved backward in time, given the terminal condition $\boldsymbol{a}(\boldsymbol{X}_{t:T}, T;\boldsymbol{u})=\nabla_{\boldsymbol{x}_T}\Phi(\boldsymbol{X}_T)$ by integrating:
    \begin{small}
    \begin{align}
        \frac{d}{dt}\boldsymbol{a}(\boldsymbol{X}_{t:T}, t; \boldsymbol{u})&=-\left[\boldsymbol{a}(\boldsymbol{X}_{t:T}, t; \boldsymbol{u})^\top\left(\nabla (\boldsymbol{f}(\boldsymbol{X}_t,t)+\sigma_t\boldsymbol{u}(\boldsymbol{X}_t,t)\right)+\nabla\left(c(\boldsymbol{X}_t,t)+\frac{1}{2}\|\boldsymbol{u}(\boldsymbol{X}_t,t)\|^2\right)\right]\label{eq:adjoint-derivative}
    \end{align}
    \end{small}
    which yields the alternative form:
    \begin{small}
    \begin{align}
        \boldsymbol{a}(\boldsymbol{X}_{t:T}, t;\boldsymbol{u})&=\int_t^T\bigg(\nabla (\boldsymbol{f}(\boldsymbol{X}_s,s)^\top\boldsymbol{a}(\boldsymbol{X}_{t:T}, t;\boldsymbol{u})+\sigma_t\boldsymbol{v}(\boldsymbol{X}_s,s))\nonumber\\
        &+\nabla \bigg(c(\boldsymbol{X}_s,s)+\frac{1}{2}\|\boldsymbol{u}(\boldsymbol{X}_s,s)\|^2\bigg)\bigg)ds+\nabla \Phi(\boldsymbol{X}_T)\label{eq:integral-adjoint-state}
    \end{align}
    \end{small}
\end{definition}

Rather than directly matching the target adjoint vector field $-\sigma_t\nabla J^\star(\boldsymbol{x},t;\boldsymbol{u}^\star)$, it considers an objective that matches the vector field generated by the \textit{current control} $-\sigma_t\nabla J(\boldsymbol{x},t;\boldsymbol{u})$, which bypasses the need for importance weighting while obtaining a optimizer gradient that is \textit{equal}, in expectation to that of the target adjoint objective.

\begin{proposition}[Basic Adjoint Matching Yields the Optimal Control (Proposition 2 in \citet{domingo2024adjoint})]
    Consider \textbf{basic adjoint matching objective} defined with the adjoint state $\boldsymbol{a}:C([0,T];\mathbb{R}^d)\times[0,T]\to \mathbb{R}^d$ as:
    \begin{small}
    \begin{align}
        \mathcal{L}_{\text{basic-AM}}(\boldsymbol{u}):=\mathbb{E}_{\boldsymbol{X}_{0:T}\sim \mathbb{P}^{\bar{u}}}\left[\frac{1}{2}\int_0^T\big\|\boldsymbol{u}(\boldsymbol{X}_t,t)+\sigma_t\nabla \boldsymbol{a}(\boldsymbol{X}_{t:T}, t; \bar{\boldsymbol{u}})\big\|^2dt\right], \quad\bar{\boldsymbol{u}}=\texttt{stopgrad}(\boldsymbol{u})\label{eq:basic-am-loss}
    \end{align}
    \end{small}
    where $\bar{\boldsymbol{u}}=\texttt{stopgrad}(\boldsymbol{u})$ is the control drift where the gradient with respect to $\boldsymbol{u}$ that generates the path are not tracked, i.e., the path $\boldsymbol{X}_{t:T}$ cannot be differentiated through. Then, $\mathcal{L}_{\text{basic-AM}}(\boldsymbol{u})$ has a \textbf{unique} minimizer that is equal to the optimal control $\boldsymbol{u}^\star$.
\end{proposition}

\textit{Proof.} To derive the minimizer of the functional objective (\ref{eq:basic-am-loss}), we can compute the first variation of $\mathcal{L}_{\text{basic-AM}}$ by defining a slightly perturbed control drift $(\boldsymbol{u}+\epsilon\boldsymbol{v})$, where $\boldsymbol{v}:\mathbb{R}^d\times[0,T]\to \mathbb{R}^d$ is an arbitrary vector field, to get:
\begin{align}
    \bluetext{\frac{d}{d\epsilon}}\mathcal{L}_{\text{basic-AM}}(\boldsymbol{u}+\bluetext{\epsilon\boldsymbol{v}})&=\bluetext{\frac{d}{d\epsilon}}\mathbb{E}_{\boldsymbol{X}_{0:T}\sim \mathbb{P}^{\bar{u}}}\left[\frac{1}{2}\int_0^T\left\|\bluetext{(\boldsymbol{u}+\epsilon\boldsymbol{v})}(\boldsymbol{X}_t,t)+\sigma_t\nabla \boldsymbol{a}(\boldsymbol{X}_{t:T}, t; \bar{\boldsymbol{u}})\right\|^2dt\right]\bigg|_{\epsilon=0}
\end{align}
Applying the property $\frac{d}{d\epsilon}\frac{1}{2}\|\boldsymbol{y}+\epsilon\boldsymbol{z}\|^2\big\vert_{\epsilon=0}=\frac{1}{2}\left[2(\boldsymbol{y}+\epsilon\boldsymbol{z})\boldsymbol{z}\right]\vert_{\epsilon =0}=\langle\boldsymbol{y}, \boldsymbol{z}\rangle$, we have:
\begin{small}
\begin{align}
    \bluetext{\frac{d}{d\epsilon}}\mathcal{L}_{\text{basic-AM}}(\boldsymbol{u}+\bluetext{\epsilon\boldsymbol{v}})&=\mathbb{E}_{\boldsymbol{X}_{t:T}\sim \mathbb{P}^{\bar{u}}}\left[\int_0^T\bluetext{\frac{d}{d\epsilon}}\frac{1}{2}\left\|\bluetext{\boldsymbol{u}(\boldsymbol{X}_t,t)+\sigma_t\boldsymbol{a}(\boldsymbol{X}_{t:T}, t; \bar{\boldsymbol{u}})}+\pinktext{\epsilon\boldsymbol{v}(\boldsymbol{X}_t,t)}\right\|^2dt\right]\bigg|_{\epsilon=0}\nonumber\\
    &=\mathbb{E}_{\boldsymbol{X}_{t:T}\sim \mathbb{P}^{\bar{u}}}\left[\int_0^T\big\langle \pinktext{\boldsymbol{v}(\boldsymbol{X}_t, t)},  \bluetext{\boldsymbol{u}(\boldsymbol{X}_t,t)+\sigma_t\boldsymbol{a}(\boldsymbol{X}_{t:T}, t; \bar{\boldsymbol{u}})}\big\rangle dt\right]\nonumber\\
    &=\mathbb{E}_{\boldsymbol{x}\sim p_t^{\bar{u}}}\bigg[\int_0^T\big\langle \boldsymbol{v}(\boldsymbol{x}, t),  \underbrace{\boldsymbol{u}(\boldsymbol{x},t)+\sigma_t\bluetext{\mathbb{E}_{\boldsymbol{X}_{t:T}\sim \mathbb{P}^{\bar{u}}}\left[\boldsymbol{a}(\boldsymbol{X}_{t:T}, t; \bar{\boldsymbol{u}})|\boldsymbol{X}_t=\boldsymbol{x}\right]}}_{\text{must vanish point-wise at optimality}}\big\rangle dt\bigg]\label{eq:basic-am-proof1}
\end{align}
\end{small}
where we use the law of total expectation given that only $\boldsymbol{a}(\boldsymbol{X}_{t:T}, t; \bar{\boldsymbol{u}})$ depends on the path $\boldsymbol{X}_{t:T}$. Given that $\boldsymbol{v}(\boldsymbol{X}_t, t)$ is \textit{arbitrary}, the only solution where the first variation (\ref{eq:basic-am-proof1}) is zero is one where the expectation evaluates to zero for \textit{all} $\boldsymbol{v}$, which occurs if and only if for all $(\boldsymbol{x},t)$, the following is satisfied:
\begin{align}
    \boldsymbol{u}(\boldsymbol{x},t)+\sigma_t\mathbb{E}_{\boldsymbol{X}_{t:T}\sim \mathbb{P}^{\bar{u}}}\left[\boldsymbol{a}(\boldsymbol{X}_{t:T}, t; \bar{\boldsymbol{u}})|\boldsymbol{X}_t=\boldsymbol{x}\right]=0
\end{align}
Therefore, we can write the functional derivative of $\mathcal{L}_{\text{basic-AM}}(\boldsymbol{u})$ with respect to $\boldsymbol{u}$ evaluated pointwise at $(\boldsymbol{x},t)$ as:
\begin{align}
    \bluetext{\frac{\delta}{\delta \boldsymbol{u}}}\mathcal{L}_{\text{basic-AM}}(\boldsymbol{u})(\boldsymbol{x},t)=\boldsymbol{u}(\boldsymbol{x},t)+\sigma_t\mathbb{E}_{\boldsymbol{X}_{t:T}\sim \mathbb{P}^{\bar{u}}}\left[\boldsymbol{a}(\boldsymbol{X}_{t:T}, t; \bar{\boldsymbol{u}})|\boldsymbol{X}_t=\boldsymbol{x}\right]\label{eq:basic-am-functional-derivative}
\end{align}
where the \textbf{critical points} satisfy:
\begin{align}
    \boldsymbol{u}(\boldsymbol{x},t)=-\sigma_t\bluetext{\mathbb{E}_{\boldsymbol{X}_{t:T}\sim \mathbb{P}^{\bar{u}}}\left[\boldsymbol{a}(\boldsymbol{X}_{t:T}, t; \bar{\boldsymbol{u}})|\boldsymbol{X}_t=\boldsymbol{x}\right]}\overset{(\ref{eq:adjoint-state-2})}{=}-\sigma_t\bluetext{\nabla J(\boldsymbol{x},t;\boldsymbol{u})}\label{eq:basic-am-proof2}
\end{align}
To prove that any $\boldsymbol{u}$ that satisfies (\ref{eq:basic-am-proof2}) for \textit{all} $(\boldsymbol{x},t)\in \mathbb{R}^d\times[0,T]$ is the optimal control $\boldsymbol{u}^\star$, we establish the following Lemma.

\begin{proposition}[Fixed Point Solution to Optimal Control (Lemma 6 in \citet{domingo2024adjoint})]\label{prop:basic-adjoint-loss}
    Consider a control drift $\boldsymbol{u}(\boldsymbol{x},t)$ that satisfies $\boldsymbol{u}(\boldsymbol{x},t)=-\sigma_t\nabla J(\boldsymbol{x},t;\boldsymbol{u})$ for all $(\boldsymbol{x},t)\in \mathbb{R}^d \times [0,T]$. Then, we have that the function $J(\cdot, \cdot; \boldsymbol{u}): \mathbb{R}^d\times [0,T]\to \mathbb{R}$ satisfies the Hamilton-Jacobi-Bellman equation defined in (\ref{eq:hjb-pde}). Since the HJB equation has a \textbf{unique solution}, we can conclude that:
    \begin{align}
        \forall (\boldsymbol{x},t)\in \mathbb{R}^d\times [0,T], \quad J(\boldsymbol{x},t;\boldsymbol{u})=V_t(\boldsymbol{x})\implies \boldsymbol{u}(\boldsymbol{x},t)\equiv\boldsymbol{u}^\star(\boldsymbol{x},t)=-\sigma_t\nabla V_t(\boldsymbol{x})
    \end{align}
\end{proposition}

\textit{Proof.} First, we decompose $J(\boldsymbol{x},t;\boldsymbol{u})$ using (\ref{eq:bellman-principle}) which states that the optimal cost of time $t$ is equal to the incremental cost over $[t, t+\Delta t]$ and the cost of time $t+\Delta t$ to get:
\begin{small}
\begin{align}
    J(\boldsymbol{x},t;\boldsymbol{u})&=\mathbb{E}\left[\left(\bluetext{\int_t^{t+\Delta t}}+\pinktext{\int_{t+\Delta t}^T}\right)\left(\frac{1}{2}\|\boldsymbol{u}(\boldsymbol{X}_s,s)\|^2+c(\boldsymbol{X}_s,s)\right)ds+\Phi(\boldsymbol{X}_T)\bigg|\boldsymbol{X}_t=\boldsymbol{x}\right]\nonumber\\
    &=\mathbb{E}\left[\bluetext{\int_t^{t+\Delta t}\left(\frac{1}{2}\|\boldsymbol{u}(\boldsymbol{X}_s,s)\|^2+c(\boldsymbol{X}_s,s)\right)ds}\bigg|\boldsymbol{X}_t=\boldsymbol{x}\right]+\mathbb{E}\left[\pinktext{J(\boldsymbol{X}_{t+\Delta t}, t+\Delta t; \boldsymbol{u})}|\boldsymbol{X}_t=\boldsymbol{x}\right]
\end{align}
\end{small}
Subtracting $J(\boldsymbol{x},t;\boldsymbol{u})$ from both sides, dividing by $\Delta t$, and taking the continuous time limit $\Delta t\to 0$, we get: 
\begin{small}
\begin{align}
    0&=\lim_{\Delta t}\left\{\frac{\mathbb{E}\left[J(\boldsymbol{X}_{t+\Delta t}, t+\Delta t; \boldsymbol{u})|\boldsymbol{X}_t=\boldsymbol{x}\right]\bluetext{-J(\boldsymbol{x},t;\boldsymbol{u})}}{\Delta t}+\frac{\mathbb{E}\left[\int_t^{t+\Delta t}\left(\frac{1}{2}\|\boldsymbol{u}(\boldsymbol{X}_s,s)\|^2+c(\boldsymbol{X}_s,s)\right)ds\bigg|\boldsymbol{X}_t=\boldsymbol{x}\right]}{\Delta t}\right\}\nonumber\\
    0&=\pinktext{\mathcal{A}^uJ(\boldsymbol{x},t; \boldsymbol{u})}+\bluetext{\frac{1}{2}\|\boldsymbol{u}(\boldsymbol{x},t)\|^2}+c(\boldsymbol{x},t)\nonumber\\
    0&=\pinktext{\mathcal{A}^uJ(\boldsymbol{x},t; \boldsymbol{u})}+\bluetext{\frac{\sigma^2_t}{2}\|\nabla J(\boldsymbol{x},t;\boldsymbol{u})\|^2}+c(\boldsymbol{x},t)\label{eq:basic-am-proof3}
\end{align}
\end{small}
Expanding the controlled generator $\mathcal{A}^uJ(\boldsymbol{x},t; \boldsymbol{u})$ with (\ref{eq:ito-controlled-generator}), we get: 
\begin{small}
\begin{align}
    \pinktext{\mathcal{A}^uJ(\boldsymbol{x},t; \boldsymbol{u})}&=\partial_tJ(\boldsymbol{x},t;\boldsymbol{u})+\langle \nabla J(\boldsymbol{x},t;\boldsymbol{u}), \boldsymbol{f}(\boldsymbol{x},t)+\sigma_t\boldsymbol{u}(\boldsymbol{x},t)\rangle+\frac{\sigma_t^2}{2}\Delta J(\boldsymbol{x},t;\boldsymbol{u})\nonumber\\
    &=\partial_tJ(\boldsymbol{x},t;\boldsymbol{u})+\langle \nabla J(\boldsymbol{x},t;\boldsymbol{u}), \boldsymbol{f}(\boldsymbol{x},t)\rangle+\langle \nabla J(\boldsymbol{x},t;\boldsymbol{u}), \sigma_t\bluetext{\boldsymbol{u}(\boldsymbol{x},t)}\rangle +\frac{\sigma_t^2}{2}\Delta J(\boldsymbol{x},t;\boldsymbol{u})\nonumber\\
    &=\partial_tJ(\boldsymbol{x},t;\boldsymbol{u})+\langle \nabla J(\boldsymbol{x},t;\boldsymbol{u}), \boldsymbol{f}(\boldsymbol{x},t)\rangle+\langle \nabla J(\boldsymbol{x},t;\boldsymbol{u}), -\sigma_t\bluetext{\sigma_t\nabla J(\boldsymbol{x},t;\boldsymbol{u})}\rangle +\frac{\sigma_t^2}{2}\Delta J(\boldsymbol{x},t;\boldsymbol{u})\nonumber\\
    &=\partial_tJ(\boldsymbol{x},t;\boldsymbol{u})+\langle \nabla J(\boldsymbol{x},t;\boldsymbol{u}), \boldsymbol{f}(\boldsymbol{x},t)\rangle\bluetext{-\sigma^2_t\|\nabla J(\boldsymbol{x},t;\boldsymbol{u})\|^2} +\frac{\sigma_t^2}{2}\Delta J(\boldsymbol{x},t;\boldsymbol{u})
\end{align}
\end{small}
Substituting this into (\ref{eq:basic-am-proof3}) and completing the square, we get:
\begin{small}
\begin{align}
    0&=\pinktext{\mathcal{A}^uJ(\boldsymbol{x},t; \boldsymbol{u})}+\bluetext{\frac{\sigma^2_t}{2}\|\nabla J(\boldsymbol{x},t;\boldsymbol{u})\|^2}+c(\boldsymbol{x},t)\nonumber\\
    0&=\partial_tJ(\boldsymbol{x},t;\boldsymbol{u})+\langle \nabla J(\boldsymbol{x},t;\boldsymbol{u}), \boldsymbol{f}(\boldsymbol{x},t)\rangle\bluetext{-\sigma^2_t\|\nabla J(\boldsymbol{x},t;\boldsymbol{u})\|^2} +\frac{\sigma_t^2}{2}\Delta J(\boldsymbol{x},t;\boldsymbol{u})+\bluetext{\frac{\sigma^2_t}{2}\|\nabla J(\boldsymbol{x},t;\boldsymbol{u})\|^2}+c(\boldsymbol{x},t)\nonumber\\
    0&=\partial_tJ(\boldsymbol{x},t;\boldsymbol{u})+\langle \nabla J(\boldsymbol{x},t;\boldsymbol{u}), \boldsymbol{f}(\boldsymbol{x},t)\rangle\bluetext{-\frac{\sigma^2_t}{2}\|\nabla J(\boldsymbol{x},t;\boldsymbol{u})\|^2} +\frac{\sigma_t^2}{2}\Delta J(\boldsymbol{x},t;\boldsymbol{u})+c(\boldsymbol{x},t)
\end{align}
\end{small}
Rearranging terms, we recover the HJB equation from (\ref{eq:soc-hjb-eq}):
\begin{align}
    \partial_tJ(\boldsymbol{x},t;\boldsymbol{u})&=\bluetext{\underbrace{-\langle \nabla J(\boldsymbol{x},t;\boldsymbol{u}), \boldsymbol{f}(\boldsymbol{x},t)\rangle-\frac{\sigma_t^2}{2}\Delta J(\boldsymbol{x},t;\boldsymbol{u})}_{\text{uncontrolled generator}}}+\frac{\sigma^2_t}{2}\|\nabla J(\boldsymbol{x},t;\boldsymbol{u})\|^2-c(\boldsymbol{x},t)\nonumber\\
    &=\bluetext{\mathcal{A}_tJ(\boldsymbol{x},t;\boldsymbol{u})}+\frac{\sigma^2_t}{2}\|\nabla J(\boldsymbol{x},t;\boldsymbol{u})\|^2-c(\boldsymbol{x},t)
\end{align}
and since we define $J(\boldsymbol{x},T;\boldsymbol{u})=\Phi(\boldsymbol{x})$, we have shown that $J(\cdot, \cdot; \boldsymbol{u})$ satisfies the HJB for all $(\boldsymbol{x},t)\in \mathbb{R}^d\times [0,T]$ given $\boldsymbol{u}(\boldsymbol{x},t)=-\sigma_t\nabla J(\boldsymbol{x},t;\boldsymbol{u})$. By uniqueness of the solution to the HJB, we can conclude that $J(\boldsymbol{x},t;\boldsymbol{u})=V_t(\boldsymbol{x})$ and $\boldsymbol{u}(\boldsymbol{x},t)=\boldsymbol{u}^\star(\boldsymbol{x},t)$ is the optimal control.
\hfill $\square$

The basic adjoint matching objective $\mathcal{L}_{\text{basic-AM}}$ provides a theoretical foundation for the adjoint matching method but remains computationally inefficient as it requires differentiation through the cost functional, which depends on the full trajectory. Since we have shown that $\boldsymbol{u}^\star(\boldsymbol{x},t)$ is the \textit{unique minimizer} of (\ref{eq:basic-am-loss}), it can be written as the conditional expectation of the regression target:
\begin{align}
    \boldsymbol{u}^\star(\boldsymbol{x},t)=\mathbb{E}_{\boldsymbol{X}_{t:T}\sim \mathbb{P}^{u^\star}}\left[-\sigma_t\boldsymbol{a}(\boldsymbol{X}_{t:T},t;\boldsymbol{u}^\star)|\boldsymbol{X}_t=\boldsymbol{x}\right]
\end{align}
At optimality, the forward control and adjoint state should be balanced everywhere, so we can multiply both sides of the equation by the Jacobian $\nabla \boldsymbol{u}(\boldsymbol{x},t)\in \mathbb{R}^{d\times d}$ to get:
\begin{align}
    \boldsymbol{u}^\star(\boldsymbol{x},t)^\top \bluetext{\nabla \boldsymbol{u}^\star(\boldsymbol{x},t)}=\mathbb{E}_{\boldsymbol{X}_{t:T}\sim \mathbb{P}^{u^\star}}\left[-\sigma_t\boldsymbol{a}(\boldsymbol{X}_{t:T},t;\boldsymbol{u}^\star)^\top\bluetext{ \nabla \boldsymbol{u}^\star(\boldsymbol{x},t)}|\boldsymbol{X}_t=\boldsymbol{x}\right]
\end{align}
Since the left hand side depends only on $\boldsymbol{x}$, we can rearrange to get:  
\begin{align}
    \mathbb{E}_{\boldsymbol{X}_{t:T}}[\boldsymbol{u}^\star(\boldsymbol{x},t)^\top \bluetext{\nabla \boldsymbol{u}^\star(\boldsymbol{x},t)}+\sigma_t\boldsymbol{a}(\boldsymbol{X}_{t:T},t;\boldsymbol{u}^\star)^\top\bluetext{ \nabla \boldsymbol{u}^\star(\boldsymbol{x},t)}|\boldsymbol{X}_t=\boldsymbol{x}]=0
\end{align}
which shows that \textit{at optimality}, both terms that depend on the control $\boldsymbol{u}(\boldsymbol{x},t)$ from the adjoint derivative (\ref{eq:adjoint-derivative}) evaluate to zero. Leveraging this, \citet{domingo2024adjoint} introduces the \boldtext{lean adjoint state}, which drops the $\boldsymbol{u}$-dependent terms to obtain a computationally more efficient objective.

\begin{definition}[Lean Adjoint State \citep{domingo2024adjoint}]\label{def:lean-adjoint-state}
    The \textbf{lean adjoint state} $\tilde{\boldsymbol{a}}:C([t,T], \mathbb{R}^d)\times[0,T]\to \mathbb{R}^d$ is defined by the following differential equation which can be solved backward in time, given the terminal condition $\tilde{\boldsymbol{a}}(\boldsymbol{X},T)=\nabla \Phi(\boldsymbol{X}_T)$ as: 
    \begin{align}
        \frac{d}{dt}\tilde{\boldsymbol{a}}(\boldsymbol{X}_{t:T},t)=-\big[\tilde{\boldsymbol{a}}(\boldsymbol{X}_{t:T}, t)^\top\nabla \boldsymbol{f}(\boldsymbol{X}_t,t)+\nabla c(\boldsymbol{X}_t,t)\big], \quad \tilde{\boldsymbol{a}}(\boldsymbol{X},T)=\nabla \Phi(\boldsymbol{X}_T)\label{eq:lean-adjoint-bwd-derivative}
    \end{align}
    Unlike the \textbf{adjoint state} defined in Definition \ref{def:adjoint-state}, the lean adjoint state $\tilde{\boldsymbol{a}}$ does not depend on the control $\boldsymbol{u}$ and does not require computing the Jacobian $\nabla \boldsymbol{u}(\boldsymbol{x},t)$. 
    \begin{align}
        \tilde{\boldsymbol{a}}(\boldsymbol{X}_{t:T}, t)=\int_t^T\left(\nabla \boldsymbol{f}(\boldsymbol{X}_s,s)^\top\tilde{\boldsymbol{a}}(\boldsymbol{X}_{s:T},s)+\nabla \boldsymbol{f}(\boldsymbol{X}_s,s)\right)ds+\nabla c(\boldsymbol{X}_T)\label{eq:lean-adjoint-state}
    \end{align}
\end{definition}
Using this lean adjoint state, we can construct an objective that directly matches the optimal control without requiring explicit computation of the value function or its gradients.

\begin{proposition}[Lean Adjoint Matching Yields the Optimal Control (Proposition 7 in \citep{domingo2024adjoint})]\label{prop:lean-adjoint-loss}
    Consider the \textbf{lean adjoint matching objective} defined with the lean adjoint state $\tilde{\boldsymbol{a}}:C([t,T], \mathbb{R}^d)\times[0,T]\to \mathbb{R}^d$ as:
    \begin{align}
        &\mathcal{L}_{\text{AM}}(\boldsymbol{u}):=\mathbb{E}_{\boldsymbol{X}_{0:T}\sim \mathbb{P}^{\bar{u}}}\left[\frac{1}{2}\int_0^T\left\|\boldsymbol{u}(\boldsymbol{X}_t,t)+\sigma_t\tilde{\boldsymbol{a}}(\boldsymbol{X}_{t:T},t)\right\|^2dt\right]\tag{Lean AM Objective}\label{eq:lean-adjoint-loss}\\
        &\bar{\boldsymbol{u}}:=\texttt{stopgrad}(\boldsymbol{u})\nonumber
    \end{align}
    where $\bar{\boldsymbol{u}}:=\texttt{stopgrad}(\boldsymbol{u})$ is the non-gradient tracking control drift. Then, $\mathcal{L}_{\text{AM}}(\boldsymbol{u})$ has a unique minimizer which is exactly the optimal control $\boldsymbol{u}^\star$.
\end{proposition}

\textit{Proof.} To prove this, we first establish the form of \textit{some} critical point $\hat{\boldsymbol{u}}$ of $\mathcal{L}_{\text{AM}}(\boldsymbol{u})$ and show that it is also a critical point of the \textbf{basic adjoint matching} loss $\mathcal{L}_{\text{basic-AM}}$. Then, we can apply the result from Proposition \ref{prop:basic-adjoint-loss} to conclude that $\hat{\boldsymbol{u}}$ is unique and is equal to the optimal control $\hat{\boldsymbol{u}}=\boldsymbol{u}^\star$.

\textbf{Step 1: Derive the Critical Point.} 
Since the form of the (\ref{eq:lean-adjoint-loss}) depends on a random variable $\boldsymbol{X}_{0:T}$, to take the functional derivative, we need to evaluate it for some deterministic state $\boldsymbol{X}_t=\boldsymbol{x}$. Since $\tilde{\boldsymbol{a}}(\boldsymbol{X}_{t:T},t)$ is the only term in $\mathcal{L}_{\text{AM}}$ that contains randomness in $\boldsymbol{X}_{t:T}\sim \mathbb{P}^{\bar{u}}$ after fixing $\boldsymbol{X}_t=\boldsymbol{x}$, we add and subtract the conditional expectation $\mathbb{E}[\tilde{\boldsymbol{a}}(\boldsymbol{X}_{t:T},t)|\boldsymbol{X}_t]$ to the expression inside the square in $\mathcal{L}_{\text{AM}}$ to get:
\begin{small}
\begin{align}
    \boldsymbol{u}(\boldsymbol{X}_t,t)+\sigma_t\tilde{\boldsymbol{a}}(\boldsymbol{X}_{t:T},t)=\boldsymbol{u}(\boldsymbol{X}_t,t)+\bluetext{\sigma_t\mathbb{E}[\tilde{\boldsymbol{a}}(\boldsymbol{X}_{t:T},t)|\boldsymbol{X}_t]}+\underbrace{\sigma_t\tilde{\boldsymbol{a}}(\boldsymbol{X}_{t:T},t)-\bluetext{\sigma_t\mathbb{E}[\tilde{\boldsymbol{a}}(\boldsymbol{X}_{t:T},t)|\boldsymbol{X}_t]}}_{\text{vanishes when evaluated at a point }\boldsymbol{X}_t=\boldsymbol{x}}
\end{align}
\end{small}
Then, substituting this expression back into the (\ref{eq:lean-adjoint-loss}), we have:
\begin{small}
\begin{align}
    &\mathcal{L}_{\text{AM}}(\boldsymbol{u})=\mathbb{E}_{\boldsymbol{X}_{0:T}\sim \mathbb{P}^{\bar{u}}}\left[\frac{1}{2}\int_0^T\left\|\boldsymbol{u}(\boldsymbol{X}_t,t)+\bluetext{\sigma_t\mathbb{E}[\tilde{\boldsymbol{a}}(\boldsymbol{X}_{t:T},t)|\boldsymbol{X}_t]}+\sigma_t\tilde{\boldsymbol{a}}(\boldsymbol{X}_{t:T},t)-\bluetext{\sigma_t\mathbb{E}[\tilde{\boldsymbol{a}}(\boldsymbol{X}_{t:T},t)|\boldsymbol{X}_t]}\right\|^2dt\right]\nonumber\\
    &=\mathbb{E}_{\boldsymbol{X}_{0:T}\sim \mathbb{P}^{\bar{u}}}\left[\frac{1}{2}\int_0^T\left\|\boldsymbol{u}(\boldsymbol{X}_t,t)+\bluetext{\sigma_t\mathbb{E}[\tilde{\boldsymbol{a}}(\boldsymbol{X}_{t:T},t)|\boldsymbol{X}_t]}\right\|^2\right]dt+\underbrace{\mathbb{E}_{\boldsymbol{X}_{0:T}\sim \mathbb{P}^{\bar{u}}}\left[\frac{1}{2}\int_0^T\left\|\sigma_t(\tilde{\boldsymbol{a}}(\boldsymbol{X}_{t:T})-\bluetext{\mathbb{E}[\tilde{\boldsymbol{a}}(\boldsymbol{X}_{t:T},t)|\boldsymbol{X}_t])}\right\|^2\right]dt}_{\text{not dependent on }\boldsymbol{u}}\nonumber
\end{align}
\end{small}
Then, computing the functional derivative $\frac{\delta}{\delta\boldsymbol{u}}\mathcal{L}_{\text{AM}}(\boldsymbol{u})(\boldsymbol{x},t)$ evaluated at $\boldsymbol{X}_t=\boldsymbol{x}$, we get: 
\begin{align}
    \frac{\delta}{\delta \boldsymbol{u}}\mathcal{L}_{\text{AM}}(\boldsymbol{u})(\boldsymbol{x},t)=\boldsymbol{u}(\boldsymbol{x},t)+\bluetext{\sigma_t\mathbb{E}[\tilde{\boldsymbol{a}}(\boldsymbol{X}_{t:T},t)|\boldsymbol{X}_t=\boldsymbol{x}]}\label{eq:lean-am-functional-derivative}
\end{align}
Since the first variation of $\frac{\delta}{\delta \boldsymbol{u}}\mathcal{L}_{\text{AM}}(\hat{\boldsymbol{u}})$ of critical points $\hat{\boldsymbol{u}}$ is zero, we get that the critical points of $\mathcal{L}_{\text{AM}}(\hat{\boldsymbol{u}})$ satisfy:
\begin{align}
    \forall \boldsymbol{x}\in \mathbb{R}^d, \quad \frac{\delta}{\delta \boldsymbol{u}}\mathcal{L}_{\text{AM}}(\hat{\boldsymbol{u}})(\boldsymbol{x},t)=0\implies \hat{\boldsymbol{u}}(\boldsymbol{x},t)=-\bluetext{\sigma_t\mathbb{E}[\tilde{\boldsymbol{a}}(\boldsymbol{X}_{t:T},t)|\boldsymbol{X}_t=\boldsymbol{x}]}\label{eq:lean-adjoint-proof1}
\end{align}
which we will show is also a critical point for the \textbf{basic adjoint matching} objective $\mathcal{L}_{\text{basic-AM}}$ in (\ref{eq:basic-am-loss}).

\textbf{Step 2: Matching Critical Point to Basic Adjoint Matching Objective.}
Using the same observation for obtaining the \textbf{lean adjoint state}, we note that at optimalty, the forward control and adjoint state can be balanced which allows us to multiply both sides of (\ref{eq:lean-adjoint-proof1}) with the Jacobian $\nabla \hat{\boldsymbol{u}}(\boldsymbol{x},t)$ to get: 
\begin{align}
    &\bluetext{\nabla \hat{\boldsymbol{u}}(\boldsymbol{X}_t,t)^\top\hat{\boldsymbol{u}}(\boldsymbol{X}_t,t)}=\pinktext{-\sigma_t\nabla \hat{\boldsymbol{u}}(\boldsymbol{X}_t,t)^\top}\mathbb{E}[\tilde{\boldsymbol{a}}(\boldsymbol{X}_{t:T},t)|\boldsymbol{X}_t]\nonumber\\
    &\implies \mathbb{E}\bigg[\int_t^T\left(\pinktext{\sigma_s\nabla \hat{\boldsymbol{u}}(\boldsymbol{X}_s,s)^{\top}\mathbb{E}[\tilde{\boldsymbol{a}}(\boldsymbol{X}_{s:T},s)|\boldsymbol{X}_s]}+\bluetext{\nabla \left(\frac{1}{2}\|\hat{\boldsymbol{u}}(\boldsymbol{X}_s,s)\|^2\right)}\right)ds\bigg|\boldsymbol{X}_t\bigg]=0\label{eq:zero-expectation-cond}
\end{align}
Then, adding this zero-expectation balancing condition to the conditional expectation of the lean adjoint state $\mathbb{E}[\tilde{\boldsymbol{a}}(\boldsymbol{X}_{t:T},t)|\boldsymbol{X}_t]$ using the definition in (\ref{eq:lean-adjoint-state}), we get: 
\begin{small}
\begin{align}
    \mathbb{E}[\tilde{\boldsymbol{a}}(\boldsymbol{X}_{t:T}, t)|\boldsymbol{X}_t]&=\mathbb{E}\left[\int_t^T\left(\nabla \boldsymbol{f}(\boldsymbol{X}_s,s)^\top\mathbb{E}[\tilde{\boldsymbol{a}}(\boldsymbol{X}_{s:T},s)|\boldsymbol{X}_s]+\nabla \boldsymbol{f}(\boldsymbol{X}_s,s)\right)ds+\nabla c(\boldsymbol{X}_T)\bigg|\boldsymbol{X}_t\right]\nonumber\\
    &\quad \quad  +\bluetext{\underbrace{\mathbb{E}\bigg[\int_t^T\left(\sigma_s\nabla \hat{\boldsymbol{u}}(\boldsymbol{X}_s,s)^{\top}\mathbb{E}[\tilde{\boldsymbol{a}}(\boldsymbol{X}_{s:T},s)|\boldsymbol{X}_s]+\nabla \left(\frac{1}{2}\|\hat{\boldsymbol{u}}(\boldsymbol{X}_s,s)\|^2\right)\right)ds\bigg|\boldsymbol{X}_t\bigg]}_{\text{zero-expectation balancing condition (\ref{eq:zero-expectation-cond})}}}\nonumber\\
    &=\mathbb{E}\bigg[\int_t^T\left(\nabla (\boldsymbol{f}(\boldsymbol{X}_s,s)+\bluetext{\sigma_s\hat{\boldsymbol{u}}(\boldsymbol{X}_s,s)})^\top\mathbb{E}[\tilde{\boldsymbol{a}}(\boldsymbol{X}_{s:T},s)|\boldsymbol{X}_s]+\nabla \boldsymbol{f}(\boldsymbol{X}_s,s)\right)ds\nonumber\\
    &\quad \quad +\nabla \left(c(\boldsymbol{X}_T)+\bluetext{\frac{1}{2}\|\hat{\boldsymbol{u}}(\boldsymbol{X}_s,s)\|^2}\right)\bigg|\boldsymbol{X}_t\bigg]
\end{align}
\end{small}
Since the \textbf{adjoint state} $\boldsymbol{a}(\boldsymbol{X}_{t:T},t;\boldsymbol{u})$ also solves an equivalent integral equation from (\ref{eq:integral-adjoint-state}) with \textit{arbitrary} $\boldsymbol{v}$ defined as:
\begin{small}
\begin{align}
    \mathbb{E}[\boldsymbol{a}(\boldsymbol{X}_{t:T},t;\boldsymbol{u})|\boldsymbol{X}_t]&=\mathbb{E}\bigg[\int_t^T\left(\nabla (\boldsymbol{f}(\boldsymbol{X}_s,s)+\bluetext{\sigma_s\boldsymbol{v}(\boldsymbol{X}_s,s)})^\top\mathbb{E}[\tilde{\boldsymbol{a}}(\boldsymbol{X}_{s:T},s)|\boldsymbol{X}_s]+\nabla \boldsymbol{f}(\boldsymbol{X}_s,s)\right)ds\nonumber\\
    &\quad \quad +\nabla \bigg(c(\boldsymbol{X}_T)+\bluetext{\frac{1}{2}\|\boldsymbol{v}(\boldsymbol{X}_s,s)\|^2}\bigg)\bigg|\boldsymbol{X}_t\bigg]\label{eq:lean-adjoint-proof2}
\end{align}
\end{small}
Setting $\boldsymbol{v}:=\hat{\boldsymbol{u}}$ in (\ref{eq:lean-adjoint-proof2}) yields the \textit{unique}\footnote{The uniqueness of the integral equation follows the expressing the difference between two arbitrary solutions as an integral inequality and observing that the largest squared distance between the two solutions is bounded by the integral of itself. By Grönwall’s inequality, this means their squared difference must equal zero, and thus the solution is unique. For full proof, see Proposition 8 in \citep{domingo2024adjoint}.} solution at $\mathbb{E}[\boldsymbol{a}(\boldsymbol{X}_{t:T},t;\hat{\boldsymbol{u}})|\boldsymbol{X}_t]=\mathbb{E}[\tilde{\boldsymbol{a}}(\boldsymbol{X}_{t:T},t)|\boldsymbol{X}_t]$ for all $t\in [0,T]$. Substituting this equality into the functional derivative of the basic adjoint matching loss $\mathcal{L}_{\text{basic-AM}}(\boldsymbol{u})$ from (\ref{eq:basic-am-functional-derivative}), we get: 
\begin{align}
    \bluetext{\frac{\delta}{\delta \boldsymbol{u}}}\mathcal{L}_{\text{basic-AM}}(\hat{\boldsymbol{u}})(\boldsymbol{x},t)&=\hat{\boldsymbol{u}}(\boldsymbol{x},t)+\sigma_t\bluetext{\mathbb{E}[\boldsymbol{a}(\boldsymbol{X}_{t:T},t; \boldsymbol{u})|\boldsymbol{X}_t]}=\hat{\boldsymbol{u}}(\boldsymbol{x},t)+\sigma_t\bluetext{\mathbb{E}[\tilde{\boldsymbol{a}}(\boldsymbol{X}_{t:T},t)|\boldsymbol{X}_t]}\nonumber\\
    &\overset{(\ref{eq:lean-adjoint-proof1})}{=}\pinktext{-\sigma_t\mathbb{E}[\tilde{\boldsymbol{a}}(\boldsymbol{X}_{t:T},t)|\boldsymbol{X}_t]}+\sigma_t\mathbb{E}[\tilde{\boldsymbol{a}}(\boldsymbol{X}_{t:T},t)|\boldsymbol{X}_t]=0
\end{align}
which implies that all critical points $\hat{\boldsymbol{u}}$ of $\mathcal{L}_{\text{AM}}$ are critical points of $\mathcal{L}_{\text{basic-AM}}$, and by Proposition \ref{prop:basic-adjoint-loss}, we conclude that $\hat{\boldsymbol{u}}$ is \textit{unique} and \textbf{equal to the optimal control} $\hat{\boldsymbol{u}}=\boldsymbol{u}^\star$.\hfill $\square$

Given this \textbf{general form} of the adjoint matching objective, we can further simplify it for the case where the reference process is pure Brownian motion ($\boldsymbol{f}:=0$) and the running cost is zero ($c:=0$), which results in the backward time evolution $\frac{d}{dt}\tilde{\boldsymbol{a}}$ defined in (\ref{eq:lean-adjoint-bwd-derivative}) to \textit{vanish}. Then, the \textbf{lean adjoint state} for all $t\in [0,T]$ reduces to the gradient of the terminal cost with respect to the current state given by $\tilde{\boldsymbol{a}}(\boldsymbol{X}_{t:T}, t)=\nabla_{\boldsymbol{x}_T}\Phi(\boldsymbol{X}_T)$. Therefore, the \boldtext{simplified adjoint matching} objective becomes:
\begin{align}
    &\mathcal{L}_{\text{simple-AM}}(\boldsymbol{u}):=\mathbb{E}_{\boldsymbol{X}_{0:T}\sim \mathbb{P}^{\bar{u}}}\left[\frac{1}{2}\int_0^T\|\boldsymbol{u}(\boldsymbol{X}_t,t)+\sigma_t\nabla_{\boldsymbol{x}_T}\Phi(\boldsymbol{X}_T)\|^2dt\right]\tag{Simplified AM Objective}\label{eq:simple-am-loss}\\
    &\bar{\boldsymbol{u}}=\texttt{stopgrad}(\boldsymbol{u})\nonumber
\end{align}
To further simplify the objective, \citet{havens2025adjoint} introduces the \boldtext{reciprocal adjoint matching} objective, where rather than taking the expectation over full trajectories $\boldsymbol{X}_{0:T}\sim \mathbb{P}^{\bar{u}}$ by repeatedly simulating the controlled SDE, we can leverage the \textbf{key property} that given $\boldsymbol{f}:=0$, $c:=0$, and $\boldsymbol{X}_0=0$, the (\ref{eq:simple-am-loss}) depends only on $(\boldsymbol{X}_t,\boldsymbol{X}_T)$ and the joint distribution of $(\boldsymbol{X}_t,\boldsymbol{X}_T)$ under the optimal Schrödinger bridge measure $\mathbb{P}^\star$ be factorized as:
\begin{small}
\begin{align}
    \mathbb{P}^\star(\boldsymbol{X}_t, \boldsymbol{X}_T)=\pi_T(\boldsymbol{X}_T)\mathbb{P}_{t|T}^\star(\boldsymbol{X}_t|\boldsymbol{X}_T)=\pi_T(\boldsymbol{X}_T)\mathbb{Q}_{t|T}(\boldsymbol{X}_t|\boldsymbol{X}_T)
\end{align}
\end{small}
since $\mathbb{P}^\star$ is in the reciprocal class $\mathcal{R}(\mathbb{Q})$, which shares the conditional bridge  $\mathbb{P}_{t|T}^\star(\boldsymbol{X}_t|\boldsymbol{X}_T)=\mathbb{Q}_{t|T}(\boldsymbol{X}_t|\boldsymbol{X}_T)$ to the reference process. This gives us the objective:
\begin{align}
    &\mathcal{L}_{\text{RAM}}(\boldsymbol{u}):=\mathbb{E}_{\boldsymbol{X}_t\sim \mathbb{Q}_{t|T}^{\bar{u}}, \boldsymbol{X}_T\sim p^{\bar{u}}_T}\left[\frac{1}{2}\int_0^T\|\boldsymbol{u}(\boldsymbol{X}_t,t)+\sigma_t\nabla_{\boldsymbol{x}_T}\Phi(\boldsymbol{X}_T)\|^2dt\right]\tag{Reciprocal AM Loss}\label{eq:reciprocal-am-loss}\\
    &\bar{\boldsymbol{u}}=\texttt{stopgrad}(\boldsymbol{u})\nonumber
\end{align}
Crucially, the samples $\boldsymbol{X}_t\sim \mathbb{Q}_{t|T}$ are \textbf{independent} conditioned on a terminal state $\boldsymbol{X}_T\sim p^{\bar{u}}_T$, enabling training on arbitrarily many intermediate samples given a single terminal state $\boldsymbol{X}_T$. To minimize this objective, \citet{havens2025adjoint} proposes an efficient \textbf{iterative two-step algorithm} called \boldtext{adjoint sampling}.

\purple[Adjoint Sampling Algorithm]{\label{alg:adjoint-sampling}
The \textbf{adjoint sampling} algorithm \citep{havens2025adjoint} is an iterative two step procedure:
\begin{enumerate}
    \item [(i)] First, simulate the forward-time controlled SDE using the non-gradient tracking control $\bar{\boldsymbol{u}}$ to obtain a fixed set of samples $\boldsymbol{X}_T\sim p^{\bar{u}}_T$ and evaluate the gradient of their terminal cost $\nabla_{\boldsymbol{x}_T}\Phi(\boldsymbol{X}_T)$. Then, store the set of samples in a replay buffer $\mathcal{B}=\{\boldsymbol{X}^i_T,\nabla_{\boldsymbol{x}^i_T}g(\boldsymbol{X}^i_T)\}_{i=1}^B$.
    \item [(ii)] Optimize the (\ref{eq:reciprocal-am-loss}) by repeatedly sampling intermediate states $\boldsymbol{X}_t$ from the posterior $\boldsymbol{X}_t\sim \mathbb{Q}_{t|T}$ conditioned on samples from the replay buffer $\mathcal{B}$. 
\end{enumerate}
This process repeats until the optimal $\boldsymbol{u}^\star$ is reached.
}

Intuitively, optimizing an arbitrary control $\boldsymbol{v}$ with (\ref{eq:reciprocal-am-loss}) can be considered a \textbf{reciprocal projection} (Proposition \ref{prop:reciprocal-proj}), where we constrain the endpoint as $p^{\bar{u}}_T(\boldsymbol{X}_T)$ \textit{and} a \textbf{Markovian projection} (Proposition \ref{prop:markov-proj}) by projecting on the Markov bridge measure $\mathbb{Q}_{t|T}(\boldsymbol{X}_t|\boldsymbol{X}_T)$, where $\mathbb{Q}$ is the Markov reference Brownian motion. 

\begin{proposition}[Convergence of Adjoint Sampling (Theorem 3.2 in \citet{havens2025adjoint})]\label{prop:convergence-reciprocal-am}
    Consider optimizing the (\ref{eq:reciprocal-am-loss}) $\mathcal{L}_{\text{RAM}}$ given the current control $\bar{\boldsymbol{u}}$ as \textit{projecting} an arbitrary control $\boldsymbol{v}$ onto the bridge $\mathbb{P}^{\bar{u}}(\boldsymbol{X}_{0:T})=p^{\bar{u}}_T(\boldsymbol{X}_T)\mathbb{Q}(\boldsymbol{X}_{0:T})$ where $p_T^{\bar{u}}$ is the target distribution: 
    \begin{align}
        \text{proj}(\boldsymbol{u})=\underset{\boldsymbol{v}}{\arg\min}\text{KL}\left(\mathbb{P}^v(\boldsymbol{X}_{0:T})\|p^{\bar{u}}_T(\boldsymbol{X}_T)\mathbb{Q}(\boldsymbol{X}_{0:T})\right)
    \end{align}
    Then, at each iteration, we obtain the update:
    \begin{align}
        \boldsymbol{u}^{n+1}=\text{proj}(\boldsymbol{u}^n)-\frac{\delta}{\delta \boldsymbol{u}}\mathcal{L}_{\text{AM}}(\text{proj}(\boldsymbol{u}^n) )\label{eq:reciprocal-am-proj}
    \end{align}
    where $\frac{\delta}{\delta \boldsymbol{u}}\mathcal{L}_{\text{AM}}$ is the functional derivative of the adjoint matching objective $\mathcal{L}_{\text{AM}}$ defined in (\ref{eq:lean-am-functional-derivative}). The \textbf{unique} fixed point $\boldsymbol{u}=\text{proj}(\boldsymbol{u})=\boldsymbol{u}--\frac{\delta}{\delta \boldsymbol{u}}\mathcal{L}_{\text{AM}}(\boldsymbol{u})$ is exactly the optimal control $\boldsymbol{u}^\star=\text{proj}(\boldsymbol{u}^\star)$.
\end{proposition}
\textit{Proof.}
For this proof, we leverage the form of the critical point in (\ref{eq:lean-adjoint-proof1}) and the functional derivative in (\ref{eq:lean-am-functional-derivative}). 

\textbf{Step 1: Unifying the Reciprocal Adjoint and Lean Adjoint Matching Objectives.} 
First, we define a more general form of the reciprocal adjoint matching objective, where the matching target is an arbitrary vector field $\boldsymbol{v}$, and the current control is $\boldsymbol{u}$. 
\begin{align}
    \mathcal{L}(\boldsymbol{u}; \boldsymbol{v})&=\mathbb{E}_{\boldsymbol{X}_{0:T}\sim \mathbb{P}^v}\left[\int_0^T\frac{1}{2}\|\boldsymbol{u}(\boldsymbol{X}_t,t)+\sigma_t\nabla_{\boldsymbol{x}_T}\Phi(\boldsymbol{X}_T)\|^2dt\right]\\
    \mathcal{L}_{\text{RAM}}(\boldsymbol{u})&:=\mathcal{L}(\boldsymbol{u}; \bluetext{\text{proj}(\bar{\boldsymbol{u}})}), \quad \mathcal{L}_{\text{simple-AM}}(\boldsymbol{u}):=\mathcal{L}(\boldsymbol{u}; \bluetext{\bar{\boldsymbol{u}}})\label{eq:ram-am-equal}
\end{align}
which are equivalent definitions of the (\ref{eq:reciprocal-am-loss}) and the (\ref{eq:simple-am-loss}) given $\bar{\boldsymbol{u}}=\texttt{stopgrad}(\boldsymbol{u})$.

\textbf{Step 2: Derive Expression for Adjoint Sampling Iteration.}
Given the relationship between the lean adjoint and reciprocal adjoint matching losses defined in (\ref{eq:ram-am-equal}), we can apply (\ref{eq:lean-adjoint-proof1}) to write an optimal iteration of minimizing $\mathcal{L}(\boldsymbol{u}; \text{proj}(\boldsymbol{u}^n))$ as satisfying:
\begin{align}
    \boldsymbol{u}^{n+1}(\boldsymbol{x},t)=-\sigma_t\mathbb{E}_{\mathbb{P}^{\text{proj}(\boldsymbol{u}^n)}}\left[\nabla_{\boldsymbol{x}_T}\Phi(\boldsymbol{X}_T)|\boldsymbol{X}=\boldsymbol{x}\right]
\end{align}
Since the $\mathcal{L}_{\text{RAM}}$ is just $\mathcal{L}_{\text{AM}}$ with the target set to $\text{proj}(\boldsymbol{u}^n)$, we also apply (\ref{eq:lean-am-functional-derivative}) to write the functional derivative of $\mathcal{L}_{\text{AM}}$ evaluated at $\text{proj}(\boldsymbol{u}^n)$ as:
\begin{align}
    \frac{\delta}{\delta\boldsymbol{u}}\mathcal{L}_{\text{AM}}(\text{proj}(\boldsymbol{u}^n))&=\text{proj}(\boldsymbol{u}^n)+\sigma_t\mathbb{E}_{\mathbb{P}^{\text{proj}(\boldsymbol{u}^n)}}\left[\nabla_{\boldsymbol{x}_T}\Phi(\boldsymbol{X}_T)|\boldsymbol{X}=\boldsymbol{x}\right]\nonumber\\
    \bluetext{\underbrace{-\sigma_t\mathbb{E}_{\mathbb{P}^{\text{proj}(\boldsymbol{u}^n)}}\left[\nabla_{\boldsymbol{x}_T}\Phi(\boldsymbol{X}_T)|\boldsymbol{X}=\boldsymbol{x}\right]}_{=:\boldsymbol{u}^{n+1}}}&=\text{proj}(\boldsymbol{u}^n)-\frac{\delta}{\delta\boldsymbol{u}}\mathcal{L}_{\text{AM}}(\text{proj}(\boldsymbol{u}^n))\nonumber\\
    \bluetext{\boldsymbol{u}^{n+1}(\boldsymbol{x},t)}&=\text{proj}(\boldsymbol{u}^n)-\frac{\delta}{\delta\boldsymbol{u}}\mathcal{L}_{\text{AM}}(\text{proj}(\boldsymbol{u}^n))
\end{align}
which concludes the proof of (\ref{eq:reciprocal-am-proj}) in the Proposition. 

\textbf{Step 3: Fixed Point of Adjoint Sampling Iteration.} 
We will now show that $\boldsymbol{u}$ is a fixed point of adjoint sampling such that $\boldsymbol{u}=\text{proj}(\boldsymbol{u})$ if and only if $\boldsymbol{u}$ is a critical point of the (\ref{eq:lean-adjoint-loss}), which implies that $\boldsymbol{u}=\boldsymbol{u}^\star$ by Proposition \ref{prop:lean-adjoint-loss}. 

First, we show $\bluetext{\boldsymbol{u}=\text{proj}(\boldsymbol{u})\implies \boldsymbol{u}=\boldsymbol{u}^\star}$. Suppose $\boldsymbol{u}=\text{proj}(\boldsymbol{u})$. By (\ref{eq:reciprocal-am-proj}), we have $\frac{\delta}{\delta \boldsymbol{u}}\mathcal{L}_{\text{AM}}(\text{proj}(\boldsymbol{u}))=\frac{\delta}{\delta \boldsymbol{u}}\mathcal{L}_{\text{AM}}(\boldsymbol{u})=0$. This means that $\boldsymbol{u}$ is a critical point of $\mathcal{L}_{\text{AM}}$ and by Proposition \ref{prop:lean-adjoint-loss}, we have that $\boldsymbol{u}$ is unique and equal to the optimal control $\boldsymbol{u}=\boldsymbol{u}^\star$.

Next, we show $\bluetext{\boldsymbol{u}=\boldsymbol{u}^\star\implies\boldsymbol{u}=\text{proj}(\boldsymbol{u})}$. Suppose $\boldsymbol{u}=\boldsymbol{u}^\star$. Then, by Proposition \ref{prop:lean-adjoint-loss}, we know $\boldsymbol{u}=\boldsymbol{u}^\star$. Since by definition, $\boldsymbol{u}^\star$ generates the optimal target of the projection $\mathbb{P}^\star(\boldsymbol{X}_{0:T})=\pi_T(\boldsymbol{X}_T)\mathbb{Q}(\boldsymbol{X}_{0:T})$, projecting onto $\mathbb{P}^\star$ would yield itself, so $\boldsymbol{u}^\star=\text{proj}(\boldsymbol{u}^\star)$. \hfill $\square$

Adjoint sampling (Box \ref{alg:adjoint-sampling}) provides a theoretically-grounded and computationally efficient method of obtaining the optimal control $\boldsymbol{u}^\star$ when the posterior under the reference dynamics can be easily sampled as $\boldsymbol{X}_t\sim \mathbb{Q}_{t|T}(\cdot|\boldsymbol{X}_T)$, such as the linear Brownian motion case where $\boldsymbol{f}:=0$ and the initial distribution is a dirac delta at zero $\pi_0:=\delta_0$. However, its restriction to the dirac delta prior prevents more general settings with \textit{informative priors}, such as Gaussians or task-specific priors. 

Additionally, adjoint matching requires the reference process to be \textbf{memoryless}, such that the joint distribution of $(\boldsymbol{X}_0, \boldsymbol{X}_T)$ can be factorized as $\mathbb{Q}(\boldsymbol{X}_0, \boldsymbol{X}_T)=q_0(\boldsymbol{X}_0)q_T(\boldsymbol{X}_T)$. This is to prevent the \textbf{initial value function bias} described in Box \ref{box:memoryless}, where the optimal joint distribution $\mathbb{P}^\star(\boldsymbol{X}_0, \boldsymbol{X}_T)$ derived in Proposition \ref{prop:soc-optimal-measure} is dependent on an intractable initial value function $V_0(\boldsymbol{X}_0)$: 
\begin{align}
    \mathbb{P}^\star(\boldsymbol{X}_0, \boldsymbol{X}_T)=\mathbb{Q}(\boldsymbol{X}_0, \boldsymbol{X}_T)e^{-\Phi(\boldsymbol{X}_T)+V_0(\boldsymbol{X}_0)}
\end{align}

\begin{figure}
    \centering
    \includegraphics[width=\linewidth]{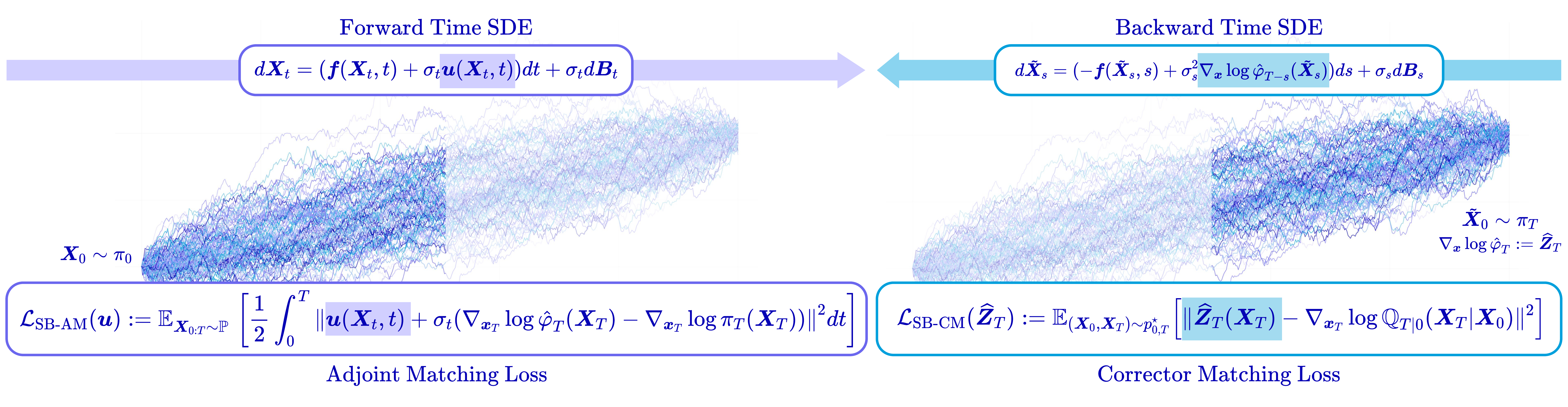}
    \caption{\textbf{Schrödinger Bridge with Adjoint Matching.} Adjoint matching learns the Schrödinger bridge by alternating between optimizing the forward half-bridge through the adjoint matching loss for $\boldsymbol{u}(\boldsymbol{x},t)$ and optimizing the backward half-bridge through the corrector matching loss for $\widehat{\boldsymbol{Z}}_T(\boldsymbol{X}_T)$.}
    \label{fig:adjoint}
\end{figure}

Although leveraging a memoryless reference drift guarantees sampling paths that generate the true target distribution $\pi_T$, it prevents the use of more informative prior distributions for sampling complex target distributions and crucially excludes all tasks where the goal is to map between distributions rather than simply sampling from a target distribution. This motivates using the (\ref{eq:sb-soc-objective}) introduced in Section \ref{subsec:sb-soc}, which leverages the SB potentials $(\varphi, \hat{\varphi})$ defined by: 
\begin{small}
\begin{align}
    &\boldsymbol{u}^\star(\boldsymbol{x},t)=-\sigma_t\nabla \log\varphi_t(\boldsymbol{x}), \quad p_t^\star(\boldsymbol{x})=\varphi_t(\boldsymbol{x})\hat{\varphi}(\boldsymbol{x})\tag{SB Optimality}\label{eq:sb-system-adjoint}\\
    &\text{where} \quad\begin{cases}
        \varphi_t(\boldsymbol{x})=\int_{\mathbb{R}^d}\mathbb{Q}_{T|t}(\boldsymbol{y}|\boldsymbol{x})\varphi_T(\boldsymbol{y})d\boldsymbol{y}, &\pi_0(\boldsymbol{x})=\varphi_0(\boldsymbol{x})\hat{\varphi}_0(\boldsymbol{x})\\
        \hat{\varphi}_t(\boldsymbol{x})=\int_{\mathbb{R}^d}\mathbb{Q}_{t|0}(\boldsymbol{x}|\boldsymbol{y})\hat{\varphi}_0(\boldsymbol{y})d\boldsymbol{y}, &\pi_T(\boldsymbol{x})=\varphi_T(\boldsymbol{x})\hat{\varphi}_T(\boldsymbol{x})
    \end{cases}\nonumber
\end{align}
\end{small}
Using these equations, we can replace the terminal cost in $\mathcal{L}_{\text{simple-AM}}$ (\ref{eq:simple-am-loss}) with the definition in (\ref{eq:terminal-cost-sb-potential}) given by $\nabla \Phi(\boldsymbol{x})=\nabla \log \frac{\hat{\varphi}_T(\boldsymbol{x})}{\pi_T(\boldsymbol{x})}=\nabla \log \hat{\varphi}_T(\boldsymbol{x})-\nabla \log \pi_T(\boldsymbol{x})$ to get the \boldtext{Schrödinger bridge adjoint matching} (SB-AM) loss.

\begin{definition}[Schrödinger Bridge Adjoint Matching Objective]
The \textbf{Schrödinger bridge adjoint matching} (SB-AM) loss which solves the (\ref{eq:sb-soc-objective}) is defined as:
\begin{small}
\begin{align}
    \mathcal{L}_{\text{SB-AM}}(\boldsymbol{u}):=\mathbb{E}_{\boldsymbol{X}_{0:T}\sim \mathbb{P}^{\bar{u}}}\left[\frac{1}{2}\int_0^T\left\|\boldsymbol{u}(\boldsymbol{X}_t,t)+\sigma_t(\nabla_{\boldsymbol{x}_T}\log \hat{\varphi}_T(\boldsymbol{X}_T)-\nabla_{\boldsymbol{x}_T}\log \pi_T(\boldsymbol{X}_T))\right\|^2dt\right]\tag{SB-AM Objective}\label{eq:sb-am-loss}
\end{align}
\end{small}
where $\bar{\boldsymbol{u}}=\texttt{stopgrad}(\boldsymbol{u})$ is the non-gradient-tracking control drift. 
\end{definition}

To tractably compute $\nabla_{\boldsymbol{x}_T}\log \hat{\varphi}_T(\boldsymbol{x}_T)$, we define the \boldtext{bridge-matching or corrector matching objective} (CM), which has been applied to both data-driven \citep{shi2023diffusion, liu20232} and sampling problems \citep{liu2025adjoint}.

\begin{proposition}[Corrector Matching Objective]\label{prop:sb-correction-match}
    The gradient of the log Schrödinger bridge potential $\nabla \log \hat{\varphi}_t(\boldsymbol{x})$ can be expressed as the minimizer of the \textbf{Schrödinger bridge corrector matching} (SB-CM) loss defined as:
    \begin{small}
    \begin{align}
        \mathcal{L}_{\text{SB-CM}}(\widehat{\boldsymbol{Z}}_T):=\mathbb{E}_{p^\star_{0,T}}\left[\|\widehat{\boldsymbol{Z}}_T(\boldsymbol{X}_T)-\nabla_{\boldsymbol{x}_T}\log \mathbb{Q}_{T|0}(\boldsymbol{X}_T|\boldsymbol{X}_0)\|^2\right]\tag{Corrector Matching Objective}\label{eq:corrector-matching-objective}
    \end{align}
    \end{small}
    where the minimizer defines the backward gradient of the log potential:
    \begin{small}
    \begin{align}
        \nabla \log \hat{\varphi}_T(\boldsymbol{x})=\widehat{\boldsymbol{Z}}^\star_T=\underset{\widehat{\boldsymbol{Z}}_T}{\arg\min}\mathbb{E}_{ p^\star_{0,T}}\left[\|\widehat{\boldsymbol{Z}}_T(\boldsymbol{X}_T)-\nabla_{\boldsymbol{x}_T}\log \mathbb{Q}_{T|0}(\boldsymbol{X}_T|\boldsymbol{X}_0)\|^2\right]
    \end{align}
    \end{small}
\end{proposition}

\textit{Proof.} Starting with the definition of $\hat{\varphi}_t$ from (\ref{eq:sb-system-adjoint}), we have:
\begin{small}
\begin{align}
    \nabla \log \bluetext{\hat{\varphi}_t(\boldsymbol{x})}&=\frac{\nabla \hat{\varphi}_t(\boldsymbol{x})}{\hat{\varphi}_t(\boldsymbol{x})}=\frac{1}{\hat{\varphi}_t(\boldsymbol{x})}\nabla \left(\bluetext{\int_{\mathbb{R}^d}\mathbb{Q}_{t|0}(\boldsymbol{x}|\boldsymbol{y})\hat{\varphi}_0(\boldsymbol{y})d\boldsymbol{y}}\right)\nonumber\\
    &=\frac{1}{\hat{\varphi}_t(\boldsymbol{x})}\int_{\mathbb{R}^d}\bluetext{\nabla \mathbb{Q}_{t|0}(\boldsymbol{x}|\boldsymbol{y})}\hat{\varphi}_0(\boldsymbol{y})d\boldsymbol{y}=\frac{\pinktext{\varphi_t(\boldsymbol{x})}}{p^\star_t(\boldsymbol{x})}\int_{\mathbb{R}^d}\bluetext{\nabla \log \mathbb{Q}_{t|0}(\boldsymbol{x}|\boldsymbol{y})}\pinktext{\mathbb{Q}_{t|0}(\boldsymbol{x}|\boldsymbol{y})\hat{\varphi}_0(\boldsymbol{y})}d\boldsymbol{y}
\end{align}
\end{small}
Recalling from (\ref{eq:sb-soc-joint-density}) that the joint density between any two timepoints $s\leq t$ is given by $p_{s,t}^\star(\boldsymbol{y}, \boldsymbol{x})=\mathbb{Q}(\boldsymbol{X}_s=\boldsymbol{y}|\boldsymbol{X}_t=\boldsymbol{x})\hat{\varphi}_s(\boldsymbol{y})\varphi_t(\boldsymbol{x})$, we observe that we can replace the highlighted terms to get:
\begin{align}
    \nabla \log \hat{\varphi}_t(\boldsymbol{x})&=\int_{\mathbb{R}^d}\nabla \log \mathbb{Q}_{t|0}(\boldsymbol{x}|\boldsymbol{y})\frac{1}{p^\star_t(\boldsymbol{x})}\underbrace{\pinktext{\mathbb{Q}_{t|0}(\boldsymbol{x}|\boldsymbol{y})\hat{\varphi}_0(\boldsymbol{y})\varphi_t(\boldsymbol{x})}}_{=:p^\star_{0,t}(\boldsymbol{y}, \boldsymbol{x})}d\boldsymbol{y}=\int_{\mathbb{R}^d}\nabla \log \mathbb{Q}_{t|0}(\boldsymbol{x}|\boldsymbol{y})\underbrace{\frac{\pinktext{p^\star_{0,t}(\boldsymbol{y}, \boldsymbol{x})}}{p^\star_t(\boldsymbol{x})}}_{=p^\star_{0|t}(\boldsymbol{y}|\boldsymbol{x})}d\boldsymbol{y}\nonumber\\
    &=\int_{\mathbb{R}^d}\nabla \log \mathbb{Q}_{t|0}(\boldsymbol{x}|\boldsymbol{y})p^\star_{0|t}(\boldsymbol{y}|\boldsymbol{x})d\boldsymbol{y}=\mathbb{E}_{\pinktext{p^\star_{0,t}}}\left[\bluetext{\nabla \log \mathbb{Q}_{t|0}(\boldsymbol{X}_t|\boldsymbol{X}_0)}|\pinktext{\boldsymbol{X}_t=\boldsymbol{x}}\right]
\end{align}
This can be rewritten as a \textbf{regression objective} where pairs $(\boldsymbol{X}_0, \boldsymbol{X}_t)\sim p^\star_{0,t}$ and we minimize a parameterized function $\widehat{\boldsymbol{Z}}_t(\boldsymbol{x}):\mathbb{R}^d\times[0,T]\to \mathbb{R}^d$ that minimizes the square loss:
\begin{align}
    \widehat{\boldsymbol{Z}}^\star_t:=\underset{h_t}{\arg\min}\mathbb{E}_{p^\star_{0,t}}\left[\|\widehat{\boldsymbol{Z}}_t(\boldsymbol{X}_t)-\bluetext{\nabla \log \mathbb{Q}_{t|0}(\boldsymbol{X}_t|\boldsymbol{X}_0)}\|^2\right]
\end{align}
where the minimizer aligns with the true log potential $\widehat{\boldsymbol{Z}}^\star_t(\boldsymbol{x})=\nabla \log \hat{\varphi}_t(\boldsymbol{x})$. For $t=T$, we obtain the expression $\nabla \log\hat{\varphi}_T(\boldsymbol{x})$ that appears in $\mathcal{L}_{\text{SB-AM}}$ (\ref{eq:sb-am-loss}) as:
\begin{align}
    \widehat{\boldsymbol{Z}}^\star_T:=\underset{\widehat{\boldsymbol{Z}}_T}{\arg\min}\mathbb{E}_{p^\star_{0,T}}\left[\|\widehat{\boldsymbol{Z}}_T(\boldsymbol{X}_T)-\bluetext{\nabla \log \mathbb{Q}_{T|0}(\boldsymbol{X}_T|\boldsymbol{X}_0)}\|^2\right]
\end{align}
which is our definition for the (\ref{eq:corrector-matching-objective}). \hfill $\square$

Although this provides a concrete variational objective for obtaining $\nabla \log \hat{\varphi}_T(\boldsymbol{x})$ required to optimize (\ref{eq:sb-am-loss}), it requires sampling $(\boldsymbol{X}_0, \boldsymbol{X}_T)\sim p^\star_{0,T}$. In the case of sampling from an energy-based distribution, we have no explicit access to the target distribution $p_T^\star$, which is exactly the challenge addressed by the \boldtext{adjoint Schrödinger bridge sampler} \citep{liu2025adjoint} algorithm. Rather than sampling pairs from the optimal joint distribution $p^\star_{0,T}$, we optimize the corrector $\widehat{\boldsymbol{Z}}_T$ using samples from the joint distribution $p^{\bar{u}}_{0,T}$ generated with the frozen control $\bar{\boldsymbol{u}}$. However, optimizing $\boldsymbol{u}$ using $\mathcal{L}_{\text{SB-AM}}$ (\ref{eq:sb-am-loss}) also requires computing $\nabla_{\boldsymbol{x}_T}\log \hat{\varphi}_T(\boldsymbol{X}_T)$, which introduces cross-dependencies between the parameterized variables.

This naturally motivates an \textbf{alternating optimization scheme}, that switches between training $\boldsymbol{u}$ with the parameters of $\widehat{\boldsymbol{Z}}_T$ fixed and optimizing $\widehat{\boldsymbol{Z}}_T$ with the parameters of $\boldsymbol{u}$ fixed. Concretely, we can interpret this as optimizing a pair of \textbf{forward and backward SDEs} that are characterized by the control drift $\boldsymbol{u}$ in the forward time coordinate $t\in [0,T]$ and correction term $\nabla \log \hat{\varphi}_t=\nabla \log \hat{\varphi}_{T-s}$ in the backward time coordinate $s=T-t\in [0,T]$, respectively:
\begin{align}
    \mathbb{P}^u: \quad d\boldsymbol{X}_t&=\left(\boldsymbol{f}(\boldsymbol{X}_t, t)+\sigma_t\boldsymbol{u}(\boldsymbol{X}_t,t)\right)dt+\sigma_td\boldsymbol{B}_t, \quad &\boldsymbol{X}_0\sim \pi_0\label{eq:sb-am-forward}\\
    \mathbb{P}^{\widehat{Z}}: \quad d\tilde{\boldsymbol{X}}_s&=\left[-\boldsymbol{f}(\tilde{\boldsymbol{X}}_s,s)+\sigma^2_s\nabla \log \hat{\varphi}_{T-s}(\tilde{\boldsymbol{X}}_s)\right]ds+\sigma_sd\boldsymbol{B}_s, \quad &\tilde{\boldsymbol{X}}_0\sim \pi_T\label{eq:sb-am-reverse}
\end{align}
where $\hat{\varphi}_{T-s}$ is defined in the forward time coordinate via (\ref{eq:sb-system-adjoint}) with the terminal constraint $\nabla_{\boldsymbol{x}_T}\log\hat{\varphi}_T(\boldsymbol{x}_T)=\widehat{\boldsymbol{Z}}_T(\boldsymbol{x}_T)$. Then, optimizing both (\ref{eq:sb-am-loss}) and (\ref{eq:corrector-matching-objective}) reduces to determining the optimal pair $(\boldsymbol{u}^\star, \widehat{\boldsymbol{Z}}^\star_T)$, where the path measures generated by both the forward and backward SDEs align with the Schrödinger bridge path $\mathbb{P}^{u^\star}=\mathbb{P}^{\widehat{Z}^\star}=\mathbb{P}^\star$. Indeed, the alternating optimization scheme achieves this goal, which we will show by establishing that optimizing (\ref{eq:sb-am-loss}) generates the \textbf{optimal forward half-bridge} (Proposition \ref{prop:sb-adjoint-match}) and that optimizing (\ref{eq:corrector-matching-objective}) generates the \textbf{optimal backward half-bridge with from the terminal constraint} (Proposition \ref{prop:corrector-match}).

\begin{proposition}[Adjoint Matching Solves the Forward Schrödinger Bridge (Theorem 4.1 in \citet{liu2025adjoint})]\label{prop:sb-adjoint-match}
    Consider optimizing the (\ref{eq:sb-am-loss}) with respect to the control drift $\boldsymbol{u}^{(k)}$ at iteration $k$, then the path measure generated by the control $\mathbb{P}^{u^{(k)}}$ solves the \textbf{forward half bridge} defined by the initial marginal constraint $p_0=\pi_0$ given by:
    \begin{align}
        \mathbb{P}^{u^{(k)}}=\underset{\mathbb{P}^u}{\arg\min}\left\{\text{KL}(\mathbb{P}^u\|\mathbb{P}^{\widehat{Z}}); \mathbb{P}^u_0=\pi_0 \right\}
    \end{align}
    where $\mathbb{P}^{\widehat{Z}}$ is the path measure generated \textbf{time-reversed SDE} defined in (\ref{eq:sb-am-reverse}).
\end{proposition}

\textit{Proof. }The goal of this proof is to show that optimizing the control drift $\boldsymbol{u}^{(k)}$ using the corrector $\widehat{\boldsymbol{Z}}_T^{(k-1)}$ from the previous iteration solves the \textbf{forward half-bridge that is closest in KL divergence to the reverse-time dynamics generated from $\widehat{\boldsymbol{Z}}_T^{(k-1)}$}. To do this, we first define the forward-time dynamics corresponding to the reverse-time SDE induced by $\widehat{\boldsymbol{Z}}_T^{(k-1)}$. Then, we apply Itô calculus to write the variational KL objective as an Itô integral which reduces the forward half-bridge matching objective to the (\ref{eq:sb-am-loss}). Throughout the proof, we will denote the density induced by $\widehat{\boldsymbol{Z}}_T^{(k-1)}$ as $p^{\widehat{Z}}\equiv p^{\widehat{Z}^{(k-1)}}$ and denote $\hat{z}_T$ as the potential which generates the gradient field  $\widehat{\boldsymbol{Z}}=\nabla \log \hat{z}_T$\footnote{While the potential $\hat{z}_T$ is not the value typically learned in practice, it provides a useful notation for defining the terminal constraint that the dynamics induced by the learned gradient field $\widehat{\boldsymbol{Z}}_T$}.

\textbf{Step 1: Define the Forward SDE from the Backward Dynamics. }
Since the corrector $h^{(k-1)}$ defines the reverse-time dynamics via (\ref{eq:sb-am-reverse}) following the time coordinate $s=T-t$, we can define the corresponding forward-time dynamics for $t$ using the (\ref{eq:time-rev-forward-back-sde}) to get:
\begin{small}
\begin{align}
    \mathbb{P}^{\widehat{Z}}:d\boldsymbol{X}_t=\bigg[\boldsymbol{f}(\boldsymbol{X}_t,t)\bluetext{\underbrace{-\sigma_t^2\nabla \log \hat{\varphi}_t(\boldsymbol{X}_t)+\sigma^2_t\nabla \log p_t^{\widehat{Z}}(\boldsymbol{X}_t)}_{(\bigstar)}}\bigg]dt + \sigma_td\boldsymbol{B}_t, \quad \boldsymbol{X}_T\sim \pi_T\label{eq:sb-am-fwd-proof1}
\end{align}
\end{small}
The (\ref{eq:corrector-matching-objective}) aims to match $\mathbb{P}^{\widehat{Z}}$ with a path measure induced by some control drift $\boldsymbol{u}$ with the forward-time SDE:
\begin{align}
    \mathbb{P}^u:\quad d\boldsymbol{X}_t=(\boldsymbol{f}(\boldsymbol{X}_t,t)+\sigma_t\bluetext{\underbrace{\boldsymbol{u}(\boldsymbol{X}_t,t)}_{(\diamond)}})dt+\sigma_td\boldsymbol{B}_t, \quad \boldsymbol{X}_0\sim \pi_0\label{eq:sb-am-fwd-proof2}
\end{align}
which minimizes the KL divergence to $\mathbb{P}^{\widehat{Z}}$. Matching $(\diamond)$ to $(\bigstar)$ can be expanded using the KL divergence derived via Girsanov's theorem in Section \ref{subsec:path-measure-rnd-kl} as:
\begin{small}
\begin{align}
    \text{KL}(\mathbb{P}^u\|\mathbb{P}^{\widehat{Z}})=\mathbb{E}_{\boldsymbol{X}^u_{0:T}\sim \mathbb{P}^u}\left[\int_0^T\frac{1}{2}\left\|\boldsymbol{u}(\boldsymbol{X}^u_t,t)+\sigma_t\left(\nabla \log \hat{\varphi}_t(\boldsymbol{X}^u_t)-\nabla \log p_t^{\widehat{Z}}(\boldsymbol{X}^u_t))\right)\right\|^2dt\right]\label{eq:sb-am-fwd-proof3}
\end{align}
\end{small}
which depends on the integral of gradient terms $\nabla \log \hat{\varphi}_t(\boldsymbol{X}_t)$ and $\nabla \log p_t^{\widehat{Z}}(\boldsymbol{X}_t)$. Since we know that the Itô integral evaluates to the difference between boundary conditions, we can rewrite the objective using Itô's calculus.

\textbf{Step 2: Apply Itô's Caclulus to Simplify Integral. }
To expand the gradient terms, we can apply (\ref{eq:ito-2}) to the Itô processes defined by $\log \hat{\varphi}_t(\boldsymbol{X}^u_t)$ and $\log p_t^{\widehat{Z}}(\boldsymbol{X}^u_t)$ which contains the desired gradient term. Since the KL divergence (\ref{eq:sb-am-fwd-proof3}) is respect to the $\mathbb{P}^u$, the underlying Itô process follows the SDE in (\ref{eq:sb-am-fwd-proof2}), which we denote as $(\boldsymbol{X}_t^u)_{t\in [0,T]}$. First, for $\log \hat{\varphi}_t(\boldsymbol{X}^u_t)$, we have:
\begin{small}
\begin{align}
    d\log \hat{\varphi}_t(\boldsymbol{X}^u_t)=&\left[\bluetext{\partial_t\log \hat{\varphi}_t(\boldsymbol{X}^u_t)}+(\boldsymbol{f}+\sigma_t \boldsymbol{u})(\boldsymbol{X}^u_t,t)^{\top}\nabla \log \hat{\varphi}_t(\boldsymbol{X}^u_t)+\frac{\sigma_t^2}{2}\Delta \log \hat{\varphi}_t(\boldsymbol{X}^u_t)\right]dt\nonumber\\
    &+\sigma_t\nabla \log \hat{\varphi}_t(\boldsymbol{X}^u_t)d\boldsymbol{B}_t\label{eq:sb-am-fwd-proof6}
\end{align}
\end{small}
To derive the expression for $\bluetext{\partial_t\log \hat{\varphi}_t(\boldsymbol{X}^u_t)}$, we recall that it follows a deterministic integral $\hat{\varphi}_t(\boldsymbol{x})=\int_{\mathbb{R}^d}\mathbb{Q}_{t|0}(\boldsymbol{x}|\boldsymbol{y})\hat{\varphi}_0(\boldsymbol{y})d\boldsymbol{y}$ with terminal condition $\hat{\varphi}_T=\hat{z}_T$. From Section \ref{subsec:sb-soc}, we show that any function applied to a Markov process defined by a terminal constraint satisfies the (\ref{eq:feynman-kac-formula}). Given the terminal constraint $\hat\varphi_T(\boldsymbol{x})=\hat{z}_T(\boldsymbol{x})$, the (\ref{eq:feynman-kac-formula}) for $\hat\varphi_t$ is given by:
\begin{align}
    \partial_t\hat{\varphi}_t(\boldsymbol{x})=-\nabla \cdot (\boldsymbol{f}(\boldsymbol{x},t)\hat\varphi_t(\boldsymbol{x}))+\frac{\sigma^2_t}{2}\Delta \hat\varphi_t(\boldsymbol{x}), \quad\varphi_T(\boldsymbol{x})=\hat{z}_T(\boldsymbol{x})\label{eq:sb-am-fwd-proof4}
\end{align}
To obtain $\partial_t\log \hat\varphi_t$ from (\ref{eq:sb-am-fwd-proof4}), we apply the chain rule and the divergence property $\nabla\cdot (\boldsymbol{f}\hat\varphi_t)=(\nabla \cdot \boldsymbol{f})\hat\varphi_t+\boldsymbol{f}^\top \nabla \hat\varphi_t$ to get:
\begin{small}
\begin{align}
    \partial_t\log \hat\varphi_t&=\frac{1}{\hat\varphi_t}\bluetext{\partial_t\hat\varphi_t}=\frac{1}{\hat\varphi_t}\bluetext{\left[-\nabla \cdot (\boldsymbol{f}\hat\varphi_t)+\frac{\sigma^2_t}{2}\Delta \hat\varphi_t\right]}\nonumber\\
    &=-\frac{1}{\hat\varphi_t}((\nabla \cdot \boldsymbol{f})\hat\varphi_t-\boldsymbol{f}^\top \nabla \hat\varphi_t)+\frac{\sigma^2_t}{2}\frac{\Delta \hat\varphi_t}{\hat\varphi_t}\nonumber\\
    &=-\nabla \cdot \boldsymbol{f}-\boldsymbol{f}^\top \bluetext{\frac{\nabla \hat\varphi_t}{\hat\varphi_t}}+\frac{\sigma^2_t}{2}\pinktext{\underbrace{\frac{\Delta \hat\varphi_t}{\hat\varphi_t}}_{\text{Laplacian trick}}}\nonumber\\
    &=-\nabla \cdot \boldsymbol{f}-\boldsymbol{f}^\top \bluetext{\nabla \log \hat\varphi_t}+\frac{\sigma^2_t}{2}\pinktext{(\Delta\log \hat\varphi_t+\|\nabla \log \hat\varphi_t\|^2)}\label{eq:sb-am-fwd-proof5}
\end{align}
\end{small}
where the final equality is obtained from applying the Laplacian trick\footnote{derived as \begin{align}
    \Delta\hat\varphi_t&=\nabla\cdot \nabla \hat\varphi_t=\nabla\cdot (\hat\varphi_t\nabla \log \hat\varphi_t)=\nabla \hat\varphi_t \cdot \nabla \log \hat\varphi_t+\hat\varphi_t\Delta\log \hat\varphi_t\nonumber\\
    &=\hat\varphi_t\left(\bluetext{\frac{\nabla \hat\varphi_t}{\hat\varphi_t}\cdot\nabla \log\hat\varphi_t}+\Delta\log\hat\varphi_t\right)=\hat\varphi_t\left(\bluetext{\|\nabla \log\hat\varphi_t\|^2}+\Delta\log\hat\varphi_t\right)\nonumber
\end{align}}. Finally, substituting the expression for $\partial_t\log \hat\varphi_t$ from (\ref{eq:sb-am-fwd-proof5}) into (\ref{eq:sb-am-fwd-proof6}) and canceling like terms, we get:
\begin{small}
\begin{align}
    d\log \hat{\varphi}_t&=\left[-\nabla \cdot \boldsymbol{f}\pinktext{-\boldsymbol{f}^\top \nabla \log \hat\varphi_t}+\bluetext{\frac{\sigma^2_t}{2}}(\bluetext{\Delta\log \hat\varphi_t}+\|\nabla \log \hat\varphi_t\|^2)+(\pinktext{\boldsymbol{f}}+\sigma_t \boldsymbol{u})\pinktext{^{\top}\nabla \log \hat{\varphi}_t}+\bluetext{\frac{\sigma_t^2}{2}\Delta \log \hat{\varphi}_t}\right]dt+\sigma_t\nabla \log \hat{\varphi}_td\boldsymbol{B}_t\nonumber\\
    &=\left[-\nabla \cdot \boldsymbol{f}+\frac{\sigma^2_t}{2}\|\nabla \log \hat\varphi_t\|^2+(\sigma_t \boldsymbol{u})^{\top}\nabla \log \hat{\varphi}_t+\sigma_t^2\Delta \log \hat{\varphi}_t\right]dt+\sigma_t\nabla \log \hat{\varphi}_td\boldsymbol{B}_t\tag{$\log \hat\varphi$-SDE}\label{eq:sb-am-fwd-proof10}
\end{align}
\end{small}
Next, we apply Itô's formula to $\partial_t\log p_t^{\widehat{Z}}(\boldsymbol{X}_t^u)$:
\begin{small}
\begin{align}
    d\log p_t^{\widehat{Z}}(\boldsymbol{X}_t^u)=&\left[\bluetext{\partial_t\log p_t^{\widehat{Z}}(\boldsymbol{X}_t^u)}+(\boldsymbol{f}+\sigma_t\boldsymbol{u})(\boldsymbol{X}_t^u,t)^\top \nabla \log p_t^{\widehat{Z}}(\boldsymbol{X}_t^u)+\frac{\sigma^2_t}{2}\Delta \log p_t^{\widehat{Z}}(\boldsymbol{X}_t^u)\right]dt\nonumber\\
    &+\sigma_t\nabla \log p_t^{\widehat{Z}}(\boldsymbol{X}_t^u)d\boldsymbol{B}_t\label{eq:sb-am-fwd-proof7}
\end{align}
\end{small}
In this case, since $p_t^{\widehat{Z}}(\boldsymbol{X}_t^u)$ is defined by the forward-time SDE (\ref{eq:sb-am-fwd-proof1}), we can apply the (\ref{eq:fokker-planck equation}) to write the time-evolution of the density $\partial_t p_t^{\widehat{Z}}(\boldsymbol{X}_t^u)$ as: 
\begin{small}
\begin{align}
    \partial_tp_t^{\widehat{Z}}&=-\nabla\cdot \bigg(\pinktext{\underbrace{(\boldsymbol{f}-\sigma^2_t\nabla\log\hat\varphi_t+ \sigma_t^2\nabla\log p_t^{\widehat{Z}})}_{\text{drift of the SDE (\ref{eq:sb-am-fwd-proof1})}}}p_t^{\widehat{Z}}\bigg)+\frac{\sigma_t^2}{2}\Delta p_t^{\widehat{Z}}\nonumber\\
    &=-\nabla\cdot ((\boldsymbol{f}-\sigma^2_t\nabla\log\hat\varphi_t)p_t^{\widehat{Z}})-\nabla \cdot (\sigma_t^2\bluetext{\underbrace{p_t^{\widehat{Z}}\nabla\log p_t^{\widehat{Z}}}_{=\nabla p_t^{\widehat{Z}} }})+\frac{\sigma_t^2}{2}\Delta p_t^{\widehat{Z}}\nonumber\\
    &=-\nabla\cdot ((\boldsymbol{f}-\sigma^2_t\nabla\log\hat\varphi_t)p_t^{\widehat{Z}})- \sigma_t^2\bluetext{\underbrace{\nabla \cdot (\bluetext{\nabla p_t^{\widehat{Z}}})}_{\Delta p_t^{\widehat{Z}}}}+\frac{\sigma_t^2}{2}\Delta p_t^{\widehat{Z}}\nonumber\\
    &=-\nabla\cdot ((\boldsymbol{f}-\sigma^2_t\nabla\log\hat\varphi_t)p_t^{\widehat{Z}})- \frac{\sigma_t^2}{2}\Delta p_t^{\widehat{Z}}\label{eq:sb-am-fwd-proof8}
\end{align}
\end{small}
Then, we can apply the chain rule to express $\partial_t\log p_t^{\widehat{Z}}(\boldsymbol{x})$ from (\ref{eq:sb-am-fwd-proof8}) as: 
\begin{small}
\begin{align}
    \bluetext{\partial_t\log p_t^{\widehat{Z}}}&=\frac{1}{p_t^{\widehat{Z}}}\bluetext{\partial_tp_t^{\widehat{Z}}}=\frac{1}{p_t^{\widehat{Z}}}\bluetext{\left[-\nabla\cdot ((\boldsymbol{f}-\sigma^2_t\nabla\log\hat\varphi_t)p_t^{\widehat{Z}})- \frac{\sigma_t^2}{2}\Delta p_t^{\widehat{Z}}\right]}\nonumber\\
    &=-\frac{1}{p_t^{\widehat{Z}}}\left((p_t^{\widehat{Z}}\nabla\cdot (\boldsymbol{f}-\sigma^2_t\nabla\log\hat\varphi_t)+(\boldsymbol{f}-\sigma^2_t\nabla\log\hat\varphi_t)^\top \nabla p_t^{\widehat{Z}}\right)- \frac{\sigma_t^2}{2}\frac{\Delta p_t^{\widehat{Z}}}{p_t^{\widehat{Z}}}\nonumber\\
    &=-\nabla\cdot (\boldsymbol{f}-\sigma^2_t\nabla\log\hat\varphi_t)-(\boldsymbol{f}-\sigma^2_t\nabla\log\hat\varphi_t)^\top \bluetext{\frac{\nabla p_t^{\widehat{Z}}}{p_t^{\widehat{Z}}}}- \frac{\sigma_t^2}{2}\pinktext{\underbrace{\frac{\Delta p_t^{\widehat{Z}}}{p_t^{\widehat{Z}}}}_{\text{Laplacian trick}}}\nonumber\\
    &=-\nabla\cdot \boldsymbol{f}+\sigma^2_t\Delta \log\hat\varphi_t-(\boldsymbol{f}-\sigma^2_t\nabla\log\hat\varphi_t)^\top \bluetext{\nabla\log p_t^{\widehat{Z}}}- \frac{\sigma_t^2}{2}\pinktext{(\Delta\log p_t^{\widehat{Z}}+\|\nabla\log p^{\widehat{Z}}_t\|^2)}\label{eq:sb-am-fwd-proof9} 
\end{align}
\end{small}
Finally, substituting the expression for $\bluetext{\partial_t\log p_t^{\widehat{Z}}}$ from (\ref{eq:sb-am-fwd-proof9}) into (\ref{eq:sb-am-fwd-proof7}) and canceling like terms, we obtain:
\begin{small}
\begin{align}
    d\log p_t^{\widehat{Z}}=&\bigg[-\nabla\cdot \boldsymbol{f}+\sigma^2_t\Delta \log\hat\varphi_t-(\pinktext{\boldsymbol{f}}-\sigma^2_t\nabla\log\hat\varphi_t)\pinktext{^\top \nabla\log p_t^{\widehat{Z}}}\bluetext{- \frac{\sigma_t^2}{2}}(\bluetext{\Delta\log p_t^{\widehat{Z}}}+\|\nabla\log p^{\widehat{Z}}_t\|^2)\nonumber\\
    &+(\pinktext{\boldsymbol{f}}+\sigma_t\boldsymbol{u})\pinktext{^\top \nabla \log p_t^{\widehat{Z}}}+\bluetext{\frac{\sigma^2_t}{2}\Delta \log p_t^{\widehat{Z}}}\bigg]dt+\sigma_t\nabla \log p_t^{\widehat{Z}}d\boldsymbol{B}_t\nonumber\\
    =&\bigg[-\nabla\cdot \boldsymbol{f}+\sigma^2_t\Delta \log\hat\varphi_t+\sigma^2_t\nabla\log\hat\varphi_t^\top \nabla\log p_t^{\widehat{Z}}- \frac{\sigma_t^2}{2}\|\nabla\log p^h_t\|^2+(\sigma_t\boldsymbol{u})^\top \nabla \log p_t^{\widehat{Z}}\bigg]dt\nonumber\\
    &+\sigma_t\nabla \log p_t^{\widehat{Z}}d\boldsymbol{B}_t\tag{$\log p_t^{\widehat{Z}}$-SDE}\label{eq:sb-am-fwd-proof11}
\end{align}
\end{small}
Observing that (\ref{eq:sb-am-fwd-proof10}) and (\ref{eq:sb-am-fwd-proof11}) have several matching terms, we can cancel them by subtracting (\ref{eq:sb-am-fwd-proof11}) from (\ref{eq:sb-am-fwd-proof10}) to get:
\begin{small}
\begin{align}
    d\log \hat\varphi&- d\log p_t^{\widehat{Z}}=\left[\bluetext{-\nabla \cdot \boldsymbol{f}}+\frac{\sigma^2_t}{2}\|\nabla \log \hat\varphi_t\|^2+\pinktext{(\sigma_t \boldsymbol{u})^{\top}\nabla \log \hat{\varphi}_t}+\bluetext{\sigma_t^2\Delta \log \hat{\varphi}_t}\right]dt+\pinktext{\sigma_t\nabla \log \hat{\varphi}_td\boldsymbol{B}_t}\nonumber\\
    &-\bigg[\bluetext{-\nabla \cdot \boldsymbol{f}}+\bluetext{\sigma^2_t\Delta \log\hat\varphi_t}+\sigma^2_t\nabla\log\hat\varphi_t^\top \nabla\log p_t^{\widehat{Z}}- \frac{\sigma_t^2}{2}\|\nabla\log p^{\widehat{Z}}_t\|^2+\pinktext{(\sigma_t\boldsymbol{u})^\top \nabla \log p_t^{\widehat{Z}}}\bigg]dt+\pinktext{\sigma_t\nabla \log p_t^{\widehat{Z}}d\boldsymbol{B}_t}\nonumber\\
    =&\left[\pinktext{(\sigma_t \boldsymbol{u})^{\top}(\nabla \log \hat{\varphi}_t-\nabla \log p_t^{\widehat{Z}})}+\frac{\sigma^2_t}{2}\|\nabla \log \hat\varphi_t\|^2+\frac{\sigma^2_t}{2}\|\nabla \log p^{\widehat{Z}}_t\|^2+\sigma^2_t\nabla\log \hat\varphi_t^\top \nabla\log p_t^{\widehat{Z}}\right]dt\nonumber\\
    &+\pinktext{\sigma_t\nabla \log \frac{\hat\varphi_t}{p_t^{\widehat{Z}}}d\boldsymbol{B}_t}
\end{align}
\end{small}
where the terms inside the bracket are almost a perfect square of $\boldsymbol{u}+\nabla \log \hat\varphi_t-\nabla \log p_t^{\widehat{Z}}$. Completing the square with $\frac{1}{2}\|\boldsymbol{u}\|^2$, we get:
\begin{small}
\begin{align}
    d\log \hat\varphi_t- d\log p_t^{\widehat{Z}}&=\underbrace{\left[\frac{1}{2}\big\|\boldsymbol{u}+\nabla \log \hat\varphi_t-\nabla \log p_t^{\widehat{Z}}\big\|^2\right]dt}_{\text{matches the KL divergence objective}}-\frac{1}{2}\|\boldsymbol{u}\|^2dt+\sigma_t\nabla \log \frac{\hat\varphi_t}{p_t^{\widehat{Z}}}d\boldsymbol{B}_t\nonumber\\
    \left[\frac{1}{2}\big\|\boldsymbol{u}+\nabla \log \hat\varphi_t-\nabla \log p_t^{\widehat{Z}}\big\|^2\right]dt&=\frac{1}{2}\|\boldsymbol{u}\|^2dt+d\log \hat\varphi_t- d\log p_t^{\widehat{Z}}-\sigma_t\nabla \log \frac{\hat\varphi_t}{p_t^{\widehat{Z}}}d\boldsymbol{B}_t\label{eq:sb-am-fwd-proof12}
\end{align}
\end{small}
which recovers the integrand from our KL divergence objective. 

\textbf{Step 3: Rewriting the KL Divergence. }
Using (\ref{eq:sb-am-fwd-proof12}), we can rewrite the KL divergence objective in (\ref{eq:sb-am-fwd-proof3}) as:
\begin{small}
\begin{align}
    \text{KL}(\mathbb{P}^u\|\mathbb{P}^{{\widehat{Z}}})&=\mathbb{E}_{\boldsymbol{X}^u_{0:T}\sim \mathbb{P}^u}\bigg[\int_0^T\left(\frac{1}{2}\|\boldsymbol{u}(\boldsymbol{X}^u_t,t)\|^2+d\log \hat\varphi_t(\boldsymbol{X}^u_t)- d\log p_t^{\widehat{Z}}(\boldsymbol{X}^u_t)\right)dt-\underbrace{\int_0^T\sigma_t\nabla \log \frac{\hat\varphi_t(\boldsymbol{X}^u_t)}{p_t^{\widehat{Z}}(\boldsymbol{X}^u_t)}d\boldsymbol{B}_t}_{=0\text{ (Itô integral)}}\bigg]\nonumber\\
    &=\mathbb{E}_{\boldsymbol{X}^u_{0:T}\sim \mathbb{P}^u}\bigg[\frac{1}{2}\|\boldsymbol{u}(\boldsymbol{X}^u_t,t)\|^2dt+\bluetext{\underbrace{\int_0^Td\log \hat\varphi_t(\boldsymbol{X}^u_t)}_{\log \hat\varphi_T-\log \hat\varphi_0}}- \pinktext{\underbrace{\int_0^Td\log p_t^{\widehat{Z}}(\boldsymbol{X}^u_t)}_{\log p^{\widehat{Z}}_T-\log p^{\widehat{Z}}_0}}\bigg]\nonumber\\
    &=\mathbb{E}_{\boldsymbol{X}^u_{0:T}\sim \mathbb{P}^u}\bigg[\frac{1}{2}\|\boldsymbol{u}(\boldsymbol{X}^u_t,t)\|^2dt+\log\frac{\bluetext{\hat\varphi_T(\boldsymbol{X}^u_T)}}{\pinktext{p_T^{\widehat{Z}}(\boldsymbol{X}^u_T)}}-\underbrace{\log\frac{\bluetext{\hat\varphi_0(\boldsymbol{X}^u_0)}}{\pinktext{p_0^{\widehat{Z}}(\boldsymbol{X}^u_0)}}}_{\text{constant}}\bigg]
\end{align}
\end{small}
where the term dependent on $\boldsymbol{X}^u_0$ is a constant with respect to $\boldsymbol{u}$ since we fix the initial distribution at $\boldsymbol{X}_0\sim \pi_0$. Therefore, substituting $\hat\varphi_T=\hat{z}_T^{(k-1)}$ into the objective, we have:
\begin{align}
    \text{KL}(\mathbb{P}^u\|\mathbb{P}^{\widehat{Z}})=\mathbb{E}_{\boldsymbol{X}_{0:T}\sim \mathbb{P}^u}\bigg[\frac{1}{2}\|\boldsymbol{u}(\boldsymbol{X}_t,t)\|^2dt+\log\frac{\bluetext{\hat{z}^{(k-1)}_T(\boldsymbol{X}_T)}}{\pi_T(\boldsymbol{X}_T)}+\text{const}\bigg]
\end{align}
which is proportional to the (\ref{eq:sb-am-loss}) up to an additive constant, and we conclude that the control $\boldsymbol{u}^{(k)}$ obtained from minimizing (\ref{eq:sb-am-loss}) solves the forward half-bridge that minimizes $\text{KL}(\mathbb{P}^u\|\mathbb{P}^{\widehat{Z}})$.\hfill $\square$

However, we have already shown in (\ref{box:memoryless}) that solving the forward half-bridge for arbitrary prior distributions and reference drifts results in a mismatch of the target distribution, which motivated the definition of (\ref{eq:corrector-matching-objective}) to define the terminal cost as $\Phi(\boldsymbol{x}):= \log \frac{\hat\varphi_T(\boldsymbol{x})}{\pi_T(\boldsymbol{x})}$. While this definition of the terminal cost provably eliminates the initial value bias as shown in Section \ref{subsec:sb-soc}, we can further show that by optimizing (\ref{eq:corrector-matching-objective}), we obtain the optimal corrector $\widehat{\boldsymbol{Z}}_T^{(k)}$ that induces the backward half-bridge which is closest in KL to the forward dynamics defined by any arbitrary control $\boldsymbol{u}$ while correcting for the bias created at the terminal distribution $\pi_T$.

\begin{proposition}[Corrector Matching Solves the Backward Schrödinger Bridge (Theorem 4.2 in \citet{liu2025adjoint})]\label{prop:corrector-match}
    Consider optimizing the (\ref{eq:corrector-matching-objective}) with the control drift $\boldsymbol{u}^{(k)}$ from the $k$th iteration, then the path measure generated by the corrector $\widehat{\boldsymbol{Z}}_T^{(k)}$ solves the backward half bridge defined by the terminal constraint $p_T=\pi_T$ as:
    \begin{align}
        \mathbb{P}^{\widehat{Z}}=\underset{\mathbb{P}^{\widehat{Z}}}{\arg\min}\left\{\text{KL}(\mathbb{P}^{u^{(k)}}\|\mathbb{P}): p^{\widehat{Z}}_T=\pi_T\right\}
    \end{align}
    where $\mathbb{P}^{u^{(k)}}$ is the path measure generated from the learned control at iteration $k$ defined in (\ref{eq:sb-am-forward}).
\end{proposition}

\textit{Proof.} This proof starts by defining the time-reversal of the forward half-bridge generated by $\boldsymbol{u}^{(k)}$ from the $k$th iteration of the algorithm and showing that optimizing (\ref{eq:corrector-matching-objective}) yields the optimal reverse-time dynamics that enforce the terminal constraint.

\textbf{Step 1: Time Reversal of Forward Controlled SDE. }
Given the control drift from the current iteration of the optimization algorithm $\boldsymbol{u}^{(k)}$, we can define the corresponding backward SDE following the reversed time coordinate $s:= T-t$ using the (\ref{eq:time-rev-forward-back-sde}) to get:
\begin{small}
\begin{align}
    \mathbb{P}^{u^{(k)}}: \quad d\tilde{\boldsymbol{X}}_s=\left(-\boldsymbol{f}(\tilde{\boldsymbol{X}}_s,s)-\sigma_s\bluetext{\boldsymbol{u}^{(k)}(\tilde{\boldsymbol{X}}_s,s)}+\sigma_s^2\nabla \log \bluetext{p^{u^{(k)}}_s(\tilde{\boldsymbol{X}}_s)}\right)ds + \sigma_sd\boldsymbol{B}_s, \quad \tilde{\boldsymbol{X}}_0\sim \pi_T\label{eq:k-control-time-reversal}
\end{align}
\end{small}
where the highlighted terms are defined by the control $\boldsymbol{u}^{(k)}$. Since we have shown that $\boldsymbol{u}^{(k)}$ solves the forward half-bridge in Proposition \ref{prop:sb-adjoint-match}, it satisfies the SB equations given by:
\begin{small}
\begin{align}
    \boldsymbol{u}^{(k)}_t(\boldsymbol{x},t)=\sigma_t\nabla \log \varphi_t(\boldsymbol{x})\quad 
    \begin{cases}
        \varphi_t(\boldsymbol{x})=\int_{\mathbb{R}^d}\mathbb{Q}_{T|t}(\boldsymbol{y}|\boldsymbol{x})\varphi_T(\boldsymbol{y})d\boldsymbol{y}, & \pi_0(\boldsymbol{x})=\varphi_0(\boldsymbol{x})\hat{\varphi}_0(\boldsymbol{x})\\
        \hat{\varphi}_t(\boldsymbol{x})=\int_{\mathbb{R}^d}\mathbb{Q}_{t|0}(\boldsymbol{x}|\boldsymbol{y})\hat{\varphi}_0(\boldsymbol{y})d\boldsymbol{y}, & \bluetext{p^{u^{(k)}}_T}(\boldsymbol{x})=\varphi_T(\boldsymbol{x})\hat{\varphi}_T(\boldsymbol{x})\label{eq:k-control-time-reversal2}
    \end{cases}
\end{align}
\end{small}
which does not necessarily satisfy the terminal constraint $p_T^\star=\pi_T$. The goal of the (\ref{eq:corrector-matching-objective}) is to generate the backward half-bridge that minimizes the divergence from the forward half-bridge while constraining the terminal marginal to $p^{\widehat{Z}}_T=\pi_T$.

\textbf{Step 2: Deriving the Matching Objective.} 
To do this, we aim to match some arbitrary control $\boldsymbol{v}(\boldsymbol{X}_s,s)$ to the time-reversal of the forward half-bridge in (\ref{eq:k-control-time-reversal}) where the initial states $\tilde{\boldsymbol{X}}_0\sim \pi_T$ are sampled from the target marginal $\pi_T$. To define the matching loss as a KL divergence, let $\boldsymbol{v}(\boldsymbol{x},s)$ be an arbitrary control drift defining the reverse-time SDE initialized at $\pi_T$:
\begin{small}
\begin{align}
    \mathbb{P}^v: &d\tilde{\boldsymbol{X}}_s=(-\boldsymbol{f}(\tilde{\boldsymbol{X}}_s,s)+\sigma_s\bluetext{\boldsymbol{v}(\tilde{\boldsymbol{X}}_s, s)}) ds+ \sigma_sd\boldsymbol{B}_s,\quad \tilde{\boldsymbol{X}}_0\sim \bluetext{\pi_T}\label{eq:sb-am-bwd-sde}\\
    \mathbb{P}^{u^{(k)}}: &d\tilde{\boldsymbol{X}}_s=\left(-\boldsymbol{f}(\tilde{\boldsymbol{X}}_s,s)+\sigma_s\bluetext{(-\sigma_s\nabla \log \varphi_s(\tilde{\boldsymbol{X}}_s)+\sigma_s\nabla \log p^{u^{(k)}}_s(\tilde{\boldsymbol{X}}_s))}\right)ds + \sigma_sd\boldsymbol{B}_s,  \quad \tilde{\boldsymbol{X}}_0\sim \bluetext{p^{u^{(k)}}_T}
\end{align}
\end{small}
which we will show matches the result obtained from minimizing (\ref{eq:corrector-matching-objective}) at optimality. Now, expanding the KL divergence $\text{KL}(\mathbb{P}^{u^{(k)}}\|\mathbb{P}^v)$ as shown in Section \ref{subsec:path-measure-rnd-kl}, we have:
\begin{small}
\begin{align}
    \text{KL}(\mathbb{P}^{u^{(k)}}\|\mathbb{P}^v)=\mathbb{E}_{\boldsymbol{X}_{0:T}\sim \mathbb{P}^{u^{(k)}}}\left[\int_0^T\frac{1}{2}\big\|\bluetext{(-\sigma_s\nabla \log \varphi_s(\tilde{\boldsymbol{X}}_s)+\sigma_s\nabla \log p^{u^{(k)}}_s(\tilde{\boldsymbol{X}}_s))}-\bluetext{\boldsymbol{v}(\tilde{\boldsymbol{X}}_s, s)}\big\|^2ds\right]\label{eq:fwb-sb-am-1}
\end{align}
\end{small}
Minimizing (\ref{eq:fwb-sb-am-1}) yields for all $(\boldsymbol{x},s)$, the following expression for $\boldsymbol{v}(\boldsymbol{x},s)$:
\begin{small}
\begin{align}
    \boldsymbol{v}^\star(\boldsymbol{x},s)=-\sigma_s\nabla \log \varphi_{T-s}(\boldsymbol{x})+\sigma_s\nabla \log p^{u^{(k)}}_s(\boldsymbol{x})=\sigma_s\nabla \log \bluetext{\underbrace{\frac{\varphi_s(\boldsymbol{x})}{p^{u^{(k)}}_s(\boldsymbol{x})}}_{=\hat{\varphi}_{T-s}(\boldsymbol{x})}}=\sigma_s\nabla \log \hat{\varphi}_{T-s}(\boldsymbol{x})
\end{align}
\end{small}
where we use the (\ref{eq:sb-system-adjoint}) which defines $p^{u^{(k)}}_t(\boldsymbol{x})= \varphi_t(\boldsymbol{x})\hat{\varphi}_t(\boldsymbol{x})$. Substituting this expression into the backward path measure $\mathbb{P}^v$ in (\ref{eq:sb-am-bwd-sde}), we have the \textbf{optimal half-bridge is generated by the SDE}:
\begin{small}
\begin{align}
    d\tilde{\boldsymbol{X}}_s=(-\boldsymbol{f}(\tilde{\boldsymbol{X}}_s,s)+\sigma_s^2\bluetext{\nabla \log \hat{\varphi}_{T-s}(\tilde{\boldsymbol{X}}_s)}) ds+ \sigma_sd\boldsymbol{B}_s, \quad \bluetext{\hat{\varphi}_T}=\frac{p_T^{u^{(k)}}}{\varphi_T}
\end{align}
\end{small}
which is \textit{fully characterized} by the terminal condition $\hat\varphi_T=\frac{p_T^{u^{(k)}}}{\varphi_T}$ from which $\hat{\varphi}_{T-s}$ can be defined using (\ref{eq:sb-system-adjoint}). From Proposition \ref{prop:sb-correction-match}, we show that optimizing (\ref{eq:corrector-matching-objective}) yields $\widehat{\boldsymbol{Z}}^{(k)}_T=\nabla \log \hat\varphi_T(\boldsymbol{x})$ given the optimal SB density $p^\star_t$, so applying the same logic, we have: 
\begin{align}
    \widehat{\boldsymbol{Z}}^{(k)}_T:=\underset{\widehat{\boldsymbol{Z}}_T}{\arg\min}\mathbb{E}_{p^{u^{(k)}}_{0,T}}\left[\left\|\widehat{\boldsymbol{Z}}_T(\boldsymbol{X}_T)-\nabla \log \mathbb{Q}_{T|0}(\boldsymbol{X}_T|\boldsymbol{X}_0)\right\|^2\right]\overset{(\ref{prop:sb-correction-match})}{=}\bluetext{\nabla \log\hat{\varphi}_T}
\end{align}
which yields the same backward time SDE through the terminal constraint as the optimal drift $\boldsymbol{v}^\star$, and we have shown that optimizing (\ref{eq:corrector-matching-objective}) is \textbf{equivalent to finding the optimal reverse-time dynamics that correct the SB forward-time SDE} such that it satisfies the terminal constraint. \hfill $\square$

The results from Proposition \ref{prop:sb-adjoint-match} and Proposition \ref{prop:corrector-match} indicate that alternating between optimizing (\ref{eq:sb-am-loss}) and (\ref{eq:corrector-matching-objective}) is equivalent to alternating between solving the forward Schrödinger half-bridge that satisfies the initial marginal to solving the backward Schrödinger half-bridge that satisfies the target marginal. This alternating scheme is reminiscent of our discussion of \textbf{Sinkhorn's algorithm} from Section \ref{subsec:sinkhorn-algorithm}, but now adapted using an efficient \textit{matching objective}. Just like the adjoint matching algorithm (Box \ref{alg:adjoint-sampling}), we can optimize (\ref{eq:sb-am-loss}) and (\ref{eq:corrector-matching-objective}) by repeatedly optimizing over samples from a replay buffer.

\begin{remark}[Adjoint and Corrector Matching Doesn't Require Target Samples]\label{remark:corrector-match-data}
    We highlight that optimizing (\ref{eq:sb-am-loss}) and (\ref{eq:corrector-matching-objective}) does not require explicit samples from $\boldsymbol{X}_T\sim\pi_T$ since the samples used to compute the objective are purely from sampling $(\boldsymbol{X}_0, \boldsymbol{X}_T)\sim p^u_{0,T}$ from the SDE induced by $\boldsymbol{u}$. The dependence on $\pi_T$ only appears when computing the loss, which can be computed as a \textbf{probability under $\pi_T$}, which can be empirical or a pre-defined potential energy function, as we will discuss further in Section \ref{subsec:sampling}.
\end{remark}

Throughout this section, we have demonstrated how the adjoint state can be used as an efficient variational framework for solving Schrödinger bridge problems by alternating between two tractable objectives that correspond to the forward and backward Schrödinger half-bridges. Rather than directly optimizing over path measures, the method reduces the problem to learning the forward control drift and the backward correction through cheap matching objectives that can be evaluated using trajectories generated by the current dynamics. This formulation yields several practical advantages:
\begin{enumerate}
    \item [(i)] It avoids the need for expensive likelihood ratios or full path-space KL computations, replacing them with local drift-matching losses that are straightforward to estimate.
    \item[(ii)] The alternating optimization naturally mirrors the structure of Sinkhorn iterations in entropic optimal transport, providing convergence guarantees to the optimal SB control $\boldsymbol{u}^\star$.
    \item[(iii)] The corrector matching objective does not require explicit samples from the target distribution, enabling training on unknown energy-based target densities.
    \item[(iv)] Since the objectives do not require backpropagation through SDE trajectories or maintaining the full trajectory in memory, it can easily scale to high-dimensional systems. 
\end{enumerate}

These properties make adjoint matching a practical and flexible approach for learning Schrödinger bridges in complex generative modeling settings.

\subsection{Closing Remarks for Section \ref{sec:generative-modeling}}
In this section, we explored several generative modeling frameworks grounded in Schrödinger bridge (SB) theory. We began with the classical score-based generative modeling paradigm (Section \ref{subsec:primer-sgm}), which can be interpreted as a specialized instance of the SB formulation in which the forward process is fixed, and learning focuses on estimating the reverse-time dynamics. Building on this perspective, we then leveraged SB theory to generalize to controlled forward processes, where both the forward and backward control drifts can be learned through likelihood-based training (Section \ref{subsec:likelihood-training}).

Next, we introduced an alternative viewpoint through the Iterative Markovian Fitting (IMF) procedure and the diffusion Schrödinger bridge matching algorithm (Section \ref{subsec:diffusionsbm}), which draws on the Markov and reciprocal projection theory developed in Section \ref{subsec:markov-reciprocal-proj}. In this formulation, solving the Schrödinger bridge problem can be understood as performing iterative projections in path space.

To overcome the limitations in optimizing over full stochastic trajectories in the path space from the previous approaches, we conclude with two approaches for learning Schrödinger bridges with efficient \textbf{matching objectives}, including score and flow matching (Section \ref{subsec:score-and-flow}) and adjoint matching (Section \ref{subsec:adjoint-matching}), which locally optimize the control drift to generate trajectories consistent with the optimal Schrödinger bridge dynamics.

Overall, this section builds the intuition behind the core generative modeling frameworks that leverage Schrödinger bridge theory. While it is not intended to be an exhaustive review of algorithmic developments in the field, it should provide the core theoretical foundations needed to understand a broad class of modern generative modeling techniques. 

So far, we have restricted our attention to the \textbf{continuous state space}, where data is represented as continuous-valued vectors in $\mathbb{R}^d$. The structure of the continuous state space is \textit{required} for many ideas developed throughout this guide, including stochastic differential equations, path measures, and Brownian motion. This naturally raises the question: \textit{How does Schrödinger bridge theory extend to the \textbf{discrete state space}, where states belong to a finite or countable set rather than a continuous vector space?} This is precisely the question that we explore in the next section:
\begin{enumerate}
    \item [(i)] We will introduce the concept of stochastic processes in the discrete state space as \textbf{continuous-time Markov chains} (CTMCs), where the control drift in SDEs takes the analogous form of a \textit{transition rate matrix} in discrete state spaces.
    \item[(ii)] While the structure of the discrete Schrödinger bridge problem remains the same, we will introduce fundamental differences in the KL divergence in discrete state spaces. This will provide the theoretical grounding for our discussion on solving the discrete Schrödinger bridge problem with generative modeling, which adapts several ideas developed in the previous sections for the discrete state space.
\end{enumerate}

In doing so, we extend the Schrödinger bridge framework from diffusion processes on continuous spaces to jump processes on discrete state spaces, laying the groundwork for an even broader class of generative modeling methods.

\newpage
\section{From Continuous to Discrete State Space}
\label{sec:discrete-state-space}
Now that we have built the foundation required to understand and construct the Schrödinger bridge where the states exist as continuous latent vectors in some state space $\mathcal{X}\subseteq \mathbb{R}^d$, we will now take a detour to the \textbf{discrete state space}, where states exist as probabilities of existing in a finite set of discrete states. We will see that rather than representing dynamics with velocity fields that transport states via smooth lines, dynamics in the discrete state spcae are represented with \textbf{transition rates} that characterize the instantaneous change in the probabilities of existing in each state. 

In this section, we introduce discrete state path measures not as SDEs but as \textbf{continuous-time Markov chains} (CTMCs) defined by their rate matrices (Section \ref{subsec:ctmc}). Leveraging the theory of CTMCs, we define the \textbf{discrete Schrödinger bridge problem}, and extend the definitions for the Radon-Nikodym derivative (RND) and KL divergence to CTMCs (Section \ref{subsec:discrete-sbp}). Then, we analyze two methods of solving the discrete SB problem which mirror the continuous state space, starting with the stochastic optimal control formulation (Section \ref{subsec:soc-ctmcs} and \ref{subsec:sb-soc}) and concluding with the discrete analog of Iterative Markovian Fitting using Markovian and reciprocal projections (Sections \ref{subsec:discrete-markov-reciprocal} and \ref{subsec:ddsbm}). 

\subsection{Continuous-Time Markov Chains}
\label{subsec:ctmc}
The discrete state space can be defined as a finite set of states $\mathcal{X}=\{1, \dots, d\}$ and the probability simplex $\Delta^{d-1}$ over the $d$ discrete states given by:
\begin{align}
    \Delta^{d-1}=\left\{\boldsymbol{x}=(x_1, \dots, x_d)\in \mathbb{R}^d\bigg|x_i\in [0,1], \sum_{i=1}^dx_i=1\right\}
\end{align}
To define a stochastic path measure $\mathbb{P}$ that lies on the simplex $\Delta^{d-1}$, we introduce the theory of \boldtext{continuous-time Markov chains} (CTMCs) in the probability space $(\Omega, \text{Pr})$, where $\Omega \in D([0,T], \mathcal{X})$ is the space of left-limited and right-continuous (cádlág)\footnote{piecewise continuous paths with jump discontinuities where the right limit is the actual value.} paths over $\mathcal{X}$ and $\text{Pr}$ is the probability measure over events. A CTMC is a stochastic process that evolves over time $\boldsymbol{X}_{0:T}$ whose probability law is defined by a time-dependent \boldtext{generator} or \boldtext{transition rate matrix} $(\boldsymbol{Q}_t\in \mathbb{R}^{\mathcal{X}\times \mathcal{X}})_{t\in [0,T]}$ of the form:
\begin{align}
    \boldsymbol{Q}_t(\boldsymbol{x},\boldsymbol{y})=\lim_{\Delta t\to 0}\frac{1}{\Delta t}\left(\text{Pr}(\boldsymbol{X}_{t+\Delta t}=\boldsymbol{y}|\boldsymbol{X}_t=\boldsymbol{x})-\boldsymbol{1}_{\boldsymbol{x}=\boldsymbol{y}}\right)
\end{align}
which defines the instantaneous rate of transitioning from state $\boldsymbol{x}\in \mathcal{X}$ to state $\boldsymbol{y}\in \mathcal{X}$ at time $t$. Since the state transitions must remain on the probability simplex, the transition rates satisfy the following conditions:
\begin{align}
 \forall \boldsymbol{x}\neq\boldsymbol{y}, \quad \boldsymbol{Q}_t(\boldsymbol{x},\boldsymbol{y})\geq 0, \quad\sum_{\boldsymbol{y}\in \mathcal{X}}\boldsymbol{Q}_t(\boldsymbol{x},\boldsymbol{y})=0
\end{align}
A generator $\boldsymbol{Q}$ \textit{uniquely} defines a path measure $\mathbb{P}\in \mathcal{P}(\Omega)$, under which we can define the transition probability over a discrete time interval $[t, t+\Delta t]$ as:
\begin{align}
    \mathbb{P}(\boldsymbol{X}_{t+\Delta t}=\boldsymbol{y}|\boldsymbol{X}_t=\boldsymbol{x})=\begin{cases}
        \Delta t\boldsymbol{Q}_t(\boldsymbol{x}, \boldsymbol{y}) +\mathcal{O}(\Delta t^2)&\boldsymbol{y}\neq \boldsymbol{x}\\
        1-\Delta t\sum_{\boldsymbol{z}\neq\boldsymbol{x}}\boldsymbol{Q}_t(\boldsymbol{x},\boldsymbol{z})+\mathcal{O}(\Delta t^2)&\boldsymbol{y}=\boldsymbol{x}\\
    \end{cases}\label{eq:ctmc-transition-prob}
\end{align}
which can be derived from the definition of the generator as the instantaneous jump rate, and multiplying that by the time interval $\Delta t$ for all states $\boldsymbol{y}\neq \boldsymbol{x}$ with $\mathcal{O}(\Delta t^2)$ accounting for the error of multiple jumps within the discrete interval. To ensure that the probabilities are normalized and sum to one, i.e., $\sum_{\boldsymbol{y}\in \mathcal{X}}\mathbb{P}(\boldsymbol{X}_{t+\Delta t}=\boldsymbol{y}|\boldsymbol{X}_t=\boldsymbol{x})=1$, we subtract all probabilities of leaving the state $\boldsymbol{x}$ as to get the probability of remaining at $\boldsymbol{x}$.

\begin{figure}
    \centering
    \includegraphics[width=\linewidth]{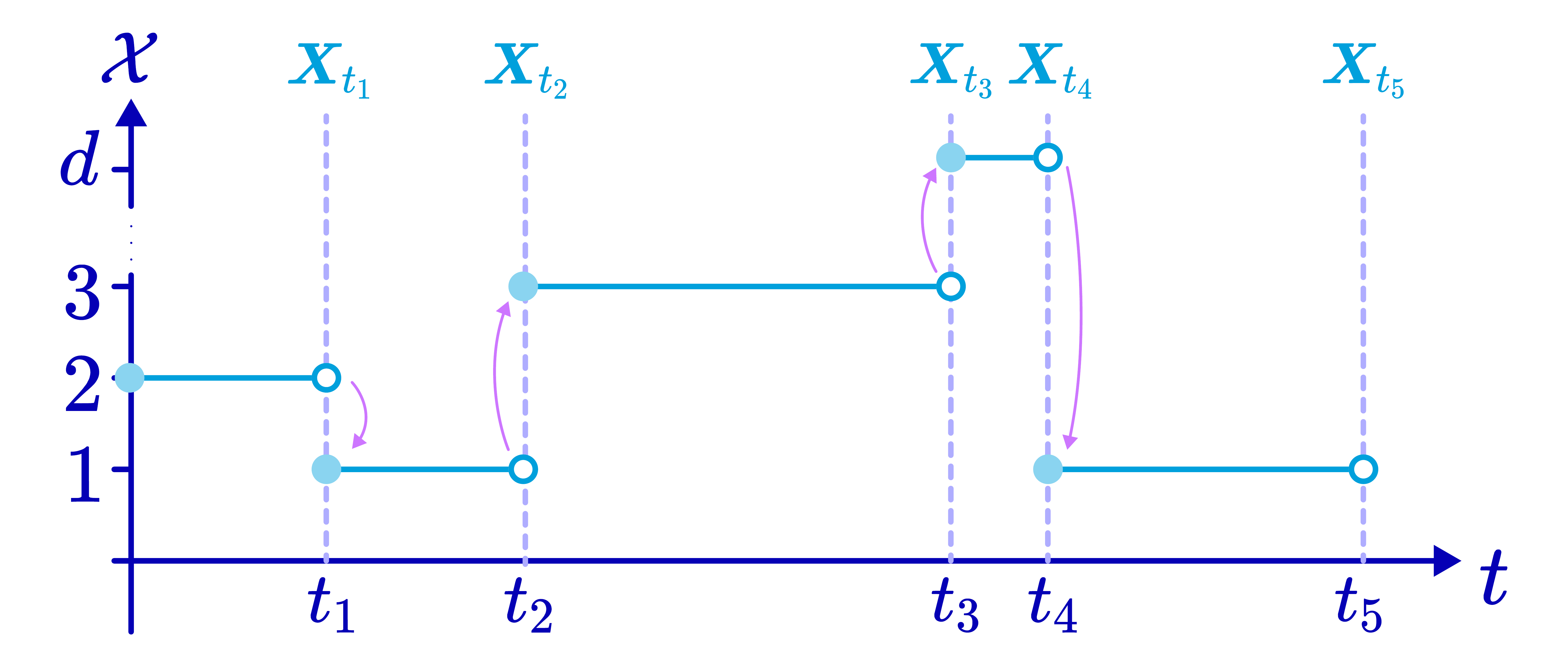}
    \caption{\textbf{Piecewise Càdlàg Path of Continuous-Time Markov Chains.} Illustration of a càdlàg stochastic trajectory, which is right-continuous with left limits. Between jump times $t_1, \dots, t_5$ the process evolves continuously, while discontinuities occur at the marked times. The filled markers indicate the value of the process $\boldsymbol{X}_t$ (the right limit), while the open markers denote the left limit $\boldsymbol{X}_{t^-}$.}
    \label{fig:ctmc-2}
\end{figure}

Given that CTMCs are cádlág paths, we denote the left limit of a state $\boldsymbol{X}_t$ as $\boldsymbol{X}_{t^-}=\lim_{s\uparrow t}\boldsymbol{X}_s$, where $\boldsymbol{X}_{t^-}\neq \boldsymbol{X}_t$ at jump times from state $\boldsymbol{X}_{t^-}$ to state $\boldsymbol{X}_t$. A \textbf{key property} of CTMCs is that they satisfy the \boldtext{Kolmogorov forward equation} which define the time evolution of the path.

\begin{lemma}[Kolmogorov Forward Equation for CTMCs]\label{lemma:ctmc-kolmogorov-fwd}
 The forward-time dynamics of a CTMC $\boldsymbol{X}_{0:T}$ with probability measure $p_t(\cdot):=\text{Pr}(\boldsymbol{X}_t=\cdot)$ and generator $\boldsymbol{Q}_t$ satisfies the \textbf{Kolmogorov forward equation} defined as:
 \begin{align}
     \forall \boldsymbol{x}\in \mathcal{X}, \quad \partial_tp_t(\boldsymbol{x})=\sum_{\boldsymbol{y}\in \mathcal{X}}\boldsymbol{Q}_t(\boldsymbol{x},\boldsymbol{y})p_t(\boldsymbol{y})=\sum_{\boldsymbol{x}\neq\boldsymbol{y}}\left(\boldsymbol{Q}_t(\boldsymbol{y}, \boldsymbol{x})p_t(\boldsymbol{y})-\boldsymbol{Q}_t(\boldsymbol{x},\boldsymbol{y})p_t(\boldsymbol{x})\right)\label{eq:ctmc-kolmogorov-fwd}
 \end{align}
 where that the probability measure $p_t(\cdot)$ is \textbf{unique} given a pair of endpoint conditions at $t\in \{0,T\}$ and $t\mapsto \boldsymbol{Q}_t$ is continuous over time $t\in [0,T]$.
\end{lemma}
\textit{Proof.} This proposition can be shown simply by defining a forward transition probability over the discrete time increment $[t,t+\Delta t]$ and taking a limit as $\Delta t\to 0$ to get the expression for $\partial_tp_t$. First, using (\ref{eq:ctmc-transition-prob}) we have:
\begin{small}
\begin{align}
    p_{t+\Delta t}(\boldsymbol{x}) &=\sum_{\boldsymbol{y}\in \mathcal{X}}\bluetext{\text{Pr}(\boldsymbol{X}_{t+\Delta t}=\boldsymbol{x}|\boldsymbol{X}_t=\boldsymbol{y})}p_t(\boldsymbol{y})\nonumber\\
    &\overset{(\ref{eq:ctmc-transition-prob})}{=}\sum_{\boldsymbol{y}\in \mathcal{X}}\bluetext{(\boldsymbol{1}_{\boldsymbol{x}=\boldsymbol{y}}+\Delta t\boldsymbol{Q}_t(\boldsymbol{x},\boldsymbol{y})+\mathcal{O}(\Delta t^2) )}p_t(\boldsymbol{y})\nonumber\\
    &=p_t(\boldsymbol{x})+\Delta t\sum_{\boldsymbol{y}\in \mathcal{X}}\boldsymbol{Q}_t(\boldsymbol{y}, \boldsymbol{x})p_t(\boldsymbol{y})+\mathcal{O}(\Delta t^2)
\end{align}
\end{small}
Taking the continuous time limit as $\Delta t\to 0$, we have:
\begin{small}
\begin{align}
\partial_tp_t(\boldsymbol{x})&=\lim_{\Delta t\to 0}\left[\Delta t\sum_{\boldsymbol{y}\in \mathcal{X}}\boldsymbol{Q}_t(\boldsymbol{y},\boldsymbol{x})p_t(\boldsymbol{y})+\mathcal{O}(\Delta t^2)\right]=\sum_{\boldsymbol{y}\in \mathcal{X}}\boldsymbol{Q}_t(\boldsymbol{y},\boldsymbol{x})p_t(\boldsymbol{y})\nonumber\\
&=\boldsymbol{Q}_t(\boldsymbol{x},\boldsymbol{x})p_t(\boldsymbol{x})+\sum_{\boldsymbol{y}\neq \boldsymbol{x}}\boldsymbol{Q}_t(\boldsymbol{y},\boldsymbol{x})p_t(\boldsymbol{y})=-\sum_{\boldsymbol{y}\neq \boldsymbol{x}}\boldsymbol{Q}_t(\boldsymbol{y},\boldsymbol{x})p_t(\boldsymbol{y})+\sum_{\boldsymbol{y}\neq \boldsymbol{x}}\boldsymbol{Q}_t(\boldsymbol{y},\boldsymbol{x})p_t(\boldsymbol{y})\nonumber\\
&=\sum_{\boldsymbol{x}\neq\boldsymbol{y}}\left(\boldsymbol{Q}_t(\boldsymbol{y}, \boldsymbol{x})p_t(\boldsymbol{y})-\boldsymbol{Q}_t(\boldsymbol{x},\boldsymbol{y})p_t(\boldsymbol{x})\right)\label{eq:ctmc-kolmogorov-fwd-proof}
\end{align}
\end{small}
which is exactly the Kolmogorov forward equation defined in (\ref{eq:ctmc-kolmogorov-fwd}). To prove \textbf{uniqueness}, we can write (\ref{eq:ctmc-kolmogorov-fwd-proof}) in vector form as: 
\begin{align}
    \forall t\in [0,T ]: \quad-\partial_t\boldsymbol{p}_t=\boldsymbol{Q}_t\boldsymbol{p}_t, \quad \text{s.t.}\quad \boldsymbol{p}_t=(p_t(\boldsymbol{x}):\boldsymbol{x}\in \mathcal{X})\in \mathbb{R}^{|\mathcal{X}|}
\end{align}
which is a linear ODE in $\mathbb{R}^{|\mathcal{X}|}$. Since $t\mapsto \boldsymbol{Q}_t$ is continuous, linear ODEs have a unique solution.\hfill $\square$

Having derived the Kolmogorov forward equation, which describes how the state distribution $p_t$ evolves \textit{forward} under the time-dependent generator $\boldsymbol{Q}_t$, we now ask how CTMC dynamics evolve \textit{backwards} from a terminal constraint. As we have seen in the continuous state space, the idea of terminal conditioning is the foundation for solving the Schrödinger bridge problem, as it allows us to define the optimal dynamics that generate a target distribution. The idea of evolving CTMC dynamics backward is exactly captured by the \boldtext{Kolmogorov backward equation}, which we define next.

\begin{figure}
    \centering
    \includegraphics[width=\linewidth]{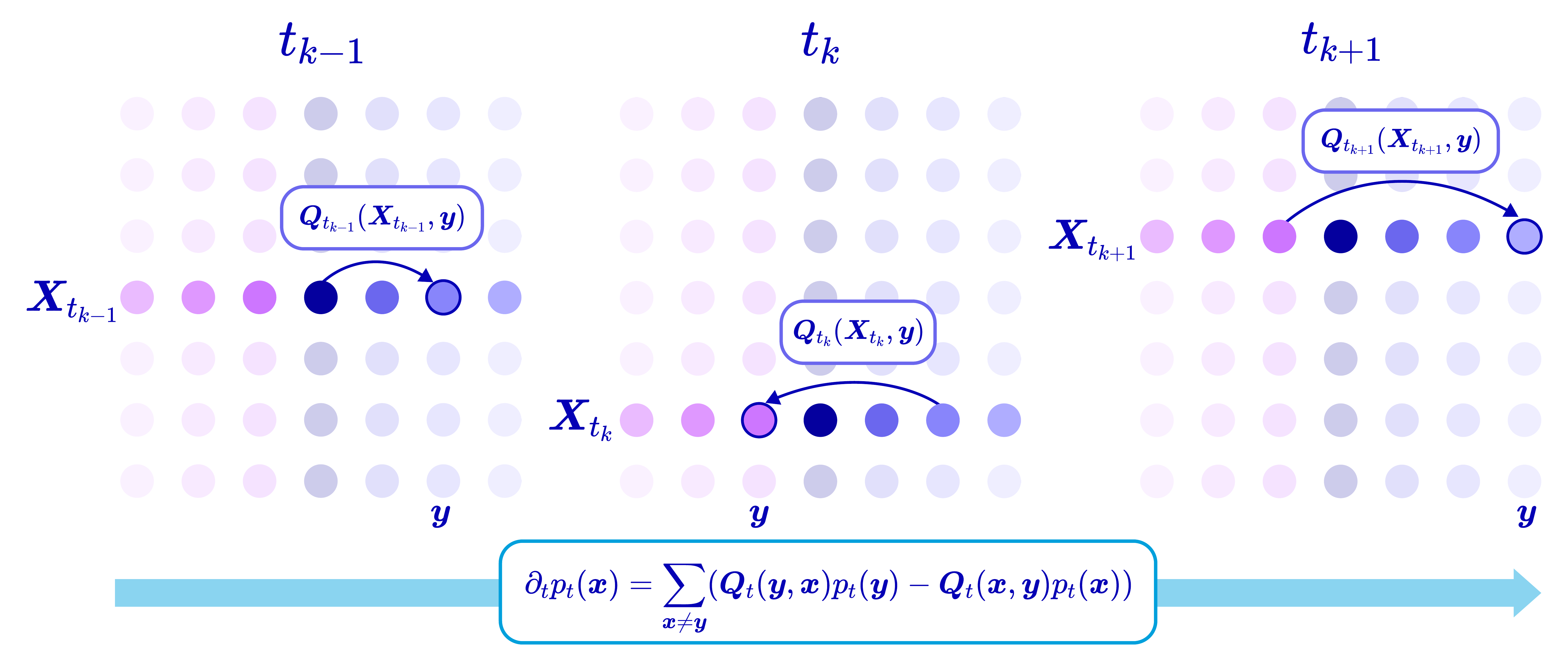}
    \caption{\textbf{Continuous-Time Markov Chains (CTMC).} A CTMC defines a stochastic process on a finite state space $\mathcal{X}=\{1, \dots d\}$ that evolves through jumps between discrete states governed by a time-dependent transition rate matrix $\boldsymbol{Q}_t$. At each jump time $t_k$, the system transitions from the current state $\boldsymbol{X}_{t_k}$ to a different state $\boldsymbol{y}$ according to the rate $\boldsymbol{Q}_t(\boldsymbol{X}_{t_k}, \boldsymbol{y})$. The evolution of the probability distribution $p_t(\boldsymbol{x})$ satisfies the Kolmogorov forward equation, which balances incoming and outgoing probability flows induced by the transition rates.}
    \label{fig:ctmc}
\end{figure}

\begin{lemma}[Kolmogorov Backward Equation for CTMCs]\label{lemma:ctmc-kolmogorov-bwd}
 Consider a CTMC $\boldsymbol{X}_{0:T}$ under the measure $\mathbb{P}$ with generator $\boldsymbol{Q}_t$ and let $\phi_t(\boldsymbol{x}):=\mathbb{E}[\phi(\boldsymbol{X}_T)|\boldsymbol{X}_t=\boldsymbol{x}]$ given an arbitrary test function $\phi: \mathcal{X}\to \mathbb{R}$. Then, the reverse-time dynamics of $\phi_t$ satisfies the \textbf{Kolmogorov backward equation} defined as:
 \begin{small}
 \begin{align}
     \forall \boldsymbol{x}\in \mathcal{X}, \quad-\partial_t\phi(\boldsymbol{x})=\sum_{\boldsymbol{y}\in \mathcal{X}}\phi_t(\boldsymbol{y})\boldsymbol{Q}_t(\boldsymbol{x}, \boldsymbol{y})=\sum_{\boldsymbol{y}\neq \boldsymbol{x}}(\phi(\boldsymbol{y})-\phi(\boldsymbol{x}))\boldsymbol{Q}_t(\boldsymbol{x}, \boldsymbol{y}), \quad \phi_T(\boldsymbol{x})=\phi(\boldsymbol{x})\label{eq:ctmc-kolmogorov-bwd}
 \end{align}
 \end{small}
 which admits a \textbf{unique} solution $\phi$ when $t\mapsto \boldsymbol{Q}_t$ is continuous over all $t\in [0,T]$.
\end{lemma}

\textit{Proof.} Since $\phi_t(\boldsymbol{x})$ can be considered some cost function to go from $\boldsymbol{X}_t=\boldsymbol{x}$ to the terminal state $\boldsymbol{X}_T$, we can expand the inner function $\phi(\boldsymbol{X}_T)$ using the law of total expectation to write $\phi_t(\boldsymbol{x})$ with respect to a discrete time step $\phi_{t+\Delta t}(\boldsymbol{x})$: 
\begin{align}
    \phi_t(\boldsymbol{x})&=\mathbb{E}[\bluetext{\phi(\boldsymbol{X}_T)}|\boldsymbol{X}_t=\boldsymbol{x}]=\mathbb{E}[\bluetext{\mathbb{E}[\phi(\boldsymbol{X}_T)|\boldsymbol{X}_{t+\Delta t}]}|\boldsymbol{X}_t=\boldsymbol{x}]=\mathbb{E}[\bluetext{\phi_{t+\Delta t}(\boldsymbol{X}_{t+\Delta t})}|\boldsymbol{X}_t=\boldsymbol{x}]
\end{align}
Then, applying the transition probability of a discrete time step defined in (\ref{eq:ctmc-transition-prob}), we have:
\begin{small}
\begin{align}
    \phi_t(\boldsymbol{x})&=\mathbb{E}[\bluetext{\phi_{t+\Delta t}(\boldsymbol{X}_{t+\Delta t})}|\boldsymbol{X}_t=\boldsymbol{x}]=\sum_{\boldsymbol{y}\in \mathcal{X}}\phi(\boldsymbol{y})\bluetext{\mathbb{P}(\boldsymbol{X}_{t+\Delta t}=\boldsymbol{y}|\boldsymbol{X}_t=\boldsymbol{x})}\nonumber\\
    &=\sum_{\boldsymbol{y}\in \mathcal{X}}\phi_{t+\Delta t}(\boldsymbol{y})\left(\bluetext{\boldsymbol{1}_{\boldsymbol{x}=\boldsymbol{y}}+\Delta t\boldsymbol{Q}_t(\boldsymbol{x}, \boldsymbol{y})+\mathcal{O}(\Delta t^2)}\right)\nonumber\\
    &=\phi_{t+\Delta t}(\boldsymbol{x})+\Delta t\sum_{\boldsymbol{y}\in \mathcal{X}}\phi_{t+\Delta t}(\boldsymbol{y})\boldsymbol{Q}_t(\boldsymbol{x}, \boldsymbol{y})+\mathcal{O}(\Delta t^2)
\end{align}
\end{small}
Now, subtracting $\phi_{t+\Delta t}(\boldsymbol{x})$ from both sides, dividing by $\Delta t$,  and taking the continuous time limit $\Delta t\to 0$, we get the reverse time dynamics $-\partial_t\phi_t(\boldsymbol{x})$ as:
\begin{small}
\begin{align}
    -\partial_t\phi_t(\boldsymbol{x})&=\lim_{\Delta t\to 0}\left[\frac{\phi_t(\boldsymbol{x})-\phi_{t+\Delta t}(\boldsymbol{x})}{\Delta t}\right]=\sum_{\boldsymbol{y}\in \mathcal{X}}\phi_{t}(\boldsymbol{y})\boldsymbol{Q}_t(\boldsymbol{x}, \boldsymbol{y})=\sum_{\boldsymbol{y}\neq \boldsymbol{x}}\phi_t(\boldsymbol{y})\boldsymbol{Q}_t(\boldsymbol{x}, \boldsymbol{y})+\bluetext{\phi_t(\boldsymbol{x})\boldsymbol{Q}_t(\boldsymbol{x}, \boldsymbol{x})}\nonumber\\
    &=\sum_{\boldsymbol{y}\neq \boldsymbol{x}}\phi_t(\boldsymbol{y})\boldsymbol{Q}_t(\boldsymbol{x}, \boldsymbol{y})\bluetext{-\sum_{\boldsymbol{y}\neq \boldsymbol{x}}\phi_t(\boldsymbol{x})\boldsymbol{Q}_t(\boldsymbol{x}, \boldsymbol{y})}=\sum_{\boldsymbol{y}\neq \boldsymbol{x}}(\phi_t(\boldsymbol{y})-\phi_t(\boldsymbol{x}))\boldsymbol{Q}_t(\boldsymbol{x}, \boldsymbol{y})\label{eq:ctmc-kolmogorov-bwd-proof}
\end{align}
\end{small}
which is the Kolmogorov backward equation. To prove \textbf{uniqueness}, we follow the proof of the forward equation and write (\ref{eq:ctmc-kolmogorov-bwd-proof}) in vector form as $-\partial_t\boldsymbol{\phi}_t=\boldsymbol{Q}_t\boldsymbol{\phi}_t$, which is a linear ODE in $\mathbb{R}^{|\mathcal{X}|}$. Since $t\mapsto \boldsymbol{Q}_t$ is continuous, linear ODEs have a unique solution. \hfill $\square$

CTMCs can be interpreted as the \textbf{discrete analog of stochastic differential equations (SDEs)} in the continuous state space, where the time-dependent generator conditioned on generator $\boldsymbol{Q}_t(\boldsymbol{x}, \cdot):\mathcal{X}\times[0,T]\to \mathbb{R}^d$ is analogous to the time-dependent \textit{drift} $\boldsymbol{u}(\boldsymbol{x},t)$ of an SDE. Since we have characterized the forward and reverse evolution of path-measure CTMCs, we are ready to formulate the discrete-state-space analog of the Schrödinger bridge problem, which aims to recover a CTMC that satisfies a pair of marginal distributions while remaining close to a reference CTMC.

\subsection{Discrete Schrödinger Bridge Problem}
\label{subsec:discrete-sbp}
In this section, we formulate the \boldtext{discrete Schrödinger bridge problem} for continuous-time Markov chains (CTMCs) on a finite state space $\mathcal{X}=\{1, \dots , d\}$. Just like the continuous-space Schrödinger bridge problem, the objective is to find a path measure $\mathbb{P}^\star$ that is closest, in relative entropy or KL divergence, to a given reference process $\mathbb{Q}$, while matching prescribed marginal constraints $p_0=\pi_0$ and $p_T=\pi_T$. 

Since the \textit{control drift} in continuous SB theory is denoted $\boldsymbol{u}(\boldsymbol{x},t):\mathbb{R}^d\times[0,T]\to\mathbb{R}^d$, we use $\boldsymbol{Q}^u_t(\boldsymbol{x}, \cdot):\mathcal{X}\times[0,T]\to \mathbb{R}^{d\times d}$ to denote the CTMC generator of the \textit{controlled path measure} $\mathbb{P}^u\in \mathcal{P}(C([0,T]; \mathcal{X}))$, which we aim to optimize to recover the Schrödinger bridge. To denote the generator of the \textit{reference process} $\mathbb{Q}\in \mathcal{P}(C([0,T]; \mathcal{X}))$, we use $\boldsymbol{Q}^0_t(\boldsymbol{x}, \cdot):\mathcal{X}\times[0,T]\to \mathbb{R}^{d\times d}$. 

\begin{definition}[Discrete Schrödinger Bridge Problem]\label{def:discrete-sbp}
    Consider a reference CTMC measure $\mathbb{Q}$ with generator $\boldsymbol{Q}^0_t$ and a pair of marginal distributions in the discrete state space $\pi_0, \pi_T\in \mathcal{P}(\mathcal{X})$. The \textbf{discrete Schrödinger bridge problem} aims to determine the optimal CTMC path measure $\mathbb{P}^\star$ with generator $\boldsymbol{Q}^\star_t$ where $p_0=\pi_0$ and $p_T=\pi_T$ that solves the minimization problem: 
    \begin{align}
        \mathbb{P}^\star=\underset{\mathbb{P}^u\in \mathcal{P}(C([0,T];\mathcal{X})}{\arg\min}\big\{\text{KL}(\mathbb{P}^u\|\mathbb{Q}):p_0=\pi_0, p_T=\pi_T\big\}\tag{Discrete SB Problem}\label{eq:discrete-sb-problem}
    \end{align}
    where the KL divergence $\text{KL}(\cdot\|\cdot)$ between CTMCs is defined as: \begin{small}
    \begin{align}
    \text{KL}(\mathbb{P}^u\|\mathbb{Q})=\mathbb{E}_{\boldsymbol{X}_{0:T}\sim \mathbb{P}^u}\int_0^T\left[\sum_{\boldsymbol{y}\neq \boldsymbol{X}_t}\left(\boldsymbol{Q}_t^u\log \frac{\boldsymbol{Q}^u_t}{\boldsymbol{Q}^0_t}+\boldsymbol{Q}^0_t-\boldsymbol{Q}^u_t\right)(\boldsymbol{X}_t,\boldsymbol{y})\right]dt
    \end{align}
    \end{small}
\end{definition}

Solving the discrete SB problem yields the CTMC path measure that minimizes the discrepancy from the reference jump process, which penalizes differences in jump intensities, jump times, and transition structure over the time interval $t\in [0,T]$.

To define the form of the KL divergence, we first derive the \boldtext{Radon-Nikodym derivative between CTMC path measures}, which defines the probability ratio of a CTMC $\boldsymbol{X}_{0:T}$ under two path measures $\mathbb{P}$ and $\mathbb{P}'$ with \textbf{generators} $\boldsymbol{Q}_t$ and $\boldsymbol{Q}_t'$. 

\begin{proposition}[Radon-Nikodym Derivative Between CTMCs]\label{prop:ctmc-rnd}
    Consider two CTMC path measures $\mathbb{P}$ and $\mathbb{P}'$ with generators $\boldsymbol{Q}$ and $\boldsymbol{Q}_t'$ and initial distributions $p_0=\pi_0$ and $\mathbb{P}'_0=\pi_0'$. Assume that the $\pi'_0\ll\pi_0$ and $\mathbb{P}'\ll \mathbb{P}$, where the generators satisfy $\boldsymbol{Q}(\boldsymbol{x},\boldsymbol{y})=0\implies \boldsymbol{Q}(\boldsymbol{x},\boldsymbol{y})=0$. Then, the logarithm of the Radon-Nikodym derivative is given by:
    \begin{small}
    \begin{align}
        \log \frac{\mathrm{d}\mathbb{P}'}{\mathrm{d}\mathbb{P}}(\boldsymbol{X}_{0:T})=\log \frac{\mathrm{d}\pi_0'}{\mathrm{d}\pi_0}(\boldsymbol{X}_0)+\sum_{t:\boldsymbol{X}_{t^-}\neq \boldsymbol{X}_t}\log \frac{\boldsymbol{Q}_t'(\boldsymbol{X}_{t^-}, \boldsymbol{X}_t)}{\boldsymbol{Q}_t(\boldsymbol{X}_{t^-}, \boldsymbol{X}_t)}+\int_0^T\sum_{\boldsymbol{y}\neq\boldsymbol{X}_t}(\boldsymbol{Q}_t-\boldsymbol{Q}'_t)(\boldsymbol{X}_t,\boldsymbol{y})\mathrm{d}t
    \end{align}
    \end{small}
\end{proposition}

\textit{Proof.} To derive the RND for CTMCs, we first consider the \textbf{discrete-time case}, where we break down the time horizon into time steps $0=t_0< t_1< \dots< t_k< \dots< t_{K-1}< t_K=T$ where the time intervals are separated by $\Delta t$. Then, we can write the log ratio of a discrete path $(\boldsymbol{X}_{t_k})_{k\in \{0, \dots, K\}}$ under the probability measures $\mathbb{P}$ and $\mathbb{P}'$ as:
\begin{small}
\begin{align}
    \log \frac{\mathrm{d}\mathbb{P}'}{\mathrm{d}\mathbb{P}}(\boldsymbol{X}_{0:T})=\log \frac{\mathrm{d}\pi_0'}{\mathrm{d}\pi_0}(\boldsymbol{X}_0)+\sum_{k=0}^{K-1}\log \frac{\mathrm{d}\mathbb{P}'(\boldsymbol{X}_{t_{k+1}}|\boldsymbol{X}_{t_k})}{\mathrm{d}\mathbb{P}(\boldsymbol{X}_{t_{k+1}}|\boldsymbol{X}_{t_k})}+\mathcal{O}(\Delta t)\label{eq:ctmc-rnd-eq1}
\end{align}
\end{small}
From (\ref{eq:ctmc-transition-prob}), we can derive the probability ratio for two distinct cases given an interval $[t_k, t_{k+1}]$: the case of at least one change in state in $[t_k, t_{k+1}]$ and the case of no change in state in $[t_k, t_{k+1}]$. First, for the case when at least one jump is made, the log ratio becomes:
\begin{small}
\begin{align}
    \log \frac{\mathbb{P}'(\boldsymbol{X}_{t_{k+1}}|\boldsymbol{X}_{t_k})}{\mathbb{P}(\boldsymbol{X}_{t_{k+1}}|\boldsymbol{X}_{t_k})}=\log \frac{\bluetext{\Delta t\boldsymbol{Q}'_{t_k}(\boldsymbol{X}_{t_k}, \boldsymbol{X}_{t_{k+1}}) +\mathcal{O}(\Delta t^2)}}{\pinktext{\Delta t\boldsymbol{Q}_{t_k}(\boldsymbol{X}_{t_k}, \boldsymbol{X}_{t_{k+1}}) +\mathcal{O}(\Delta t^2)}}=\log \frac{\bluetext{\boldsymbol{Q}'_{t_k}(\boldsymbol{X}_{t_k}, \boldsymbol{X}_{t_{k+1}})}}{\pinktext{\boldsymbol{Q}_{t_k}(\boldsymbol{X}_{t_k}, \boldsymbol{X}_{t_{k+1}})}}+\mathcal{O}(\Delta t^2)\label{eq:ctmc-rnd-case1}
\end{align}
\end{small}
For the case when no jumps are made, the log ratio becomes:
\begin{small}
\begin{align}
    \log \frac{\mathbb{P}'(\boldsymbol{X}_{t_{k+1}}|\boldsymbol{X}_{t_k})}{\mathbb{P}(\boldsymbol{X}_{t_{k+1}}|\boldsymbol{X}_{t_k})}&=\log \frac{\bluetext{1-\Delta t\sum_{\boldsymbol{y}\neq\boldsymbol{X}_{t_k}}\boldsymbol{Q}'_{t_k}(\boldsymbol{X}_{t_k},\boldsymbol{y})+\mathcal{O}(\Delta t^2)}}{\pinktext{1-\Delta t\sum_{\boldsymbol{y}\neq\boldsymbol{X}_{t_k}}\boldsymbol{Q}_{t_k}(\boldsymbol{X}_{t_k},\boldsymbol{y})+\mathcal{O}(\Delta t^2)}}\nonumber\\
    &=\Delta t\sum_{\boldsymbol{y}\neq \boldsymbol{X}_{t_k}}\left(\pinktext{\boldsymbol{Q}_{t_k}(\boldsymbol{X}_{t_k}, \boldsymbol{y})}-\bluetext{\boldsymbol{Q}'_{t_k}(\boldsymbol{X}_{t_k}, \boldsymbol{y})}\right)+\mathcal{O}(\Delta t^2)\label{eq:ctmc-rnd-case2}
\end{align}
\end{small}
Substituting (\ref{eq:ctmc-rnd-case1}) and (\ref{eq:ctmc-rnd-case2}) into (\ref{eq:ctmc-rnd-eq1}) and taking the continuous time limit $\Delta t\to 0$ and $K\to \infty$, we have:
\begin{small}
\begin{align}
    \log \frac{\mathrm{d}\mathbb{P}'}{\mathrm{d}\mathbb{P}}(\boldsymbol{X}_{0:T})&=\lim_{\Delta t\to 0}\bigg\{\log \frac{\mathrm{d}\pi_0'}{\mathrm{d}\pi_0}(\boldsymbol{X}_0)+\sum_{k=0}^{K-1}\log \frac{\boldsymbol{Q}'_{t_k}(\boldsymbol{X}_{t_k}, \boldsymbol{X}_{t_{k+1}})}{\boldsymbol{Q}_{t_k}(\boldsymbol{X}_{t_k}, \boldsymbol{X}_{t_{k+1}})}\nonumber\\
    &+\sum_{k=0}^{K-1}\Delta t\sum_{\boldsymbol{y}\neq \boldsymbol{X}_{t_k}}\left(\boldsymbol{Q}_{t_k}(\boldsymbol{X}_{t_k}, \boldsymbol{y})-\bluetext{\boldsymbol{Q}'_{t_k}(\boldsymbol{X}_{t_k}, \boldsymbol{y})}\right)+\mathcal{O}(\Delta t)\bigg\}\nonumber\\
    &=\log \frac{\mathrm{d}\pi_0'}{\mathrm{d}\pi_0}(\boldsymbol{X}_0)+\sum_{t:\boldsymbol{X}_{t^-}\neq \boldsymbol{X}_t}\log \frac{\boldsymbol{Q}_t'(\boldsymbol{X}_{t^-}, \boldsymbol{X}_t)}{\boldsymbol{Q}_t(\boldsymbol{X}_{t^-}, \boldsymbol{X}_t)}+\int_0^T\sum_{\boldsymbol{y}\neq\boldsymbol{X}_t}(\boldsymbol{Q}_t-\boldsymbol{Q}'_t)(\boldsymbol{X}_t,\boldsymbol{y})\mathrm{d}t
\end{align}
\end{small}
which is the exact form of the log RND between $\mathbb{P}'$ and $\mathbb{P}$. \hfill $\square$

Using this result, we can easily derive the \boldtext{KL divergence between CTMC path measures}, which is the foundation of the discrete Schrödinger bridge objective.

\begin{corollary}[KL Divergence Between CTMCs]\label{corollary:ctmc-kl}
    The KL divergence between two CTMC path measures $\mathbb{P}$ and $\mathbb{P}'$ with generators $\boldsymbol{Q}$ and $\boldsymbol{Q}_t'$ is defined as:
    \begin{small}
    \begin{align}
        \text{KL}(\mathbb{P}'\|\mathbb{P})=\text{KL}(\pi_0'\|\pi_0)+\mathbb{E}_{\boldsymbol{X}_{0:T}\sim \mathbb{P}'}\int_0^T\left[\sum_{\boldsymbol{y}\neq \boldsymbol{X}_t}\left(\boldsymbol{Q}_t'\log \frac{\boldsymbol{Q}'_t}{\boldsymbol{Q}_t}+\boldsymbol{Q}_t-\boldsymbol{Q}_t'\right)(\boldsymbol{X}_t,\boldsymbol{y})\right]dt\label{eq:kl-ctmc-1}
    \end{align}
    \end{small}
    and can be equivalently written as:
    \begin{small}
    \begin{align}
        \text{KL}(\mathbb{P}'\|\mathbb{P})
        &=\text{KL}(\pi'_0\|\pi_0)+\mathbb{E}_{\boldsymbol{X}_{0:T}\sim \mathbb{P}'}\int_0^T\left[\sum_{\boldsymbol{y}\neq \boldsymbol{X}_t}\left(\boldsymbol{Q}_t'\log \frac{\boldsymbol{Q}_{t}'}{\boldsymbol{Q}_{t}}\right)(\boldsymbol{X}_t, \boldsymbol{y})+(\boldsymbol{Q}'_t-\boldsymbol{Q}_t)(\boldsymbol{X}_t,\boldsymbol{X}_t)\right]dt\label{eq:kl-ctmc-2}
    \end{align}
    \end{small}
\end{corollary}

\textit{Proof.} Recalling that the KL divergence $\text{KL}(\mathbb{P}'\|\mathbb{P})$ is simply the expectation of the log RND over paths from the first argument, we have:
\begin{small}
\begin{align}
    &\text{KL}(\mathbb{P}'\|\mathbb{P})=\mathbb{E}_{\boldsymbol{X}_{0:T}\sim \mathbb{P}'}\left[\log \frac{\mathrm{d}\mathbb{P}'}{\mathrm{d}\mathbb{P}}\right]\nonumber\\
    &=\mathbb{E}_{\boldsymbol{X}_{0:T}\sim \mathbb{P}'}\left[\bluetext{\log \frac{\mathrm{d}\pi_0'}{\mathrm{d}\pi_0}(\boldsymbol{X}_0)+\sum_{t:\boldsymbol{X}_{t^-}\neq \boldsymbol{X}_t}\log \frac{\boldsymbol{Q}_t'(\boldsymbol{X}_{t^-}, \boldsymbol{X}_t)}{\boldsymbol{Q}_t(\boldsymbol{X}_{t^-}, \boldsymbol{X}_t)}+\int_0^T\sum_{\boldsymbol{y}\neq\boldsymbol{X}_t}(\boldsymbol{Q}_t-\boldsymbol{Q}'_t)(\boldsymbol{X}_t,\boldsymbol{y})dt}\right]\nonumber\\
    &=\underbrace{\mathbb{E}_{\boldsymbol{X}_{0:T}\sim \mathbb{P}'}\left[\log \frac{\mathrm{d}\pi_0'}{\mathrm{d}\pi_0}(\boldsymbol{X}_0)\right]}_{(\bigstar)}+\underbrace{\mathbb{E}_{\boldsymbol{X}_{0:T}\sim \mathbb{P}'}\left[\sum_{t:\boldsymbol{X}_{t^-}\neq \boldsymbol{X}_t}\log \frac{\boldsymbol{Q}_t'(\boldsymbol{X}_{t^-}, \boldsymbol{X}_t)}{\boldsymbol{Q}_t(\boldsymbol{X}_{t^-}, \boldsymbol{X}_t)}\right]}_{(\blacklozenge)}+\underbrace{\mathbb{E}_{\boldsymbol{X}_{0:T}\sim \mathbb{P}'}\left[\int_0^T\sum_{\boldsymbol{y}\neq\boldsymbol{X}_t}(\boldsymbol{Q}_t-\boldsymbol{Q}'_t)(\boldsymbol{X}_t,\boldsymbol{y})dt\right]}_{(\diamond)}\label{eq:ctmc-kl-div1}
\end{align}
\end{small}
The first term ($\bigstar$) can be written as:
\begin{align}
    \mathbb{E}_{\bluetext{\boldsymbol{X}_{0:T}\sim \mathbb{P}'}}\underbrace{\left[\log \frac{\mathrm{d}\pi_0'}{\mathrm{d}\pi_0}(\boldsymbol{X}_0)\right]}_{\text{only depends on }\boldsymbol{X}_0}=\mathbb{E}_{\bluetext{\boldsymbol{X}_{0}\sim \pi'_0}}\left[\log \frac{\mathrm{d}\pi_0'}{\mathrm{d}\pi_0}(\boldsymbol{X}_0)\right]=\text{KL}(\pi'_0\|\pi_0)\tag{$\bigstar$}
\end{align}
To write the second term ($\blacklozenge$) as an integral, we can consider the discrete time case for $0=t_0< \dots< t_k< \dots< t_K=T$, distribute the expectation, and take the continuous time limit as follows:
\begin{align}
    \mathbb{E}_{(\boldsymbol{X}_{t_k})_{k\in \{0,\dots, K\}}\sim \mathbb{P}'}\bigg[&\sum_{k=0}^{K-1}\bluetext{\boldsymbol{1}[\boldsymbol{X}_{t_{k+1}}\neq \boldsymbol{X}_{t_k}]}\log \frac{\boldsymbol{Q}_{t_k}'(\boldsymbol{X}_{t_k}, \boldsymbol{X}_{t_{k+1}})}{\boldsymbol{Q}_{t_k}(\boldsymbol{X}_{t_k}, \boldsymbol{X}_{t_{k+1}})}\bigg]\nonumber\\
    &=\sum_{k=0}^{K-1}\bluetext{\mathbb{E}_{\mathbb{P}'(\boldsymbol{X}_{t_k}), \mathbb{P}'(\boldsymbol{X}_{t_{k+1}}|\boldsymbol{X}_{t_k})}}\left[\boldsymbol{1}[\boldsymbol{X}_{t_{k+1}}\neq \boldsymbol{X}_{t_k}]\log \frac{\boldsymbol{Q}_{t_k}'(\boldsymbol{X}_{t_k}, \boldsymbol{X}_{t_{k+1}})}{\boldsymbol{Q}_{t_k}(\boldsymbol{X}_{t_k}, \boldsymbol{X}_{t_{k+1}})}\right]\nonumber\\
    &=\sum_{k=0}^{K-1}\bluetext{\mathbb{E}_{\mathbb{P}'(\boldsymbol{X}_{t_k})}}\sum_{\boldsymbol{y}\neq \boldsymbol{X}_{t_k}}\left[\bluetext{\mathbb{P}'(\boldsymbol{y}|\boldsymbol{X}_{t_k})}\log \frac{\boldsymbol{Q}_{t_k}'(\boldsymbol{X}_{t_k}, \boldsymbol{y})}{\boldsymbol{Q}_{t_k}(\boldsymbol{X}_{t_k}, \boldsymbol{y})}\right]\nonumber\\
    &=\sum_{k=0}^{K-1}\bluetext{\mathbb{E}_{\mathbb{P}'(\boldsymbol{X}_{t_k})}}\sum_{\boldsymbol{y}\neq \boldsymbol{X}_{t_k}}\left[\bluetext{\Delta t\boldsymbol{Q}_{t_k}'(\boldsymbol{X}_{t_k}, \boldsymbol{y})}\log \frac{\boldsymbol{Q}_{t_k}'(\boldsymbol{X}_{t_k}, \boldsymbol{y})}{\boldsymbol{Q}_{t_k}(\boldsymbol{X}_{t_k}, \boldsymbol{y})}+\bluetext{\mathcal{O}(\Delta t^2)}\right]\nonumber\\
    &\underset{\bluetext{\Delta t\to 0}}{=}\mathbb{E}_{\boldsymbol{X}_{0:T}\sim \mathbb{P}'}\left[\int_0^T\sum_{\boldsymbol{y}\neq \boldsymbol{X}_t}\left(\boldsymbol{Q}_t'\log \frac{\boldsymbol{Q}_{t}'}{\boldsymbol{Q}_{t}}\right)(\boldsymbol{X}_t, \boldsymbol{y})\right]\tag{$\blacklozenge$}
\end{align}
Plugging these expressions for ($\bigstar$) and ($\blacklozenge$) into (\ref{eq:ctmc-kl-div1}), we get:
\begin{small}
\begin{align}
    \text{KL}(\mathbb{P}'\|\mathbb{P})&=\underbrace{\text{KL}(\pi'_0\|\pi_0)}_{(\bigstar)}+\underbrace{\mathbb{E}_{\boldsymbol{X}_{0:T}\sim \mathbb{P}'}\left[\int_0^T\sum_{\boldsymbol{y}\neq \boldsymbol{X}_t}\left(\bluetext{\boldsymbol{Q}_t'\log \frac{\boldsymbol{Q}_{t}'}{\boldsymbol{Q}_{t}}}\right)(\boldsymbol{X}_t, \boldsymbol{y})dt\right]}_{(\blacklozenge)}+\underbrace{\mathbb{E}_{\boldsymbol{X}_{0:T}\sim \mathbb{P}'}\left[\int_0^T\sum_{\boldsymbol{y}\neq\boldsymbol{X}_t}(\pinktext{\boldsymbol{Q}_t-\boldsymbol{Q}'_t})(\boldsymbol{X}_t,\boldsymbol{y})dt\right]}_{(\diamond)}\nonumber\\
    &=\text{KL}(\pi'_0\|\pi_0)+\mathbb{E}_{\boldsymbol{X}_{0:T}\sim \mathbb{P}'}\left[\sum_{\boldsymbol{y}\neq \boldsymbol{X}_t}\left(\bluetext{\boldsymbol{Q}_t'\log \frac{\boldsymbol{Q}_{t}'}{\boldsymbol{Q}_{t}}}+\pinktext{\boldsymbol{Q}_t-\boldsymbol{Q}'_t}\right)(\boldsymbol{X}_t, \boldsymbol{y})\right]dt
    &=\text{KL}(\pi'_0\|\pi_0)+\mathbb{E}_{\boldsymbol{X}_{0:T}\sim \mathbb{P}'}\left[\sum_{\boldsymbol{y}\neq \boldsymbol{X}_t}\left(\bluetext{\boldsymbol{Q}_t'\log \frac{\boldsymbol{Q}_{t}'}{\boldsymbol{Q}_{t}}}+\pinktext{\boldsymbol{Q}_t-\boldsymbol{Q}'_t}\right)(\boldsymbol{X}_t, \boldsymbol{y})\right]dt
\end{align}
\end{small}
which is the simplified KL divergence between $\mathbb{P}'$ and $\mathbb{P}$. We can also use the equality $\boldsymbol{Q}_t(\boldsymbol{x},\boldsymbol{x})=1-\sum_{\boldsymbol{y}\neq \boldsymbol{x}}\boldsymbol{Q}_t(\boldsymbol{x},\boldsymbol{y})$ to write $\sum_{\boldsymbol{y}\neq \boldsymbol{X}_t}(\boldsymbol{Q}_t-\boldsymbol{Q}'_t)(\boldsymbol{X}_t,\boldsymbol{y})=(\boldsymbol{Q}'_t-\boldsymbol{Q}_t)(\boldsymbol{X}_t,\boldsymbol{X}_t)$, which gives us:
\begin{small}
\begin{align}
    \text{KL}(\mathbb{P}'\|\mathbb{P})
    &=\text{KL}(\pi'_0\|\pi_0)+\mathbb{E}_{\boldsymbol{X}_{0:T}\sim \mathbb{P}'}\int_0^T\left[\sum_{\boldsymbol{y}\neq \boldsymbol{X}_t}\left(\bluetext{\boldsymbol{Q}_t'\log \frac{\boldsymbol{Q}_{t}'}{\boldsymbol{Q}_{t}}}\right)(\boldsymbol{X}_t, \boldsymbol{y})+\int _0^T\pinktext{(\boldsymbol{Q}'_t-\boldsymbol{Q}_t)(\boldsymbol{X}_t,\boldsymbol{X}_t)}\right]dt
\end{align}
\end{small}
which is an equivalent expression for KL divergence between CTMCs using the rate of remaining at a position or the \textbf{stay rate}. \hfill $\square$

\begin{figure}
    \centering
    \includegraphics[width=\linewidth]{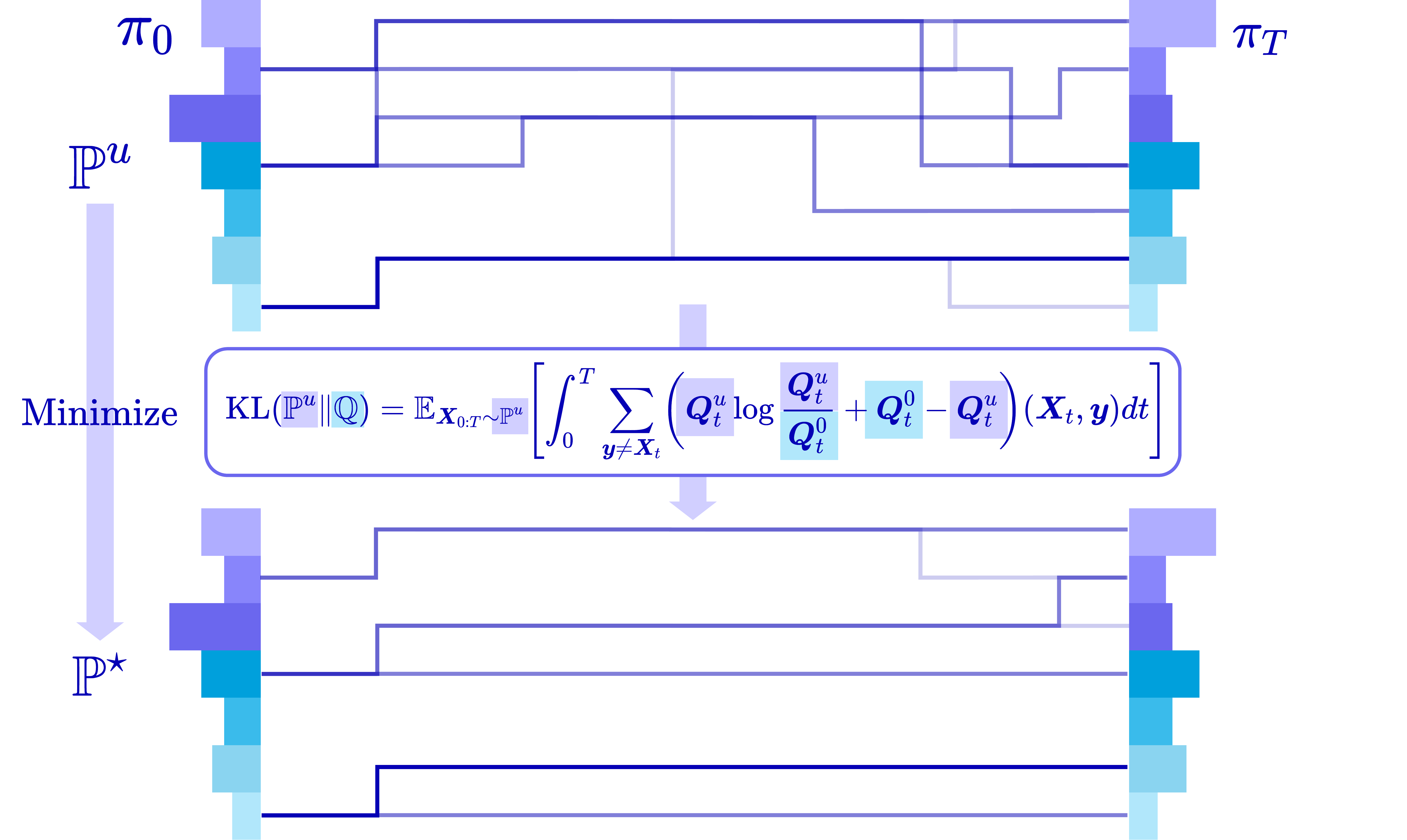}
    \caption{\textbf{Discrete Schrödinger Bridge Problem.} The discrete SB problem seeks a controlled CTMC path measure $\mathbb{P}^\star$ that is closest, in KL divergence, to a reference CTMC $\mathbb{Q}$ while matching the prescribed marginal distributions $\pi_0$ and $\pi_T$. \textbf{Top}: CTMC trajectories under the sub-optimal controlled path measure $\mathbb{P}^u$ with generator $\boldsymbol{Q}^u$. \textbf{Bottom}: CTMC trajectories under the Schrödinger bridge path measure $\mathbb{P}^\star$ with generator $\boldsymbol{Q}^\star$ obtained by minimizing the KL divergence with the reference generator $\boldsymbol{Q}^0$ while maintaining the terminal marginals.}
    \label{fig:discrete-sb}
\end{figure}

This decomposition reveals that the KL divergence between two CTMC path measures is the sum of the discrepancy between their initial marginals and a time-integrated KL divergence of the generator matrices at \textit{jump times}, where the state $\boldsymbol{X}_t$ jumps to a new state $\boldsymbol{y}\neq \boldsymbol{X}_t$. This provides an intuitive interpretation for the discrete Schrödinger bridge problem, which selects, among all Markov processes matching the marginals, the one whose jump dynamics deviates minimally in KL divergence from the reference dynamics. While we defined the canonical reference process in the continuous state space as Brownian motion with a reference drift $\boldsymbol{f}$, which is often set to $\boldsymbol{f}:= 0$, we can also define two common forms of the reference generator $\boldsymbol{Q}^0$ below.

\purple[Forms of the Reference Generator for CTMCs]{
The \textbf{reference generator} $\boldsymbol{Q}^0$ defining the CTMC $\mathbb{Q}$ is the baseline discrete stochastic dynamics that is minimally reweighted to match the prescribed marginal constraints. The choice of $\boldsymbol{Q}^0$ fundamentally changes the solution to the (\ref{eq:discrete-sb-problem}). Here, we will highlight the \boldtext{uniform generator} and the \boldtext{pre-trained generator} as common choices. 

\textbf{Uniform Generator. }The uniform generator $\boldsymbol{Q}^0_t$ is defined as:
\begin{small}
\begin{align}
    \boldsymbol{Q}^0_t(\boldsymbol{x},\boldsymbol{y})=\begin{cases}
    \frac{\gamma(t)}{|\mathcal{X}|-1}&\boldsymbol{y}\neq \boldsymbol{x}\\
    -\gamma(t)&\boldsymbol{y}=\boldsymbol{x}\tag{Uniform Generator}
\end{cases}
\end{align}
\end{small}
which corresponds to a homogeneous jump process such that from any state $\boldsymbol{x}\in \mathcal{X}$ at time $t$, the state remains unchanged with rate $-\gamma(t)$ and jumps uniformly to any other state $\boldsymbol{y}\neq \boldsymbol{x}$. This means that the optimal generator $\boldsymbol{Q}^\star$ is entirely determined by the marginal constraints. It can be interpreted as the discrete state space analog of pure Brownian motion and is useful for settings where prior dynamics are unknown. 

\textbf{Baseline Transition Generator. } A more structured alternative is to define the reference generator $\boldsymbol{Q}^0_t$ with baseline transition rates of the system's natural dynamics or from a \textbf{pretrained model}, where the model is trained to capture the dynamics of the system from data samples. This choice embeds prior knowledge about plausible transitions into the reference process. The Schrödinger bridge then acts as a minimal correction of the pretrained dynamics to match the prescribed marginals.
}

While we have defined the (\ref{eq:discrete-sb-problem}) objective and provided common forms of the reference generator, it remains unknown how to optimize the generator such that it yields the optimal CTMC while preserving the marginal constraints. In the remainder of Section \ref{sec:discrete-state-space}, we introduce two concrete methods of solving the discrete SB problem, each of which has deep connections to methods in the continuous state space. Specifically, we introduce \textbf{stochastic optimal control (SOC) of CTMCs}, which extends the SOC ideas from Section \ref{sec:sb-optimal-control} to the discrete state space, and \textbf{discrete diffusion Schrödinger bridge matching}, which extends the ideas of Markov and reciprocal projections from Section \ref{subsec:markov-reciprocal-proj} and diffusion Schrödinger bridge matching algorithm from Section \ref{subsec:diffusionsbm} to the discrete state space. 

\subsection{Stochastic Optimal Control of CTMCs}
\label{subsec:soc-ctmcs}
\textit{Prerequisite: Section \ref{sec:sb-optimal-control}}

Just like in the continuous state space, we can reframe the SB problem as a \boldtext{stochastic optimal control} (SOC) problem, where the Schrödinger bridge aligns with the lowest cost path from any intermediate state to a state in the target distribution. In this section, we build the theoretical foundations of SOC in the discrete state space, highlighting the deviations from the continuous state-spacee formulation, which will lead us to explicitly defining objectives for tractably solving the discrete SB problem with SOC in Section \ref{subsec:discrete-sb-soc}.

First, we define the \boldtext{cost functional} $J(\boldsymbol{x},t;\boldsymbol{u}):\mathcal{X}\times [0,T]\to \mathbb{R}$ which returns the \textit{cost-to-go} from a intermediate state $\boldsymbol{x}\in \mathcal{X}$ to the target distribution $\pi_T$ under the \textbf{controlled} path measure $\mathbb{P}^u$:
\begin{align}
    J(\boldsymbol{x},t;\boldsymbol{u})&:=\mathbb{E}_{\boldsymbol{X}^u_{0:T}\sim \mathbb{P}^u}\left[\int_0^T\sum_{\boldsymbol{y}\neq \boldsymbol{X}^u_s}\bluetext{C_t(\boldsymbol{X}^u_s,\boldsymbol{y})}ds +\pinktext{\Phi(\boldsymbol{X}^u_T)}\bigg|\boldsymbol{X}^u_t=\boldsymbol{x}\right]
\end{align}
which measures the expected cumulative running cost $C_t(\boldsymbol{x},\boldsymbol{y})$ incurred from time $t$ until the terminal time $T$ and the terminal cost $\Phi(\boldsymbol{x}):\mathcal{X}\to \mathbb{R}$. The running cost is defined as the instantaneous KL divergence between the controlled generator $\boldsymbol{Q}^u_t$ and the reference generator $\boldsymbol{Q}^0_t$ given by:
\begin{align}
    \bluetext{C_t(\boldsymbol{x},\boldsymbol{y})}&:=\text{KL}(p^u_t\|q_t)=\left(\boldsymbol{Q}_t^u\log \frac{\boldsymbol{Q}_t^u}{\boldsymbol{Q}_t^0}-\boldsymbol{Q}_t^u+\boldsymbol{Q}^0_t\right)(\boldsymbol{x}, \boldsymbol{y})
\end{align}

The objective of the SOC problem is to determine the \boldtext{optimal control} generator $\boldsymbol{Q}^\star:=\boldsymbol{Q}^{u^\star}$ that minimizes the cost-to-go functional:
\begin{align}
    J^\star(\boldsymbol{x},t;\boldsymbol{u}^\star):= \inf_{\mathbb{P}^u}J(\boldsymbol{x},t;\boldsymbol{u}), \quad \forall (\boldsymbol{x},t)\in \mathcal{X}\times[0,T]
\end{align}
which generates the \textbf{optimal CTMC path measure} $\mathbb{P}^\star:= \mathbb{P}^{u^\star}$ with generator $\boldsymbol{Q}^\star_t:=\boldsymbol{Q}^{u^\star}_t$. Analogous to the continuous setting, the optimal cost-to-go satisfies (\ref{eq:bellman-principle}). For a small time increment $\Delta t$, the cost decomposes into the cost accumulated over $[t,t+\Delta t]$ and the optimal cost from $t+\Delta t$ onward:
\begin{small}
\begin{align}
    J^\star(\boldsymbol{x},t;\boldsymbol{u}^\star)&=\inf_{\mathbb{P}^u}\mathbb{E}_{\boldsymbol{X}^u_{t:T}\sim \mathbb{P}^u}\left[\left(\bluetext{\int_t^{t+\Delta t}}+\pinktext{\int_{t+\Delta t}^T}\right)\sum_{\boldsymbol{y}\neq \boldsymbol{X}^u_s}\bluetext{C_t(\boldsymbol{X}^u_s,\boldsymbol{y})}ds +\Phi(\boldsymbol{X}^u_T)\bigg|\boldsymbol{X}^u_t=\boldsymbol{x}\right]\nonumber\\
    &=\inf_{\mathbb{P}^u}\bigg\{\bluetext{\underbrace{\Delta t\sum_{\boldsymbol{y}\neq \boldsymbol{x}}C_t(\boldsymbol{x}, \boldsymbol{y})+\mathcal{O}(\Delta t^2)}_{\text{optimal cost of jumps over }t\to t+\Delta t}}+\pinktext{\underbrace{\mathbb{E}_{\boldsymbol{X}^u_{t+\Delta t:T}\sim \mathbb{P}^u}\left[J^\star(\boldsymbol{X}^u_{t+\Delta t},t+\Delta t)|\boldsymbol{X}^u_t=\boldsymbol{x}\right]}_{\text{optimal cost of all jumps over }t+\Delta t\to T}}\bigg\}\label{eq:ctmc-soc-prob}
\end{align}
\end{small}
The optimally controlled measure $\mathbb{P}^\star$ can be obtained by defining the \textbf{value function} $V_t$ as the \textit{optimal cost-to-go} $V_t(\boldsymbol{x}):=J^\star(\boldsymbol{x},t;\boldsymbol{u}^\star)$, which yields the \textbf{dynamic programming relation}:
\begin{small}
\begin{align}
    \bluetext{V_t(\boldsymbol{x})}&=\inf_{\mathbb{P}^u}\bigg\{\Delta t\sum_{\boldsymbol{y}\neq \boldsymbol{x}}C_t(\boldsymbol{x}, \boldsymbol{y})+\mathcal{O}(\Delta t^2)+\mathbb{E}_{\boldsymbol{X}^u_{t+\Delta t:T}\sim \mathbb{P}^u}\left[\bluetext{V_{t+\Delta t}(\boldsymbol{X}^u_{t+\Delta t})}|\boldsymbol{X}^u_t=\boldsymbol{x}\right]\bigg\}\label{eq:ctmc-value-func}
\end{align}
\end{small}
Since no more running cost $C_t$ can be incurred at time $t=T$, the terminal value function is equal to the terminal cost $V_T(\boldsymbol{x})=\Phi(\boldsymbol{x})$. 

To obtain an explicit characterization of the optimal CTMC dynamics solving the stochastic optimal control problem in (\ref{eq:ctmc-soc-prob}), we derive the structure of the optimal process in a sequence of steps. The key goal is to connect the dynamic programming formulation of the SOC problem with the change-of-measure perspective underlying Schrödinger bridges. Concretely, we first determine the form of the optimal controlled generator $\boldsymbol{Q}^\star$, then characterize the value function through the Hamilton–Jacobi–Bellman equation, and finally use these results to recover the optimal path measure $\mathbb{P}^\star$ and its likelihood ratio with respect to the reference process. \textbf{The derivation proceeds as follows:}
\begin{enumerate}
    \item [(i)] We derive the form of the optimal generator $\boldsymbol{Q}^\star_t$ that defines the law of paths under the path measure $\mathbb{P}^\star$ that solves the SOC problem in (\ref{eq:ctmc-soc-prob}) (Proposition \ref{prop:ctmc-optimal-generator}).
    \item[(ii)] We show that the value function in (\ref{eq:ctmc-value-func}) satisfies the Hamilton-Jacobi-Bellman equation (Corollary \ref{corollary:ctmc-hjb}).
    \item[(iv)] We derive the form of the optimal path measure $\mathbb{P}^\star$ that solves the SOC problem (Proposition \ref{prop:ctmc-optimal-path}).
    \item[(iii)] Finally, we derive the Radon-Nikodym derivative (RND) between the optimal path measure $\mathbb{P}^\star$, and the reference path measure $\mathbb{Q}$ (Proposition \ref{prop:ctmc-rnd-optimal}).
    
\end{enumerate}
Together, these theoretical ideas will form the basis for solving the discrete SB using SOC. We start with the derivation of the optimal generator $\boldsymbol{Q}^\star_t$ with respect to the value function $V_t$ (\ref{eq:ctmc-value-func}) which will naturally lead to the proof that the value function satisfies the HJB equation.

% sign changed
\begin{proposition}[Optimal Generator]\label{prop:ctmc-optimal-generator}
    Given the generator $\boldsymbol{Q}^0_t$ of the reference process $\mathbb{Q}$, the \textbf{optimal generator} $\boldsymbol{Q}^\star_t$ of the process $\mathbb{P}^\star$ that solves the SOC problem in (\ref{eq:ctmc-soc-prob}) takes the form:
    \begin{align}
        \boldsymbol{Q}_t^\star(\boldsymbol{x}, \boldsymbol{y})=\boldsymbol{Q}^0_t(\boldsymbol{x},\boldsymbol{y})e^{V_t(\boldsymbol{x})-V_t(\boldsymbol{y})}, \quad \forall \boldsymbol{x},\boldsymbol{y}\in \mathcal{X}\tag{Optimal Generator}\label{eq:prop-optimal-generator}
    \end{align}
    where $V_t:\mathcal{X}\to \mathbb{R}$ is the value function defined in (\ref{eq:ctmc-value-func}).
\end{proposition}

\textit{Proof.} First, we expand the second term of the value function in (\ref{eq:ctmc-value-func}), which defines the Bellman recursion, to get:
\begin{small}
\begin{align}
    \inf_{\mathbb{P}^u}\; &\mathbb{E}_{\boldsymbol{X}^u_{t+\Delta t:T}\sim \mathbb{P}^u}\left[V_{t+\Delta t}(\boldsymbol{X}^u_{t+\Delta t})\big |\boldsymbol{X}^u_t=\boldsymbol{x}\right]=\inf_{\mathbb{P}^u}\left[\sum_{\boldsymbol{y}\in \mathcal{X}}V_{t+\Delta t}(\boldsymbol{y})\bluetext{\mathbb{P}^u(\boldsymbol{X}^u_{t+\Delta t}=\boldsymbol{y}|\boldsymbol{X}^u_t=\boldsymbol{x})}\right]\nonumber\\
    &\overset{(\ref{eq:ctmc-transition-prob})}{=}\inf_{\mathbb{P}^u}\left[\sum_{\boldsymbol{y}\in \mathcal{X}}V_{t+\Delta t}(\boldsymbol{y})\bluetext{\left(\boldsymbol{1}_{\boldsymbol{x}=y}+\Delta t\boldsymbol{Q}_t^u(\boldsymbol{x},\boldsymbol{y})+\mathcal{O}(\Delta t^2)\right)}\right]\nonumber\\
    &=\inf_{\mathbb{P}^u}\bigg[\bluetext{\underbrace{V_{t+\Delta t}(\boldsymbol{x})(\boldsymbol{1}_{\boldsymbol{x}=y}+\Delta t\boldsymbol{Q}_t^u(\boldsymbol{x},\boldsymbol{x}))}_{\text{Case: }\boldsymbol{y}=\boldsymbol{x}}}+\pinktext{\underbrace{\sum_{\boldsymbol{y}\neq \boldsymbol{x}}V_{t+\Delta t}(\boldsymbol{y})\Delta t\boldsymbol{Q}_t^u(\boldsymbol{x},\boldsymbol{y})}_{\text{Case: }\boldsymbol{y}\neq\boldsymbol{x}}}+\mathcal{O}(\Delta t^2))\bigg]\label{eq:optimal-gen-proof1}
\end{align}
\end{small}
For any CTMC generator, the sum of transition rates from a given state $\boldsymbol{x}$ to all states $\boldsymbol{y}\in \mathcal{X}$ must sum to zero, we have  $\boldsymbol{Q}_t(\boldsymbol{x},\boldsymbol{x})=-\sum_{\boldsymbol{y}\neq \boldsymbol{x}}\boldsymbol{Q}_t(\boldsymbol{x},\boldsymbol{y})$, which we can substitute into (\ref{eq:optimal-gen-proof1}) to get:
\begin{small}
\begin{align}
    \inf_{\mathbb{P}^u}\; &\mathbb{E}_{\boldsymbol{X}^u_{t+\Delta t:T}\sim \mathbb{P}^u}\big[V_{t+\Delta t}(\boldsymbol{X}^u_{t+\Delta t})\big |\boldsymbol{X}^u_t=\boldsymbol{x}\big]\nonumber\\
    &=\inf_{\mathbb{P}^u}\left[V_{t+\Delta t}(\boldsymbol{x})\left(\boldsymbol{1}_{\boldsymbol{x}=y}\bluetext{-}\Delta t\bluetext{\sum_{\boldsymbol{y}\neq \boldsymbol{x}}\boldsymbol{Q}^u_t(\boldsymbol{x},\boldsymbol{y})}\right)+\sum_{\boldsymbol{y}\neq \boldsymbol{x}}V_{t+\Delta t}(\boldsymbol{y})\Delta t\boldsymbol{Q}_t^u(\boldsymbol{x},\boldsymbol{y})+\mathcal{O}(\Delta t^2))\right]\nonumber\\
    &=\inf_{\mathbb{P}^u}\left[V_{t+\Delta t}(\boldsymbol{x})\bluetext{-\Delta t\sum_{\boldsymbol{x}\neq \boldsymbol{y}}V_{t+\Delta t}(\boldsymbol{x})\boldsymbol{Q}_t^u(\boldsymbol{x},\boldsymbol{y})}+\Delta t\sum_{\boldsymbol{x}\neq \boldsymbol{y}}V_{t+\Delta t}(\boldsymbol{y})\boldsymbol{Q}_t^u(\boldsymbol{x},\boldsymbol{y})+\mathcal{O}(\Delta t^2)\right]\nonumber\\
    &=V_{t+\Delta t}(\boldsymbol{x})+\Delta t\inf_{\mathbb{P}^u}\left[\sum_{\boldsymbol{x}\neq \boldsymbol{y}}\boldsymbol{Q}_t^u(\boldsymbol{x},\boldsymbol{y})\bluetext{(V_{t+\Delta t}(\boldsymbol{y})-V_{t+\Delta t}(\boldsymbol{x}))}\right]+\mathcal{O}(\Delta t^2)
\end{align}
\end{small}
Substituting this back into (\ref{eq:ctmc-value-func}), we have:
\begin{small}
\begin{align}
    V_t(\boldsymbol{x})&=\inf_{\mathbb{P}^u}\bigg\{\Delta t\sum_{\boldsymbol{y}\neq \boldsymbol{x}}C_t(\boldsymbol{x}, \boldsymbol{y})+\mathcal{O}(\Delta t^2)\bluetext{+V_{t+\Delta t}(\boldsymbol{x})+\Delta t\sum_{\boldsymbol{y}\neq \boldsymbol{x}}\boldsymbol{Q}_t^u(\boldsymbol{x},\boldsymbol{y})(V_{t+\Delta t}(\boldsymbol{y})-V_{t+\Delta t}(\boldsymbol{x}))+\mathcal{O}(\Delta t^2)}\bigg\}\nonumber\\
    V_t(\boldsymbol{x})-V_{t+\Delta t}(\boldsymbol{x})&=\inf_{\mathbb{P}^u}\bigg\{\Delta t\sum_{\boldsymbol{y}\neq \boldsymbol{x}}C_t(\boldsymbol{x}, \boldsymbol{y})+\Delta t\sum_{\boldsymbol{y}\neq \boldsymbol{x}}\boldsymbol{Q}_t^u(\boldsymbol{x},\boldsymbol{y})(V_{t+\Delta t}(\boldsymbol{y})-V_{t+\Delta t}(\boldsymbol{x}))\bigg\}+\mathcal{O}(\Delta t^2)\nonumber\\
    V_{t+\Delta t}(\boldsymbol{x})-V_t(\boldsymbol{x})&=-\inf_{\mathbb{P}^u}\bigg\{\Delta t\sum_{\boldsymbol{y}\neq \boldsymbol{x}}C_t(\boldsymbol{x}, \boldsymbol{y})+\Delta t\sum_{\boldsymbol{y}\neq \boldsymbol{x}}\boldsymbol{Q}_t^u(\boldsymbol{x},\boldsymbol{y})(V_{t+\Delta t}(\boldsymbol{y})-V_{t+\Delta t}(\boldsymbol{x}))\bigg\}+\mathcal{O}(\Delta t^2)
\end{align}
\end{small}
where we rearrange the Bellman recursion to express the forward difference $V_{t+\Delta t}(\boldsymbol{x})-V_t(\boldsymbol{x})$, which introduces a minus sign before the infimum. This expression is used to get the time derivative $\partial_tV_t$, by dividing both sides by $\Delta t$ and take the limit $\Delta t\to 0$:
\begin{small}
\begin{align}
\partial_tV_t(\boldsymbol{x})=\lim_{\Delta t\to 0}\left\{\frac{V_{t+\Delta t}(\boldsymbol{x})-V_{t}(\boldsymbol{x})}{\Delta t}\right\}&=-\inf_{\mathbb{P}^u}\bigg\{\sum_{\boldsymbol{y}\neq \boldsymbol{x}}\bluetext{\underbrace{\bigg(C_t(\boldsymbol{x}, \boldsymbol{y})+\boldsymbol{Q}_t^u(\boldsymbol{x},\boldsymbol{y})(V_t(\boldsymbol{y})-V_t(\boldsymbol{x}))\bigg)}_{f(\boldsymbol{Q}^u)}}\bigg\}\label{eq:partial-t-value}
\end{align}
\end{small}
Defining $f(\boldsymbol{Q}_t^u)$ as the function inside the infimum for all $\boldsymbol{x}\neq \boldsymbol{y}$ and expanding $C_t(\boldsymbol{x},\boldsymbol{y})$ as the KL divergence in (\ref{eq:ctmc-soc-prob}), we can take the derivative $f'(\boldsymbol{Q}_t^u)$ with respect to $\boldsymbol{Q}_t^u$ to get:
\begin{align}
    f(\boldsymbol{Q}^u_t)&=\left(\boldsymbol{Q}^u_t\log \frac{\boldsymbol{Q}_t^u}{\boldsymbol{Q}_t^0}\right)(\boldsymbol{x}, \boldsymbol{y})-\boldsymbol{Q}^u_t(\boldsymbol{x},\boldsymbol{y})+\boldsymbol{Q}^0_t(\boldsymbol{x}, \boldsymbol{y})+\boldsymbol{Q}^u_t(\boldsymbol{x}, \boldsymbol{y})(V_t(\boldsymbol{y})-V_t(\boldsymbol{x}))\\
    f'(\boldsymbol{Q}^u_t)&=\log \frac{\boldsymbol{Q}_t^u}{\boldsymbol{Q}_t^0}(\boldsymbol{x}, \boldsymbol{y})+V_t(\boldsymbol{y})-V_t(\boldsymbol{x})
\end{align}
Setting $f'(\boldsymbol{Q}^u_t)=0$ to obtain the minimizer $\boldsymbol{Q}^\star_t$:
\begin{align}
    \log \frac{\boldsymbol{Q}_t^\star}{\boldsymbol{Q}_t^0}(\boldsymbol{x}, \boldsymbol{y})=V_t(\boldsymbol{x})-V_t(\boldsymbol{y})\implies \boxed{\boldsymbol{Q}^\star_t(\boldsymbol{x},\boldsymbol{y})=\boldsymbol{Q}^0_t(\boldsymbol{x},\boldsymbol{y})e^{V_t(\boldsymbol{x})-V_t(\boldsymbol{y})}}\label{eq:end-optimal-generator}
\end{align}
which is the form of the optimal generator. \hfill $\square$

From the result (\ref{eq:end-optimal-generator}), we can also observe that by rewriting the exponential as a fraction, we recover a form analogous the \boldtext{Doob's $h$-transform} described in Section \ref{subsec:doob-transform} for CTMCs. 

% sign changed
\begin{remark}[Doob's $h$-Transform of CTMCs]\label{remark:ctmc-doob}
The optimal generator $\boldsymbol{Q}^\star_t$ is the Doob $h$-transform of the reference generator $\boldsymbol{Q}^0_t$ where the $h$ function is defined as the exponentiated value function $h(\boldsymbol{x},t):=e^{-V_t(\boldsymbol{x})}$:
\begin{align}
\boldsymbol{Q}^\star_t(\boldsymbol{x},\boldsymbol{y})=\boldsymbol{Q}^0_t(\boldsymbol{x},\boldsymbol{y})\frac{e^{V_t(\boldsymbol{x})}}{e^{V_t(\boldsymbol{y})}}=\boldsymbol{Q}^0_t(\boldsymbol{x},\boldsymbol{y})\frac{e^{-V_t(\boldsymbol{y})}}{e^{-V_t(\boldsymbol{x})}}=:\boldsymbol{Q}^0_t(\boldsymbol{x},\boldsymbol{y})\frac{h(\boldsymbol{y},t)}{h(\boldsymbol{x},t)}
\end{align}
which can be interpreted as tilting the generator toward states $\boldsymbol{y}$ that minimize the optimal cost-to-go defined by $V_t(\boldsymbol{x}):= J^\star(\boldsymbol{x},t;\boldsymbol{u}^\star)$.
\end{remark}

Now we will show that the value function satisfies the HJB equation, which is analagous to our derivation in Section \ref{subsec:nonlinear-sbp} in continuous state spaces. Crucially, this defines the discrete SOC problem with a non-linear PDE which can be transformed to a linear equation via exponentiation, which acts as a \textbf{discrete analog of the Hopf-Cole transform} discussed in Section \ref{subsec:hopf-cole-transform}.

\begin{corollary}[Hamilton-Jacobi-Bellman Equation]\label{corollary:ctmc-hjb}
    The value function in (\ref{eq:ctmc-value-func}) satisfies the \textbf{Hamilton-Jacobi-Bellman equations}, defined as:
    \begin{align}
        \partial_tV_t(\boldsymbol{x})=\sum_{\boldsymbol{y}\neq \boldsymbol{x}}\boldsymbol{Q}_t^0(\boldsymbol{x},\boldsymbol{y})\left(e^{V_t(\boldsymbol{x})-V_t(\boldsymbol{y})}-1\right)\iff\partial_te^{-V_t(\boldsymbol{x})}=\sum_{\boldsymbol{y}\ne\boldsymbol{x}}\boldsymbol{Q}^0_t(\boldsymbol{x},\boldsymbol{y})\left(e^{-V_t(\boldsymbol{x})}-e^{-V_t(\boldsymbol{y})}\right)\nonumber
    \end{align}
\end{corollary}

\textit{Proof.} To prove this, we substitute the final form of the optimal generator $\boldsymbol{Q}^\star_t=\boldsymbol{Q}^0_te^{V_t(\boldsymbol{x})-V_t(\boldsymbol{y})}$ defined in (\ref{eq:prop-optimal-generator}) into the equation for $\partial_tV_t$ in (\ref{eq:partial-t-value}) to get:
\begin{small}
\begin{align}
&\partial_tV_t(\boldsymbol{x})=-\sum_{\boldsymbol{y}\neq \boldsymbol{x}}\bigg[\bluetext{\boldsymbol{Q}^\star_t}\log \frac{\bluetext{\boldsymbol{Q}^\star_t}}{\boldsymbol{Q}_t^0}-\boldsymbol{Q}^\star_t+\boldsymbol{Q}^0_t+\bluetext{\boldsymbol{Q}^\star_t}(V_t(\boldsymbol{y})-V_t(\boldsymbol{x}))\bigg]\nonumber\\
&=-\sum_{\boldsymbol{y}\neq \boldsymbol{x}}\bigg[\bluetext{\boldsymbol{Q}^0_te^{V_t(\boldsymbol{x})-V_t(\boldsymbol{y})}}\log \frac{\bluetext{\boldsymbol{Q}^0_te^{V_t(\boldsymbol{x})-V_t(\boldsymbol{y})}}}{\boldsymbol{Q}_t^0}-\boldsymbol{Q}^0_te^{V_t(\boldsymbol{x})-V_t(\boldsymbol{y})}+\boldsymbol{Q}^0_t+\bluetext{\boldsymbol{Q}^0_te^{V_t(\boldsymbol{x})-V_t(\boldsymbol{y})}}(V_t(\boldsymbol{y})-V_t(\boldsymbol{x}))\bigg]\nonumber\\
&=-\sum_{\boldsymbol{y}\neq \boldsymbol{x}}\bigg[\bluetext{-\boldsymbol{Q}^0_te^{V_t(\boldsymbol{x})-V_t(\boldsymbol{y})}(V_t(\boldsymbol{y})-V_t(\boldsymbol{x}))}-\boldsymbol{Q}^0_te^{V_t(\boldsymbol{y})-V_t(\boldsymbol{x})}+\boldsymbol{Q}^0_t+\bluetext{\boldsymbol{Q}^0_t(\boldsymbol{x},\boldsymbol{y})e^{V_t(\boldsymbol{x})-V_t(\boldsymbol{y})}(V_t(\boldsymbol{y})-V_t(\boldsymbol{x}))}\bigg]\nonumber\\
&=-\sum_{\boldsymbol{y}\neq \boldsymbol{x}}\boldsymbol{Q}^0_t\left(1-e^{V_t(\boldsymbol{x})-V_t(\boldsymbol{y})}\right)=\boxed{\sum_{\boldsymbol{y}\neq \boldsymbol{x}}\boldsymbol{Q}^0_t\left(e^{V_t(\boldsymbol{x})-V_t(\boldsymbol{y})}-1\right)}\label{eq:ctmc-partial-t-value}
\end{align}
\end{small}
which gives us the first HJB equation in the Corollary. To get the second expression, we can differentiate $e^{-V_t(\boldsymbol{x})}$ and apply the chain rule to get:
\begin{small}
\begin{align}
    \partial_te^{-V_t(\boldsymbol{x})}=e^{-V_t(\boldsymbol{x})}\bluetext{\partial_tV_t(\boldsymbol{x})}&\overset{(\ref{eq:ctmc-partial-t-value})}{=}e^{-V_t(\boldsymbol{x})}\bluetext{\sum_{\boldsymbol{y}\neq \boldsymbol{x}}\boldsymbol{Q}^0_t(\boldsymbol{x},\boldsymbol{y})\left(e^{V_t(\boldsymbol{x})-V_t(\boldsymbol{y})}-1\right)}\nonumber\\
    &=\boxed{\sum_{\boldsymbol{y}\neq \boldsymbol{x}}\boldsymbol{Q}^0_t(\boldsymbol{x},\boldsymbol{y})\left(e^{-V_t(\boldsymbol{y})}-e^{-V_t(\boldsymbol{x})}\right)}
\end{align}
\end{small}
which is the second HJB equation in the Corollary. \hfill $\square$

Using this Corollary, we can derive the \boldtext{optimal path measure}, which can be defined in terms of the value function $V_t(\boldsymbol{x})$.

\begin{proposition}[Optimal Path Measure]\label{prop:ctmc-optimal-path}
    The optimal path measure $\mathbb{P}^\star$ given the value function $V_t(\boldsymbol{x})$ can be expressed as:
    \begin{align}
        p_t^\star(\boldsymbol{x})=\frac{1}{Z_t}q_t(\boldsymbol{x})e^{-V_t(\boldsymbol{x})}, \quad Z_t:=\mathbb{E}_{\boldsymbol{x}\sim q_t}\left[e^{-V_t(\boldsymbol{x})}\right]
    \end{align}
\end{proposition}

\textit{Proof.} From Remark \ref{remark:ctmc-doob}, we can consider the probability of a state $\boldsymbol{x}\in \mathcal{X}$ under the optimal path measure as its probability under the reference measure tilted by the $h$-function defined as $h(\boldsymbol{x},t):=e^{-V_t(\boldsymbol{x})}$ which yields $\xi_t(\boldsymbol{x})=\frac{1}{Z}q_t(\boldsymbol{x})e^{-V_t(\boldsymbol{x})}$, where $Z$ is the normalization factor. To show that $\xi(\boldsymbol{x})$ is indeed the optimal path measure, we must check that it satisfies \textbf{Kolmogorov's forward equation} defined in Proposition \ref{lemma:ctmc-kolmogorov-fwd} for the optimal generator $\boldsymbol{Q}^\star_t$. First, taking the partial derivative, we get:
\begin{align}
    \partial_t\xi_t(\boldsymbol{x})=\partial_t\left(\frac{1}{Z}q_t(\boldsymbol{x})e^{-V_t(\boldsymbol{x})}\right)=\frac{1}{Z}\left(e^{-V_t(\boldsymbol{x})}\bluetext{\partial_tq_t(\boldsymbol{x})}+q_t(\boldsymbol{x})\pinktext{\partial_te^{-V_t(\boldsymbol{x})}}\right)\label{eq:ctmc-optimal-path-eq1}
\end{align}
Applying the Kolgomorov forward equation from Lemma \ref{lemma:ctmc-kolmogorov-fwd} to the reference path measure $\mathbb{Q}$ and the HJB equations from Corollary \ref{corollary:ctmc-hjb}, we have:
\begin{align}
    \bluetext{\partial_tp_t^0(\boldsymbol{x})}&\overset{(\ref{lemma:ctmc-kolmogorov-fwd})}{=}\sum_{\boldsymbol{x}\neq \boldsymbol{y}}\left(\boldsymbol{Q}^0_t(\boldsymbol{y}, \boldsymbol{x})q_t(\boldsymbol{y})-\boldsymbol{Q}_t^0(\boldsymbol{x}, \boldsymbol{y})q_t(\boldsymbol{x})\right)\\
    \pinktext{\partial_te^{-V_t(\boldsymbol{x})}}&\overset{(\ref{corollary:ctmc-hjb})}{=}\sum_{\boldsymbol{x}\neq \boldsymbol{y}}\boldsymbol{Q}_t^0(\boldsymbol{x},\boldsymbol{y})\left(e^{-V_t(\boldsymbol{x})}-e^{-V_t(\boldsymbol{y})}\right)
\end{align}
and substituting this back into (\ref{eq:ctmc-optimal-path-eq1}), we get:
\begin{small}
\begin{align}
    &\partial_t\xi_t(\boldsymbol{x})=\frac{1}{Z}\left[e^{-V_t(\boldsymbol{x})}\bluetext{\sum_{\boldsymbol{x}\neq \boldsymbol{y}}\left(\boldsymbol{Q}^0_t(\boldsymbol{y}, \boldsymbol{x})q_t(\boldsymbol{y})-\boldsymbol{Q}_t^0(\boldsymbol{x}, \boldsymbol{y})q_t(\boldsymbol{x})\right)}+q_t(\boldsymbol{x})\pinktext{\sum_{\boldsymbol{x}\neq \boldsymbol{y}}\boldsymbol{Q}_t^0(\boldsymbol{x},\boldsymbol{y})\left(e^{-V_t(\boldsymbol{x})}-e^{-V_t(\boldsymbol{y})}\right)}\right]\nonumber\\
    &=\sum_{\boldsymbol{x}\neq\boldsymbol{y}}\bigg[\boldsymbol{Q}^0_t(\boldsymbol{y}, \boldsymbol{x})\frac{1}{Z}q_t(\boldsymbol{y})\bluetext{e^{-V_t(\boldsymbol{x})}}\underbrace{-\boldsymbol{Q}_t^0(\boldsymbol{x}, \boldsymbol{y})\frac{1}{Z}q_t(\boldsymbol{x})e^{-V_t(\boldsymbol{x})}+\boldsymbol{Q}_t^0(\boldsymbol{x}, \boldsymbol{y})\frac{1}{Z}q_t(\boldsymbol{x})e^{-V_t(\boldsymbol{x})}}_{=0}-\boldsymbol{Q}^0_t(\boldsymbol{x}, \boldsymbol{y})\frac{1}{Z}q_t(\boldsymbol{x})\bluetext{e^{-V_t(\boldsymbol{y})}}\bigg]\nonumber\\
    &=\sum_{\boldsymbol{x}\neq\boldsymbol{y}}\bigg[\boldsymbol{Q}^0_t(\boldsymbol{y}, \boldsymbol{x})\bluetext{\underbrace{\frac{1}{Z}q_t(\boldsymbol{y})e^{-V_t(\boldsymbol{y})}}_{=:\xi _t(\boldsymbol{y)}}}e^{V_t(\boldsymbol{x})-V_t(\boldsymbol{y})}-\boldsymbol{Q}^0_t(\boldsymbol{x}, \boldsymbol{y})\bluetext{\underbrace{\frac{1}{Z}q_t(\boldsymbol{x})e^{-V_t(\boldsymbol{x})}}_{=:\xi_t(\boldsymbol{x})}}e^{V_t(\boldsymbol{x})-V_t(\boldsymbol{y})}\bigg]\nonumber\\
    &=\sum_{\boldsymbol{x}\neq\boldsymbol{y}}\bigg[\pinktext{\underbrace{\boldsymbol{Q}^0_t(\boldsymbol{y}, \boldsymbol{x})e^{V_t(\boldsymbol{x})-V_t(\boldsymbol{y})}}_{=:\boldsymbol{Q}^\star_t(\boldsymbol{y}, \boldsymbol{x})}}\xi_t(\boldsymbol{y})-\pinktext{\underbrace{\boldsymbol{Q}^0_t(\boldsymbol{x}, \boldsymbol{y})e^{V_t(\boldsymbol{x})-V_t(\boldsymbol{y})}}_{=:\boldsymbol{Q}^\star_t(\boldsymbol{x}, \boldsymbol{y})}}\xi_t(\boldsymbol{x})\bigg]\nonumber\\
    &=\sum_{\boldsymbol{x}\neq\boldsymbol{y}}\left(\boldsymbol{Q}^\star_t(\boldsymbol{y}, \boldsymbol{x})\xi_t(\boldsymbol{y})-\boldsymbol{Q}^\star_t(\boldsymbol{x}, \boldsymbol{y})\xi_t(\boldsymbol{x})\right)
\end{align}
\end{small}
which is exactly the Kolmogorov forward equation for the generator $\boldsymbol{Q}^\star_t$ that defines the optimal path measure $\mathbb{P}^\star$. Since we prove in Lemma \ref{lemma:ctmc-kolmogorov-fwd} that the solution to the Kolmogorov forward equation is \textbf{unique}, we have shown that $p^\star_t(\boldsymbol{x})=\frac{1}{Z}q_t(\boldsymbol{x})e^{-V_t(\boldsymbol{x})}$. We derive $Z_t$ such that the probability distribution is normalized, i.e. $\sum_{\boldsymbol{x}\in \mathcal{X}}p^\star_t(\boldsymbol{x})=1$, as follows:
\begin{small}
\begin{align}
    \sum_{\boldsymbol{x}\in \mathcal{X}}p^\star_t(\boldsymbol{x})=\frac{1}{\bluetext{Z_t}}\sum_{\boldsymbol{x}\in \mathcal{X}}q_t(\boldsymbol{x})e^{-V_t(\boldsymbol{x})}=1\implies \bluetext{Z_t}=\sum_{\boldsymbol{x}\in \mathcal{X}}q_t(\boldsymbol{x})e^{-V_t(\boldsymbol{x})}= \mathbb{E}_{\boldsymbol{x}\sim q_t}\left[e^{-V_t(\boldsymbol{x})}\right]
\end{align}
\end{small}
which concludes our proof of the optimal path measure. \hfill $\square$

Using this form of the optimal path measure $p^\star_t(\boldsymbol{x})=\frac{1}{Z}q_t(\boldsymbol{x})e^{-V_t(\boldsymbol{x})}$, we can now derive the RND between the optimal and reference path measures.

\begin{proposition}[Radon-Nikodym Derivative of Optimal and Reference Path Measure]\label{prop:ctmc-rnd-optimal}
    The \textbf{Radon-Nikodym Derivative} (RND) of the optimal path measure $\mathbb{P}^\star$ with generator $\boldsymbol{Q}^\star_t$ and the reference path measure $\mathbb{Q}$ with generator $\boldsymbol{Q}^0_t$ is given by:
    \begin{align}
        \frac{\mathrm{d}\mathbb{P}^\star}{\mathrm{d}\mathbb{Q}}(\boldsymbol{X}_{0:T})=\frac{1}{Z}e^{-\Phi(\boldsymbol{X}_T)}, \quad Z:=\mathbb{E}_{\boldsymbol{x}\sim q_T}\left[e^{-\Phi(\boldsymbol{X}_T)}\right]\tag{Optimal RND}\label{eq:soc-optimal-rnd}
    \end{align}
    where $\Phi(\boldsymbol{x}):\mathcal{X}\to \mathbb{R}$ is the terminal cost function.
\end{proposition}

\textit{Proof.} Starting from the definition of RND between two CTMC path measures in Proposition \ref{prop:ctmc-rnd}, we can write:
\begin{small}
\begin{align}
    \log \frac{\mathrm{d}\mathbb{P}^\star}{\mathrm{d}\mathbb{Q}}(\boldsymbol{X}_{0:T})=\log \frac{p_0^\star(\boldsymbol{X}_0)}{q_0(\boldsymbol{X}_0)}+\sum_{t:\boldsymbol{X}_{t^-}\neq \boldsymbol{X}_t}\log \frac{\boldsymbol{Q}_t^\star(\boldsymbol{X}_{t^-}, \boldsymbol{X}_t)}{\boldsymbol{Q}^0_t(\boldsymbol{X}_{t^-}, \boldsymbol{X}_t)}+\int_0^T\sum_{\boldsymbol{y}\neq\boldsymbol{X}_t}(\boldsymbol{Q}^0_t-\boldsymbol{Q}^\star_t)(\boldsymbol{X}_t,\boldsymbol{y})\mathrm{d}t
\end{align}
\end{small}

Now, we can substitute the expression for the optimal path probability $p^\star_0(\boldsymbol{x})=\frac{1}{Z}q_0(\boldsymbol{x})e^{-V_0(\boldsymbol{x})}$ from Proposition \ref{prop:ctmc-optimal-path} and the optimal generator $\boldsymbol{Q}^\star(\boldsymbol{x},\boldsymbol{y})=\boldsymbol{Q}^0_t(\boldsymbol{x},\boldsymbol{y})e^{V_t(\boldsymbol{x})-V_t(\boldsymbol{y})}$ from Proposition \ref{prop:ctmc-optimal-generator} to get:
\begin{small}
\begin{align}
    \log \frac{\mathrm{d}\mathbb{P}^\star}{\mathrm{d}\mathbb{Q}}(\boldsymbol{X}_{0:T})&=\log \frac{\bluetext{\frac{1}{Z_0}q_0e^{-V_0(\boldsymbol{X}_0)}}}{q_0(\boldsymbol{X}_0)}+\sum_{t:\boldsymbol{X}_{t^-}\neq \boldsymbol{X}_t}\log \frac{\bluetext{\boldsymbol{Q}^0_t(\boldsymbol{X}_{t^-},\boldsymbol{X}_t)e^{V_t(\boldsymbol{X}_{t^-})-V_t(\boldsymbol{X}_t)}}}{\boldsymbol{Q}^0_t(\boldsymbol{X}_{t^-}, \boldsymbol{X}_t)}\nonumber\\
    &+\int_0^T\sum_{\boldsymbol{y}\neq\boldsymbol{X}_t}\left(\boldsymbol{Q}^0_t(\boldsymbol{X}_t,\boldsymbol{y})-\bluetext{\boldsymbol{Q}^0_t(\boldsymbol{X}_t,\boldsymbol{y})e^{V_t(\boldsymbol{X}_t)-V_t(\boldsymbol{y})}}\right)\mathrm{d}t\nonumber\\
    =-V_0(\boldsymbol{X}_0)-\log &Z_0+\sum_{t:\boldsymbol{X}_{t^-}\neq \boldsymbol{X}_t}\left(V_t(\boldsymbol{X}_{t^-})-V_t(\boldsymbol{X}_{t})\right)+\int_0^T\sum_{\boldsymbol{y}\neq\boldsymbol{X}_t}\boldsymbol{Q}^0_t(\boldsymbol{X}_t,\boldsymbol{y})\left(1-e^{V_t(\boldsymbol{X}_t )-V_t(\boldsymbol{y})}\right)\mathrm{d}t\label{eq:ctmc-rnd-eq2}
\end{align}
\end{small}

Since a CTMC is a piecewise càdlàg function, we can define jump times $0=t_0< t_1< \dots< t_k< \dots <t_{K-1}<t_K=T$. Then, we have that in the time interval $[t_k, t_{k+1}]$ the state $\boldsymbol{X}_{t_k}$ stays fixed until the left limit $\boldsymbol{X}_{t_{k+1}^-}$ where it jumps to state $\boldsymbol{X}_{t_{k+1}}$. Therefore, we can define the value difference over all time steps as the sum of changes in value at state $\boldsymbol{X}_{t_k}$ over the time interval $[t_k, t_{k+1}]$ and the value change over the jump between states $\boldsymbol{X}_{t_k}$ and $\boldsymbol{X}_{t_{k+1}}$, given by:
\begin{align}
    V_T(\boldsymbol{X}_T)-V_0(\boldsymbol{X}_0)&=\sum_{k=0}^{K-1}\left(\bluetext{V_{t_{k+1}}}(\boldsymbol{X}_{t_k})-\bluetext{V_{t_k}}(\boldsymbol{X}_{t_k})\right)+\sum_{k=1}^{K-1}\left(V_{t_k}(\pinktext{\boldsymbol{X}_{t_k}})-V_{t_k}(\pinktext{\boldsymbol{X}_{t_{k-1}}}) \right)\nonumber\\
    &=\sum_{k=0}^{K-1}\int_{t_k}^{t_{k+1}}\partial_tV_t(\boldsymbol{X}_{t})dt+ \sum_{t:\boldsymbol{X}_{t^-}\neq \boldsymbol{X}_t}\left(V_t(\boldsymbol{X}_t)-V_t(\boldsymbol{X}_{t^-})\right)\nonumber\\
    &=\int_0^T\partial_tV_t(\boldsymbol{X}_{t})dt+ \sum_{t:\boldsymbol{X}_{t^-}\neq \boldsymbol{X}_t}\left(V_t(\boldsymbol{X}_t)-V_t(\boldsymbol{X}_{t^-})\right)
\end{align}
Isolating $V_0(\boldsymbol{X}_0)$ and substituting the HJB equation from Corollary \ref{corollary:ctmc-hjb}, we get the expression: 
\begin{small}
\begin{align}
    -V_0(\boldsymbol{X}_0)&=-V_T(\boldsymbol{X}_T)+\int_0^T\bluetext{\partial_tV_t(\boldsymbol{X}_{t})}dt+ \sum_{t:\boldsymbol{X}_{t^-}\neq \boldsymbol{X}_t}\left(V_t(\boldsymbol{X}_t)-V_t(\boldsymbol{X}_{t^-})\right)\nonumber\\
    &=-V_T(\boldsymbol{X}_T)+\int_0^T\bluetext{\sum_{\boldsymbol{y}\neq \boldsymbol{X}_t}\boldsymbol{Q}_t^0(\boldsymbol{X}_t,\boldsymbol{y})\left(e^{V_t(\boldsymbol{X}_t)-V_t(\boldsymbol{y})}-1\right)}dt+ \sum_{t:\boldsymbol{X}_{t^-}\neq \boldsymbol{X}_t}\left(V_t(\boldsymbol{X}_t)-V_t(\boldsymbol{X}_{t^-})\right)\label{eq:ctmc-rnd-eq-soc}
\end{align}
\end{small}
Finally, substituting (\ref{eq:ctmc-rnd-eq-soc}) back into (\ref{eq:ctmc-rnd-eq2}) and cancelling terms, we get: 
\begin{align}
    \log \frac{\mathrm{d}\mathbb{P}^\star}{\mathrm{d}\mathbb{Q}}(\boldsymbol{X}_{0:T})=-V_T(\boldsymbol{X}_T)-\log Z_0\implies \boxed{\frac{\mathrm{d}\mathbb{P}^\star}{\mathrm{d}\mathbb{Q}}(\boldsymbol{X}_{0:T})=\frac{1}{Z_0}e^{-V_T(\boldsymbol{X}_T)}}
\end{align}
which yields the form of the RND between the CTMC that solves the SOC problem and the reference path measure. Substituting $V_T(\boldsymbol{X}_T)=\Phi(\boldsymbol{X}_T)$ yields the final result. \hfill $\square$

Having derived the stochastic optimal control formulation for CTMCs, we have shown that the optimal dynamics arise from an exponential tilting of the reference generator and that the resulting optimal path measure $\mathbb{P}^\star$ admits a simple Radon–Nikodym derivative with respect to the reference CTMC $\mathbb{Q}$. In particular, the likelihood ratio depends only on the terminal value function $V_T(\boldsymbol{X}_T)$, revealing that the SOC solution can be interpreted as an entropy-regularized change of measure on path space. 

From this perspective, solving the SOC problem is equivalent to computing a KL projection of path measures, where the optimal controlled process is the closest process to the reference dynamics that satisfies the desired terminal value constraint. This observation provides a direct connection to the Schrödinger bridge problem. In the next section, we explicitly formulate discrete Schrödinger bridges using the SOC framework, allowing the objectives developed in Section \ref{subsec:soc-objectives} to be adapted to discrete state spaces and enabling practical algorithms for learning optimal CTMC dynamics.

\subsection{Discrete Schrödinger Bridges with Stochastic Optimal Control}
\label{subsec:discrete-sb-soc}
\textit{Prerequisite: Section \ref{subsec:soc-objectives}}

Given our construction of SOC for CTMCs in Section \ref{subsec:soc-ctmcs}, we can now explicitly write the (\ref{eq:discrete-sb-problem}) in the form of a stochastic optimal control functional, which contains a running cost that corresponds to the KL divergence between the controlled bridge measure $\mathbb{P}^u$ and the reference measure $\mathbb{Q}$ and a terminal cost that ensures the optimal process satisfies the terminal constraint $\boldsymbol{X}_T\sim \pi_T$.

\begin{definition}[Discrete Schrödinger Bridge Problem with SOC]\label{def:discrete-sb-soc}
    Consider the discrete SB problem where $\boldsymbol{Q}^0_t$ is the generator of the reference CTMC $\mathbb{Q}$ and $\pi_0, \pi_T\in \mathcal{P}(\mathcal{X})$ are the initial and terminal constraints on the finite state space $\mathcal{X}$. The \textbf{discrete Schrödinger bridge problem} can be formulated as a \textbf{stochastic optimal control} (SOC) problem which seeks a controlled generator $\boldsymbol{Q}^u_t$ of a controlled CTMC $\mathbb{P}^u$ that minimizes:
    \begin{small}
    \begin{align}
        \inf_{\mathbb{P}^u}\mathbb{E}_{\boldsymbol{X}^u_{0:T}\sim \mathbb{P}^u}\left[\int_0^T\sum_{\boldsymbol{y}\neq \boldsymbol{X}^u_t}C_t(\boldsymbol{X}^u_t,\boldsymbol{y})dt +\Phi(\boldsymbol{X}^u_T)\right], \quad \boldsymbol{X}^u_0\sim \pi_0\tag{Discrete SB-SOC Objective}\label{eq:discrete-sb-soc-loss}
    \end{align}
    \end{small}
    where the running cost $C_t(\boldsymbol{x},\boldsymbol{y}):\mathcal{X}\times \mathcal{X}\to \mathbb{R}$ and terminal cost $\Phi(\boldsymbol{x}):\mathcal{X}\to \mathbb{R}$ are defined as:
    \begin{small}
    \begin{align}
        C_t(\boldsymbol{X}^u_t,\boldsymbol{y})&:=\text{KL}(p^u_t\|q_t)=\left(\boldsymbol{Q}_t^u\log \frac{\boldsymbol{Q}_t^u}{\boldsymbol{Q}_t^0}-\boldsymbol{Q}_t^u+\boldsymbol{Q}^0_t\right)(\boldsymbol{X}^u_t,\boldsymbol{y}), \quad \Phi(\boldsymbol{X}^u_T)=\log \frac{\hat\varphi(\boldsymbol{X}^u_T)}{\pi_T(\boldsymbol{X}^u_T)}
    \end{align}
    \end{small}
    The optimal controlled process $\mathbb{P}^\star$ defines the Schrödinger bridge between $\pi_0$ and $\pi_T$ relative to the reference dynamics $\boldsymbol{Q}_t^0$.
\end{definition}

Now that we have established that the SOC problem can be solved in the discrete state space, we can easily adapt the objectives defined in Section \ref{subsec:soc-objectives} to solve the SB-SOC problem for discrete variables \citep{nusken2021solving, zhu2025mdns}.

\begin{definition}[Discrete Relative Entropy (RE) Loss]\label{def:discrete-relative-entropy}
    The \textbf{relative entropy} (RE) loss between the controlled path measure $\mathbb{P}^u$ and the optimal path measure $\mathbb{P}^\star$ is defined as the KL divergence:
    \begin{align}
        \mathcal{L}_{\text{RE}}(\mathbb{P}^u, \mathbb{P}^\star):=\text{KL}(\mathbb{P}^u\|\mathbb{P}^\star)=\mathbb{E}_{\mathbb{P}^u}\left[\log \frac{\mathrm{d}\mathbb{P}^u}{\mathrm{d}\mathbb{P}^\star}\right]=\mathbb{E}_{\mathbb{P}^u}\left[\log \frac{\mathrm{d}\mathbb{P}^u}{\mathrm{d}\mathbb{Q}}-\log \frac{\mathrm{d}\mathbb{P}^\star}{\mathrm{d}\mathbb{Q}}\right]\tag{Discrete RE Objective}\label{eq:discrete-relative-entropy-path}
    \end{align}
    which can be written in terms of the controlled and reference generator $\boldsymbol{Q}^u$ and $\boldsymbol{Q}^0$ as:
    \begin{small}
    \begin{align}
        \mathcal{L}_{\text{RE}}(\boldsymbol{Q}^u):=\mathbb{E}_{\boldsymbol{X}^u_{0:T}\sim \mathbb{P}^u}\bigg[\underbrace{\int_0^T\sum_{\boldsymbol{y}\neq \boldsymbol{X}^u_t}\left(\boldsymbol{Q}_t^u\log \frac{\boldsymbol{Q}^u_t}{\boldsymbol{Q}^0_t}+\boldsymbol{Q}_t^0-\boldsymbol{Q}_t^u\right)(\boldsymbol{X}^u_t,\boldsymbol{y})}_{\log \frac{\mathrm{d}\mathbb{P}^u}{\mathrm{d}\mathbb{Q}}}+\underbrace{V_T(\boldsymbol{X}^u_T)+\log Z_0}_{-\log \frac{\mathrm{d}\mathbb{P}^\star}{\mathrm{d}\mathbb{Q}}}\bigg]\tag{Discrete RE Loss}\label{eq:relative-entropy-discrete}
    \end{align}
    \end{small}
\end{definition}

As discussed in Section \ref{subsec:soc-objectives}, the (\ref{eq:discrete-relative-entropy-path}) requires backpropagating through the full stochastic trajectory simulations, so we can use (\ref{eq:rerf-loss}) as a practical surrogate loss for (\ref{eq:discrete-relative-entropy-path}).

Next, we will adapt the (\ref{eq:soc-cross-entropy}) to the discrete state space. Similarly to Definition \ref{def:cross-entropy}, the expectation over the optimal path measure $\mathbb{P}^\star$ is generally intractable during training, so we define the expectation over an arbitrary path measure $\mathbb{P}^v$ and \textit{reweighting} the log RND by the RND between $\mathbb{P}^\star$ and $\mathbb{P}^v$.

\begin{definition}[Cross-Entropy Objective]\label{def:discrete-cross-entropy}
    The \textbf{cross-entropy} (CE) loss that optimizes a controlled path measure $\mathbb{P}^u$ to match the optimal path measure $\mathbb{P}^\star$ is defined as the reverse KL divergence:
    \begin{align}
        \mathcal{L}_{\text{CE}}(\mathbb{P}^u, \mathbb{P}^\star)&:=\text{KL}(\mathbb{P}^\star\|\mathbb{P}^u)=\mathbb{E}_{\mathbb{P}^\star}\left[\log \frac{\mathrm{d}\mathbb{P}^\star}{\mathrm{d}\mathbb{P}^u}\right]=\mathbb{E}_{\mathbb{P}^v}\left[\frac{\mathrm{d}\mathbb{P}^\star }{\mathrm{d}\mathbb{P}^v}\log \frac{\mathrm{d}\mathbb{P}^\star}{\mathrm{d}\mathbb{P}^u}\right]\nonumber\\
        &=\mathbb{E}_{\mathbb{P}^v}\bigg[\frac{\mathrm{d}\mathbb{P}^\star }{\mathrm{d}\mathbb{P}^v}\bigg(\underbrace{\log \frac{\mathrm{d}\mathbb{P}^\star}{\mathrm{d}\mathbb{Q}}}_{\text{constant w.r.t. }\mathbb{P}^u}+\log \frac{\mathrm{d}\mathbb{Q}}{\mathrm{d}\mathbb{P}^u}\bigg)\bigg]=\mathbb{E}_{\mathbb{P}^v}\bigg[\frac{\mathrm{d}\mathbb{P}^\star }{\mathrm{d}\mathbb{P}^v}\log \frac{\mathrm{d}\mathbb{Q}}{\mathrm{d}\mathbb{P}^u}\bigg]+C
    \end{align}
    where $\log \frac{\mathrm{d}\mathbb{P}^\star}{\mathrm{d}\mathbb{Q}}$ vanishes in the gradient as it is an additive constant with respect to $\mathbb{P}^u$. The CE objective can be written in terms of the controlled and reference generator $\boldsymbol{Q}^u$ and $\boldsymbol{Q}^0$ as:
    \begin{small}
    \begin{align}
        \mathcal{L}_{\text{CE}}(\boldsymbol{Q}^u)=\frac{1}{Z}\mathbb{E}_{\boldsymbol{X}_{0:T}^v\sim\mathbb{P}^v}\bigg[e^{W(\boldsymbol{X}_{0:T}^v)}\bigg(\int_0^T\sum_{\boldsymbol{y}\neq \boldsymbol{X}^v_t}\left(\boldsymbol{Q}^0_t\log \frac{\boldsymbol{Q}^0_t}{\boldsymbol{Q}^u_t}+\boldsymbol{Q}_t^u-\boldsymbol{Q}_t^0\right)(\boldsymbol{X}_t,\boldsymbol{y})\bigg)\bigg]+C\label{eq:discrete-cross-entropy}
    \end{align}
    \end{small}
    where $W(\boldsymbol{X}_{0:T}^v):=\log \frac{\mathrm{d}\mathbb{P}^\star}{\mathrm{d}\mathbb{P}^v}(\boldsymbol{X}_{0:T}^v)$ is a weight that can be expanded as:
    \begin{small}
    \begin{align}
        W(\boldsymbol{X}_{0:T}^v)&:=\log\frac{\mathrm{d}\mathbb{P}^\star}{\mathrm{d}\mathbb{P}^v}(\boldsymbol{X}_{0:T}^v)=\log \frac{\mathrm{d}\mathbb{P}^\star}{\mathrm{d}\mathbb{Q}}(\boldsymbol{X}_{0:T}^v)-\frac{\mathrm{d}\mathbb{P}^v}{\mathrm{d}\mathbb{Q}}(\boldsymbol{X}_{0:T}^v)\nonumber\\
        &=\int_0^T\sum_{\boldsymbol{y}\neq \boldsymbol{X}^v_t}\left(\boldsymbol{Q}_t^v\log \frac{\boldsymbol{Q}^v_t}{\boldsymbol{Q}^0_t}+\boldsymbol{Q}_t^0-\boldsymbol{Q}_t^v\right)(\boldsymbol{X}^v_t,\boldsymbol{y})+V_T(\boldsymbol{X}^v_T)+\log Z_0
    \end{align}
    \end{small}
    The sampling law is commonly defined in practice as the stop-gradient sampling measure $\mathbb{P}^{\bar{u}}$ like in $\mathcal{L}_{\text{RERF}}$, where $\boldsymbol{Q}^{\bar{u}}:=\texttt{stopgrad}(\boldsymbol{Q}^u)$ is the non-gradient-tracking controlled generator. 
\end{definition}

The final objective introduced for continuous SOC in Section \ref{subsec:soc-objectives} is the \boldtext{variance and log-variance objectives}. Recall that the RND between the optimal and controlled path measure is high when the two distributions are dissimilar, in the sense that paths with high probability under $\mathbb{P}^\star$ have a low probability under $\mathbb{P}^u$ and vice versa. On the other hand, the variance is low when the distributions are similar and is minimized at zero exactly when the RND evaluates to a constant regardless of the stochastic path, indicating that the two measures are equal. Since the variance objective is generally unstable, we only define the log-variance objective below.

\begin{definition}[Log-Variance Objectives]
    The \textbf{log-variance} (LV) loss that optimizes a controlled path measure $\mathbb{P}^u$ to match the optimal path measure $\mathbb{P}^\star$ is defined as:
    \begin{align}
        \mathcal{L}_{\text{LV}}(\mathbb{P}^u, \mathbb{P}^\star)&:=\text{Var}_{\mathbb{P}^v}\left(\log \frac{\mathrm{d}\mathbb{P}^\star}{\mathrm{d}\mathbb{P}^u}\right)=\text{Var}_{\mathbb{P}^v}\left(\log \frac{\mathrm{d}\mathbb{P}^\star}{\mathrm{d}\mathbb{Q}}-\log \frac{\mathrm{d}\mathbb{P}^u}{\mathrm{d}\mathbb{Q}}\right)
    \end{align}
    which can be written in terms of the controlled and reference generator $\boldsymbol{Q}^u$ and $\boldsymbol{Q}^0$ as:
    \begin{small}
    \begin{align}
        \mathcal{L}_{\text{LV}}(\boldsymbol{Q}^u)=\text{Var}_{\boldsymbol{X}_{0:T}^v\sim \mathbb{P}^v}\bigg(\underbrace{\int_0^T\sum_{\boldsymbol{y}\neq \boldsymbol{X}^v_t}\left(\boldsymbol{Q}_t^u\log \frac{\boldsymbol{Q}^u_t}{\boldsymbol{Q}^0_t}+\boldsymbol{Q}_t^0-\boldsymbol{Q}_t^u\right)(\boldsymbol{X}^v_t,\boldsymbol{y})}_{\log \frac{\mathrm{d}\mathbb{P}^u}{\mathrm{d}\mathbb{Q}}(\boldsymbol{X}^v_{0:T})}+\underbrace{V_T(\boldsymbol{X}^v_T)+\log Z_0}_{-\log \frac{\mathrm{d}\mathbb{P}^\star}{\mathrm{d}\mathbb{Q}}(\boldsymbol{X}^v_{0:T})}\bigg)\label{eq:discrete-log-variance}
    \end{align}
    \end{small}
\end{definition}
These objectives allow us to tractably solve the (\ref{eq:discrete-sb-soc-loss}) by simulating the controlled CTMC $\mathbb{P}^u$ over discrete time steps while simultaneously evaluating the transition rates of each jump under the reference generator $\boldsymbol{Q}^0$, computing the SOC objective using the time-discretized form of the log RND, and optimizing $\boldsymbol{u}$ using standard gradient descent. For uniform discrete time steps $0=t_0< \dots< t_k< \dots< t_K=T$ with increments $\Delta t$, the log RND over the discrete states $(\boldsymbol{X}_{t_k})_{k\in \{0, \dots, K\}}$ used to compute the losses can be computed as: 
\begin{small}
\begin{align}
    \log \frac{\mathrm{d}\mathbb{P}^u}{\mathrm{d}\mathbb{Q}}((\boldsymbol{X}_{t_k})_{k\in \{0, \dots, K\}})=\sum_{k=0}^{K-1}\left[\Delta t\left(\boldsymbol{Q}_{t_k}^u\log \frac{\boldsymbol{Q}^u_{t_k}}{\boldsymbol{Q}^0_{t_k}}+\boldsymbol{Q}_{t_k}^0- \boldsymbol{Q}_{t_k}^u\right)(\boldsymbol{X}_{t_k},\boldsymbol{X}_{t_{k+1}})\right]+\mathcal{O}(\Delta t^2)\tag{Discretized Log RND}\label{eq:discretized-logrnd}
\end{align}
\end{small}
Then, we can define a general form of the (\ref{eq:discrete-sb-soc-loss}) training procedure \citep{zhu2025mdns, tang2025tr2} below:
\purple[Discrete SB-SOC Training Framework]{\label{box:discrete-sbsoc}
Training a parameterized generator $\boldsymbol{Q}^u$ using one of the SOC objectives defined above typically iterates through the following steps:
\begin{enumerate}
    \item [(i)] Sample descretized trajectories $(\boldsymbol{X}^v_{t_k})_{t\in \{0, \dots, K\}}$ using the proposal generator $\boldsymbol{Q}^v$ for $K$ steps. Alternatively, if the proposal generator is not $\boldsymbol{u}$ and trajectories can be reused, then samples can be stored and subsequently sampled from a replay buffer $\mathcal{B}$.
    \item[(ii)] For each trajectory, compute the transition rates $\boldsymbol{Q}_{t_k}^0(\boldsymbol{X}_{t_k}, \boldsymbol{X}_{t_{k+1}})$ and $\boldsymbol{Q}_{t_k}^u(\boldsymbol{X}_{t_k}, \boldsymbol{X}_{t_{k+1}})$ over every interval $[t_k ,t_{k+1}]$.
    \item[(ii)] Compute the (\ref{eq:discretized-logrnd}) for each trajectory and use it to compute one of the SOC losses $\mathcal{L}(\boldsymbol{Q}^u)$.
    \item[(iii)] Optimize the parameterized generator $\boldsymbol{Q}^u$ using gradient steps of $\nabla_{\boldsymbol{u}}\mathcal{L}(\boldsymbol{Q}^u)$
    \item [(iv)] Repeat from Step (i).

\end{enumerate}
In practice, the proposal generator $\boldsymbol{Q}^v$ is commonly defined as the same generator being optimized $\boldsymbol{Q}^u$ but without gradient-tracking, i.e., $\boldsymbol{Q}^v:=\boldsymbol{Q}^{\bar{u}}=\texttt{stopgrad}(\boldsymbol{Q}^u)$. 
}

The SOC framework provides a practical procedure for learning the controlled generator $\boldsymbol{Q}^u$ of a CTMC through sampling discrete trajectories from a proposal generator $\boldsymbol{Q}^v$ and optimizing the SOC objective with respect to $\boldsymbol{Q}^u$. While this process is well-suited for tasks where we have no access to samples from the target distribution $\pi_T$ but only a way of evaluating the likelihood of a sample under it, optimizing a Schrödinger bridge on \textit{paired samples} or an explicit optimal transport map requires a different approach. Next, we will describe the extension of Markovian and reciprocal projections in the discrete state space, which allows us to optimize discrete Schrödinger bridges directly on samples from the initial and target distributions, or pairs from an optimal transport coupling $\pi^\star_{0,T}$.

\subsection{Discrete Markov and Reciprocal Projections}
\label{subsec:discrete-markov-reciprocal}
\textit{Prerequisite: Section \ref{subsec:markov-reciprocal-proj}}

In this section, we extend the theory of Markov and reciprocal projections to discrete state spaces, where stochastic processes are characterized by the generators of continuous-time Markov chains (CTMCs). Recall from Section \ref{subsec:markov-reciprocal-proj} that the \boldtext{Markovian projection} identifies the Markov process that is closest in KL divergence to a given reciprocal process within the class of Markov dynamics, while the \boldtext{reciprocal projection} enforces the endpoint constraints through conditioning on the boundary distributions. 
\begin{small}
\begin{align}
    \mathbb{M}^\star&:=\text{proj}_{\mathcal{M}}(\Pi)=\underset{\mathbb{M}\in \mathcal{M}}{\arg\min}\text{KL}\left(\Pi\|\mathbb{M}\right)\tag{Markovian Projection}\\
    \Pi^\star&:=\mathbb{Q}_{\cdot|0,T}\mathbb{M}_{0,T}=\text{proj}_{\mathcal{R}(\mathbb{Q})}(\mathbb{P})=\int_{\mathbb{R}^d\times \mathbb{R}^d}\mathbb{Q}_{\cdot|0,T}(\cdot|\boldsymbol{x}_0, \boldsymbol{x}_T)\mathrm{d}\mathbb{M}_{0,T}(\boldsymbol{x}_0, \boldsymbol{x}_T)\tag{Reciprocal Projection}
\end{align}
\end{small}

Since this section is quite notation-heavy, we will first establish the meaning behind some and shorthand for the notation. 

\begingroup
\begin{center}
\begin{longtable}{@{\extracolsep{\fill}}>{\centering\arraybackslash}p{1.8in} p{4.0in}@{}}
\hline
\multicolumn{2}{l}{\textbf{Notation for Section \ref{subsec:discrete-markov-reciprocal} and \ref{subsec:ddsbm}}} \\
\hline
\textit{Notation} & \textit{Meaning} 
\\
\hline
$\mathbb{Q}$ & reference CTMC path measure \\
$\boldsymbol{Q}^0_t(\boldsymbol{x},\boldsymbol{y})$ & generator of the reference process $\mathbb{Q}$ \\
$\mathbb{M}$ & Markov CTMC path measure in the Markov class $\mathcal{M}$ \\
$\Pi$ & CTMC path measure in the reciprocal class $\mathcal{R}(\mathbb{Q})$ \\
$\mathbb{M}_t, \Pi_t$ & marginal distributions at time $t$ under a Markov or reciprocal path measure \\
$\mathbb{M}^\star$ & Markov projection of a reciprocal bridge measure $\Pi$ \\
$\boldsymbol{Q}^{\mathbb{M}^\star}(\boldsymbol{x},\boldsymbol{y})$ & generator of Markovian projection $\mathbb{M}^\star$ \\
$\Pi^\star$ & reciprocal projection of a Markov measure $\mathbb{M}$ \\
$\Pi^{\cdot|0}=\Pi^{\cdot|0=\boldsymbol{x}_0}$ & mixture of bridges conditioned on initial state $\boldsymbol{X}_0=\boldsymbol{x}_0$. Transitions under the conditioned measure, denoted $\mathbb{Q}^{\cdot|0=\boldsymbol{x}_0}_{t|s}(\boldsymbol{y}|\boldsymbol{x})$, are also conditioned on $\boldsymbol{x}_0$ \\
$\boldsymbol{Q}^{\Pi^{\cdot|0}}(\boldsymbol{x},\boldsymbol{y})=\boldsymbol{Q}^{\Pi^{\cdot|0=\boldsymbol{x}_0}}(\boldsymbol{x},\boldsymbol{y})$ & generator of reciprocal process conditioned on some initial state $\boldsymbol{X}_0=\boldsymbol{x}_0$\\
$\mathbb{Q}^{\cdot|T}=\mathbb{Q}^{\cdot|T=\boldsymbol{x}_T}$ & reference path measure conditioned on terminal state $\boldsymbol{X}_T=\boldsymbol{x}_T$. Transitions under the conditioned measure, denoted $\mathbb{Q}^{\cdot|T=\boldsymbol{x}_T}_{t|s}(\boldsymbol{y}|\boldsymbol{x})$ are also conditioned on $\boldsymbol{x}_T$ \\
$\boldsymbol{Q}_t^0(\boldsymbol{x}, \boldsymbol{y}; \boldsymbol{x}_T)$ & generator of the reference process $\mathbb{Q}$ conditioned on the terminal state $\boldsymbol{X}_T=\boldsymbol{x}_T$ \\
$\mathbb{Q}^{\cdot|0}=\mathbb{Q}^{\cdot|0=\boldsymbol{x}_0}$ & reference path measure conditioned on terminal state $\boldsymbol{X}_0=\boldsymbol{x}_0$. Transitions under the conditioned measure, denoted $\mathbb{Q}^{\cdot|0=\boldsymbol{x}_0}_{t|s}(\boldsymbol{y}|\boldsymbol{x})$ are also conditioned on $\boldsymbol{x}_0$ \\
$\tilde{\boldsymbol{Q}}_s^0(\boldsymbol{x}, \boldsymbol{y}; \boldsymbol{x}_0)$ & reverse-time generator of the reference process $\mathbb{Q}$ conditioned on the state $\boldsymbol{X}_0=\boldsymbol{x}_0$ \\
$\mathbb{Q}_{t|s}(\boldsymbol{y}|\boldsymbol{x}), \Pi_{t|s}(\boldsymbol{y}|\boldsymbol{x})$ & transition density from state $\boldsymbol{x}$ at time $s$ from state $\boldsymbol{y}$ at time $t$ under the specified path measure. Shorthand for $\mathbb{Q}_{t|s}(\boldsymbol{X}_t=\boldsymbol{y}|\boldsymbol{X}_s=\boldsymbol{x}), \Pi_{t|s}(\boldsymbol{X}_t=\boldsymbol{y}|\boldsymbol{X}_s=\boldsymbol{x})$. \\
\hline
\end{longtable}
\end{center}
\endgroup

We will derive the explicit form of the Markovian and reciprocal projections in terms of the generators of CTMCs, which will serve as the theoretical basis for extending the iterative Markovian fitting (IMF) procedure from Section \ref{subsec:diffusionsbm} to the finite state space.
\begin{enumerate}
    \item [(i)] First, we define the Markovian projection $\mathbb{M}^\star$ of a CTMC given a bridge measure in the reciprocal class $\Pi\in \mathcal{R}(\mathbb{Q})$. We define the explicit form of its generator $\boldsymbol{Q}^{\mathbb{M}^\star}$ and the KL divergence with the reciprocal process (Definition \ref{def:ctmc-markov-proj}).
    \item[(ii)] We derive the definition for the reference generator $\boldsymbol{Q}^0_t$ conditioned on a terminal state $\boldsymbol{x}_T$, which appears in the expression for the generator of the Markovian projection (Lemma \ref{lemma:ctmc-cond-generator-markov}).
    \item[(iii)] We show that conditioning the bridge measure in the reciprocal class $\Pi\in \mathcal{R}(\mathbb{Q})$ on an initial state $\boldsymbol{X}_0=\boldsymbol{x}_0$ yields a Markov measure, which we denote $\Pi^{\cdot|0=\boldsymbol{x}_0}$ (Lemma \ref{lemma:ctmc-cond-generator-reciprocal}).
    \item[(iv)] We derive the form of the KL divergence $\text{KL}(\Pi\|\mathbb{M})$ using the generator of the conditioned reciprocal process $\boldsymbol{Q}_t^{\Pi^{\cdot|0=\boldsymbol{x}_0}}$ and the generator of the Markov process $\mathbb{M}$ (Lemma \ref{lemma:generator-markov-proj}).
    \item [(v)] Finally, we define the \textbf{reverse-time} Markovian projection, which will allow us to condition on both the initial distribution $\pi_0$ and target distribution $\pi_T$, reducing error accumulation during the IMF procedure (Definition \ref{def:reverse-time-markov-proj}).
\end{enumerate}

We start by defining the \textbf{Markovian projection $\mathbb{M}^\star$ of a CTMC path measure}, its explicit form in terms of the generator matrices and the KL minimization objective that yields $\mathbb{M}^\star$ given a reciprocal measure $\Pi$.

\begin{definition}[Markovian Projection of CTMC Path Measure]\label{def:ctmc-markov-proj}
    Consider a reference path measure $\mathbb{Q}$ with generator $\boldsymbol{Q}^0_t$ and a measure $\Pi\in \mathcal{R}(\mathbb{Q})$ in the reciprocal class of $\mathbb{Q}$ that preserves the bridge. The \textbf{Markovian projection} $\mathbb{M}^\star:=\text{proj}_{\mathcal{M}}(\Pi)$ has a generator $\boldsymbol{Q}^{\mathbb{M}^\star}_t$ defined as the expectation over \textbf{conditional generators} $\boldsymbol{Q}^0_t(\cdot, \cdot;\boldsymbol{x})$ under the reference process $\mathbb{Q}$ of the form:
    \begin{small}
    \begin{align}
        \boldsymbol{Q}^{\mathbb{M}^\star}_t(\boldsymbol{x},\boldsymbol{y})&=\mathbb{E}_{\boldsymbol{x}_T\sim \mathbb{Q}_{T|t}(\cdot|\boldsymbol{x})}\left[\bluetext{\boldsymbol{Q}^0_t(\boldsymbol{x},\boldsymbol{y}; \boldsymbol{x}_T)}\big|\boldsymbol{X}_t=\boldsymbol{x}\right]\tag{Generator of Markovian Projection}\label{eq:ctmc-markov-projection-generator}
    \end{align}
    \end{small}
    where each conditional generator is defined as:
    \begin{small}
    \begin{align}
        \bluetext{\boldsymbol{Q}^0_t(\boldsymbol{x},\boldsymbol{y}; \boldsymbol{x}_T)}&=\boldsymbol{Q}^0_t(\boldsymbol{x}, \boldsymbol{y})\frac{\mathbb{Q}_{T|t}(\boldsymbol{x}_T| \boldsymbol{y})}{\mathbb{Q}_{T|t}(\boldsymbol{x}_T| \boldsymbol{x})}-\boldsymbol{1}_{\boldsymbol{x}=\boldsymbol{y}}\sum_{\boldsymbol{z}\in \mathcal{X}}\boldsymbol{Q}^0_t(\boldsymbol{x}, \boldsymbol{z})\frac{\mathbb{Q}_{T|t}(\boldsymbol{x}_T| \boldsymbol{z})}{\mathbb{Q}_{T|t}(\boldsymbol{x}_T| \boldsymbol{x})}\tag{Endpoint-Conditioned Generator}\label{eq:ctmc-endpoint-cond-gen}
    \end{align}
    \end{small}
    where $\mathbb{Q}_{T|t}(\cdot|\cdot)$ is the conditional transition probability under the reference measure $\mathbb{Q}$. We can also define the KL divergence between $\Pi$ and its Markovian projection $\mathbb{M}^\star$ using Corollary \ref{corollary:ctmc-kl} as:
    \begin{align}
        \text{KL}(\Pi\|\mathbb{M}^\star)=\int_0^T\mathbb{E}_{\Pi_{0,t}}\left[\sum_{\boldsymbol{y}\neq \boldsymbol{X}_t}\left(\bluetext{\boldsymbol{Q}_t^{\Pi^{\cdot|0}}}\log \frac{\bluetext{\boldsymbol{Q}_t^{\Pi^{\cdot|0}}}}{\boldsymbol{Q}_t^{\mathbb{M}^\star}}+\boldsymbol{Q}_t^{\mathbb{M}^\star}-\bluetext{\boldsymbol{Q}_t^{\Pi^{\cdot|0}}}\right)(\boldsymbol{X}_t,\boldsymbol{y})\right]dt\label{eq:ctmc-markov-kl}
    \end{align}
    where the KL divergence between the initial distributions vanishes as we assume they are aligned. Specifically, we define $\Pi^{\cdot|0=\boldsymbol{x}_0}$ as the conditional bridge measure with generator $\boldsymbol{Q}^{\Pi^{\cdot|0=\boldsymbol{x}_0}}_t$ defined as:
    \begin{align}
        \bluetext{\boldsymbol{Q}^{\Pi^{\cdot|0=\boldsymbol{x}_0}}_t(\boldsymbol{x}, \boldsymbol{y})}=\mathbb{E}_{\boldsymbol{x}_T\sim \Pi_{T|0,t}}\left[\boldsymbol{Q}^0_t(\boldsymbol{x},\boldsymbol{y}; \boldsymbol{x}_T)|\boldsymbol{X}_0=\boldsymbol{x}_0, \boldsymbol{X}_t=\boldsymbol{x}\right]
    \end{align}
    where for any $t\in [0,T]$, the marginal distributions match $\mathbb{M}_t=\Pi_t$.
\end{definition}

Breaking down Definition \ref{def:ctmc-markov-proj}, we introduce several \textit{unfamiliar} definitions, including the \textbf{endpoint-conditioned generator} of the reference process $\boldsymbol{Q}^0_t(\boldsymbol{x},\boldsymbol{y}; \boldsymbol{x}_T)$, the reciprocal bridge measure $\Pi$ conditioned on $\boldsymbol{X}_0=\boldsymbol{x}_0$ denoted $\Pi^{\cdot|0=\boldsymbol{x}_0}$ and its \textbf{generator} defined as $\boldsymbol{Q}^{\Pi^{\cdot|0=\boldsymbol{x}_0}}_t$. In the following sequence of Lemmas, we will break down these ideas more concretely to better understand the discrete analogue of the Markovian projection.

\begin{lemma}[Conditional Generator of Markov Process]\label{lemma:ctmc-cond-generator-markov}
    Consider a CTMC $\boldsymbol{X}_{0:T}$ under the reference path measure $\mathbb{Q}$ with generator $\boldsymbol{Q}^0_t$. Conditioning $\mathbb{Q}$ on a terminal state $\boldsymbol{X}_T=\boldsymbol{x}_T$ gives the conditioned path measure, denoted $\mathbb{Q}^{\cdot|T=\boldsymbol{x}_T}$, which is \textbf{Markov} and is defined by the generator:
    \begin{small}
    \begin{align}
        \boldsymbol{Q}^0_t(\boldsymbol{x},\boldsymbol{y}; \boldsymbol{x}_T)=\boldsymbol{Q}^0_t(\boldsymbol{x},\boldsymbol{y})\frac{\mathbb{Q}_{T|t}(\boldsymbol{x}_T|\boldsymbol{y})}{\mathbb{Q}_{T|t}(\boldsymbol{x}_T|\boldsymbol{x})}-\boldsymbol{1}_{\boldsymbol{x}=\boldsymbol{y}}\sum_{\boldsymbol{z}\in \mathcal{X}}\boldsymbol{Q}^0_t(\boldsymbol{x}, \boldsymbol{z})\frac{\mathbb{Q}_{T|t}( \boldsymbol{x}_T|\boldsymbol{z})}{\mathbb{Q}_{T|t}(\boldsymbol{x}_T|\boldsymbol{x})}
    \end{align}
    \end{small}
\end{lemma}
\textit{Proof.} To confirm the Markov property of $\mathbb{Q}^{\cdot|T=\boldsymbol{x}_T}$, we first apply Bayes' rule\footnote{The standard form of Bayes' rule states $p(y|z) =\frac{p(z|y) \cdot p(y)}{p(z)}$, which we extend to add conditioning on a third variable $x$.} to get:
\begin{small}
\begin{align}
    \mathbb{Q}_{t|s,T}(\boldsymbol{y}|\boldsymbol{x}, \bluetext{\boldsymbol{x}_T})=\underbrace{\frac{\mathbb{Q}_{t|s}(\boldsymbol{y}|\boldsymbol{x})\mathbb{Q}_{T|t}(\bluetext{\boldsymbol{x}_T}|\boldsymbol{y})}{\mathbb{Q}_{T|s}(\bluetext{\boldsymbol{x}_T}|\boldsymbol{x})}}_{\text{Bayes' rule}}=:\bluetext{\mathbb{Q}^{\cdot|T=\boldsymbol{x}_T}_{t|s}(\boldsymbol{y}|\boldsymbol{x})}\label{eq:ctmc-endpoint-cond-prob}
\end{align}
\end{small}
which defines the transition kernel of the bridge process conditioned on $\boldsymbol{X}_T=\boldsymbol{x}_T$. This shows that the conditioned process is still Markov, that is given the current state $\boldsymbol{X}_T=\boldsymbol{x}_T$ and the fixed endpoint $\boldsymbol{X}_T=\boldsymbol{x}_T$, the law of the future state $\boldsymbol{X}_t$ depends on the past only through the current state $\boldsymbol{X}_s$.

Now, applying the \textbf{Kolmogorov forward equation} from Lemma \ref{lemma:ctmc-kolmogorov-fwd} and the \textbf{Kolmogorov backward equation} from Lemma \ref{lemma:ctmc-kolmogorov-bwd}, we obtain the generator of the conditioned process as:
\begin{small}
\begin{align}
    \boldsymbol{Q}^0_t(\boldsymbol{x}, \boldsymbol{y};\boldsymbol{x}_T) &=\partial_s\mathbb{Q}^{\cdot|T=\boldsymbol{x}_T}_{t|s}(\boldsymbol{y}|\boldsymbol{x})\big\vert_{s=t}\nonumber\\
    &=\partial_s\left(\frac{\mathbb{Q}_{t|s}(\boldsymbol{y}|\boldsymbol{x})\mathbb{Q}_{T|t}(\bluetext{\boldsymbol{x}_T}|\boldsymbol{y})}{\mathbb{Q}_{T|s}(\bluetext{\boldsymbol{x}_T}|\boldsymbol{x})}\right)\bigg|_{s=t}\nonumber\\
    &=\bluetext{\partial_s\mathbb{Q}_{t|s}(\boldsymbol{y}|\boldsymbol{x})}\frac{\mathbb{Q}_{T|t}(\boldsymbol{x}_T|\boldsymbol{y})}{\bluetext{\mathbb{Q}_{T|t}}(\boldsymbol{x}_T|\boldsymbol{x})}+\underbrace{\mathbb{Q}_{t|t}(\boldsymbol{y}|\boldsymbol{x})}_{\boldsymbol{1}_{\boldsymbol{x}=\boldsymbol{y}}}\frac{\bluetext{\partial_t\mathbb{Q}_{T|t}(\bluetext{\boldsymbol{x}_T}|\boldsymbol{y})}}{\mathbb{Q}_{T|t}(\bluetext{\boldsymbol{x}_T}|\boldsymbol{x})}\nonumber\\
    &=\bluetext{\boldsymbol{Q}_t^0(\boldsymbol{x}, \boldsymbol{y})}\frac{\mathbb{Q}_{T|t}(\bluetext{\boldsymbol{x}_T}|\boldsymbol{y})}{\mathbb{Q}_{T|t}(\bluetext{\boldsymbol{x}_T}|\boldsymbol{x})}+\boldsymbol{1}_{\boldsymbol{x}=\boldsymbol{y}}\left[\frac{\bluetext{-\sum_{\boldsymbol{z}\in\mathcal{X}}\boldsymbol{Q}_t(\boldsymbol{x}, \boldsymbol{y})\mathbb{Q}_{T|t}(\boldsymbol{x}_T|\boldsymbol{z})}}{\mathbb{Q}_{T|t}(\boldsymbol{x}_T|\boldsymbol{x})}\right]\nonumber\\
    &=\boldsymbol{Q}^0_t(\boldsymbol{x},\boldsymbol{y})\frac{\mathbb{Q}_{T|t}( \boldsymbol{x}_T|\boldsymbol{y})}{\mathbb{Q}_{T|t}( \boldsymbol{x}_T|\boldsymbol{x})}-\boldsymbol{1}_{\boldsymbol{x}=\boldsymbol{y}}\sum_{\boldsymbol{z}\in \mathcal{X}}\boldsymbol{Q}^0_t(\boldsymbol{x}, \boldsymbol{z})\frac{\mathbb{Q}_{T|t}(\boldsymbol{x}_T|\boldsymbol{z})}{\mathbb{Q}_{T|t}(\boldsymbol{x}_T|\boldsymbol{x})}
\end{align}
\end{small}
which recovers for the form of the generator $\boldsymbol{Q}^0_t(\boldsymbol{x}, \boldsymbol{y} ; \boldsymbol{x}_T)$ given the terminal state $\boldsymbol{X}_T=\boldsymbol{x}_T$ under the reference process $\mathbb{Q}$. \hfill $\square$

Next, we will derive the generator for the reciprocal process $\Pi\in \mathcal{R}(\mathbb{Q})$ \textit{conditioned} on the initial state $\boldsymbol{X}_0=\boldsymbol{x}_0$, which is \textit{Markov} as shown in Proposition \ref{eq:prop:reciprocal-process}. This allows us to define the KL divergence between $\Pi$ and its Markovian projection $\mathbb{M}^\star=\text{proj}_{\mathcal{M}}(\Pi)$ which is necessary for optimizing the discrete SB.

\begin{lemma}[KL Divergence Between Reciprocal and Markov CTMCs]\label{lemma:ctmc-cond-generator-reciprocal}
    Given the reciprocal process $\Pi\in \mathcal{R}(\mathbb{Q})$, \textbf{conditioning} on an initial state $\boldsymbol{X}_0=\boldsymbol{x}_0$, yields a \textbf{Markov} process $\Pi^{\cdot|0=\boldsymbol{x}_0}$ that is defined by the generator:
    \begin{align}
        \boldsymbol{Q}^{\Pi^{\cdot|0=\boldsymbol{x}_0}}_t(\boldsymbol{x}, \boldsymbol{y})=\mathbb{E}_{\boldsymbol{x}_T\sim \Pi_{T|0,t}}\left[\boldsymbol{Q}^0_t(\boldsymbol{x}, \boldsymbol{y};\boldsymbol{x}_T)|\boldsymbol{X}_0=\boldsymbol{x}_0, \boldsymbol{X}_t=\boldsymbol{x}\right]
    \end{align}
    Then, for any Markov measure $\mathbb{M}\in \mathcal{M}$ where $\mathbb{M}_0=\Pi_0$ and $\Pi\ll\mathbb{M}$, the KL divergence between $\Pi$ and $\mathbb{M}$ is defined as:
    \begin{small}
    \begin{align}
        \text{KL}(\Pi\|\mathbb{M})&=\mathbb{E}_{\Pi_0}\left[\text{KL}(\Pi^{\cdot|0=\boldsymbol{x}_0}\|\mathbb{M}^{\cdot|0=\boldsymbol{x}_0})\right]
    \end{align}
    \end{small}
    where $\text{KL}(\Pi^{\cdot|0=\boldsymbol{x}_0}\|\mathbb{M}^{\cdot|0=\boldsymbol{x}_0})$ is a KL divergence between the Markov conditioned reciprocal process $\Pi^{\cdot|0=\boldsymbol{x}_0}$ and the conditioned Markov measure $\mathbb{M}^{\cdot|0=\boldsymbol{x}_0}$, which expands to:
    \begin{small}
    \begin{align}
       \text{KL}(\Pi^{\cdot|0}\|\mathbb{M}^{\cdot|0})= &\int_0^T\mathbb{E}_{\Pi_{0,t}}\left[\sum_{\boldsymbol{y}\neq \boldsymbol{X}_t}\left(\boldsymbol{Q}^{\bluetext{\Pi^{\cdot|0}}}_t\log \frac{\boldsymbol{Q}^{\bluetext{\Pi^{\cdot|0}}}_{t}}{\boldsymbol{Q}^{\pinktext{\mathbb{M}^{\cdot|0}}}_{t}}\right)(\boldsymbol{X}_t, \boldsymbol{y})+\left(\boldsymbol{Q}^{\bluetext{\Pi^{\cdot|0}}}_t-\boldsymbol{Q}^{\pinktext{\mathbb{M}^{\cdot|0}}}_t\right)(\boldsymbol{X}_t,\boldsymbol{X}_t)\right]dt\label{eq:reciprocal-generator-3}
    \end{align}
    \end{small}
\end{lemma}

\textit{Proof.} We prove each part of the Lemma in steps. First, we derive the  generator of the conditioned reciprocal measure, which takes the form of an expectation of the conditional reference generators derived in Lemma \ref{lemma:ctmc-cond-generator-markov}.

\textbf{Step 1: Derive the Conditioned Generator. }
First, we establish that the pinned down bridge measure $\Pi^{\cdot|0=\boldsymbol{x}_0}$ is in the reciprocal class $\mathcal{R}(\mathbb{Q})$ and satisfies the (\ref{eq:reciprocal-property}):
\begin{align}
    \Pi ^{\cdot|0=\boldsymbol{x}_0}=\int\mathbb{Q}_{\cdot|0,T}(\cdot|\boldsymbol{x}_0, \boldsymbol{x}_T)\Pi_{T|0}(d\boldsymbol{x}_T|\boldsymbol{x}_0)\tag{Pinned-Down Bridge}
\end{align}
We first write the transition probability between states under the conditional process using the law of total probability:
\begin{align}
\Pi^{\cdot|0=\boldsymbol{x}_0}_{t|s, 0}(\boldsymbol{y}|\boldsymbol{x})&=\sum_{\boldsymbol{x}_T\in \mathcal{X}}\bluetext{\Pi_{T|s}^{\cdot|0=\boldsymbol{x}_0}(\boldsymbol{x}_T|\boldsymbol{x})}\pinktext{\Pi^{\cdot |0=\boldsymbol{x}_0}_{t|s,T}(\boldsymbol{y}|\boldsymbol{x}, \boldsymbol{x}_T)}\tag{Law of Total Probability}
\end{align}
which samples the terminal state $\boldsymbol{X}_T=\boldsymbol{x}_T$ and then samples the intermediate time $\boldsymbol{X}_t=\boldsymbol{y}$ from the bridge. Given that $\Pi\in \mathcal{R}(\mathbb{Q})$ is in the reciprocal class with a shared bridge measure as $\mathbb{Q}$, by the (\ref{eq:reciprocal-property}), the second term in the sum is equal to:
\begin{align}
    \pinktext{\Pi^{\cdot |0=\boldsymbol{x}_0}_{t|s,T}(\boldsymbol{y}|\boldsymbol{x}, \boldsymbol{x}_T)}=\mathbb{Q}_{t|s,T}(\boldsymbol{y}|\boldsymbol{x}, \boldsymbol{x}_T)=\underbrace{\frac{\mathbb{Q}_{t|s}(\boldsymbol{y}|\boldsymbol{x})\mathbb{Q}_{T|t}(\boldsymbol{x}_T|\boldsymbol{y})}{\mathbb{Q}_{T|s}(\boldsymbol{x}_T|\boldsymbol{x})}}_{\text{Bayes' rule}}
\end{align}
Then, substituting this back into the expression for $\Pi^{\cdot|0=\boldsymbol{x}_0}_{t|s, 0}(\boldsymbol{y}|\boldsymbol{x})$ and recognizing $\Pi_{T|s}^{\cdot|0=\boldsymbol{x}_0}(\boldsymbol{x}_T|\boldsymbol{x})=\Pi_{T|s,0}(\boldsymbol{x}_T|\boldsymbol{x}, \boldsymbol{x}_0)$, we have:
\begin{small}
\begin{align}
    \Pi^{\cdot|0=\boldsymbol{x}_0}_{t|s, 0}(\boldsymbol{y}|\boldsymbol{x})&=\sum_{\boldsymbol{x}_T\in \mathcal{X}}\bluetext{\Pi_{T|s}^{\cdot|0=\boldsymbol{x}_0}(\boldsymbol{x}_T|\boldsymbol{x})}\pinktext{\frac{\mathbb{Q}_{t|s}(\boldsymbol{y}|\boldsymbol{x})\mathbb{Q}_{T|t}(\boldsymbol{x}_T|\boldsymbol{y})}{\mathbb{Q}_{T|s}(\boldsymbol{x}_T|\boldsymbol{x})}}\nonumber\\
    &=\mathbb{Q}_{t|s}(\boldsymbol{y}|\boldsymbol{x})\sum_{\boldsymbol{x}_T\in \mathcal{X}}\bluetext{\underbrace{\Pi_{T|s,0}(\boldsymbol{x}_T|\boldsymbol{x}, \boldsymbol{x}_0)}_{(\bigstar)}}\frac{\mathbb{Q}_{T|t}(\boldsymbol{x}_T|\boldsymbol{y})}{\mathbb{Q}_{T|s}(\boldsymbol{x}_T|\boldsymbol{x})}\label{eq:pi-proof-1}
\end{align}
\end{small}
We can break down $(\bigstar)$ using Bayes' rule and applying the reciprocal property to get:
\begin{align}
    \Pi_{T|s,0}(\boldsymbol{x}_T|\boldsymbol{x},\boldsymbol{x}_0)&=\frac{\bluetext{\Pi_{s|0,T}(\boldsymbol{x}|\boldsymbol{x}_0, \boldsymbol{x}_T)}\Pi_{T|0}(\boldsymbol{x}_T|\boldsymbol{x}_0)}{\Pi_{s|0}(\boldsymbol{x}|\boldsymbol{x}_0)}\tag{Bayes' Rule}\\
    &=\frac{\bluetext{\mathbb{Q}_{s|0,T}(\boldsymbol{x}|\boldsymbol{x}_0, \boldsymbol{x}_T)}\Pi_{T|0}(\boldsymbol{x}_T|\boldsymbol{x}_0)}{\Pi_{s|0}(\boldsymbol{x}|\boldsymbol{x}_0)}\tag{Reciprocal Property}\\
    &=\bluetext{\frac{\mathbb{Q}_{s|0}(\boldsymbol{x}|\boldsymbol{x}_0)\mathbb{Q}_{T|s}(\boldsymbol{x}_T|\boldsymbol{x})}{\mathbb{Q}_{T|0}(\boldsymbol{x}_T|\boldsymbol{x}_0)}}\frac{\Pi_{T|0}(\boldsymbol{x}_T|\boldsymbol{x}_0)}{\Pi_{s|0}(\boldsymbol{x}|\boldsymbol{x}_0)}\tag{Bayes' Rule}\\
    &=\frac{\mathbb{Q}_{s|0}(\boldsymbol{x}|\boldsymbol{x}_0)}{\mathbb{Q}_{T|0}(\boldsymbol{x}_T|\boldsymbol{x}_0)}\frac{\Pi_{T|0}(\boldsymbol{x}_T|\boldsymbol{x}_0)}{\Pi_{s|0}(\boldsymbol{x}|\boldsymbol{x}_0)}\mathbb{Q}_{T|s}(\boldsymbol{x}_T|\boldsymbol{x})
\end{align}
Substituting $(\bigstar)$ this back into (\ref{eq:pi-proof-1}), we can cancel terms and factor out terms not dependent on $\boldsymbol{x}_T$ to get: 
\begin{align}
    \Pi^{\cdot|0=\boldsymbol{x}_0}_{t|s}(\boldsymbol{y}|\boldsymbol{x})&=\mathbb{Q}_{t|s}(\boldsymbol{y}|\boldsymbol{x})\sum_{\boldsymbol{x}_T\in \mathcal{X}}\bluetext{\left[\frac{\mathbb{Q}_{s|0}(\boldsymbol{x}|\boldsymbol{x}_0)}{\mathbb{Q}_{T|0}(\boldsymbol{x}_T|\boldsymbol{x}_0)}\frac{\Pi_{T|0}(\boldsymbol{x}_T|\boldsymbol{x}_0)}{\Pi_{s|0}(\boldsymbol{x}|\boldsymbol{x}_0)}\mathbb{Q}_{T|s}(\boldsymbol{x}_T|\boldsymbol{x})\right]}\frac{\mathbb{Q}_{T|t}(\boldsymbol{x}_T|\boldsymbol{y})}{\mathbb{Q}_{T|s}(\boldsymbol{x}_T|\boldsymbol{x})}\nonumber\\
    &=\frac{\mathbb{Q}_{s|0}(\boldsymbol{x}|\boldsymbol{x}_0)}{\Pi_{s|0}(\boldsymbol{x}|\boldsymbol{x}_0)}\mathbb{Q}_{t|s}(\boldsymbol{y}|\boldsymbol{x})\sum_{\boldsymbol{x}_T\in \mathcal{X}}\frac{\Pi_{T|0}(\boldsymbol{x}_T|\boldsymbol{x}_0)}{\mathbb{Q}_{T|0}(\boldsymbol{x}_T|\boldsymbol{x}_0)}\mathbb{Q}_{T|t}(\boldsymbol{x}_T|\boldsymbol{y})\label{eq:pi-proof-2}
\end{align}
To derive the generator using (\ref{eq:pi-proof-2}), we take the time derivative evaluated at $t=s$ to get: 
\begin{small}
\begin{align}
    \boldsymbol{Q}_s^{\Pi^{\cdot|0=\boldsymbol{x}_0}}(\boldsymbol{x},\boldsymbol{y})&=\partial_t\Pi^{\cdot|0=\boldsymbol{x}_0}(\boldsymbol{X}_t=\boldsymbol{y}|\boldsymbol{X}_s=\boldsymbol{x})\big\vert_{t=s}\nonumber\\
    &\overset{(\ref{eq:pi-proof-2})}{=}\partial_t\left(\frac{\mathbb{Q}_{s|0}(\boldsymbol{x}|\boldsymbol{x}_0)}{\Pi_{s|0}(\boldsymbol{x}|\boldsymbol{x}_0)}\mathbb{Q}_{t|s}(\boldsymbol{y}|\boldsymbol{x})\sum_{\boldsymbol{x}_T\in \mathcal{X}}\frac{\Pi_{T|0}(\boldsymbol{x}_T|\boldsymbol{x}_0)}{\mathbb{Q}_{T|0}(\boldsymbol{x}_T|\boldsymbol{x}_0)}\mathbb{Q}_{T|t}(\boldsymbol{x}_T|\boldsymbol{y})\right)\bluetext{\bigg|_{t=s}}\nonumber\\
    &=\frac{\mathbb{Q}_{s|0}(\boldsymbol{x}|\boldsymbol{x}_0)}{\Pi_{s|0}(\boldsymbol{x}|\boldsymbol{x}_0)}\bluetext{\partial_t\mathbb{Q}_{t|s}(\boldsymbol{y}|\boldsymbol{x})\big|_{t=s}}\sum_{\boldsymbol{x}_T\in \mathcal{X}}\left[\frac{\Pi_{T|0}(\boldsymbol{x}_T|\boldsymbol{x}_0)}{\mathbb{Q}_{T|0}(\boldsymbol{x}_T|\boldsymbol{x}_0)}\mathbb{Q}_{T|t}(\boldsymbol{x}_T|\boldsymbol{y})\right]\nonumber\\
    &\quad \quad +\frac{\mathbb{Q}_{s|0}(\boldsymbol{x}|\boldsymbol{x}_0)}{\Pi_{s|0}(\boldsymbol{x}|\boldsymbol{x}_0)}\mathbb{Q}_{\bluetext{s}|s}(\boldsymbol{y}|\boldsymbol{x})\sum_{\boldsymbol{x}_T\in \mathcal{X}}\left[\frac{\Pi_{T|0}(\boldsymbol{x}_T|\boldsymbol{x}_0)}{\mathbb{Q}_{T|0}(\boldsymbol{x}_T|\boldsymbol{x}_0)}\bluetext{\partial_t\mathbb{Q}_{T|t}(\boldsymbol{x}_T|\boldsymbol{y})\big\vert_{t=s}}\right]
\end{align}
\end{small}
Applying the \textbf{backward Kolmogorov equation} from Lemma \ref{lemma:ctmc-kolmogorov-bwd} to $\partial_t\mathbb{Q}_{T|t}(\boldsymbol{x}_T|\boldsymbol{y})$, we have:
\begin{small}
\begin{align}
    \boldsymbol{Q}_s^{\Pi^{\cdot|0=\boldsymbol{x}_0}}(\boldsymbol{x},\boldsymbol{y})&=\frac{\mathbb{Q}_{s|0}(\boldsymbol{x}|\boldsymbol{x}_0)}{\Pi_{s|0}(\boldsymbol{x}|\boldsymbol{x}_0)}\bluetext{\boldsymbol{Q}^0_t(\boldsymbol{x},\boldsymbol{y})}\sum_{\boldsymbol{x}_T\in \mathcal{X}}\left[\frac{\Pi_{T|0}(\boldsymbol{x}_T|\boldsymbol{x}_0)}{\mathbb{Q}_{T|0}(\boldsymbol{x}_T|\boldsymbol{x}_0)}\mathbb{Q}_{T|t}(\boldsymbol{x}_T|\boldsymbol{y})\right]\nonumber\\
    &\quad \quad +\frac{\mathbb{Q}_{s|0}(\boldsymbol{x}|\boldsymbol{x}_0)}{\Pi_{s|0}(\boldsymbol{x}|\boldsymbol{x}_0)}\bluetext{\boldsymbol{1}_{\boldsymbol{x}=\boldsymbol{y}}}\sum_{\boldsymbol{x}_T\in \mathcal{X}}\left[\frac{\Pi_{T|0}(\boldsymbol{x}_T|\boldsymbol{x}_0)}{\mathbb{Q}_{T|0}(\boldsymbol{x}_T|\boldsymbol{x}_0)}\bluetext{\left(-\sum_{\boldsymbol{z}\in \mathcal{X}}\boldsymbol{Q}_t^0(\boldsymbol{y}, \boldsymbol{z})\mathbb{Q}_{T|s}(\boldsymbol{x}_T|\boldsymbol{z})\right)}\right]\nonumber\\
    &=\frac{\mathbb{Q}_{s|0}(\boldsymbol{x}|\boldsymbol{x}_0)}{\Pi_{s|0}(\boldsymbol{x}|\boldsymbol{x}_0)}\bluetext{\boldsymbol{Q}^0_t(\boldsymbol{x},\boldsymbol{y})}\sum_{\boldsymbol{x}_T\in \mathcal{X}}\left[\frac{\Pi_{T|0}(\boldsymbol{x}_T|\boldsymbol{x}_0)}{\mathbb{Q}_{T|0}(\boldsymbol{x}_T|\boldsymbol{x}_0)}\mathbb{Q}_{T|t}(\boldsymbol{x}_T|\boldsymbol{y})\right]\nonumber\\
    &\quad \quad \bluetext{-\boldsymbol{1}_{\boldsymbol{x}=\boldsymbol{y}}}\frac{\mathbb{Q}_{s|0}(\boldsymbol{x}|\boldsymbol{x}_0)}{\Pi_{s|0}(\boldsymbol{x}|\boldsymbol{x}_0)}\sum_{\boldsymbol{x}_T\in \mathcal{X}}\left[\sum_{\boldsymbol{z}\in \mathcal{X}}\big(\boldsymbol{Q}_t^0(\boldsymbol{y}, \boldsymbol{z})\mathbb{Q}_{T|s}(\boldsymbol{x}_T|\boldsymbol{z})\big)\frac{\Pi_{T|0}(\boldsymbol{x}_T|\boldsymbol{x}_0)}{\mathbb{Q}_{T|0}(\boldsymbol{x}_T|\boldsymbol{x}_0)}\right]
\end{align}
\end{small}
Factoring out $\frac{\mathbb{Q}_{s|0}(\boldsymbol{x}|\boldsymbol{x}_0)}{\Pi_{s|0}(\boldsymbol{x}|\boldsymbol{x}_0)}$ and $\frac{\Pi_{T|0}(\boldsymbol{x}_T|\boldsymbol{x}_0)}{\mathbb{Q}_{T|0}(\boldsymbol{x}_T|\boldsymbol{x}_0)}$ and recognizing the (\ref{eq:ctmc-endpoint-cond-gen}) of the Markov process defined in Lemma \ref{lemma:ctmc-cond-generator-markov}, we get:
\begin{small}
\begin{align}
    \boldsymbol{Q}_s^{\Pi^{\cdot|0=\boldsymbol{x}_0}}&(\boldsymbol{x},\boldsymbol{y})
    =\sum_{\boldsymbol{x}_T\in \mathcal{X}}\frac{\mathbb{Q}_{s|0}(\boldsymbol{x}|\boldsymbol{x}_0)}{\Pi_{s|0}(\boldsymbol{x}|\boldsymbol{x}_0)}\bigg[\bluetext{\underbrace{\boldsymbol{Q}^0_t(\boldsymbol{x},\boldsymbol{y})\mathbb{Q}_{T|t}(\boldsymbol{x}_T|\boldsymbol{y})-\boldsymbol{1}_{\boldsymbol{x}=\boldsymbol{y}}\sum_{\boldsymbol{z}\in \mathcal{X}}\boldsymbol{Q}_t^0(\boldsymbol{y}, \boldsymbol{z})\mathbb{Q}_{T|s}(\boldsymbol{x}_T|\boldsymbol{z})}_{=:\boldsymbol{Q}_s^0(\boldsymbol{x}, \boldsymbol{y};\boldsymbol{x}_T)\mathbb{Q}_{T|s}(\boldsymbol{x}_T|\boldsymbol{x})}}\bigg]\frac{\Pi_{T|0}(\boldsymbol{x}_T|\boldsymbol{x}_0)}{\mathbb{Q}_{T|0}(\boldsymbol{x}_T|\boldsymbol{x}_0)}\nonumber\\
    &=\sum_{\boldsymbol{x}_T\in \mathcal{X}}\underbrace{\frac{\mathbb{Q}_{s|0}(\boldsymbol{x}|\boldsymbol{x}_0)}{\mathbb{Q}_{T|0}(\boldsymbol{x}_T|\boldsymbol{x}_0)}\frac{\Pi_{T|0}(\boldsymbol{x}_T|\boldsymbol{x}_0)}{\Pi_{s|0}(\boldsymbol{x}|\boldsymbol{x}_0)}\bluetext{\mathbb{Q}_{T|s}(\boldsymbol{x}_T|\boldsymbol{x})}}_{=:\Pi^{\cdot|0=\boldsymbol{x}_0}(\boldsymbol{X}_T=\boldsymbol{x}_T|\boldsymbol{X}_s=\boldsymbol{x})}\bluetext{\boldsymbol{Q}_s^0(\boldsymbol{x}, \boldsymbol{y};\boldsymbol{x}_T)}\nonumber\\
    &=\sum_{\boldsymbol{x}_T\in \mathcal{X}}\bluetext{\Pi^{\cdot|0=\boldsymbol{x}_0}(\boldsymbol{X}_T=\boldsymbol{x}_T|\boldsymbol{X}_s=\boldsymbol{x})}\bluetext{\boldsymbol{Q}_s^0(\boldsymbol{x}, \boldsymbol{y};\boldsymbol{x}_T)}\nonumber\\
    &=\boxed{\mathbb{E}_{\boldsymbol{x}_T\sim \Pi_{T|0,t}}\left[\boldsymbol{Q}^0_t(\boldsymbol{x}, \boldsymbol{y};\boldsymbol{x}_T)|\boldsymbol{X}_0=\boldsymbol{x}_0, \boldsymbol{X}_t=\boldsymbol{x}\right]}
\end{align}
\end{small}
which gives us our final form for the conditional generator. \hfill $\square$

\textbf{Step 2: Derive the KL Divergence. }
In Corollary \ref{corollary:ctmc-kl}, we derived the expression for the KL divergence between CTMCs, which relies on the Markov structure. To derive the KL divergence between a \textit{non-Markov repciprocal process} $\Pi$ and a Markov measure $\mathbb{M}$, we apply the (\ref{eq:chain-rule-kl}) from Lemma \ref{lemma:chain-rule-kl} to decompose the KL divergence as the sum of the KL divergence of the initial marginal and the expectation of the KL divergence of the conditioned measures:
\begin{small}
\begin{align}
    \text{KL}(\Pi\|\mathbb{M})&=\bluetext{\underbrace{\text{KL}(\Pi_0\|\mathbb{M}_0)}_{=0\; (\Pi_0=\mathbb{M}_0)}}+ \mathbb{E}_{\Pi_0}\left[\text{KL}(\Pi^{\cdot|0=\boldsymbol{x}_0}\|\mathbb{M}^{\cdot|0=\boldsymbol{x}_0})\right]=\mathbb{E}_{\Pi_0}\left[\text{KL}(\Pi^{\cdot|0=\boldsymbol{x}_0}\|\mathbb{M}^{\cdot|0=\boldsymbol{x}_0})\right]\label{eq:markov-generator-proof1}
\end{align}
\end{small}
where the KL between the initial marginals vanishes by our definition $\Pi_0=\mathbb{M}_0$. We have shown in Proposition \ref{eq:prop:reciprocal-process} that conditioning a reciprocal process on an endpoint $\boldsymbol{X}_0=\boldsymbol{x}_0$ yields a \textbf{Markov process} and the KL divergence between two Markov measures is defined in Corollary \ref{corollary:ctmc-kl}, we can further expand (\ref{eq:markov-generator-proof1}) as:
\begin{small}
\begin{align}
    \text{KL}(\Pi\|\mathbb{M})&\overset{(\ref{eq:kl-ctmc-2})}{=} \int_0^T\mathbb{E}_{\Pi_{0,t}}\left[\sum_{\boldsymbol{y}\neq \boldsymbol{X}_t}\left(\boldsymbol{Q}^{\bluetext{\Pi^{\cdot|0}}}_t\log \frac{\boldsymbol{Q}^{\bluetext{\Pi^{\cdot|0}}}_{t}}{\boldsymbol{Q}^{\pinktext{\mathbb{M}^{\cdot|0}}}_{t}}\right)(\boldsymbol{X}_t, \boldsymbol{y})+\left(\boldsymbol{Q}^{\bluetext{\Pi^{\cdot|0}}}_t-\boldsymbol{Q}^{\pinktext{\mathbb{M}^{\cdot|0}}}_t\right)(\boldsymbol{X}_t,\boldsymbol{X}_t)\right]ds
\end{align}
\end{small}
which is exactly the form of the KL divergence between the reciprocal and Markov measures defined in the Proposition. 

\begin{lemma}[Generator of Markovian Projection]\label{lemma:generator-markov-proj}
    The Markovian projection $\mathbb{M}^\star=\text{proj}_{\mathcal{M}}(\Pi)$ of the reciprocal measure $\Pi\in \mathcal{R}(\mathbb{Q})$ that minimizes the KL divergence $\text{KL}(\Pi\|\mathbb{M})$ is defined by the generator:
    \begin{align}
        \boldsymbol{Q}^{\mathbb{M}^\star}_t(\boldsymbol{x}, \boldsymbol{y})=\mathbb{E}_{\boldsymbol{x}_T\sim\Pi_{T|t}}\left[\boldsymbol{Q}_t^0(\boldsymbol{x}, \boldsymbol{y};\boldsymbol{x}_T)|\boldsymbol{X}_t=\boldsymbol{x}\right]\tag{Markovian Projection Generator}\label{eq:markov-proj-generator}
    \end{align}
\end{lemma}

\textit{Intuition.} The goal is to find the closest Markov process to the reciprocal measure $\Pi\in \mathcal{R}(\mathbb{Q})$, which is not Markov and depends on the \textit{future state} $\boldsymbol{X}_T$. Since Markov measures can only depend on the present, we can remove the dependence on $\boldsymbol{X}_T$ by \textit{taking the expectation over the conditional distribution of the endpoint.} This yields the following \textbf{candidate form of the Markov generator}:
\begin{small}
\begin{align}
    \boldsymbol{Q}^{\mathbb{M}}_t(\boldsymbol{x}, \boldsymbol{y})=\mathbb{E}_{\boldsymbol{x}_T\sim\Pi_{T|t}}\left[\boldsymbol{Q}_t^0(\boldsymbol{x}, \boldsymbol{y};\boldsymbol{x}_T)|\boldsymbol{X}_t=\boldsymbol{x}\right]\label{eq:markov-generator-candidate}
\end{align}
\end{small}
Now, we will show that this is indeed the generator of the \textbf{Markovian projection} $\mathbb{M}^\star$ which minimizes the KL divergence with $\Pi$.

\textit{Proof.} First, we assume that for all $\boldsymbol{x}\neq \boldsymbol{y}, \boldsymbol{x}_T\in \mathcal{X}$ and $t\in [0,T)$, we have $\boldsymbol{Q}^0(\boldsymbol{x},\boldsymbol{y}; \boldsymbol{x}_T)=0\iff \boldsymbol{Q}_t^0(\boldsymbol{x}, \boldsymbol{y})$. To prove that (\ref{eq:markov-generator-candidate}) is the generator of $\mathbb{M}^\star$, we first decompose it as follows:
\begin{small}
\begin{align}
    \boldsymbol{Q}^{\mathbb{M}}_t(\boldsymbol{x}, \boldsymbol{y})&=\mathbb{E}_{\boldsymbol{x}_T\sim\Pi_{T|t}}\left[\boldsymbol{Q}_t^0(\boldsymbol{x}, \boldsymbol{y};\boldsymbol{x}_T)|\boldsymbol{X}_t=\boldsymbol{x}\right]\nonumber\\
    &= \sum_{\boldsymbol{x}_T\in \mathcal{X}}\bluetext{\Pi_{T|t}(\boldsymbol{x}_T|\boldsymbol{x})}\boldsymbol{Q}_t^0(\boldsymbol{x}, \boldsymbol{y};\boldsymbol{x}_T)
\end{align}
\end{small}
Since the reciprocal process depends on both the initial state in addition to the target state, we apply the law of total expectation to insert conditioning on $\boldsymbol{X}_0=\boldsymbol{x}_0$ to get:
\begin{small}
\begin{align}
    \boldsymbol{Q}^{\mathbb{M}}_t(\boldsymbol{x}, \boldsymbol{y})&=\sum_{\boldsymbol{x}_T\in \mathcal{X}}\bluetext{\sum_{\boldsymbol{x}_0\in \mathcal{X}}}\Pi_{0,T|t}(\bluetext{\boldsymbol{x}_0},\boldsymbol{x}_T|\boldsymbol{x})\boldsymbol{Q}_t^0(\boldsymbol{x}, \boldsymbol{y};\boldsymbol{x}_T)\nonumber\\
    &=\bluetext{\sum_{\boldsymbol{x}_0\in \mathcal{X}}\Pi_{0|t}\left(\boldsymbol{x}_0|\boldsymbol{x}\right)}\bigg[\underbrace{\sum_{\boldsymbol{x}_T\in \mathcal{X}}\Pi_{T|0,t}(\boldsymbol{x}_T|\boldsymbol{x}_0,\boldsymbol{x})\boldsymbol{Q}_t^0(\boldsymbol{x}, \boldsymbol{y};\boldsymbol{x}_T)}_{=:\mathbb{E}_{\Pi_{T|0,t}}[\boldsymbol{Q}_t^0(\boldsymbol{x},\boldsymbol{y};\boldsymbol{x}_T)|\boldsymbol{X}_0=\boldsymbol{x}_0, \boldsymbol{x}]}\bigg]\nonumber\\
    &=\bluetext{\sum_{\boldsymbol{x}_0\in \mathcal{X}}\Pi_{0|t}\left(\boldsymbol{x}_0|\boldsymbol{x}\right)}\boldsymbol{Q}_t^{\Pi^{\cdot|0=\boldsymbol{x}_0}}(\boldsymbol{x},\boldsymbol{y})=\boxed{\mathbb{E}_{\Pi_{0|t}}\left[\boldsymbol{Q}_t^{\Pi^{\cdot|0=\boldsymbol{x}_0}}(\boldsymbol{X}_t,\boldsymbol{y})\big|\boldsymbol{X}_t=\boldsymbol{x}\right]}
\end{align}
\end{small}
and we have shown that the generator $\boldsymbol{Q}^{\mathbb{M}}_t(\boldsymbol{x}, \boldsymbol{y})$ is the expectation of bridge generators conditioned on the initial state or a \textbf{mixture of bridges} evaluated at $\boldsymbol{x}$. 

\textbf{Step 2: Show the Markov Generator Minimizes the KL Divergence. }
Let $\mathbb{M}\in \mathcal{M}$ be an arbitrary Markov measure with generator defined in (\ref{eq:markov-proj-generator}), then by optimality of $\mathbb{M}^\star$, we have $\text{KL}(\Pi\|\mathbb{M})\geq\text{KL}(\Pi\|\mathbb{M}^\star)$. We aim to show that this inequality reduces to an \textbf{equality} where $\text{KL}(\Pi\|\mathbb{M})=\text{KL}(\Pi\|\mathbb{M}^\star)$.

Using the definition of $\text{KL}(\Pi\|\mathbb{M})$ from (\ref{eq:reciprocal-generator-3}) and some algebra, we can write the difference $\text{KL}(\Pi\|\mathbb{M}^\star)-\text{KL}(\Pi\|\mathbb{M})$ as:
\begin{small}
\begin{align}
\text{KL}(\Pi\|\mathbb{M}^\star)-\text{KL}(\Pi\|\mathbb{M})&=\mathbb{E}_{\Pi_0}\left[\int_0^T \sum_{\boldsymbol{y}\neq \boldsymbol{X}_t}\left(\boldsymbol{Q}^{\Pi^{\cdot|0=\boldsymbol{x}_0}}\log \frac{\boldsymbol{Q}_t^{\mathbb{M}}}{\boldsymbol{Q}_t^{\mathbb{M}^\star}}+\boldsymbol{Q}_t^{\mathbb{M}^\star}-\boldsymbol{Q}^{\mathbb{M}}_t\right)(\boldsymbol{X}_t,\boldsymbol{y})dt\right]\nonumber\\
&=\int_0^T \sum_{\boldsymbol{y}\neq \boldsymbol{X}_t}\bigg(\bluetext{\underbrace{\mathbb{E}_{\Pi_{0|t}}\left[\boldsymbol{Q}^{\Pi^{\cdot|0=\boldsymbol{x}_0}}\big|\boldsymbol{X}_t\right]}_{(\bigstar)}}\log \frac{\boldsymbol{Q}_t^{\mathbb{M}}}{\boldsymbol{Q}_t^{\mathbb{M}^\star}}+\boldsymbol{Q}_t^{\mathbb{M}^\star}-\boldsymbol{Q}^{\mathbb{M}}_t\bigg)(\boldsymbol{X}_t,\boldsymbol{y})dt\nonumber\\
&=\int_0^T \sum_{\boldsymbol{y}\neq \boldsymbol{X}_t}\bigg(\bluetext{\boldsymbol{Q}_t^{\mathbb{M}}}\log \frac{\boldsymbol{Q}_t^{\mathbb{M}}}{\boldsymbol{Q}_t^{\mathbb{M}^\star}}+\boldsymbol{Q}_t^{\mathbb{M}^\star}-\boldsymbol{Q}^{\mathbb{M}}_t\bigg)(\boldsymbol{X}_t,\boldsymbol{y})dt=\boxed{\text{KL}(\mathbb{M}\|\mathbb{M}^\star)}\geq 0
\end{align}
\end{small}
Since $\mathbb{M}^\star$ is optimal, by definition $\text{KL}(\Pi\|\mathbb{M}^\star)-\text{KL}(\Pi\|\mathbb{M})\leq 0$. However, we have shown that $\text{KL}(\Pi\|\mathbb{M}^\star)-\text{KL}(\Pi\|\mathbb{M})=\text{KL}(\mathbb{M}\|\mathbb{M}^\star)$ where $\text{KL}(\mathbb{M}\|\mathbb{M}^\star)\geq 0$ by non-negativity of the KL divergence. Therefore, we can conclude that $\text{KL}(\Pi\|\mathbb{M}^\star)=\text{KL}(\Pi\|\mathbb{M})$ and $\mathbb{M}=\mathbb{M}^\star$ almost surely. \hfill $\square$

Now, we have seen that the Markovian projection $\mathbb{M}^\star:=\text{proj}_{\mathcal{R}(\mathbb{Q})}(\Pi)$ which minimizes the KL divergence from a reciprocal measure $\Pi$ is explicitly defined with (\ref{eq:ctmc-markov-projection-generator}). However, recall from our discussion of Iterative Markovian Fitting from Section \ref{subsec:diffusionsbm} that simulating the Markov SDE in the forward-time can accumulate errors in matching the target distribution $\pi_T$, motivating the simulation a corresponding reverse-time Markovian projection. Therefore, we will also extend the idea of a reverse-time Markovian projection for CTMCs.

\begin{definition}[Reverse-Time Markovian Projection]\label{def:reverse-time-markov-proj}
    The generator of the \textbf{reverse-time Markovian projection} $\tilde{\mathbb{M}}^\star$ is given by:
    \begin{small}
    \begin{align}
        \tilde{\boldsymbol{Q}}_t^{\mathbb{M}^\star}(\boldsymbol{x},\boldsymbol{y})=\mathbb{E}_{\boldsymbol{x}_0\sim \mathbb{Q}_{0|t}(\cdot|\boldsymbol{x})}\left[\bluetext{\tilde{\boldsymbol{Q}}_t^0(\boldsymbol{x},\boldsymbol{y};\boldsymbol{x}_0)}\big|\tilde{\boldsymbol{X}}_t=\boldsymbol{x}\right]\tag{Reverse-Time Markov Generator}\label{eq:reverse-time-generator}
    \end{align}
    \end{small}
    where each conditional generator is defined as:
    \begin{small}
    \begin{align}
        \bluetext{\tilde{\boldsymbol{Q}}_t^0(\boldsymbol{x},\boldsymbol{y};\boldsymbol{x}_0)}&=\partial_s\mathbb{Q}^{\cdot |0=\boldsymbol{x}_0}\left(\boldsymbol{X}_t=\boldsymbol{y}|\boldsymbol{X}_s=\boldsymbol{x}\right)\big|_{s=t}\nonumber\\
        &=\boldsymbol{Q}^0_{T-t}(\boldsymbol{x},\boldsymbol{y})\frac{\mathbb{Q}_{T-t|0}(\boldsymbol{y}|\boldsymbol{x}_0)}{\mathbb{Q}_{T-t|0}(\boldsymbol{x}|\boldsymbol{x}_0)}-\boldsymbol{1}_{\boldsymbol{x}=\boldsymbol{y}}\sum_{\boldsymbol{z}\in \mathcal{X}}\boldsymbol{Q}^0_{T-t}(\boldsymbol{z},\boldsymbol{x})\frac{\mathbb{Q}_{T-t|0}(\boldsymbol{z}|\boldsymbol{x}_0)}{\mathbb{Q}_{T-t|0}(\boldsymbol{y}|\boldsymbol{x}_0)}
    \end{align}
    \end{small}
    where $\mathbb{Q}_{T-t|0}(\cdot|\cdot)=\mathbb{Q}(\boldsymbol{X}_{T-t}=\cdot|\boldsymbol{X}_0=\cdot)$ is the conditional transition probability under the reference measure $\mathbb{Q}$. Then, the reverse-time KL divergence that is minimized under the Markovian projection $\mathbb{M}^\star$ is given by:
    \begin{small}
    \begin{align}
        \text{KL}(\Pi\|\mathbb{M}^\star)=\int_0^T\mathbb{E}_{\Pi_{t,T}}\left[\sum_{\boldsymbol{y}\neq \boldsymbol{X}_t}\left(\bluetext{\tilde{\boldsymbol{Q}}_t^{\Pi^{\cdot |T}}}\log \frac{\bluetext{\tilde{\boldsymbol{Q}}_t^{\Pi^{\cdot|T}}}}{\tilde{\boldsymbol{Q}}_t^{\mathbb{M}^\star}}\right)(\tilde{\boldsymbol{X}}_t,\boldsymbol{y})+\left(\tilde{\boldsymbol{Q}}_t^{\mathbb{M}^\star}-\bluetext{\tilde{\boldsymbol{Q}}_t^{\Pi^{\cdot|T}}}\right)(\tilde{\boldsymbol{X}}_t,\tilde{\boldsymbol{X}}_t)\right]dt\label{eq:ctmc-markov-kl-reverse}
    \end{align}
    \end{small}
    where the KL divergence between the terminal distributions at time $T$ vanishes as we initialize $\mathbb{M}_T=\Pi_T$. Specifically, we define $\Pi^{\cdot|T}$ is the bridge measure conditioned on $\boldsymbol{X}_T=\boldsymbol{x}_T$ with the reverse-time generator $\tilde{\boldsymbol{Q}}^{\Pi^{\cdot|T=\boldsymbol{x}_T}}_t$ defined as:
    \begin{align}
        \bluetext{\tilde{\boldsymbol{Q}}^{\Pi^{\cdot|T=\boldsymbol{x}_T}}_t(\boldsymbol{x}, \boldsymbol{y})}=\mathbb{E}_{\boldsymbol{x}_T\sim \Pi_{T|0,t}}\left[\tilde{\boldsymbol{Q}}^0_t(\boldsymbol{x},\boldsymbol{y}; \boldsymbol{x}_T)|\tilde{\boldsymbol{X}}_t=\boldsymbol{x}, \tilde{\boldsymbol{X}}_0=\boldsymbol{x}_T\right]
    \end{align}
    where for any $t\in [0,T]$, the marginal distributions match $\mathbb{M}^\star_t=\Pi_t$.
\end{definition}

\textit{Proof Sketch.} The result follows from the same proof sequence as the forward-time Markovian projection, which consists of deriving the form of the conditional generator in reverse-time coordinate (Lemma \ref{lemma:ctmc-cond-generator-markov}), showing that the reciprocal process conditioned on an initial state $\tilde{\boldsymbol{X}}_0=\boldsymbol{x}_T$ is Markov (Lemma \ref{lemma:ctmc-cond-generator-reciprocal}), and deriving the form of the optimal reverse-time generator of the Markovian projection (Lemma \ref{lemma:generator-markov-proj}). 

This analogous proof works because the probability of a path is the same under time reversal, and the only difference is the \textit{direction of the generator}. Therefore, by simply defining the reverse-time generator on the time-reversed CTMC process $\tilde{\boldsymbol{X}}_t:=\boldsymbol{X}_{T-t}$ as in (\ref{eq:reverse-time-generator}), we can compute the KL divergence between the generators of the reciprocal process $\Pi^{\cdot|T}$ conditioned on a state from time $T$ and the Markov process $\mathbb{M}$ evaluated along the reversed trajectory $\tilde{\boldsymbol{X}}_{0:T}$. Taking expectation with respect to the trajectories from the reversed path measure $\tilde{\boldsymbol{X}}_{0:T}\sim \Pi$ yields the reverse KL divergence. \hfill $\square$

Now that we have written the Markovian and reciprocal projections explicitly with respect to the generators of CTMCs, we can extend the Iterative Markovian Fitting (IMF) procedure described in Section \ref{subsec:diffusionsbm} to the discrete state space, where instead of parameterizing the forward and reverse Markov drifts, we parameterize the forward and reverse Markov generators $\boldsymbol{Q}^{\mathbb{M}^\theta}$ and $\boldsymbol{Q}^{\mathbb{M}^\phi}$.

\subsection{Discrete Diffusion Schrödinger Bridge Matching}
\label{subsec:ddsbm}
\textit{Prerequisite: Section \ref{subsec:markov-reciprocal-proj}, \ref{subsec:diffusionsbm}}

Using these theoretical foundations of Markov and reciprocal projections of CTMCs, we can define the discrete state space analog of the Iterative Markovian Fitting (IMF) algorithm from Section \ref{subsec:diffusionsbm} called the \boldtext{Discrete Diffusion Schrödinger Bridge Matching} (DDSBM) \citep{kim2024discrete}. Just like the IMF algorithm, DDSBM alternates between performing a Markovian projection to obtain a Markov generator and reciprocal projections to preserve the marginal constraints. Concretely, the algorithm is outlined as follows.

\purple[Discrete Diffusion Schrödinger Bridge Matching]{\label{box:ddsbm}
DDSBM generates a sequence of Markov and reciprocal measures $(\mathbb{M}^n, \Pi^n)_{n\in \mathbb{N}}$ initialized at $\Pi^0:=\pi_{0,T}\mathbb{Q}_{\cdot|0,T}$ by alternating between the following steps:
\begin{enumerate}
    \item [(1a)] Solve the forward-time Markovian projection $\mathbb{M}^{2n+1}:=\text{proj}_{\mathcal{M}}(\Pi^{2n})$ by updating a parameterized generator $\boldsymbol{Q}^{\mathbb{M}^\theta}$ to minimize $\mathcal{L}_{\text{DDSBM}}(\theta):=\text{KL}(\Pi^{2n}\|\mathbb{M}^\theta)$
    \item[(1b)] Define the reciprocal projection as $\Pi^{2n+1}:=\mathbb{M}^{2n+1}\mathbb{Q}_{\cdot|0,T}$
    \item[(2a)] Solve the backward-time Markovian projection $\mathbb{M}^{2n+2}:=\text{proj}_{\mathcal{M}}(\Pi^{2n+1})$ by updating a parameterized generator $\tilde{\boldsymbol{Q}}^{\mathbb{M}^\phi}$ to minimize $\mathcal{L}_{\text{DDSBM}}(\phi):=\text{KL}(\Pi^{2n+1}\|\mathbb{M}^\phi)$
    \item[(2b)] Define the reciprocal projection as $\Pi^{2n+2}:=\mathbb{M}^{2n+2}\mathbb{Q}_{\cdot|0,T}$
\end{enumerate}
}
While the high-level form of this algorithm is straightforward, we will build a deeper understanding of how the algorithm is implemented in practice by analyzing each step below.

\textbf{Step 1a: Forward-Time Markovian Projection.} 
To obtain the Markovian projection $\mathbb{M}^{2n+1}=\text{proj}_{\mathcal{M}}(\Pi^{2n})$, we \textbf{minimize the KL divergence} $\text{KL}(\Pi^{2n}\|\mathbb{M}^\theta)$ derived in Lemma \ref{lemma:ctmc-cond-generator-reciprocal} between the reciprocal measure from the previous iteration $\Pi^{2n}=\mathbb{M}^\phi_{0,T}\mathbb{Q}_{\cdot|0,T}$.

To do this, we define the loss function $\mathcal{L}(\theta)$ which yields $\mathbb{M}_\theta=\text{proj}_{\mathcal{M}}(\Pi^{2n})$ at optimality:
\begin{small}
\begin{align}
    \mathcal{L}_{\text{DDSBM}}(\theta) :=\text{KL}(\Pi^{2n}\|\mathbb{M}^\theta)=\int_0^T\mathbb{E}_{\Pi_{t,T}^{2n}}\left[\sum_{\boldsymbol{y}\neq \boldsymbol{X}_t}\boldsymbol{Q}^{\bluetext{\mathbb{Q}_{\cdot|T}}}_t\log \frac{\boldsymbol{Q}^{\bluetext{\mathbb{Q}_{\cdot|T}}}_t}{\boldsymbol{Q}_t^{\pinktext{\mathbb{M}^\theta}}}(\boldsymbol{X}_t, \boldsymbol{y})+(\boldsymbol{Q}^{\bluetext{\mathbb{Q}_{\cdot|T}}}_{t}-\boldsymbol{Q}_t^{\pinktext{\mathbb{M}^\theta}})(\boldsymbol{X}_t, \boldsymbol{X}_t)\right]dt
\end{align}
\end{small}
where $\Pi^{2n}=\mathbb{M}^\phi_{0,T}\mathbb{Q}_{\cdot|0,T}$. To sample from $\Pi^{2n}$, we first sample endpoints by simulating the reverse-time Markov generator $\mathbb{M}^\phi$ to obtain $(\boldsymbol{x}_0, \boldsymbol{x}_T)\sim \mathbb{M}^\phi_{0,T}$ and then sampling the reference bridge $\boldsymbol{X}_t\sim \mathbb{Q}(\cdot|\boldsymbol{x}_0, \boldsymbol{x}_T)$

For each intermediate sample $\boldsymbol{X}_t\sim \mathbb{Q}(\cdot|\boldsymbol{x}_0, \boldsymbol{x}_T)$, we condition the forward-time reference generator to $\boldsymbol{X}_T=\boldsymbol{x}_T$ defined in Lemma \ref{lemma:ctmc-cond-generator-markov} as:
\begin{align}
    \bluetext{\boldsymbol{Q}^{\mathbb{Q}_{\cdot|T}}_t(\boldsymbol{X}_t, \boldsymbol{y})}=\mathbb{E}_{\boldsymbol{X}_T\sim \Pi_{T|0,t}}\left[\boldsymbol{Q}^0_t(\boldsymbol{X}_t,\boldsymbol{y}; \boldsymbol{X}_T)|\boldsymbol{X}_t, \boldsymbol{X}_0\right]
\end{align}

\textbf{Step 1b: Forward-Time Reciprocal Projection.}
Then, we define the corresponding reciprocal projection $\Pi^{2n+1}=\text{proj}_{\mathcal{R}(\mathbb{Q})}(\mathbb{M}^{2n+1})$ as the mixture of bridges under the reference measure conditioned on the endpoint law of $\mathbb{M}^\theta$ defined as:
\begin{align}
    \Pi^{2n+1}=\mathbb{M}^\theta_{0,T}\mathbb{Q}_{\cdot|0,T}
\end{align}

While we show that the Markovian projection preserves the bridge measure $\mathbb{M}_t=\Pi_t$ in theory (Proposition \ref{prop:markov-proj}), in practice,  parameterizing the generator of the forward CTMC with a neural network $\theta$ can result in mismatches in the terminal marginal constraint at time $t=T$. Therefore, similarly to the DSBM algorithm in Section \ref{subsec:diffusionsbm}, we also parameterize the \textit{reverse-time} Markovian projection that explicitly constrains the terminal marginal.

\textbf{Step 2a: Reverse-Time Markovian Projection.} 
To obtain the reverse-time Markovian projection $\mathbb{M}^{2n+2}=\text{proj}_{\mathcal{M}}(\Pi^{2n+1})$ parameterized by the reverse generator $\phi$, we \textbf{minimize the KL divergence} $\text{KL}(\Pi^{2n+1}\|\mathbb{M}^\phi)$ with reciprocal measure from the previous iteration $\Pi^{2n+1}=\mathbb{M}^\theta_{0,T}\mathbb{Q}_{\cdot|0,T}$ and define the loss function $\mathcal{L}(\phi)$ which yields $\mathbb{M}^\phi=\text{proj}_{\mathcal{M}}(\Pi^{2n+1})$ at optimality:
\begin{small}
\begin{align}
    \mathcal{L}_{\text{DDSBM}}(\phi) :=\text{KL}(\Pi^{2n+1}\|\pinktext{\mathbb{M}^\phi})=\int_0^T\mathbb{E}_{\Pi^{2n+1}_{0,t}}\left[\sum_{\boldsymbol{y}\neq \boldsymbol{X}_t}\tilde{\boldsymbol{Q}}^{\bluetext{\mathbb{Q}_{\cdot|0}}}_t\log \frac{\tilde{\boldsymbol{Q}}^{\bluetext{\mathbb{Q}_{\cdot|0}}}_t}{\tilde{\boldsymbol{Q}}_t^{\pinktext{\mathbb{M}^\phi}}}(\boldsymbol{X}_t, \boldsymbol{y})+(\tilde{\boldsymbol{Q}}^{\bluetext{\mathbb{Q}_{\cdot|0}}}_{t}-\tilde{\boldsymbol{Q}}_t^{\pinktext{\mathbb{M}^\phi}})(\boldsymbol{X}_t, \boldsymbol{X}_t)\right]dt
\end{align}
\end{small}
where the expectation is over samples from $\Pi^{2n+1}=\mathbb{M}^\theta_{0,T}\mathbb{Q}_{\cdot|0,T}$. Each sample from $\Pi^{2n+1}$ is obtained by simulating the \textit{forward-time} Markov generator $\mathbb{M}^\theta$ to obtain $(\boldsymbol{x}_0, \boldsymbol{x}_T)\sim \mathbb{M}^\theta_{0,T}$ and then sampling the reference bridge $\boldsymbol{X}_t\sim \mathbb{Q}(\cdot|\boldsymbol{x}_0, \boldsymbol{x}_T)$. 

For each intermediate sample $\boldsymbol{X}_t\sim \mathbb{Q}(\cdot|\boldsymbol{x}_0, \boldsymbol{x}_T)$, we condition the reverse-time reference generator to $\boldsymbol{X}_0=\boldsymbol{x}_0$ defined analogously to the forward-time generator as:
\begin{align}
    \bluetext{\tilde{\boldsymbol{Q}}^{\mathbb{Q}_{\cdot|0}}_t(\tilde{\boldsymbol{X}}_t, \boldsymbol{y})}=\mathbb{E}_{\boldsymbol{X}_0\sim \Pi_{0|t,T}}\left[\boldsymbol{Q}^0_t(\tilde{\boldsymbol{X}}_t,\boldsymbol{y}; \boldsymbol{X}_T)|\tilde{\boldsymbol{X}}_t, \boldsymbol{X}_0\right]
\end{align}

\textbf{Step 2b: Reverse-Time Reciprocal Projection.}
Then, we define the corresponding reciprocal projection $\Pi^{2n+2}=\text{proj}_{\mathcal{R}(\mathbb{Q})}(\mathbb{M}^{2n+2})$ as the mixture of bridges under the reference measure conditioned on the endpoint law of $\mathbb{M}^\phi$ defined as:
\begin{align}
    \Pi^{2n+2}=\mathbb{M}^\phi_{0,T}\mathbb{Q}_{\cdot|0,T}
\end{align}

Now that we understand how the discrete diffusion SBM algorithm is implemented in practice, it is natural to question whether it yields the same convergence guarantees as the IMF procedure. Since we already establish the intuition and formal proofs for the convergence of the IMF procedure in $\mathbb{R}^d$ in Section \ref{subsec:markov-reciprocal-proj}, the discrete state space analog follows similar intuition, where the IMF sequence results in monotonically decreasing KL divergence with the optimal bridge $\mathbb{P}^\star$ which converges to zero at the limit when the number of iterations goes to infinity.\footnote{for a rigorous proof that explicitly uses the CTMC generators, see Theorem 3.3 in \citep{kim2024discrete}.}

\begin{remark}[Comparison of Discrete SB-SOC and DDSBM]
    A \textbf{key distinction} between the discrete diffusion Schrödinger bridge method (DDSBM) and the discrete stochastic optimal control (SOC) formulation lies in their dependence on target samples. The DDSBM algorithm initializes the reciprocal measure as $\Pi^0:= \pi_{0,T}\mathbb{Q}_{\cdot|0,T}$, which requires access pairs from both marginals, and then samples from $\pi_0$ when optimizing the forward-time Markovian projection and $\pi_T$ when optimizing the reverse-time Markovian projection. 
    
    In contrast, the discrete SB-SOC framework (Box \ref{box:discrete-sbsoc}) only requires computing the RND between the optimally-controlled path measure and the one generated from the current control, enabling matching of unknown target distributions, such as temperature-annealed or reward-tilted distributions. 
\end{remark}

\subsection{Closing Remarks for Section \ref{sec:discrete-state-space}}
In this section, we introduced the theory of \textbf{continuous-time Markov chains} (CTMCs), which provide the natural analogue of stochastic differential equations for modeling stochastic processes in discrete state spaces $\mathcal{X}=\{1,\dots,d\}$. In this setting, the role of the control drift is replaced by a \textit{generator} or \textit{transition rate} matrix $\boldsymbol{Q}_t^u$, which governs the jump dynamics of the process. Using this representation, we defined the \textbf{discrete Schrödinger bridge problem}, which seeks the controlled CTMC path measure $\mathbb{P}^u$ with generator $\boldsymbol{Q}_t^u$ that remains closest, in KL divergence, to a reference path measure $\mathbb{Q}$ with generator $\boldsymbol{Q}_t^0$, while satisfying the marginal constraints $p_0=\pi_0$ and $p_T=\pi_T$.

Building on this formulation, we introduced two frameworks for solving the discrete Schrödinger bridge problem: \textbf{stochastic optimal control} (SOC) and \textbf{iterative Markovian fitting} (IMF). These approaches mirror their continuous-state counterparts, translating the ideas of controlled SDEs and path-space projections to the setting of jump processes governed by CTMC generators.

Taken together, we have shown that the Schrödinger bridge framework extends naturally from diffusion processes in continuous spaces to jump processes in discrete state spaces. With this theoretical foundation in place, we are now prepared to explore how all the ideas developed throughout this guide can be applied in practice to solve real data-driven and scientific problems. In the final section, we examine several \textbf{applications of generative modeling with Schrödinger bridges}, illustrating how the theoretical tools developed throughout this guide can be used to construct practical generative models across a variety of domains.

\newpage
\section{Applications of Generative Modeling with Schrödinger Bridges}
\label{sec:applications}
While Schrödinger bridges can be seen as a unified framework that encompasses a large portion of generative modeling techniques, from denoising diffusion to score-based generative modeling to flow matching, there are several applications where Schrödinger bridge frameworks are specialized to solve. In this section, we highlight three prominent applications where advances in Schrödinger bridge–based generative modeling have led to principled and practically effective solutions, including \textbf{data translation} (Section \ref{subsec:data-translation}), \textbf{single-cell modeling} (Section \ref{subsec:cell-modeling}), and \textbf{sampling high-Boltzmann energy distributions} (Section \ref{subsec:sampling}). Our goal is not to provide an exhaustive review of applications, but to illustrate how learning entropy-regularized stochastic paths between structured distributions can be applied to diverse scientific and data-driven problems. 

\subsection{Data Translation}
\label{subsec:data-translation}

\textit{Prerequisite: Section \ref{subsec:primer-sgm}}

\begin{figure}
    \centering
    \includegraphics[width=\linewidth]{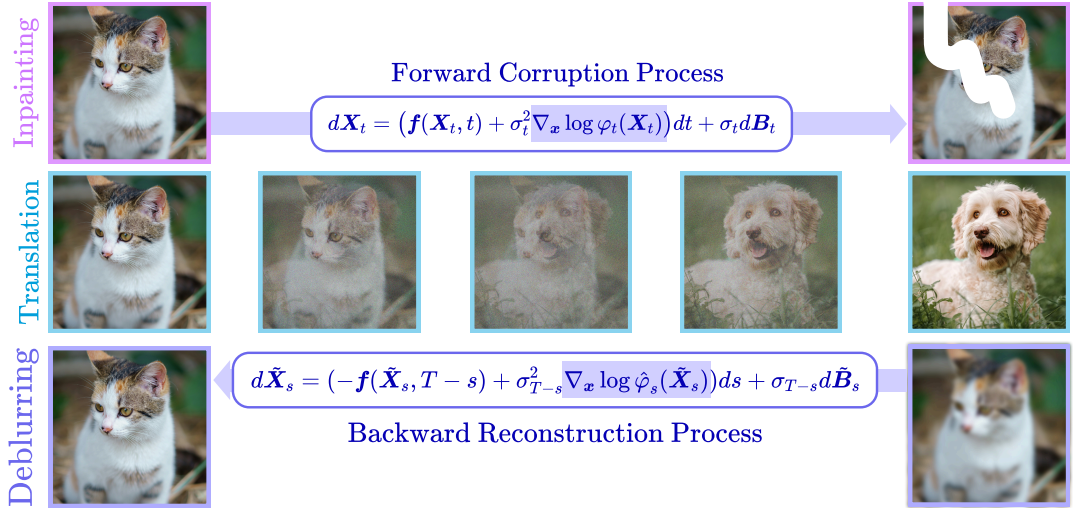}
    \caption{\textbf{Image to Image Translation with Schrödinger Bridges.} Schrödinger bridges enable image-to-image translation by learning a stochastic process that transports samples between a structured prior distribution $\pi_0$ (e.g., corrupted or partially observed images) and a clean target distribution $\pi_T$ (right to left). Rather than generating images from pure Gaussian noise, the model learns the optimal stochastic interpolation conditioned on the endpoints, corresponding to the dynamic optimal transport map when diffusion is small. This yields generation trajectories that progressively restore structure, for tasks such as inpainting, deblurring, or translation, leading to more efficient and interpretable transformations than denoising from noise alone.}
    \label{fig:data-translation}
\end{figure}

The goal of data translation is to transform samples from one unknown distribution to another. Specifically, we will focus on \boldtext{image-to-image translation}, which aims to learn a mapping between pairs of images and can be applied to many tasks such as deblurring, inpainting, and image editing, as illustrated in Figure \ref{fig:data-translation}. In this setting, the source and target datasets correspond to two probability distributions defined over the space of images, and the goal is to learn a transformation that transports samples from the source distribution to the target distribution while \textit{preserving the underlying structure of the data}.

The Schrödinger bridge framework provides a principled probabilistic formulation of this problem. Given empirical samples from the source distribution $\pi_0$ and the target distribution $\pi_T$, the Schrödinger bridge constructs the most likely stochastic process that connects the two distributions while remaining close to a reference diffusion. Rather than learning a deterministic mapping between individual samples, the bridge defines a stochastic transport process that transforms an image in the source distribution into a many possible images in the target distribution.

In many image restoration tasks the degraded image already contains \textbf{substantial structural information about the target image}. Rather than generating from random noise, Schrödinger bridges allow the generative process to start directly from the degraded image distribution and gradually transform it into the clean image distribution. 

Since the data translation task assumes \textit{paired samples}, we can consider the form of the SB problem, where the clean distribution $\pi_T=\delta_{x}$ is a dirac Delta function at the target data point (e.g., clean image), which corresponds to some distribution of initial samples $\boldsymbol{x}_T\sim \pi_T(\cdot|\boldsymbol{x}_0)$ (e.g. degraded or corrupted images). This produces a \textit{joint distribution} of paired samples defined as:
\begin{align}
    \pi(\boldsymbol{x}_0, \boldsymbol{x}_T)=\pi_0(\boldsymbol{x}_0)\pi(\boldsymbol{x}_T|\boldsymbol{x}_0)
\end{align}

Given this joint distribution, we can define a form of the SB problem which samples pairs from the joint data distribution and learns a \textbf{Gaussian interpolation} between the pair \citep{liu20232}. 
\begin{definition}[Data Translation with Schrödinger Bridge]\label{def:data-translation}
    Given data pairs sampled from the joint distribution $(\boldsymbol{x}_0, \boldsymbol{x}_T)\sim \pi_0(\boldsymbol{x}_0)\pi_T(\boldsymbol{x}_T|\boldsymbol{x}_0)$, the optimal Schrödinger bridge density at intermediate times $t\in [0,T]$ takes the Gaussian form:
    \begin{small}
    \begin{align}
        \boldsymbol{X}_t \sim p_t(\boldsymbol{X}_t|\boldsymbol{x}_0, \boldsymbol{x}_T)=\mathcal{N}(\boldsymbol{X}_t;\boldsymbol{\mu}_t(\boldsymbol{x}_0, \boldsymbol{x}_T), \boldsymbol{\Sigma}_t)
    \end{align}
    \end{small}
    where $\boldsymbol{\mu}_t\in \mathbb{R}^d$ is the conditional mean at time $t$, $\boldsymbol{\Sigma}_t\in \mathbb{R}^{d\times d}$ is the covariance matrix. A choice of parameterization used in image restoration \citep{liu20232} defines $\sigma_t\in \mathbb{R}$ as the variance accumulated in forward time, and $\bar{\sigma}_t\in \mathbb{R}$ as the variance accumulated in the reverse time coordinate, which explicitly define the mean and covariance as:
    \begin{align}
        \boldsymbol{\mu}_t=\frac{\bar{\sigma}_t^2}{\bar{\sigma}_t^2+\sigma_t}\boldsymbol{x}_0+\frac{\sigma_t^2}{\bar{\sigma}_t^2+\sigma_t}\boldsymbol{x}_T, \quad \boldsymbol{\Sigma}_t=\frac{\sigma^2_t\bar{\sigma}_t^2}{\bar{\sigma}_t^2+\sigma_t^2}\boldsymbol{I}_d, \quad \sigma_t^2:=\int_0^t\beta_sds, \quad \bar{\sigma}_t^2:=\int_t^T\beta_sds
    \end{align}
    When $\beta_t\equiv \beta$ is constant over $t\in [0,T]$ and sufficiently small, the optimal SB reduces to the \textbf{dynamic optimal transport} map between the pair $(\boldsymbol{x}_0, \boldsymbol{x}_T)$ defined as:
    \begin{small}
    \begin{align}
        \boldsymbol{v}_t(\boldsymbol{X}_t)=\frac{\boldsymbol{X}_t-\boldsymbol{x}_0}{t}, \quad \boldsymbol{\mu}_t=\left(1-\frac{t}{T}\right)\boldsymbol{x}_0+\frac{t}{T}\boldsymbol{x}_T
    \end{align}
    \end{small}
    To learn the velocity $\boldsymbol{v}_\theta$ that optimally transports samples from $\pi_0$ to $\pi_T$, we optimize the matching loss defined as:
    \begin{small}
    \begin{align}
        \mathcal{L}(\theta):=\mathbb{E}_{\boldsymbol{x}_0\sim \pi_0(\boldsymbol{X}_0), \boldsymbol{x}_T\sim \pi_T(\boldsymbol{X}_T|\boldsymbol{X}_T)}\int_0^T\bigg\|\boldsymbol{v}_\theta(\boldsymbol{X}_t,t)-\underbrace{\frac{\boldsymbol{X}_t-\boldsymbol{x}_0}{\sigma_t}}_{\nabla \log \hat\varphi_t(\boldsymbol{X}_t|\boldsymbol{x}_0)}\bigg\|^2 
    \end{align}
    \end{small}
    where $\nabla \log \hat\varphi_t(\boldsymbol{X}_t|\boldsymbol{X}_0)$ is the SB potential drift which is equivalent to the \textbf{conditional score function} $\nabla \log p_t(\boldsymbol{X}_t|\boldsymbol{x}_0)$ where $\boldsymbol{X}_t\sim p_t(\cdot|\boldsymbol{x}_0, \boldsymbol{x}_T)$.
\end{definition}

We observe that this objective resembles the standard \textbf{conditional score matching objective}, where the score function is given by $\nabla \log p_t(\boldsymbol{X}_t|\boldsymbol{y})$, where $\boldsymbol{y}$ is the conditioning information, which in the case of data translation is the initial or corrupted sample. However, in conditional score matching, the generation process starts from Gaussian noise $\boldsymbol{x}_0\sim \mathcal{N}(\boldsymbol{0}, \boldsymbol{I}_d)$ and learns the score function using the corrupted sample as just another input to the model. 

When the corrupted sample is already close to the target distribution, generating the clean target from pure noise is highly inefficient, which reveals the advantages of Schrödinger bridges as a way to learn the optimal interpolating bridge between a \textbf{structurally informative prior} $\pi_0$ to the target distribution $\pi_T$. Since the SB dynamics explicitly connect corrupted and clean image distributions, the generative trajectories correspond to progressive restoration processes rather than denoising from pure noise, leading to more interpretable and efficient generation. 

\subsection{Modeling Single-Cell State Dynamics}
\label{subsec:cell-modeling}
\textit{Prerequisites: Sections \ref{subsec:generalized-sb}, \ref{subsec:multi-marginal-sb}, \ref{subsec:unbalanced-sbp}, \ref{subsec:branched-sbp}}

Cellular systems undergo dynamic state transitions from cell differentiation to responses under genetic or drug perturbations to adaptive changes during development and disease. These processes can be viewed as trajectories through a high-dimensional cell state space $\mathbb{R}^d$, typically defined as single-cell RNA sequencing (scRNA-seq) measurements of gene expression.

Due to the destructive nature of single-cell sequencing technologies, which kill the cell after measurement, data is often a collection of static snapshots of populations at discrete time points rather than continuous-time measurements. As a result, the underlying stochastic dynamics governing how cells transition between states are not directly observed. Instead, the problem becomes one of \textbf{inferring the dynamical process that transports a distribution of cells at an initial condition $\pi_0\in \mathcal{P}(\mathbb{R}^d)$ to a distribution observed at a later time $\pi_T\in \mathcal{P}(\mathbb{R}^d)$}.
\begin{align}
    \underbrace{\pi_0(\boldsymbol{x})}_{\text{initial cell distribution}}\xrightarrow{\text{differentiation, perturbation, etc.}}\underbrace{\pi_T(\boldsymbol{x})}_{\text{terminal cell distribution}}, \quad \boldsymbol{x}\in \mathbb{R}^d
\end{align}
Since cell states naturally lie on a non-linear biological manifold, the Schrödinger bridge framework provides a natural approach to this problem. Given empirical distributions of cellular states at sequential time points $\{\pi_{t_k}\}_{k=1}^K$, the Schrödinger bridge identifies the most likely stochastic process that connects these distributions while remaining close to a chosen reference diffusion. In this formulation, the inferred bridge represents a probabilistic model of cell state evolution that captures both the deterministic drift driving differentiation and the stochastic variability inherent in biological systems.

\begin{figure}
    \centering
    \includegraphics[width=\linewidth]{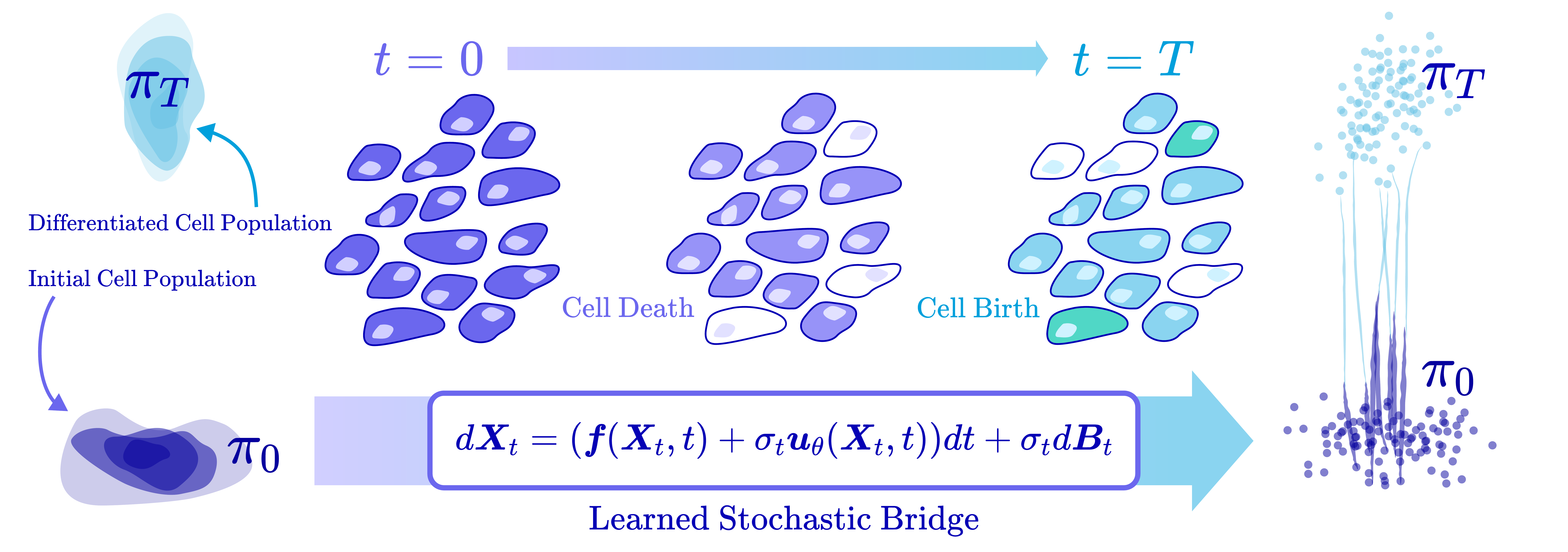}
    \caption{\textbf{Modeling Cellular Dynamics with Schrödinger Bridges.} Cellular processes such as differentiation, perturbation responses, and development can be viewed as stochastic trajectories through a high-dimensional cell state space inferred from single-cell measurements. Because single-cell sequencing provides only static snapshots of cell populations at discrete times, the underlying dynamics are not directly observed. Schrödinger bridges infer the most likely stochastic process that transports an initial cell distribution $\pi_0$ to a later observed distribution $\pi_T$, yielding a probabilistic model of cell state evolution that captures both deterministic differentiation trends and stochastic biological variability.}
    \label{fig:cell-state}
\end{figure}

\purple[Modeling Cellular Dynamics with Schrödinger Bridges]{\label{def:cell-dynmamics}
Consider a sequence of snapshots of a cell population at discrete time intervals $\{\pi_{t_k}\in \mathcal{P}(\mathbb{R}^d)\}_{k=1}^K$ for $0=t_0< \dots < t_k< \dots< t_K=T$ where $\mathbb{R}^d$ represents a $d$-dimensional gene expression space. Then, the problem of modeling cellular dynamics between snapshots can framed as optimizing a parameterized control $\boldsymbol{u}_\theta(\boldsymbol{x},t)$ that minimizes a combination of the (\ref{eq:multi-marginal-sbp}) and the (\ref{eq:generalized-sbp}):
\begin{align}
&\inf_{\boldsymbol{u}_\theta}\mathbb{E}_{\boldsymbol{X}_{0,T}\sim \mathbb{P}^u}\bigg[\int_0^T\left(\frac{1}{2}\|\boldsymbol{u}_\theta(\boldsymbol{X},t)\|^2+\mathcal{I}(\boldsymbol{X},p_t,t)\right)dt\bigg]\tag{Generalized SB Problem}\\
&\text{s.t.}\quad\begin{cases}
    d\boldsymbol{X}_t=(\boldsymbol{f}(\boldsymbol{x},t)+\sigma_t\boldsymbol{u}_\theta(\boldsymbol{X}_t,t))dt+\sigma_td\boldsymbol{B}_t\\
    \boldsymbol{X}_0\sim \pi_0, \quad \boldsymbol{X}_T\sim\pi_T
\end{cases}
\end{align}
where $\boldsymbol{f}(\boldsymbol{x},t)$ is the underlying drift of the cell population, $\sigma_t$ is the diffusion coefficient representing stochastic in cell evolution, and $\mathcal{I}(\boldsymbol{x},p_t):\mathbb{R}^d\times\mathcal{P}(\mathbb{R}^d)\to \mathbb{R}$ is the interaction cost that measures how the cells evolve with respect to the overall population. 

When the population of cells \textit{changes over time}, we can formulate the cell modeling problem as a (\ref{eq:unbalanced-sbp}) with unbalanced marginal constraints:
\begin{align}
    &\inf_{p_t, \boldsymbol{u}_\theta, g_\phi}\int_0^T\int_{\mathbb{R}^d}\left[\frac{1}{2}\|\boldsymbol{u}_\theta(\boldsymbol{x},t)\|^2+\alpha\Psi(g_\phi(\boldsymbol{x},t))\right]p_t(\boldsymbol{x})d\boldsymbol{x}dt\\
    &\textbf{s.t.} \quad \begin{cases}
        \partial_tp_t=-\nabla\cdot (p_t(\boldsymbol{f}+\sigma_t\boldsymbol{u})+\frac{\sigma^2_t}{2} \Delta p_t+gp_t\\
        \forall k\in \{1, \dots, K\}, \quad p_{t_k}= \pi_{t_k}
    \end{cases}\nonumber
\end{align}
where $g_\phi(\boldsymbol{x},t)$ parametrizes the growth rate that determines how the cell population increase or decrease in mass and $\Psi(g_\phi(\boldsymbol{x},t))$ is the scalar function that penalizes cell growth and death.

Finally, we can consider the case of \textbf{branching cell dynamics}, where the terminal cell population contains multiple distinct modes $\pi_T=\{\pi_{T,k}\}_{k=1}^K$ defining different sub-populations that have diverged as a result of differentiation or cellular perterbation. This problem can naturally be framed as the (\ref{eq:branched-sbp}), which can be solved by defining a \textit{set} of parameterized control drifts and growth rates $\{\boldsymbol{u}_{\theta, k}, g_{\phi, k}\}_{k=1}^K$ that minimize:
\begin{small}
\begin{align}
    &\inf_{\{\boldsymbol{u}_{\theta, k}, g_{\phi, k}\}_{k=1}^K}\int_0^T\bigg\{\mathbb{E}_{p_{t, 0}}\left[\frac{1}{2}\|\boldsymbol{u}_{\theta, 0}(\boldsymbol{X}_{t, 0},t)\|^2+c(\boldsymbol{X}_{t, 0},t)\right]w_{t, 0}+\nonumber\\
    &\quad\quad \sum_{k=1}^K\mathbb{E}_{p_{t, k}}\left[\frac{1}{2}\|\boldsymbol{u}_{\theta, k}(\boldsymbol{X}_{t, k},t)\|^2+c(\boldsymbol{X}_{t,k},t)\right]w_{t, k}\bigg\}dt\\
    &\quad \quad \text{s.t.}\quad \begin{cases}
        d\boldsymbol{X}_{t, k}=(\boldsymbol{f}(\boldsymbol{X}_{t,k},t)+\sigma_t\boldsymbol{u}_{\theta, k}(\boldsymbol{X}_{t,k},t))dt+\sigma_td\boldsymbol{B}_t\\
        \boldsymbol{X}_0\sim \pi_0, \quad \boldsymbol{X}_{T, k}\sim \pi_{T, k}\\
        w_{0, k}=\delta_{k=0}, \quad w_{T,k}=w^\star_{T,k}
    \end{cases}\nonumber
\end{align}
\end{small}
where the weight of the primary branch is given by $w_{t, 0}=1+\int_0^tg_{t,\phi}(\boldsymbol{X}_{s,0},s)ds$ and the weights of the $K$ secondary branches is given by $w_{t, k}=\int_0^tg_{k,\phi}(\boldsymbol{X}_{s,k},s)ds$.
}

This shows that Schrödinger bridges defines a unified problem that can be specialized to several biologically relevant settings. In the balanced setting, the bridge captures smooth state transitions between multiple cell population distributions. In the unbalanced setting, it incorporates cell proliferation and death through growth terms, allowing the total population mass to vary over time, and finally, the branched Schrödinger bridge formulation models lineage diversification and diverse responses to perturbations by decomposing the population into multiple sub-trajectories, each governed by its own dynamics and weights.

Together, these formulations provide a principled and flexible framework for modeling complex cellular processes, including differentiation, perturbation responses, and lineage branching \citep{zhang2024learning, zhang2025modeling, tang2026branchsbm}. By learning the underlying bridge dynamics, we not only interpolate intermediate cellular states but also recover the latent structure of population evolution, enabling predictive modeling of how cell populations evolve under changing biological conditions.

\subsection{Sampling Boltzmann Distributions}
\label{subsec:sampling}
The \textbf{sampling} problem aims to generate a complete reconstruction of a probability distribution. A specific class of distributions of particular interest in computational sciences are \boldtext{Boltzmann distributions}, which are unnormalized and often high-dimensional and defined only by an \textbf{energy function} $U(\boldsymbol{x}):\mathbb{R}^d\to \mathbb{R}$ without explicit samples \citep{binder1992monte, tuckerman2023statistical, yang2019enhanced}.  

\begin{definition}[Sampling Boltzmann Distributions]\label{problem:boltzmann}
    Given an energy function $U(\boldsymbol{x}): \mathbb{R}^d\to \mathbb{R}$ where $\mathbb{R}^d$, the \textbf{Boltzmann distribution} is defined as:
    \begin{align}
        \pi_T(\boldsymbol{x}):=\frac{e^{-U(\boldsymbol{x})}}{Z}, \quad \text{where}\quad Z:=\int_{\mathbb{R}^d}e^{-U(\boldsymbol{x})}d\boldsymbol{x}
    \end{align}
    such that $\int_{\mathbb{R}^d}\pi_T(\boldsymbol{x})d\boldsymbol{x}=1$ and $Z$ is some, often intractable, normalization factor.
\end{definition}

This class of distributions are particularly difficult to sample due to the computational cost of evaluating the energy function of the molecular systems. Traditional approaches such as Markov-chain Monte Carlo (MCMC) \citep{metropolis1953equation, neal2001annealed} and Sequential Monte Carlo (SMC) \citep{doucet2001introduction} while provably converges to the target distribution $\pi_T$, exhibit \textbf{slow mixing times}, where the sampled distribution requires a significant number of simulation steps to converge to the target distribution, and can \textbf{remain trapped in local minima} with poor reconstruction of the global energy landscape, especially for large molecules with high-dimensional representations and multi-modal energy landscapes.

To overcome these challenges, generative modeling frameworks have been developed that approach the problem as transporting a easy-to-sample source distribution $\boldsymbol{X}_0\sim \pi_0$ (e.g., a Gaussian or Dirac delta at the origin) to the target Boltzmann distribution $\pi_T(\boldsymbol{x})=\frac{e^{-U(\boldsymbol{x})}}{Z}$. Speciically, these frameworks define parameterized control drift $\boldsymbol{u}_\theta(\boldsymbol{x},t)$ that steers the sample toward the target Boltzmann distribution $\pi_T(\boldsymbol{x})=\frac{e^{-U(\boldsymbol{x})}}{Z}$ through the controlled SDE:
\begin{align}
    d\boldsymbol{X}_t=[\boldsymbol{f}_t(\boldsymbol{X}_t)+\sigma_t\boldsymbol{u}_\theta(\boldsymbol{X}_t,t)]dt+\sigma_td\boldsymbol{B}_t, \quad \boldsymbol{X}_0\sim \pi_0
\end{align}
In this setting, we do not have access to explicit samples from the target distribution $\pi_T$ and can only evaluate its unnormalized density through an energy function $\pi_T(\boldsymbol{x})=\frac{e^{-U(\boldsymbol{x})}}{Z}$. This implicit specification makes direct sampling intractable, and naturally motivates a formulation using \textbf{stochastic optimal control} (SOC) as described in Section \ref{sec:sb-optimal-control}, where we optimize the control $\boldsymbol{u}_\theta(\boldsymbol{x},t)$ such that the terminal distribution generated from the optimal control $\boldsymbol{u}^\star(\boldsymbol{x},t)$ matches the Boltzmann distribution $p^\star_T=\pi_T$.

Specifically, we consider the (\ref{eq:sb-soc-objective}) described in Section \ref{subsec:sb-soc}, where the terminal constraint is defined as $\log \frac{\hat\varphi_T(\boldsymbol{X}_T)}{\pi_T(\boldsymbol{X}_T)}$, where $\hat\varphi_T$ is the backward Schrödinger potential evaluated at time $t=T$. When sampling from a Boltzmann distribution, this constraint becomes:
\begin{align}
    \log \frac{\hat\varphi_T(\boldsymbol{X}_T)}{\pi_T(\boldsymbol{X}_T)}=\log\hat\varphi_T(\boldsymbol{X}_T)-\log\bluetext{\left(\frac{e^{-U(\boldsymbol{X}_T)}}{Z}\right)}=\log\hat\varphi_T(\boldsymbol{X}_T)+U(\boldsymbol{X}_T)+\log Z
\end{align}

Using this definition for the terminal constraint, we can define the Boltzmann SB-SOC problem.

\begin{figure}
    \centering
    \includegraphics[width=\linewidth]{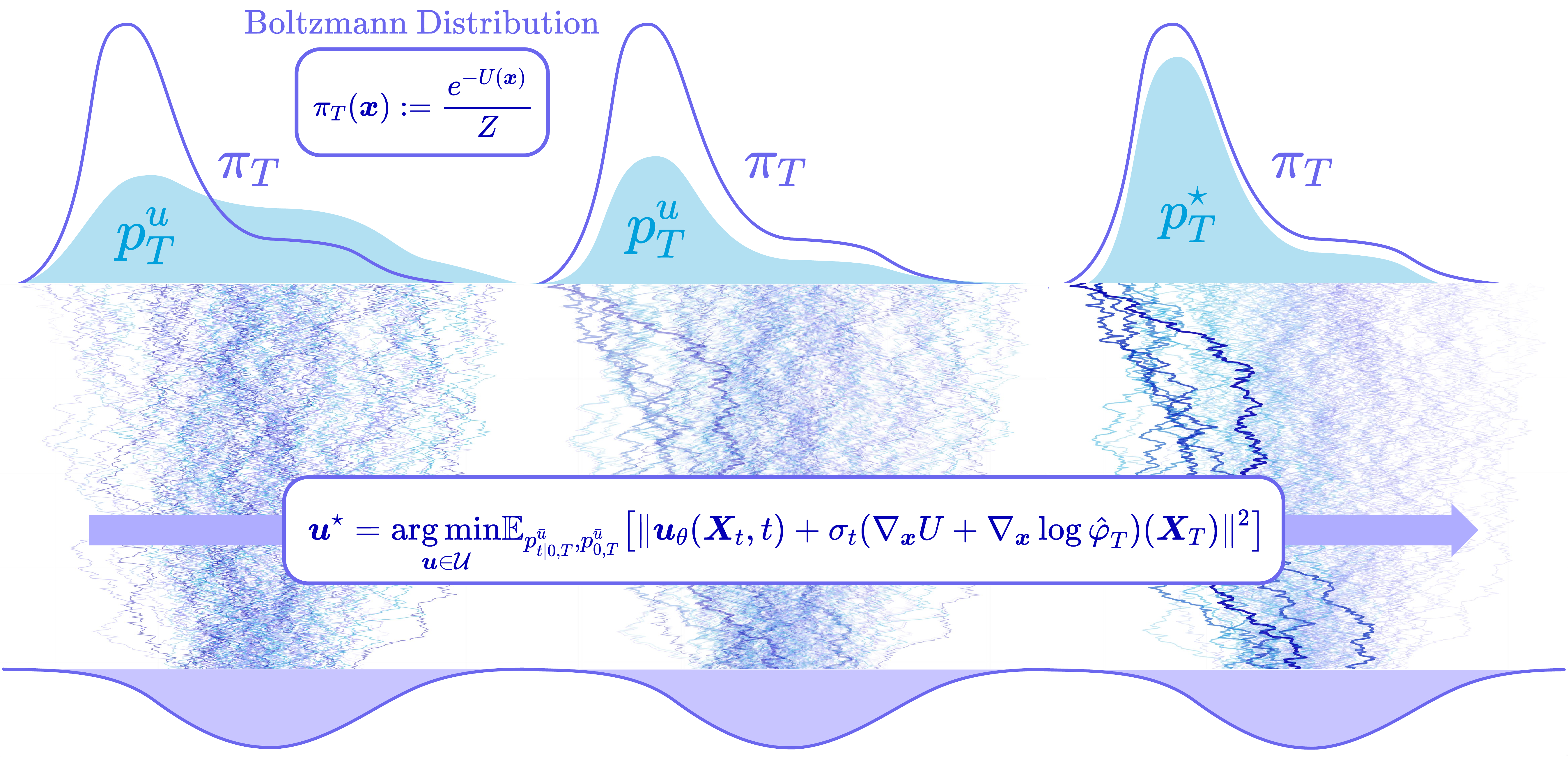}
    \caption{\textbf{Sampling the Boltzmann Distribution with Schrödinger Bridges.} Boltzmann distributions $\pi_T(\boldsymbol{x})\propto e^{-U(\boldsymbol{x})}$ arise in many scientific applications but are often difficult to sample from due to the intractable normalization constant $Z$. Schrödinger bridges address this by learning a controlled stochastic process that transports samples from an easy-to-sample prior $\pi_0$ (e.g., a Gaussian) to the target Boltzmann distribution. The learned control drift guides trajectories so that their terminal distribution matches $\pi_T$, enabling efficient sampling from energy-based models.}
    \label{fig:boltzmann}
\end{figure}

\begin{proposition}[Sampling the Boltzmann Distribution with Schrödinger Bridges and Stochastic Optimal Control (\citep{liu2025adjoint})]\label{prop:soc-sampling-boltzmann}
    The optimal control drift $\boldsymbol{u}^\star(\boldsymbol{x},t)$ that generates samples from the Boltzmann distribution $\pi_T(\boldsymbol{X}_T)= \frac{e^{-U(\boldsymbol{X}_T)}}{Z}$ solves the \textbf{stochastic optimal control} (SOC) problem defined as:
    \begin{align}
        \inf_{\boldsymbol{u}}&\mathbb{E}_{\boldsymbol{X}_{0:T}\sim \mathbb{P}^{u_\theta}}\left[\int_0^T\frac{1}{2}\|\boldsymbol{u}_\theta(\boldsymbol{X}_t,t)\|^2dt+\log\hat\varphi_T(\boldsymbol{X}_T)+U(\boldsymbol{X}_T)+\log Z\right]\label{eq:bolzmann-sb-soc}\\
        &\text{s.t.}\quad d\boldsymbol{X}_t=(\boldsymbol{f}(\boldsymbol{X}_t,t)+\sigma_t\boldsymbol{u}_\theta(\boldsymbol{X}_t,t))dt+\sigma_td\boldsymbol{B}_t, \quad \boldsymbol{X}_0\sim \pi_0\nonumber
    \end{align}
    where $\hat\varphi_T(\boldsymbol{x})$ is the backward Schrödinger potential. Under the optimal control $\boldsymbol{u}^\star(\boldsymbol{x},t)$, the generated distribution $p^\star_T$ exactly reconstructs the Boltzmann distribution $p_T^\star=\pi_T$.
\end{proposition}

\textit{Proof.} This result follows immediately from our proof of Proposition \ref{prop:initial-value-bias} by replacing the terminal value constraint with $\log \frac{\hat{\varphi}_T(\boldsymbol{X}_T)}{\pi_T(\boldsymbol{X}_T)}=\log\hat\varphi_T(\boldsymbol{X}_T)+U(\boldsymbol{X}_T)+\log Z$ to get:
\begin{align}
    \mathbb{P}^\star(\boldsymbol{X}_0, \boldsymbol{X}_T)&=\frac{1}{Z}\mathbb{Q}(\boldsymbol{X}_0, \boldsymbol{X}_T)\exp\left(\bluetext{-\log\hat\varphi_T(\boldsymbol{X}_T)-U(\boldsymbol{X}_T)-\log Z}-\log \varphi_0(\boldsymbol{X}_0)\right)
\end{align}
and integrating over $\boldsymbol{X}_T$ to get the terminal distribution generated from the optimal control drift $\boldsymbol{u}^\star$:
\begin{align}
    p_T^\star(\boldsymbol{X}_T)&=\exp\left(-U(\boldsymbol{X}_T)-\log Z\right)=\pi_T(\boldsymbol{X}_T)
\end{align}
which exactly matches the target Boltzmann density without the initial value function bias. \hfill $\square$

One approach to solving this problem is using the (\ref{eq:sb-am-loss}) and (\ref{eq:corrector-matching-objective}) \citep{liu2025adjoint}, defined speciically for sampling Boltzmann density as:
\begin{align}
    \mathcal{L}_{\text{SB-AM}}(\boldsymbol{u})&:=\mathbb{E}_{p^{\bar{u}}_{0,T}}\left[\frac{1}{2}\int_0^T\bluetext{\mathbb{E}_{p^{\bar{u}}_{t|0,T}}}\left[\left\|\boldsymbol{u}(\boldsymbol{X}_t,t)+\sigma_t(\bluetext{\nabla U(\boldsymbol{X}_T)+ \nabla \log\hat\varphi_T(\boldsymbol{X}_T)})\right\|^2\right]dt\right]\\
    \mathcal{L}_{\text{SB-CM}}(\widehat{\boldsymbol{Z}}_T)&:=\mathbb{E}_{p^{\bar{u}}_{0,T}}\left[\|\widehat{\boldsymbol{Z}}_T(\boldsymbol{X}_T)-\nabla_{\boldsymbol{x}_T}\log \mathbb{Q}_{T|0}(\boldsymbol{X}_T|\boldsymbol{X}_0)\|^2\right]
\end{align}
Since the prior distribution for the sampling problem is typically Gaussian and the intermediate dynamics can be arbitrarily defined as long as it converges to the target distribution, we consider the case where $\mathbb{Q}$ is pure Brownian motion with zero drift $\boldsymbol{f}_t:=0$ and the conditional distribution $p^{\bar{u}}_{t|0,T}$ takes a tractable form without requiring sampling the full path\footnote{See \citep{liu2025adjoint} Appendix A for explicit definition} $\boldsymbol{X}_{0:T}\sim \mathbb{P}^{\bar{u}}$.

This perspective reveals that sampling from Boltzmann distributions can be interpreted as learning the optimal Schrödinger bridge between a simple prior distribution and the target energy-based distribution. We show that formulation yields efficient algorithms that significantly reduce the number of energy evaluations and enables fast inference by simulating an SDE with a parameterized control drift. Notably, these examples highlight the broader potential of Schrödinger bridge methods as a \textbf{scalable framework for sampling in high-dimensional and complex energy landscapes}, with promising applications in modeling and understanding diverse molecular properties.

\subsection{Closing Remarks for Section \ref{sec:applications}}
This section illustrates notable examples of how Schrödinger bridge formulations extend beyond a unifying theoretical framework to become powerful, task-specific tools for generative modeling. Across data translation, single-cell dynamics, and sampling from complex energy landscapes, the common principle is the learning of entropy-regularized stochastic paths that respect both the structure of the data and the underlying physical or statistical constraints. By framing these problems as controlled stochastic processes, Schrödinger bridges provide a flexible mechanism for incorporating domain knowledge, handling distributional mismatch, and enabling efficient inference in high-dimensional settings. These successes highlight the versatility of the framework and suggest that Schrödinger bridge–based methods will continue to play a central role in advancing generative modeling across scientific and data-driven domains.

\newpage 
\section{Conclusion}

This guide aims to serve as a \textit{self-contained} theoretical exploration of the foundations of Schrödinger bridges and their role in modern generative modeling. By starting from fundamental concepts in optimal transport, stochastic processes, and information theory, we progressively build toward the theory of constructing Schrödinger bridge and modern generative modeling algorithms leveraging Schrödinger bridge theory to model real-world, high-dimensional systems and data. Throughout this guide, we illustrate how the Schrödinger bridge problem provides a unifying perspective on entropy-regularized optimal transport, stochastic optimal control, and score-based generation. Rather than viewing generative modeling as a collection of disconnected algorithmic techniques, the \textbf{Schrödinger bridge framework reveals a unified mathematical structure underlying these methods}. Building on top of this foundation exist extensions to diverse problem settings and algorithmic innovations.

One of the \textbf{central themes} of this guide is that generative modeling can be understood as a problem of probability transport in the space of path measures. In this view, the goal is not just to map one distribution into another through a deterministic map, but to identify the \textit{most likely stochastic process} that forms a continuous-time bridge between distributions them under a given reference process. This perspective unifies several seemingly distinct paradigms—including score-based generative models, flow matching, and stochastic optimal control—under a \textbf{common variational principle based on relative entropy between path measures}. 

Looking ahead, Schrödinger bridge methods are likely to play an increasingly important role in the development of generative modeling and probabilistic machine learning. Ongoing work on scalable bridge solvers \citep{gushchin2024light, zhang2025learning}, stochastic control formulations \citep{liu2025adjoint, tang2025entangledsbm}, application-focused algorithms \citep{li2025diffusion, li2025bridge}, and extensions to the discrete state space \citep{guo2026discrete, ksenofontov2025categorical} suggests the adaptibility and scalability of this framework. Notably, emerging applications of Schrödinger bridges for scientific applications—ranging from prediction of drug perturbations on cell populations \citep{tang2026branchsbm} and cell differentiation modeling \citep{zhang2024learning} to molecular dynamics simulation \citep{du2024doob, tang2025entangledsbm} and sampling high-dimensional molecular energy landscapes \citep{liu2025adjoint}—highlight the potential of bridge-based methods for modeling complex stochastic processes.

We hope that this guide serves both as a conceptual foundation and as a practical reference for researchers, practitioners, and learners interested in the theory and applications of Schrödinger bridges. By building the mathematical principles that underpin modern generative models, we aim to provide a framework for understanding existing methods and for developing new approaches that \textit{bridge} the gap between theory and application-focused design.

\paragraph{Note from the Author} I deeply thank all of my research mentors and collaborators for all the educational and inspiring conversations throughout my time in this field. As someone who enjoys writing and breaking down theory from first-principles, the process of writing this guide has been an extremely fulfilling experience, and I hope it becomes a core resource for researchers and learners interested in the foundations of generative modeling.

\newpage
\printbibliography

\end{document}